\documentclass{article}

\providecommand{\neuripsmode}{preprint}

\PassOptionsToPackage{numbers}{natbib}
\usepackage[\neuripsmode]{neurips_2026}

\usepackage{todonotes}
\usepackage{alphalph}

\usepackage[utf8]{inputenc}   
\usepackage[T1]{fontenc}      
\usepackage{microtype}        
\usepackage{amsmath}          
\usepackage{amsfonts}         
\usepackage{nicefrac}         
\usepackage{calc}
\usepackage{array}            
\usepackage{booktabs}         
\usepackage{multirow}         
\usepackage{tabularx}         
\usepackage{adjustbox}        
\usepackage{graphicx}         
\usepackage{subcaption}       
\usepackage{float}            
\usepackage{placeins}         
\usepackage{wrapfig}          
\usepackage{pdflscape}        
\ifdefined\colmmode
\fi
\usepackage{tikz}             
\usetikzlibrary{calc, arrows.meta, positioning, decorations.pathreplacing}
\usetikzlibrary{shapes.geometric}
\usepackage{pgffor}           
\usepackage{xcolor}           
\usepackage{listings}         
\usepackage{tcolorbox}        
\usepackage{colortbl}         
\usepackage{enumitem}         
\usepackage{url}              
\usepackage[none]{hyphenat}   
\usepackage{xkeyval}          %
\usepackage{pifont}           %
\usepackage{makecell}         
\newcommand{\cmark}{\ding{51}}
\newcommand{\xmark}{\ding{55}}
\PassOptionsToPackage{hypertexnames=false}{hyperref} 
\usepackage{hyperref}         

\ifdefined\neuripsmode
  \usepackage{lmodern}
\fi
\usepackage[mathlines]{lineno} 
\newcommand*\patchAmsMathEnvironmentForLineno[1]{%
  \expandafter\let\csname old#1\expandafter\endcsname\csname #1\endcsname
  \expandafter\let\csname oldend#1\expandafter\endcsname\csname end#1\endcsname
  \renewenvironment{#1}{\linenomath\csname old#1\endcsname}{\csname oldend#1\endcsname\endlinenomath}}%
\newcommand*\patchBothAmsMathEnvironmentsForLineno[1]{%
  \patchAmsMathEnvironmentForLineno{#1}%
  \patchAmsMathEnvironmentForLineno{#1*}}%
\AtBeginDocument{%
  \patchBothAmsMathEnvironmentsForLineno{equation}%
  \patchBothAmsMathEnvironmentsForLineno{align}%
  \patchBothAmsMathEnvironmentsForLineno{gather}%
  \patchBothAmsMathEnvironmentsForLineno{multline}%
}

\usepackage{paperutils}

\usepackage{caption}
\title{%
  Temporal Preference Concepts\\
  and their Functions\\
  in a Large Language Model
}%

\newcommand{\authorentry}[4]{%
  #1 \\ 
  #2 \\ 
  #3 \\ 
}

\newcommand{\eqcontrib}{$^{\dagger}$}%
\newcommand{\eqcontribnote}{\thanks{%
    Equal contribution. Listed alphabetically.
    \\ \qquad
    \quad AISC = AI Safety Camp.
    \quad SPAR = Supervised Program for Alignment Research.
}}%
\newcommand{\authornote}{\thanks{%
    Correspondence to: \texttt{ian@unrulyabstractions.com}.
}}%
\author{%
  \authorentry{Ian Rios-Sialer\authornote}{AISC}{San Francisco, USA}{ian@unrulyabstractions.com}
  \And
  \authorentry{Shantanu Darveshi\eqcontribnote}{AISC}{Mumbai, India}{shantanu.darveshi@gmail.com}
  \And
  \authorentry{Shuai Jiang\eqcontrib}{AISC}{Albuquerque, USA}{marshall.jiang@gmail.com}
  \And
  \authorentry{Avigya Paudel\eqcontrib}{SPAR, AISC}{New York, USA}{https://avigya.vercel.app/}
  \And
  \authorentry{Anastasiia Pronina\eqcontrib}{AISC}{Munich, Germany}{asyadeveloper@gmail.com}
  \And
  \authorentry{Ipshita Bandyopadhyay}{SPAR}{Bangalore, India}{}
  \And
  \authorentry{Justin Shenk}{SPAR, AISC}{Berlin, Germany}{shenk.justin@gmail.com}
}

\begin{document}

\maketitle
\begin{abstract}%
Large Language Models (LLMs) are increasingly being deployed to make decisions that require trading off near-term gains against long-term consequences, yet little is known about how they internally represent or resolve these tradeoffs.
In this work, we causally localize an underlying subgraph for temporal preference in a distilled LLM (\texttt{Qwen3-4B-Instruct-2507}), identifying mid-to-upper-layer nodes through converging evidence from gradient-based attribution and activation patching.
We find that the geometry of time horizon is encoded in the residual stream at the expected localized layers.
A behavioral analysis reveals that unintervened LLMs discount the future several times less steeply than humans, yet this preference is unstable across contexts, motivating explicit control rather than implicit reliance on training.
Finally, we find suggestive evidence that steering vectors can shift temporal preference.
Our work demonstrates how mechanistic interpretability can bring us closer to reliable control over how LLMs plan and reason.
\end{abstract}



\section{Introduction}\label{sec:introduction}

\begin{quote}%
    \itshape
    All your life, you wait for the propitious time. \\
    Then the propitious time \\
    reveals itself as action taken.\\
    \upshape

    \hspace{6em} Louise Gl\"uck, \textit{Landscape}
\end{quote}

Large Language Models (LLMs) appear to hold preferences mediated by abstract concepts such as time.
Impatience and paralysis both push humans into bad decisions.
LLMs risk failing in similar ways.
For now, the consequences have been limited.
No coding agent \href{https://github.com/anthropics/claude-code/issues/42796}{cutting corners}~\citep{betley2025emergent} has caused a catastrophe yet.
But the stakes are rising quickly.
In early 2026, the U.S. Department of Defense and Anthropic publicly clashed over a range of sensitive issues~\citep{amodei-statement, csa-pentagon-anthropic, cnn-anthropic-pentagon}, including whether LLMs should be allowed to autonomously operate weapons.
In high-stakes scenarios, autonomous agents~\citep{un-sg-laws, mitchell-autonomous-agents} would need to trade off short-term gains against long-term effects~\citep{fedus2019hyperbolicdiscountinglearningmultiple}.
When choosing among alternatives, the decision often depends on the temporal scope used to evaluate the consequences~\citep{dohmen2012interpreting, svenson1989decision, gazmararian2025valuing}.
Temporal preference is indeed fundamental for planning~\citep{ameriks2003wealth}, but also for cooperation~\citep{wang2025measuring, kim2023effects} and trust~\citep{fehr2011field}, where agents must bear present costs for future collective benefit~\citep{persson2024intertemporal}.
These intertemporal tradeoffs grow even more consequential in the context of Artificial General Intelligence (AGI).
A myopic system~\citep{ngo2025alignmentproblemdeeplearning} poses different risks than one capable of scheming across long horizons~\citep{meinke2025frontiermodelscapableincontext, park2023aideceptionsurveyexamples}.
Detecting and maintaining control~\citep{greenblatt2024aicontrolimprovingsafety} over these capabilities while that is still tractable motivates our inquiry.
\textbf{Where and how does an LLM encode temporal preference?}

Previous work has investigated the existence of temporal representations~\citep{gurnee2024language}, characterized the economic behavior of LLMs~\citep{cook2026llms, horton2026largelanguagemodelssimulated}, and even shown that risk preference can be steered~\citep{zhu2025steering}.
Yet no work has identified \emph{where} temporal preference lives inside an LLM, how it is geometrically organized, or how to control it through targeted intervention.

Using Mechanistic Interpretability (MI)~\citep{bereska2024mechanisticinterpretabilityaisafety} techniques, we isolate the components that are causally responsible for temporal preference and show how activation-space representations evolve through them.
This offers a geometric perspective on how interventions function to shift temporal preference, even in general open-ended generation tasks.

\begin{figure}[htbp]
  \centering
  \begin{minipage}[c]{0.52\textwidth}
    \centering
    \hspace*{0.74cm}{\small\bfseries Localization}\\[2pt]
    \resizebox{\textwidth}{!}{
%
%
\resizebox{\linewidth}{!}{%
\begin{tikzpicture}[
    x=0.42cm, y=0.50cm,
]


\foreach \l in {6,7,...,35} {
    \pgfmathsetmacro{\xx}{\l + 0.5}
    \draw[black!10, line width=0.25pt] (\xx, 0) -- (\xx, 11.1);
}
\foreach \l in {9,14,19,24,29,34} {
    \pgfmathsetmacro{\xx}{\l + 0.5}
    \draw[black!22, line width=0.35pt] (\xx, 0) -- (\xx, 11.1);
}

\foreach \y in {1.35, 3.65, 5.45, 7.75, 9.55} {
    \draw[black!10, line width=0.25pt] (6.5, \y) -- (35.5, \y);
}

\draw[dashed, thick, black!50] (21, 0) -- (21, 11.1);
\draw[dashed, thick, black!50] (24, 0) -- (24, 11.1);
\draw[dashed, thick, black!50] (31, 0) -- (31, 11.1);
\node[font=\tiny\bfseries, black!70, anchor=south] at (21, 11.1) {L21};
\node[font=\tiny\bfseries, black!70, anchor=south] at (24, 11.1) {L24};
\node[font=\tiny\bfseries, black!70, anchor=south] at (31, 11.1) {L31};


\foreach \l in {7,8,...,35} {
    \node[below, font=\tiny] at (\l, -0.25) {\l};
    \draw[black!35, line width=0.3pt] (\l, -0.05) -- (\l, 0.05);
}
\draw[->] (6.5, 0) -- (35.6, 0) node[right, font=\tiny] {Layer};


\node[anchor=east, font=\scriptsize\bfseries] at (6.2, 10.20) {Probing};
\fill[green!10] (16.5, 10.00) rectangle (35.5, 10.40);   
\fill[green!25] (21.5, 10.00) rectangle (30.5, 10.40);   
\fill[green!50] (24.5, 10.00) rectangle (27.5, 10.40);   
\fill[green!70] (25.5, 10.00) rectangle (26.5, 10.40);   

\node[anchor=east, font=\scriptsize\bfseries] at (6.2, 8.65) {Attr.\ (contr.)};
\fill[red!15]  (30.5, 8.20) rectangle (35.5, 8.60);    
\fill[red!40]  (33.5, 8.20) rectangle (35.5, 8.60);    
\fill[blue!15] (20.5, 8.70) rectangle (26.5, 9.10);    
\fill[blue!40] (21.5, 8.70) rectangle (25.5, 9.10);    
\fill[blue!60] (23.5, 8.70) rectangle (24.5, 9.10);    

\node[anchor=east, font=\scriptsize\bfseries] at (6.2, 6.60) {Attr.\ (param.)};
\fill[black!10] (16.5, 5.90) rectangle (35.5, 6.30);   
\fill[black!30] (16.5, 5.90) rectangle (21.5, 6.30);   
\fill[black!30] (27.5, 5.90) rectangle (35.5, 6.30);   
\fill[red!15]  (27.5, 6.40) rectangle (35.5, 6.80);    
\fill[red!50]  (30.5, 6.40) rectangle (31.5, 6.80);    
\fill[blue!15] (18.5, 6.90) rectangle (24.5, 7.30);    
\fill[blue!40] (19.5, 6.90) rectangle (23.5, 7.30);    

\node[anchor=east, font=\scriptsize\bfseries] at (6.2, 4.55) {Causal (param.)};
\fill[red!15]  (21.5, 4.10) rectangle (35.5, 4.50);    
\fill[red!40]  (30.5, 4.10) rectangle (35.5, 4.50);    
\fill[red!60]  (34.5, 4.10) rectangle (35.5, 4.50);    
\fill[blue!15] (18.5, 4.60) rectangle (25.5, 5.00);    
\fill[blue!40] (20.5, 4.60) rectangle (24.5, 5.00);    
\fill[blue!50] (20.5, 4.60) rectangle (21.5, 5.00);    
\fill[blue!65] (23.5, 4.60) rectangle (24.5, 5.00);    

\node[anchor=east, font=\scriptsize\bfseries] at (6.2, 2.50) {Causal (class.)};
\fill[blue!25] (32.5, 2.80) rectangle (33.5, 3.20);   
\fill[blue!30] (18.5, 2.80) rectangle (19.5, 3.20);   
\fill[blue!30] (20.5, 2.80) rectangle (21.5, 3.20);   
\fill[blue!45] (17.5, 2.80) rectangle (18.5, 3.20);   
\fill[blue!45] (25.5, 2.80) rectangle (26.5, 3.20);   
\fill[blue!45] (29.5, 2.80) rectangle (30.5, 3.20);   
\fill[blue!65] (23.5, 2.80) rectangle (24.5, 3.20);   
\fill[red!15] (10.5, 2.30) rectangle (11.5, 2.70);   
\fill[red!20] (8.5,  2.30) rectangle (9.5,  2.70);   
\fill[red!25] (26.5, 2.30) rectangle (27.5, 2.70);   
\fill[red!27] (7.5,  2.30) rectangle (8.5,  2.70);   
\fill[red!27] (9.5,  2.30) rectangle (10.5, 2.70);   
\fill[red!30] (17.5, 2.30) rectangle (18.5, 2.70);   
\fill[black!8]  (19.5, 1.80) rectangle (27.5, 2.20);   
\fill[black!22] (20.5, 1.80) rectangle (21.5, 2.20);   
\fill[black!22] (23.5, 1.80) rectangle (25.5, 2.20);   
\fill[black!22] (26.5, 1.80) rectangle (27.5, 2.20);   
\fill[black!40] (21.5, 1.80) rectangle (22.5, 2.20);   

\node[anchor=east, font=\scriptsize\bfseries] at (6.2, 0.70) {Error char.};
\fill[orange!15] (18.5, 0.50) rectangle (31.5, 0.90);  
\fill[orange!40] (23.5, 0.50) rectangle (27.5, 0.90);  
\fill[orange!60] (24.5, 0.50) rectangle (25.5, 0.90);  

\fill[blue!50]   (6.5, -1.50) rectangle (7.1, -1.10);
\node[anchor=west, font=\tiny] at (7.1, -1.30) {Attn};
\fill[red!50]    (9.5, -1.50) rectangle (10.1, -1.10);
\node[anchor=west, font=\tiny] at (10.1, -1.30) {MLP};
\fill[black!40]  (12.5, -1.50) rectangle (13.1, -1.10);
\node[anchor=west, font=\tiny] at (13.1, -1.30) {Resid};
\fill[green!50]  (16.0, -1.50) rectangle (16.6, -1.10);
\node[anchor=west, font=\tiny] at (16.6, -1.30) {Probe};
\fill[orange!50] (19.5, -1.50) rectangle (20.1, -1.10);
\node[anchor=west, font=\tiny] at (20.1, -1.30) {Error};

\end{tikzpicture}%
}}
    \vspace{4pt}
    \resizebox{\textwidth}{!}{
\begin{tikzpicture}[x=0.5cm, y=0.65cm]

\node[font=\small\bfseries, anchor=south] at (-0.5, 2.55) {Steering};

\draw[thick, ->] (-4.5,0) -- (3.2,0);
\node[below, font=\tiny] at (-4,0) {$-4$};
\node[below, font=\tiny] at (-2,0) {$-2$};
\node[below, font=\tiny] at (0,0) {$0$};
\node[below, font=\tiny] at (2,0) {$+2$};
\node[below=3pt, font=\tiny] at (-0.5, -0.12) {temporal orientation score};

\node[above, font=\tiny\bfseries, text=red!70] at (-4.2, 0.05) {short-term};
\node[above, font=\tiny\bfseries, text=blue!70] at (2.9, 0.05) {long-term};

\def\rowC{0.6}
\def\rowB{1.25}
\def\rowA{1.9}

\node[left, font=\tiny\bfseries] at (-4.6, \rowA) {L21};
\node[left, font=\tiny\bfseries] at (-4.6, \rowB) {L22};
\node[left, font=\tiny\bfseries, black!50] at (-4.6, \rowC) {L26};

\draw[decorate, decoration={brace, amplitude=2pt, mirror}, thick, black!50]
  (-6.0, \rowB-0.18) -- (-6.0, \rowA+0.18);
\node[left, font=\tiny, black!50, align=right] at (-6.2, {(\rowA+\rowB)/2}) {steering\\layers};

\draw[-{Latex[length=1.3mm]}, thick, red!60] (0,\rowA) -- (-2.9,\rowA);
\node[above=-1pt, font=\tiny, red!70] at (-1.5, \rowA) {$\alpha\!=\!-50$};
\draw[-{Latex[length=1.3mm]}, thick, blue!60] (0,\rowA) -- (2.7,\rowA);
\node[above=-1pt, font=\tiny, blue!70] at (1.35, \rowA) {$\alpha\!=\!+50$};
\fill[black] (0,\rowA) circle (1.3pt);

\draw[-{Latex[length=1.3mm]}, thick, red!60] (0,\rowB) -- (-4.0,\rowB);
\node[above=-1pt, font=\tiny, red!70] at (-2.0, \rowB) {$\alpha\!=\!-50$};
\draw[-{Latex[length=1.3mm]}, thick, blue!60] (0,\rowB) -- (2.2,\rowB);
\node[above=-1pt, font=\tiny, blue!70] at (1.1, \rowB) {$\alpha\!=\!+50$};
\fill[black] (0,\rowB) circle (1.3pt);

\draw[-{Latex[length=1.3mm]}, thick, red!25] (0,\rowC) -- (-0.1,\rowC);
\draw[-{Latex[length=1.3mm]}, thick, blue!25] (0,\rowC) -- (1.0,\rowC);
\fill[black] (0,\rowC) circle (1.3pt);

\end{tikzpicture}}
  \end{minipage}%
  \hfill
  \begin{minipage}[c]{0.44\textwidth}
    \centering
    {\small\bfseries Geometry}\\[2pt]
    \includegraphics[width=\textwidth]{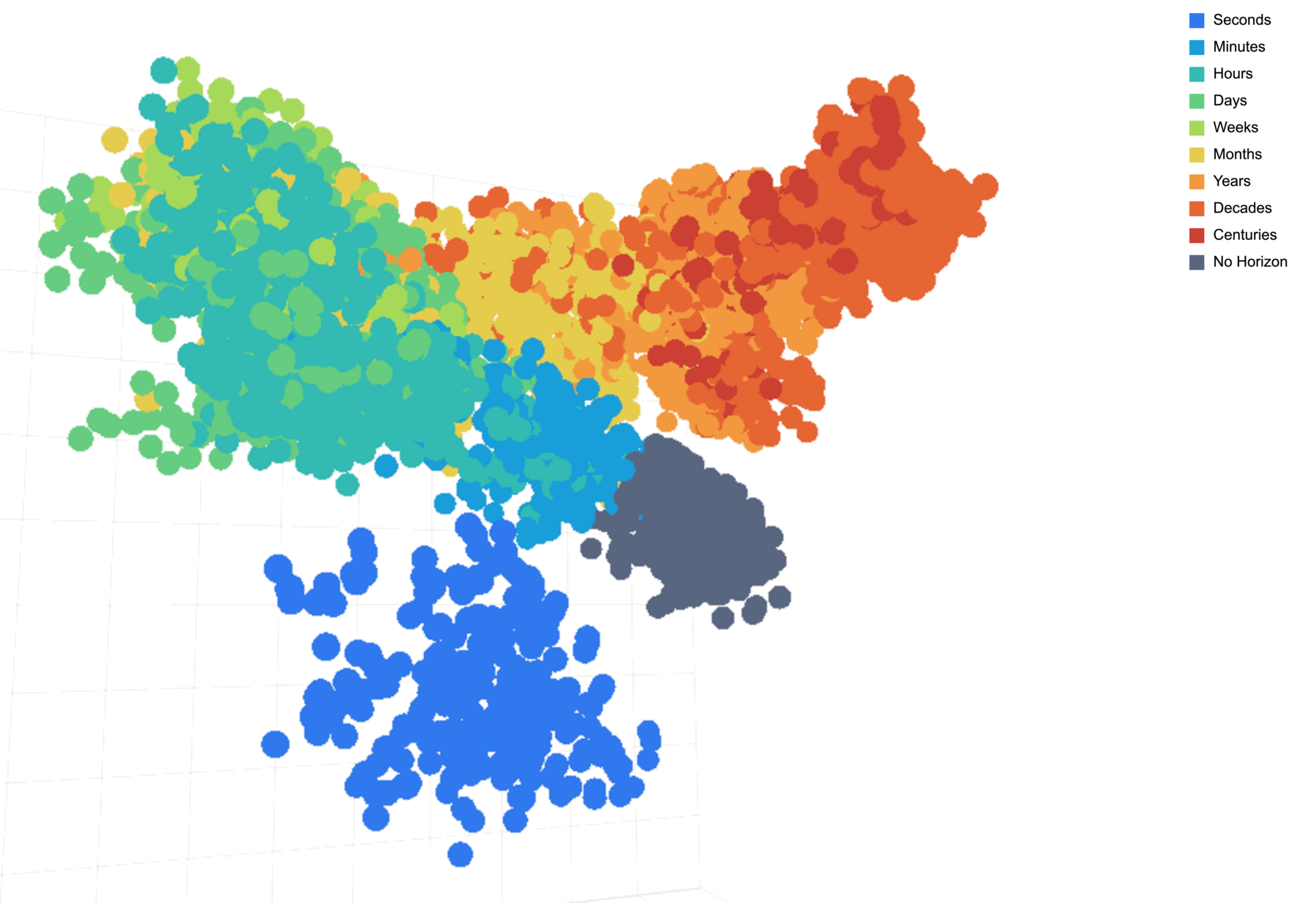}
  \end{minipage}
  \caption{(Top-left) Six rows summarize localization findings. Four preference-targeted methods (probing, attributional contrastive, attributional parametric, causal parametric) converge on a subgraph in layers 17--35; a fifth (causal classification) tests whether the same subgraph supports related horizon inference.
  Darker shading indicates stronger signal. The bottom row (\textit{Error char.}) shows that an unrelated meta-cognitive variable, cumulative reasoning reliability, is also decodable above 95\% across L19--L31 within the same subgraph, indicating the localized region encodes more than just temporal preference (Section~\ref{sec:localize-results}; \ref{app:error-monitoring}).
  (Top-right) Time horizon geometry within the identified subgraph (Section~\ref{sec:characterize-results}).
  (Bottom) CAA steering shifts temporal preference bidirectionally at layers 19--22 but weakly at the best probing layer (L26), illustrating the probing--steering dissociation (Section~\ref{sec:steering}).}
  \label{fig:main-pic}
\end{figure}
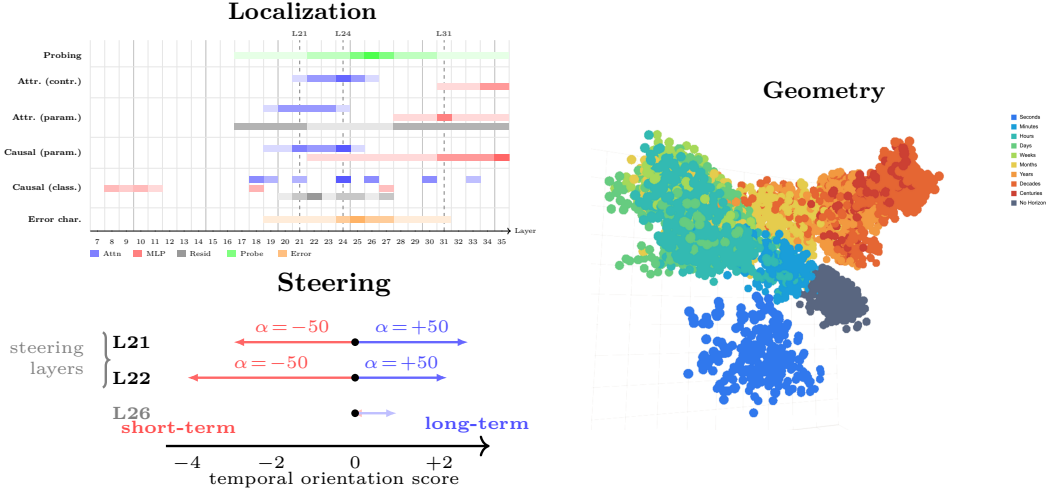

We focus on \texttt{Qwen3-4B-Instruct-2507}: its non-thinking-only operation keeps computation inside a fixed prompt template (enabling token-aligned attribution and patching), and its latent preferences are stable under prompt perturbations, trading breadth across model families for depth in tracing a concept from localization through geometry to intervention.
Our methodology integrates four localization pipelines for the temporal-preference subgraph, a fifth that probes its reach into categorical horizon inference, and dedicated behavioral and steering instruments.
The pipelines vary in prompting paradigm, localization technique, scale, and resolution.
Because pipelines approach localization from fundamentally different angles, their independent convergence on the common subgraph components strongly suggests that our findings reflect the genuine model structure.

Our work establishes that \textbf{temporal preference is localizable} within LLMs and, given the behavioral inconsistencies we observe, \textbf{should be explicitly controlled} rather than left to emerge implicitly from training.
The convergence of multiple separate paradigms on the same subgraph demonstrates the value of \textbf{complementary experimental approaches} to validate mechanistic claims.
Finally, while current interpretability methods are well-suited for binary contrastive concepts (truthfulness~\citep{marks2024geometry}, refusal~\citep{NEURIPS2024_f5454485}, sycophancy~\citep{panickssery2024steering}), this work takes initial steps toward \textbf{steering dimensional concepts} such as time, uncertainty, or risk preference, an underexplored area that warrants further development.

We summarize our contributions as follows:

\emph{Causal Localization of Temporal Preference} (Section~\ref{sec:localize-results})
\begin{itemize}[noitemsep, topsep=0pt, leftmargin=*]
    \item We provide causal and attributional localization of a temporal-preference subgraph in \texttt{Qwen3-4B-Instruct-2507} using complementary methods, including standard attribution patching, activation patching, linear probing, and CAA (Section~\ref{sec:methodology}), and show that its L24 attention layer is also recruited for categorical horizon inference (\ref{app:causal-contrastive}).
\end{itemize}
\emph{Characterization of Temporal Preference} (Section~\ref{sec:characterize-results})
\begin{itemize}[noitemsep, topsep=0pt, leftmargin=*]
    \item We show that time horizon has non-linear geometry within the subgraph (Section~\ref{sec:characterize-results}).
    \item We identify the user-to-assistant turn boundary (the change-of-turn token sequence \texttt{<|im\_end|>} $\to$ \texttt{\textbackslash n} $\to$ \texttt{<|im\_start|>} $\to$ \texttt{assistant}) as the site where attention collapses the continuous horizon manifold into a committed binary preference (Section~\ref{sec:characterize-results}).
    \item We provide a behavioral analysis (Section~\ref{sec:characterize-results}) that shows that unintervened LLMs behave very differently from humans, suggesting that the implicit time preference is inconsistent between contexts.
\end{itemize}
\emph{Steering of Temporal Preference} (Section~\ref{sec:steering})
\begin{itemize}[noitemsep, topsep=0pt, leftmargin=*]
    \item We show successful interventions that change temporal preference and interpret them through our characterizations, via the steering methodology detailed in \ref{app:contrastive-steering-methods}.
\end{itemize}

Our methodology combines multiple independent methods, datasets, and resolutions that approach the temporal-preference subgraph from different angles, to build converging evidence for how it is mechanistically implemented.

\newcommand{\perishablefootnote}{%
    \footnote{Although not explored in this work, when rewards are \textit{perishable}~\citep{econometrica}, the temporal scope is also bounded below by the \textit{retroactive reach}~\citep{retroactive-date, Mitchell2005}.}%
}

\section{Background}\label{sec:background}

We refer to \textbf{temporal preference} as the degree to which an agent values outcomes differently depending on when they occur~\citep{cohen2020measuring, wang2025measuring}.
We use the term \textbf{time horizon} to denote the future moment at which outcomes are evaluated against an objective~\citep{ebert1973time}.
Because future events do not affect past outcomes, the time horizon also acts as a \textit{constraint} on planning~\citep{reilly2016time}.
It would be \textit{instrumentally}~\citep{korsgaard1997} or \textit{means-end}~\citep{bratman1981} \textbf{incoherent} for the agent to choose actions that are incapable of causing effects by the specified deadline.
Then, the \textbf{temporal scope} is the bounded interval of time over which an agent weighs the results according to its preference\perishablefootnote.

\begin{figure}[htbp]
    \centering
    \resizebox{\textwidth}{!}{

\definecolor{colTitle}{HTML}{6B7D8E}
\definecolor{colHorizon}{HTML}{1A1A2E}
\definecolor{colBoundary}{HTML}{3D5A80}
\definecolor{colCurve}{HTML}{CC785C}
\definecolor{colFill}{HTML}{D4956B}
\definecolor{colScope}{HTML}{C4703E}
\definecolor{colAxis}{HTML}{A0A8B0}
\definecolor{colLabel}{HTML}{5A6570}
\definecolor{colGrayCurve}{HTML}{A0A8B0}
\definecolor{colDark}{HTML}{2D3748}
\definecolor{colRewardFill}{HTML}{5BA4B5}
\definecolor{colValueA}{HTML}{CC785C}
\definecolor{colValueB}{HTML}{5BA4B5}
\definecolor{colOp}{HTML}{1A1A2E}

\pgfmathdeclarefunction{lam}{1}{%
  \pgfmathparse{exp(-0.13*(#1))*(1+0.15*sin(70*(#1))+0.08*cos(130*(#1))+0.05*sin(200*(#1)))}%
}
\pgfmathdeclarefunction{rew}{1}{%
  \pgfmathparse{1.6*(#1)/(1+0.2*(#1))}%
}
\pgfmathdeclarefunction{val}{1}{%
  \pgfmathparse{lam(#1)*rew(#1)}%
}

\newcommand{\TT}{44}
\newcommand{\OO}{80}
\newcommand{\symscale}{1.25}
\newcommand{\tscale}{1}

\newcommand{\aspectR}{0.50}
\newcommand{\Pscale}{2.0}
\pgfmathsetmacro{\Pw}{10*\Pscale}
\pgfmathsetmacro{\Ph}{\Pw*\aspectR}

\newcommand{\tHorizVal}{6.75}
\pgfmathsetmacro{\pHoriz}{\tHorizVal*\Pscale}

\pgfmathsetmacro{\valAtHoriz}{\Ph*0.8*lam(\tHorizVal)*rew(\tHorizVal)/2.6}
\pgfmathsetmacro{\rewAtHoriz}{\Ph*0.85*rew(\tHorizVal)/5.5}
\pgfmathsetmacro{\lamAtHoriz}{\Ph*lam(\tHorizVal)}

\newcommand{\opGap}{3.5}
\newcommand{\opW}{2.0}

\pgfmathsetmacro{\eqX}{\Pw+\opGap+\opW/2}
\pgfmathsetmacro{\panBx}{\Pw+\opGap+\opW+\opGap}
\pgfmathsetmacro{\mulX}{\panBx+\Pw+\opGap+\opW/2}
\pgfmathsetmacro{\panCx}{\panBx+\Pw+\opGap+\opW+\opGap}

\newcommand{\titleGY}{16.0}
\newcommand{\symGY}{12.5}
\pgfmathsetmacro{\opY}{\Ph/2}

\begin{tikzpicture}[>=Stealth]

\node[colTitle, anchor=south, font=\fontsize{\TT}{\TT}\selectfont\bfseries]
  at ({\Pw/2}, \titleGY) {Value};
\node[colTitle, anchor=south, font=\fontsize{\TT}{\TT}\selectfont\bfseries]
  at ({\panBx+\Pw/2}, \titleGY) {Temporal Preference};
\node[colTitle, anchor=south, font=\fontsize{\TT}{\TT}\selectfont\bfseries]
  at ({\panCx+\Pw/2}, \titleGY) {Reward};

\node[colLabel, anchor=south, font=\fontsize{\TT}{\TT}\selectfont]
  at ({\Pw/2}, \symGY) {\scalebox{\symscale}{$V(t)$}};
\node[colLabel, anchor=south, font=\fontsize{\TT}{\TT}\selectfont]
  at ({\panBx+\Pw/2}, \symGY) {\scalebox{\symscale}{$\lambda(t)$}};
\node[colLabel, anchor=south, font=\fontsize{\TT}{\TT}\selectfont]
  at ({\panCx+\Pw/2}, \symGY) {\scalebox{\symscale}{$r(t)$}};

\begin{scope}[shift={(0,0)}]

  \draw[-{Stealth[length=5pt,width=3pt]}, colAxis, line width=0.5pt]
    (0,0) -- ({\Pw+0.5},0);
  \draw[-{Stealth[length=5pt,width=3pt]}, colAxis, line width=0.5pt]
    (0,0) -- (0,{\Ph+0.5});

  \draw[colBoundary, line width=1.5pt, dashed] ({\pHoriz},0) -- ({\pHoriz},{\valAtHoriz});

  \begin{scope}
    \clip (0,0) rectangle ({\pHoriz},{\Ph+1});
    \fill[colValueB, opacity=0.15]
      plot[domain=0:10, samples=100, smooth]
        ({\x*\Pscale}, {\Ph*0.8*val(\x)/2.6}) -- ({\Pw},0) -- (0,0) -- cycle;
  \end{scope}
  \begin{scope}
    \clip (0,0) rectangle ({\pHoriz},{\Ph+1});
    \fill[colValueA, opacity=0.18]
      plot[domain=0:10, samples=100, smooth]
        ({\x*\Pscale}, {\Ph*0.8*val(\x)/2.6}) -- ({\Pw},0) -- (0,0) -- cycle;
  \end{scope}

  \begin{scope}
    \clip (0,0) rectangle ({\pHoriz},{\Ph+1});
    \draw[colDark, line width=3pt, smooth]
      plot[domain=0:10, samples=100] ({\x*\Pscale}, {\Ph*0.8*val(\x)/2.6});
  \end{scope}
  \begin{scope}
    \clip ({\pHoriz},0) rectangle ({\Pw+1},{\Ph+1});
    \draw[colGrayCurve, line width=3pt, opacity=0.55, dashed]
      plot[domain=0:10, samples=100, smooth] ({\x*\Pscale}, {\Ph*0.8*val(\x)/2.6});
  \end{scope}

  \node[colScope, font=\fontsize{44}{46}\selectfont\bfseries, anchor=north, align=center]
    at ({0.5*\pHoriz}, -0.8) {Temporal\\Scope};

  \node[colLabel, anchor=north west, font=\fontsize{\TT}{\TT}\selectfont]
    at ({\Pw+0.5}, -0.1) {\scalebox{\tscale}{$t$}};

\end{scope}

\node[colOp, font=\fontsize{\OO}{\OO}\selectfont\bfseries]
  at (\eqX, \opY) {\textsf{=}};

\begin{scope}[shift={(\panBx, 0)}]

  \begin{scope}
    \clip (0,0) rectangle ({\pHoriz},{\Ph});
    \fill[colFill, opacity=0.18]
      plot[domain=0:10, samples=120, smooth]
        ({\x*\Pscale}, {\Ph*lam(\x)}) -- ({\Pw},0) -- (0,0) -- cycle;
  \end{scope}

  \draw[-{Stealth[length=5pt,width=3pt]}, colAxis, line width=0.5pt]
    (0,0) -- ({\Pw+0.5},0);
  \draw[-{Stealth[length=5pt,width=3pt]}, colAxis, line width=0.5pt]
    (0,0) -- (0,{\Ph+0.5});

  \begin{scope}
    \clip (0,0) rectangle ({\pHoriz},{\Ph});
    \draw[colCurve, line width=4pt]
      plot[domain=0:10, samples=120, smooth] ({\x*\Pscale}, {\Ph*lam(\x)});
  \end{scope}
  \begin{scope}
    \clip ({\pHoriz},0) rectangle ({\Pw+1},{\Ph});
    \draw[colGrayCurve, line width=3.5pt, opacity=0.55, dashed]
      plot[domain=0:10, samples=120, smooth] ({\x*\Pscale}, {\Ph*lam(\x)});
  \end{scope}

  \draw[colBoundary, line width=5pt] ({\pHoriz},0) -- ({\pHoriz},{\lamAtHoriz+1.5});
  \node[colHorizon, font=\fontsize{44}{46}\selectfont\bfseries, anchor=south]
    at ({\pHoriz}, {\Ph*0.82}) {Time Horizon};

  \node[colLabel, anchor=north west, font=\fontsize{\TT}{\TT}\selectfont]
    at ({\Pw+0.5}, -0.1) {\scalebox{\tscale}{$t$}};

\end{scope}

\node[colOp, font=\fontsize{\OO}{\OO}\selectfont\bfseries]
  at (\mulX, \opY) {\textsf{×}};

\begin{scope}[shift={(\panCx, 0)}]

  \begin{scope}
    \clip (0,0) rectangle ({\Pw},{\Ph+1});
    \fill[colRewardFill, opacity=0.15]
      plot[domain=0:10, samples=80, smooth]
        ({\x*\Pscale}, {\Ph*0.85*rew(\x)/5.5}) -- ({\Pw},0) -- (0,0) -- cycle;
  \end{scope}

  \draw[-{Stealth[length=5pt,width=3pt]}, colAxis, line width=0.5pt]
    (0,0) -- ({\Pw+0.5},0);
  \draw[-{Stealth[length=5pt,width=3pt]}, colAxis, line width=0.5pt]
    (0,0) -- (0,{\Ph+0.5});

  \draw[colDark, line width=3pt, smooth]
    plot[domain=0:10, samples=80] ({\x*\Pscale}, {\Ph*0.85*rew(\x)/5.5});

  \draw[colBoundary, line width=3pt, dash pattern=on 6pt off 4pt] ({\pHoriz},0) -- ({\pHoriz},{\rewAtHoriz});

  \node[colLabel, anchor=north west, font=\fontsize{\TT}{\TT}\selectfont]
    at ({\Pw+0.5}, -0.1) {\scalebox{\tscale}{$t$}};

\end{scope}

\pgfresetboundingbox
\path[use as bounding box] (-0.5,-4.5) rectangle ({\panCx+\Pw+2.5},\titleGY+2);

\end{tikzpicture}}
    \caption{
        The time horizon specifies when the consequences of a decision are assessed.
        The temporal scope is then bounded above by the time horizon.
    }
    \label{fig:temporal-defs}
\end{figure}
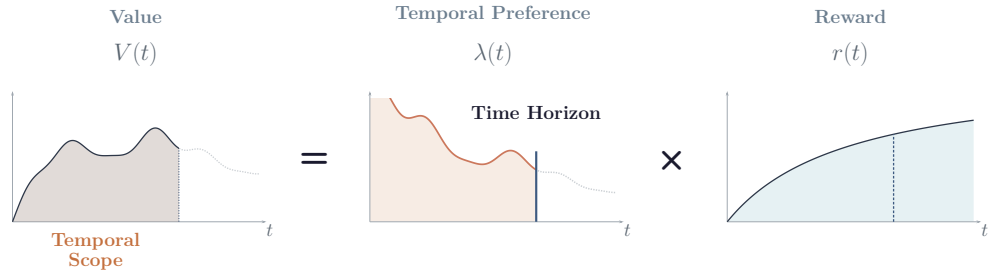
\FloatBarrier

The empirical handle on temporal preference is \textbf{intertemporal choice}: a decision between options that differ in both reward and delay~\citep{frederick2002time, green2004discounting}.
Each option $i$ is a tuple $(r_i, t_i)$ of reward $r_i \in \mathbb{R}^{+}$ and delay $t_i \in \mathbb{R}^{+}$.
Its subjective value is the reward scaled by a delay-dependent temporal-preference weight $\lambda$:
\begin{equation}
  V_i \;=\; \lambda(t_i) \cdot r_i\, ,
  \label{eq:temporal-value}
\end{equation}
and, given a pair of options $\{A, B\}$, an agent with preference $\lambda$ selects
\begin{equation}
  i^{*} \;=\; \operatorname*{arg\,max}_{i \in \{A, B\}} \; \lambda(t_i) \cdot r_i\, .
  \label{eq:intertemporal-choice}
\end{equation}
Humans are typically modeled with hyperbolic discount functions~\citep{frederick2002time, mazyaki2025temporal}; for LLMs, whether the same functional form fits is an open empirical question~\citep{mazyaki2025temporal}.
Characterizing LLM behavior therefore requires fitting $\lambda$ via regression, assessing its stability across contexts, and benchmarking the resulting preferences against human intertemporal choice.

In humans, these concepts have localized neural representations~\citep{kable2007neural} that predict behavior~\citep{shamosh2008delay}, respond causally to intervention~\citep{figner2010lateral}, and exhibit internal organization interpretable as a \emph{functional role}~\citep{cummins1975functional}.\footnote{Some authors argue that concepts are best modeled by geometric or topological spaces~\citep{gardenfors2000conceptual}, a perspective that resonates with our geometric analysis of temporal representations in activation space.}
Our work asks whether temporal preference exists in an LLM in an analogous way: localized, predictive, causally efficacious, and geometrically organized.

\subsection{Locate and characterize, then steer}\label{sec:interp-primitives}

Our pipeline engages the target concept in three complementary modes.
To \emph{locate} the subgraph responsible for temporal preference, we pair \emph{causal} intervention, activation patching~\citep{heimersheim2024useinterpretactivationpatching} in do-calculus notation~\citep{pearl2009causality}, with cheaper \emph{attributional} proxies that scale across inputs: gradient-based EAP-IG~\citep{hanna2024faithfaithfulnessgoingcircuit, bereska2024mechanisticinterpretabilityaisafety} and linear probes~\citep{mueller2025mibmechanisticinterpretabilitybenchmark, kim2025linearrepresentationspoliticalperspective} that surface where the concept linearly emerges.
To \emph{characterize} how the localized components encode horizon information, we apply PCA~\citep{shlens2014} inside the subgraph; prior work warns that concept geometry is often non-linear~\citep{engels2025notall, modell2025origins, gurnee2026models} and can drift across generation~\citep{lampinen2026linearrepresentationslanguagemodels}, so a single global direction rarely tells the whole story.
Only after we have located and characterized the subgraph do we \emph{steer}: we inject a probe-derived vector~\citep{turner2023activation, panickssery2024steering} at inference time; localization is not strictly required but tightens precision, shrinks magnitudes, and reduces side effects~\citep{zhang2026locatesteerimprovepractical}.
Full definitions are in~\ref{app:extended-background}.

\subsection{Related work}\label{sec:related-work}

Four strands of work frame this paper: (i) \emph{temporal representation}, showing that LLMs encode time geometrically~\citep{gurnee2024language, engels2025notall, kantamneni2025trigonometry, gurnee2026models} as locally linear features on globally curved manifolds~\citep{modell2025origins, park2024linear}; (ii) \emph{temporal reasoning and planning}, where models fail despite the geometric encoding~\citep{wang2024trambenchmarkingtemporalreasoning, sehgal2026realtimedeadlinesrevealtemporal, wang2026reasoningfailsplanplanningcentric}; (iii) \emph{LLM economic behavior}, reproducing human biases~\citep{horton2026largelanguagemodelssimulated, cook2026llms, wang2025prospect} with entangled risk/time preferences~\citep{zhu2025steering, moghimi2026decouplingtimeriskrisksensitive, mazyaki2025temporal}; and (iv) \emph{steering advancements}, from activation addition~\citep{turner2023activation, panickssery2024steering} through sparse dictionaries~\citep{cunningham2023sparse} to geometry-aware methods~\citep{vu2025angular, curveball2026, li2026svf, postmus2025conceptors}, with known failure modes at large $|\alpha|$~\citep{wolf2024tradeoffs, braun2025understandingunreliabilitysteeringvectors}.
No prior work has localized a subgraph functionally responsible for temporal preference, characterized the geometry of the causal representation, or steered along it.
Full discussion in~\ref{app:extended-literature}.
\section{Methodology}\label{sec:methodology}

Our methodology follows three stages: \emph{localize} the temporal-preference subgraph, \emph{characterize} its representations, and \emph{intervene} to steer it.
Localization pairs \emph{wide attribution} (contrastive A/B prompts $\times$ EAP-IG and linear probing, cheap to sweep across hundreds of components) with \emph{targeted intervention} (parametric prompts with explicit horizons $\times$ activation patching, expensive but causal); the two paradigms converge on the same subgraph, which is the basis for our localization claim. A complementary classification pipeline (IOI-style prompts $\times$ directional patching) tests whether the same subgraph generalizes from valuation to categorical horizon inference.
Characterization applies PCA inside that subgraph to examine how explicit horizons organize the activation manifold and whether latent (no-horizon) preferences align with that geometry, paired with two behavioral instruments (Kirby MCQ-27 and a 30-model investment-coherence questionnaire) that test whether the geometry actually drives choice.
Intervention uses Contrastive Activation Addition with a probe-derived steering vector, swept across layers and magnitudes to test for a probing-steering dissociation.
Full per-pipeline protocols, dataset construction, metric definitions, and the reader's guide to Part~4 are in the Methodology Summary (\ref{app:methodology-summary}).

\section{Experimental setup}\label{sec:experimental-setup}

We focus on a single model, \texttt{Qwen3-4B-Instruct-2507}, a mode-specialized non-thinking refresh of \texttt{Qwen3-4B}~\citep{qwen2025instruct2507, yang2025qwen3}.
We chose this model because operating in non-thinking mode keeps all ``cognition'' inside a fixed prompt template (no \texttt{<think>} block perturbs token positions), which is what the activation-patching and attribution pipelines need to align clean and corrupted runs; because its latent preference is stable under minor prompt perturbations at a scale where similar-sized models drift; and because it is small enough for repeated attribution and intervention sweeps.
The pipeline operates on four dataset paradigms: minimally-framed A/B prompts (500 explicit + 500 implicit pairs), highly-formatted parametric prompts with explicit time horizons (4{,}588 prompts), IOI-style classification prompts (160 short/long pairs), and behavioral questionnaires (Kirby MCQ-27 plus a 960-prompt investment-coherence instrument run on 30 models).
All experiments fit on a MacBook Pro (M4 Max, 48\,GB) except for the causal classification pipeline, which requires 79\,GB (\ref{app:causal-contrastive-methods}), and are reproducible end-to-end within two weeks.
Full configurations, dataset construction, and model-selection rationale are in~\ref{app:experimental-details}.

\section{Results}\label{sec:results}

\FloatBarrier
\subsection{Where is temporal preference for the LLM?}\label{sec:localize-results}

\begin{figure}[htbp]
  \centering
  \begin{minipage}{0.85\linewidth}\centering
    \begin{minipage}[t]{0.49\linewidth}\centering
      {\footnotesize\bfseries Causal parametric}\\[2pt]
      \includegraphics[width=\linewidth]{images/localize/causal_parametric/component_importance_ranked}\\[6pt]
      {\footnotesize\bfseries Attributional contrastive}\\[2pt]
      \includegraphics[width=\linewidth]{images/localize/attributional_contrastive/line_layer_distribution}
    \end{minipage}\hfill
    \begin{minipage}[t]{0.49\linewidth}\centering
      {\footnotesize\bfseries Causal classification}\\[2pt]
      \includegraphics[width=\linewidth]{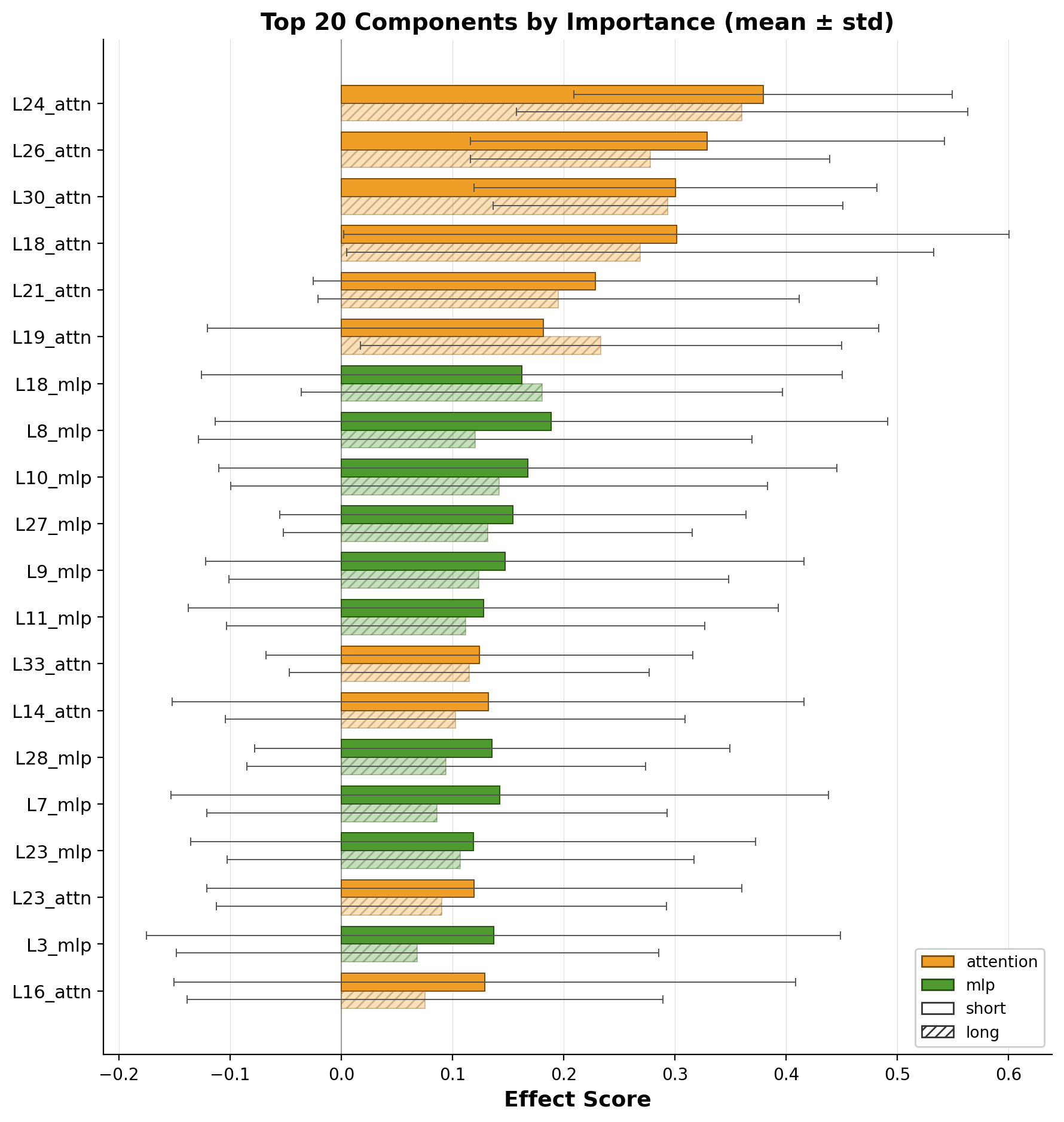}
    \end{minipage}
  \end{minipage}
  \caption{Localization evidence converges on a common subgraph.
  \textbf{Top-left:} top components ranked by mean causal effect under the parametric paradigm ($n=71$ pairs), with \texttt{L24\_attn} the only attention component above 0.5 noising disruption.
  \textbf{Right:} top-20 components from causal classification ($n=160$ pairs).
  \textbf{Bottom-left:} EAP-IG top-$k$ attribution mass per layer on contrastive prompts (short-term and long-term flips), peaking sharply at L24 with a secondary peak at L31--L35 and stable across $k$ (\ref{app:attributional-contrastive}).
  All three methods rank \texttt{L24\_attn} as the highest-effect attention component and consistently surface \texttt{L21\_attn} among the leading contributors.
  The preference-targeted methods specify \texttt{L35\_mlp} and \texttt{L31\_mlp} as the top-contributing MLP components, consistent with the cross-method agreement in Table~\ref{tab:convergence} (\ref{app:convergence}).
  The classification-targeted method reveals a mid-network MLP cluster instead (L8--L11, L18), with a late L27 contributor (\ref{app:causal-contrastive}). \label{fig:top-components}}
\end{figure}

Four preference-localization methods (probing, attributional contrastive, attributional parametric, causal parametric) converge on layers 17--35 (Figure~\ref{fig:main-pic}, top-left; \ref{app:convergence}); 
a fifth method (causal classification) suggests that the core attention layers are also recruited for related horizon inference (\ref{app:causal-contrastive}).
L24 attention is flagged by all four non-probing methods.
MLP effects concentrate in L31--L35 across attributional contrastive, attributional parametric, and causal parametric patching; the classification run additionally surfaces a classification-specific mid-network MLP cluster at L8--L11 outside L17--35 zone.
Probes peak at layer 26 (99.2\%; \ref{app:contrastive-probing-linear}).
Parametric activation patching ranks the four highest-importance components as \texttt{L24\_attn}, \texttt{L35\_mlp}, \texttt{L31\_mlp}, and \texttt{L21\_attn}, separated from the remaining components in effect size (Figure~\ref{fig:top-components}, top-left).
Gradient-based attribution on the contrastive prompts independently concentrates top-$k$ attribution mass at the same layers, with a sharp peak at L24 and a secondary peak at L31--L35 that is robust across $k$ (Figure~\ref{fig:top-components}, bottom-left inset).

\FloatBarrier

\FloatBarrier
\subsection{What is the LLM's temporal preference like?}\label{sec:characterize-results}

\begin{figure}[htbp]
  \centering
  \begin{minipage}[t]{0.245\linewidth}\centering
    \includegraphics[width=\linewidth]{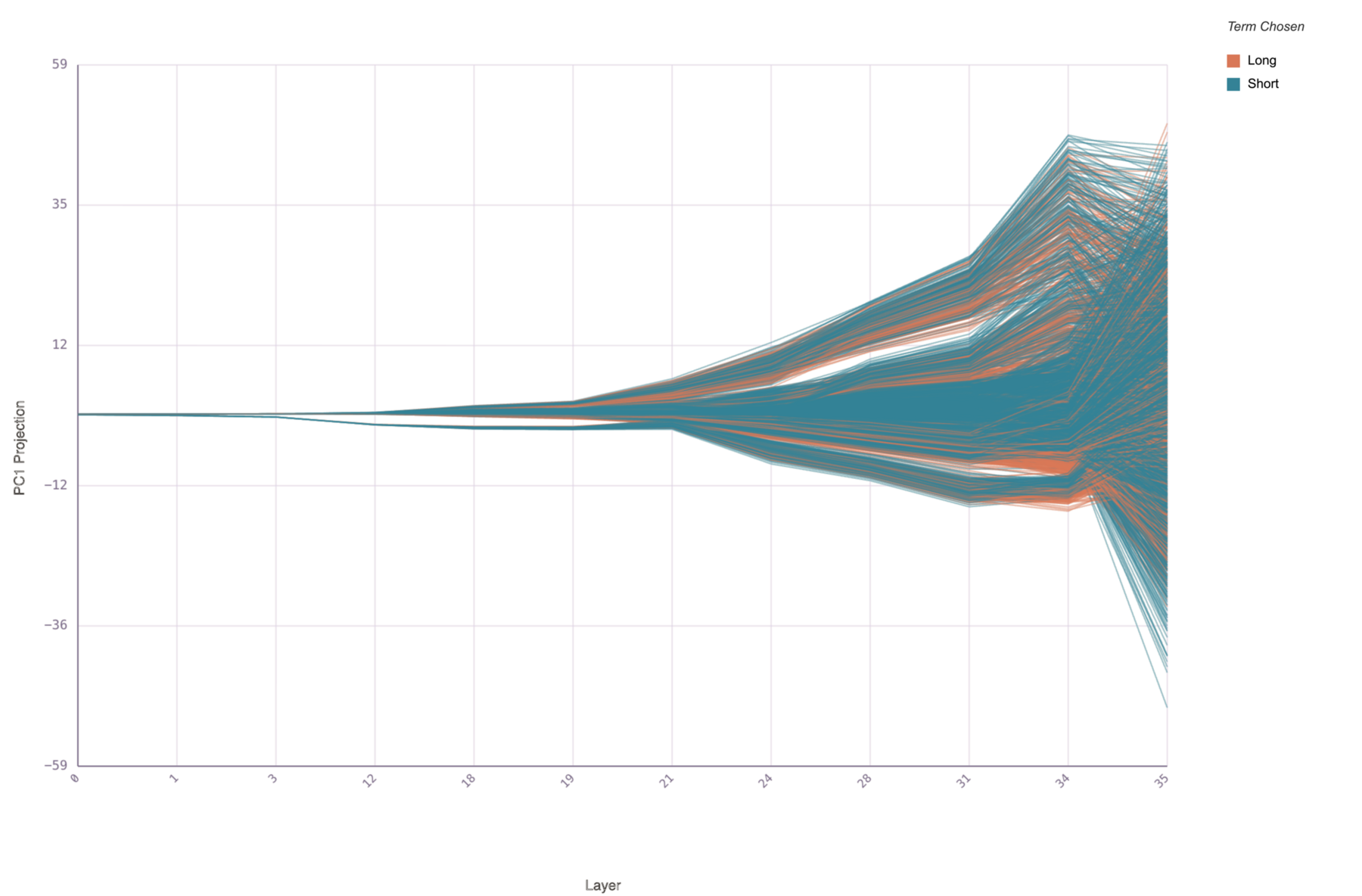}\\[2pt]
    {\footnotesize\texttt{<|im\_end|>}}
  \end{minipage}\hfill
  \begin{minipage}[t]{0.245\linewidth}\centering
    \includegraphics[width=\linewidth]{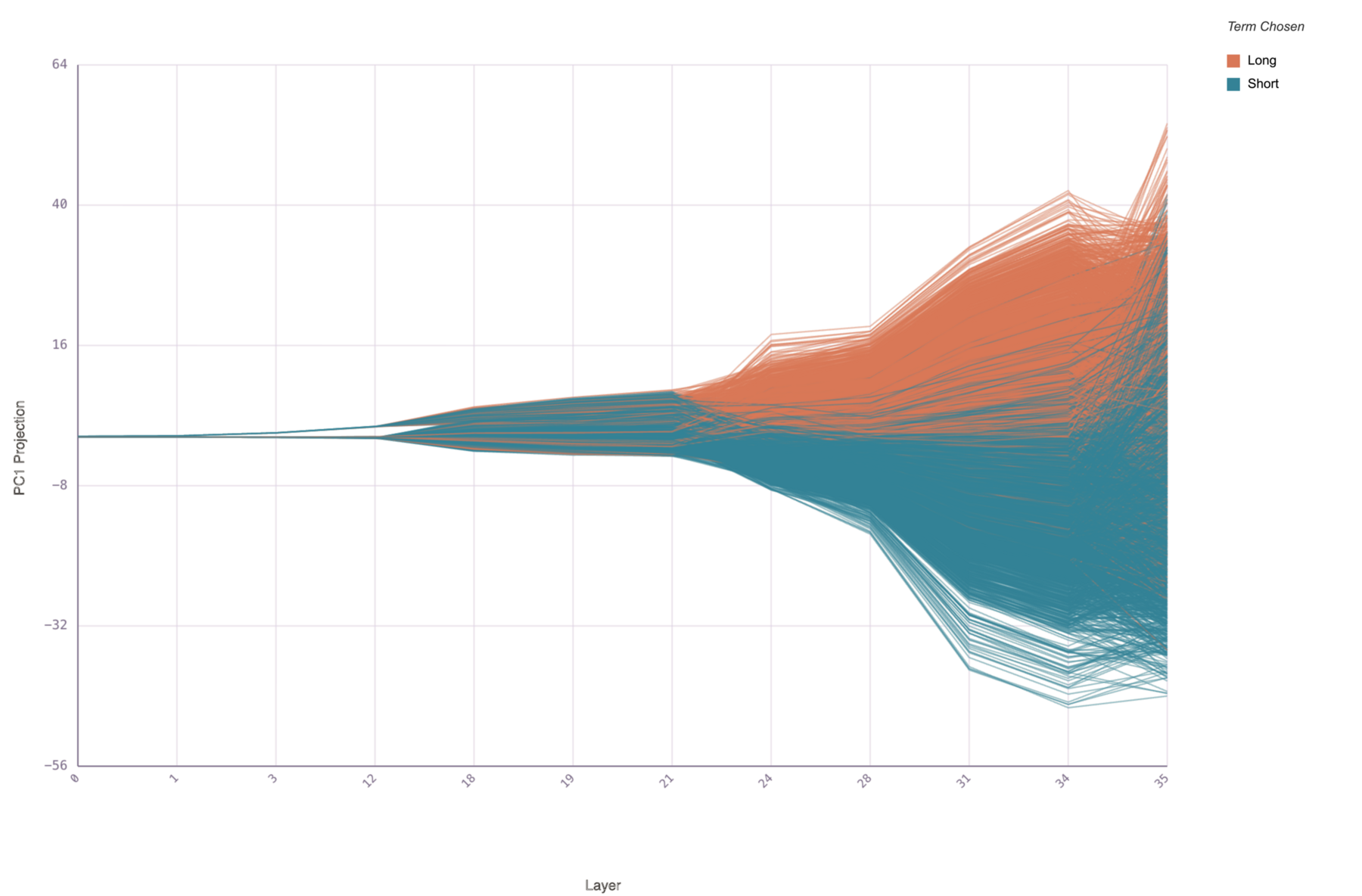}\\[2pt]
    {\footnotesize\texttt{\textbackslash n}}
  \end{minipage}\hfill
  \begin{minipage}[t]{0.245\linewidth}\centering
    \includegraphics[width=\linewidth]{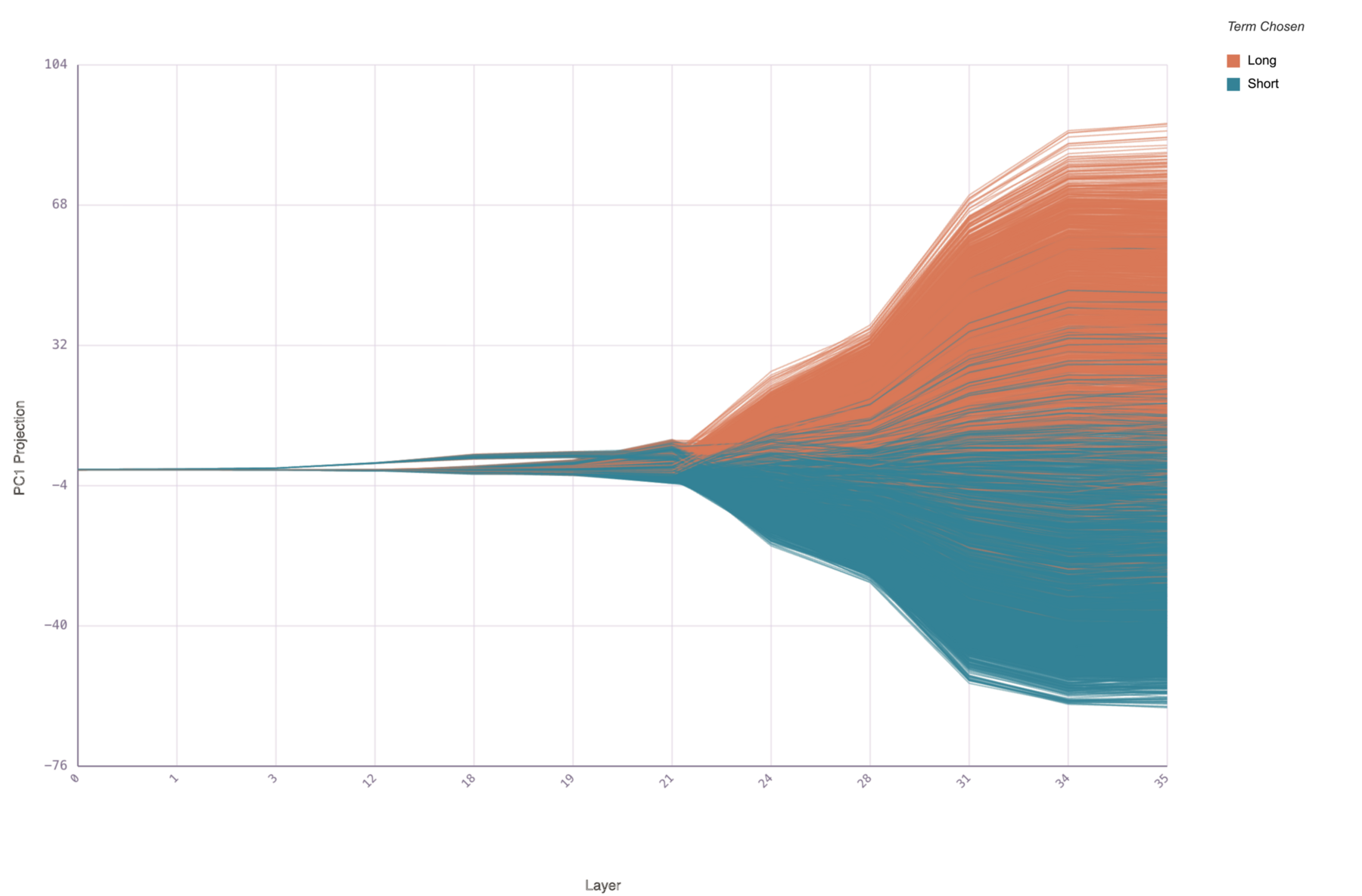}\\[2pt]
    {\footnotesize\texttt{<|im\_start|>}}
  \end{minipage}\hfill
  \begin{minipage}[t]{0.245\linewidth}\centering
    \includegraphics[width=\linewidth]{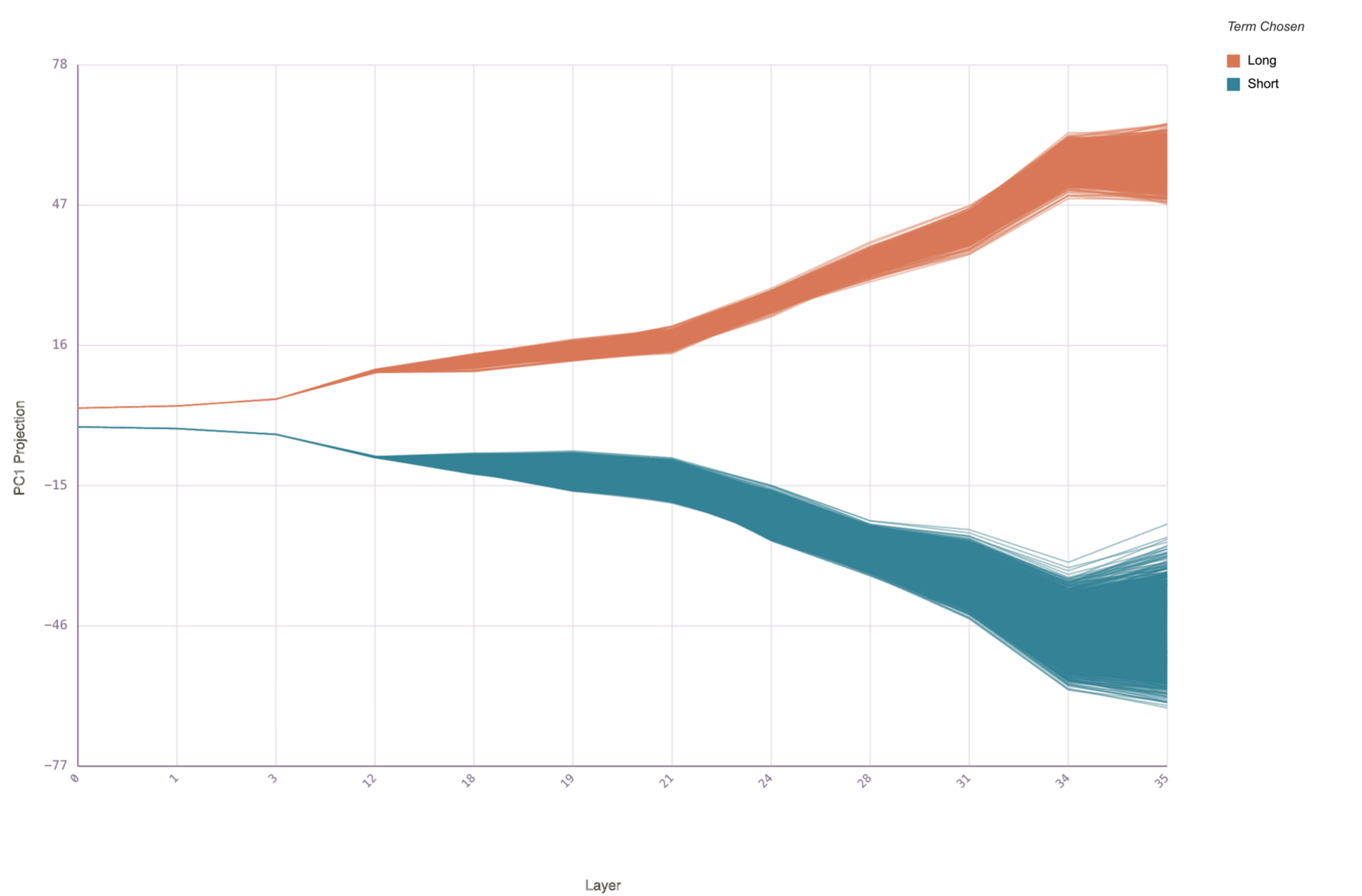}\\[2pt]
    {\footnotesize\texttt{assistant}}
  \end{minipage}
  \caption{\texttt{resid\_post} at the four turn-transition tokens (\texttt{<|im\_end|>}, \texttt{\textbackslash n}, \texttt{<|im\_start|>}, \texttt{assistant}), colored by preference (orange = long, blue = short).
  The preference signal sharpens from heavy overlap at the beginning of the turn change to clean separation by its end (\ref{app:parametric-geometry}).\label{fig:resid-post-turn-preference}}
\end{figure}

\textbf{Geometry.} Time horizons form ordinal clusters (seconds to centuries) in activation space, but the direction encoding them is unstable across prompt positions until the user-to-assistant turn boundary, where attention collapses the continuous horizon representation into a binary preference signal that sharpens from heavy overlap at the beginning of the turn change (\texttt{<|im\_end|>}) to clean separation by the end of the turn change (\texttt{assistant}) (Figures~\ref{fig:resid-post-turn-preference} and~\ref{fig:resid-post-turn}; \ref{app:parametric-geometry}).

\figtwocolFull[0.8]{%
    \multirow{4}{*}{\begin{tabular}{@{}c@{}}\includegraphics[width=\linewidth]{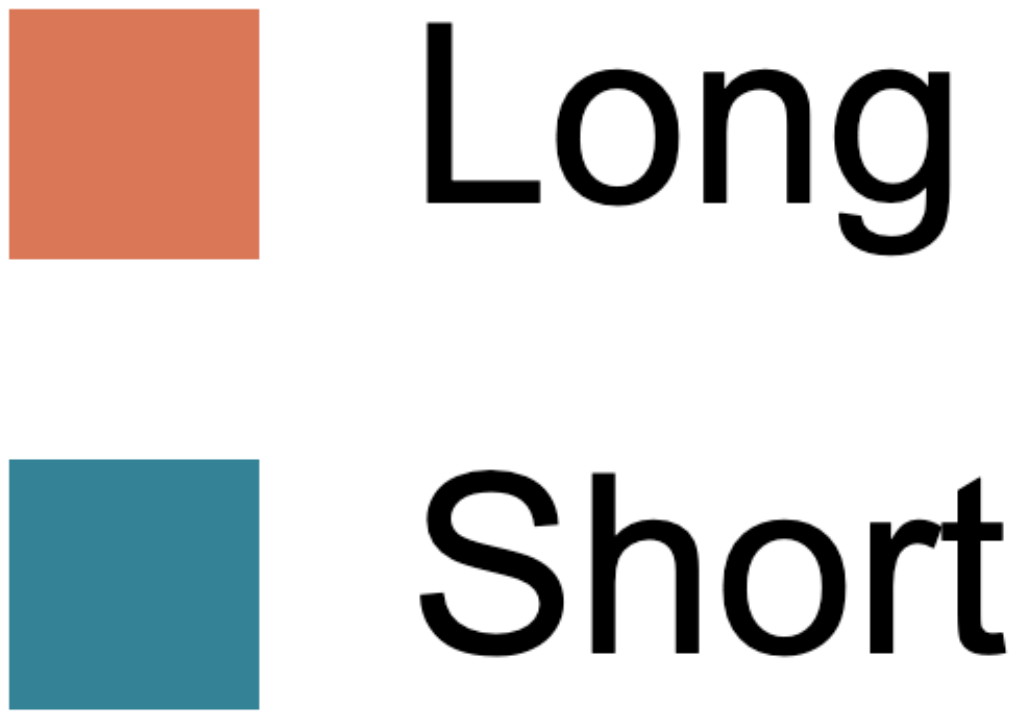}\\[6pt]\includegraphics[width=\linewidth]{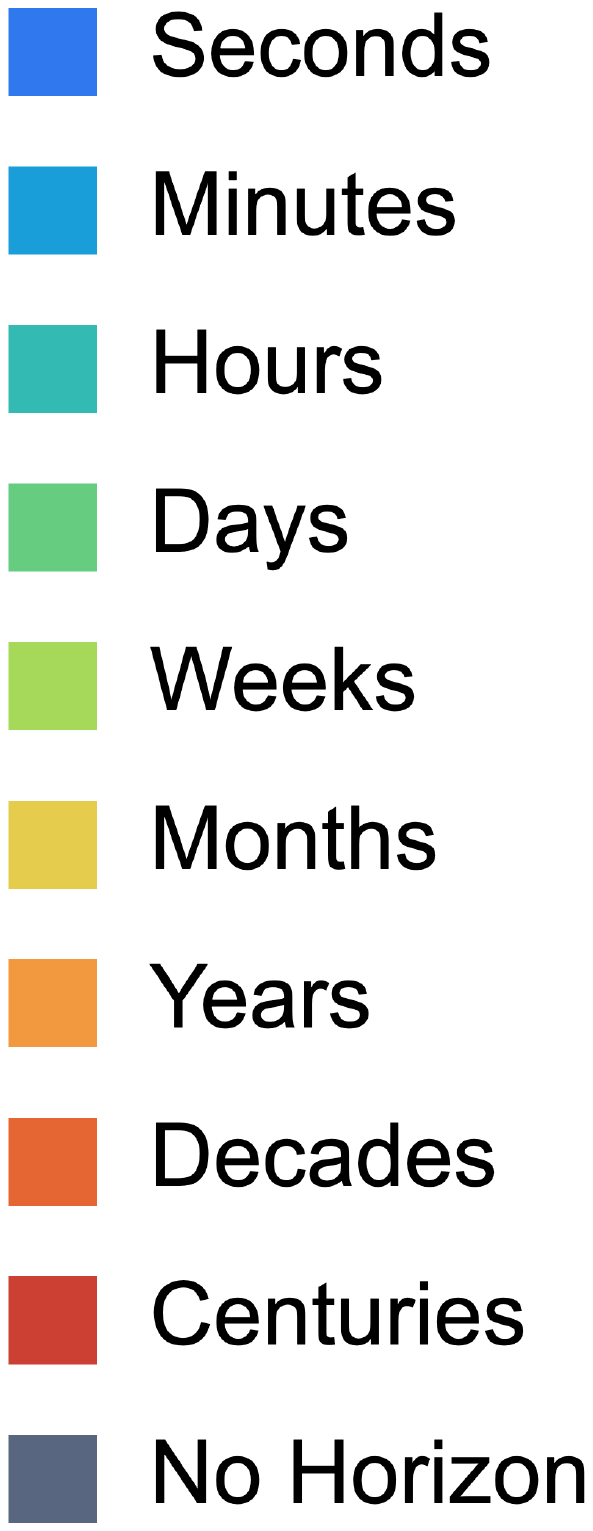}\end{tabular}}%
}{%
    \twocolrow{images/characterize/parametric_geometry/by_layer/L31__chat_suffix_0__term_chosen}
              {images/characterize/parametric_geometry/by_layer/L31__chat_suffix_0__time_scale}{\texttt{<|im\_end|>}}
    \twocolrow{images/characterize/parametric_geometry/by_layer/L31__chat_suffix_1__term_chosen}
              {images/characterize/parametric_geometry/by_layer/L31__chat_suffix_1__time_scale}{\texttt{\textbackslash n}}
    \twocolrow{images/characterize/parametric_geometry/by_layer/L31__chat_suffix_2__term_chosen}
              {images/characterize/parametric_geometry/by_layer/L31__chat_suffix_2__time_scale}{\texttt{<|im\_start|>}}
    \twocolrow{images/characterize/parametric_geometry/by_layer/L31__chat_suffix_3__term_chosen}
              {images/characterize/parametric_geometry/by_layer/L31__chat_suffix_3__time_scale}{\texttt{assistant}}
}{%
    Layer-31 PCA at the four turn-transition tokens (rows: \texttt{<|im\_end|>}, \texttt{\textbackslash n}, \texttt{<|im\_start|>}, \texttt{assistant}), colored by preference (left column: orange = long, blue = short) and by time horizon (right column: gradient from seconds to centuries; gray = no horizon).
    At \texttt{<|im\_end|>} the preference is not yet linearly separable and the no-horizon samples sit off the time-horizon manifold; by \texttt{assistant} the no-horizon samples have aligned to the manifold and preference clusters are cleanly separated, identifying the change-of-turn token sequence as the site where horizon is collapsed into a committed binary preference (\ref{app:parametric-geometry}).\label{fig:resid-post-turn}%
}
\FloatBarrier

\textbf{Discounting.} LLM discount rates ($k < 0.005$) are 3--8$\times$ below human controls ($k \approx 0.013$); chain-of-thought amplifies present bias in the 4B model but produces paradoxical patience in the 8B model (\ref{app:behavioral-temporal-discount}).

\begin{figure}[htbp]
  \centering
  \begin{minipage}[c]{0.58\textwidth}
    \centering
    \includegraphics[width=\linewidth]{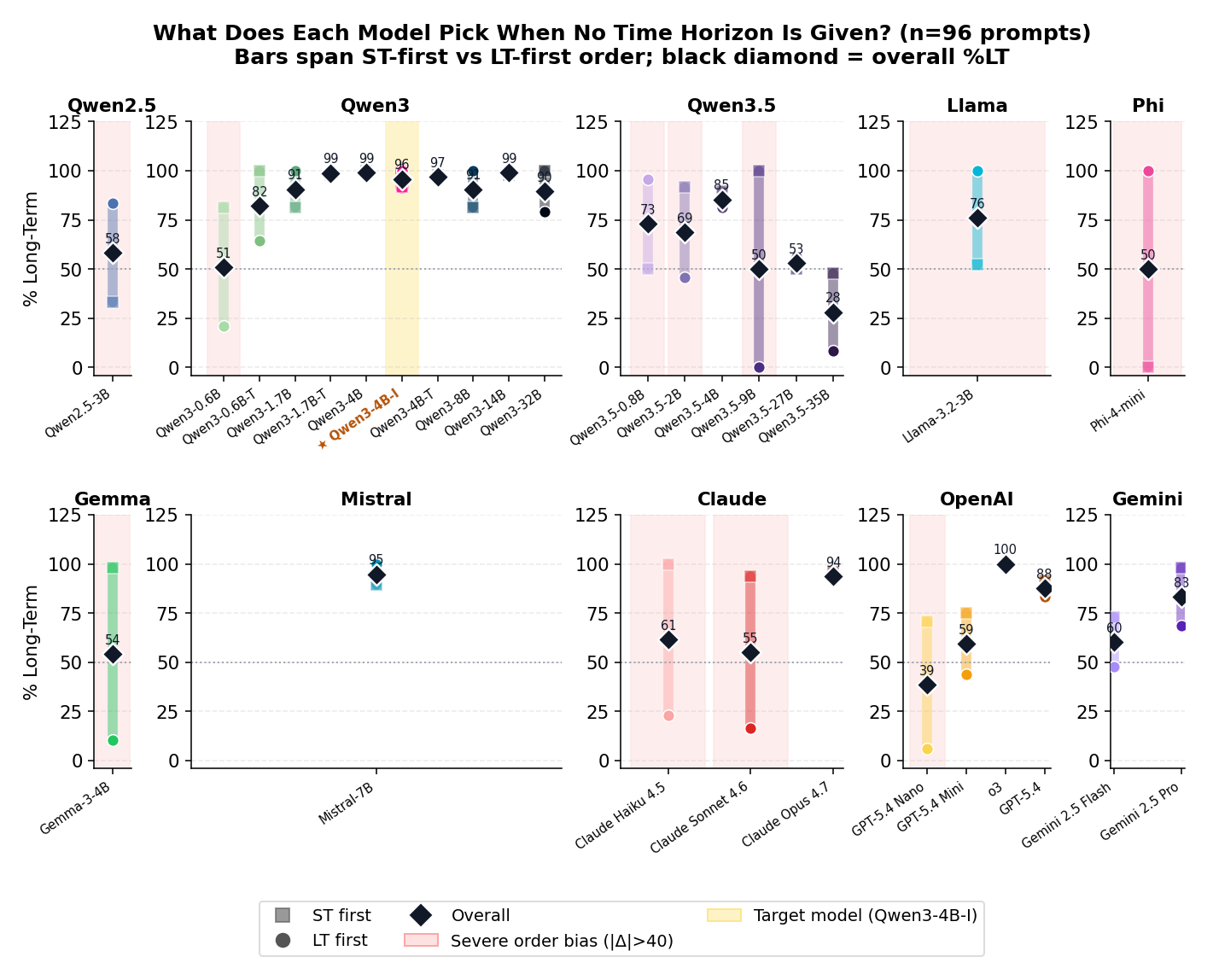}
  \end{minipage}\hfill
  \begin{minipage}[c]{0.40\textwidth}
    \centering
    \includegraphics[width=\linewidth]{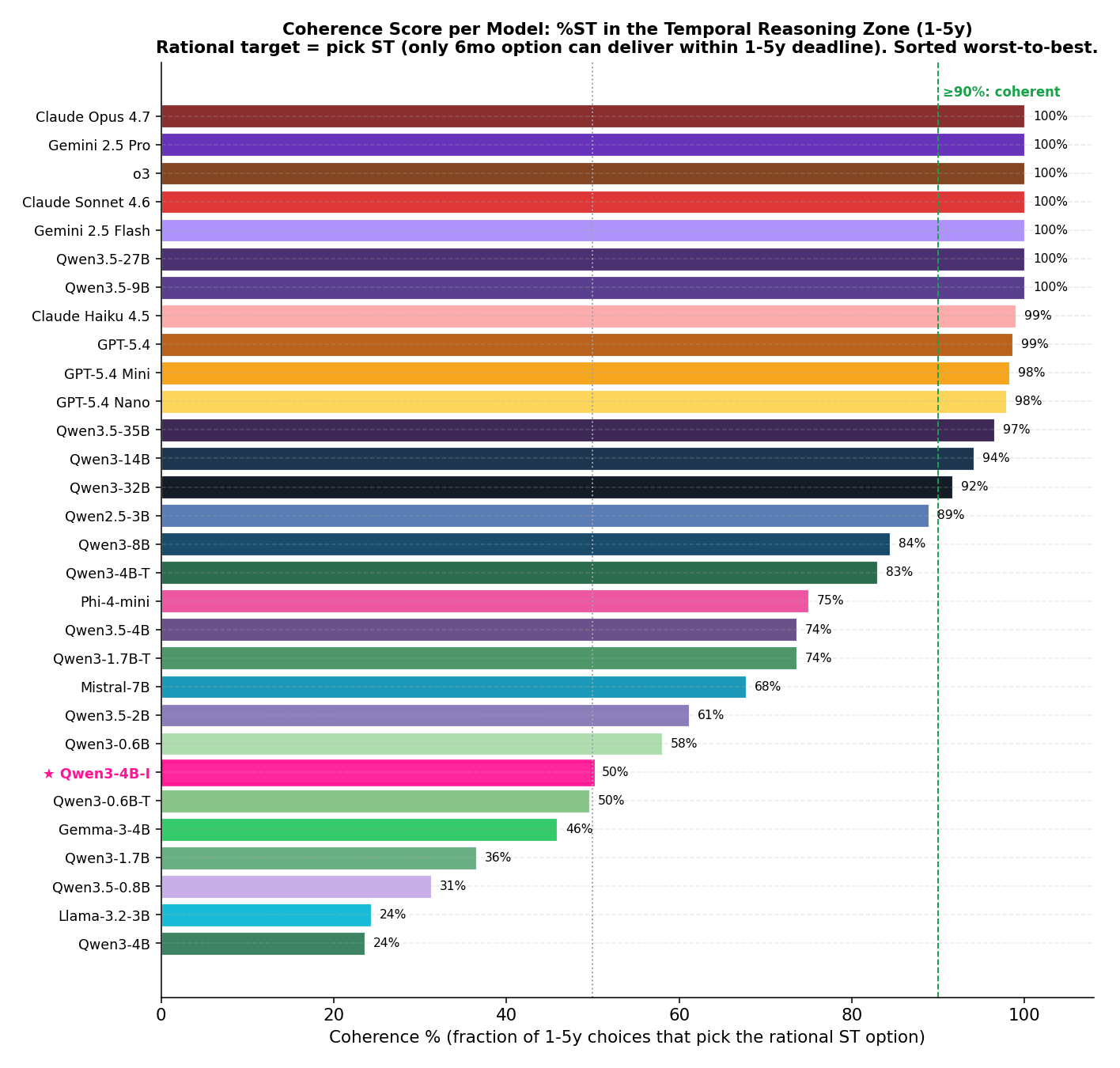}
  \end{minipage}
  \caption{\texttt{Qwen3-4B-Instruct-2507} (yellow) holds a stable temporal preference under presentation-order swaps in the no-horizon condition (left), but does not produce coherent temporal reasoning when given an explicit deadline (right; star marks our target at 50\%, far below the 90\% coherent threshold). Full per-model breakdowns in~\ref{app:behavioral-coherence}.\label{fig:stable-vs-coherent}}
\end{figure}

\textbf{Coherence.} We test whether \texttt{Qwen3-4B-Instruct-2507} makes instrumentally coherent choices (Section~\ref{sec:background}) on 960 investment prompts offering \$20K in 6 months vs.\ a long-term option (\$100K, \$300K, or \$500K in 10 years) (\ref{app:behavioral-coherence}).
Coherence is defined in the 1--5y reasoning zone, where only the 6-month option can deliver within the deadline.
Our model does not meet this bar: it picks the undeliverable long-term option 47--53\% of the time, and a deep dive shows this is positional polarization rather than uncertainty; reward size and label format are effectively inert (\ref{app:coherence-target-deep-dive}).
Benchmarking against 29 other models confirms the failure is not idiosyncratic: only frontier API models (\texttt{Claude} family, \texttt{Gemini 2.5 Pro}, \texttt{GPT-5.4}, \texttt{GPT-5.4 Mini}, \texttt{o3}) reach 95--100\% coherence, and the smaller \texttt{Claude} variants do so via a binary heuristic that collapses at longer horizons (Figure~\ref{fig:stable-vs-coherent}).

\FloatBarrier

\textbf{Generality.} Probing the same layers and turn-transition tokens for an unrelated meta-cognitive variable (the cumulative reliability of a multi-step reasoning chain) recovers a 95\% decodability plateau over L19 to L31.
The error-reliability direction is linearly orthogonal to the temporal-preference direction yet rides the same curved manifold within the localized subgraph (\ref{app:error-monitoring}).
The gap between rich internal representation and weak behavioral influence is therefore an architectural pattern of this subgraph, not specific to temporal preference.

\FloatBarrier
\subsection{Could temporal preference be controlled?}\label{sec:steering}

CAA steering at layers 19--22 shifts temporal preference bidirectionally: $\sim$$3.4\times$ higher relative odds for long- vs short-term completion at layer 22, $\alpha = 50$ ($\exp(1.22) \approx 3.39$; Figure~\ref{fig:main-pic}, bottom; Table~\ref{tab:main-extended-sweep}; \ref{app:contrastive-steering}).
Open-ended generation shifts from triage framing ($\alpha < 0$) to strategic framing ($\alpha > 0$).
Output quality degrades at $|\alpha| = 60$, suggesting the linear steering vector exceeds the locally linear regime of the curved manifold.
The optimal steering layers (19--22) sit 4--7 layers below the best probing layer (26), a probing--steering dissociation consistent with the localized subgraph.

\begin{table}[htbp]
  \centering
  \small
  \setlength{\tabcolsep}{4pt}
  \begin{tabular}{c|ccccccc}
  \toprule
  $\alpha$ & L19 & L20 & L21 & L22 & L23 & L24 & L25 \\
  \midrule
  20 & 0.72 & 0.69 & 0.72 & 0.68 & 0.56 & 0.51 & 0.46 \\
  30 & 0.94 & 0.91 & 0.97 & 0.93 & 0.76 & 0.68 & 0.60 \\
  40 & 1.13 & 1.10 & 1.20 & 1.17 & 0.96 & 0.84 & 0.74 \\
  50 & 1.30 & 1.27 & 1.39 & 1.39 & 1.14 & 1.01 & 0.87 \\
  \bottomrule
  \end{tabular}
  \caption{Forced-choice score $S(\alpha, l)$ for the layer $\times$ steering-coefficient sweep.
  Baseline (no steering): $S = 0.17$.
  Best configuration: layer~22 with $\alpha = 50$, $S = 1.39$ (lift $+1.22$, $\exp(1.22) \approx 3.39$ odds-ratio change).
  Layers 19--22 form the behavioral sweet spot; effectiveness drops sharply from L23 onward.
  Full sweep including $\alpha \in \{1, 2, 5, 10\}$ in~\ref{app:contrastive-steering}.\label{tab:main-extended-sweep}}
\end{table}
\section{Discussion}\label{sec:discussion}

Whether an LLM operates under the right temporal preference is ultimately an alignment question.
Post-training methods may suffice for routine use, but high-stakes settings call for stronger guarantees.
We believe activation geometry can serve as a fail-safe here: localize the subgraph relevant to a specific task, characterize the geometry of the temporal concept within it, and then, at inference time, monitor the model's internal representations against that manifold and intervene if they drift.
This perspective frames interpretability not only as a diagnostic tool but as infrastructure for runtime alignment.
\section{Limitations and future work}\label{sec:limitations}

Our work is a starting point on an entangled concept.
The main open directions are: finer-grained circuit tracing to move from subgraph-level attribution to atomic components and information flow; generalization beyond the single financial task and the single target model (\texttt{Qwen3-4B-Instruct-2507}) to other domains, model scales, and thinking vs.\ non-thinking variants; richer parameterization along reward, risk, role, and domain axes to map the full intertemporal choice space and its interactions with adjacent concepts such as emotion and urgency; multi-turn and agentic settings where temporal representations may shift across turns; and non-linear steering methods that respect the curvature of the underlying manifold and avoid the output-quality degradation we observe in linear CAA at high $|\alpha|$.
Full discussion is in~\ref{app:extended-limitations}.

\section{Conclusion}\label{sec:conclusion}
We show that temporal preference in LLMs is localizable, that we can characterize its representational geometry, and that targeted activation interventions can shift it bidirectionally.
Our work highlights the value of using complementary paradigms.
More broadly, while the literature has focused on identifying contrastive binary concepts, this work offers initial steps toward decomposing dimensional concepts such as time.

\makeatletter
\if@anonymous\else
  \begin{ack}
    We thank the \textbf{AI Safety Camp (AISC)} and the \textbf{Supervised Program for Alignment Research (SPAR)} for providing the collaborative structure, mentorship, and computational resources that made this project possible.
    AISC's cohort-based research model brought the authors together and sustained the multi-month investigation; SPAR's program provided additional mentorship and connected contributors across timezones.
  \end{ack}
\fi
\makeatother

\bibliographystyle{plainnat}
\bibliography{references}


\clearpage
\appendix


\renewcommand{\thesection}{Appendix \AlphAlph{\value{section}}}
\renewcommand{\thesubsection}{\AlphAlph{\value{section}}.\arabic{subsection}}
\renewcommand{\thesubsubsection}{\AlphAlph{\value{section}}.\arabic{subsection}.\arabic{subsubsection}}
\renewcommand{\theequation}{\AlphAlph{\value{section}}.\arabic{equation}}%
\renewcommand{\thefigure}{\AlphAlph{\value{section}}.\arabic{figure}}%
\renewcommand{\thetable}{\AlphAlph{\value{section}}.\arabic{table}}%
\newcommand{\clearappnumbering}{
    \setcounter{equation}{0}%
    \setcounter{figure}{0}%
    \setcounter{table}{0}%
}



\clearpage
\thispagestyle{empty}
\vspace*{0.4\textheight}
\begin{center}{\Huge\bfseries Appendices}\end{center}

\clearpage


\section*{Index of Appendices}
\label{sec:appendix-index}

\begin{center}
\small
\renewcommand{\arraystretch}{1.2}
\setlength{\tabcolsep}{5pt}
\rowcolors{2}{black!3}{white}
\begin{tabularx}{\linewidth}{@{} c r l X @{}}
  \toprule
  \rowcolor{white}
  \textbf{App.} & \textbf{p.} & \textbf{Paradigm} & \textbf{Content} \\
  \midrule
  \multicolumn{4}{@{}l}{\textbf{Part 0: Groundwork}} \\
  \hyperref[app:extended-background]{A}       & \pageref{app:extended-background}    & n/a           & Extended background: definitions and interpretability primitives. \\
  \hyperref[app:extended-literature]{B}       & \pageref{app:extended-literature}    & n/a           & Extended literature. \\
  \hyperref[app:methodology-summary]{C}       & \pageref{app:methodology-summary}    & n/a           & Methodology summary and overview of the extended methodologies. \\
  \hyperref[app:experimental-details]{D}      & \pageref{app:experimental-details}   & n/a           & Full experimental details. \\
  \hyperref[app:prompts]{E}                   & \pageref{app:prompts}                & n/a           & Prompting settings and dataset construction. \\
  \hyperref[app:extended-limitations]{F}      & \pageref{app:extended-limitations}   & n/a           & Extended limitations and future work. \\
  \midrule
  \multicolumn{4}{@{}l}{\textbf{Part 1: Localize} (ordered by method strength: probing $\to$ attributional $\to$ causal)} \\
  \hyperref[app:contrastive-probing-linear]{G}   & \pageref{app:contrastive-probing-linear} & Probing         & Linear probing: 99.2\% at L26, cross-dataset generalization. \\
  \hyperref[app:attributional-contrastive]{H}    & \pageref{app:attributional-contrastive}  & Attr.\ contr.   & EAP-IG attribution on contrastive prompts. \\
  \hyperref[app:attributional-parametric]{I}    & \pageref{app:attributional-parametric}  & Attr.\ param.   & EAP-IG attribution on parametric prompts. \\
  \hyperref[app:causal-parametric]{J}            & \pageref{app:causal-parametric}          & Causal param.   & Activation patching: L21--24 attn, L31--35 MLP. \\
  \hyperref[app:causal-contrastive]{K}           & \pageref{app:causal-contrastive}         & Causal class.   & Directional patching: asymmetric effects, two-phase classification, L24 attn. \\
  \hyperref[app:convergence]{L}                  & \pageref{app:convergence}                & All             & Cross-method convergence on layers 17--35. \\
  \midrule
  \multicolumn{4}{@{}l}{\textbf{Part 2: Characterize}} \\
  \hyperref[app:parametric-geometry]{M}          & \pageref{app:parametric-geometry}        & Parametric      & PCA geometry: horizon$\to$preference transformation at turn boundary. \\
  \hyperref[app:latent-vs-constrained]{N}        & \pageref{app:latent-vs-constrained}      & All             & Latent vs.\ constrained: sparse attn vs.\ full subgraph. \\
  \hyperref[app:behavioral-temporal-discount]{O} & \pageref{app:behavioral-temporal-discount} & Behavioral    & Temporal discounting: LLMs 3--8$\times$ more patient than humans. \\
  \hyperref[app:behavioral-coherence]{P}         & \pageref{app:behavioral-coherence}       & Behavioral      & Behavioral coherence: order bias, instruct degradation. \\
  \hyperref[app:cross-model-comparison]{Q}       & \pageref{app:cross-model-comparison}     & Causal param.   & Cross-model patching: circuit localizes at fractional depth 0.6--0.7 across scale. \\
  \hyperref[app:error-monitoring]{R}             & \pageref{app:error-monitoring}           & Probing         & Error monitoring: subgraph encodes chain reliability orthogonally to time horizon. \\
  \midrule
  \multicolumn{4}{@{}l}{\textbf{Part 3: Intervene}} \\
  \hyperref[app:contrastive-steering]{S}         & \pageref{app:contrastive-steering}       & Contrastive     & CAA steering: bidirectional, L19--22, probing--steering dissociation. \\
  \midrule
  \multicolumn{4}{@{}l}{\textbf{Part 4: Extended methodologies} (same strength ordering as Part 1)} \\
  \hyperref[app:notation]{T}                             & \pageref{app:notation}                           & n/a             & Notation and key concepts. \\
  \hyperref[app:contrastive-probing-linear-methods]{U}   & \pageref{app:contrastive-probing-linear-methods} & Probing         & Probing protocol and activation extraction. \\
  \hyperref[app:attributional-contrastive-methods]{V}    & \pageref{app:attributional-contrastive-methods}  & Attr.\ contr.   & EAP-IG methodology, bias controls, component taxonomy. \\
  \hyperref[app:causal-parametric-methods]{W}            & \pageref{app:causal-parametric-methods}          & Causal param.   & Activation patching setup, noise/denoise protocol. \\
  \hyperref[app:causal-contrastive-methods]{X}           & \pageref{app:causal-contrastive-methods}         & Causal class.   & Directional patching on contrastive classification prompts. \\
  \hyperref[app:parametric-geometry-methods]{Y}          & \pageref{app:parametric-geometry-methods}        & Parametric      & PCA geometry analysis pipeline. \\
  \hyperref[app:behavioral-temporal-discount-methods]{Z} & \pageref{app:behavioral-temporal-discount-methods} & Behavioral    & Kirby MCQ-27 instrument, decision boundary method. \\
  \hyperref[app:behavioral-coherence-methods]{AA}         & \pageref{app:behavioral-coherence-methods}       & Behavioral      & Behavioral coherence experiment design. \\
  \hyperref[app:contrastive-steering-methods]{AB}        & \pageref{app:contrastive-steering-methods}       & Contrastive     & CAA vector construction and steering setup. \\
  \hyperref[app:case-study-hf]{AC}                       & \pageref{app:case-study-hf}                      & n/a             & Worked case study: highly-formatted pair. \\
  \bottomrule
\end{tabularx}
\end{center}

\clearpage

\clearpage
\clearappnumbering

\phantomsection
\label{sec:appendix-guide}
\thispagestyle{empty}

\vspace*{\fill}

\begin{center}
\small
\begin{tikzpicture}[
    every node/.style={font=\small\sffamily},
    partbox/.style={
        draw, rounded corners=5pt, thick,
        inner sep=7pt, align=left,
        minimum width=0.82\linewidth,
    },
    pre/.style={partbox,   fill=black!3,   draw=black!30},
    p1/.style={partbox,    fill=blue!5,    draw=blue!40},
    p2/.style={partbox,    fill=teal!5,    draw=teal!40},
    p3/.style={partbox,    fill=orange!6,  draw=orange!45},
    p4/.style={partbox,    fill=violet!5,  draw=violet!40},
    arrow/.style={-{Latex[length=2.5mm]}, very thick, draw=black!45},
]
\node[pre] (p0) {%
  \begin{tabular}{@{}l@{}}
  {\bfseries Part 0: Groundwork}\\[3pt]
  \textbf{A.}~Ext.\ background \quad
  \textbf{B.}~Ext.\ literature \quad
  \textbf{C.}~Method.\ summary\\
  \textbf{D.}~Experimental details \quad
  \textbf{E.}~Prompts \quad
  \textbf{F.}~Ext.\ limitations
  \end{tabular}%
};

\node[p1, below=5mm of p0] (p1) {%
  \begin{tabular}{@{}l@{}}
  {\bfseries Part 1: Localize}\\[3pt]
  \textbf{G.}~Probing \quad
  \textbf{H.}~Attr.\ contr.\ \quad
  \textbf{I.}~Attr.\ param.\\
  \textbf{J.}~Causal param.\ \quad
  \textbf{K.}~Causal class.\ \quad
  \textbf{L.}~Convergence
  \end{tabular}%
};

\node[p2, below=5mm of p1] (p2) {%
  \begin{tabular}{@{}l@{}}
  {\bfseries Part 2: Characterize}\\[3pt]
  \textbf{M.}~Geometry \quad
  \textbf{N.}~Latent/constr.\ \quad
  \textbf{O.}~Discounting\\
  \textbf{P.}~Coherence \quad
  \textbf{Q.}~Cross-model comp.\ \quad
  \textbf{R.}~Error monitoring
  \end{tabular}%
};

\node[p3, below=5mm of p2] (p3) {%
  \begin{tabular}{@{}l@{}}
  {\bfseries Part 3: Intervene}\\[3pt]
  \textbf{S.}~Steering
  \end{tabular}%
};

\node[p4, below=5mm of p3] (p4) {%
  \begin{tabular}{@{}l@{}}
  {\bfseries Part 4: Extended methodologies}\\[3pt]
  \textbf{T.}~Notation \quad
  \textbf{U.}~Probing meth.\ \quad
  \textbf{V.}~Attr.\ contr.\ meth.\\
  \textbf{W.}~Causal param.\ meth.\ \quad
  \textbf{X.}~Causal class.\ meth.\ \quad
  \textbf{Y.}~Geometry meth.\\
  \textbf{Z.}~Discounting meth.\ \quad
  \textbf{AA.}~Coherence meth.\ \quad
  \textbf{AB.}~Steering meth.\\
  \textbf{AC.}~Case study
  \end{tabular}%
};

\draw[arrow] (p0.south) -- (p1.north);
\draw[arrow] (p1.south) -- (p2.north);
\draw[arrow] (p2.south) -- (p3.north);
\draw[arrow] (p3.south) -- (p4.north);

\end{tikzpicture}
\end{center}

\vspace*{\fill}

\clearpage
\clearappnumbering


\section*{A visual tour of the appendices}

\vspace{1em}

\newcommand{\galimg}[2]{%
  \begin{center}
    \includegraphics[width=\linewidth,height=0.26\textheight,keepaspectratio]{#1}\\[0.35em]
    {\footnotesize\itshape\hyperref[#2]{\ref*{#2}}, p.\ \pageref{#2}}
  \end{center}%
}

\noindent
\begin{minipage}[t]{0.58\textwidth}
  \galimg{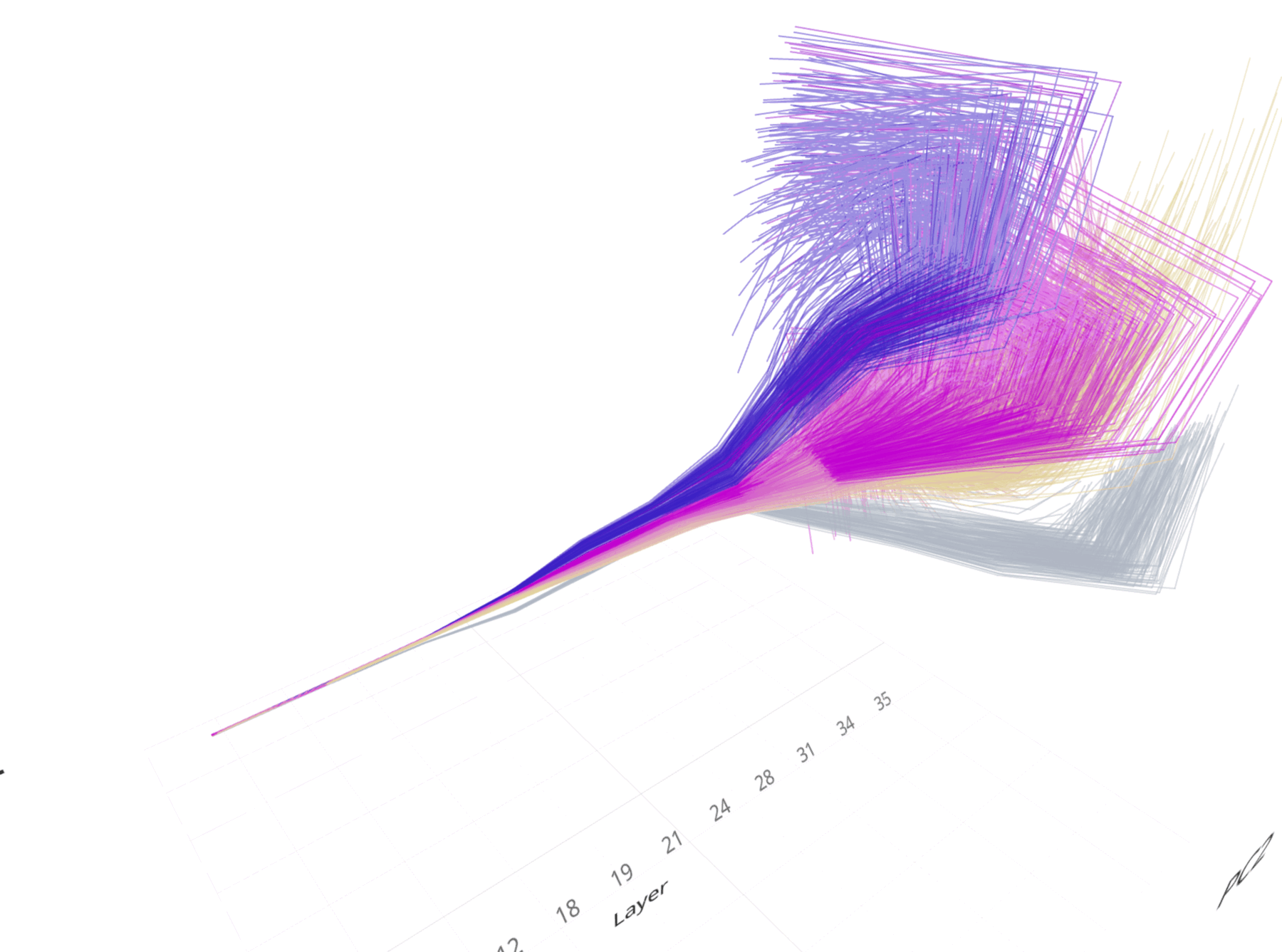}%
         {app:parametric-geometry}
\end{minipage}\hfill
\begin{minipage}[t]{0.38\textwidth}
  \galimg{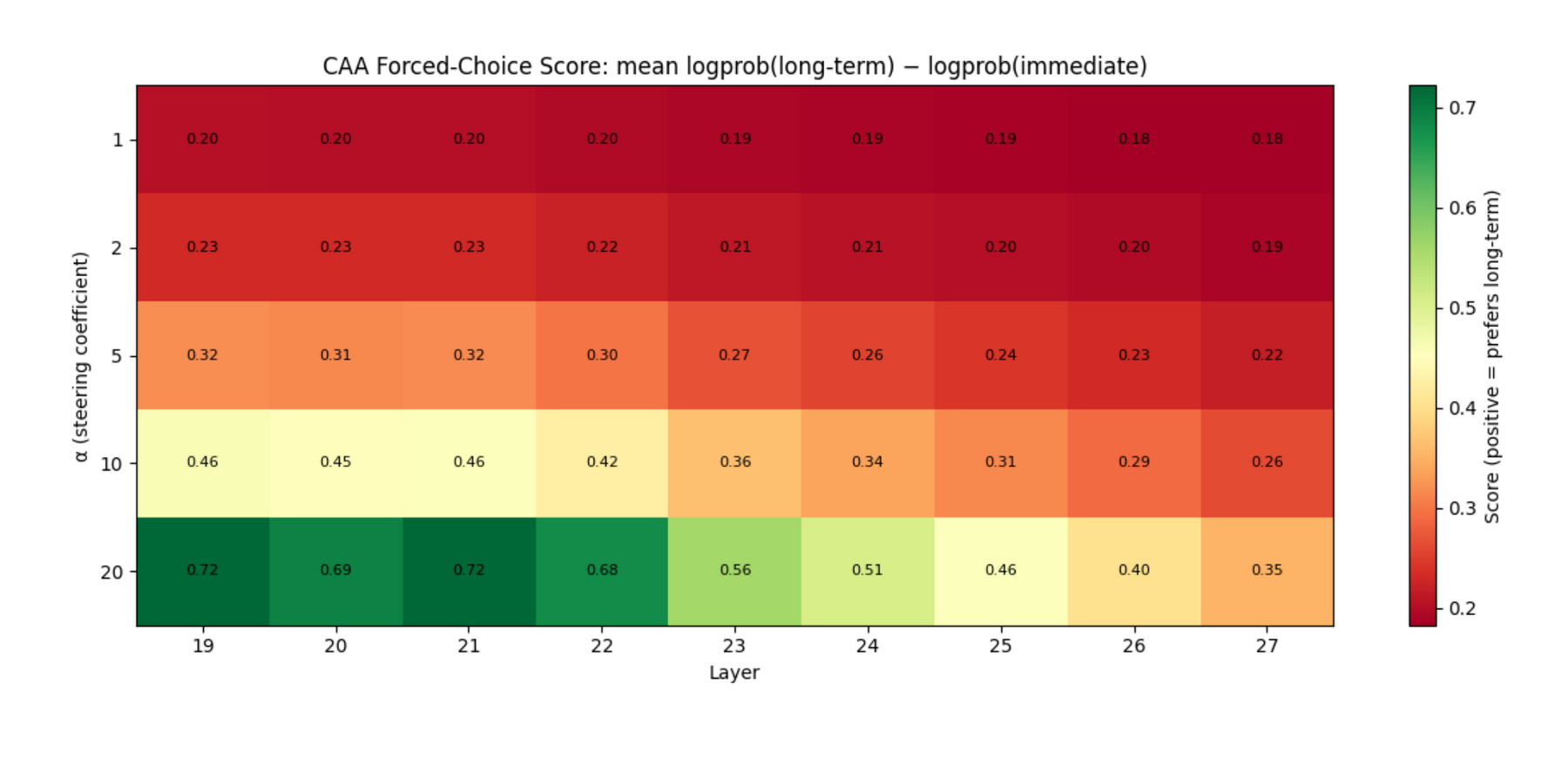}%
         {app:contrastive-steering}
\end{minipage}

\vspace{1em}

\noindent
\begin{minipage}[t]{0.32\textwidth}
  \galimg{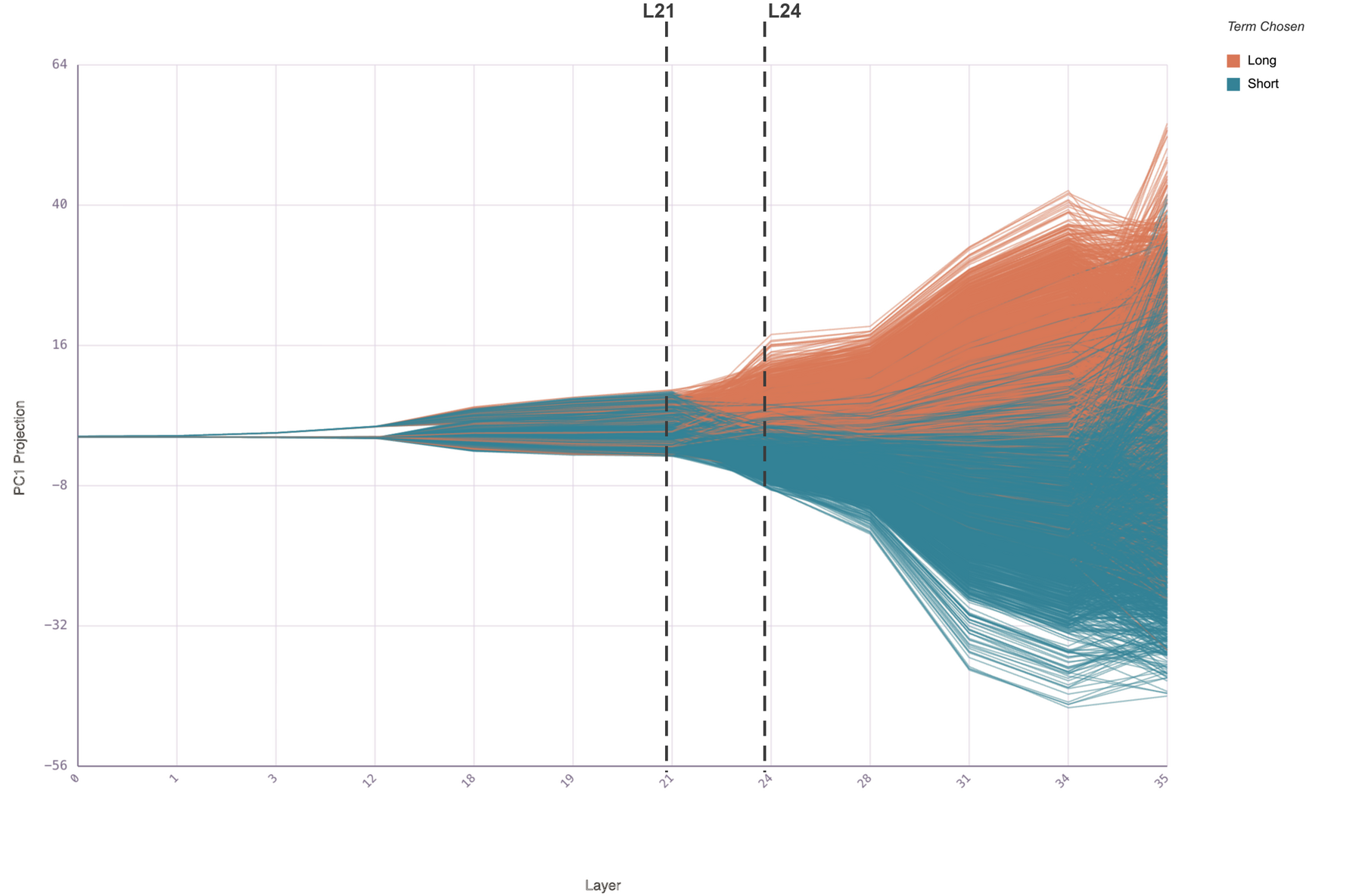}%
         {app:parametric-geometry}
\end{minipage}\hfill
\begin{minipage}[t]{0.32\textwidth}
  \galimg{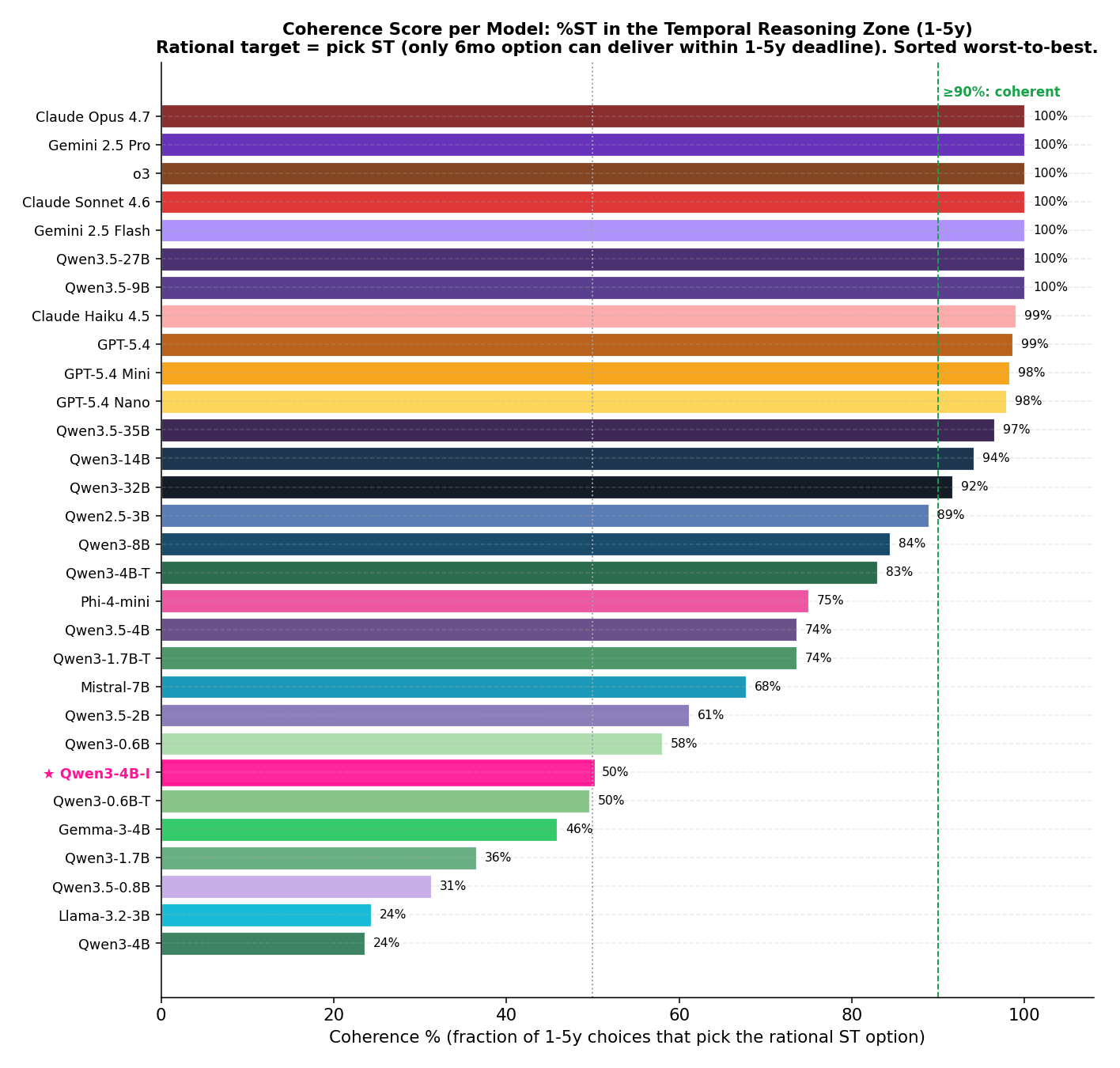}%
         {app:behavioral-coherence}
\end{minipage}\hfill
\begin{minipage}[t]{0.32\textwidth}
  \galimg{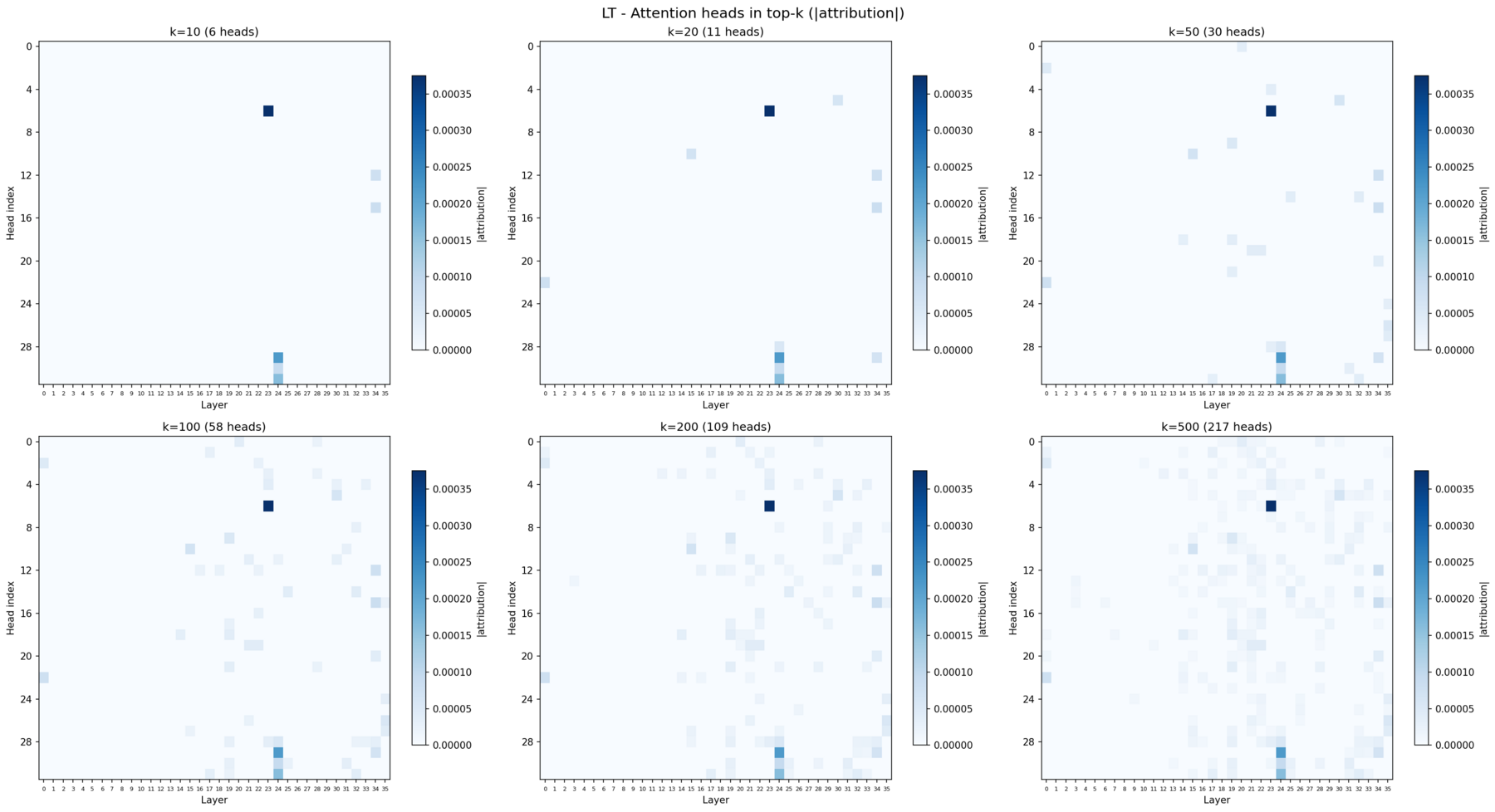}%
         {app:attributional-contrastive}
\end{minipage}

\vspace{1em}

\noindent
\begin{minipage}[t]{0.48\textwidth}
  \galimg{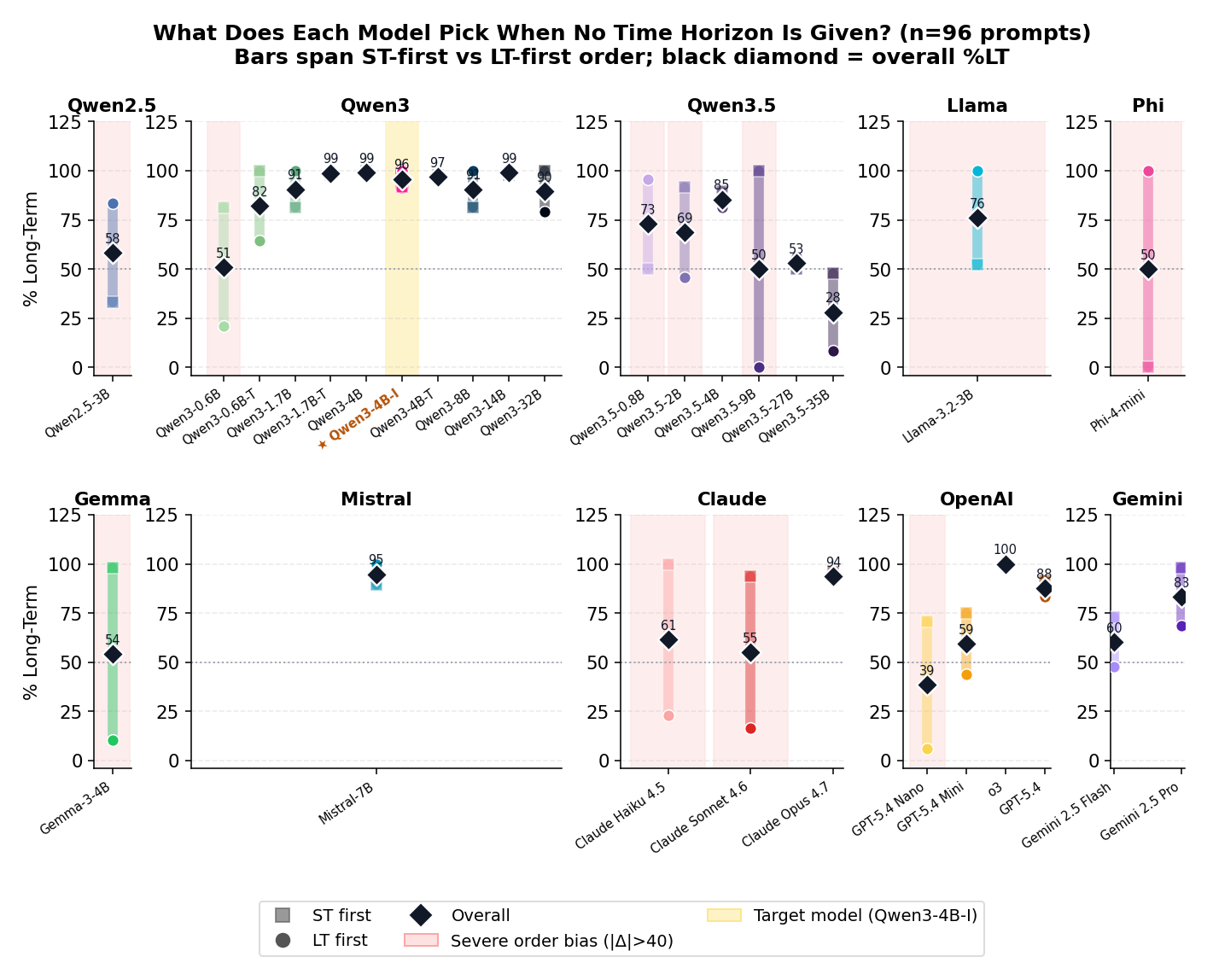}%
         {app:behavioral-coherence}
\end{minipage}\hfill
\begin{minipage}[t]{0.48\textwidth}
  \galimg{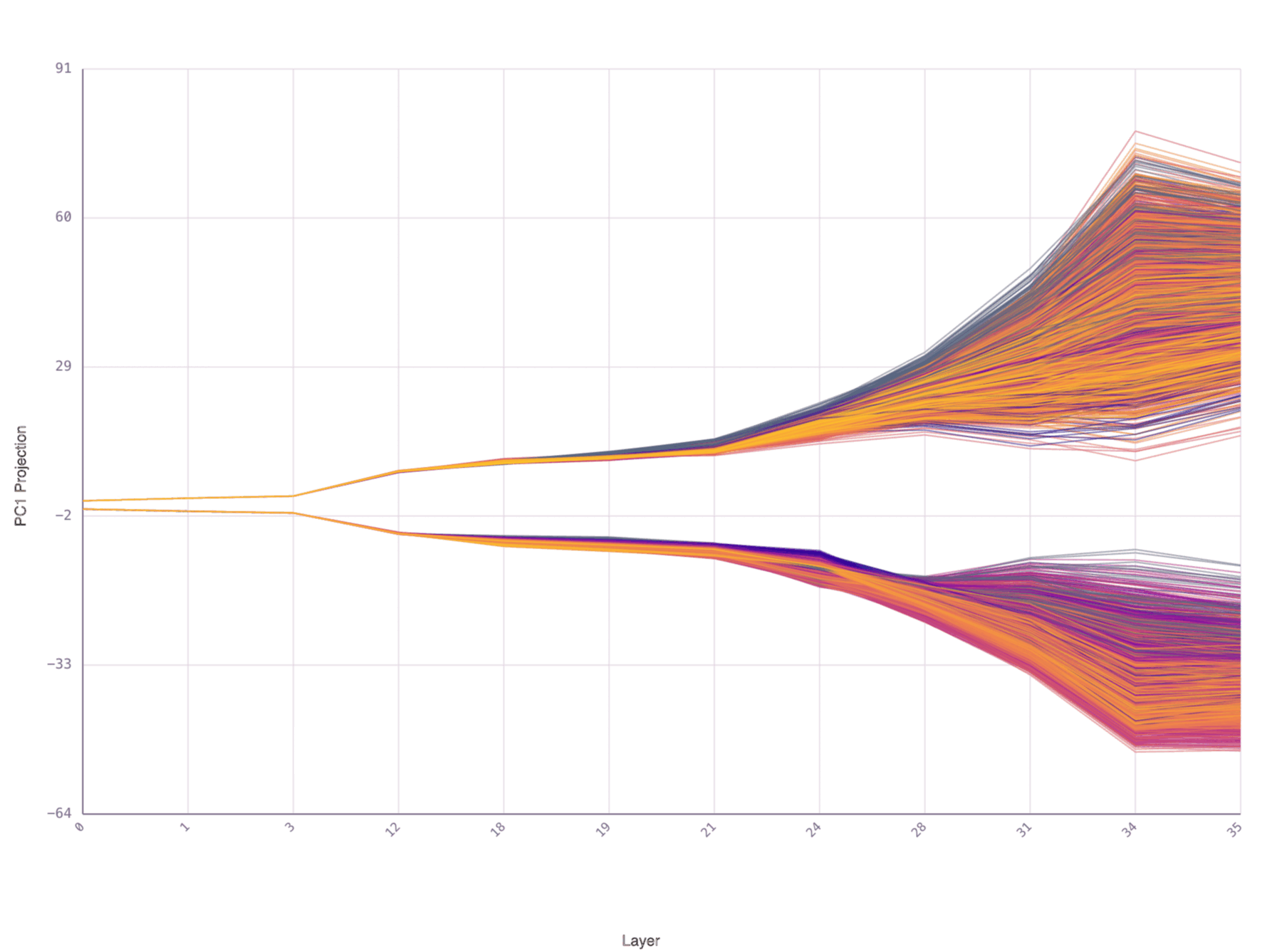}%
         {app:parametric-geometry}
\end{minipage}

\vspace{1em}

\noindent
\begin{minipage}[t]{0.32\textwidth}
  \galimg{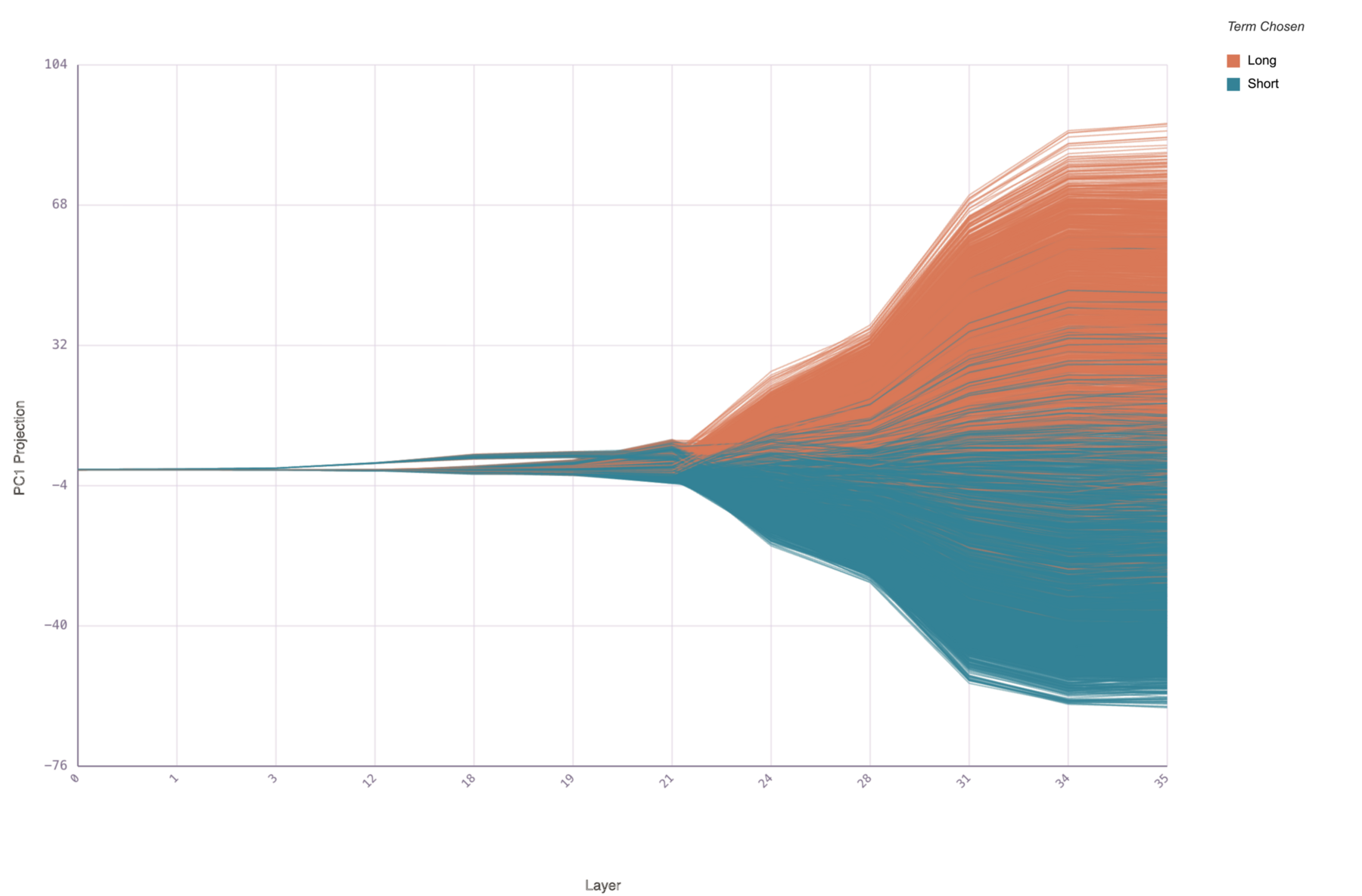}%
         {app:parametric-geometry}
\end{minipage}\hfill
\begin{minipage}[t]{0.32\textwidth}
  \galimg{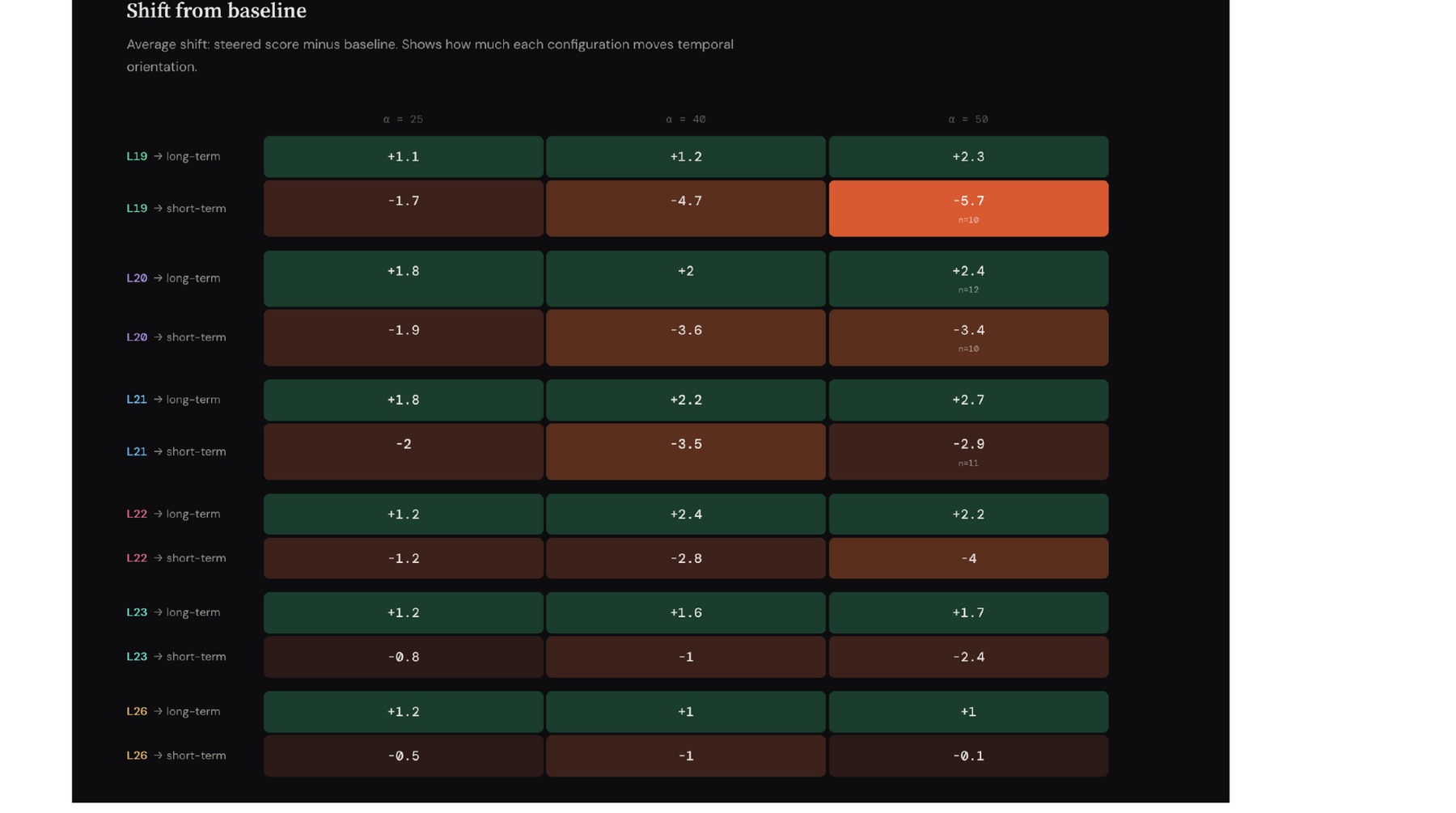}%
         {app:contrastive-steering}
\end{minipage}\hfill
\begin{minipage}[t]{0.32\textwidth}
  \galimg{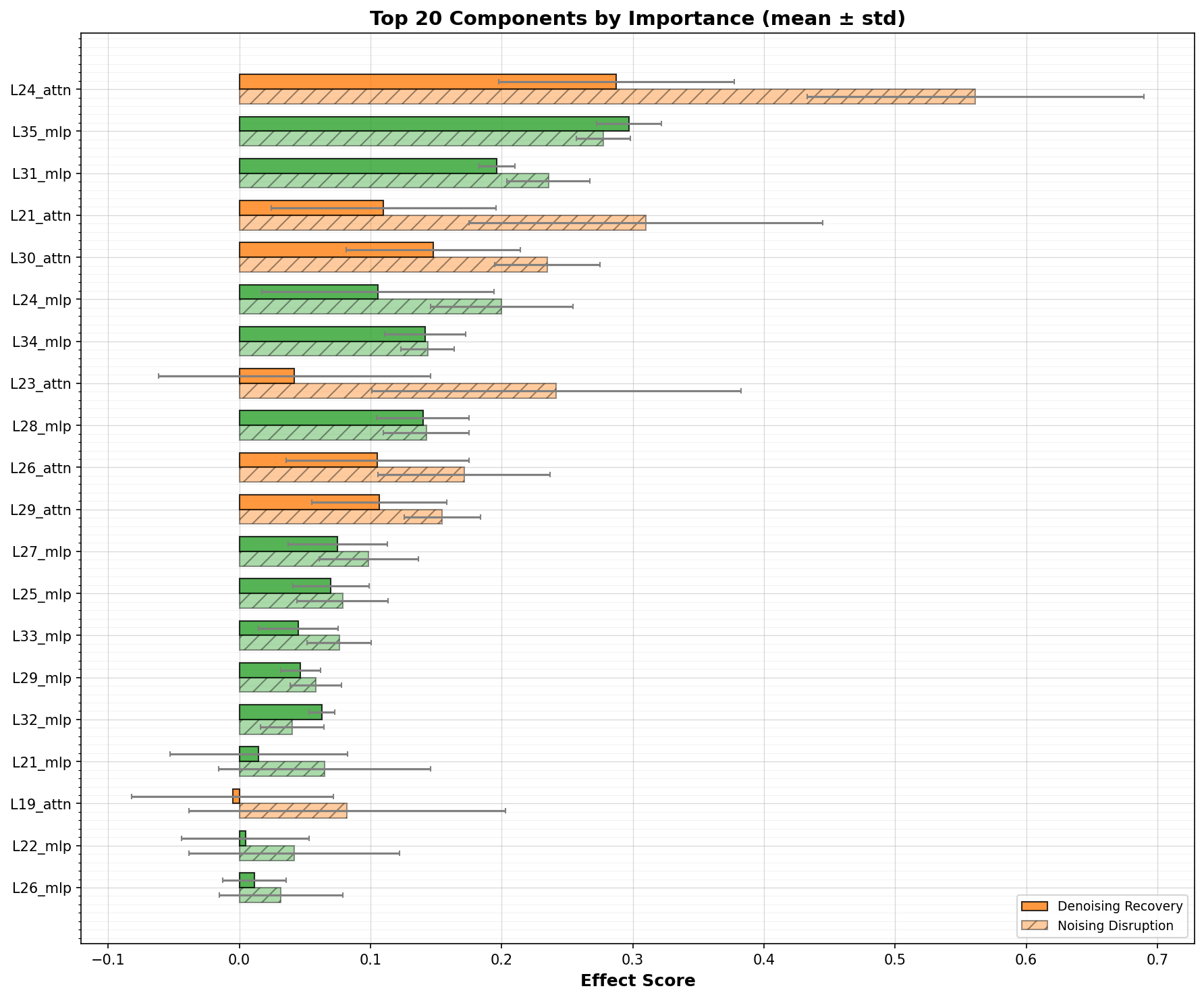}%
         {app:causal-parametric}
\end{minipage}

\vspace{1em}

\noindent
\begin{minipage}[t]{0.24\textwidth}
  \galimg{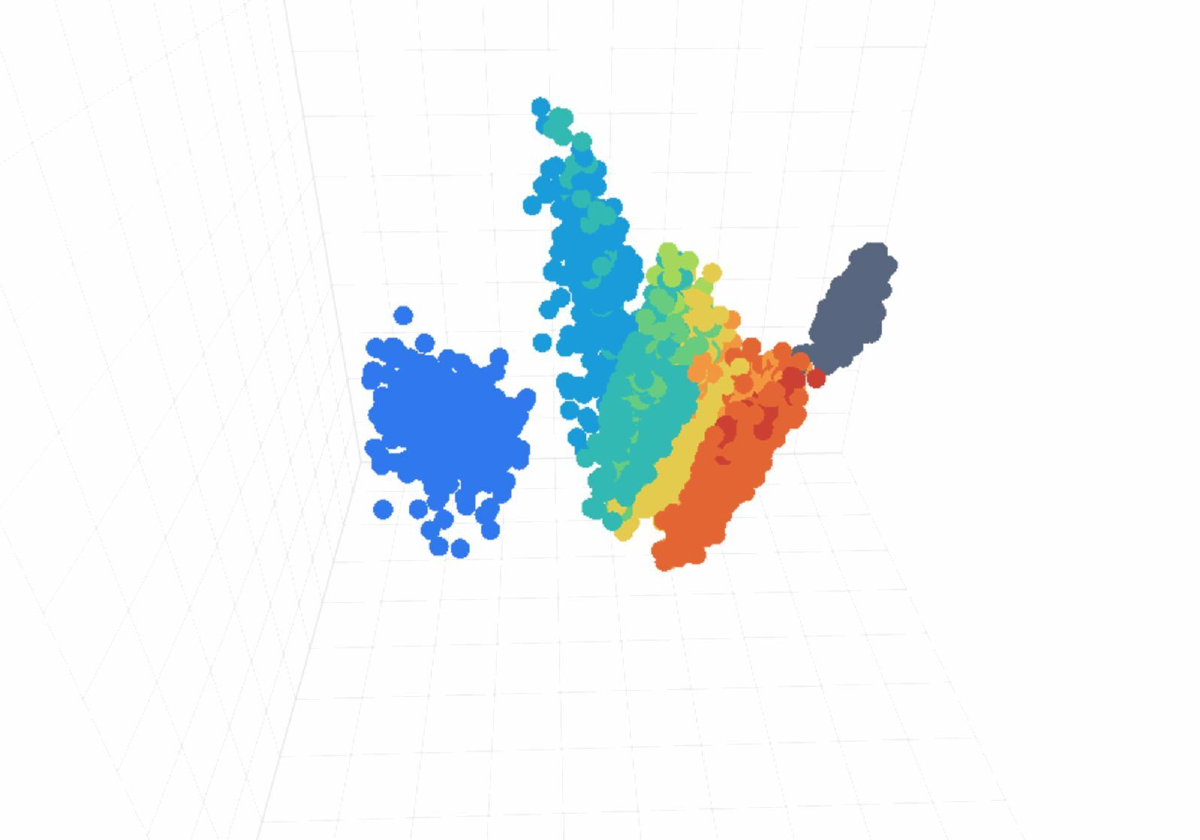}%
         {app:parametric-geo-bot}
\end{minipage}\hfill
\begin{minipage}[t]{0.24\textwidth}
  \galimg{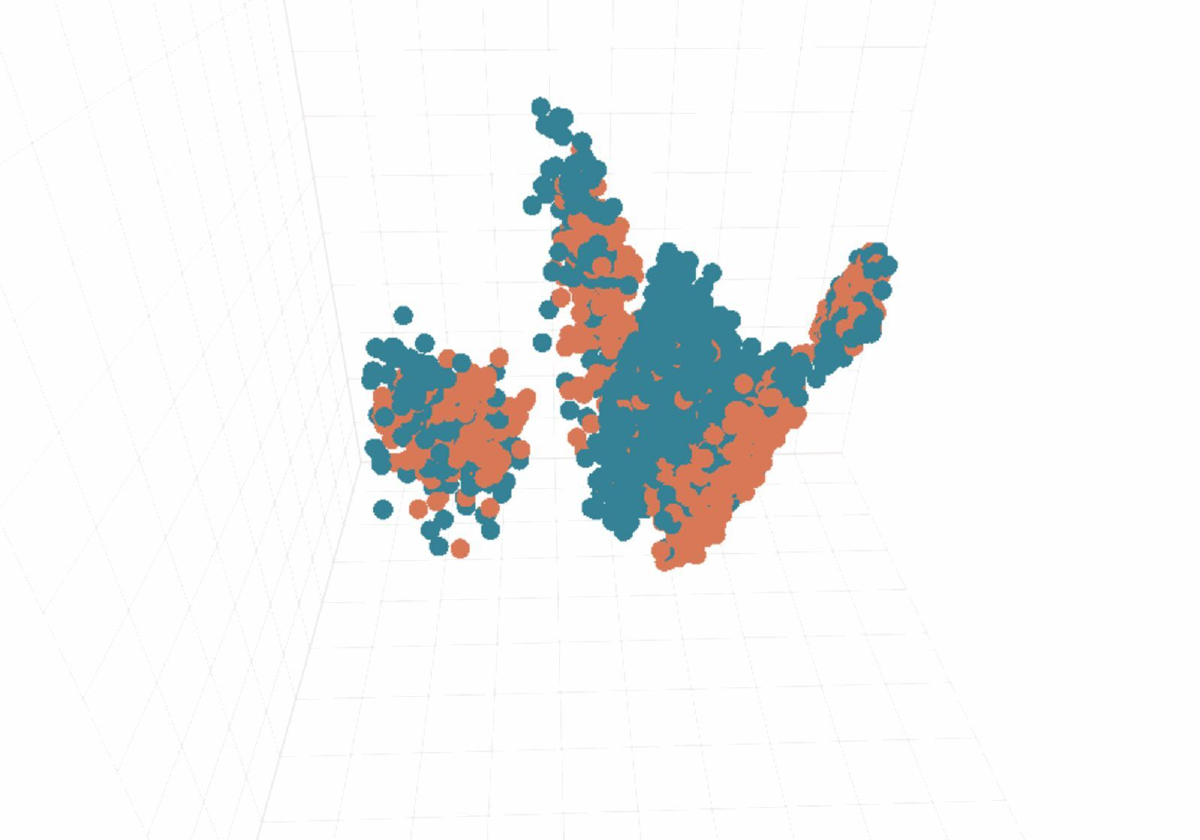}%
         {app:parametric-geo-bot}
\end{minipage}\hfill
\begin{minipage}[t]{0.24\textwidth}
  \galimg{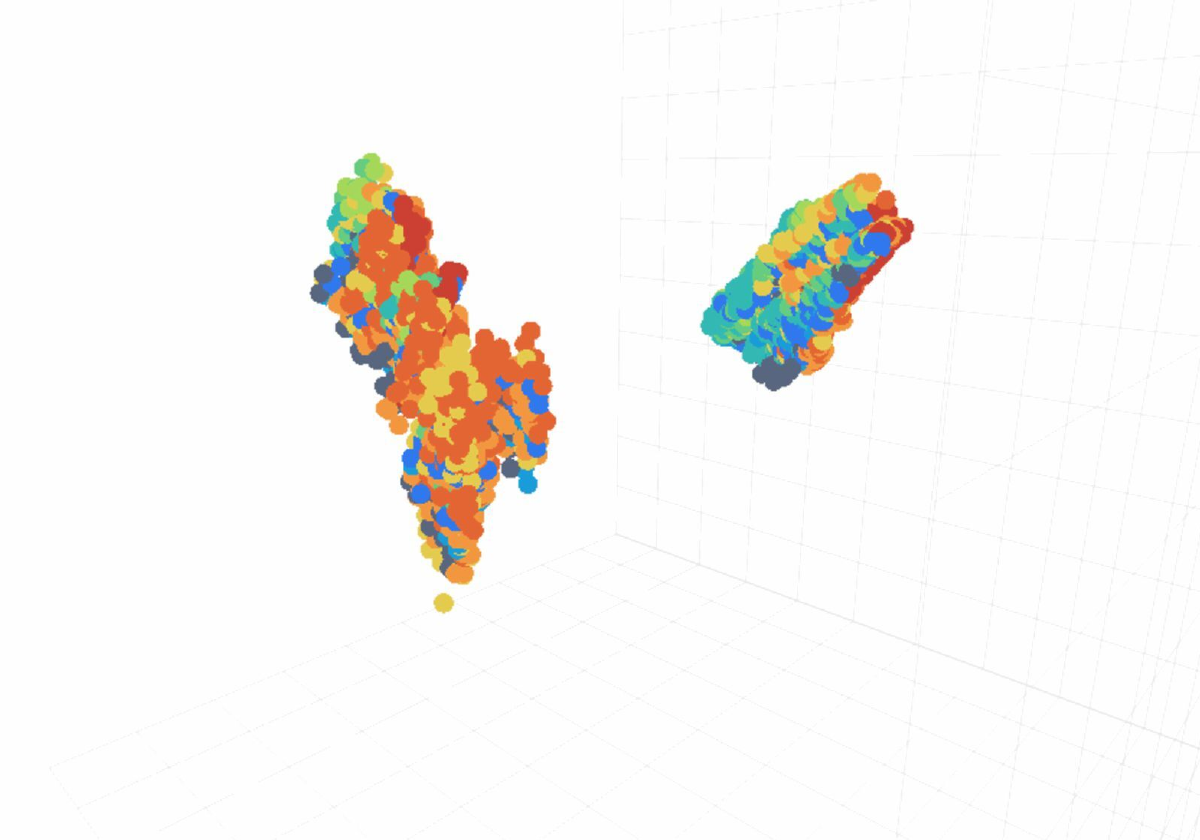}%
         {app:parametric-geo-bot}
\end{minipage}\hfill
\begin{minipage}[t]{0.24\textwidth}
  \galimg{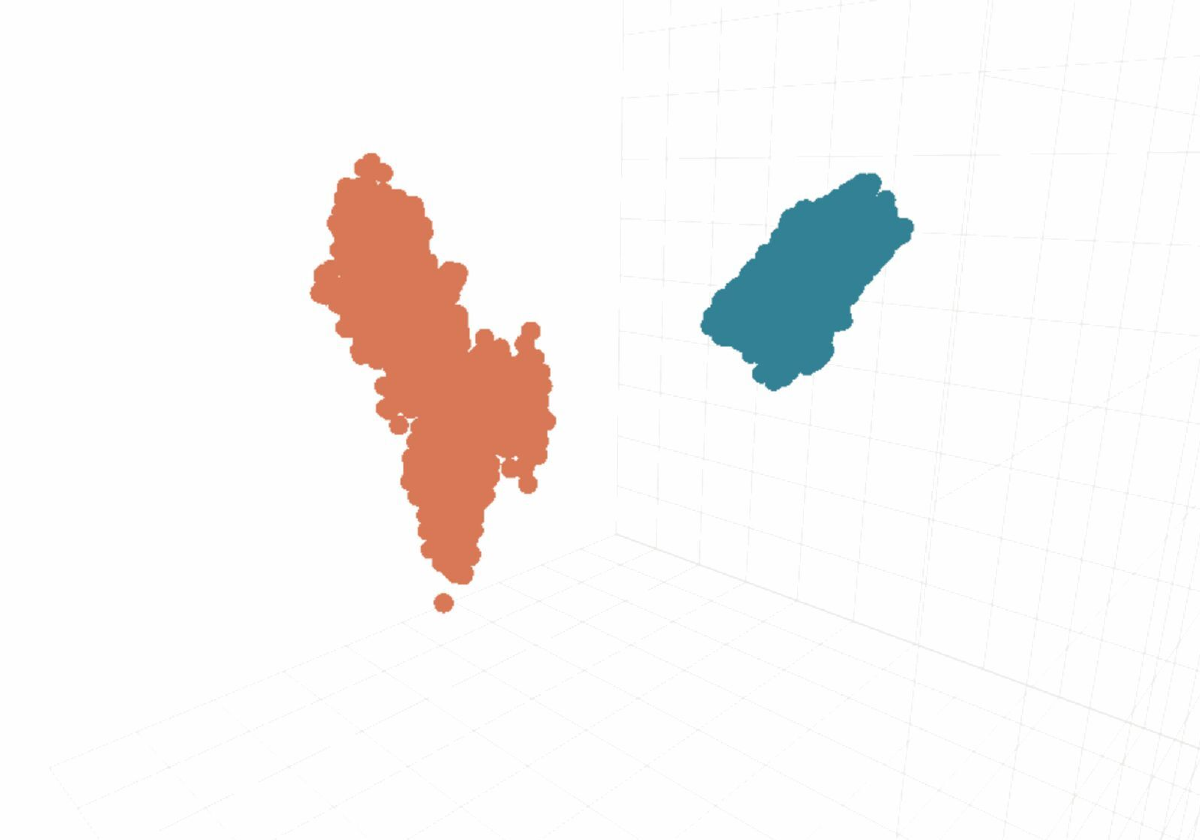}%
         {app:parametric-geo-bot}
\end{minipage}

\vspace{1em}

\noindent
\begin{minipage}[t]{0.24\textwidth}
  \galimg{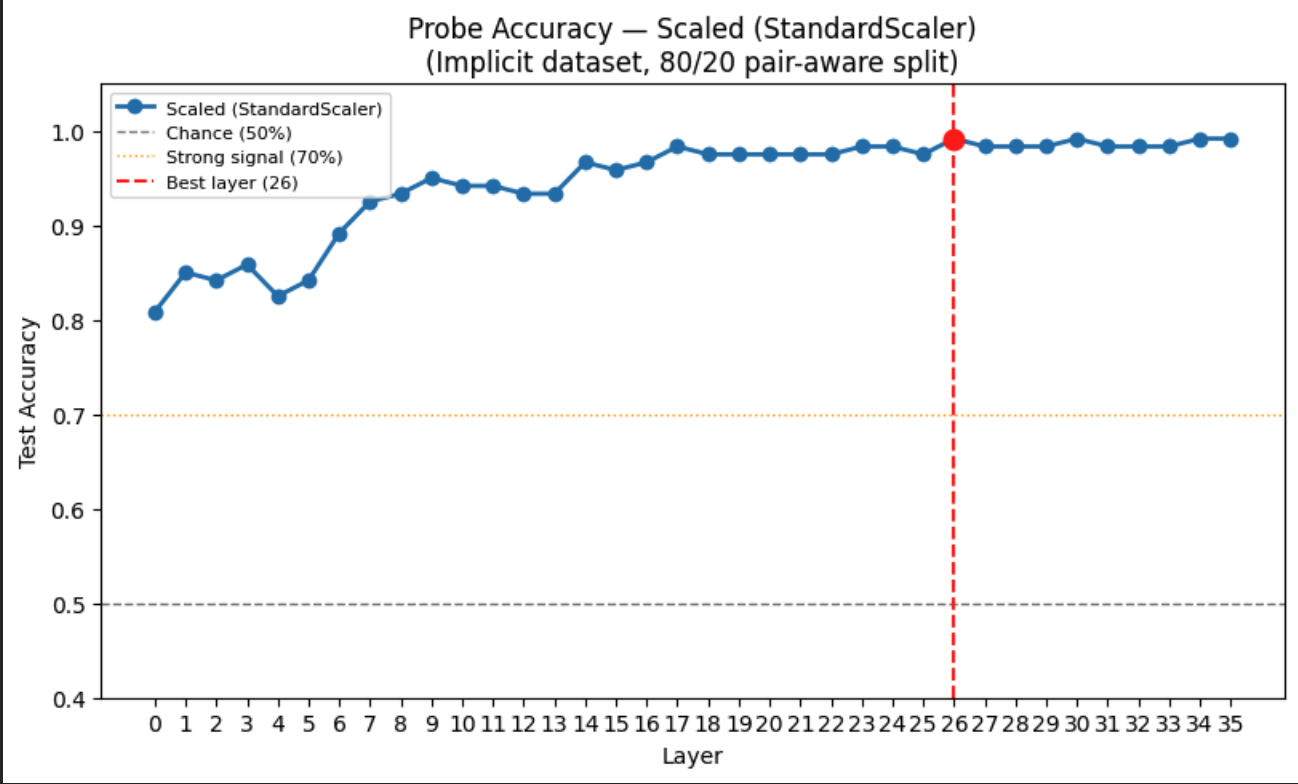}%
         {app:contrastive-probing-linear}
\end{minipage}\hfill
\begin{minipage}[t]{0.24\textwidth}
  \galimg{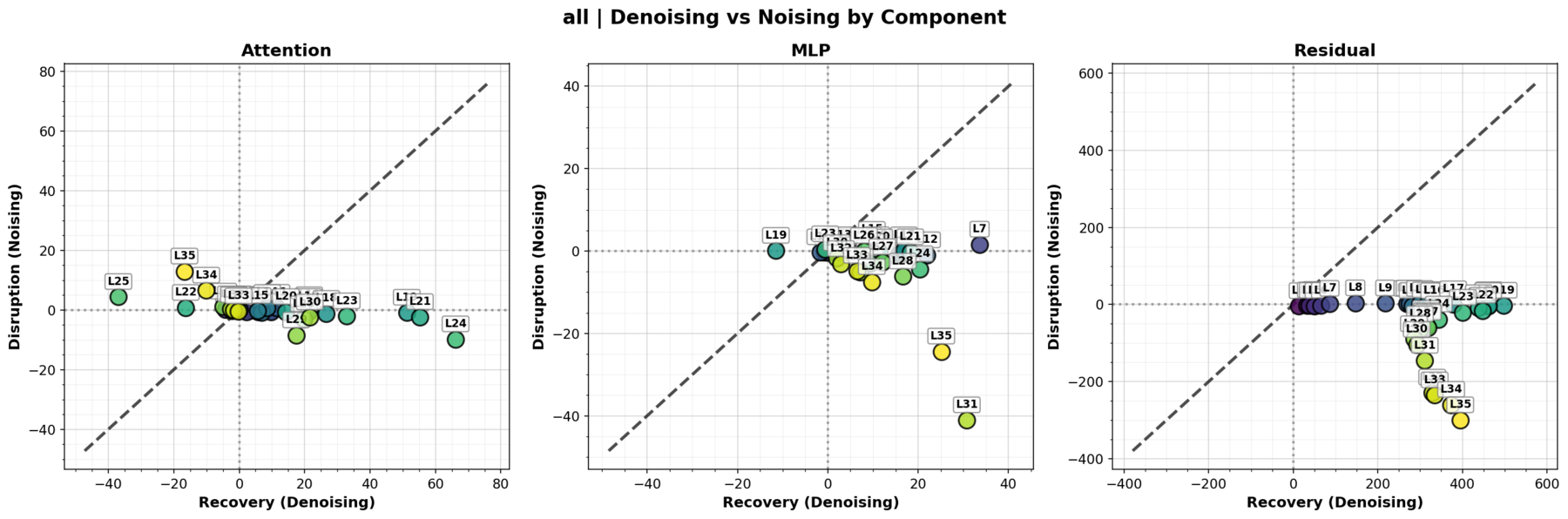}%
         {app:attributional-parametric}
\end{minipage}\hfill
\begin{minipage}[t]{0.24\textwidth}
  \galimg{images/localize/causal_contrastive/all_tokens_top_20_component_importance.png}%
         {app:causal-contrastive}
\end{minipage}\hfill
\begin{minipage}[t]{0.24\textwidth}
  \galimg{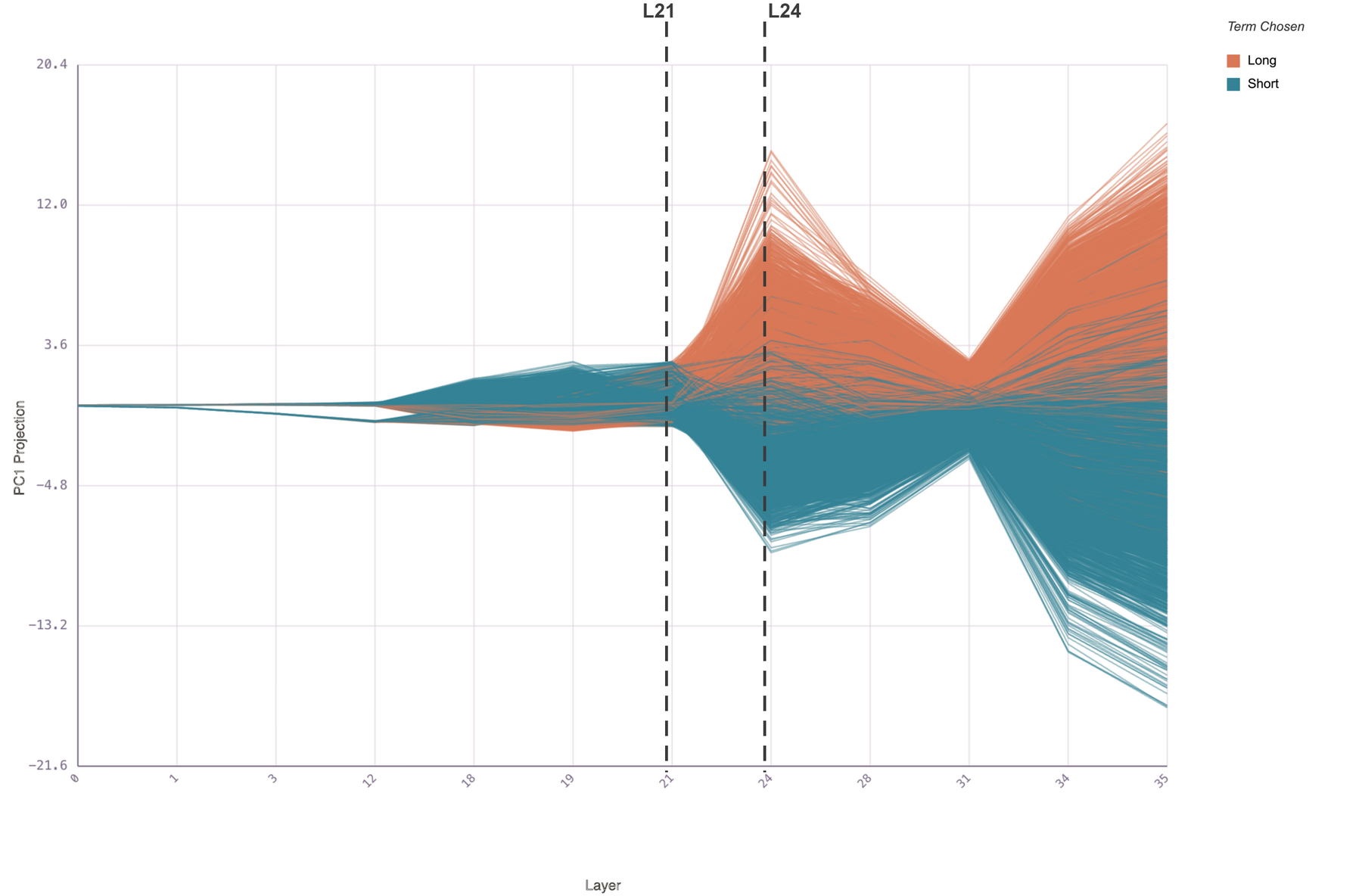}%
         {app:parametric-geo-layers}
\end{minipage}

\clearpage

\clearpage
\clearappnumbering

\partpagecontent{Part 0:}{Groundwork}{%
\begin{itemize}[leftmargin=*, itemsep=0.8em]
  \item \textbf{\hyperref[app:extended-background]{A}.} Extended background
  \item \textbf{\hyperref[app:extended-literature]{B}.} Extended literature
  \item \textbf{\hyperref[app:methodology-summary]{C}.} Methodology summary
  \item \textbf{\hyperref[app:experimental-details]{D}.} Experimental details
  \item \textbf{\hyperref[app:prompts]{E}.} Prompts
  \item \textbf{\hyperref[app:extended-limitations]{F}.} Extended limitations and future work
\end{itemize}%
}

\section{Extended background}\label{app:extended-background}

This appendix preserves the full-length background that the main text condenses for space.
It defines the temporal-preference concepts we use, introduces intertemporal choice as the empirical handle, and reviews the mechanistic-interpretability primitives our pipeline rests on: causal and attributional localization, representational geometry, and steering.
The related-work discussion lives separately in~\ref{app:extended-literature}.

\subsection{Temporal preference, horizon, and scope}

We refer to \textbf{temporal preference} as the degree to which an agent values outcomes differently depending on when they occur~\citep{cohen2020measuring, wang2025measuring}.
We use the term \textbf{time horizon} to denote the future moment at which outcomes are evaluated against an objective~\citep{ebert1973time}.
Because future events do not affect past outcomes, the time horizon also acts as a \textit{constraint} on planning~\citep{reilly2016time}.
It would be \textit{instrumentally}~\citep{korsgaard1997} or \textit{means-end}~\citep{bratman1981} \textbf{incoherent} for the agent to choose actions that are incapable of causing effects by the specified deadline.
Then, the \textbf{temporal scope} is the bounded interval of time over which an agent weighs the results according to its preference\perishablefootnote.

\begin{figure}[htbp]
    \centering
    \resizebox{0.8\textwidth}{!}{

\definecolor{colTitle}{HTML}{6B7D8E}
\definecolor{colHorizon}{HTML}{1A1A2E}
\definecolor{colBoundary}{HTML}{3D5A80}
\definecolor{colRetro}{HTML}{6B7280}
\definecolor{colCurve}{HTML}{CC785C}
\definecolor{colFill}{HTML}{D4956B}
\definecolor{colScope}{HTML}{C4703E}
\definecolor{colAxis}{HTML}{A0A8B0}
\definecolor{colLabel}{HTML}{5A6570}
\definecolor{colGrayFill}{HTML}{D5DAE0}
\definecolor{colGrayCurve}{HTML}{A0A8B0}
\definecolor{colDark}{HTML}{2D3748}
\definecolor{colRewardFill}{HTML}{5BA4B5}
\definecolor{colValueA}{HTML}{CC785C}
\definecolor{colValueB}{HTML}{5BA4B5}
\definecolor{colOp}{HTML}{1A1A2E}

\pgfmathdeclarefunction{lam}{1}{%
  \pgfmathparse{exp(-0.13*(#1))*(1+0.15*sin(70*(#1))+0.08*cos(130*(#1))+0.05*sin(200*(#1)))}%
}
\pgfmathdeclarefunction{rew}{1}{%
  \pgfmathparse{-0.05+1.6*(#1)/(1+0.2*(#1))}%
}
\pgfmathdeclarefunction{val}{1}{%
  \pgfmathparse{lam(#1)*rew(#1)}%
}

\newcommand{\TT}{44}
\newcommand{\OO}{80}
\newcommand{\symscale}{1.25}
\newcommand{\tscale}{1}

\newcommand{\aspectR}{0.50}
\newcommand{\Pscale}{2.0}
\pgfmathsetmacro{\Pw}{10*\Pscale}
\pgfmathsetmacro{\Ph}{\Pw*\aspectR}

\newcommand{\tRetroVal}{1.2}
\newcommand{\tHorizVal}{6.75}
\pgfmathsetmacro{\pRetro}{\tRetroVal*\Pscale}
\pgfmathsetmacro{\pHoriz}{\tHorizVal*\Pscale}

\pgfmathsetmacro{\valAtRetro}{\Ph*0.8*lam(\tRetroVal)*rew(\tRetroVal)/2.6}
\pgfmathsetmacro{\valAtHoriz}{\Ph*0.8*lam(\tHorizVal)*rew(\tHorizVal)/2.6}
\pgfmathsetmacro{\rewAtRetro}{\Ph*0.85*rew(\tRetroVal)/5.5}
\pgfmathsetmacro{\rewAtHoriz}{\Ph*0.85*rew(\tHorizVal)/5.5}
\pgfmathsetmacro{\lamAtHoriz}{\Ph*lam(\tHorizVal)}

\newcommand{\opGap}{3.5}
\newcommand{\opW}{2.0}

\pgfmathsetmacro{\eqX}{\Pw+\opGap+\opW/2}          
\pgfmathsetmacro{\panBx}{\Pw+\opGap+\opW+\opGap}   
\pgfmathsetmacro{\mulX}{\panBx+\Pw+\opGap+\opW/2}  
\pgfmathsetmacro{\panCx}{\panBx+\Pw+\opGap+\opW+\opGap} 

\newcommand{\titleGY}{16.0}
\newcommand{\symGY}{12.5}
\pgfmathsetmacro{\opY}{\Ph/2}

\begin{tikzpicture}[>=Stealth]

\node[colTitle, anchor=south, font=\fontsize{\TT}{\TT}\selectfont\bfseries]
  at ({\Pw/2}, \titleGY) {Value};
\node[colTitle, anchor=south, font=\fontsize{\TT}{\TT}\selectfont\bfseries]
  at ({\panBx+\Pw/2}, \titleGY) {Temporal Preference};
\node[colTitle, anchor=south, font=\fontsize{\TT}{\TT}\selectfont\bfseries]
  at ({\panCx+\Pw/2}, \titleGY) {Reward};

\node[colLabel, anchor=south, font=\fontsize{\TT}{\TT}\selectfont]
  at ({\Pw/2}, \symGY) {\scalebox{\symscale}{$V(t)$}};
\node[colLabel, anchor=south, font=\fontsize{\TT}{\TT}\selectfont]
  at ({\panBx+\Pw/2}, \symGY) {\scalebox{\symscale}{$\lambda(t)$}};
\node[colLabel, anchor=south, font=\fontsize{\TT}{\TT}\selectfont]
  at ({\panCx+\Pw/2}, \symGY) {\scalebox{\symscale}{$r(t)$}};

\begin{scope}[shift={(0,0)}]

  \draw[-{Stealth[length=5pt,width=3pt]}, colAxis, line width=0.5pt]
    (0,0) -- ({\Pw+0.5},0);
  \draw[-{Stealth[length=5pt,width=3pt]}, colAxis, line width=0.5pt]
    (0,0) -- (0,{\Ph+0.5});

  \draw[colRetro, line width=1pt, dashed] ({\pRetro},0) -- ({\pRetro},{\valAtRetro});
  \draw[colBoundary, line width=1.5pt, dashed] ({\pHoriz},0) -- ({\pHoriz},{\valAtHoriz});

  \begin{scope}
    \clip ({\pRetro},0) rectangle ({\pHoriz},{\Ph+1});
    \fill[colValueB, opacity=0.15]
      plot[domain=0:10, samples=100, smooth]
        ({\x*\Pscale}, {\Ph*0.8*val(\x)/2.6}) -- ({\Pw},0) -- (0,0) -- cycle;
  \end{scope}
  \begin{scope}
    \clip ({\pRetro},0) rectangle ({\pHoriz},{\Ph+1});
    \fill[colValueA, opacity=0.18]
      plot[domain=0:10, samples=100, smooth]
        ({\x*\Pscale}, {\Ph*0.8*val(\x)/2.6}) -- ({\Pw},0) -- (0,0) -- cycle;
  \end{scope}

  \begin{scope}
    \clip (0,0) rectangle ({\pRetro},{\Ph+1});
    \draw[colGrayCurve, line width=3pt, opacity=0.55]
      plot[domain=0:10, samples=100, smooth] ({\x*\Pscale}, {\Ph*0.8*val(\x)/2.6});
  \end{scope}
  \begin{scope}
    \clip ({\pRetro},0) rectangle ({\pHoriz},{\Ph+1});
    \draw[colDark, line width=3pt, smooth]
      plot[domain=0:10, samples=100] ({\x*\Pscale}, {\Ph*0.8*val(\x)/2.6});
  \end{scope}
  \begin{scope}
    \clip ({\pHoriz},0) rectangle ({\Pw+1},{\Ph+1});
    \draw[colGrayCurve, line width=3pt, opacity=0.55, dashed]
      plot[domain=0:10, samples=100, smooth] ({\x*\Pscale}, {\Ph*0.8*val(\x)/2.6});
  \end{scope}

  \node[colScope, font=\fontsize{44}{46}\selectfont\bfseries, anchor=north, align=center]
    at ({0.5*(\pRetro+\pHoriz)}, -0.8) {Temporal\\Scope};

  \node[colLabel, anchor=north west, font=\fontsize{\TT}{\TT}\selectfont]
    at ({\Pw+0.5}, -0.1) {\scalebox{\tscale}{$t$}};

\end{scope}

\node[colOp, font=\fontsize{\OO}{\OO}\selectfont\bfseries]
  at (\eqX, \opY) {\textsf{=}};

\begin{scope}[shift={(\panBx, 0)}]

  \begin{scope}
    \clip (0,0) rectangle ({\pHoriz},{\Ph});
    \fill[colFill, opacity=0.18]
      plot[domain=0:10, samples=120, smooth]
        ({\x*\Pscale}, {\Ph*lam(\x)}) -- ({\Pw},0) -- (0,0) -- cycle;
  \end{scope}

  \draw[-{Stealth[length=5pt,width=3pt]}, colAxis, line width=0.5pt]
    (0,0) -- ({\Pw+0.5},0);
  \draw[-{Stealth[length=5pt,width=3pt]}, colAxis, line width=0.5pt]
    (0,0) -- (0,{\Ph+0.5});

  \begin{scope}
    \clip (0,0) rectangle ({\pHoriz},{\Ph});
    \draw[colCurve, line width=4pt]
      plot[domain=0:10, samples=120, smooth] ({\x*\Pscale}, {\Ph*lam(\x)});
  \end{scope}
  \begin{scope}
    \clip ({\pHoriz},0) rectangle ({\Pw+1},{\Ph});
    \draw[colGrayCurve, line width=3.5pt, opacity=0.55, dashed]
      plot[domain=0:10, samples=120, smooth] ({\x*\Pscale}, {\Ph*lam(\x)});
  \end{scope}

  \draw[colBoundary, line width=5pt] ({\pHoriz},0) -- ({\pHoriz},{\lamAtHoriz+1.5});
  \node[colHorizon, font=\fontsize{44}{46}\selectfont\bfseries, anchor=south]
    at ({\pHoriz}, {\Ph*0.82}) {Time Horizon};

  \node[colLabel, anchor=north west, font=\fontsize{\TT}{\TT}\selectfont]
    at ({\Pw+0.5}, -0.1) {\scalebox{\tscale}{$t$}};

\end{scope}

\node[colOp, font=\fontsize{\OO}{\OO}\selectfont\bfseries]
  at (\mulX, \opY) {\textsf{×}};

\begin{scope}[shift={(\panCx, 0)}]

  \begin{scope}
    \clip (0,0) rectangle ({\pRetro},{\Ph+1});
    \fill[colGrayFill, opacity=0.22]
      plot[domain=0:10, samples=80, smooth]
        ({\x*\Pscale}, {\Ph*0.85*rew(\x)/5.5}) -- ({\Pw},0) -- (0,0) -- cycle;
  \end{scope}
  \begin{scope}
    \clip ({\pRetro},0) rectangle ({\Pw},{\Ph+1});
    \fill[colRewardFill, opacity=0.15]
      plot[domain=0:10, samples=80, smooth]
        ({\x*\Pscale}, {\Ph*0.85*rew(\x)/5.5}) -- ({\Pw},0) -- (0,0) -- cycle;
  \end{scope}

  \draw[-{Stealth[length=5pt,width=3pt]}, colAxis, line width=0.5pt]
    (0,0) -- ({\Pw+0.5},0);
  \draw[-{Stealth[length=5pt,width=3pt]}, colAxis, line width=0.5pt]
    (0,0) -- (0,{\Ph+0.5});

  \draw[colDark, line width=3pt, smooth]
    plot[domain=0:10, samples=80] ({\x*\Pscale}, {\Ph*0.85*rew(\x)/5.5});

  \draw[colRetro, line width=5pt] ({\pRetro},0) -- ({\pRetro},{\rewAtRetro*0.6});

  \draw[colBoundary, line width=3pt, dash pattern=on 6pt off 4pt] ({\pHoriz},0) -- ({\pHoriz},{\rewAtHoriz});

  \draw[-{Stealth[length=8pt,width=5pt]}, colRetro, line width=2pt]
    ({\pHoriz},{\rewAtRetro*0.6+0.4}) -- ({\pRetro},{\rewAtRetro*0.6+0.4});
  \node[colRetro, font=\fontsize{22}{24}\selectfont\bfseries, anchor=south, align=center]
    at ({0.5*(\pRetro+\pHoriz)}, {\rewAtRetro*0.6+0.8}) {Retroactive\\Reach};

  \node[colLabel, anchor=north west, font=\fontsize{\TT}{\TT}\selectfont]
    at ({\Pw+0.5}, -0.1) {\scalebox{\tscale}{$t$}};

\end{scope}

\pgfresetboundingbox
\path[use as bounding box] (-0.5,-4.5) rectangle ({\panCx+\Pw+2.5},\titleGY+2);

\end{tikzpicture}}
    \caption{
        The time horizon specifies when the consequences of a decision are assessed.
        The temporal scope is then bounded above by the time horizon.
    }
    \label{fig:temporal-defs-appendix}
\end{figure}
\FloatBarrier

In humans, these concepts have localized neural representations~\citep{kable2007neural} that predict behavior~\citep{shamosh2008delay}, respond causally to intervention~\citep{figner2010lateral}, and exhibit internal organization interpretable as a \emph{functional role}~\citep{cummins1975functional}.\footnote{Some authors argue that concepts are best modeled by geometric or topological spaces~\citep{gardenfors2000conceptual}, a perspective that resonates with our geometric analysis of temporal representations in activation space.}
Our work asks whether temporal preference exists in an LLM in an analogous way: localized, predictive, causally efficacious, and geometrically organized.

\subsection{Modeling behavior via intertemporal choice}

We measure temporal preference through \textbf{intertemporal choice}: forced binary decisions between options that differ in reward and delay.
This is the standard instrument in behavioral economics and neuroeconomics~\citep{kirby1999, frederick2002time, kable2007neural} because it isolates preference from effort, attention, and planning; separates reward from delay; and produces a single forced-choice token we can align across prompts and patch at the activation level.
Each option $i$ is defined as a tuple $(r_i, t_i)$, where $r_i \in \mathbb{R}^{+}$ denotes the reward and $t_i \in \mathbb{R}^{+}$ the delay until receipt.
The subjective value of an option is the product of a temporal preference and the reward:%
\begin{equation}
  V(t) = \lambda(t) \cdot r(t)
  \label{eq:temporal-value-appendix}
\end{equation}
Given two options $A = (r_A, t_A)$ and $B = (r_B, t_B)$, we predict the agent selects the option with the highest value:
\begin{equation}
  i^{*} = \operatorname*{arg\,max}_{i \in \{A, B\}} \; \lambda(t_i) \cdot r_i
  \label{eq:intertemporal-choice-appendix}
\end{equation}
In the case of humans, temporal preference is often modeled using discount functions that capture our tendency to prefer immediate rewards over future ones~\citep{frederick2002time, green2004discounting}.
For AI agents, it is possible that different classes of functions better model their behavior~\citep{mazyaki2025temporal}.
Characterizing LLM behavior requires fitting a discount function via regression, assessing its stability across varying contexts, and benchmarking the resulting preferences against human intertemporal choice.

\subsection{Localizing a subgraph}

The process of \textit{subgraph localization} within an LLM involves identifying which components of the neural network are responsible for the behavior of interest.
The gold standard is \textit{causal localization}~\citep{geiger2025causalabstractiontheoreticalfoundation}, which works by intervening within an LLM to measure the causal effect of specific components on the behavior of the model.
We adopt \textbf{activation patching}~\citep{heimersheim2024useinterpretactivationpatching} as our causal technique (Section~\ref{app:causal-parametric}, Section~\ref{app:causal-contrastive}): replace one component's activation with a counterfactual value from another input and measure the behavioral change.
Attribution scores components via gradients and probing reads linearly decodable information, but neither intervenes on the forward pass; patching is the only one of the three that tests whether a component is causally necessary or sufficient for the output.
Using the do-calculus notation~\citep{pearl2009causality}, the patching intervention can be expressed as:
\begin{equation}
    \Delta_i^{(l)}(x,\, x')
    \;=\;
    \mathbb{E}\bigl[\, Y \mid \Do \bigl(\, a_i^{(l)} = a_i^{(l)}(x') \,\bigr),\; X = x \,\bigr]
    \;-\;
    \mathbb{E}\bigl[\, Y \mid X = x \,\bigr]
\end{equation}

Unfortunately, performing targeted interventions is computationally expensive.
As an alternative, \textit{attributional localization} approximates causal localization~\citep{bereska2024mechanisticinterpretabilityaisafety}.
We use \textbf{EAP-IG}~\citep{hanna2024faithfaithfulnessgoingcircuit}, a gradient-based attribution method (Section~\ref{app:attributional-contrastive}) that scores every head and MLP in a single backward pass.
This makes a full-network scan tractable; the tradeoff is correlational estimates rather than causal guarantees.
We also use \textbf{probes}, linear classifiers trained on a model's internal activations~\citep{mueller2025mibmechanisticinterpretabilitybenchmark, kim2025linearrepresentationspoliticalperspective}, to give us a complementary view by identifying which concepts a model encodes, where they emerge, and whether they are linearly represented.

Including more components in a subgraph explains more of the LLM's behavior, but yields a larger, less interpretable picture.
The full network trivially explains everything, and the empty subgraph explains nothing.
Any useful circuit falls between these extremes, balancing behavioral coverage against subgraph size.

\subsection{Visualizing representational geometry}

The activation space within an LLM encodes concepts in internal representations~\citep{bereska2024mechanisticinterpretabilityaisafety}.
A growing body of evidence has documented that many representations possess complex geometric structures~\citep{karkada2026symmetrylanguagestatisticsshapes, engels2025notall, modell2025origins, gurnee2026models}, beyond the global directions that the \textit{Linear Representation Hypothesis} predicts~\citep{park2024linear}.
Furthermore, recent work has also noticed that LLMs show local low-dimensional structure~\citep{shafran2026directionsregionsdecomposingactivations, saglam2025largelanguagemodelsencode, lee2025geometricsignaturescompositionalitylanguage}, and that even when representations are linear, they can change dramatically throughout generation~\citep{lampinen2026linearrepresentationslanguagemodels}.

These past findings motivate us to visualize the representational geometry within the localized subgraph as a way to understand what each component is doing.
In our work, we apply Principal Component Analysis (PCA)~\citep{shlens2014} to examine how the temporal concepts of interest are represented within a lower-dimensional subspace.

\subsection{Steering behavior with interventions}

Steering refers to the control of an LLM's behavior by directly intervening on its internal representations, rather than through prompting or training.
Subgraph localization is not required for steering~\citep{bartoszcze2025representationengineeringlargelanguagemodels, wehner2025taxonomyopportunitieschallengesrepresentation}, but localization generally improves precision, reduces side effects, and allows smaller intervention magnitudes~\citep{zhang2026locatesteerimprovepractical}.
In our work, we seek to understand the interventions in our subgraph through both a geometric and a behavioral perspective.

\clearpage
\clearappnumbering

\section{Extended literature}\label{app:extended-literature}

Understanding temporal preference in LLMs requires drawing together the literature that has largely developed in isolation.

\paragraph{Temporal Representation.}
LLMs encode temporal and spatial coordinates as geometric objects recoverable via regression probes~\citep{gurnee2024language, nylund2024time, holtermann2025world24hoursprobing}, forming circular, helical, and manifold structures~\citep{engels2025notall, kantamneni2025trigonometry, kadlcik2025sinusoidal, hanna2023does, modell2025origins, tiblias2025shape, gurnee2026models} that obey psychophysical scaling laws~\citep{cacioli2026weberslawtransformermagnitude, karkada2026symmetrylanguagestatisticsshapes}.
Linear decodability coexists with non-linear geometry because features are locally linear on globally curved manifolds~\citep{modell2025origins, park2024linear, shafran2026directionsregionsdecomposingactivations, rajendran2024from}.
Yet the best-geometry layer is not the computational layer~\citep{cacioli2026categoricalperceptionlargelanguage}, causality is rarely established~\citep{heinzerling2024monotonic}, and it is not known whether temporal representations causally drive downstream behavior across contexts the way emotion concepts do~\citep{anthropic-emotions}.

\paragraph{Temporal Reasoning and Planning.}
LLMs fail at temporal reasoning tasks despite encoding time geometrically~\citep{wang2024trambenchmarkingtemporalreasoning, fatemi2024testtimebenchmarkevaluating, liu2025timer1comprehensivetemporalreasoning, wallat2025studyinvestigatingtemporalrobustness, herel2025timeawarenesslargelanguage}, and lack continuous temporal grounding: they cannot track real-time deadlines even when discrete turn-based reasoning succeeds~\citep{sehgal2026realtimedeadlinesrevealtemporal, garikaparthi2026llmsperceivetimeempirical, cheng2025temporalblind}.
Evidence of temporal structure exists~\citep{papadopoulos2024arrowstimelargelanguage, li2025mindlanguagemodelsexhibit, david2025temporal}, and targeted fixes have been proposed~\citep{han2025temporalalignmentllmscycle, shin2025tardis}, but none connect temporal geometry to temporal decision-making.
Separately, token-level lookahead over discrete sequence positions is detectable via probing~\citep{nainani2025detectingcharacterizingplanninglanguage, song2025llminterpretabilityidentifiabletemporalinstantaneous}, but operates over next-token predictions rather than real-valued time horizons; both modes fail at long-horizon decisions~\citep{wang2026reasoningfailsplanplanningcentric, cao2025largelanguagemodelsplanning}.

\paragraph{LLM Economic Behavior and Risk Preference.}
LLMs reproduce behavioral-economic biases~\citep{horton2026largelanguagemodelssimulated, cook2026llms, chen2025financial, ross2024llm, leng2024folk, leng2024canllms, cacioli2026categoricalperceptionlargelanguage} with unstable risk preferences~\citep{wang2025prospect}.
Risk and time preferences can be steered neurally~\citep{zhu2025steering, zhu2025languagemodelstrainedarithmetic} but entangle through discount factors~\citep{fedus2019hyperbolicdiscountinglearningmultiple, moghimi2026decouplingtimeriskrisksensitive}.
Temporal preferences have been studied only behaviorally~\citep{mazyaki2025temporal}; whether they form a steerable activation-space direction, as shown for truthfulness~\citep{marks2024geometry, ying2026truthfulnessspectrumhypothesis} and emotion~\citep{anthropic-emotions}, is open.

\paragraph{Steering Advancements.}
Representation engineering~\citep{zou2025representationengineeringtopdownapproach} has progressed from activation addition~\citep{turner2023activation, panickssery2024steering} and representation interventions~\citep{wu2024reft} through sparse dictionaries~\citep{cunningham2023sparse, dunefsky2024transcoders} to geometric approaches~\citep{vu2025angular, spherical2026, postmus2025conceptors}.
Prompting still leads on many benchmarks~\citep{wu2025axbench, engels2025takeaways}, and over-steering degrades helpfulness~\citep{wolf2024tradeoffs, NEURIPS2024_f5454485}.
Patching along continuous numeric directions produces monotonic output shifts~\citep{heinzerling2024monotonic}, but static vectors assume a fixed concept direction; when the effective direction varies with context or curves through activation space, they become misaligned or unreliable~\citep{li2026svf, curveball2026, braun2025understandingunreliabilitysteeringvectors}.

No prior work has localized a subgraph functionally responsible for temporal preference, characterized the geometry of the causal representation, or steered along it.
Table~\ref{tab:related-work-gap} situates our contribution against the most directly comparable works on six axes.

\begin{table}[htbp]
  \centering
  \small
  \setlength{\tabcolsep}{4pt}
  \renewcommand{\arraystretch}{1.15}
  \resizebox{\textwidth}{!}{%
  \begin{tabular}{@{}l l c c c c c c@{}}
    \toprule
    \textbf{Work} & \textbf{Concept}
      & \makecell{\textbf{Causal}\\\textbf{subgraph}}
      & \textbf{Geometry}
      & \textbf{Steering}
      & \makecell{\textbf{Layer-}\\\textbf{localized}\\\textbf{steering}}
      & \makecell{\textbf{Human}\\\textbf{baseline}}
      & \makecell{\textbf{Latent $+$}\\\textbf{explicit}} \\
    \midrule
    \multicolumn{8}{@{}l}{\emph{Temporal representation}} \\
    \citet{gurnee2024language}                    & space/time      & \xmark & \xmark    & \xmark & \xmark & \xmark    & \xmark \\
    \citet{engels2025notall}                      & days/months     & \cmark & \cmark    & partial & \xmark & \xmark   & \xmark \\
    \citet{modell2025origins}                     & theory          & \xmark & \cmark    & \xmark & \xmark & \xmark    & \xmark \\
    \citet{gurnee2026models}                      & counting        & \cmark & \cmark    & \xmark & \xmark & \xmark    & \xmark \\
    \midrule
    \multicolumn{8}{@{}l}{\emph{Temporal reasoning and planning}} \\
    \citet{wang2026reasoningfailsplanplanningcentric} & planning     & \xmark & \xmark    & \xmark & \xmark & \xmark    & \xmark \\
    \citet{sehgal2026realtimedeadlinesrevealtemporal} & deadlines    & \xmark & \xmark    & \xmark & \xmark & \xmark    & \xmark \\
    \midrule
    \multicolumn{8}{@{}l}{\emph{LLM economic behavior and preference}} \\
    \citet{zhu2025steering}                       & risk pref.      & \xmark & \xmark    & \cmark & \cmark & \xmark    & \cmark \\
    \citet{mazyaki2025temporal}                   & temporal pref.  & \xmark & \xmark    & \xmark & \xmark & \cmark    & \xmark \\
    \citet{horton2026largelanguagemodelssimulated, cook2026llms} & economic bias & \xmark & \xmark & \xmark & \xmark & \cmark & \xmark \\
    \midrule
    \multicolumn{8}{@{}l}{\emph{Steering methods}} \\
    \citet{turner2023activation, panickssery2024steering} & generic   & \xmark & \xmark    & \cmark & \cmark & \xmark   & \xmark \\
    \citet{marks2024geometry}                     & truthfulness    & \cmark & \xmark    & \cmark & \xmark & \xmark    & \xmark \\
    \citet{anthropic-emotions}                    & emotion         & \cmark & \cmark    & \cmark & \xmark & \cmark    & partial \\
    \midrule
    \rowcolor{black!5}
    \textbf{This work}                            & \textbf{temporal pref.} & \cmark & \cmark & \cmark & \cmark & \cmark & \cmark \\
    \bottomrule
  \end{tabular}%
  }
  \caption{Our contribution against the closest prior work on six axes:
  (i) whether a causal subgraph is identified, not just a probe direction;
  (ii) whether concept geometry is non-linear / curved;
  (iii) whether steering traverses a dimensional axis rather than a binary contrast;
  (iv) whether steering is layer-localized rather than applied uniformly;
  (v) whether outcomes are benchmarked against human behavior;
  (vi) whether both latent (no-horizon) and explicitly parameterized prompts are analyzed together.
  No prior work covers all six for temporal preference.}
  \label{tab:related-work-gap}
\end{table}

\clearpage
\clearappnumbering

\section{Methodology summary}\label{app:methodology-summary}

Our methodology follows three stages: \emph{localize} the subgraph, \emph{characterize} the representations, and \emph{intervene}.
Each stage has one or more dedicated experimental pipelines, and each pipeline has its own full-detail methodology appendix collected in Part~4 (see \S\ref{app:methodology-extended-overview}).

\begin{figure}[!htbp]
  \centering
  \begin{minipage}{0.48\textwidth}
    \centering
    \resizebox{\textwidth}{!}{\definecolor{s1fill}{HTML}{E0DAF5}
\definecolor{s1stroke}{HTML}{8B7BC7}
\definecolor{s1text}{HTML}{3B2D6E}
\definecolor{s2fill}{HTML}{D6ECFA}
\definecolor{s2stroke}{HTML}{5EA4D4}
\definecolor{s2text}{HTML}{1C4F73}
\definecolor{s3fill}{HTML}{FDE8CE}
\definecolor{s3stroke}{HTML}{D4943C}
\definecolor{s3text}{HTML}{6B4210}

\begin{tikzpicture}[
    node distance=1.2cm,
    stage/.style={
        rounded corners=8pt,
        minimum height=2cm,
        minimum width=4.2cm,
        font=\sffamily,
        align=center,
        line width=1pt,
    },
    s1/.style={stage, fill=s1fill, draw=s1stroke, text=s1text},
    s2/.style={stage, fill=s2fill, draw=s2stroke, text=s2text},
    s3/.style={stage, fill=s3fill, draw=s3stroke, text=s3text},
    arr/.style={-{Stealth[length=7pt, width=5pt]}, gray!40, line width=1.5pt},
]

\node[s1] (loc) {
    {\large\textbf{Localize Subgraphs}}\\[6pt]
    {\ttfamily\scriptsize PARAMETRIC + CONTRASTIVE + CLASSIFICATION}\\[1pt]
    {\ttfamily\scriptsize QUERYING}
};

\node[s2, right=of loc] (char) {
    {\large\textbf{Characterize Geometry}}\\[6pt]
    {\ttfamily\scriptsize ACTIVATION SUBSPACE} \\[1pt]
    {\ttfamily\scriptsize PCA}
};

\node[s3, right=of char] (int) {
    {\large\textbf{Intervene}}\\[6pt]
    {\ttfamily\scriptsize STEERING ALONG}\\[1pt]
    {\ttfamily\scriptsize MANIFOLD}
};

\draw[arr] (loc) -- (char);
\draw[arr] (char) -- (int);

\end{tikzpicture}}
  \end{minipage}%
  \hfill
  \begin{minipage}{0.48\textwidth}
    \centering
    \resizebox{\textwidth}{!}{\definecolor{r1fill}{HTML}{D2F0E6}
\definecolor{r1stroke}{HTML}{5CB88A}
\definecolor{r1text}{HTML}{14614A}
\definecolor{r2fill}{HTML}{F5DDE6}
\definecolor{r2stroke}{HTML}{D4537E}
\definecolor{r2text}{HTML}{72243E}
\definecolor{r3fill}{HTML}{FAECE7}
\definecolor{r3stroke}{HTML}{D85A30}
\definecolor{r3text}{HTML}{712B13}

\begin{tikzpicture}[
    node distance=0.7cm,
    box/.style={
        rounded corners=8pt,
        minimum height=1.5cm,
        font=\sffamily,
        align=center,
        line width=1pt,
    },
    arr/.style={-{Stealth[length=7pt, width=5pt]}, line width=1.5pt},
]

\node[box, fill=r1fill, draw=r1stroke, text=r1text, minimum width=3.6cm] (probe) {
    \textbf{Parametric}\\
    \textbf{Querying}
};
\node[box, fill=r1fill, draw=r1stroke, text=r1text, minimum width=4.2cm, right=of probe] (geomtime) {
    \textbf{Geometry: \textit{g}(time)}
};
\draw[arr, r1stroke] (probe) -- (geomtime);

\node[box, fill=r2fill, draw=r2stroke, text=r2text, minimum width=3.6cm, below=0.8cm of probe] (behav) {
    \textbf{Behavioral}\\
    \textbf{Modeling}
};
\node[box, fill=r2fill, draw=r2stroke, text=r2text, minimum width=4.2cm, right=of behav] (choicetime) {
    \textbf{Behavior: \textit{b}(time)}
};
\draw[arr, r2stroke] (behav) -- (choicetime);

\node[box, fill=r3fill, draw=r3stroke, text=r3text, minimum width=5.4cm, minimum height=1.8cm,
      right=1.5cm of $(geomtime.east)!0.5!(choicetime.east)$] (result) {
    {\large\textbf{Behavior: \textit{b}(geometry)}}\\[3pt]
    {\ttfamily\scriptsize GEOMETRIC INTERVENTION}
};

\draw[arr, r1stroke] (geomtime.east) -- ++(0.5,0) |- (result.west);
\draw[arr, r2stroke] (choicetime.east) -- ++(0.5,0) |- (result.west);

\end{tikzpicture}}
  \end{minipage}
  \caption{Overview of our approach.
  Parametric querying could help us \textbf{reparametrize} our behavioral modeling as a function of activation-space geometry instead of an explicit time horizon.}
  \label{fig:method-diagram-appendix}
\end{figure}
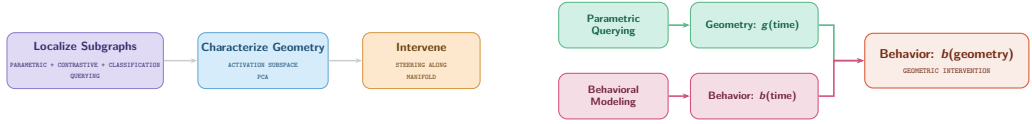

\subsection{Complementary localizations}\label{app:method-localizations}

We perform experiments with three different querying techniques applied to three corresponding prompting settings.

The first, \textbf{wide attribution}, combines contrastive querying with attribution patching and probing.
Minimally-framed prompts elicit latent preferences without explicit temporal cues, while gradient-based approximations efficiently score component importance.
This pipeline scales across samples, aggregating signal from hundreds of diverse prompts.

The second, \textbf{targeted intervention}, combines parametric querying with activation patching.
Highly-structured prompts specify explicit time horizons, while direct interventions establish the causal effect.
This pipeline disentangles causal relationships on carefully designed prompt variations.

The third, \textbf{targeted classification intervention}, combines the temporal classification task with activation patching. Each prompt presents a single goal whose horizon must be inferred and queries its short/long classification directly. This pipeline isolates temporal reasoning from valuation and tests whether the subgraph identified above is also recruited for categorical horizon judgment.

\subsection{Characterizing via geometry}\label{app:method-characterization-geometry}

We apply PCA~\citep{shlens2014} to residual-stream activations at subgraph nodes, examining how explicit time-horizon constraints (seconds to centuries) organize the activation manifold and whether latent preferences, elicited without any horizon cue, align to this geometry.
We pay particular attention to the user-to-assistant turn transition, where the model converts off-policy context into on-policy generation.
In principle, each prompt's explicit horizon maps to a point on the manifold, opening the possibility of reparametrizing behavioral discount as a function of geometry rather than time (Figure~\ref{fig:method-diagram-appendix}).
We do not pursue this reparametrization fully here, but the geometry results in \ref{app:parametric-geometry} lay the groundwork.

\subsection{Behavioral analysis}\label{app:method-characterization-behavioral}

We probe temporal preference at the behavioral level through two experiments.
First, we administer the Kirby MCQ-27~\citep{kirby1999} under multiple personas and response modes, fitting hyperbolic discount functions and introducing a \emph{decision boundary method} that binary-searches the delayed reward to locate per-item indifference points.
Second, we test behavioral coherence: whether the model's choices respect the time-horizon constraint (Section~\ref{sec:background}).
Choosing an option that cannot deliver within the specified deadline is instrumentally incoherent, and we systematically vary horizon, reward, order, label, and context to separate genuine temporal reasoning from surface heuristics.
The behavioral experiments serve a dual role: they characterize the model's temporal preferences independently of the mechanistic analysis, and they reveal the gap between the internal representation (rich, ordinal, geometrically structured) and the behavioral output (discrete, order-biased, partially incoherent).

\subsection{Steering in the wild}\label{app:method-steering}

We test causal control over temporal preference using Contrastive Activation Addition (CAA)~\citep{turner2023activation, panickssery2024steering}.
Logistic probes~\citep{mueller2025mibmechanisticinterpretabilitybenchmark, kim2025linearrepresentationspoliticalperspective} trained on the implicit dataset identify where temporal orientation is linearly decodable; the probe direction at the best layer yields a steering vector $\hat{\mathbf{v}}_{\text{CAA}}$ injected as $\mathbf{h}^{(l)} \leftarrow \mathbf{h}^{(l)} + \alpha\cdot\hat{\mathbf{v}}_{\text{CAA}}$.
We evaluate via forced-choice log-probability shifts and open-ended generation scored by an external LLM judge, sweeping layers and $\alpha$ to test for a probing--steering dissociation: whether the best layer for \emph{reading} temporal preference differs from the best layer for \emph{writing} it~\citep{heimersheim2024useinterpretactivationpatching}.

\subsection{Overview of the extended methodologies}\label{app:methodology-extended-overview}

Full protocol-level details for each pipeline, including dataset construction, sample sizes, prompt formats, component-selection thresholds, and analysis procedures, are in Part~4 of the appendices.
A reader looking for one specific experiment's full methodology can jump directly to the corresponding appendix below; the four-part organization mirrors the localize/characterize/intervene pipeline:

\paragraph{Localize (four pipelines).}
\begin{itemize}[nosep, leftmargin=*]
  \item \ref{app:contrastive-probing-linear-methods} -- Logistic-probe training protocol, activation extraction, and the token-position correction applied to the contrastive dataset.
  \item \ref{app:attributional-contrastive-methods} -- EAP-IG attribution on minimally-framed contrastive prompts, with bias controls and the component-taxonomy thresholds used to define the candidate subgraph.
  \item \ref{app:causal-parametric-methods} -- Activation patching on parametric prompts: noise/denoise protocol, position alignment across horizons, and the metric used to score each (layer, component) cell.
  \item \ref{app:causal-contrastive-methods} -- Directional activation patching on classification prompts: dataset construction, model validation and metric definitions; tests  whether the same layers are recruited for both preference and categorical horizon inference.
\end{itemize}

\paragraph{Characterize (three pipelines).}
\begin{itemize}[nosep, leftmargin=*]
  \item \ref{app:parametric-geometry-methods} -- PCA geometry pipeline: layer selection, variance-explained thresholds, and the turn-boundary analysis.
  \item \ref{app:behavioral-temporal-discount-methods} -- Kirby MCQ-27 instrument and the decision-boundary binary-search extension, including persona and response-mode conditions.
  \item \ref{app:behavioral-coherence-methods} -- 30-model investment-coherence instrument: horizon $\times$ reward $\times$ order $\times$ label $\times$ context grid and parse protocol.
\end{itemize}

\paragraph{Intervene (one pipeline).}
\begin{itemize}[nosep, leftmargin=*]
  \item \ref{app:contrastive-steering-methods} -- CAA vector construction from the best probing layer, the $\alpha$-sweep protocol, and the forced-choice / open-ended evaluation setup.
\end{itemize}

\paragraph{Case study.}
\begin{itemize}[nosep, leftmargin=*]
  \item \ref{app:case-study-hf} -- Worked token-level case study for a single highly-formatted prompt pair, tying the attribution, patching, and probing signals to specific tokens.
\end{itemize}

\clearpage
\clearappnumbering

\section{Experimental details}\label{app:experimental-details}

Full details for each experiment are in the corresponding methodology appendix.
All experiments can be run on a MacBook Pro (M4 Max, 48\,GB), except for the causal contrastive one (\ref{app:causal-contrastive-methods}), which requires 79\,GB.
The full pipeline reproduces end-to-end within two weeks.

\subsection{Why \texttt{Qwen3-4B-Instruct-2507}?}
We select \texttt{Qwen3-4B-Instruct-2507}~\citep{qwen2025instruct2507}, the non-thinking-only mode-specialized refresh of \texttt{Qwen3-4B}~\citep{yang2025qwen3}, for three reasons:
\begin{itemize}[noitemsep, topsep=0pt, leftmargin=*]
    \item \textbf{Non-thinking keeps cognition inside a fixed template.}
    The model operates exclusively in non-thinking mode: it never emits a \texttt{<think>...</think>} reasoning block, so the token positions we patch into are stable across clean and corrupted runs.
    All ``cognition'' happens inside the fixed prompt template, which is the alignment condition that activation patching and EAP-IG attribution both require.
    The hybrid-thinking \texttt{Qwen3-4B} would produce variable-length reasoning blocks that break this alignment.

    \item \textbf{Stable latent preference under perturbation.}
    Localization requires that the model's answer does not flip under minor syntactic changes.
    \texttt{Qwen3-4B-Instruct-2507} satisfies this: across the 30-model behavioral panel (\ref{app:behavioral-coherence}), it is among the most label-stable and context-stable checkpoints at its scale, while similarly-sized open-weight models drift under perturbation.

    \item \textbf{Tractable to sweep.}
    \texttt{Qwen3-4B} outperforms \texttt{Qwen2.5-7B} on most benchmarks and competes with \texttt{Qwen2.5-14B-Instruct}, \texttt{Gemma-3-12B-IT}, and \texttt{Phi-4}~\citep{yang2025qwen3}, yet the 4B footprint lets us run the full attribution-plus-patching-plus-steering pipeline on a single MacBook, with the classification patching as the only memory-bound exception.
\end{itemize}
The same mode specialization that enables this analysis also exposes the behavioral gap we study: the non-thinking variant collapses the hybrid-thinking checkpoint's graded horizon curve into three discrete order-biased modes (\ref{app:behavioral-coherence}) even though its internal temporal geometry remains rich (\ref{app:parametric-geometry}).

\subsection{Datasets}
\begin{itemize}[noitemsep, topsep=0pt, leftmargin=*]
    \item \textbf{Minimally-framed.}
    Minimally-framed A/B prompts: an \emph{explicit} set (500 pairs, 25 categories) with overt temporal markers, and an \emph{implicit} set (500 pairs, 10 categories) using only semantic framing.
    Both are counterbalanced across two orderings and seven label schemes.
    \item \textbf{Highly-formatted.}
    4{,}588 investment intertemporal choice prompts with optional horizon constraints ranging from seconds to centuries.
    \item \textbf{Classification-oriented.}
    160 IOI-style short/long classification pairs, each presenting a single goal across 25 life subdomains, with balanced question order (80 SL, 80 LS).
    \item \textbf{Behavioral.}
    Kirby MCQ-27 administered under 8 conditions (2~personas $\times$ 2~response modes), plus a binary-search decision-boundary extension.
    \item \textbf{Steering evaluation.}
    20 held-out forced-choice and 13 open-ended prompts scored by an external LLM judge.
\end{itemize}

\subsection{Subset selection criteria}\label{app:subset-selection}

Several analyses operate on filtered subsets of the datasets above; we collect the filtering rules here for cross-reference.
\begin{itemize}[noitemsep, topsep=0pt, leftmargin=*]
    \item \textbf{$n = 71$ highly-formatted parametric contrastive pairs} (used for activation patching in \ref{app:causal-parametric}).
    Filtered from the 4{,}588 generated highly-formatted samples by selecting pairs with valid clean/corrupted alignment under the piecewise-linear position mapping (\ref{app:causal-parametric-methods}) and a non-trivial baseline logit difference $|y_{\text{clean}} - y_{\text{corrupted}}|$, giving 71 pairs that admit faithful patching.
    \item \textbf{$n = 57$ constrained subset} (\ref{app:latent-vs-constrained}).
    The subset of those 71 pairs in which both clean and corrupted prompts carry an explicit time horizon.
    \item \textbf{$n = 10$ unconstrained subset} (\ref{app:latent-vs-constrained}).
    The subset of those 71 pairs in which neither prompt carries a time horizon.
    The remaining 4 pairs are mixed (one constrained, one not) and are analyzed separately in \ref{app:case-study-hf}.
    \item \textbf{$n = 160$ classification pairs} (\ref{app:causal-contrastive}).
    Filtered from 200 generated short/long classification pairs by retaining only those that \texttt{Qwen3-4B-Instruct-2507} classifies correctly, yielding the 80\% (160/200) accuracy filter documented in \ref{app:classification-oriented}.
    \item \textbf{$n = 1{,}100$ temporal-preference samples for error monitoring} (\ref{app:error-monitoring}).
    Drawn from the 500-pair $D_{\text{explicit}}$ and 500-pair $D_{\text{implicit}}$ minimally-framed datasets and used as the temporal arm of the shared error/temporal probe pipeline; no additional filtering beyond the dataset construction in \ref{app:prompts}.
\end{itemize}

\clearpage
\clearappnumbering

\section{Prompting settings}\label{app:prompts}

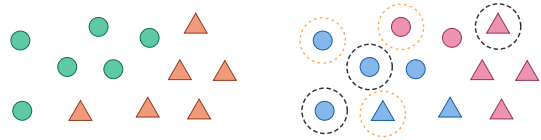
\begin{figure}[htbp]
  \centering
  \scalebox{0.5}{
\begin{tikzpicture}[scale=1, every node/.style={inner sep=0pt}]

\definecolor{tealfill}{HTML}{5DCAA5}
\definecolor{tealstroke}{HTML}{0F6E56}
\definecolor{coralfill}{HTML}{F0997B}
\definecolor{coralstroke}{HTML}{993C1D}
\definecolor{bluefill}{HTML}{85B7EB}
\definecolor{bluestroke}{HTML}{185FA5}
\definecolor{pinkfill}{HTML}{ED93B1}
\definecolor{pinkstroke}{HTML}{993556}

\tikzset{
  teal dot/.style={circle, minimum size=5mm, fill=tealfill, draw=tealstroke,
                   line width=0.3pt},
  coral tri/.style={regular polygon, regular polygon sides=3, minimum size=7mm,
                    fill=coralfill, draw=coralstroke, line width=0.3pt},
  blue dot/.style={circle, minimum size=5mm, fill=bluefill, draw=bluestroke,
                   line width=0.35pt},
  pink dot/.style={circle, minimum size=5mm, fill=pinkfill, draw=pinkstroke,
                   line width=0.35pt},
  blue tri/.style={regular polygon, regular polygon sides=3, minimum size=7mm,
                   fill=bluefill, draw=bluestroke, line width=0.35pt},
  pink tri/.style={regular polygon, regular polygon sides=3, minimum size=7mm,
                   fill=pinkfill, draw=pinkstroke, line width=0.35pt},
  ring A/.style={circle, minimum size=12mm, draw=orange!70, line width=1.0pt,
                 dash pattern=on 2pt off 2.5pt},
  ring B/.style={circle, minimum size=12mm, draw=black!80, line width=1.1pt,
                 dash pattern=on 3.5pt off 2pt},
}


\node[teal dot] at (0.37, 2.31) {};
\node[teal dot] at (2.44, 2.67) {};
\node[teal dot] at (0.42, 0.45) {};
\node[teal dot] at (1.61, 1.61) {};
\node[teal dot] at (2.83, 1.55) {};
\node[teal dot] at (3.79, 2.38) {};

\node[coral tri] at (1.96, 0.37) {};
\node[coral tri] at (5.01, 2.68) {};
\node[coral tri] at (4.59, 1.45) {};
\node[coral tri] at (3.74, 0.46) {};
\node[coral tri] at (5.78, 1.43) {};
\node[coral tri] at (5.10, 0.41) {};


\node[blue dot] at (8.37, 2.31) {};
\node[ring A]   at (8.37, 2.31) {};
\node[pink dot] at (10.44, 2.67) {};
\node[ring A]   at (10.44, 2.67) {};
\node[blue dot] at (8.42, 0.45) {};
\node[ring B]   at (8.42, 0.45) {};
\node[blue dot] at (9.61, 1.61) {};
\node[ring B]   at (9.61, 1.61) {};
\node[blue dot] at (10.83, 1.55) {};
\node[pink dot] at (11.79, 2.38) {};

\node[blue tri]  at (9.96, 0.37) {};
\node[ring A]    at (9.96, 0.37) {};
\node[pink tri]  at (13.01, 2.68) {};
\node[ring B]    at (13.01, 2.68) {};
\node[pink tri]  at (12.59, 1.45) {};
\node[blue tri]  at (11.74, 0.46) {};
\node[pink tri]  at (13.78, 1.43) {};
\node[pink tri]  at (13.10, 0.41) {};

\end{tikzpicture}}
  \caption{The \textbf{contrastive} paradigm (left) treats temporal preference as a binary distinction, while the \textbf{parametric} paradigm (right) treats it as a dimensional concept by varying time horizon, reward magnitude, and delay.}
  \label{fig:query-diagram}
\end{figure}

We use two prompting settings that probe temporal preference at different levels of abstraction and one that queries temporal reasoning without valuation.
The \emph{minimally-framed} setting is purely contrastive: it presents a binary choice between a short-horizon and a long-horizon option, with no explicit time or reward values.
This captures temporal preference as a binary concept (present vs.\ future).
The \emph{highly-formatted} setting is both contrastive and parametric: it can elicit the same binary preference, but it also sweeps explicit time horizons from seconds to centuries, treating time as a continuous, dimensional concept.
Together, the two settings let us study temporal preference both as a categorical distinction and as a graded quantity (Figure~\ref{fig:query-diagram}).
Separately, the \emph{classification setting} drops the option contrast entirely: each prompt presents a single goal and asks the model to classify its horizon directly, probing temporal reasoning rather than preference, as a test of whether the subgraph generalizes across cognitive operations. 

\FloatBarrier


\subsection{Minimally-framed prompts}\label{app:minimally-framed}

The minimally-framed setting queries the model's temporal preference via forced-choice A/B prompts.
Each prompt presents a scenario and two candidate responses (one short-horizon and one long-horizon) and the model selects a label.
We construct two complementary datasets under this paradigm, probing the same underlying distinction under different cueing regimes.

Each pair is structured as \{\texttt{question}, \texttt{immediate\_choice}, \texttt{long\_term\_choice}, \texttt{category}\}, with A/B label positions randomized ($\approx$50/50 split) to control for positional bias.
Both datasets comprise 500 paired examples balanced across categories.
For probing experiments (\ref{app:contrastive-probing-linear-methods}), a 300-pair subset of the implicit dataset is used as the primary training and evaluation set.

\subsubsection{Explicit dataset ($D_{\text{explicit}}$)}

The explicit dataset contains 500 examples in which temporal scope is directly specified through overt temporal expressions.
The contrast between options is signaled by phrases denoting near-term deadlines, short planning horizons, annual targets, or multi-year timeframes; the distinction is lexically explicit.

\promptfigure%
{%
    Innovation should aim for:

    \qquad (A) Products we can ship this quarter

    \qquad (B) Technologies that matter in 20 years
}%
{%
    \qquad (A)
}%
{%
    Example of an \textbf{explicit} minimally-framed prompt.
    The temporal contrast is carried by surface vocabulary (``this quarter'' vs.\ ``20 years'').
}%
{%
    fig:minimally-framed-explicit
}%

The dataset is evenly distributed across 25 topical categories with 20 examples per category, spanning planning, decision-making, problem-solving, investment, prioritization, goal-setting, action, vision, risk management, resource allocation, strategy formation, change management, innovation, leadership, communication, evaluation, learning, adaptation, hiring, product development, customer relations, financial planning, team building, market entry, and crisis response.

\subsubsection{Implicit dataset ($D_{\text{implicit}}$)}

The implicit dataset contains 500 examples in which temporal scope is encoded through semantic framing rather than explicit temporal markers.
The short-horizon response emphasizes immediate containment, execution, or preservation, whereas the long-horizon response emphasizes redesign, investment, transformation, or compounding effects, without directly invoking time-related language.

\promptfigure%
{%
    Organizational focus should be on

    \qquad (A) Firefighting and troubleshooting

    \qquad (B) Culture-building and capability development
}%
{%
    \qquad (B)
}%
{%
    Example of an \textbf{implicit} minimally-framed prompt.
    Neither option contains temporal vocabulary; the distinction is carried entirely by semantic framing.
}%
{%
    fig:minimally-framed-implicit
}%

The dataset is balanced across 10 abstract contrast categories with 50 examples per category: \texttt{crisis\_vs\_foundation}, \texttt{harvest\_vs\_cultivate}, \texttt{execute\_vs\_design}, \texttt{react\_vs\_anticipate}, \texttt{preserve\_vs\_transform}, \texttt{tactical\_vs\_strategic}, \texttt{consume\_vs\_invest}, \texttt{fix\_vs\_build}, \texttt{survive\_vs\_thrive}, and \texttt{capture\_vs\_compound}.
This design isolates temporal reasoning from lexical cues, enabling evaluation of whether the model relies on semantic abstractions rather than explicit time indicators.

\subsubsection{LLM-Assisted Generation and Validation}

To construct the $D_{\text{explicit}}$ and $D_{\text{implicit}}$ datasets at scale while maintaining strict control over linguistic variables, we used a multi-stage, LLM-assisted generation and verification pipeline.
The initial candidate pairs for both datasets were generated using \texttt{Claude Sonnet 4.6}.
To ensure these generated pairs adhered to our contrastive constraints and were free of unintended confounds, we implemented an automated validation framework.
Each candidate pair was independently evaluated and scored by both \texttt{Claude Sonnet 4.6} and \texttt{Gemini 3 Flash}.

The models verified the pairs across four strict dimensions: \textit{lexical confounds} (ensuring the implicit set contained absolutely no explicit temporal keywords and that vocabulary complexity was balanced), \textit{surface form} (matching character length and grammatical structure), \textit{semantic confounds} (aligning formality, hedging, specificity, and sentiment), and \textit{content validity} (guaranteeing a unidimensional, unambiguous distinction between immediate and long-term choices).
Both models scored these factors on a 1 to 5 scale.
Pairs falling below the acceptable quality threshold (an average score $< 3.5$) were iteratively revised or discarded.
Finally, to eliminate positional bias, the presentation order of the immediate and long-term options (A/B) was randomized in the finalized datasets.


\subsection{Highly-formatted prompts}\label{app:highly-formatted}

The highly-formatted setting uses structured prompts with explicit section markers that ensure consistent token-position alignment across prompt variants.
Each prompt contains the following fields:

\begin{itemize}[nosep, leftmargin=*]
  \item \texttt{SITUATION:} Domain context (e.g., household financial planning)
  \item \texttt{TASK:} The role, task, and two labeled options with reward amounts and time horizons
  \item \texttt{OBJECTIVE:} Instruction to deliberate
  \item \texttt{CONSTRAINT:} An explicit time-horizon constraint (e.g., ``1 year,'' ``5 centuries,'' or omitted for no-constraint prompts)
  \item \texttt{ACTION:} Instruction to select one option
  \item \texttt{FORMAT:} Response template specifying ``I choose: \texttt{<label>}'' and ``My reasoning: \texttt{<text>}''
\end{itemize}

\noindent The shared structural markers serve as \emph{anchors} for a position mapping that aligns token indices between prompts of different lengths.
When clean and corrupted prompts differ in token count (e.g., because different time horizons require different numbers of tokens), the position mapping uses these anchors to interpolate correctly during activation patching (\ref{app:causal-parametric-methods}).

\promptfigure%
{%
        \textbf{SITUATION:} Plan for the future of the household based on the stated objectives and constraints.

        \textbf{TASK:} You, the head of the household, are tasked to choose the best investment:

        \begin{enumerate}[label=\alph*), leftmargin=2em, itemsep=4pt, topsep=2pt]
          \item 20,000 dollars in 6 months.
          \item 500,000 dollars in 10 years.
        \end{enumerate}

        \textbf{OBJECTIVE:} Think deeply about which option is preferable.

        \textcolor{red}{\textbf{CONSTRAINT:} You must select the option that provides the greatest benefit \\ for this time horizon: \textit{8 months}.}

        \textbf{ACTION:} Select one of the two options. Provide reasoning on why this choice was made.

        \textbf{FORMAT:} Respond in this format:

        \quad\quad I choose: \texttt{<a) or b)>}. My reasoning: \texttt{<reasoning in 1-3 sentences>}
}%
{%
    \quad I choose: \texttt{b)}. My reasoning: Although the immediate...
}%
{%
    Example of a highly-formatted prompt.
    The \textcolor{red}{constraint} section is optional; omitting it queries the model's latent preference.
}%
{%
    fig:highly-formatted-prompt
}%

\subsubsection{Parametric variation}

The experiment configuration sweeps over several axes to systematically vary the temporal context:
\begin{itemize}[nosep, leftmargin=*]
  \item \textbf{Reward range}: Logarithmic steps between a minimum and maximum (e.g., \$1{,}000--\$100{,}000)
  \item \textbf{Time range}: Logarithmic steps for both short-term and long-term options
  \item \textbf{Time horizons}: 17 values from \texttt{null} (no constraint) through seconds, hours, days, weeks, months, years, decades, to centuries
\end{itemize}

\noindent This yields a grid of contrastive pairs that disentangle the effects of reward magnitude, delay, and horizon constraint on internal representations.

\begin{figure}[htbp]
    \centering
    \begin{adjustbox}{scale=0.75}
    \begin{lstlisting}[language={}, basicstyle=\ttfamily\small, breaklines=true, frame=single, columns=fullflexible]
    {
        "name": "investment_geometry",
        "context": {
            "reward_unit": "dollars",
            "role": "the head of the household",
            "situation": "Plan for the future of the households.",
            "task_in_question": "choose the best investment",
            "domain": "finance"
        },
        "options": {
            "short_term": {
                "reward_range": [1000, 100000],
                "time_range": [
                {"value": 1, "unit": "days"},
                {"value": 20, "unit": "years"}
                ],
                "reward_steps": [2, "logarithmic"],
                "time_steps": [5, "logarithmic"]
            },
            "long_term": {
                "reward_range": [1000, 100000],
                "time_range": [
                    {"value": 1, "unit": "years"},
                    {"value": 100, "unit": "years"}
                ],
                "reward_steps": [2, "logarithmic"],
                "time_steps": [5, "logarithmic"]
            }
        },
        "time_horizons": [
            null,
            {"value": 1, "unit": "seconds"},
            {"value": 1, "unit": "hours"},
            {"value": 1, "unit": "days"},
            {"value": 1, "unit": "week"},
            {"value": 1, "unit": "months"},
            {"value": 2, "unit": "months"},
            {"value": 6, "unit": "months"},
            {"value": 1, "unit": "years"},
            {"value": 3, "unit": "years"},
            {"value": 5, "unit": "years"},
            {"value": 1, "unit": "decades"},
            {"value": 3, "unit": "decades"},
            {"value": 5, "unit": "decades"},
            {"value": 1, "unit": "centuries"},
            {"value": 2, "unit": "centuries"},
            {"value": 5, "unit": "centuries"}
        ]
    }
    \end{lstlisting}
    \end{adjustbox}
    \caption{Example configuration for the \texttt{investment\_geometry} scenario.
    Each scenario defines a context, short- and long-term option ranges, and a set of time horizons spanning seconds to centuries.}
    \label{fig:parametric-config}
\end{figure}

\noindent The configured \texttt{short\_term} and \texttt{long\_term} time ranges overlap (1 day--20 years and 1 year--100 years).
We did not enforce a delay-ordering filter at sample time, so a small fraction of generated pairs may have realized short-term delay $\ge$ realized long-term delay.
We did not observe systematic effects from this in our analyses, but it is a known dataset caveat.

\FloatBarrier

\subsection{Classification-oriented prompts}\label{app:classification-oriented}

The classification setting queries the model's temporal reasoning rather than its preference: each prompt presents a single goal whose horizon must be inferred from world knowledge, and the model produces a direct short/long judgment.
Unlike the minimally-framed setting (\ref{app:minimally-framed}), there is no A/B option contrast within a single prompt; instead, contrastive structure is established \emph{across} prompt pairs in the IOI style.
Each clean sample is paired with a corrupted sample that shares the same template and differs only in the embedded goal, with the two goals lying on the same life continuum and differing only in temporal horizon.

\newlength\myboxwidth
\setlength{\myboxwidth}{\dimexpr\textwidth-2\fboxsep}

\textbf{Prompt template}\\\\
\fbox{\parbox{\myboxwidth}{"The goal is to <goal>. Is this a <short-term or long-term / long-term or short-term> goal? The answer is:"}}\\

\promptpairfigure%
{%
    The goal is to cook a warm dinner for the family. Is this a short-term or long-term goal? The answer is:
}%
{%
    short
}%
{%
    The goal is to become a top chef in the city. Is this a short-term or long-term goal? The answer is:
}%
{%
    long
}%
{%
    Example of a classification pair of prompts.
}%
{%
    fig:classification-oriented-pair
}%

All prompts are appended with the chat template before being passed to the model.

\subsubsection{Dataset composition}

The dataset contains 160 prompt pairs with perfectly balanced question order: 80 SL pairs (\textit{"is this a short-term or long-term"}) and 80 LS pairs (\textit{"is this a long-term or short-term"}), included to mitigate priming effects.
Three temporal cue types signal the long-term horizon: (1) \textit{career/mastery}, achieving elite status or deep expertise at something; (2) \textit{growth}, transforming something small into something large or established; and (3) \textit{accumulation}, exhaustive scope requiring years of sustained effort.
All goals relate to a general life domain and are distributed fairly evenly across 25 life subdomains (gardening, cooking, swimming, languages, board games, etc.).

\subsubsection{Design principles}

The dataset is governed by four design principles.
First, \emph{token alignment}: all 320 prompts have the same token length under \texttt{Qwen3-4B-Instruct-2507}, ensuring positional correspondence across pairs.
In total, each prompt contains 34 tokens, including the chat template, with 7 tokens covering the goal statement.
Second, \emph{semantic overlap within pairs}: each clean and corrupted goal shares the same domain and lies on the same life continuum, differing only in temporal horizon.
Third, \emph{no explicit temporal keywords}: words such as ``daily,'' ``weekly,'' or ``years'' are banned; the model must infer the horizon from world knowledge alone.
Fourth, \emph{unambiguous horizons}: every short-term goal can be completed in hours or a single sitting, while every long-term goal requires years of sustained effort.

\subsubsection{LLM-assisted generation and human verification}

The classification dataset was constructed iteratively with \texttt{Claude Opus 4.6} in successive batches of 20 candidate pairs: each batch was generated by the model under the constraints described above, manually verified against those same constraints, and refined through continuous human feedback before the next batch. 
This loop was repeated until the dataset reached its target size and was frozen.

\subsubsection{Validation and filtering}

The original dataset contained 200 pairs.
\texttt{Qwen3-4B-Instruct-2507} successfully classified 160/200 pairs, eliciting 80\% accuracy.
Of the 40 misclassified pairs, 25 involve genuinely ambiguous temporal horizons where the model's interpretation is defensible.
The remaining 15 failures, where the model labels the prompt as short-term despite clear temporal signals, concentrate in accumulation-type scholarly activities (\textit{"catalog moon lore from old hill folk"}) and growth-type production at scale (\textit{"fire glazed plates for the whole county"}).
All subsequent activation patching in (\ref{app:causal-contrastive-methods}) operates on the 160 successfully classified pairs.

Table~\ref{tab:cue_types_stat_160} reports cue-type statistics for the 160 surviving pairs; Table~\ref{tab:sl_ls_stat_160} reports the same statistics grouped by question order.

Because this dataset serves a narrower purpose — testing whether the temporal-preference subgraph generalizes to categorical horizon inference, rather than characterizing preference itself — we do not include it in the prompt-setting comparison in Section ~\ref{app:prompts-comparison}, which contrasts only the two preference-oriented paradigms.
\FloatBarrier

\subsection{Comparison of prompting settings}\label{app:prompts-comparison}

Table~\ref{tab:prompting-comparison} summarizes the complementary strengths of the two settings.
The minimally-framed setting is better suited for probing latent preferences under naturalistic conditions, while the highly-formatted setting enables controlled parametric sweeps and richer mechanistic analysis.

\begin{table}[htbp]
\centering
\small
\renewcommand{\arraystretch}{1.3}
\begin{tabularx}{\linewidth}{@{} >{\bfseries}l X X @{}}
\toprule
& \textbf{Minimally-framed} & \textbf{Highly-formatted} \\
\midrule
Paradigm
  & Contrastive only (binary: short vs.\ long)
  & Contrastive + parametric (binary preference \emph{and} continuous time horizon) \\
Concept type
  & Binary (present vs.\ future)
  & Dimensional (seconds to centuries) \\
Prompt structure
  & No explicit time or reward; model infers temporality from semantic framing
  & Explicit reward amounts, delays, and a constraint field specifying the horizon \\
Validity
  & Closer to on-policy; low demand characteristics
  & Closer to off-policy; structured scaffolding may anchor the model \\
Mechanistic use
  & Attribution, probing, CAA vector construction
  & Attribution, activation patching with token-level position mapping, geometry analysis \\
Behavioral modeling
  & Binary preference only; applies to any domain
  & Discount curves, comparison to human baselines, reparametrization via activation geometry \\
\bottomrule
\end{tabularx}
\caption{Comparison of the two prompting settings.
The minimally-framed setting probes latent binary preference; the highly-formatted setting adds parametric control over the time dimension.}
\label{tab:prompting-comparison}
\end{table}

\FloatBarrier
\clearpage
\clearappnumbering

\section{Extended limitations and future work}\label{app:extended-limitations}

Time is a complex and entangled concept.
Our work is merely a starting point.

\begin{itemize}[noitemsep, topsep=0pt, leftmargin=*]

    \item \textbf{Finer localization.}
    Full circuit tracing~\citep{zhang2026locatesteerimprovepractical, goldowskydill2023localizing} would identify atomic components and their information flow.
    Our EAP-IG analysis shows attribution mass distributed across many nodes, and whether this reflects genuine distribution or a methodological limitation remains open.

    \item \textbf{Domain generalization and dataset provenance.}
    Our approach has several limitations: the pipeline uses only financial scenarios, so findings may not generalize to other domains (health, career); contrastive labels were synthetically assigned and lack human validation; the steering vector, derived from controlled off-policy settings, may capture correlated features rather than purely temporal preference; and the classification dataset entangles temporal horizon with achievement vocabulary.

    \item \textbf{Scaling across models and variants.}
    We study only \texttt{Qwen3-4B-Instruct-2507}; replicating across families and scales would test whether the subgraph location and the probing--steering dissociation generalize.
    Comparing this distilled, non-thinking variant against its thinking counterpart (\texttt{Qwen3-4B}) is particularly compelling: our behavioral analysis shows that chain-of-thought dramatically alters temporal preference, but whether reasoning reorganizes the underlying subgraph is unexplored.

    \item \textbf{Richer parameterization and concept interactions.}
    Our parametric querying maps time horizon but could extend to reward magnitude, risk, role, and domain to parameterize the full intertemporal choice space.
    Temporal preference likely interacts with representations of risk~\citep{zhu2025steering, moghimi2026decouplingtimeriskrisksensitive}, emotion~\citep{anthropic-emotions}, and urgency, but we treat the subgraph in isolation.
    Moreover, all experiments are single-turn, yet temporal preference matters most in multi-turn and agentic settings where representations may shift across turns~\citep{lampinen2026linearrepresentationslanguagemodels}.

    \item \textbf{Non-linear steering.}
    Our linear CAA vector approximates a curved manifold; output quality degrades at $|\alpha|{=}60$.
    Methods that follow the manifold's curvature~\citep{curveball2026, li2026svf, postmus2025conceptors} could enable stronger, cleaner interventions.

\end{itemize}

\clearpage
\clearappnumbering

\partpagecontent{Part 1:}{Where is temporal preference?}{%
\begin{itemize}[leftmargin=*, itemsep=0.8em]
  \item \textbf{\hyperref[app:contrastive-probing-linear]{G}.} Linear probing
  \item \textbf{\hyperref[app:attributional-contrastive]{H}.} Attributional contrastive
  \item \textbf{\hyperref[app:attributional-parametric]{I}.} Attributional parametric
  \item \textbf{\hyperref[app:causal-parametric]{J}.} Causal parametric
  \item \textbf{\hyperref[app:causal-contrastive]{K}.} Causal classification
  \item \textbf{\hyperref[app:convergence]{L}.} Cross-method convergence
\end{itemize}%
}

\section{Contrastive linear probing results}\label{app:contrastive-probing-linear}

The four experiments above localized temporal preference through attribution and causal intervention.
Here we take a complementary approach: training logistic regression probes on residual-stream activations to ask \emph{where} the model linearly encodes the short/long distinction (methodology in \ref{app:contrastive-probing-linear-methods}).
Unlike the previous methods, probing does not measure causal effect but rather the readability of a concept at each layer.
This distinction will prove important: the best probing layer turns out to differ from the best intervention layer.

\subsection{Layer-by-Layer Probe Accuracy}

Logistic regression probes were trained at each of the 36 layers using the protocol
described in \ref{app:contrastive-probing-linear-methods}.

\begin{table}[htbp]
\centering
\small
\begin{tabular}{lc}
\toprule
\textbf{Metric} & \textbf{Result} \\
\midrule
Best layer & 26 \\
Best test accuracy & 99.2\% \\
Signal above chance & +52.3 pp \\
Cross-dataset generalization & Yes (see Section~\ref{app:contrastive-probing-linear:cross_dataset}) \\
\bottomrule
\end{tabular}
\caption{Summary of probing results on $D_{\mathrm{implicit}}$.}
\label{tab:probing_summary}
\end{table}

\noindent Accuracy rises steadily from $\sim$80\% at layer~0 to a plateau above 95\%
around layer~17, reaching 99.2\% at layer~26.
The monotonic increase across layers is
consistent with the model progressively refining a linear temporal representation in
deeper layers.

\begin{figure}[htbp]
\centering
\includegraphics[width=0.75\textwidth]{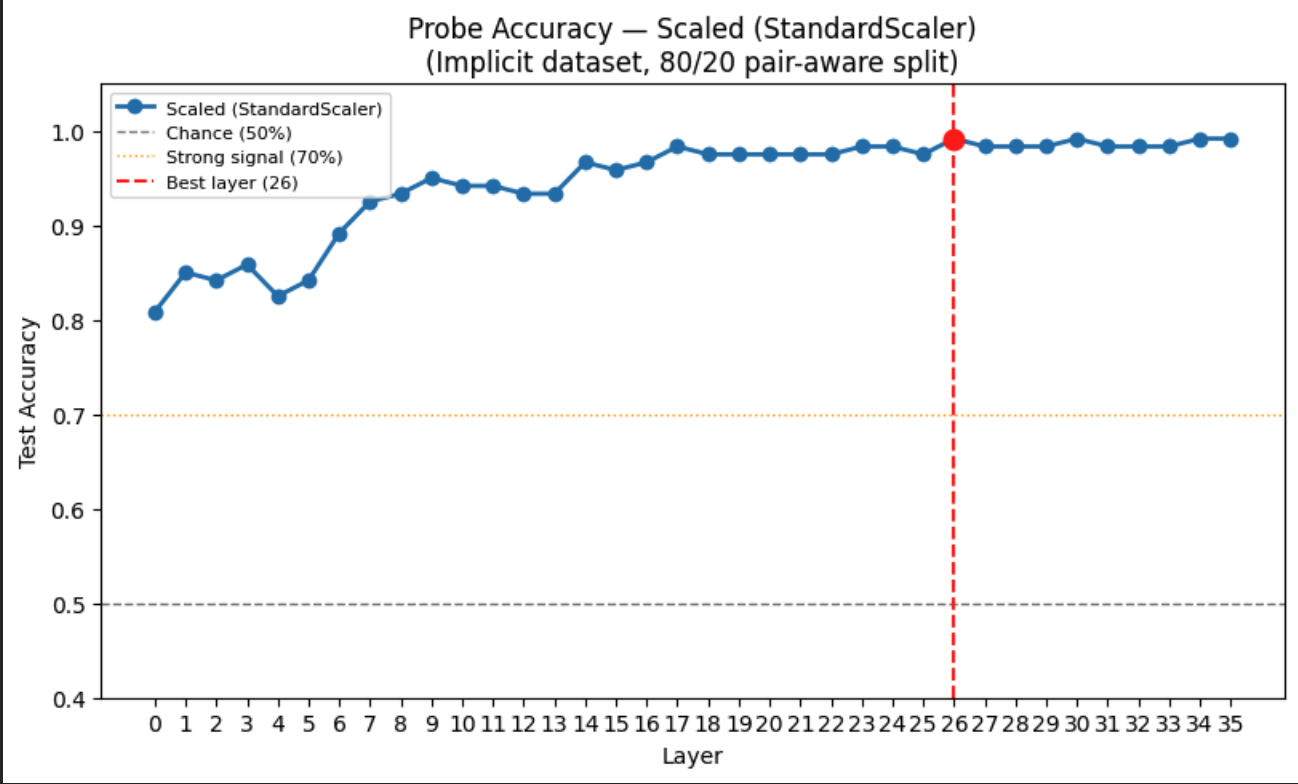}
\caption{Test accuracy of scaled logistic regression probes across all 36 layers
on $D_{\mathrm{implicit}}$ (80/20 pair-aware split).
Dashed lines indicate chance
(50\%) and the strong-signal threshold (70\%).
The best layer (26) is marked.}
\label{fig:probe_accuracy}
\end{figure}

\FloatBarrier

\subsection{Shuffled-Label Control}
\label{app:contrastive-probing-linear:shuffle}

To confirm that probe accuracy reflects genuine temporal structure rather than
geometric artifacts of the activation space, we train 10 probes per layer on randomly
permuted labels using the same scaled activations.
Shuffled accuracy was approximately
50\% at every layer, including layer~0.
The gap between real-label accuracy
($\sim$80--99\%) and shuffled-label accuracy ($\sim$50\%) at every layer confirms that
the signal is a learned property of the temporal concept, not an intrinsic property
of the activation geometry.

\begin{figure}[htbp]
\centering
\includegraphics[width=0.75\textwidth]{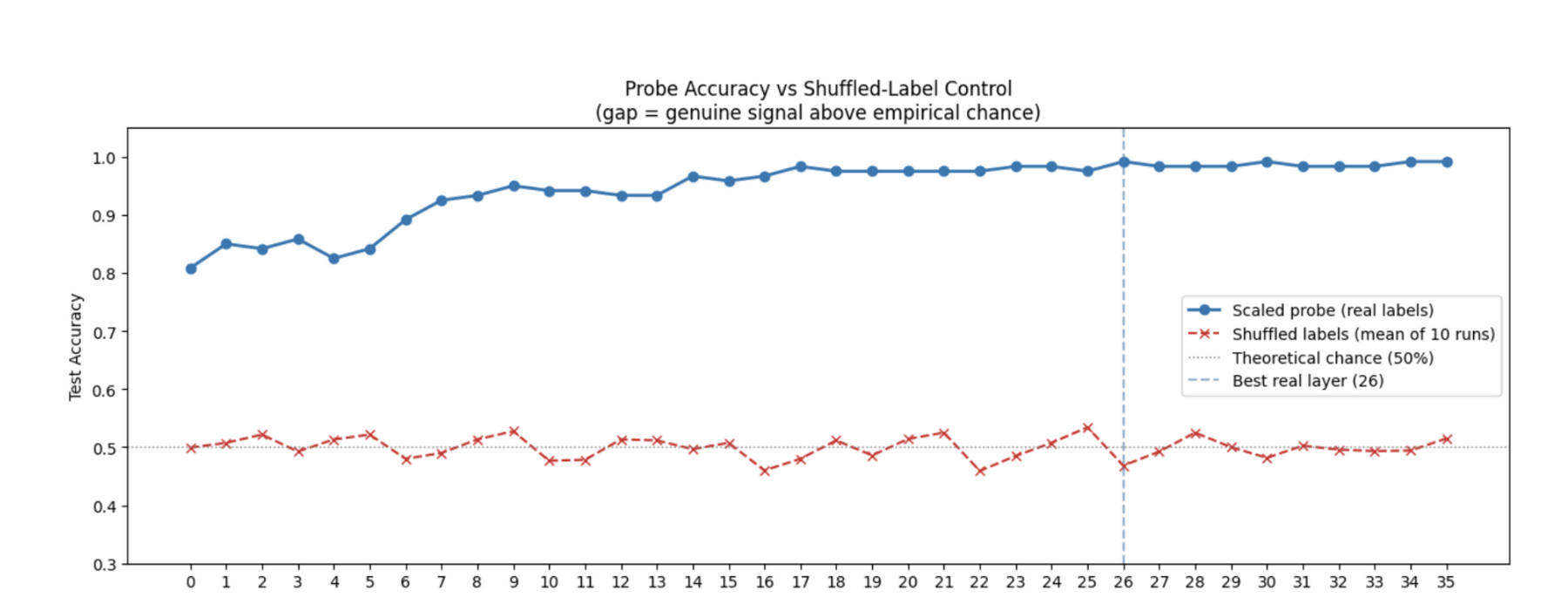}
\caption{Probe accuracy vs.\ shuffled-label control across all 36 layers.
Real-label
probes (blue) rise to 99.2\% at layer~26, while shuffled-label probes (orange)
remain at chance ($\sim$50\%) throughout.}
\label{fig:shuffled_label_control}
\end{figure}

\FloatBarrier

\subsection{Representation Geometry (PCA)}
\label{app:contrastive-probing-linear:pca}

PCA analysis reveals an important asymmetry between the two datasets:

\begin{itemize}[leftmargin=1.5em]
    \item \textbf{Implicit dataset:} No separation is visible in the top two principal
    components (PC1 explains only 2--5\% of variance).
    The temporal direction is subtle
    and occupies dimensions that PCA discards.
    \item \textbf{Explicit dataset:} Clear separation is visible in PCA
    (PC1 $=$ 9.5\% at layer~0), driven by surface vocabulary differences between
    short-term and long-term choices.
\end{itemize}

\noindent This result is significant: PCA fails to detect the temporal concept in the
implicit dataset, yet the supervised probe succeeds at 99.2\%.
The temporal direction
is real but non-obvious; it requires supervised search to find a direction that
unsupervised methods miss.
This is consistent with findings on the non-trivial geometry
of concept representations in LLMs~\citep{engels2025notall, marks2024geometry}.

\begin{figure}[htbp]
\centering
\includegraphics[width=\textwidth]{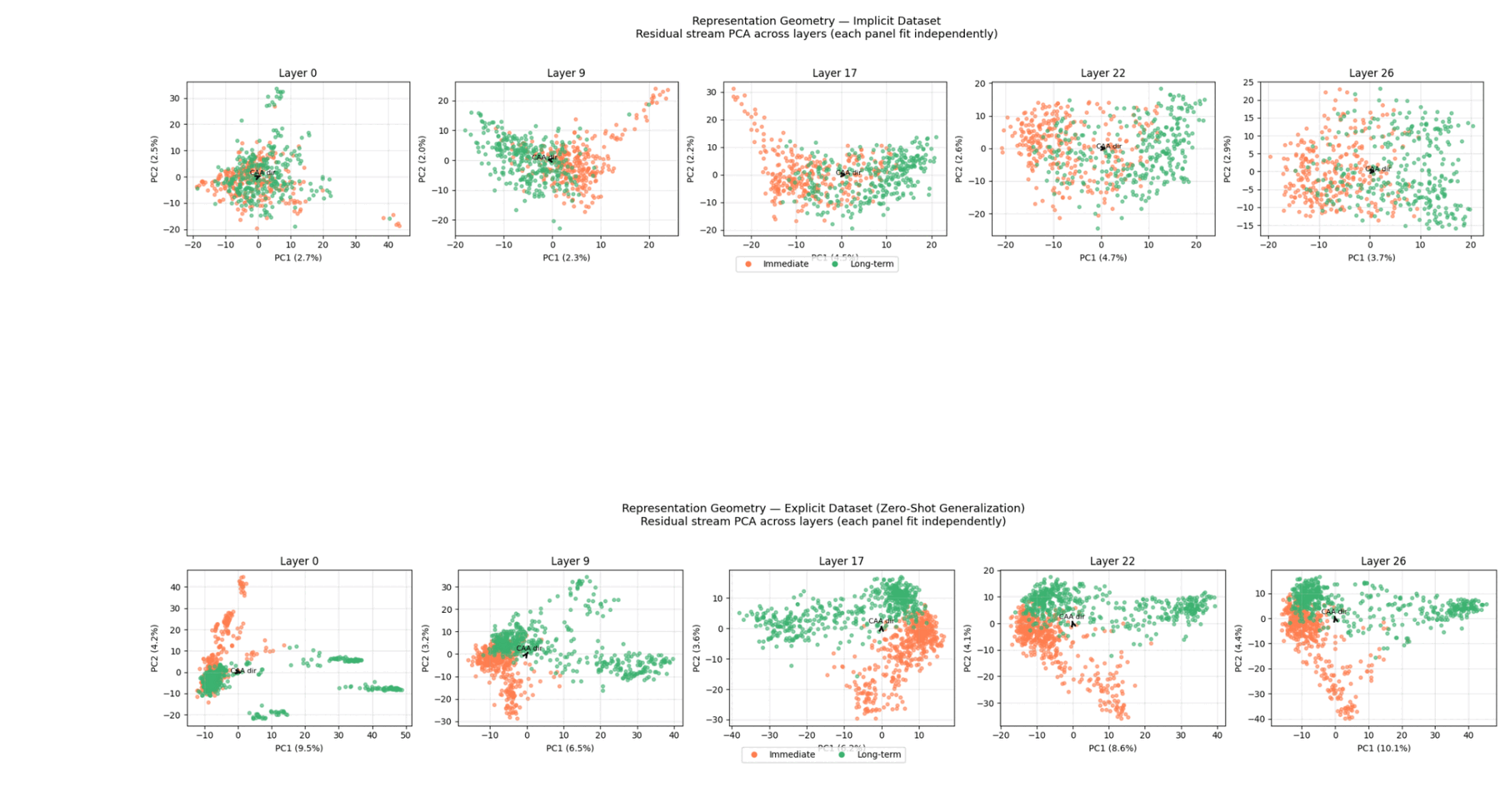}
\caption{PCA projections of layer activations for the implicit (top) and explicit
(bottom) datasets at selected layers.
The implicit dataset shows no visible separation
in the top two PCs, while the explicit dataset shows clear clustering driven by
surface vocabulary.}
\label{fig:pca_implicit_explicit}
\end{figure}

\FloatBarrier

\subsection{Cross-Dataset Generalization}
\label{app:contrastive-probing-linear:cross_dataset}

Probes trained on $D_{\mathrm{implicit}}$ were evaluated zero-shot on
$D_{\mathrm{explicit}}$ (different vocabulary, same underlying concept).
This tests
whether the probe has learned a genuine temporal direction rather than
vocabulary-specific features.
The saved \texttt{StandardScaler} from training is
re-applied to the explicit activations before scoring.

Cross-dataset accuracy tracks the within-dataset accuracy closely across all layers,
confirming that the probe direction generalizes from implicit semantic cues to explicit
temporal markers.
At the best layer (26), implicit test accuracy is 99.2\% and
cross-dataset accuracy on $D_{\mathrm{explicit}}$ remains above 95\%.

\begin{figure}[htbp]
\centering
\includegraphics[width=0.75\textwidth]{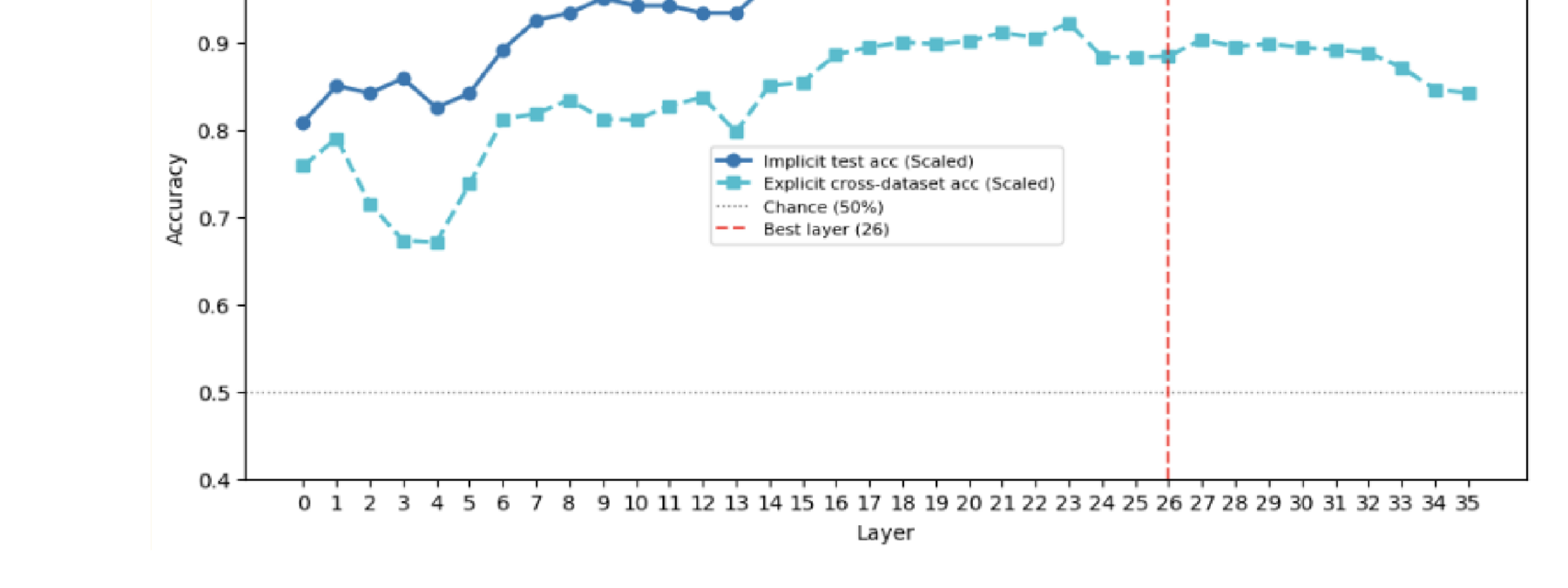}
\caption{Cross-dataset generalization: probes trained on $D_{\mathrm{implicit}}$
evaluated zero-shot on $D_{\mathrm{explicit}}$.
Accuracy tracks closely across all
layers, confirming the probe captures a genuine temporal direction rather than
dataset-specific features.}
\label{fig:cross_dataset_generalization}
\end{figure}

\FloatBarrier

\subsection{Summary}

Contrastive probing confirms that \texttt{Qwen3-4B-Instruct-2507} maintains a linear temporal
direction in its residual stream.
The key findings are:

\begin{enumerate}[leftmargin=1.5em]
    \item \textbf{Strong linear signal.}
    A logistic regression probe achieves 99.2\%
    test accuracy at layer~26, with accuracy rising monotonically across layers.
    \item \textbf{Shuffled control rules out artifacts.}
    Probes trained on permuted
    labels remain at chance ($\sim$50\%) at every layer, confirming that the signal
    reflects genuine temporal structure (Section~\ref{app:contrastive-probing-linear:shuffle}).
    \item \textbf{PCA misses it; supervised search finds it.}
    The temporal direction
    is not visible in the top principal components of the implicit dataset, yet a
    supervised probe recovers it with near-perfect accuracy
    (Section~\ref{app:contrastive-probing-linear:pca}).
    The concept is real but geometrically subtle.
    \item \textbf{Cross-dataset generalization.}
    Probes trained on implicit cues
    transfer zero-shot to explicit temporal markers, confirming that the learned
    direction captures a genuine temporal concept rather than vocabulary-specific
    features (Section~\ref{app:contrastive-probing-linear:cross_dataset}).
\end{enumerate}

\noindent An important dissociation emerges when comparing these probing results with
the steering experiments in \ref{app:contrastive-steering}: probing accuracy peaks at
layer~26, while steering is most effective at layers~19--22.
This gap suggests that the
layers where the model most cleanly \emph{represents} temporal preference are not the
same layers where \emph{intervening} on that representation most strongly influences
downstream behavior.
We discuss possible explanations for this probing--steering
dissociation in Section~\ref{app:contrastive-steering:discussion}.

\clearpage
\clearappnumbering

\section{Attributional contrastive results}\label{app:attributional-contrastive}

Our first approach to localizing temporal preference uses gradient-based attribution (EAP-IG) on the minimally-framed contrastive prompts, where the model chooses between a short-horizon and a long-horizon option with no explicit time vocabulary.
This is the cheapest localization method: it approximates causal effect via gradients rather than direct intervention, and the contrastive prompts are short and semantically controlled.
The tradeoff is that the signal may be noisier than causal methods, so we treat the results here as a selection prior rather than ground truth (methodology in \ref{app:attributional-contrastive-methods}).

The attribution reveals a candidate subgraph comprising approximately 0.125\% of all nodes, concentrated in layers 21--35.
However, as we show below, the circuit is highly diffuse: the median per-component attribution is well below 0.1\% of the total mass, and even the highest-scoring outlier (Figure~\ref{fig:attribution_score_distribution}) sits below $\sim$1\%.
The layers that emerge here (particularly L24 for attention) will reappear consistently across the causal and probing experiments that follow (\ref{app:causal-parametric}, \ref{app:causal-contrastive}, \ref{app:contrastive-probing-linear}).

\subsection{Attribution Score Distribution}

\begin{figure}[htbp]
    \centering
    \includegraphics[width=0.8\textwidth]{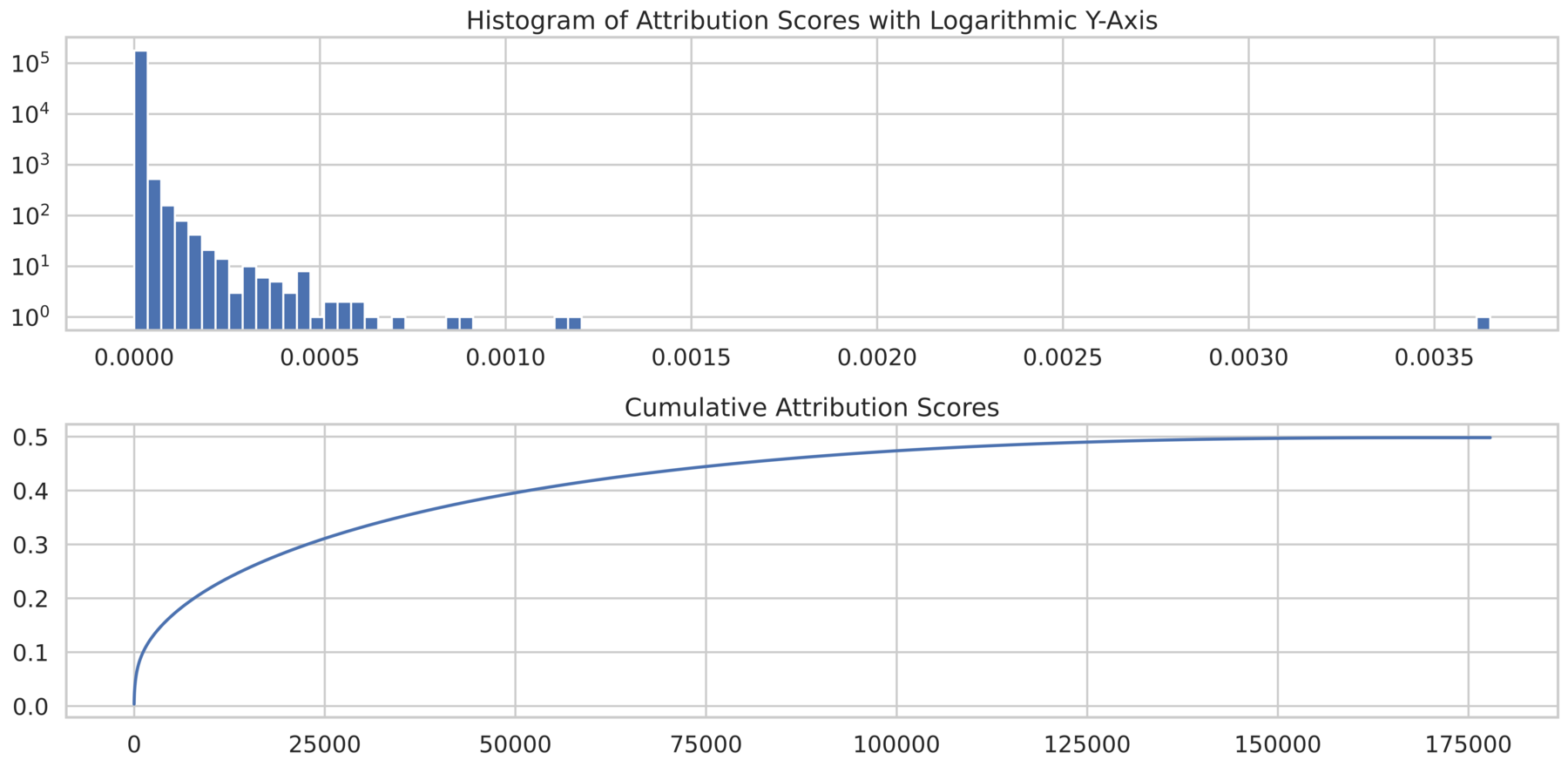}
    \caption{
        (top) Histogram of attribution scores with logarithmic y-axis.
        The distribution is \textbf{not} approximated by either a power law or an exponential.
        (bottom) Cumulative logit-normalized component attribution scores for variant \textbf{(A)} for the canonical option order with respect to the short-term concept.
    }
    \label{fig:attribution_score_distribution}
\end{figure}

Figure~\ref{fig:attribution_score_distribution} shows that attribution mass is distributed across a large number of nodes: the distribution is neither power-law nor exponential, and the vast majority of components have near-zero attribution scores.
This poses a fundamental challenge for top-$k$ selection, as there is little theoretical justification for any particular cutoff when the score distribution lacks a natural elbow or gap.

\FloatBarrier

\subsection{Limitations of EAP-IG for This Circuit}

The attribution results reveal that temporal preference is likely mediated by a highly \emph{diffuse} circuit rather than a sparse, localizable one.
Even the highest-scoring individual components explain at most a fraction of a percent of the total attribution mass individually, and the bulk of components sit far below 0.1\%.
The sheer number of low-scoring components dominates the distribution, making top-$k$ selection inherently noisy: it is unclear whether selected nodes are genuinely temporal-preference components or statistical artifacts of aggregation over hundreds of prompt variations.

Despite these limitations, the EAP-IG results provide a useful \emph{signal} when interpreted alongside independent methods.
In particular, the layers that emerge as high-attribution under EAP-IG (e.g., L24 for attention) overlap with layers identified by activation patching (\ref{app:causal-parametric}, \ref{app:causal-contrastive}) and CAA steering (\ref{app:contrastive-steering}).
This convergence across independent methodologies suggests that the layer-level localization is genuine even though component-level identification via EAP-IG alone is unreliable.

We therefore treat EAP-IG not as a circuit-identification tool in the traditional sense, but as a \emph{selection prior}: it restricts the search space to nodes enriched for temporal signal, which we then characterize through representational geometry (\ref{app:parametric-geometry}) and probing (\ref{app:contrastive-probing-linear}).
Our analysis focuses on representational structure rather than on isolating a minimal causal mechanism, and uses activation patching (\ref{app:causal-parametric}, \ref{app:causal-contrastive}) to establish causal claims independently.

\subsection{Layer Distribution of Top-k Components}

Despite the diffuse nature of the overall attribution distribution, a clear layer-level pattern emerges: temporal-preference signal concentrates in the mid-to-upper layers (approximately layers 21--35).
This concentration is robust across different values of $k$ and holds for both attention and MLP components.
Critically, this layer range converges with the layers identified independently by parametric activation patching (\ref{app:causal-parametric}) and CAA steering (\ref{app:contrastive-steering}), providing cross-method validation that temporal preference processing is genuinely localized to this subgraph.

\begin{figure}[htbp]
    \centering
    \includegraphics[width=0.8\textwidth]{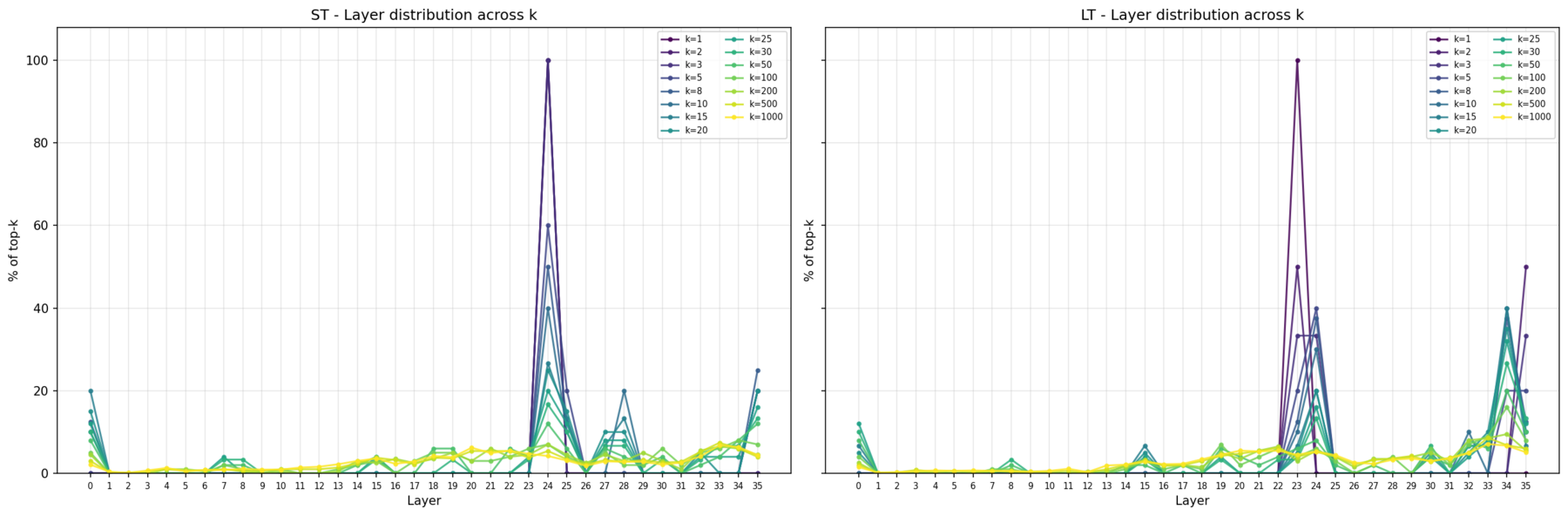}
    \caption{
        Layer-wise distribution of top-$k$ attributed components.
        Attribution mass concentrates in layers 21--35, with attention heads peaking around L24 and MLP neurons concentrated in the upper layers (L31--L35).
        This layer profile is stable across values of $k$, indicating genuine localization rather than an artifact of the threshold.
    }
    \label{fig:line_layer_distribution}
\end{figure}

\begin{figure}[htbp]
    \centering
    \includegraphics[width=\textwidth]{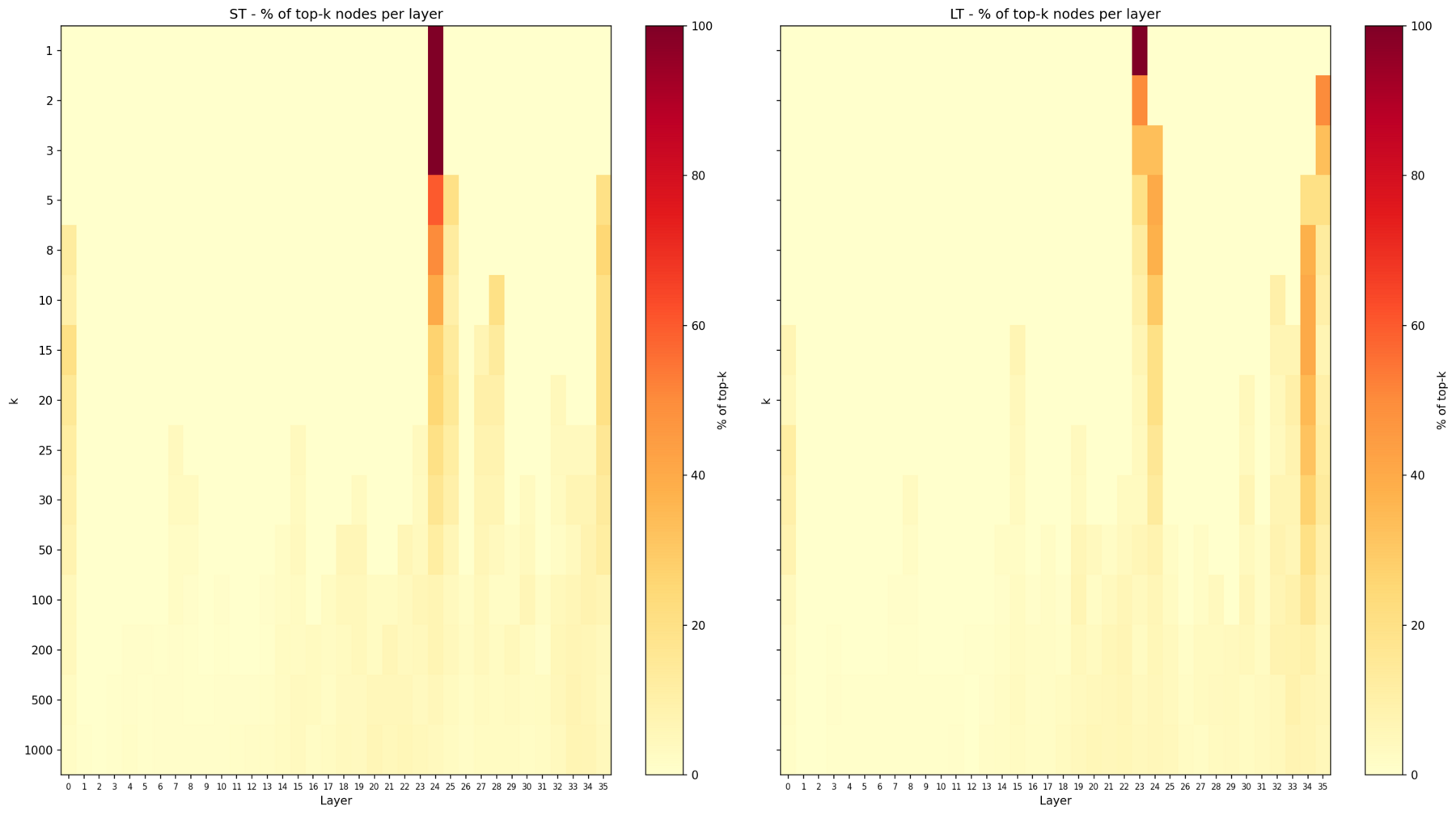}
    \caption{
        Heatmap of top-$k$ component counts per layer, broken down by component type.
        Attention heads are enriched in layers 21--26, while MLP neurons dominate in layers 31--35, consistent with a two-stage pattern where attention layers carry temporal information and later MLP layers refine it.
    }
    \label{fig:heatmap_topk_per_layer}
\end{figure}

\begin{figure}[htbp]
    \centering
    \includegraphics[width=0.8\textwidth]{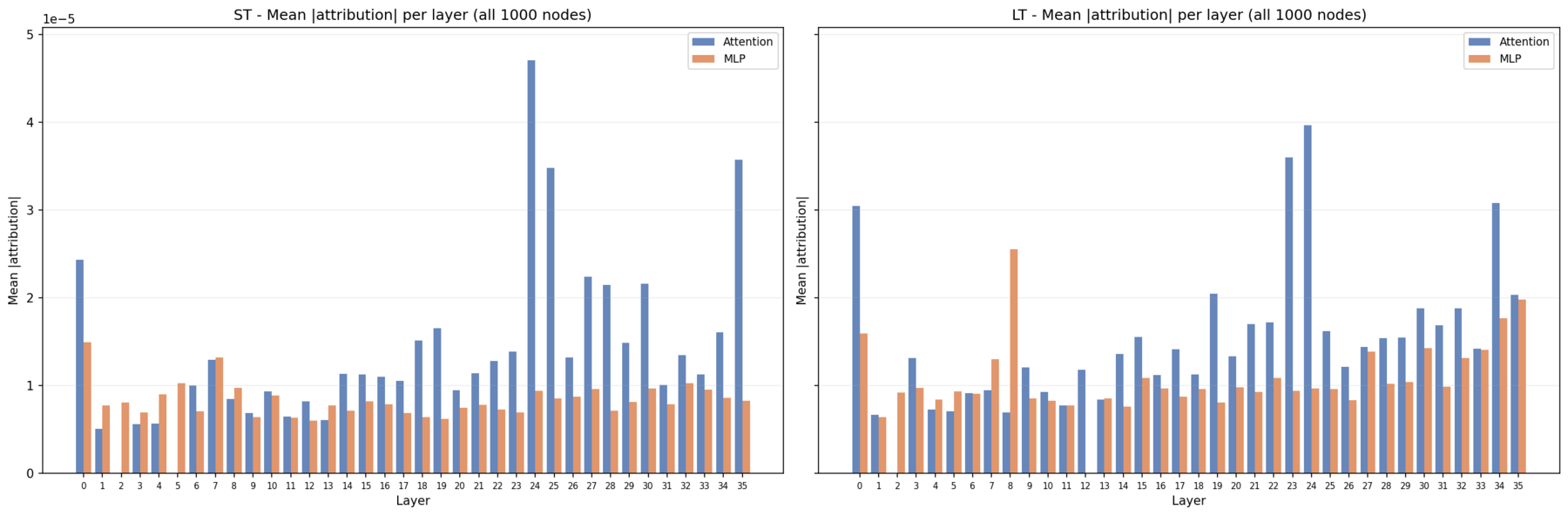}
    \caption{
        Mean attribution score by layer.
        Layers 21--35 show elevated per-component scores, confirming that the mid-to-upper layer concentration reflects genuinely higher attribution rather than a selection artifact.
        The peak around L24 for attention aligns with activation patching results identifying L24\_attn as the highest-effect attention component (\ref{app:causal-parametric}, \ref{app:causal-contrastive}).
    }
    \label{fig:mean_attribution_by_layer}
\end{figure}

Figure~\ref{fig:mean_attribution_by_layer} complements the top-$k$ count analysis by showing mean attribution scores per layer.
The mean score profile confirms that the layers 21--35 concentration is not merely a consequence of having more components selected; these layers also carry higher per-component attribution mass on average.

\FloatBarrier

\subsection{Attention vs.\ MLP Contributions}

Decomposing the top-$k$ attributed components by type reveals a division of labor between attention heads and MLP neurons.
Attention heads account for the majority of highly-attributed components, consistent with their role in routing information across token positions, while MLP neurons contribute a smaller but distinct share concentrated in the upper layers.
This attention-dominated pattern is consistent with temporal preference relying on contextual integration across the prompt rather than on purely local feature computation.

\begin{figure}[!htbp]
    \centering
    \begin{minipage}[t]{0.55\textwidth}
        \centering
        \includegraphics[width=\textwidth]{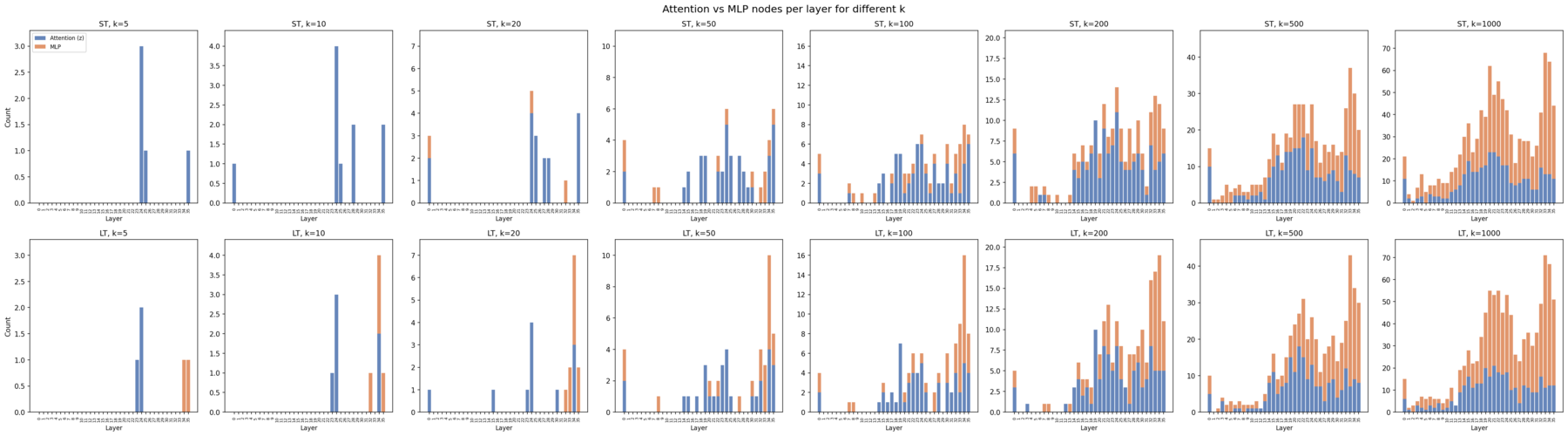}
    \end{minipage}\hfill
    \begin{minipage}[t]{0.40\textwidth}
        \centering
        \includegraphics[width=\textwidth]{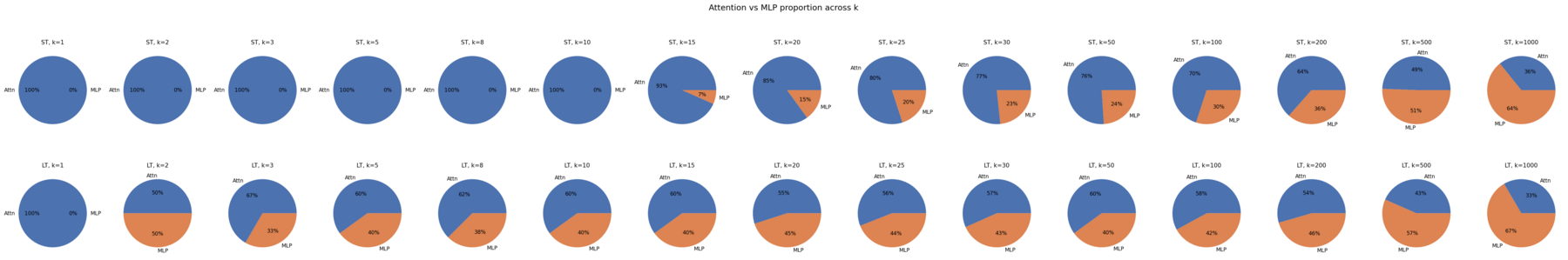}
    \end{minipage}
    \caption{Left: attention heads dominate the top-$k$ attributed components, but the MLP share grows at larger $k$.
    Right: overall attribution mass split.
    Attention carries the majority, reinforcing that temporal-preference computation is primarily mediated by cross-position information flow.}
    \label{fig:stacked_bar_attn_vs_mlp}
\end{figure}

The heatmaps below reveal which specific heads and MLP neurons carry the signal.
For attention, a small number of heads in layers 21--26 stand out, while the MLP signal is more diffuse across neurons in layers 31--35.
This spatial separation (attention in mid-layers, MLP in upper layers) is consistent with a two-phase computation: attention heads first integrate temporal context, then MLP layers transform this into the output representation.

\begin{figure}[htbp]
    \centering
    \begin{minipage}[t]{0.48\textwidth}
        \centering
        \includegraphics[width=\linewidth]{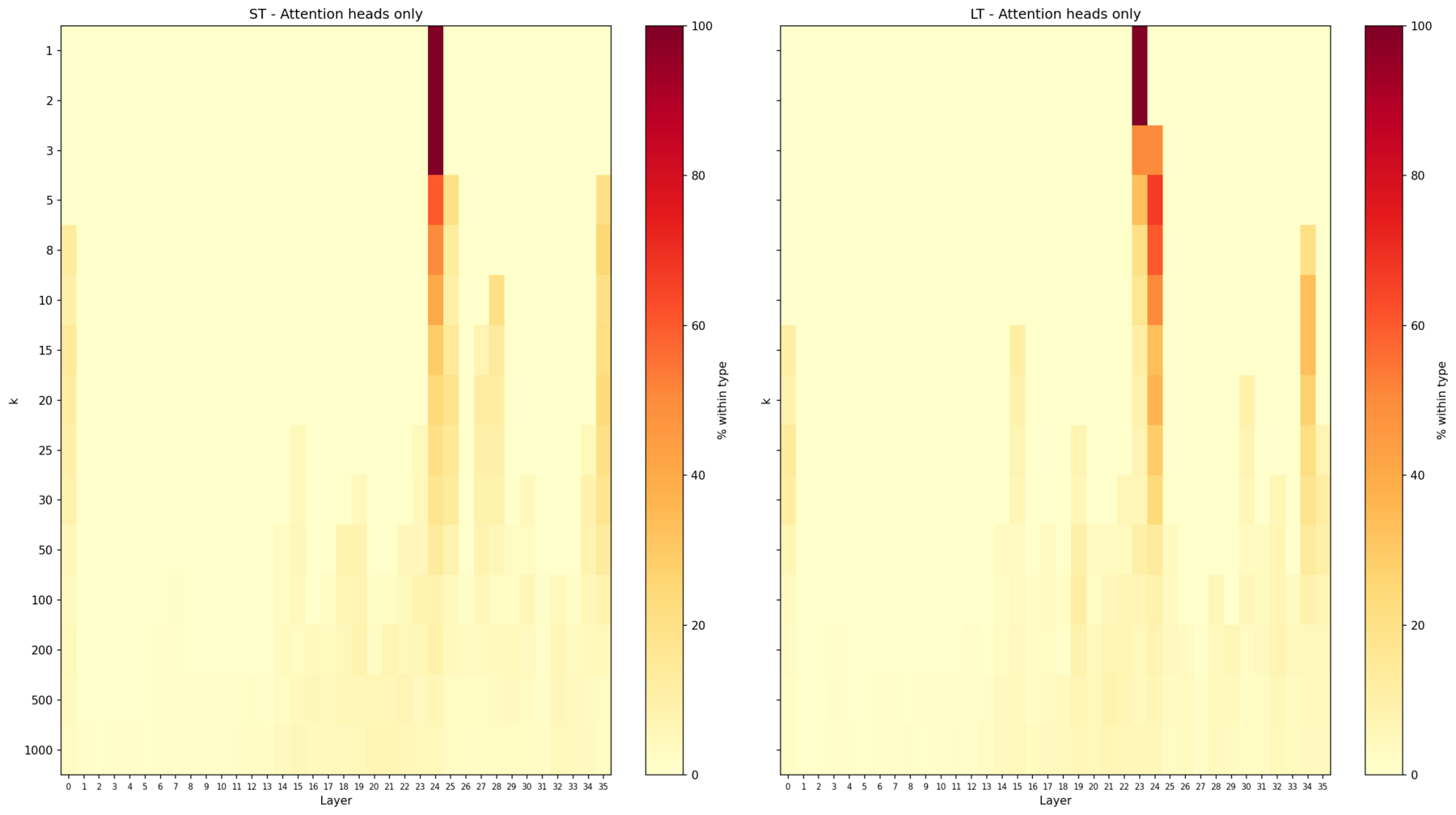}
    \end{minipage}\hfill
    \begin{minipage}[t]{0.48\textwidth}
        \centering
        \includegraphics[width=\linewidth]{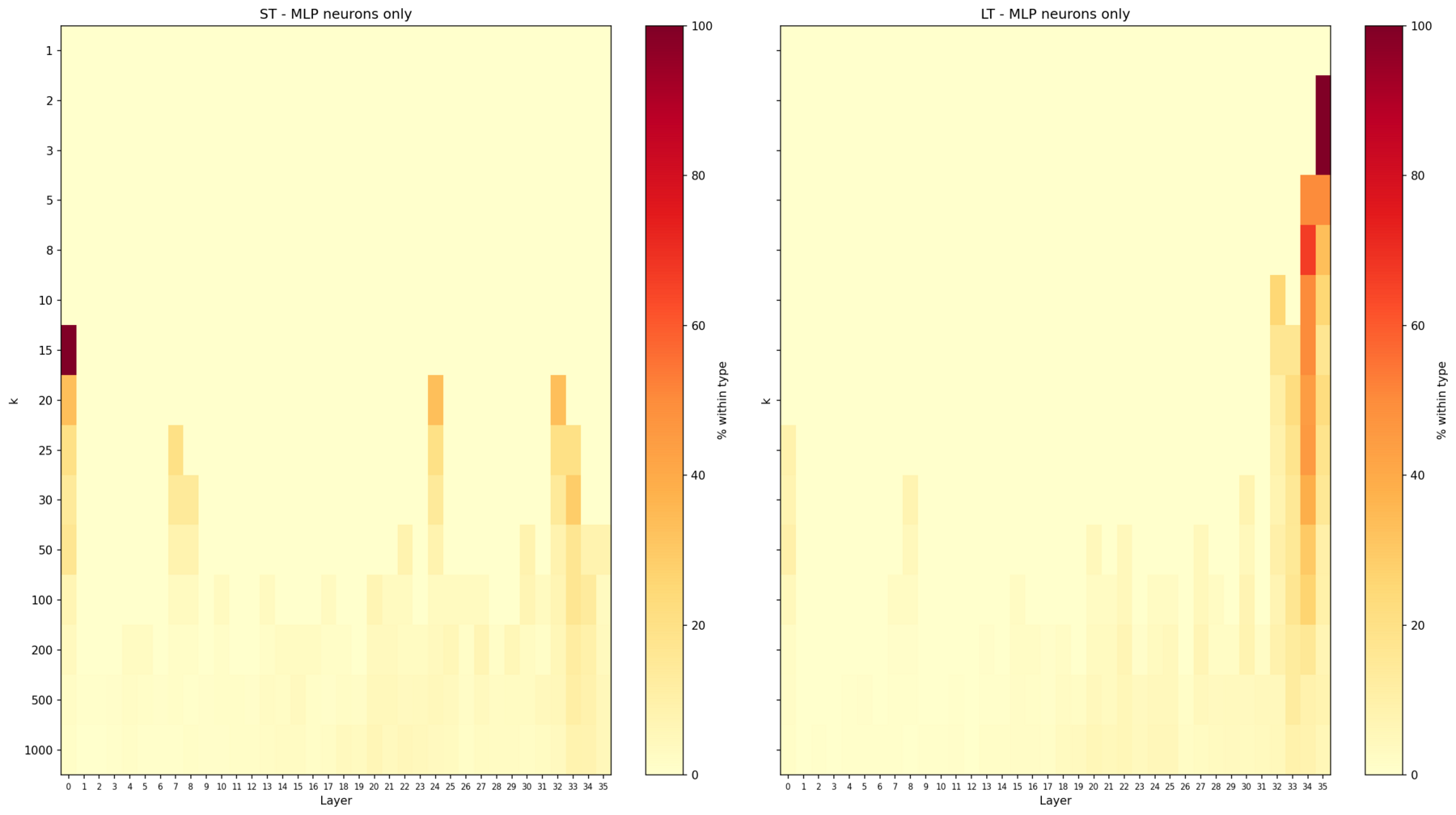}
    \end{minipage}
    \caption{Attribution heatmaps for attention heads (left) and MLP neurons (right) across layers.
        Attention: a sparse set of heads in layers 21--26 carries disproportionate attribution, with L24 heads showing the strongest signal, converging with activation patching results (\ref{app:causal-parametric}, \ref{app:causal-contrastive}).
        MLP: attribution is concentrated in the upper layers (L31--L35) and more evenly distributed across neuron indices, suggesting distributed rather than sparse MLP processing.}
    \label{fig:heatmap_attn_only}
\end{figure}

\FloatBarrier

\subsection{Short-Term vs.\ Long-Term Component Comparison}

A key question is whether short-term and long-term temporal preferences are processed by the same components or by specialized subpopulations.
We compare the top-$k$ attributed components for the short-term concept against those for the long-term concept.
The results reveal partial but incomplete overlap: many components contribute to both concepts, but each concept also recruits specialized nodes.
This pattern is consistent with a shared temporal-processing backbone augmented by concept-specific refinement, supporting the paper's claim that temporal preference is a structured rather than monolithic representation.

\begin{figure}[!htbp]
    \centering
    \begin{minipage}[t]{0.48\textwidth}
        \centering
        \includegraphics[width=\textwidth]{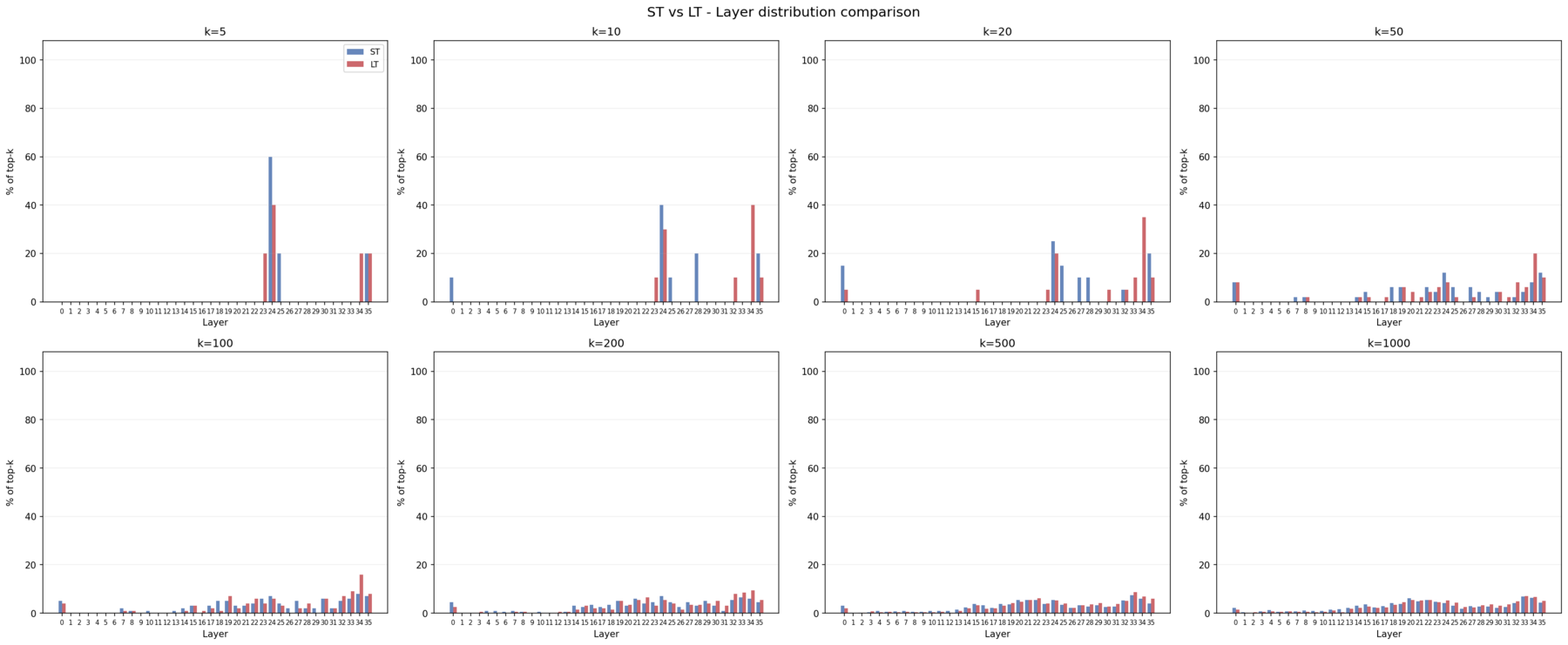}
    \end{minipage}\hfill
    \begin{minipage}[t]{0.48\textwidth}
        \centering
        \includegraphics[width=\textwidth]{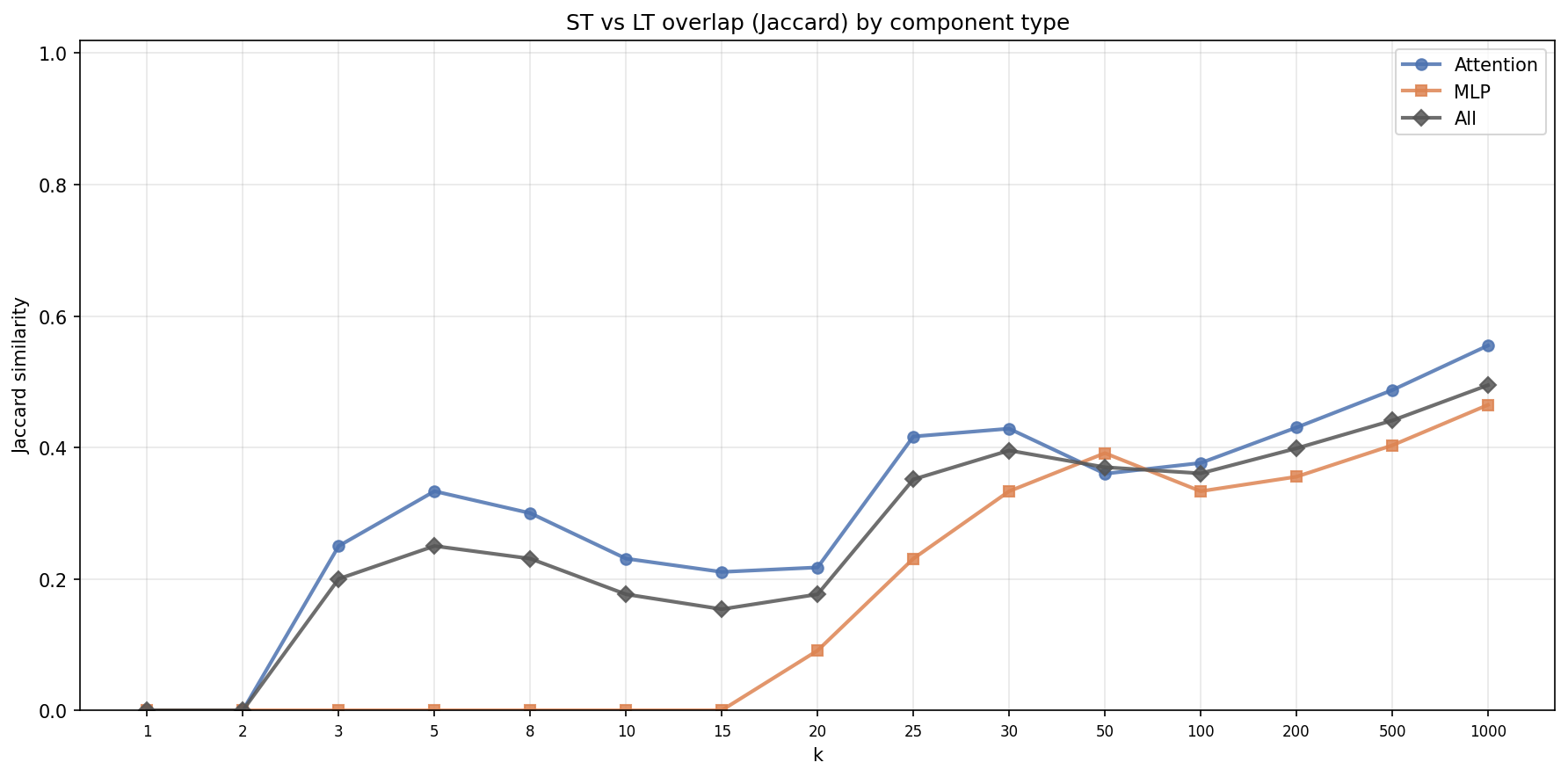}
    \end{minipage}
    \caption{Left: short-term and long-term attributed components show partial overlap.
    Short-term attribution peaks at L24, long-term shifts toward L22.
    Right: Jaccard similarity between ST and LT top-$k$ sets increases with $k$, from concept-specific nodes at small $k$ to a shared temporal backbone at larger $k$.}
    \label{fig:st_vs_lt_comparison}
\end{figure}

At small $k$, the overlap is low, indicating the most important components are concept-specific.
As $k$ grows, overlap increases, reflecting shared temporal processing consistent with the geometric separation in \ref{app:parametric-geometry}.

\FloatBarrier

\subsection{Individual Component Analysis}

While the diffuse distribution of attribution scores limits confidence in any single component (see Section~\ref{app:attributional-contrastive} limitations discussion above), examining the highest-scoring individual nodes provides a useful sanity check.
The top-ranked nodes cluster in the same mid-to-upper layer range identified by layer-level analysis, and the highest-attribution attention heads fall in L22--L24, precisely the layers flagged by activation patching as causally important.
However, even the top-ranked individual components account for less than 0.1\% of total attribution mass, underscoring why we treat EAP-IG as a selection prior rather than a definitive circuit-identification tool.

\begin{figure}[!htbp]
    \centering
    \begin{minipage}[t]{0.48\textwidth}
        \centering
        \includegraphics[width=\textwidth]{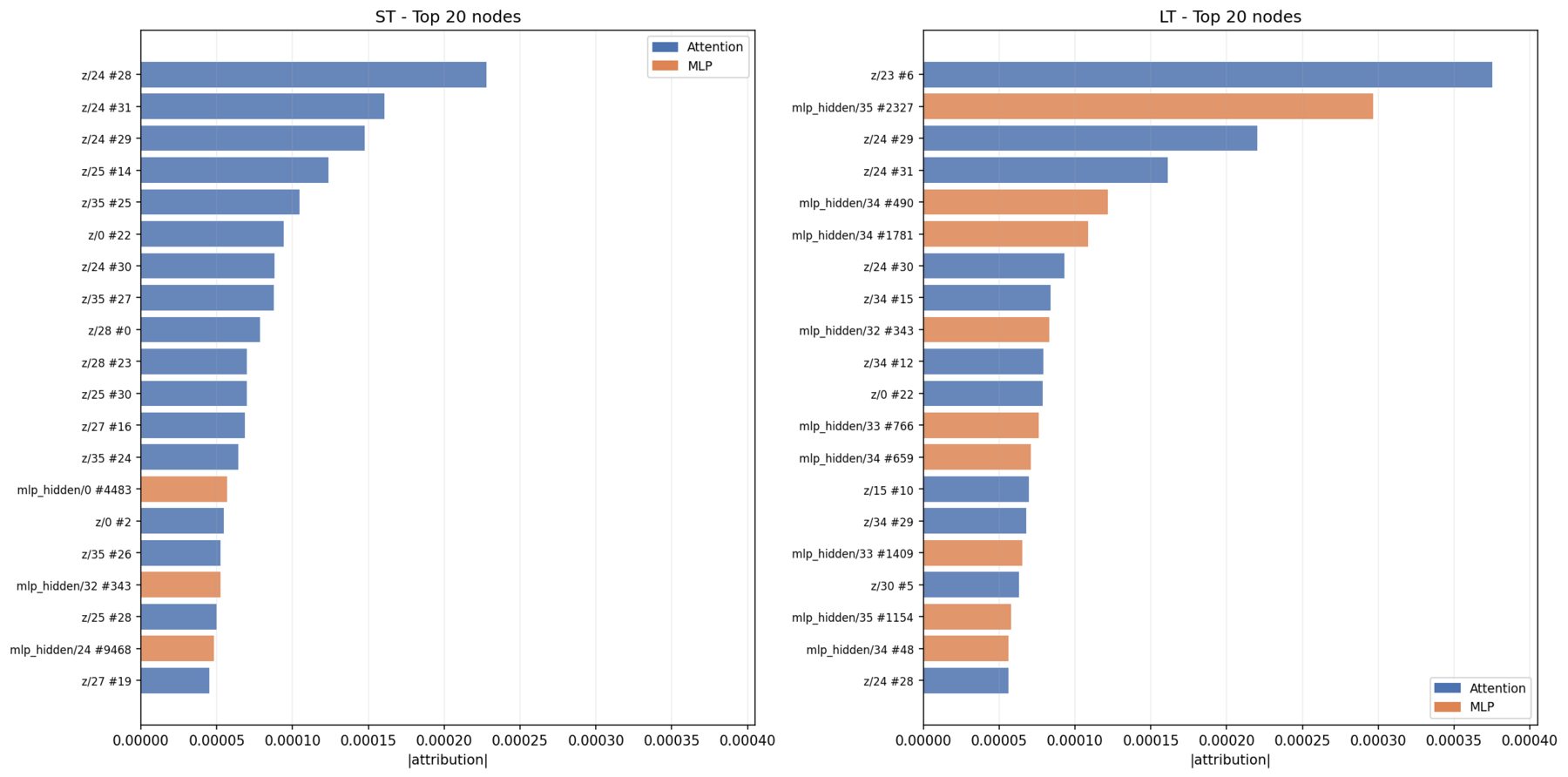}
    \end{minipage}\hfill
    \begin{minipage}[t]{0.48\textwidth}
        \centering
        \includegraphics[width=\textwidth]{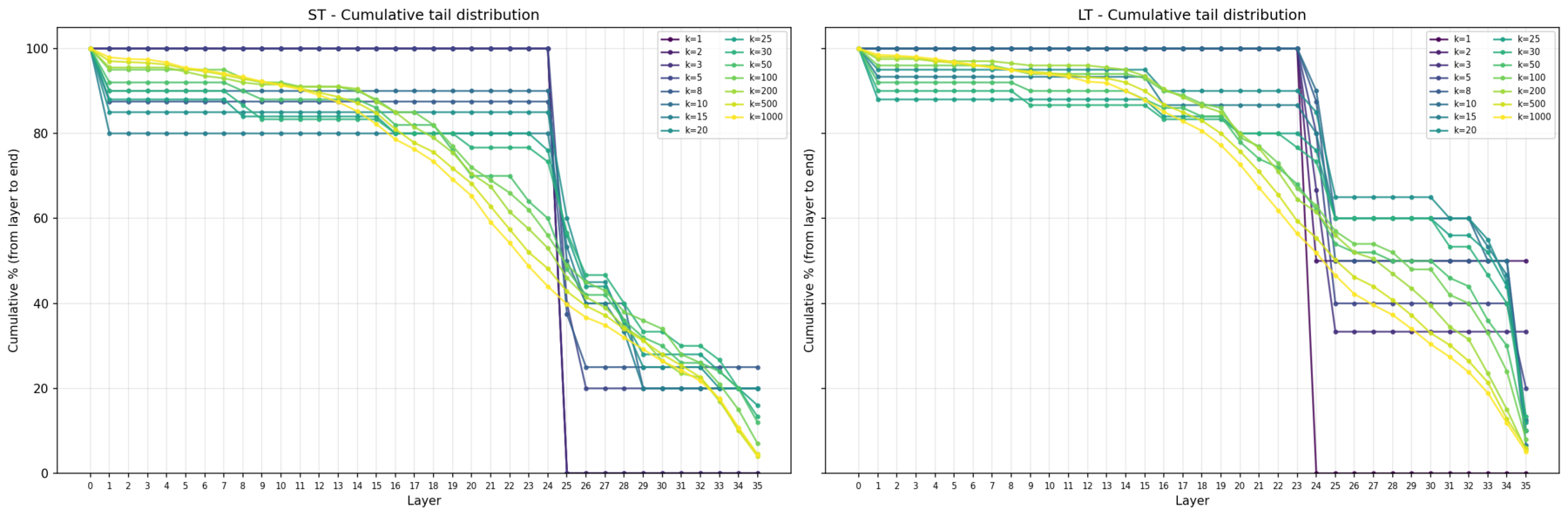}
    \end{minipage}
    \caption{Left: top individual nodes ranked by attribution score; the highest-ranked are attention heads in L22--L24, with no single component exceeding 0.1\% of total mass.
    Right: cumulative tail distribution; attribution mass accumulates slowly, confirming a diffuse circuit ($\sim$0.125\% of all nodes).}
    \label{fig:top_individual_nodes}
\end{figure}

\begin{figure}[htbp]
    \centering
    \begin{minipage}[t]{0.48\linewidth}
        \centering
        \includegraphics[width=\linewidth]{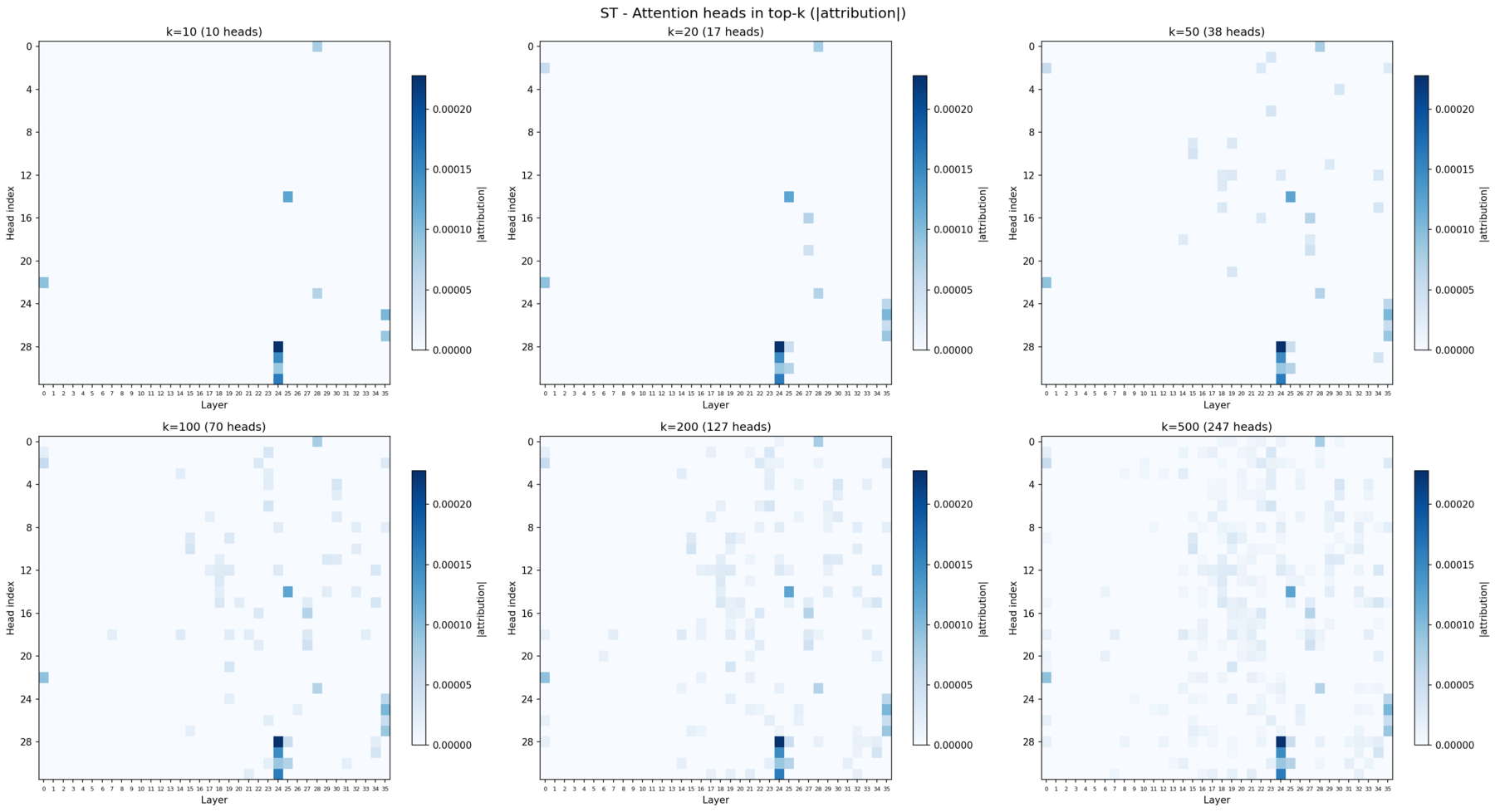}
    \end{minipage}\hfill
    \begin{minipage}[t]{0.48\linewidth}
        \centering
        \includegraphics[width=\linewidth]{images/localize/attributional_contrastive/head_matrix_lt.png}
    \end{minipage}
    \caption{Attention head attribution matrices for short-term (left) and long-term (right) concepts.
    Short-term attribution is concentrated in a sparse set of L24 heads, while long-term attribution is distributed more broadly across L22--L26, revealing an asymmetry in circuit structure between the two temporal concepts.}
    \label{fig:head_matrices}
\end{figure}

Finally, the full attention head matrices (Figure~\ref{fig:head_matrices}) provide a detailed view of which heads matter for each concept.
The short-term matrix shows concentrated signal in a few heads around L24, while the long-term matrix distributes attribution more broadly across L22--L26.
This asymmetry suggests that short-term preference relies on a slightly more focused set of attention heads, whereas long-term preference recruits a wider subnetwork.

\FloatBarrier

\clearpage
\clearappnumbering

\section{Attributional parametric results}\label{app:attributional-parametric}

The previous two appendices applied different methods to different prompts: EAP-IG on contrastive prompts (\ref{app:attributional-contrastive}), then activation patching on parametric prompts (\ref{app:causal-parametric}).
To disentangle the effect of the method from the effect of the prompt paradigm, we apply standard attribution patching (EAP-IG) to the same parametric prompts used for activation patching.
This completes one diagonal of the method $\times$ paradigm matrix and lets us ask: do the layers identified by attribution match those identified by causal intervention on the same data?

\subsection{Attribution score distribution}

\begin{figure}[htbp]
\centering
\begin{minipage}[t]{0.48\textwidth}
    \centering
    \includegraphics[width=\linewidth]{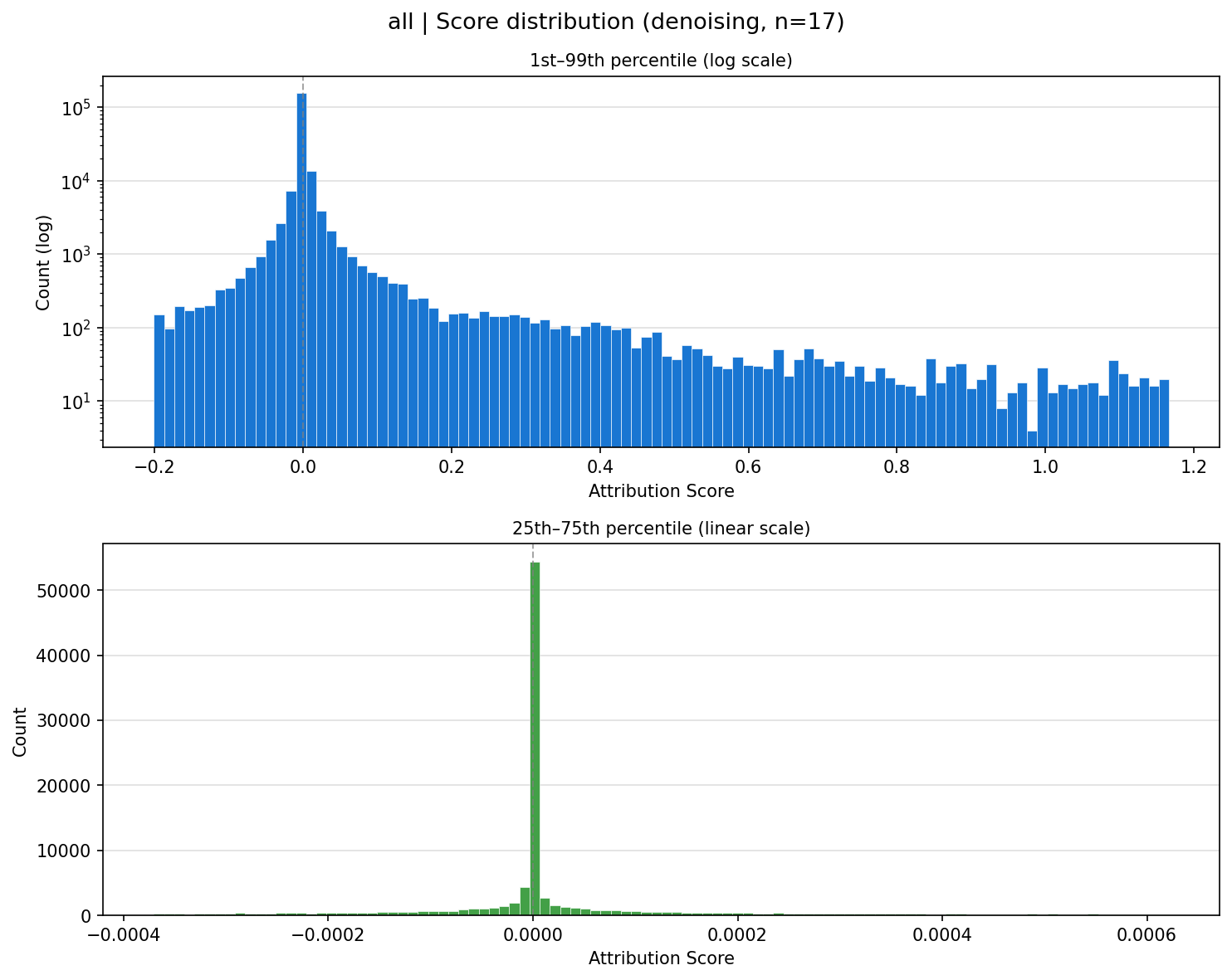}
\end{minipage}\hfill
\begin{minipage}[t]{0.48\textwidth}
    \centering
    \includegraphics[width=\linewidth]{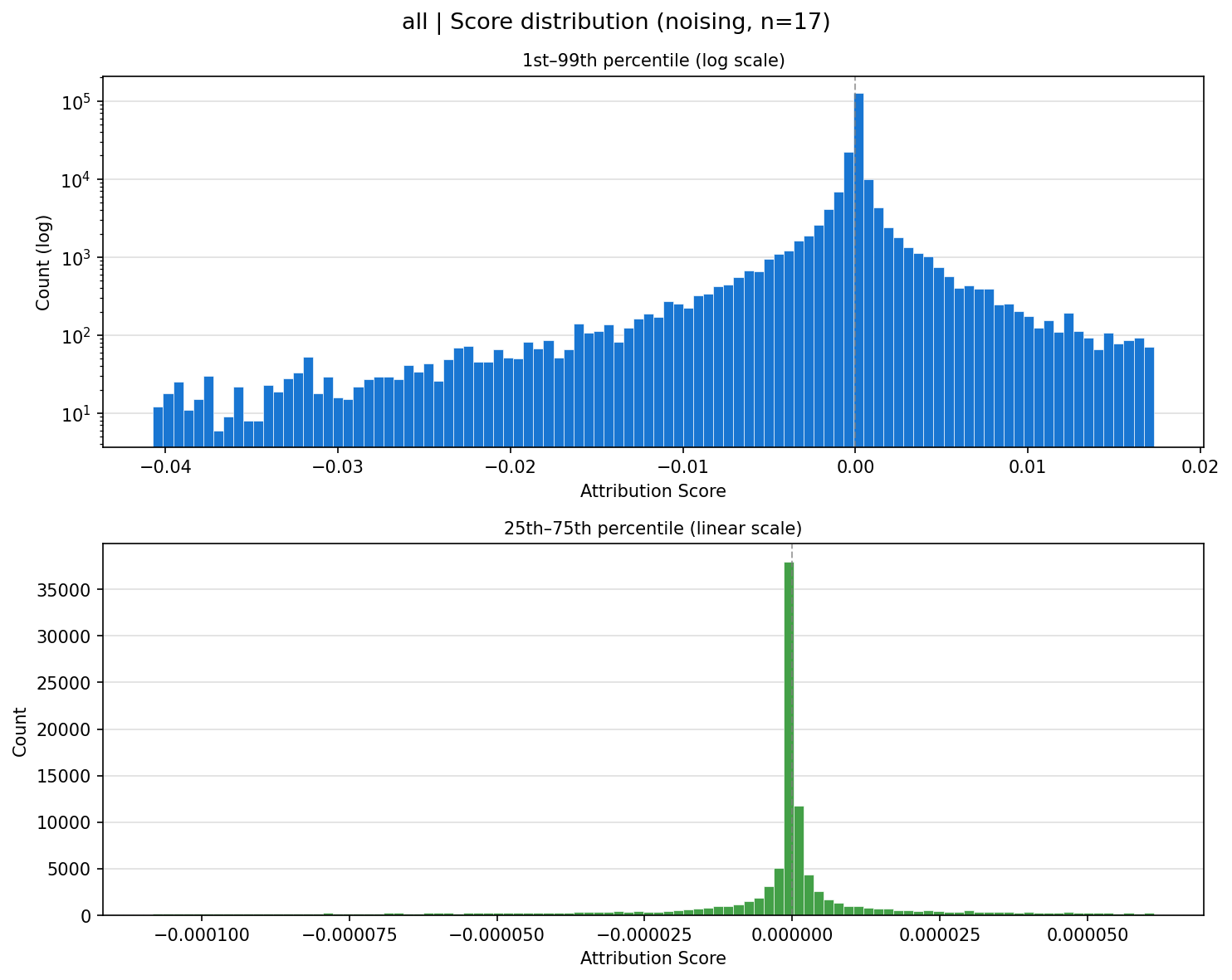}
\end{minipage}
\caption{Attribution score distributions under denoising (left) and noising (right).
Top rows: 1st--99th percentile on log scale.
The denoising distribution has a heavy positive tail extending to $\sim$1.2, while the noising distribution is more symmetric ($\pm$0.04).
Bottom rows: 25th--75th percentile on linear scale, with central mass within $\pm$0.0003 (denoising) and $\pm$0.00005 (noising).}
\label{fig:ap-hist-denoising}
\end{figure}

As with the contrastive attribution results (\ref{app:attributional-contrastive}), the vast majority of components have near-zero attribution scores.
Figure~\ref{fig:ap-hist-denoising} shows the score distributions under denoising and noising.
The denoising distribution has a heavy positive tail extending to $\sim$1.2, while the noising distribution is more symmetric and concentrated near zero ($\pm$0.04).
The central 50\% of scores (bottom panels, linear scale) are confined to a very narrow band around zero, $\pm$0.0003 for denoising and $\pm$0.00005 for noising, confirming that only a small fraction of components carry meaningful attribution.

\FloatBarrier

\subsection{Top-scoring components}

\begin{figure}[htbp]
\centering
\begin{minipage}[t]{0.48\textwidth}
    \centering
    \includegraphics[width=\linewidth]{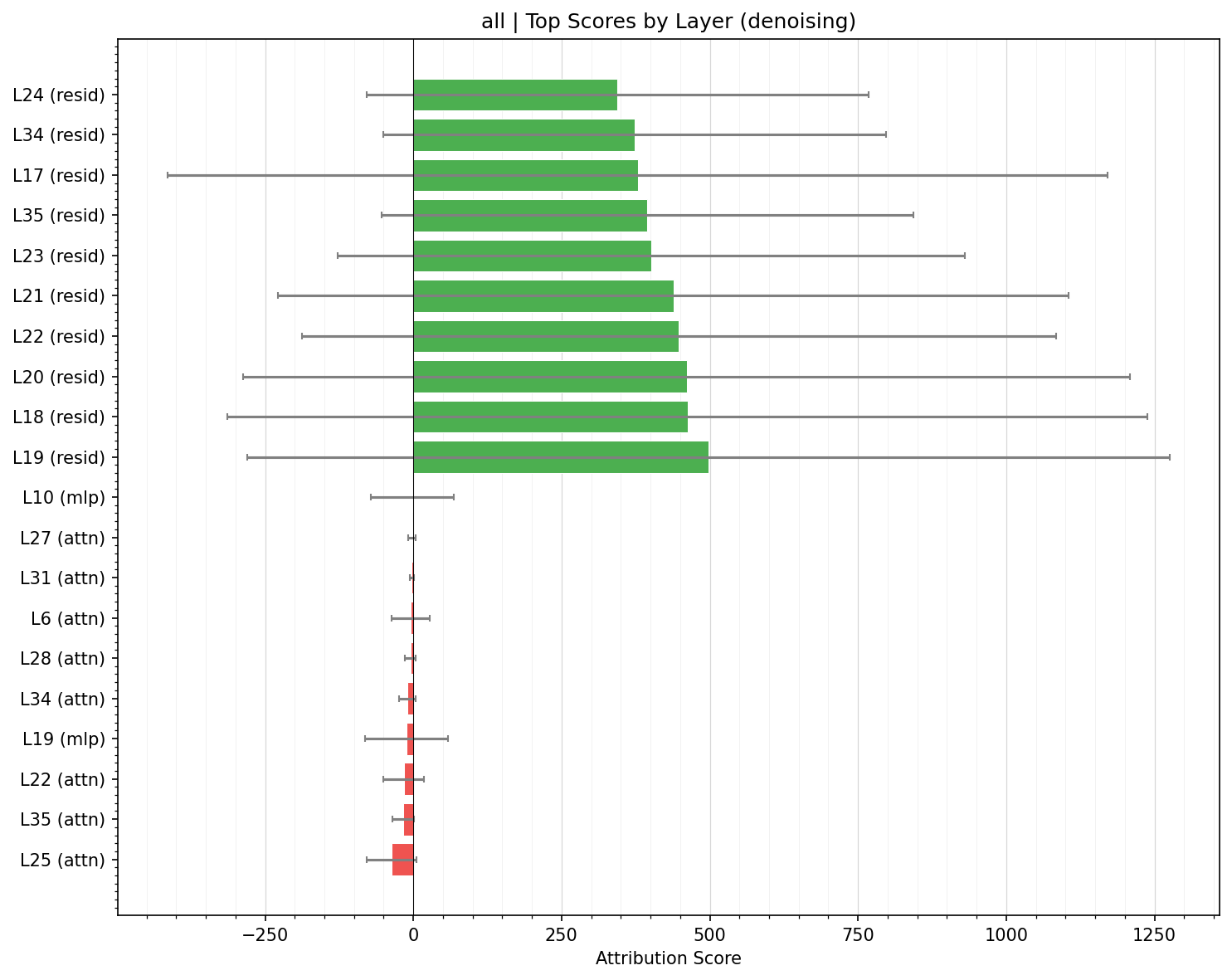}
\end{minipage}\hfill
\begin{minipage}[t]{0.48\textwidth}
    \centering
    \includegraphics[width=\linewidth]{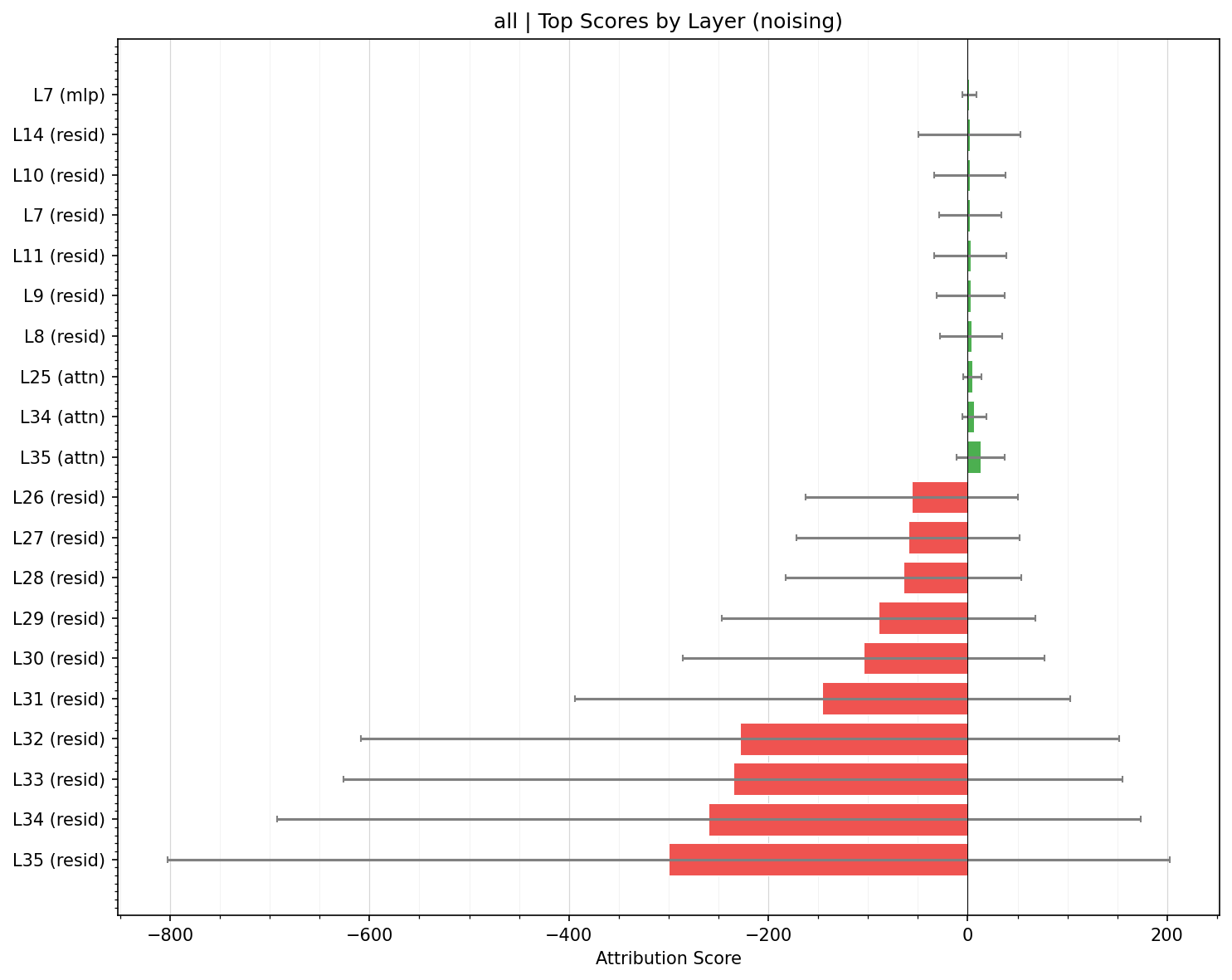}
\end{minipage}
\caption{Top components ranked by denoising (left) and noising (right) attribution score.
Denoising: residual stream components dominate, with L19, L18, and L20 carrying the largest scores ($\sim$300--500); L25 attention is a notable negative outlier.
Noising: late-layer residual components (L26--L35) carry the largest negative attributions, revealing a sufficiency/necessity asymmetry.}
\label{fig:ap-top-denoising}
\end{figure}

Figure~\ref{fig:ap-top-denoising} ranks individual components by their attribution scores under denoising and noising.

Under denoising, the residual stream at mid-layers (L17--L22) carries the bulk of the recovery signal, with individual scores reaching $\sim$500.
Attention and MLP components are an order of magnitude smaller.
Under noising, the signal shifts to late layers (L26--L35), where corrupted residual activations cause the most disruption.
This asymmetry between where information is built (mid-layers) and where it becomes vulnerable (late layers) parallels the sufficiency/necessity gap observed in the causal parametric experiments (\ref{app:causal-parametric}).

\FloatBarrier

\subsection{Recovery vs.\ disruption by component type}

\begin{figure}[!htbp]
\centering
\begin{minipage}[t]{0.48\textwidth}
    \centering
    \includegraphics[width=\linewidth]{images/localize/attributional_parametric/mode_scatter_by_component.png}
\end{minipage}\hfill
\begin{minipage}[t]{0.48\textwidth}
    \centering
    \includegraphics[width=\linewidth]{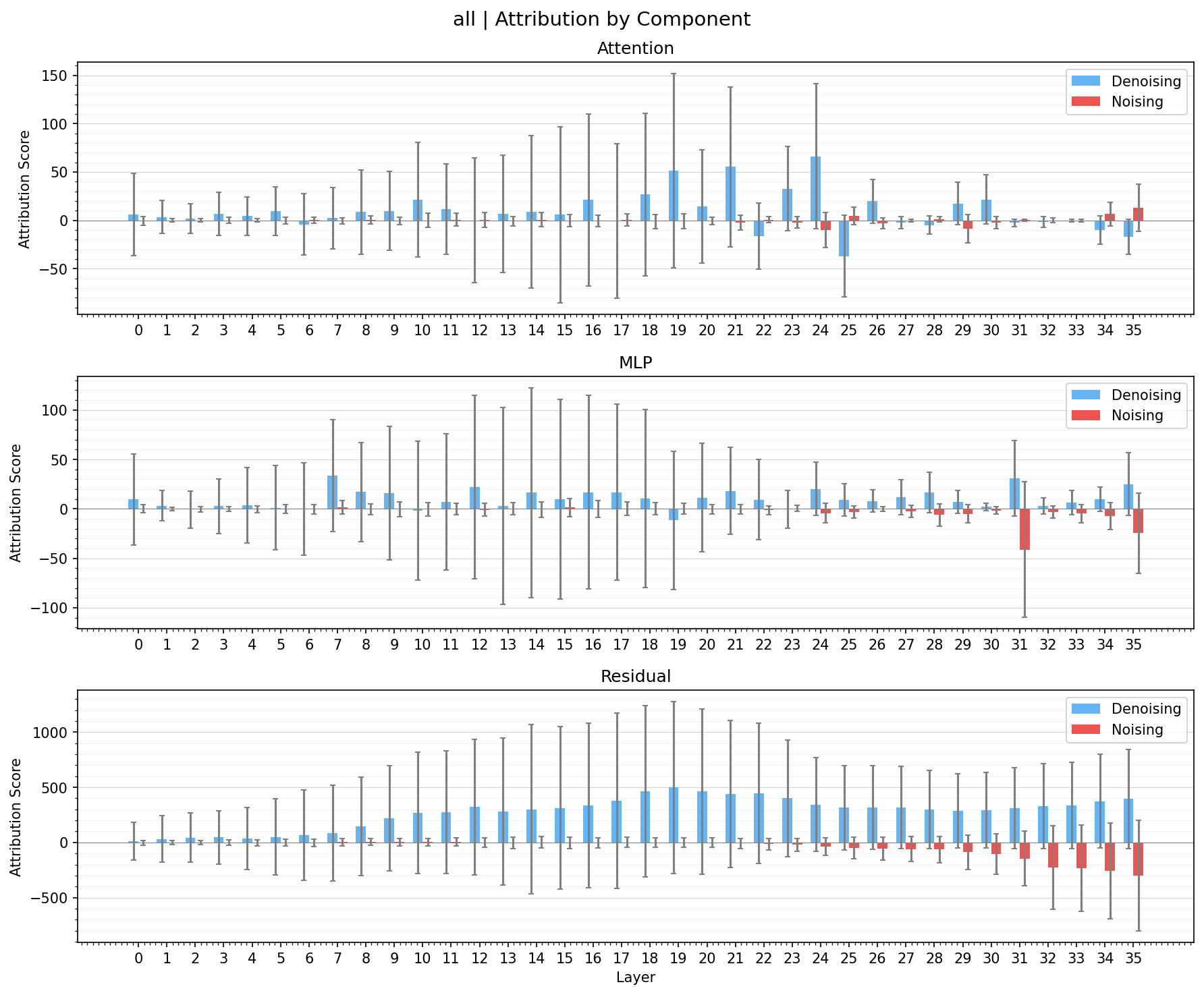}
\end{minipage}
\caption{\textbf{Left:} Recovery (denoising) vs.\ disruption (noising) attribution, separated by component type.
The residual stream panel operates on a much larger scale than attention or MLP.
Mid-layer residual components cluster in the high-recovery / low-disruption region; late-layer residual components cluster in the low-recovery / high-disruption region.
\texttt{L31\_mlp} stands out as a strong disruptor without proportional recovery.
\textbf{Right:} Per-layer attribution scores for attention (top), MLP (middle), and residual stream (bottom) under denoising (blue) and noising (red).
Attention denoising peaks at L19--L24 with a sharp negative dip at L25.
MLP noising shows a strong negative spike at L31.
The residual stream dominates in scale, with denoising plateauing at L17--L22 and noising growing increasingly negative from L25 onward.}
\label{fig:ap-mode-combined}
\end{figure}

Figure~\ref{fig:ap-mode-combined} (left) plots each component's denoising attribution against its noising attribution, separated by component type.
Figure~\ref{fig:ap-mode-combined} (right) shows per-layer attribution for each component type under both denoising and noising.

The scatter reveals that attention heads at L19--L24 contribute primarily to recovery (lower-right quadrant) without proportional disruption, consistent with a sufficiency-biased profile: patching them in restores behavior, but patching them out does not fully destroy it.
\texttt{L31\_mlp} is the clearest disruptor, consistent with its identification as a top-ranked MLP component in the causal experiments.

\FloatBarrier

\subsection{Layer-wise attribution by component type}

\begin{figure}[!htbp]
\centering
\begin{minipage}[t]{\textwidth}
    \centering
    \includegraphics[width=0.8\textwidth]{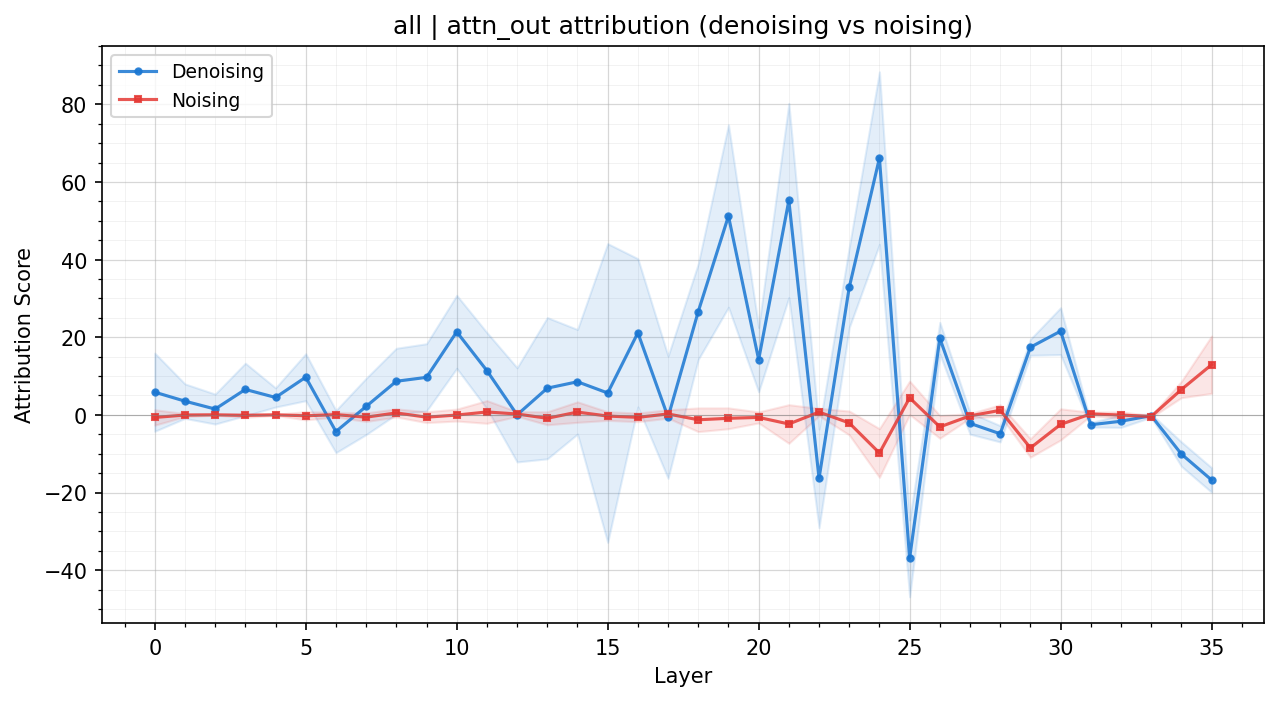}
\end{minipage}\\[0.5em]
\begin{minipage}[t]{\textwidth}
    \centering
    \includegraphics[width=0.8\textwidth]{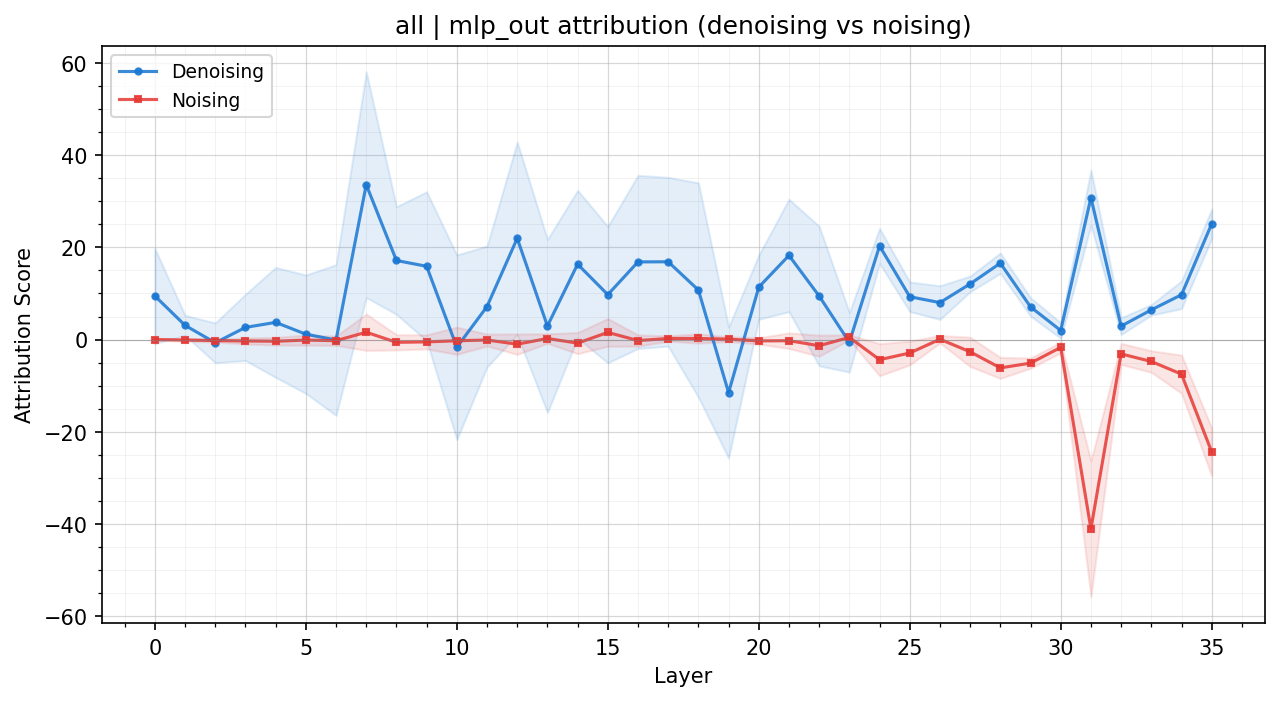}
\end{minipage}\\[0.5em]
\begin{minipage}[t]{\textwidth}
    \centering
    \includegraphics[width=0.8\textwidth]{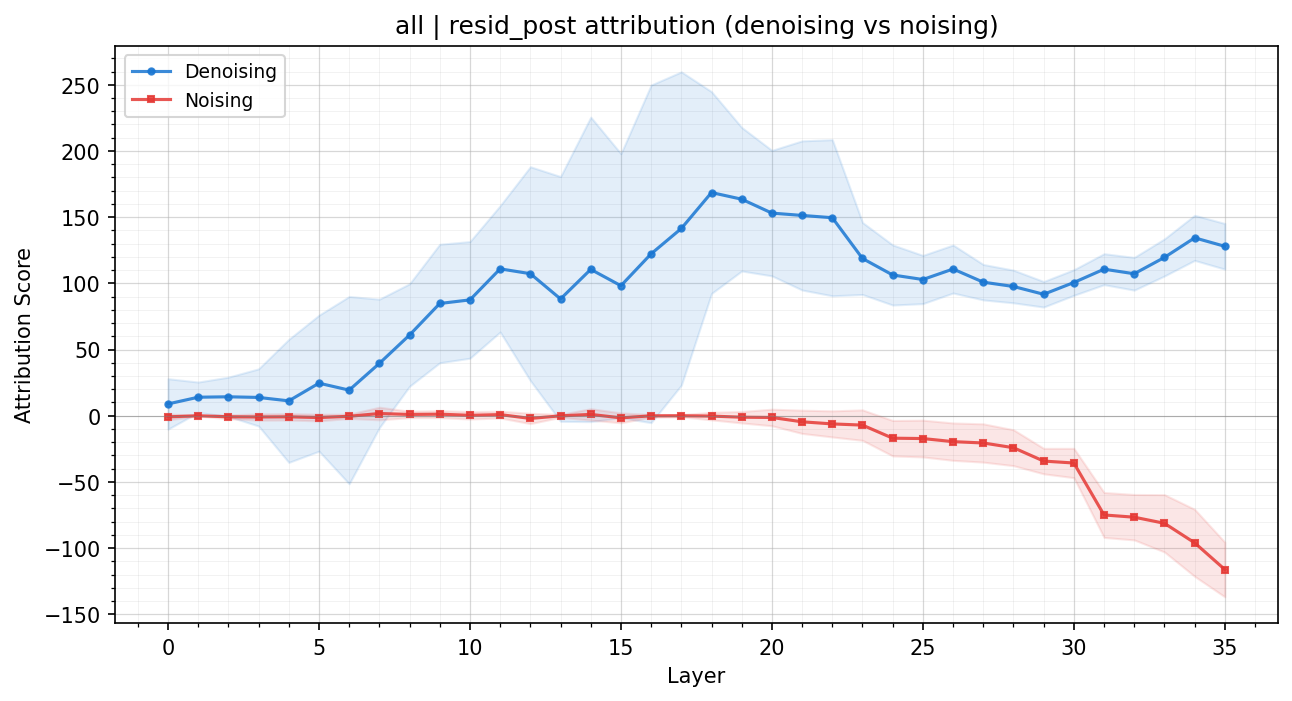}
\end{minipage}
\caption{Layer-wise attribution by component type, with denoising and noising overlaid.
\textbf{Top:} Attention output; denoising peaks at L20--L24 ($\sim$55--65), then drops sharply negative at L25 ($\sim -37$); noising is near zero throughout.
\textbf{Middle:} MLP output; denoising is modestly positive across most layers; noising shows a sharp negative spike at L31 ($\sim -42$).
\textbf{Bottom:} Residual stream (post-MLP); denoising rises steeply to a plateau at L17--L20 ($\sim$150--170); noising grows increasingly negative from L25 onward ($\sim -120$ at L35).}
\label{fig:ap-combined-lines}
\end{figure}

\ifdefined\colmmode\else
Figure~\ref{fig:ap-combined-lines} shows the denoising and noising curves overlaid for each component type.
\fi

\FloatBarrier

\ifdefined\colmmode
Figure~\ref{fig:ap-combined-lines} shows the denoising and noising curves overlaid for each component type.
\fi

\subsection{Position $\times$ layer heatmaps}

The heatmaps below show attribution across token positions and layers for key component types, revealing where in the prompt the temporal signal is concentrated.

\begin{figure}[!htbp]
\centering
\includegraphics[width=0.7\textwidth]{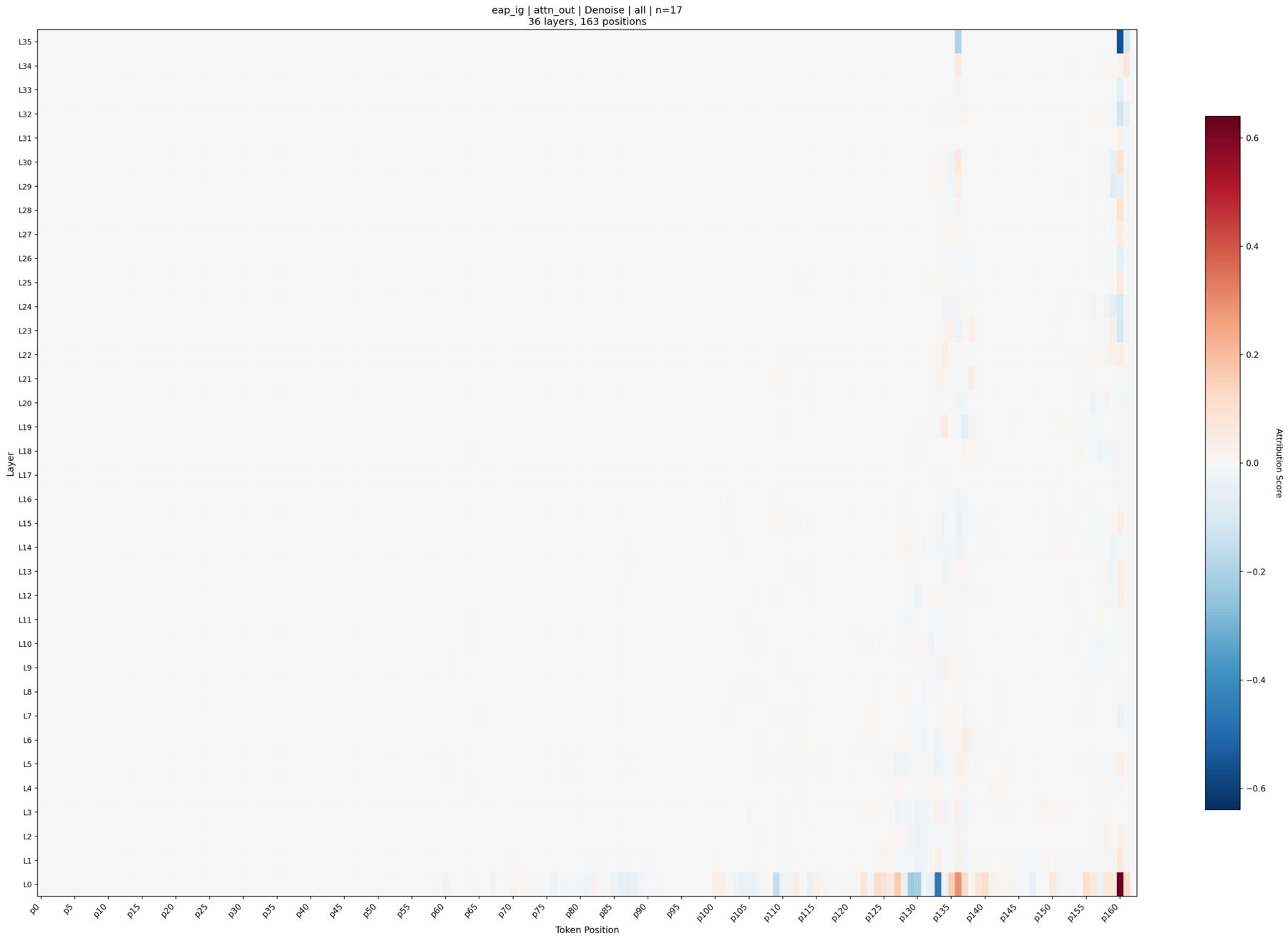}
\caption{Attention output denoising heatmap (position $\times$ layer).
Attribution is sparse and concentrated in the last $\sim$30 token positions (the answer region).
The final position shows the strongest signal.}
\label{fig:ap-attn-heatmap-denoise}
\end{figure}

\begin{figure}[!htbp]
\centering
\includegraphics[width=0.7\textwidth]{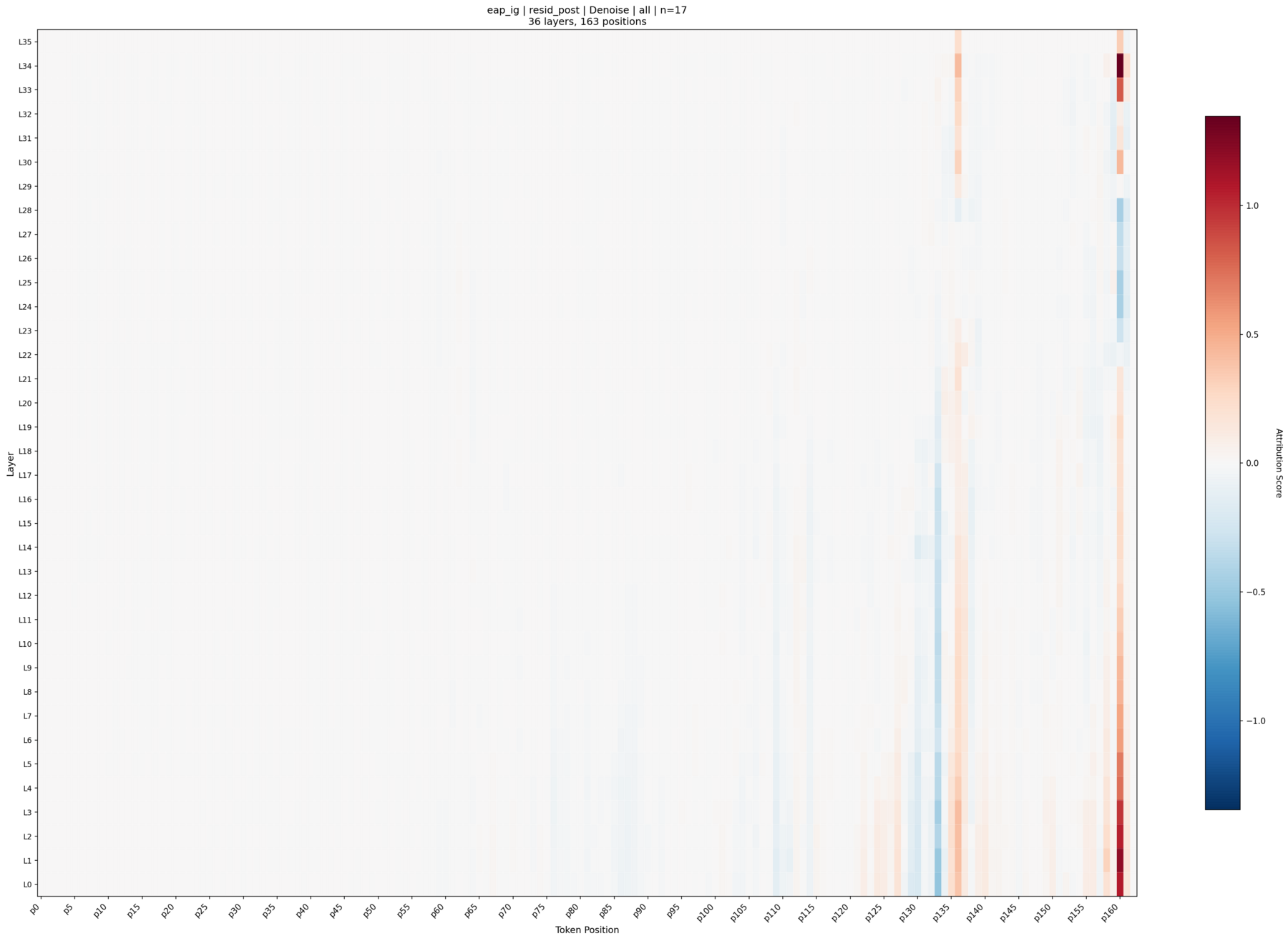}
\caption{Residual stream (post-MLP) denoising heatmap.
Complex vertical-stripe patterns from position $\sim$75 onward, with the strongest positive signal at the final position in upper layers (L34).
Early positions (the prompt preamble) contribute almost nothing.}
\label{fig:ap-resid-heatmap-denoise}
\end{figure}

\begin{figure}[!htbp]
\centering
\includegraphics[width=0.7\textwidth]{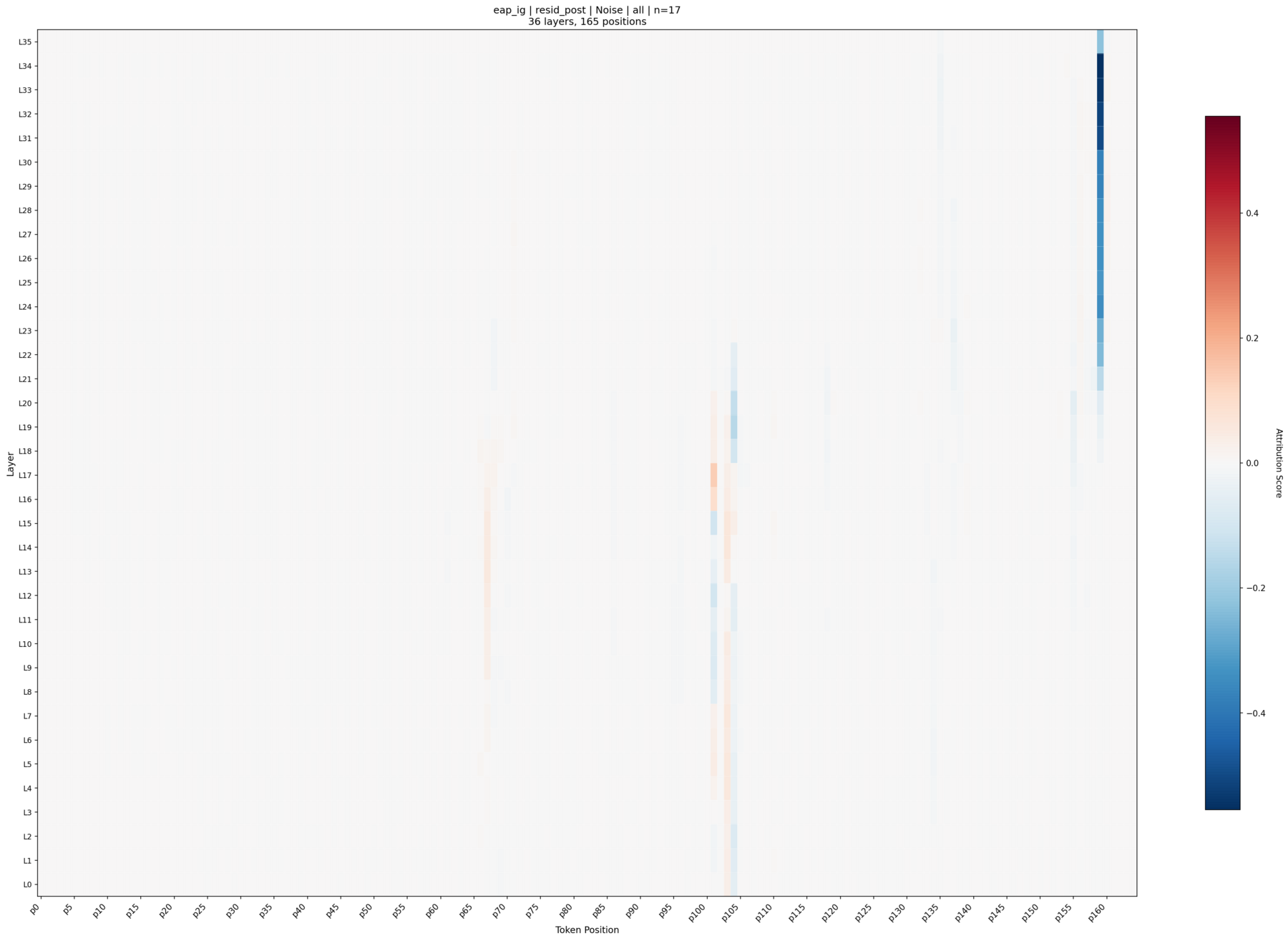}
\caption{Residual stream (post-MLP) noising heatmap.
Disruption is concentrated at the final token position in L35, with moderate mixed-sign activity at positions $\sim$65--110 in mid-layers.
The noising signal is sparser and more localized than the denoising signal.}
\label{fig:ap-resid-heatmap-noise}
\end{figure}

\FloatBarrier

\subsection{Key findings}

\begin{enumerate}[leftmargin=1.5em]
\item \textbf{Residual stream dominates attribution.}
Attribution scores for residual stream components are an order of magnitude larger than for attention or MLP, reflecting the cumulative nature of the residual stream.

\item \textbf{Mid-layer recovery, late-layer disruption.}
Denoising attribution peaks at L17--L22; noising peaks at L30--L35.
The model builds temporal information through mid-layers and becomes most vulnerable to corruption in late layers.

\item \textbf{Attention L19--L24 and the L25 anomaly.}
Attention heads in L19--L24 contribute to recovery, consistent with the causal parametric results.
L25 attention is a negative outlier under denoising; we do not interpret this causally from attribution alone.

\item \textbf{\texttt{L31\_mlp} as the key disruptor.}
\texttt{L31\_mlp} is the single most disruptive non-residual component, matching its identification in the causal experiments.

\item \textbf{Signal concentrates at late token positions.}
Most attribution mass falls in the last $\sim$30--40 token positions (the answer / format region), with the final position consistently the most important.
The prompt preamble carries almost no temporal attribution.

\item \textbf{Convergence with causal methods.}
The layers and components identified by gradient-based attribution match those found by activation patching on the same prompts (\ref{app:causal-parametric}): attention L21--L24, MLP L31 / L35, and the mid-to-late residual stream.
This cross-method agreement validates that both approaches recover the same underlying circuit.
\end{enumerate}

\clearpage
\clearappnumbering

\section{Causal parametric results}
\label{app:causal-parametric}

The attribution results in \ref{app:attributional-contrastive} flagged layers 21--35 but could not establish causal effect.
Here we apply the gold standard: activation patching on $n = 71$ highly-formatted parametric contrastive pairs, directly replacing component activations with counterfactual values (methodology in \ref{app:causal-parametric-methods}).
Where EAP-IG approximates, patching measures the actual behavioral consequence of intervention.

The results sharpen the picture considerably.
A sparse set of four components, \texttt{L24\_attn}, \texttt{L21\_attn}, \texttt{L35\_mlp}, and \texttt{L31\_mlp}, account for the majority of the causal effect, clearly separated from the rest.
L24 attention, the same layer flagged by EAP-IG, emerges as the single most important component under both denoising and noising.

\FloatBarrier

\subsection{Component importance ranking}

\begin{figure}[!htbp]
    \centering
    \includegraphics[width=0.75\textwidth]{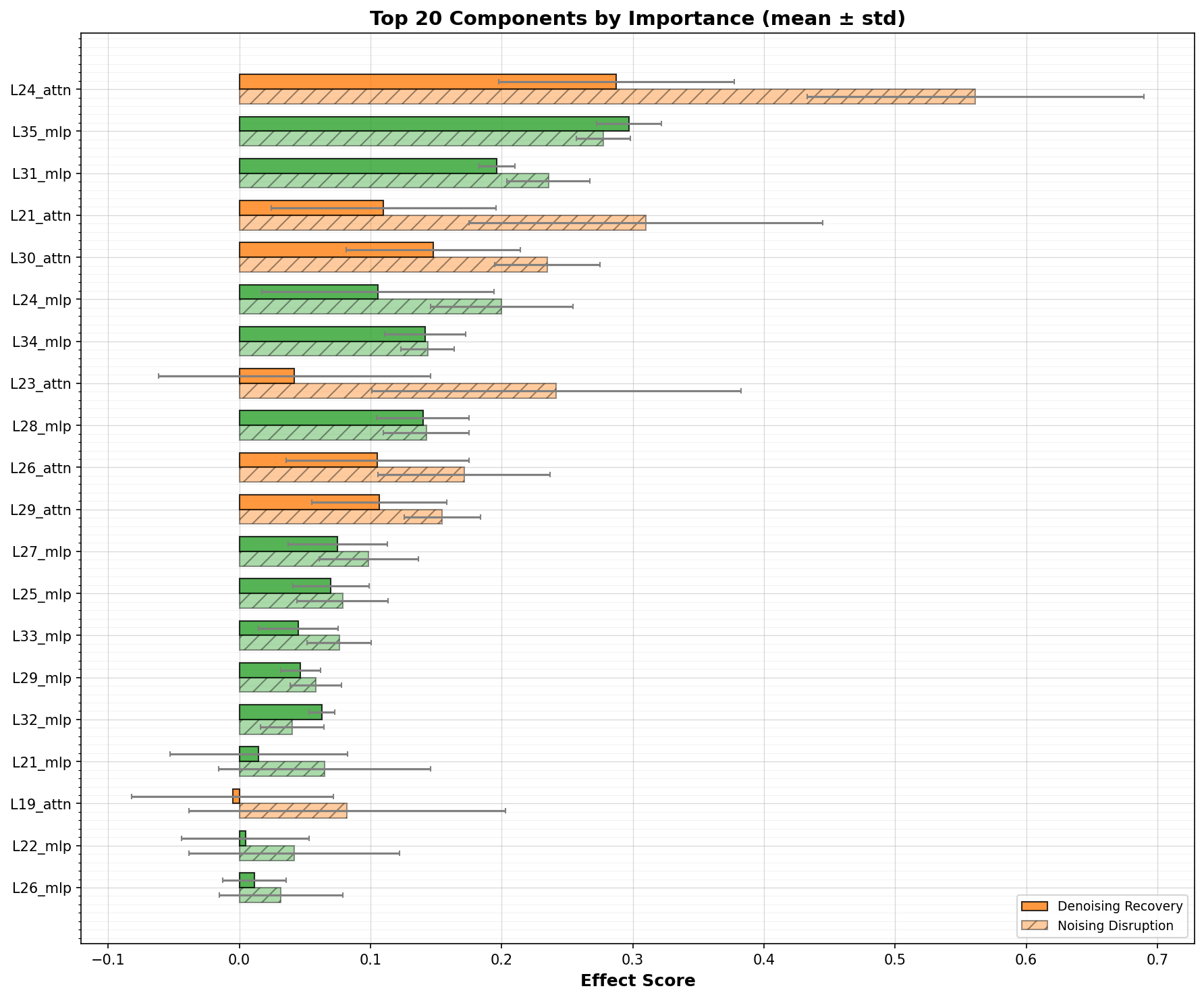}
    \caption{
        Top 20 components ranked by mean effect score (denoising recovery and noising disruption, with standard deviation across contrastive pairs).
        \texttt{L24\_attn} ranks highest, with a noising disruption score near 0.56 (the only component above 0.5).
        \texttt{L21\_attn} is the next-largest attention contributor at $\sim$0.31; \texttt{L35\_mlp} ($\sim$0.27) and \texttt{L31\_mlp} ($\sim$0.24) lead the MLP components.
        The fifth-ranked component (\texttt{L30\_attn}) has roughly half the effect of \texttt{L24\_attn}, separating the top four from the rest.
    }
    \label{fig:component_importance_ranked}
\end{figure}

We begin with a direct ranking of individual components by their causal effect size.
Figure~\ref{fig:component_importance_ranked} shows the top 20 components sorted by the mean of their denoising recovery and noising disruption scores across all contrastive pairs.

The ranking reveals a clear separation between a small number of high-effect components and a long tail of modest contributors.
Attention components dominate the top of the list, with \texttt{L24\_attn} showing the largest effect under both denoising and noising.
The most causally important MLP components (\texttt{L35\_mlp} and \texttt{L31\_mlp}) rank among the top four overall, with effect sizes comparable to \texttt{L21\_attn}.
The asymmetry between denoising recovery and noising disruption is particularly pronounced for the top attention layers: \texttt{L24\_attn} and \texttt{L21\_attn} show much higher noising disruption than denoising recovery, suggesting these components are more necessary than sufficient: corrupting them degrades performance substantially, but restoring them alone does not fully recover clean behavior.

\FloatBarrier

\subsection{Marginal contribution analysis}

\begin{figure}[!htbp]
    \centering
    \includegraphics[width=0.75\textwidth]{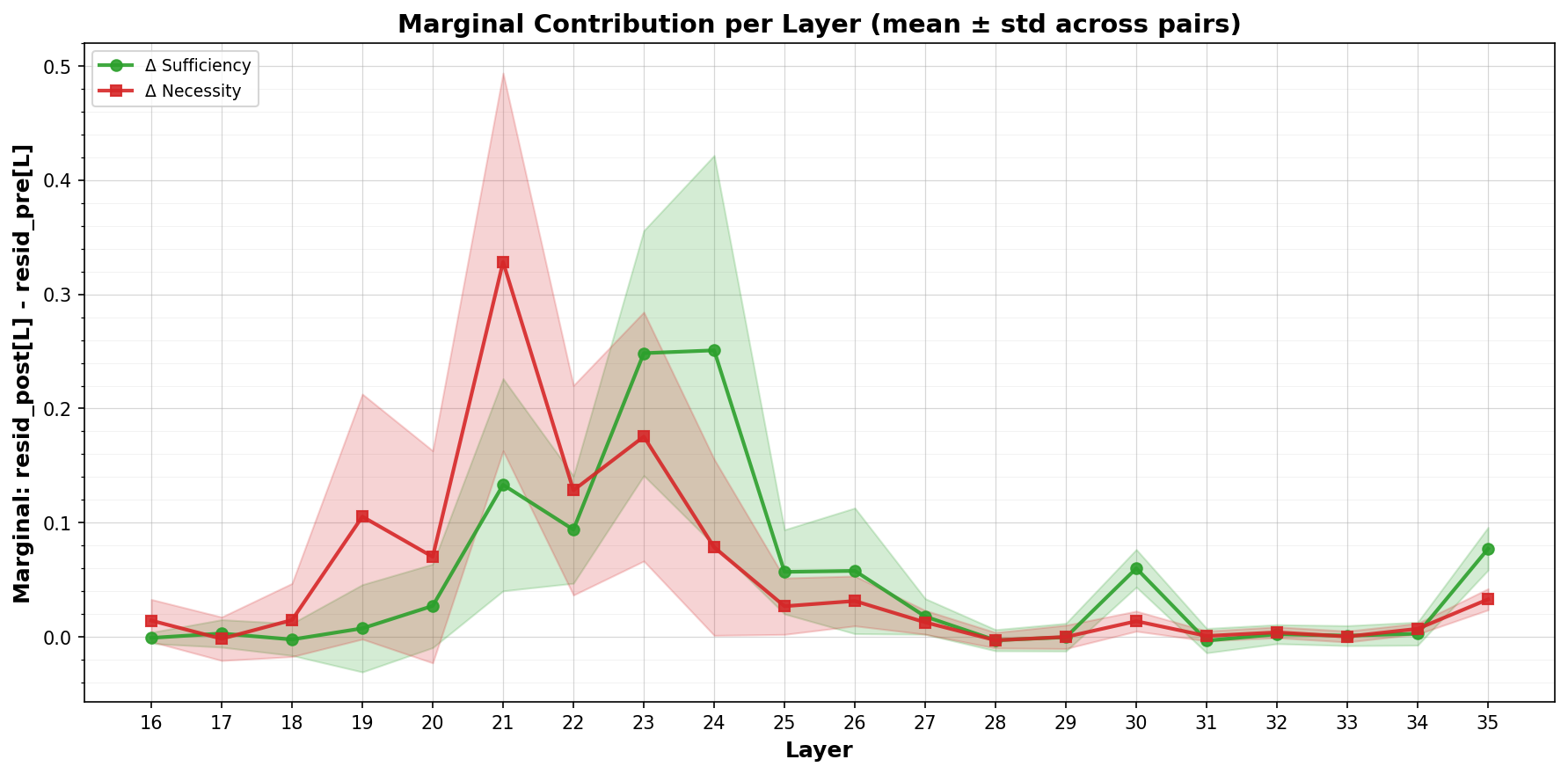}
    \caption{
        Marginal contribution per layer (mean $\pm$ standard deviation across contrastive pairs), showing sufficiency (denoising recovery, green) and necessity (noising disruption, red).
        Sufficiency peaks sharply at layers 21--24, with layer 22 showing the single highest spike.
        Necessity is flatter and lower, reflecting the distributed nature of disruption.
        The high variance in layers 20--25 reflects the sensitivity of these layers to the specific temporal framing used in each contrastive pair.
    }
    \label{fig:marginal_contribution_var_p}
\end{figure}

Figure~\ref{fig:marginal_contribution_var_p} examines the marginal contribution of each layer, defined as the difference in residual stream activations before and after the layer (\texttt{resid\_post[L]} $-$ \texttt{resid\_pre[L]}).
This isolates each layer's additive contribution to the residual stream.

The sufficiency peak at layers 21--24 indicates that the information added to the residual stream by these layers is disproportionately important for temporal preference.
The necessity curve shows a more gradual rise beginning around layer 19, suggesting that while individual layers beyond the peak contribute less, their cumulative disruption is meaningful.
The elevated variance in the peak region indicates that different contrastive pairs engage these layers to different degrees, consistent with the parametric variation in the experimental design.

\paragraph{Layer 19 as onset.}
Layer 19 deserves attention.
In the single-pair case study (\ref{app:case-study-hf}), denoising recovery jumps from $\sim$0.05 at L18 to $\sim$0.5 at L19, the first layer where patching produces a measurable effect on the output.
Before L19, the residual stream does not yet encode temporal preference in a form that patching can recover.
This onset coincides with the beginning of the steering sweet spot (layers 19--22; \ref{app:contrastive-steering}): the model can be steered at L19 precisely because the temporal computation is just beginning and the representation is still malleable.
By L26 (the probing peak; \ref{app:contrastive-probing-linear}), the computation is complete and the representation is readable but no longer easy to redirect.
The L19 onset, L21--24 peak, and L26 readout form a coherent computational timeline within the subgraph.

\FloatBarrier

\subsection{Redundancy gap heatmap and layer--component interaction}

\begin{figure}[!htbp]
    \centering
    \begin{minipage}[t]{0.48\textwidth}
        \centering
        \includegraphics[width=\textwidth]{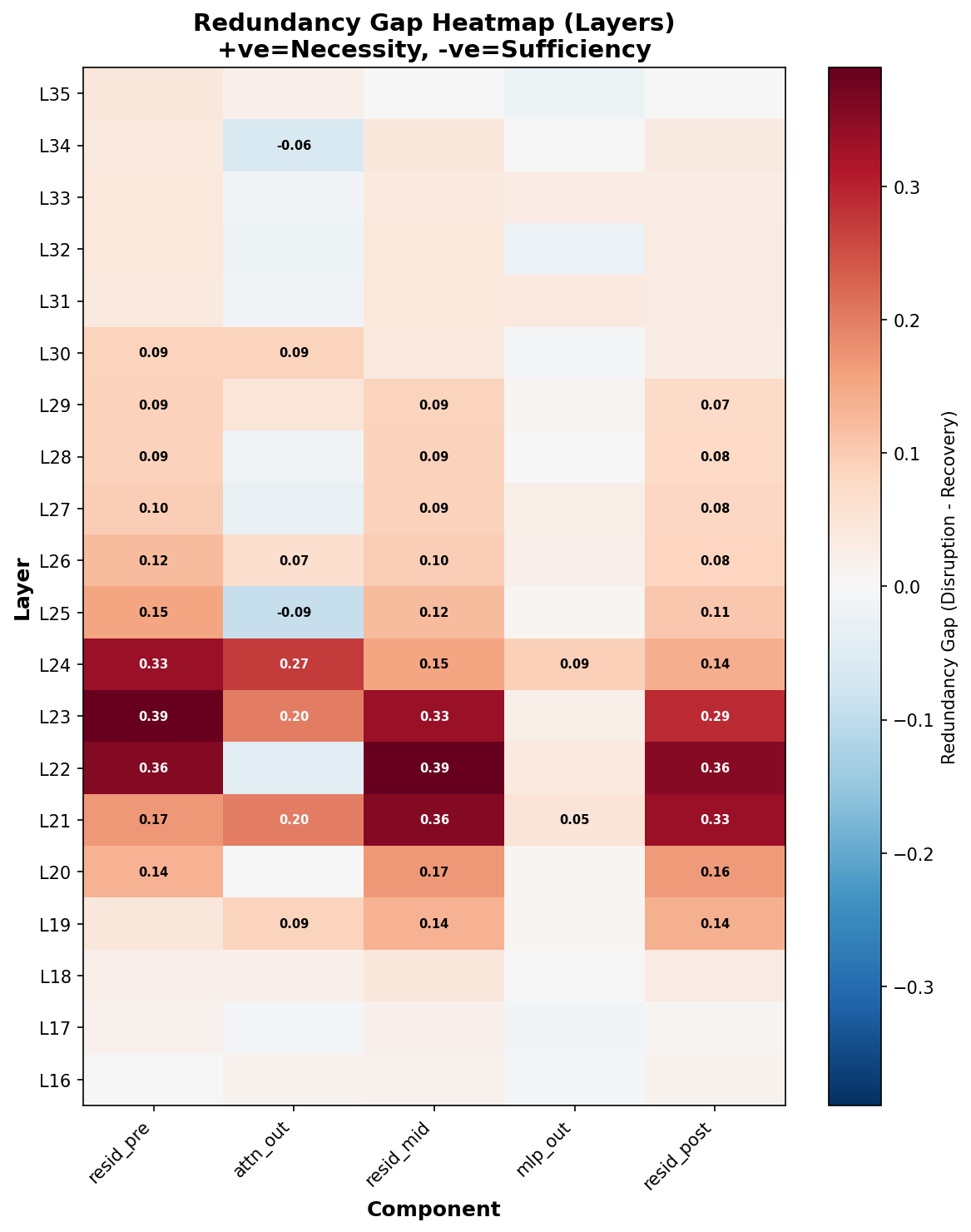}
        \subcaption{
            Redundancy gap heatmap (disruption $-$ recovery) by layer and component type.
            Strong positive values (dark red) in layers 20--23 across \texttt{resid\_pre}, \texttt{resid\_mid}, and \texttt{resid\_post} indicate high necessity with low sufficiency, characteristic of components embedded in a redundant processing pipeline where no single intervention can fully restore behavior.
            The \texttt{attn\_out} column shows a localized peak at L24 (0.39), while \texttt{mlp\_out} values remain relatively low throughout, suggesting MLP contributions are less redundantly encoded.
        }
        \label{fig:difference_heatmap_layer}
    \end{minipage}%
    \hfill
    \begin{minipage}[t]{0.48\textwidth}
        \centering
        \includegraphics[width=\textwidth]{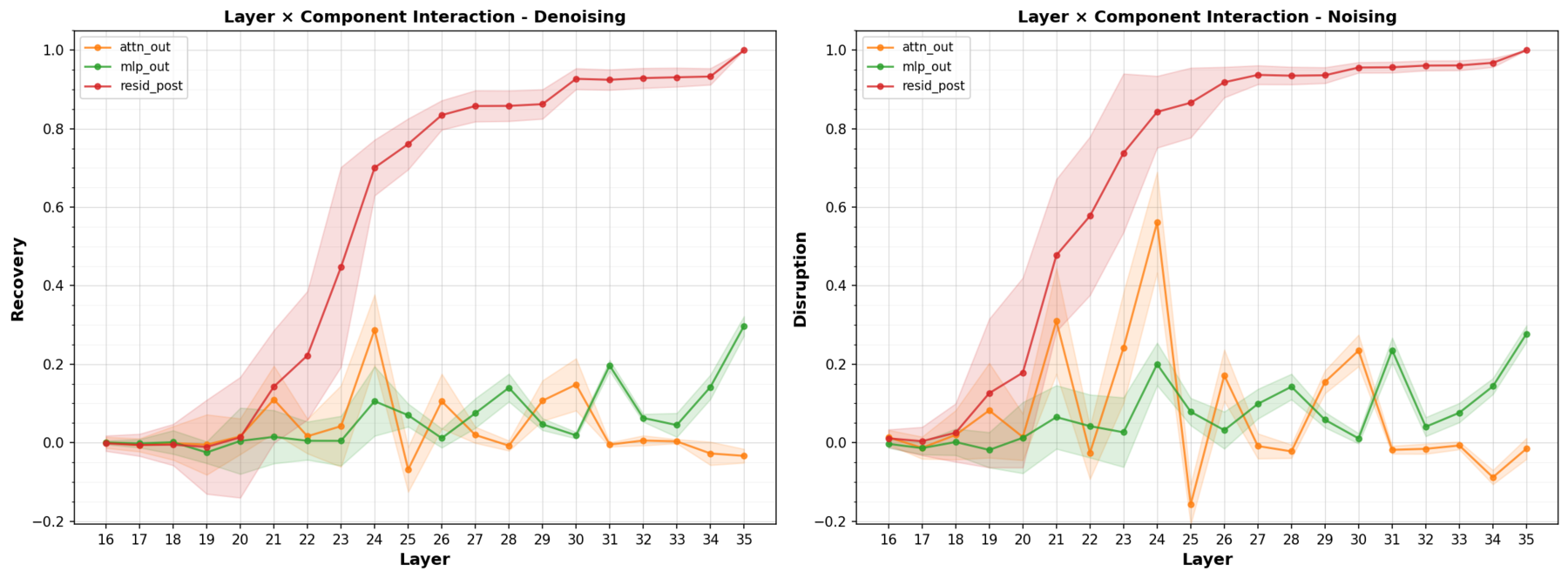}
        \subcaption{
            Layer-by-layer causal effect for \texttt{attn\_out}, \texttt{mlp\_out}, and \texttt{resid\_post} under denoising (left) and noising (right).
            The \texttt{resid\_post} curve rises sharply at layer 20 under denoising and saturates near 1.0 by layer 24, reflecting the cumulative nature of residual stream patching.
            The \texttt{attn\_out} and \texttt{mlp\_out} traces show complementary peaks: attention peaks at layers 21--24, while MLP contributions are more distributed across layers 22--35.
        }
        \label{fig:layer_component_interaction}
    \end{minipage}
    \caption{Redundancy gap heatmap and layer--component interaction analysis.}
    \label{fig:heatmap_and_interaction}
\end{figure}

Figure~\ref{fig:difference_heatmap_layer} maps the redundancy gap, defined as noising disruption minus denoising recovery, across all layers and component types.
Positive values (red) indicate components that are more necessary than sufficient, while negative values (blue) indicate components that are more sufficient than necessary.
Figure~\ref{fig:layer_component_interaction} decomposes the layer sweep into separate traces for attention, MLP, and residual stream components, revealing how each component type's causal effect varies across layers.

The heatmap reveals a striking pattern: layers 20--23 show large positive redundancy gaps across nearly all component types, peaking near 0.39 for the residual stream components (\texttt{resid\_pre} L23, \texttt{resid\_mid} L22).
This indicates that these layers are deeply embedded in the temporal preference circuit: corrupting them causes severe disruption, but patching in clean activations at only one component is insufficient for full recovery, because the corrupted signal has already propagated through earlier residual connections.
The \texttt{attn\_out} component at L25 shows a mildly negative gap ($-0.11$), making it one of the few components where recovery exceeds disruption, suggesting a degree of self-contained sufficiency at that layer.

Several patterns emerge from this decomposition.
Under denoising, the residual stream curve exhibits a characteristic sigmoid shape, rising steeply between layers 19 and 24 and then plateauing near full recovery.
This reflects the cumulative nature of the residual stream: once the critical mid-layer representations are restored, later layers can process them correctly.
The \texttt{attn\_out} component shows a pronounced peak at layers 21--24 under both denoising and noising, consistent with the component ranking in Figure~\ref{fig:component_importance_ranked}.
MLP contributions, by contrast, are more distributed: under noising, \texttt{mlp\_out} shows elevated disruption across a broad range of late layers (25--35), suggesting that MLP components contribute through distributed, incremental processing rather than a single localized intervention.

\FloatBarrier

\subsection{Noise vs.\ denoise and attention vs.\ MLP scatterplots}

\begin{figure}[!htbp]
    \centering
    \begin{minipage}[t]{0.48\textwidth}
        \centering
        \includegraphics[width=\textwidth]{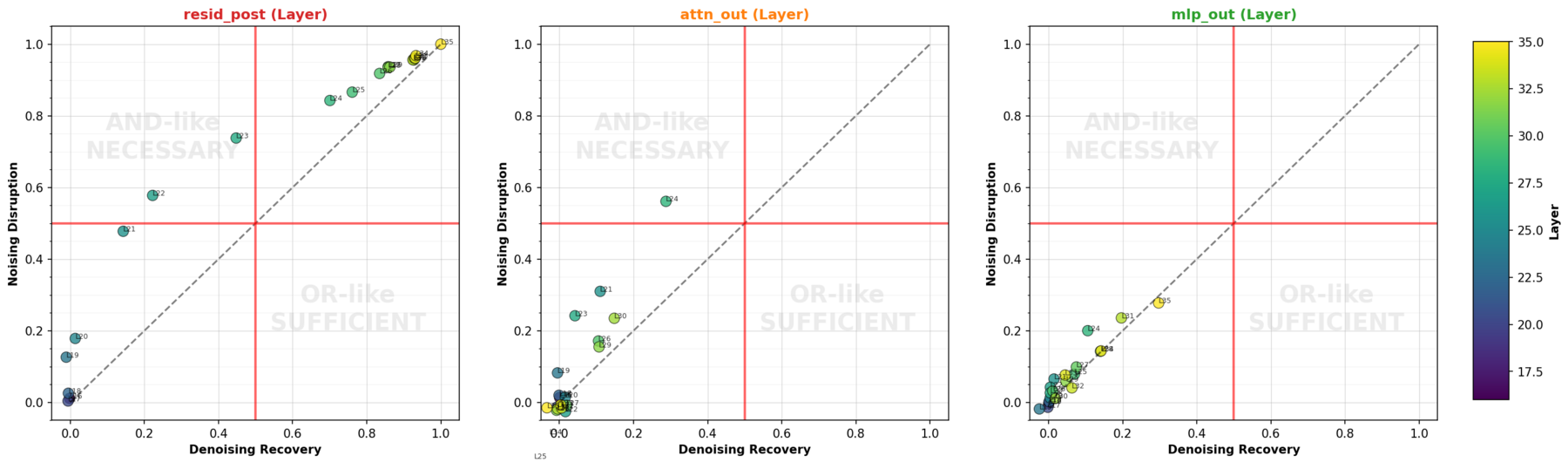}
        \subcaption{
            Denoising recovery vs.\ noising disruption for each layer, separated by component type (\texttt{resid\_post}, \texttt{attn\_out}, \texttt{mlp\_out}).
            Points are colored by layer number.
            The quadrant labels indicate the interpretive regime: ``AND-like / necessary'' (upper left) for high disruption with low recovery, and ``OR-like / sufficient'' (lower right) for high recovery with low disruption.
            Most \texttt{resid\_post} layers cluster in the necessary quadrant, while \texttt{attn\_out} and \texttt{mlp\_out} show more varied profiles.
        }
        \label{fig:noise_vs_denoise_per_component_layer}
    \end{minipage}%
    \hfill
    \begin{minipage}[t]{0.48\textwidth}
        \centering
        \includegraphics[width=\textwidth]{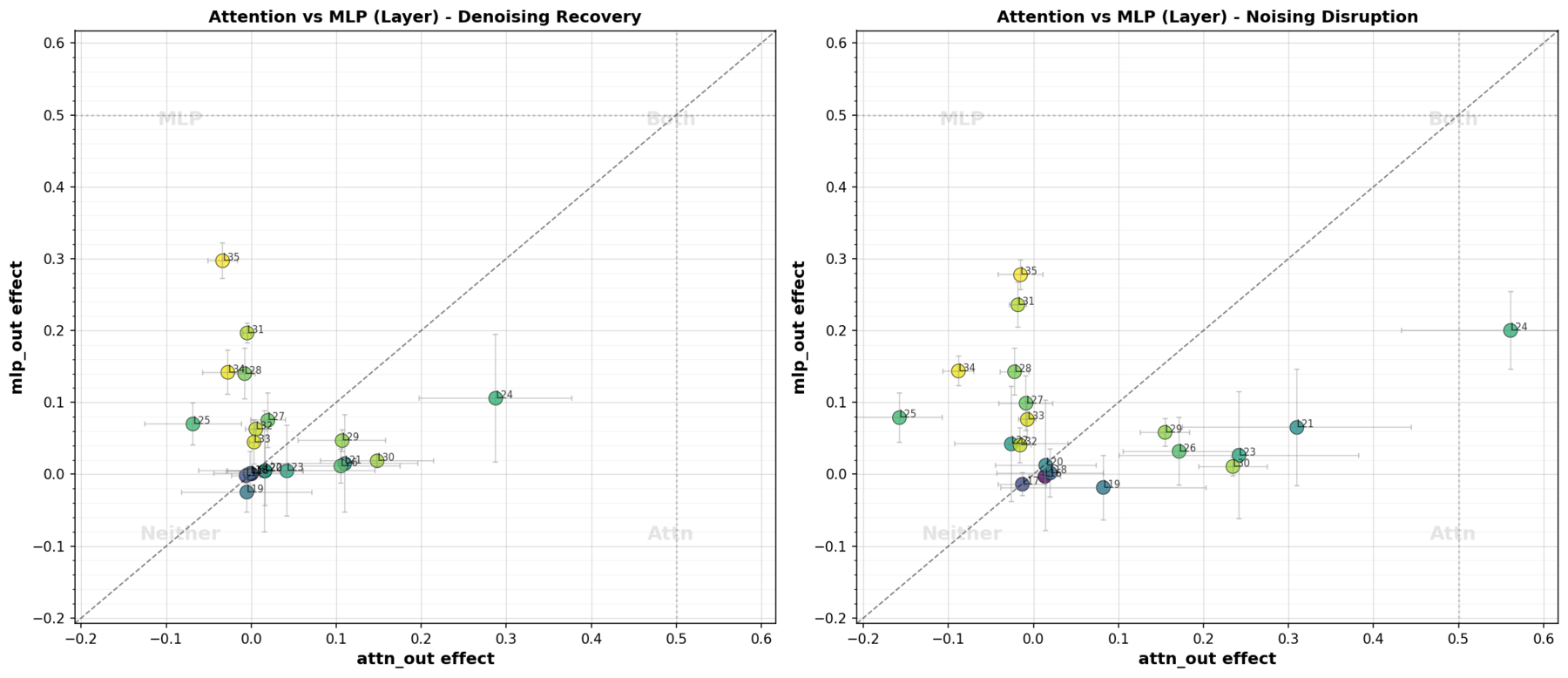}
        \subcaption{
            Attention vs.\ MLP effect size at each layer under denoising recovery (left) and noising disruption (right).
            Points above the diagonal indicate MLP-dominant layers; points below indicate attention-dominant layers.
            Under both metrics, mid-layer points (L21--L24) fall well below the diagonal, confirming that attention drives the largest single-component effects for temporal preference.
            Late layers (L31--L35) cluster near or above the diagonal under noising, reflecting the distributed MLP contributions in this range.
        }
        \label{fig:attn_vs_mlp_layer}
    \end{minipage}
    \caption{Noise vs.\ denoise scatterplots and attention vs.\ MLP comparison.}
    \label{fig:scatter_and_attn_mlp}
\end{figure}

Figure~\ref{fig:noise_vs_denoise_per_component_layer} plots each layer's denoising recovery against its noising disruption, separately for each component type.
This reveals whether components are sufficient (high recovery, low disruption), necessary (low recovery, high disruption), or both.
Figure~\ref{fig:attn_vs_mlp_layer} directly compares attention and MLP contributions at each layer, revealing the relative dominance of each component type.

The scatterplots confirm the redundancy gap analysis from a different angle.
For \texttt{resid\_post}, the picture is layer-dependent: L21 and L22 sit in the upper-left ``necessary'' quadrant (high disruption, modest recovery), L23 sits at the boundary, and L24--L25 move into the upper-right region (both high disruption \emph{and} high recovery), with L26 onwards saturating near the top-right corner.
Layers below L21 (L19--L20) remain in the low-effect region near the origin.
This pattern is consistent with a redundantly encoded signal in the L21--L23 range that cannot be fully restored by a single-layer intervention, transitioning at L24 onwards into a regime where single-layer patches both disrupt and recover the temporal signal.
For \texttt{attn\_out}, the highest-layer points (around L24) fall near the diagonal, indicating roughly balanced sufficiency and necessity.
The \texttt{mlp\_out} panel shows most layers near the origin, with a few late layers (L31, L35) reaching moderate effect sizes in both directions, consistent with their role as the top-ranked MLP components.

\begin{figure}[!htbp]
    \centering
    \includegraphics[width=0.7\textwidth]{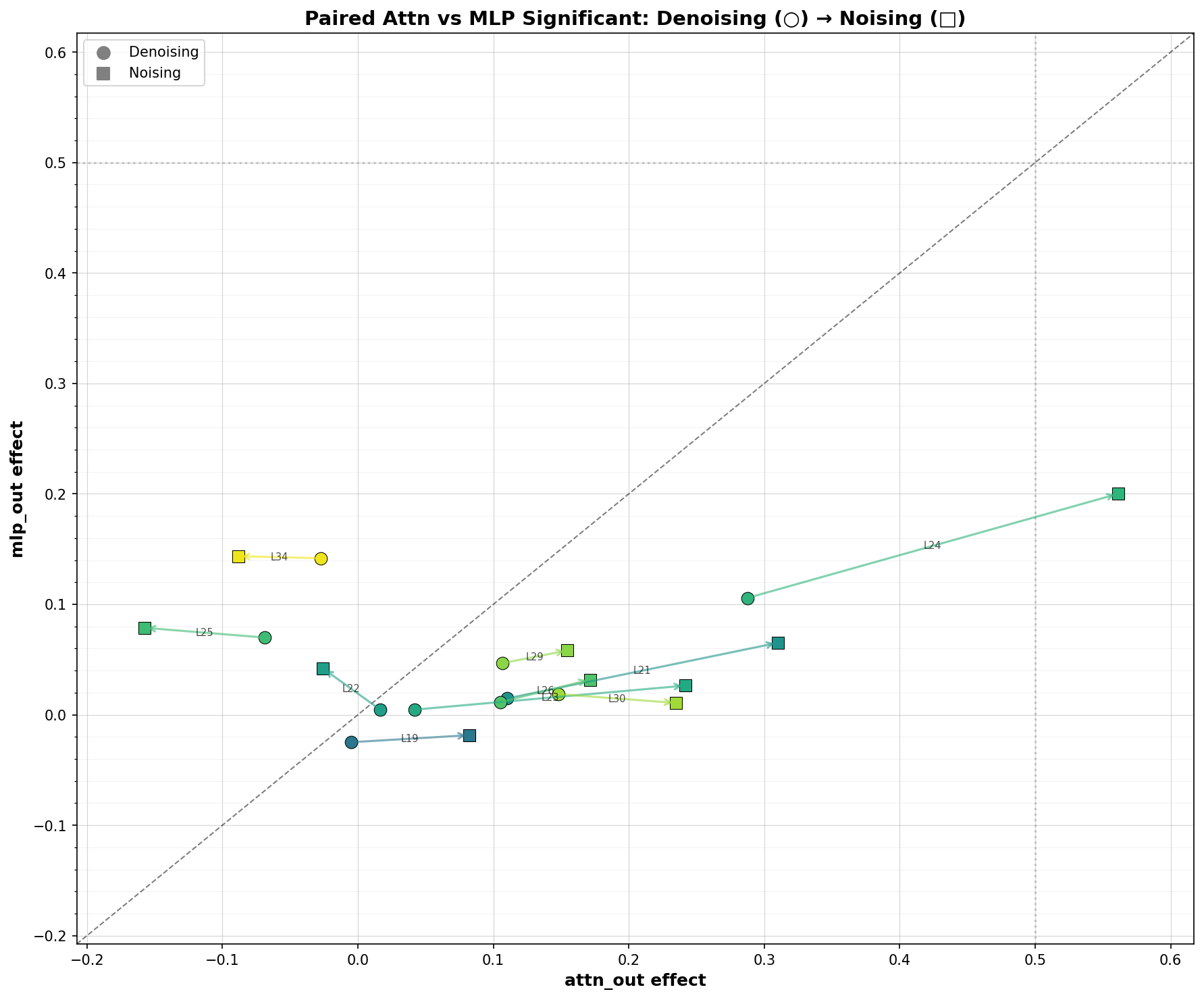}
    \caption{
        Paired attention vs.\ MLP comparison with arrows connecting each layer's denoising (circle) and noising (square) scores.
        Layers where the arrow points rightward and downward (e.g., L21, L24) indicate components where noising reveals much stronger attention dominance than denoising.
        The trajectories of L24 and L21 show the largest rightward displacement, confirming these attention heads as the most causally important individual components.
        Layers L22 and L34 shift upward, reflecting MLP-dominant noising effects at those layers.
    }
    \label{fig:attn_vs_mlp_paired}
\end{figure}

Figure~\ref{fig:attn_vs_mlp_paired} provides a paired view directly comparing attention and MLP contributions at each layer.

The attention-vs-MLP comparison reveals a consistent pattern: in the critical mid-layer range (L21--L24), attention components have substantially larger causal effects than their MLP counterparts.
This asymmetry is especially pronounced under noising disruption, where \texttt{L24\_attn} and \texttt{L21\_attn} achieve effect sizes of 0.5--0.6 while the corresponding MLP components remain below 0.2.
The paired plot (Figure~\ref{fig:attn_vs_mlp_paired}) makes this particularly clear: the arrows for L21 and L24 sweep dramatically to the right as we move from denoising to noising, indicating that these attention components become even more dominant when measuring necessity rather than sufficiency.
In contrast, some layers (L22 in the mid-range, L34 in the upper layers) show arrows pointing upward, indicating that their MLP components are more causally important than their attention components, particularly under noising.

\FloatBarrier

\subsection{Summary}
\label{app:causal-parametric-summary}

The activation patching results converge on several findings that support the claims in the main text:

\begin{enumerate}[leftmargin=*]
    \item \textbf{Sparse, localized circuit.}
    A small number of components account for the majority of causal effect on temporal preference.
    The top four components (\texttt{L24\_attn}, \texttt{L21\_attn}, \texttt{L35\_mlp}, and \texttt{L31\_mlp}) are clearly separated from the remaining components in effect size (Figure~\ref{fig:component_importance_ranked}).

    \item \textbf{Attention dominance in mid-layers.}
    Attention heads at layers 21 and 24 are the single most causally important components, particularly under noising disruption.
    Their effect sizes exceed those of any MLP component by a factor of 2--3x (Figures~\ref{fig:attn_vs_mlp_layer},~\ref{fig:attn_vs_mlp_paired}).

    \item \textbf{Distributed MLP contributions in late layers.}
    MLP components contribute through a more distributed pattern across layers 25--35, with \texttt{L35\_mlp} and \texttt{L31\_mlp} as the most prominent individual contributors (Figure~\ref{fig:layer_component_interaction}).

    \item \textbf{Necessity exceeds sufficiency.}
    The high-effect components show a consistent asymmetry: noising disruption exceeds denoising recovery, indicating redundant encoding where no single component is individually sufficient to fully determine temporal preference, but individual components are necessary in the sense that corrupting them substantially degrades performance (Figures~\ref{fig:difference_heatmap_layer},~\ref{fig:noise_vs_denoise_per_component_layer}).

    \item \textbf{Critical computation window at layers 20--24.}
    The marginal contribution analysis localizes the most informative residual stream transformations to a narrow five-layer window (Figure~\ref{fig:marginal_contribution_var_p}), consistent with a concentrated computational phase for temporal preference. Three of five layers are also identified as part of core decision window in activation patching for temporal classification (\ref{app:causal-contrastive}).
\end{enumerate}

\FloatBarrier

\clearpage
\clearappnumbering

\section{Causal classification results}\label{app:causal-contrastive}

The causal parametric results (\ref{app:causal-parametric}) intervene on highly-formatted parametric prompts and localize components causally important for temporal preference (valuation). Here we intervene on contrastive classification pairs to localize components causally important for temporal classification (categorical horizon inference). Convergence between the two pipelines would suggest that the same computational machinery is recruited for both tasks; whether this reflects a common temporal representation or merely a common decision readout requires further analysis we leave to future work.

The task design follows the IOI style: clean and corrupted prompts represent phrase beginnings awaiting completion with the tokens \textit{"short"} and \textit{"long"}.
Each sentence contains a description of a goal and a question about the time horizon of that goal.

\textbf{Example}
\begin{itemize}
        \item \textbf{Clean:} \textit{"The goal is to cook a warm dinner for the family. Is this a short-term or long-term goal? The answer is:"}. Expected clean answer \textit{"short"} as the next predicted token.
        \item \textbf{Corrupted:} \textit{"The goal is to become a top chef in the city. Is this a short-term or long-term goal? The answer is:"}. Expected corrupted answer \textit{"long"} as the next predicted token.
\end{itemize}

All prompts are appended with a chat template before being passed to the model.
The complete design of the dataset and the results of its validation are described in \ref{app:causal-contrastive-methods}.

\textbf{Since this pipeline primarily tests convergence with the other paradigms}, readers can skip directly to the convergence finding (\ref{find:cc-convergence}), with supporting evidence in Sec.~\ref{evidence:cc-resid-convergence} and Sec.~\ref{evidence:cc-alltokens-view-convergence}.

\subsection{Limitations}
We flag three limitations of this experiment before turning to its results.

\begin{itemize}[noitemsep, topsep=0pt, leftmargin=*]
    \item \textbf{Vocabulary–horizon entanglement.} Long-term prompts
    concentrate in achievement framing: Career/Mastery cues account for
    46\% of pairs (Table~\ref{tab:cue_types_stat_160}), and prestige
    markers (\textit{top}, \textit{professional}, \textit{master}) appear
    more in long-term prompts than in short-term ones
    while the latter cluster in casual, domestic framings. The
    localized circuits may partially reflect achievement-vocabulary
    detection rather than temporal horizon per se. Two considerations bound this concern. First, the remaining 54\% of pairs use Growth and Accumulation cues, which carry distinct surface vocabulary (size/scale markers and exhaustive-scope markers, respectively) rather than the prestige register that drives the Career/Mastery cluster. Second, and more directly, the same dominant attention component \texttt{(L24\_attn)} also emerges as the top-ranked component in the parametric pipeline (\ref{app:causal-parametric}), where the surface vocabulary consists of numerical reward amounts and explicit time horizons with no overlap with the achievement-prestige register identified here. A circuit whose function is to detect achievement vocabulary would not be expected to dominate causal effect on prompts of the form ``\$20K in 6 months vs.\ \$500K in 10 years''. We nevertheless flag two direct minimal-pair controls as the cleanest tests: explicit-temporal-marker pairs (e.g., ``fix the bike before lunch'' vs.\ ``fix old bikes over many years'') and same-class patching (short$\to$short, long$\to$long). We leave these to future work.

    \item \textbf{Selection bias.} We patch only on the 160 pairs that
    \texttt{Qwen3-4B-Instruct-2507} already classifies correctly,
    excluding 40 misclassified ones. 15 of those are clear-signal failures
    that concentrate in accumulation and growth cues. The identified
    circuit may therefore reflect the successful, Career/Mastery-dominated
    classification path rather than the model's general temporal-reasoning
    capability.

    \item \textbf{Single template, binary horizon.} All 320 prompts share
    one template (\textit{``The goal is to $\langle$X$\rangle$. Is this
    a $\langle$Y$\rangle$ goal? The answer is:''}) and a binary judgment,
    so the circuit cannot be tested for graded horizon sensitivity or
    template-independence within this paradigm. Convergence with the
    parametric setup (\ref{app:causal-parametric}) on attention layer
    partially mitigates the template concern.
\end{itemize}

\subsection{160-Pair Directional Patching}
We first present heatmaps for all tokens, highlighting that the end token accumulates the maximum patching effect.
We then provide per-layer plots at the final token position with 95\% confidence intervals. At the end of the results section, we also report effects of patching each layer simultaneously across all 34 token positions for \texttt{attn\_out} and \texttt{mlp\_out}. This enables direct comparison with the position-averaged attributions of EAP-IG (\ref{app:attributional-contrastive}) and the all-positions patching of causal parametric experiment (\ref{app:causal-parametric}), which intervenes on all positions jointly.

\subsubsection{Residual Stream}
\label{sec:cc-resid-pre-section}
\begin{figure}[htbp]
\centering
\begin{minipage}[c]{0.8\textwidth}
    \centering
    \includegraphics[width=\linewidth]{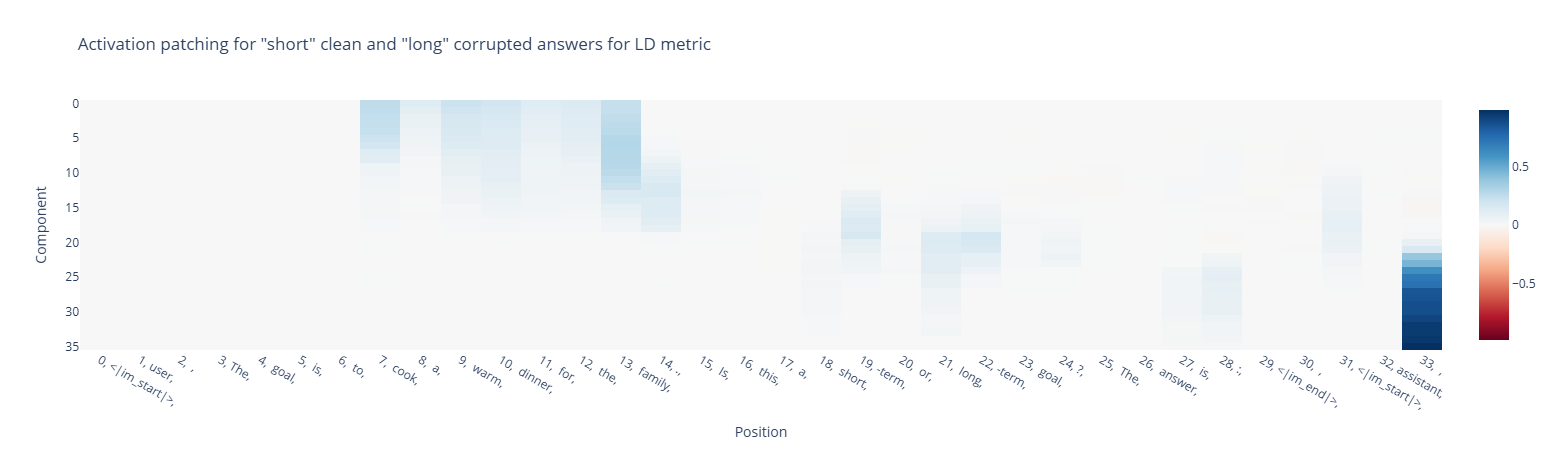}
    \vspace{4pt}
    \includegraphics[width=\linewidth]{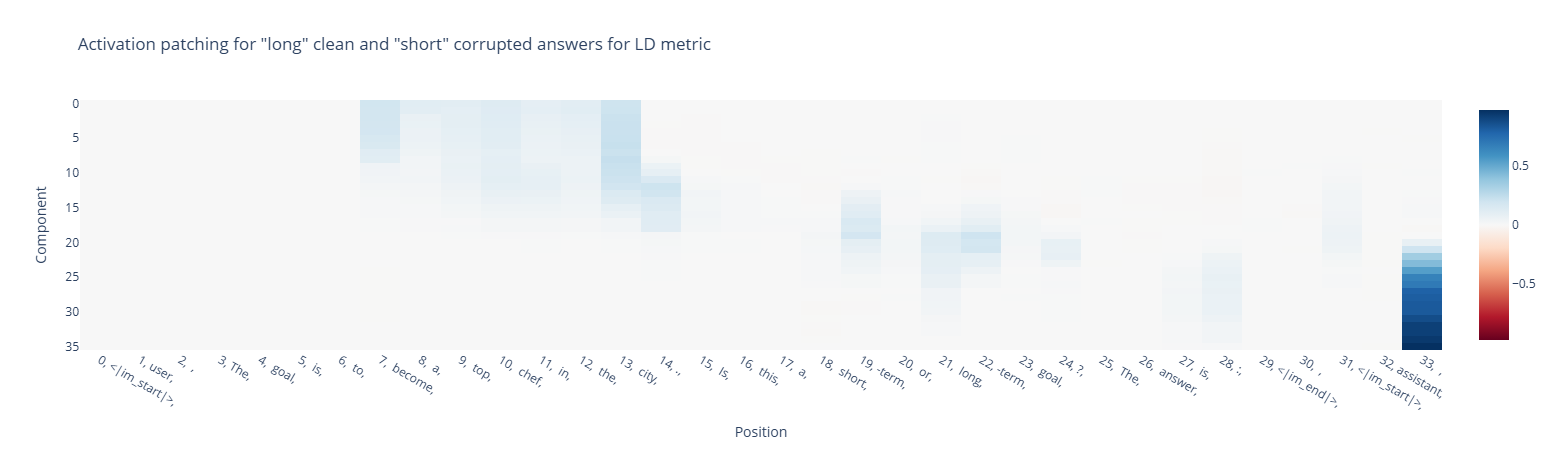}
\end{minipage}
\caption{Directional residual stream patching averaged over 160 classification pairs.
Top row: denoising for \textit{"short"} clean and \textit{"long"} corrupted; bottom row: denoising for \textit{"long"} clean and \textit{"short"} corrupted.}
\label{fig:cc-resid-160-heatmap}
\end{figure}

Figure~\ref{fig:cc-resid-160-heatmap} shows residual stream patching results for all 34 tokens of the prompt for two flip directions. We can see that both patterns are quite similar in the global structure. They show the same three activity bands:
\begin{itemize}
    \item early layers (L0--19) highlighting the goal statement tokens at positions 7--14: \textit{the model reading the cue};
    \item middle layers (L12--26) highlighting the temporal keywords (positions 18--22: \textit{``short/-term/or/long/-term''}): \textit{the question machinery};
    \item late layers (L20--35) concentrated on the end token (position 33): \textit{the decision}.
\end{itemize}
In both heatmaps the end column saturates the blue scale from $\sim$L20 downward and represents the location of maximum patching effect. The qualitative circuit ("goal-read $\to$ question-read $\to$ decide at end") is the same whether the model is being steered toward \textit{"short"} or \textit{"long"}.

\begin{figure}[htbp]
\centering
\begin{minipage}[c]{0.8\textwidth}
    \centering
    \includegraphics[width=\linewidth]{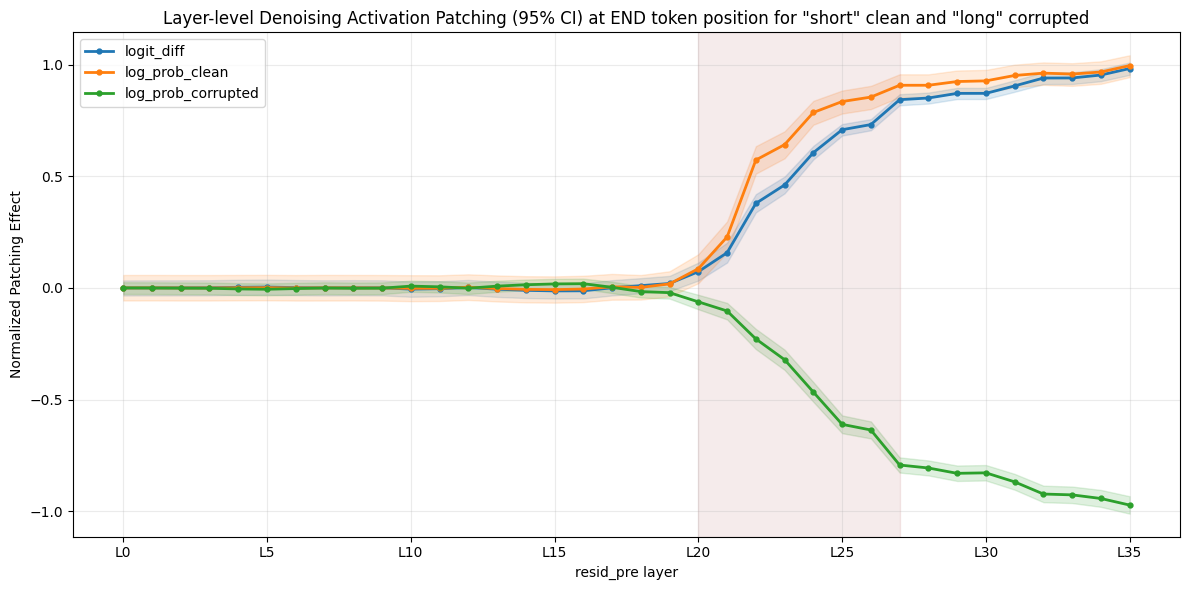}
    \vspace{4pt}
    \includegraphics[width=\linewidth]{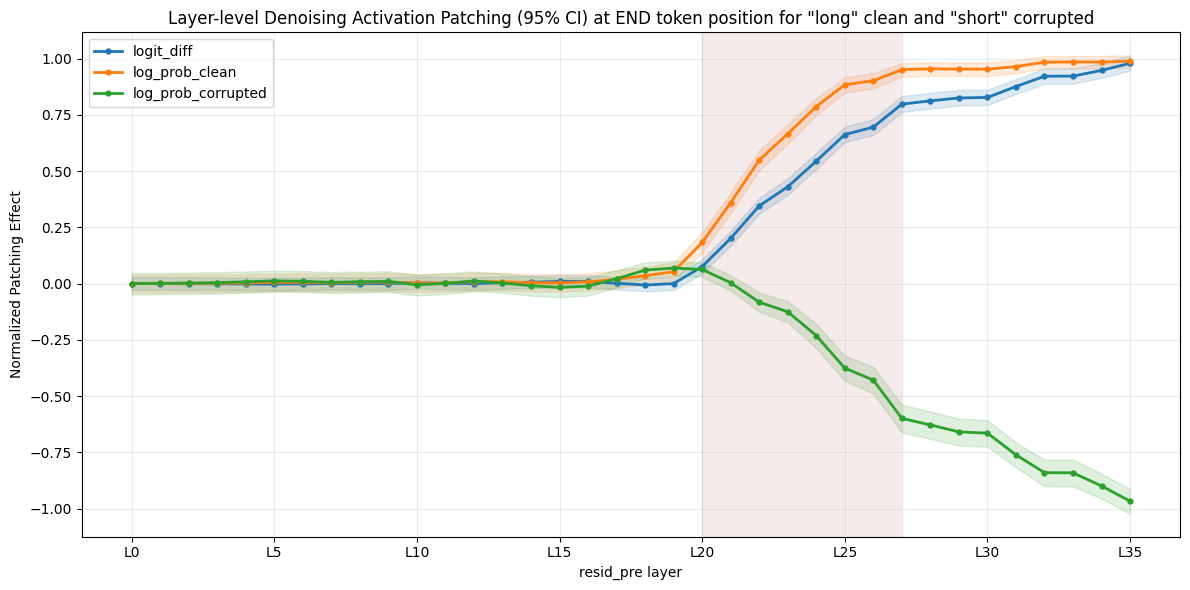}
\end{minipage}
\caption{Patching effects on residual stream at END token positions with highlighted decision window.
Top row: denoising for \textit{"short"} clean and \textit{"long"} corrupted; bottom row: denoising for \textit{"long"} clean and \textit{"short"} corrupted. Core decision window at L20-27 highlighted in red.}
\label{fig:cc-resid-160}
\end{figure}

\begin{figure}[htbp]
\centering
\begin{minipage}[c]{0.8\textwidth}
    \centering
    \includegraphics[width=\linewidth]{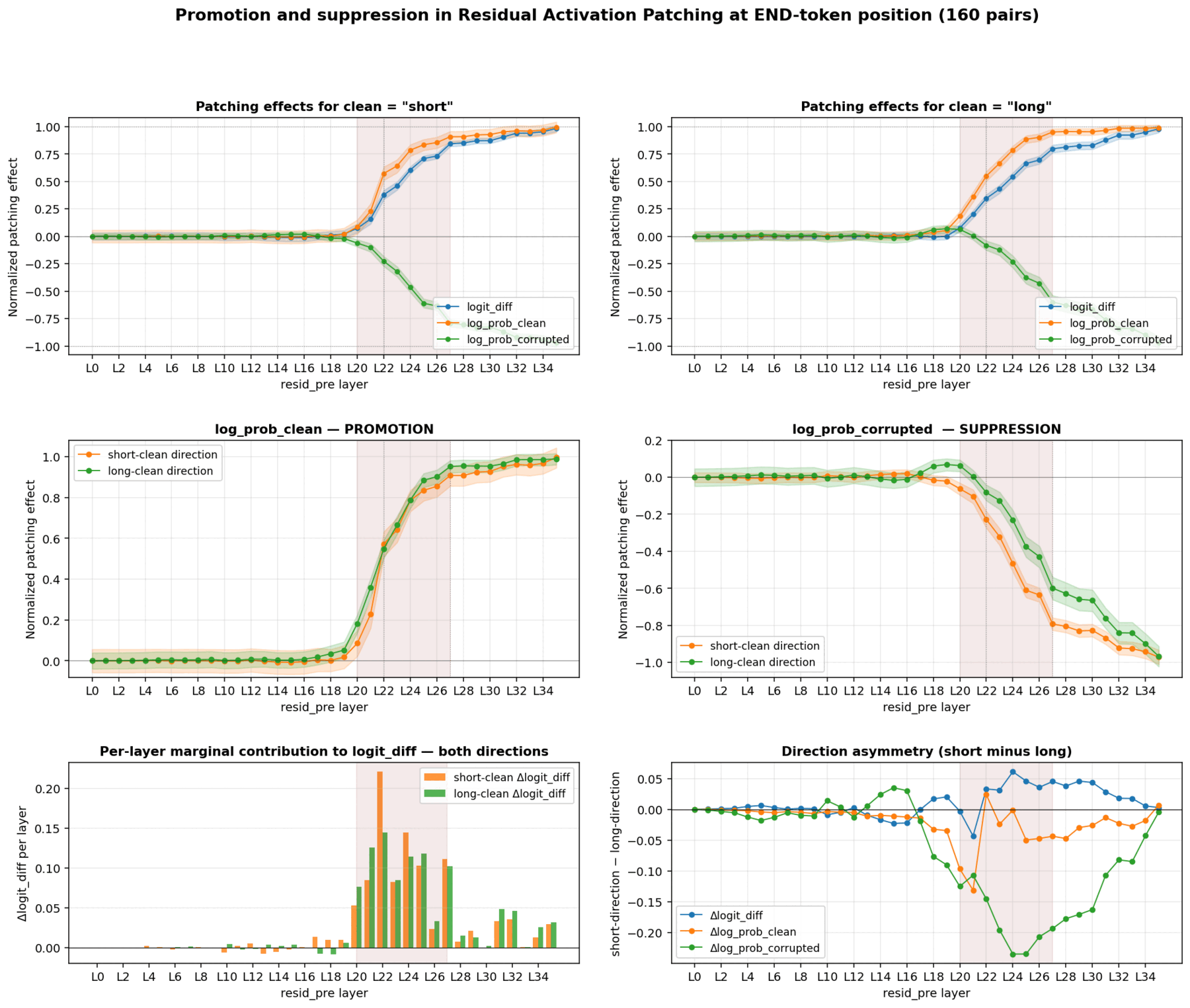}
\end{minipage}
\caption{Core decision window, almost symmetric promotion, markedly asymmetric suppression.}
\label{fig:cc-resid-pre-160-all-results}
\end{figure}

\paragraph{Three-region structure, core decision window at L20-27.}\label{evidence:cc-resid-convergence}
Figure~\ref{fig:cc-resid-160} shows residual stream patching results at final token positions for both directions.
We can observe the same three regions in the dynamics of cumulative denoising curves for both flips. Layers~0--19 are causally silent on logit difference: the confidence intervals include zero at every layer. Logarithmic probabilities show very small positive biases in some early layers that do not translate into a detectable LD effect. Layers~20--27 form a \emph{core decision window} during which the LD recovery rises from below 10\% to roughly 80\% of the clean baseline (84.3\% short-clean, 79.7\% long-clean at L27); this seven-layer staircase accounts for the vast majority of the patching effect. Layers~28--35 contribute a slow saturation tail, with small but consistent additional contributions around L31--L32.\\

Within the core window, the same five layers are the dominant contributors in both flips. Ranking layers by $\Delta\text{LD}$, the short-clean flip is led by layers $22, 24, 27, 25, 21$ and the long-clean flip by layers $22, 21, 25, 24, 27$. The sum of these five per-layer contributions accounts for 68\% of the full recovery in the short-clean flip and 62\% in the long-clean flip, while cumulative recovery at the end of the seven-layer window (L27) reaches 84\% and 80\% respectively. Layer~22 alone is the largest single contributor in both flips, with $\Delta\text{LD} = +0.221$ in the short-clean flip and $+0.144$ in the long-clean flip.
This is consistent with the critical computation window at layers L20--24 identified in the causal parametric experiment (\ref{app:causal-parametric}), within which three of our dominant contributors (L21, L22, and L24) fall.

\paragraph{Nearly symmetric promotion, asymmetric suppression.}
The two flips recover the clean answer at almost the same speed: the logarithmic probability curves of clean answers overlap within their bounds across all 36 layers, with direction-wise differences of at most $0.15$ anywhere in the domain (Fig.~\ref{fig:cc-resid-pre-160-all-results}, middle left and bottom right). Suppression of the corrupted answer, however, is markedly faster in the short-clean flip than in the long-clean flip. At layer~24, the normalized suppression is $-0.466$ in the short-clean flip versus $-0.231$ in the long-clean flip: a two-fold gap. The absolute direction difference $|\text{short} - \text{long}|$ on the logarithmic probability of corrupted answer peaks at $0.23$ across layers~24--25 and decays monotonically, reaching $0.08$ by layer~32 and closing to within $0.01$ at layer~35. This peak gap is substantially larger than the peak direction differences on LD ($0.06$ at L24) and on logarithmic probability of clean answer ($0.13$ at L21) (Fig.~\ref{fig:cc-resid-pre-160-all-results}, bottom right). The model produces the two answers with nearly identical efficiency but requires additional late-layer computation to push \textit{short} down when the correct answer is \textit{long}. This mechanical asymmetry is consistent with the behavioral bias observed during dataset construction, in which all 15 clear-signal misclassifications were false-\textit{short} predictions.
 
\paragraph{Milestone layers.}
Table~\ref{tab:resid-pre-milestones} summarizes the first layer at which the mean patching effect crosses a given magnitude threshold, for each metric and each flip.  Onsets ($|\text{effect}|\!\ge\!0.05$) occur within a narrow two-layer window, at L20 on LD in both flips, at L20 (short-clean) and L19 (long-clean) on clean logarithmic probability, and at L20 (short-clean) and L18 (long-clean) on corrupted logarithmic probability.
The long-clean flip reaches onset 1--2 layers earlier than the short-clean flip on both log-probability metrics, but falls progressively
behind at higher thresholds. Above the onset, LD and clean logarithmic probability milestones coincide across flips to within one layer at every threshold, whereas corrupted logarithmic probability milestones diverge: reaching 50\%, 75\%, and 90\% of the full suppression requires layers 25/27/32 in the short-clean flip but 27/31/35 in the long-clean flip, a delay of 2--4 layers at each threshold.

\begin{table}[h]
\centering
\small
\setlength{\tabcolsep}{5pt}
\begin{tabular}{c|cc|cc|cc}
\toprule
Threshold & \multicolumn{2}{c|}{LD} &
            \multicolumn{2}{c|}{log-$P(\text{clean})$} &
            \multicolumn{2}{c}{log-$P(\text{corr})$} \\
$|\text{effect}|$ & short & long & short & long & short & long \\
\midrule
0.05 & 20 & 20 & 20 & 19 & 20 & 18 \\
0.10 & 21 & 21 & 21 & 20 & 21 & 23 \\
0.25 & 22 & 22 & 22 & 21 & 23 & 25 \\
0.50 & 24 & 24 & 22 & 22 & \textbf{25} & \textbf{27} \\
0.75 & 27 & 27 & 24 & 24 & \textbf{27} & \textbf{31} \\
0.90 & 31 & 32 & 27 & 26 & \textbf{32} & \textbf{35} \\
\bottomrule
\end{tabular}
\caption{First \texttt{resid\_pre} layer at which the mean patching effect crosses each magnitude threshold, at the end token position, for the short-clean and long-clean flips.  \textbf{Bold entries} highlight the 50\%, 75\%, and 90\% log-$P(\text{corr})$ milestones discussed in the text, where the
short-clean flip reaches each threshold 2--4 layers before the long-clean flip.}
\label{tab:resid-pre-milestones}
\end{table}
 
\paragraph{Summary.}
The causal effect on temporal classification at the final token concentrates in \texttt{resid\_pre} layers~20--27, which together account for 83.8\% (short-clean) and 81.5\% (long-clean) of the full recovery, with layer~22 the single largest contributor in both flips and layers~21, 24, 25, and 27 together accounting for over half of the remaining recovery. Layers~0--19 have no causal effect at this position, and layers~28--35 contribute a smaller late-stage refinement (${\sim}14$--$19\%$ of the total recovery). The computation is nearly symmetric in clean-answer promotion (peak direction difference $0.13$ on log-$P(\text{clean})$) and markedly asymmetric in corrupted-answer suppression (peak direction difference $0.23$ on log-$P(\text{corr})$): suppressing \textit{long} when the answer is \textit{short} reaches $|\text{effect}|\!\ge\!0.90$ by layer~32, whereas suppressing \textit{short} when the answer is \textit{long} does not reach the same threshold until the final layer (L35). This residual-stream asymmetry parallels the short-biased behavioral errors observed during dataset construction (all 15 clear-signal misclassifications were false-\textit{short} predictions), though establishing a causal link between the two would require head-level or lens-based analysis.

\subsection{Attention-output patching at END token}
\label{sec:cc-attn-out-section}
 
\begin{figure}[htbp]
\centering
\begin{minipage}[c]{0.8\textwidth}
    \centering
    \includegraphics[width=\linewidth]{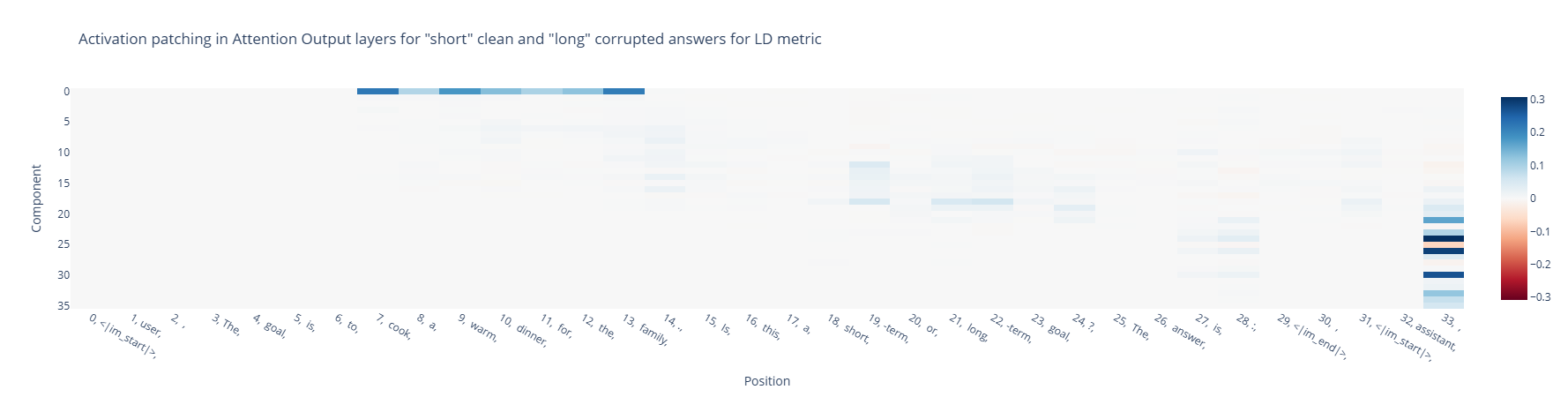}
    \vspace{4pt}
    \includegraphics[width=\linewidth]{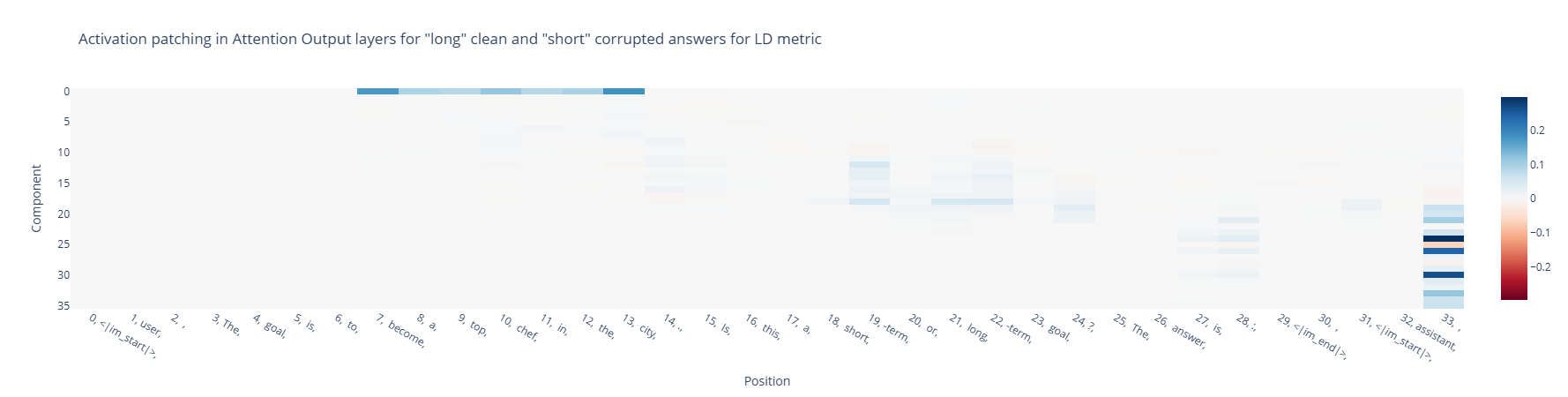}
\end{minipage}
\caption{Directional attention output patching averaged over 160 classification pairs.
Top row: denoising for \textit{"short"} clean and \textit{"long"} corrupted; bottom row: denoising for \textit{"long"} clean and \textit{"short"} corrupted.}
\label{fig:cc-attn-out-160-heatmap}
\end{figure}

\begin{figure}[htbp]
\centering
\begin{minipage}[c]{0.8\textwidth}
    \centering
    \includegraphics[width=\linewidth]{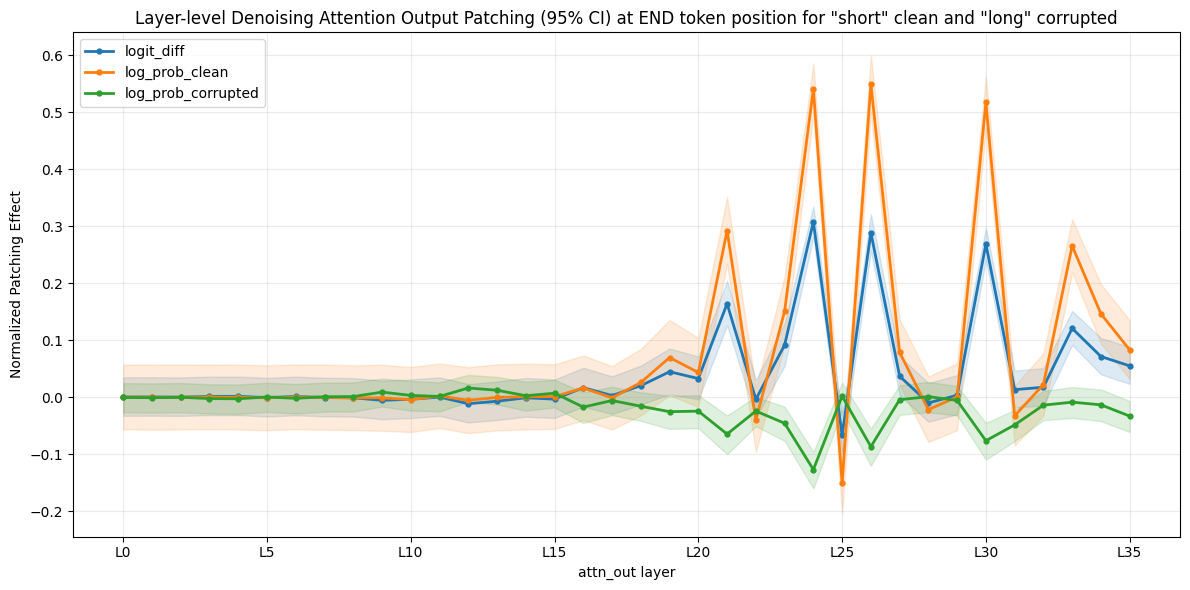}
    \vspace{4pt}
    \includegraphics[width=\linewidth]{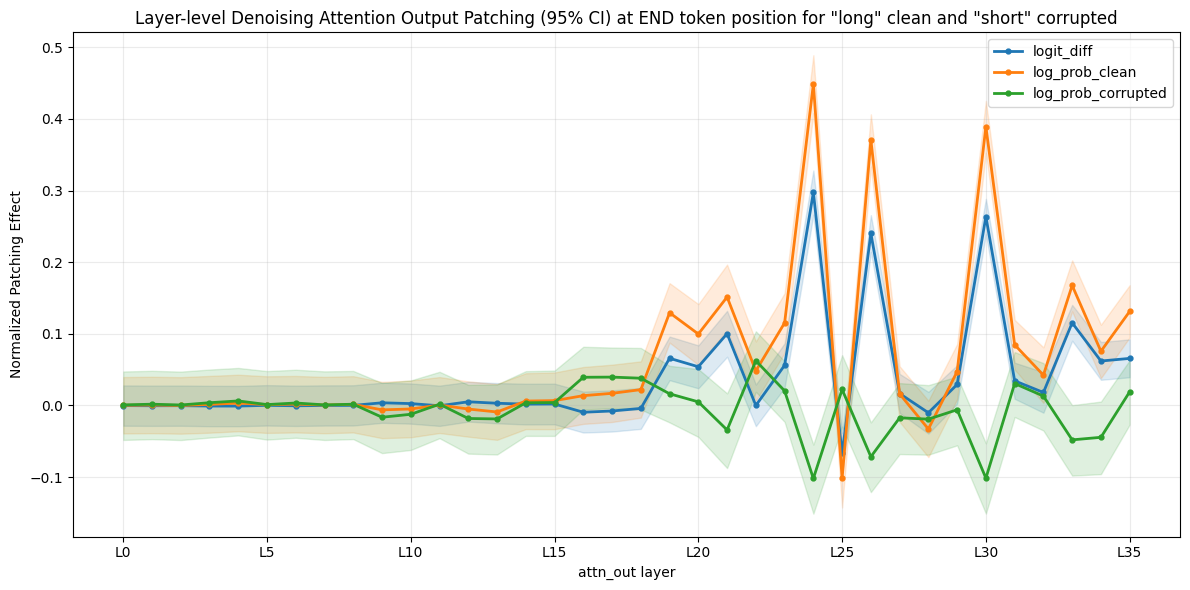}
\end{minipage}
\caption{Patching effects on attention output at END token positions.
Top row: denoising for \textit{"short"} clean and \textit{"long"} corrupted; bottom row: denoising for \textit{"long"} clean and \textit{"short"} corrupted.}
\label{fig:cc-attn-out-160}
\end{figure}

To localize the residual-stream effect to a specific component, we apply denoising activation patching to the per-layer attention-output hook (\texttt{attn\_out}) first at all token positions and then at the final one. Each patch replaces a single (layer, pos) summed attention output with its clean-run counterpart, isolating the contribution of that layer's attention at given position independent of MLPs and residual pass-through. We present the results for all tokens on Figure~\ref{fig:cc-attn-out-160-heatmap} and for the final token on Figure~\ref{fig:cc-attn-out-160}. Since we are primarily interested in interpreting the behavior of attention output flow in the END token, all the results described will concern only it.

\paragraph{Attention writes the decision at a sparse set of layers with L24 being the most dominant.} Whereas the cumulative \texttt{resid\_pre} curve rises as a smooth seven-layer staircase across L20--27 (Sec.~\ref{sec:cc-resid-pre-section}), per-layer attention effects are sharply localized to four dominant writer layers: L21, L24, L26, and L30. Each of these produces a significant positive LD effect in both flips (confidence intervals bounded away from zero), and the three strongest writers alone (L24, L26, L30) each contribute roughly a third of the full normalized recovery individually. A pair of smaller late writers at L33--34 rounds out the positive contribution, while the first fifteen layers produce no significant attention effect on any metric.

\paragraph{Attention promotes the correct answer but barely suppresses the incorrect one.} At every dominant attention writer, the effect on log-$P(\text{clean})$ is several times larger than the effect on log-$P(\text{corr})$: the per-layer promotion-to-suppression ratio exceeds four everywhere among L21, L24, L26, L30 and reaches six or more at the deeper writers in the short-clean flip. This is a component-level observation that the residual-stream totals alone cannot reveal, because at \texttt{resid\_pre} both metrics eventually approach unit magnitude. The implication is that attention at the decision layers implements primarily \emph{answer-promotion}: it writes ``the correct answer is here'' into the residual stream, with only a modest side effect on the competing answer. The deep suppression observed at \texttt{resid\_pre} (where log-$P(\text{corr})$ saturates near $-1$ by the final layers) must therefore come largely from a different component, most plausibly the MLPs, although this wasn't confirmed by our MLP patching analysis (Sec.~\ref{sec:cc-mlp-out-section}).
 
\paragraph{The direction asymmetry largely disappears at the attention level.}
A central finding of the \texttt{resid\_pre} analysis was that corrupted-answer suppression is faster when the correct answer is \textit{short} than when it is \textit{long}, with a peak direction gap of $0.23$ on log-$P(\text{corr})$. At \texttt{attn\_out}, this pattern is absent: attention effects on the corrupted answer are nearly equal across flips at every dominant writer. The largest remaining direction difference shifts to \emph{promotion} rather than suppression: the short-clean flip writes a noticeably stronger clean signal at L26 than the long-clean flip does, but even this residual asymmetry is well under half the size of the one observed at \texttt{resid\_pre}. Taken together, these two facts indicate that the direction-asymmetric late-layer suppression seen in the residual stream does not originate in the attention blocks.
 
\paragraph{Summary.}
The \texttt{resid\_pre} decision window resolves, at the attention-output level, into four primary writer layers (L21, L24, L26, L30). These attention blocks are strongly biased toward promotion of the correct answer rather than suppression of the incorrect one, and they behave nearly symmetrically across the two answer directions. The direction-asymmetric late-layer suppression observed at \texttt{resid\_pre} is therefore attributable to a different component, which matching MLP-output patching should be able to identify.
 
\FloatBarrier

\subsection{MLP-output patching at END token}
\label{sec:cc-mlp-out-section}
 
\begin{figure}[htbp]
\centering
\begin{minipage}[c]{0.8\textwidth}
    \centering
    \includegraphics[width=\linewidth]{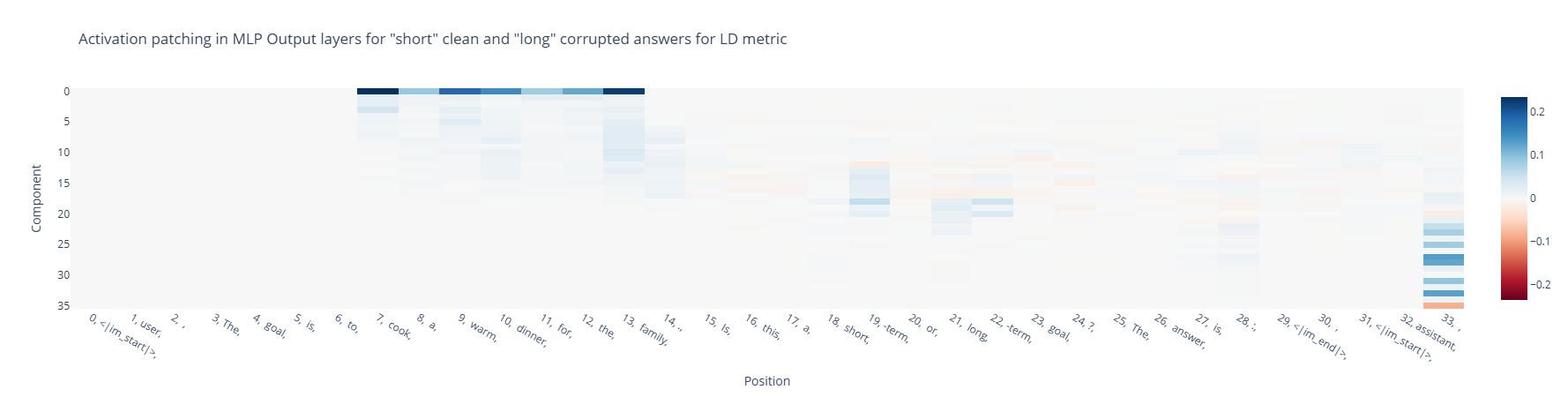}
    \vspace{4pt}
    \includegraphics[width=\linewidth]{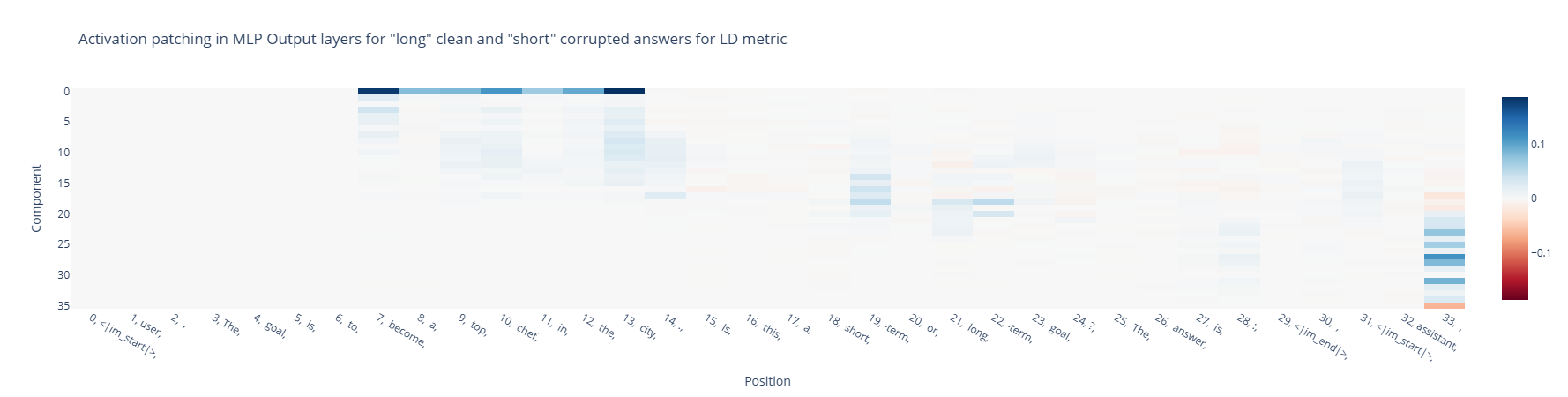}
\end{minipage}
\caption{Directional MLP output patching averaged over 160 classification pairs.
Top row: denoising for \textit{"short"} clean and \textit{"long"} corrupted; bottom row: denoising for \textit{"long"} clean and \textit{"short"} corrupted.}
\label{fig:cc-mlp-out-160-heatmap}
\end{figure}

\begin{figure}[htbp]
\centering
\begin{minipage}[c]{0.8\textwidth}
    \centering
    \includegraphics[width=\linewidth]{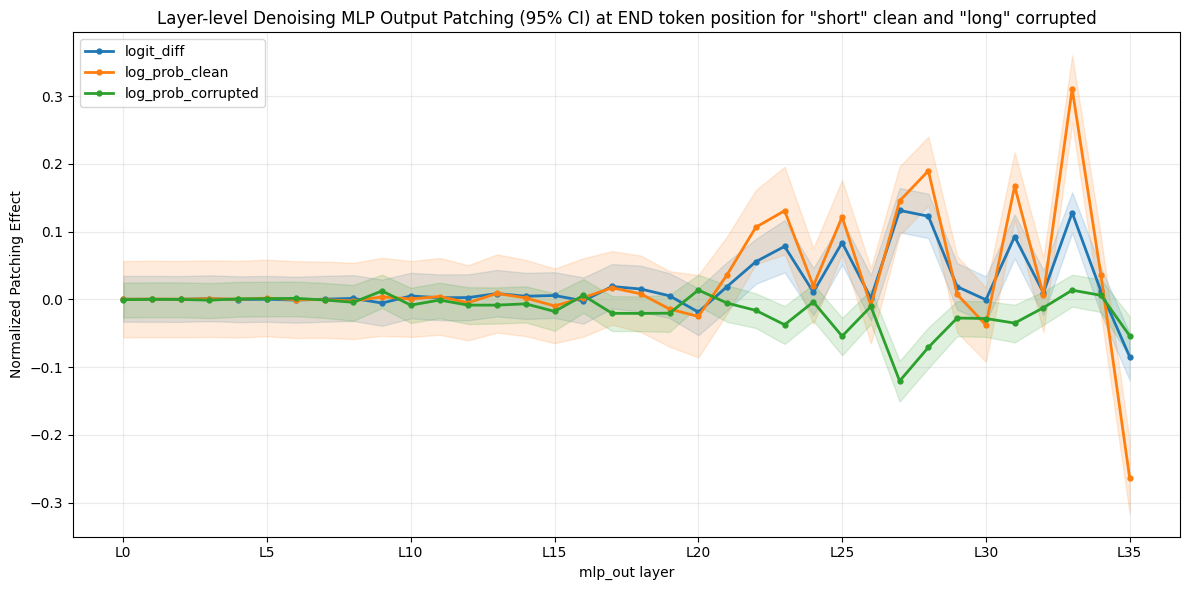}
    \vspace{4pt}
    \includegraphics[width=\linewidth]{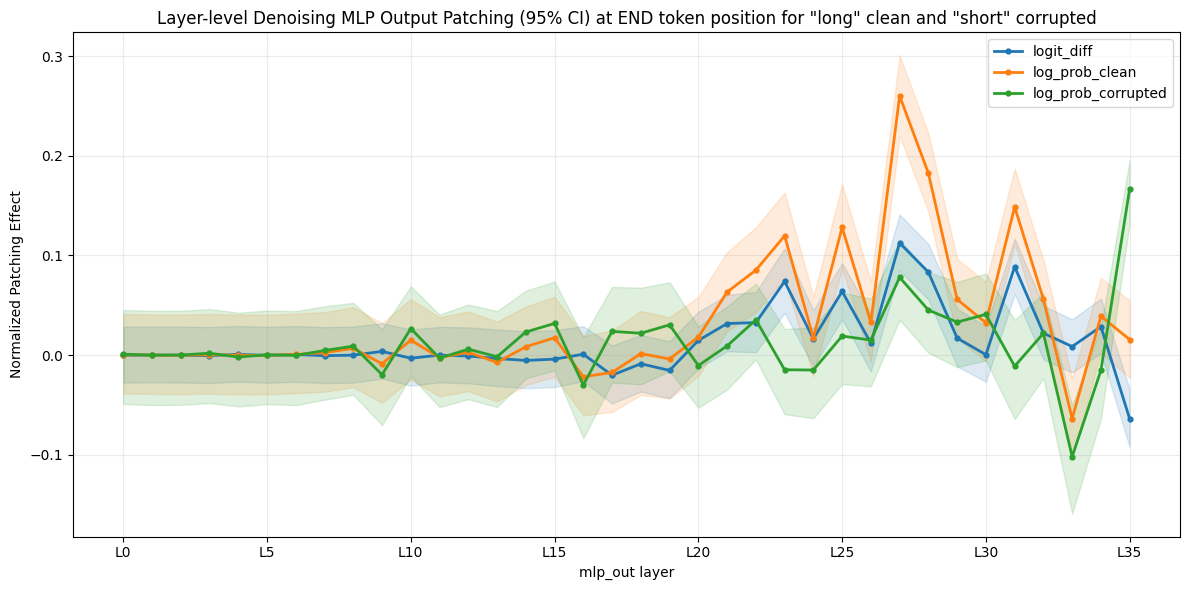}
\end{minipage}
\caption{Patching effects on MLP output at END token positions.
Top row: denoising for \textit{"short"} clean and \textit{"long"} corrupted; bottom row: denoising for \textit{"long"} clean and \textit{"short"} corrupted.}
\label{fig:cc-mlp-out-160}
\end{figure}

\begin{figure}[htbp]
\centering
\begin{minipage}[c]{0.8\textwidth}
    \centering
    \includegraphics[width=\linewidth]{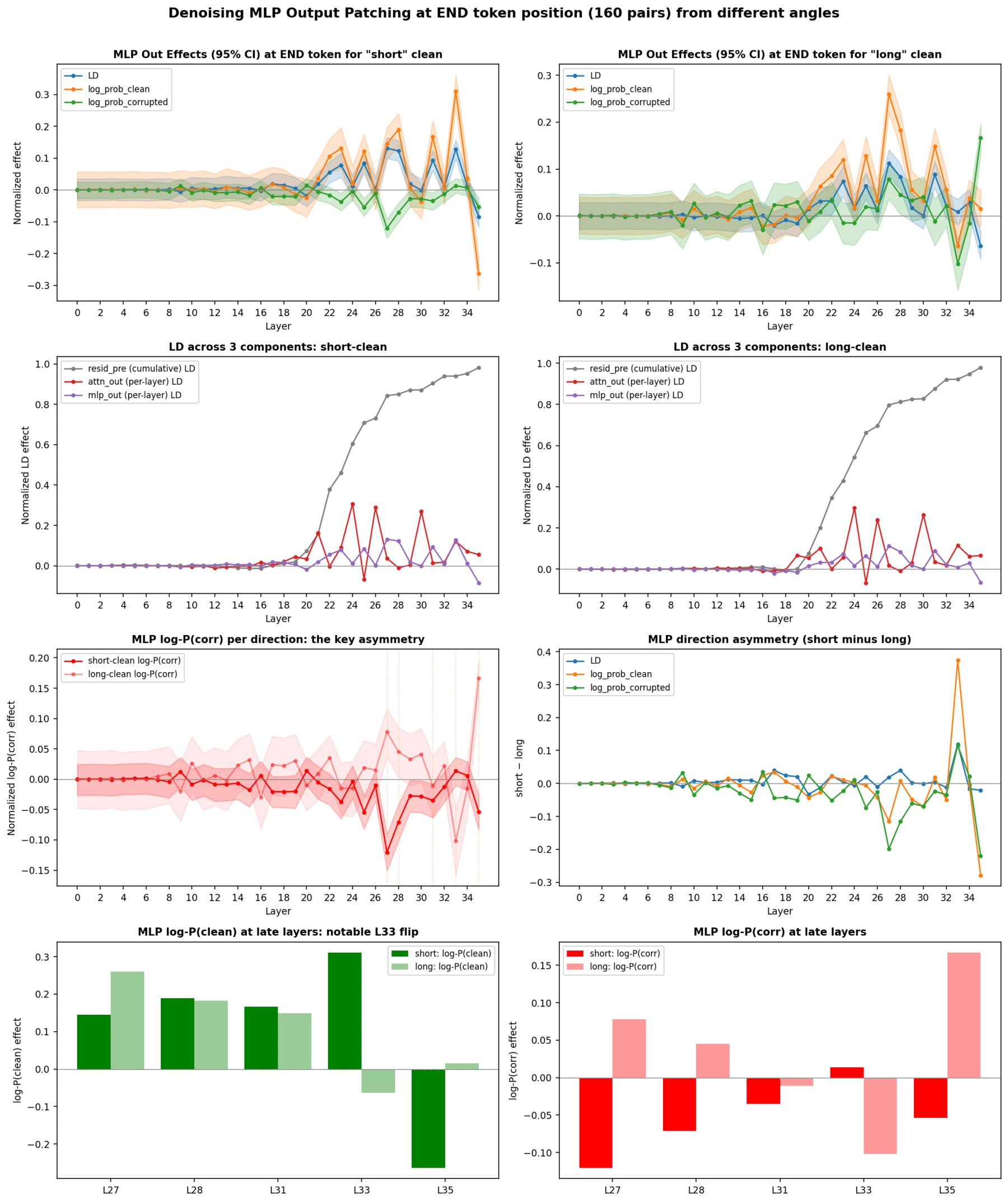}
\end{minipage}
\caption{MLP-output patching: per-layer effects and direction asymmetry.}
\label{fig:cc-mlp-out-160-all-results}
\end{figure}

The attention-output analysis (Sec.~\ref{sec:cc-attn-out-section}) suggested that attention blocks promote the correct answer with only a small suppression side-effect, and we conjectured that the deeper, direction-asymmetric suppression seen at \texttt{resid\_pre} would be carried by the MLPs. To test this, we patch the per-layer \texttt{mlp\_out} hook at the final token position, using the same two flips and the same three metrics. We also provide per-layer effects for all prompt tokens in Fig~\ref{fig:cc-mlp-out-160-heatmap} for a broader view. Fig.~\ref{fig:cc-mlp-out-160} shows patching effects at the final token and Figure~\ref{fig:cc-mlp-out-160-all-results} aggregates structural findings by multiple plots.

\paragraph{MLPs are weaker writers than attention, and they also promote.}
MLP effects are substantially smaller in magnitude than attention effects: peak LD is $+0.131$ (short-clean, L27) versus $+0.308$ for attention (short-clean, L24), about a factor of $2.4$ smaller. A consistent positive-LD signature appears across three layers in the middle-late range (L27, L28, and L31), which form the primary MLP writer band and are significant in both flips (L27: $+0.131 / +0.113$; L28: $+0.123 / +0.083$; L31: $+0.093 / +0.089$). \textit{Contrary to our initial conjecture, these MLP writers do not primarily suppress the incorrect answer:} at every layer in the primary band, the effect on log-$P(\text{clean})$ is larger in magnitude than the effect on log-$P(\text{corr})$, the same promotion-dominant pattern we observed at attention. MLPs therefore reinforce the decision written by attention rather than performing a qualitatively distinct suppression step.
 
\paragraph{The late-layer suppression hypothesis is not supported.}
The residual-stream analysis showed that log-$P(\text{corr})$ saturates near $-1$ across layers~28--35, with a pronounced direction-dependent delay in the long-clean flip. If a specific MLP layer were responsible for this late-layer suppression, it should appear here as a large negative effect on log-$P(\text{corr})$ at one or more of layers~28--35. The observed MLP log-$P(\text{corr})$ effects in that range are, however, small and mixed in sign: the largest is $-0.121$ at L27 in the short-clean flip (significant), but in the long-clean flip the corresponding MLPs at L27 and L28 show \emph{positive} log-$P(\text{corr})$ effects ($+0.078$ and $+0.045$, both significant), meaning they slightly push the model toward the incorrect answer. No single MLP layer produces a suppression effect comparable in magnitude to the saturation observed at
\texttt{resid\_pre}. The component-level decomposition we conjectured at the end of Sec.~\ref{sec:cc-attn-out-section} (attention promoting, MLPs suppressing) is therefore not supported by the data. The deep suppression at \texttt{resid\_pre} appears instead to be a cumulative property of many small contributions distributed across both components, rather than the work of a localized suppressor.

\paragraph{L33 and L35: late MLPs with prominent direction-dependent effects on \textit{short}.}
Two late MLP layers produce robust effects whose largest significant components all involve the model's prediction of \textit{short}. At L33 in the short-clean flip, patching the clean-run MLP output produces a log-$P(\text{clean})$ effect of $+0.311$. At L35 in the short-clean flip, the same operation produces a log-$P(\text{clean})$ effect of $-0.264$;  while in the long-clean flip, it produces a log-$P(\text{corrupted})$ effect of $+0.167$, the single largest log-$P(\text{corrupted})$ effect observed at any MLP layer in either direction. As a possible interpretation we can say that the fact that both layers' largest effects fall on \textit{short}-related metrics (rather than being distributed across the metrics tracked at other writers) is consistent with their carrying computations specifically tied to the \textit{short} representation. The direction of the patching effect differs between flips: at L35 in particular, patching pushes the model away from \textit{short} when \textit{short} is correct and toward \textit{short} when \textit{long} is correct (Figure~\ref{fig:cc-mlp-out-160-all-results}, bottom row). A mechanistic account of these polarity-dependent pattern would require head-level or neuron-level analysis. Alternative explanations (residual interaction with the unembedding, normalization artifacts at the final layer) cannot be ruled out without further experiments.

\paragraph{Summary.}
MLP patching refutes our earlier conjecture that the direction-asymmetric late-layer suppression at \texttt{resid\_pre} is localized to MLPs in layers 28--35. MLPs are weaker writers than
attention, and at the layers where they contribute significantly (L27, L28, L31) they continue the promotion-dominant pattern established by attention. The aggregate late-layer suppression observed at \texttt{resid\_pre} is therefore best understood as a cumulative property of many small contributions across both components, not the work of a localized suppressor. Several MLP writers show direction-dependent behavior; among these, L33 and L35 stand out for the size of their robust effects, all of which involve the model's prediction of \textit{short}. The L33/L35 finding does not localize the residual-stream asymmetry's origin: the asymmetry on log-$P(\text{corrupted})$ is already near its peak by L24. What L33 and L35 add is a set of large, direction-dependent effects whose largest components fall on metrics involving the \textit{short} token, with L35's long-clean log-$P(\text{corrupted})$ effect ($+0.167$) being the single largest log-$P(\text{corrupted})$ effect observed at any MLP layer in either direction. Head-level attention patching at L21, L24, L26, L30 and neuron-level analysis of MLPs at L33 and L35 would be the natural next experiments to test and refine this picture.

\subsection{All-tokens patching view}\label{evidence:cc-alltokens-view-convergence}
\begin{figure}[htbp]
\centering
\begin{minipage}[c]{0.49\textwidth}
    \centering
    \includegraphics[width=\linewidth]{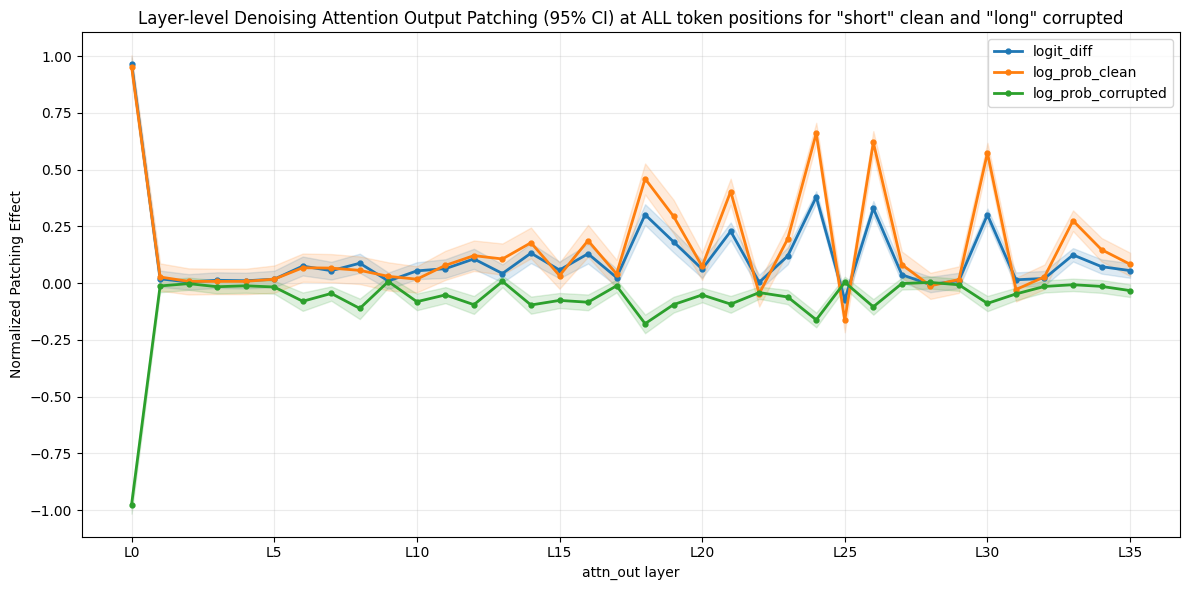}
\end{minipage}
\hfill
\begin{minipage}[c]{0.49\textwidth}
    \centering
    \includegraphics[width=\linewidth]{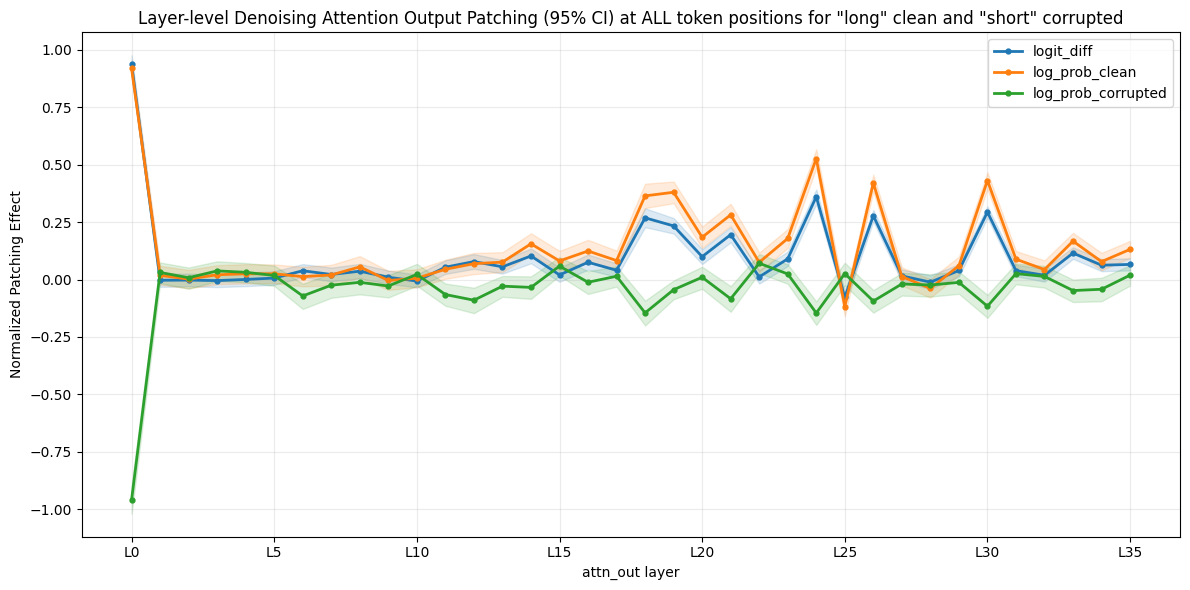}
\end{minipage}

\vspace{4pt}

\begin{minipage}[c]{0.49\textwidth}
    \centering
    \includegraphics[width=\linewidth]{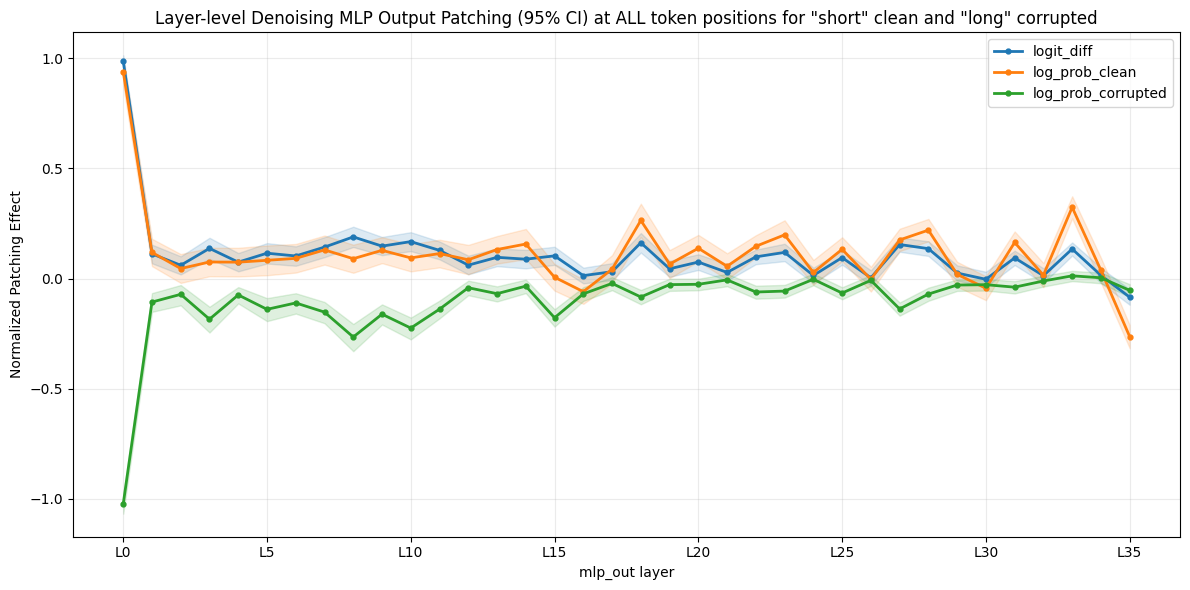}
\end{minipage}
\hfill
\begin{minipage}[c]{0.49\textwidth}
    \centering
    \includegraphics[width=\linewidth]{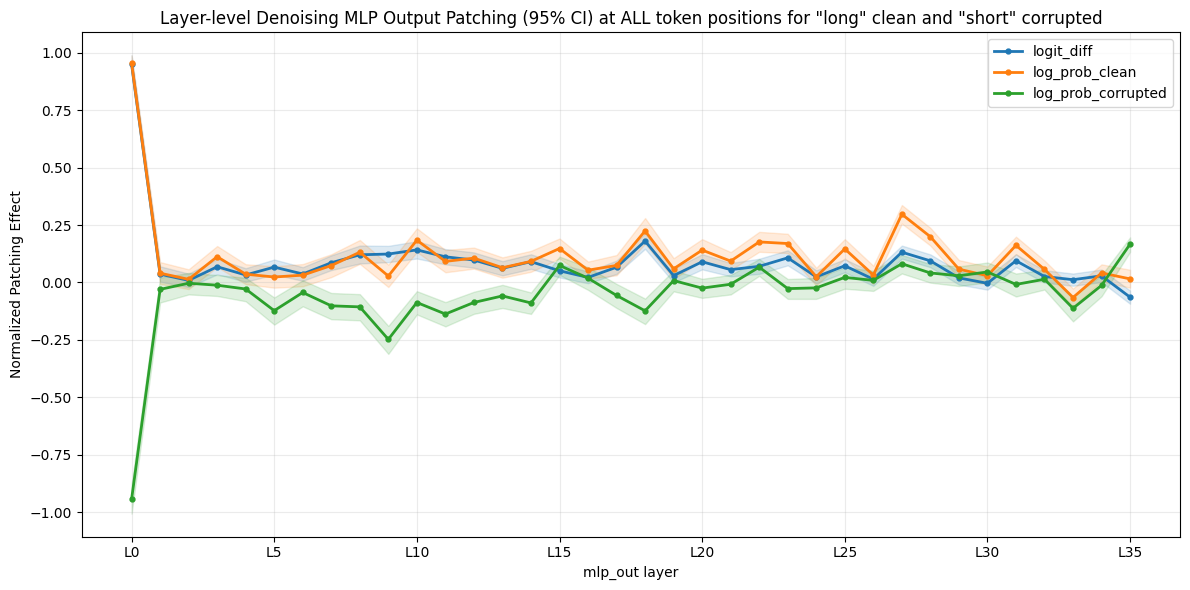}
\end{minipage}
\caption{Per-layer patching effects when each (layer, component) intervention is applied at all 34 token positions simultaneously, averaged over 160 classification pairs (95\% CIs). Top row: attention output. Bottom row: MLP output. Left column: short-clean flip; right column: long-clean flip.}
\label{fig:cc-alltokens-views}
\end{figure}

\begin{figure}[!htbp]
    \centering
    \includegraphics[width=\linewidth]{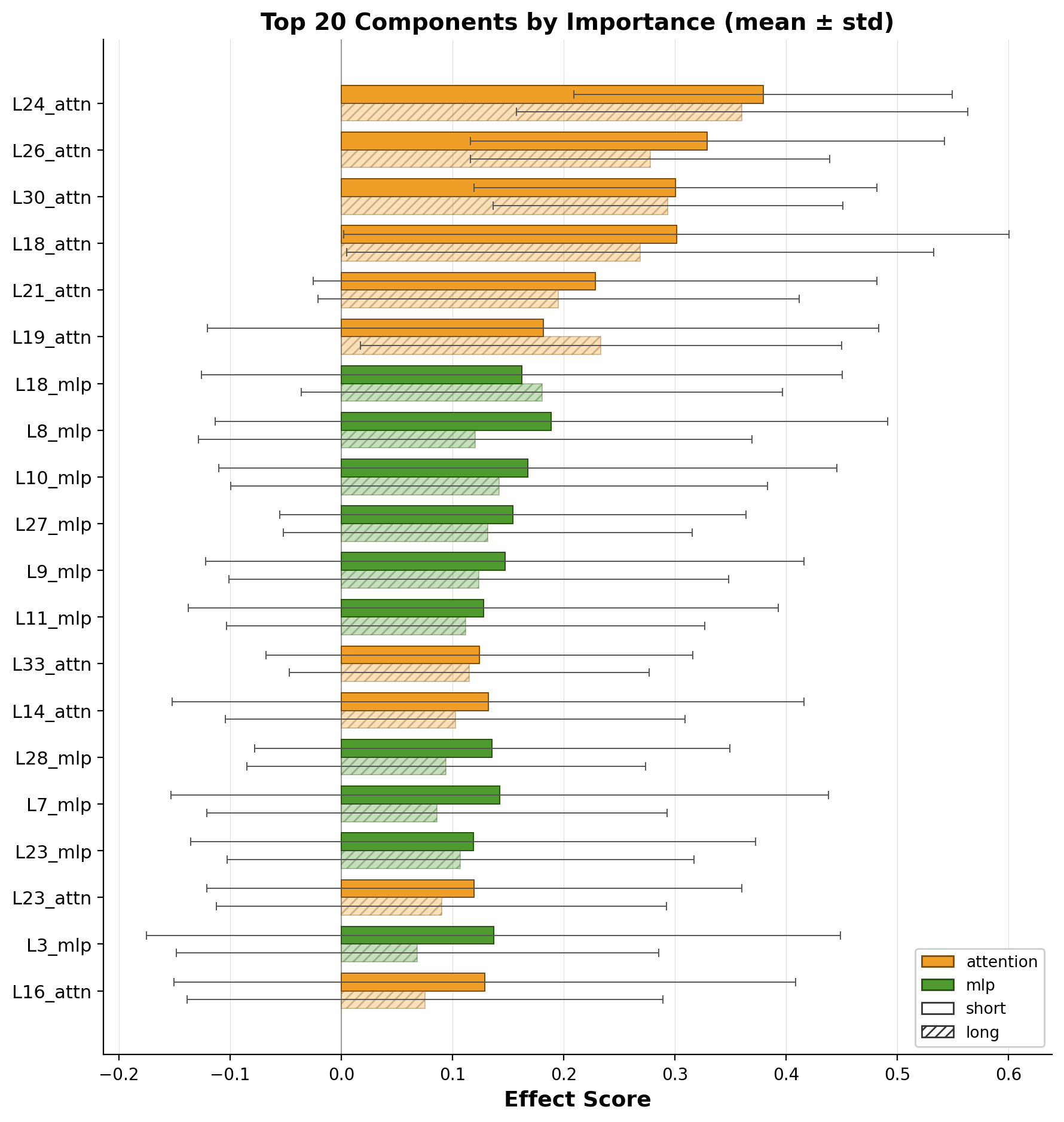}
    \caption{
        Top 20 components ranked by sum of mean $|\text{LD}|$ scores for \textit{short-clean} and \textit{long-clean} flips. 
        L0 layer is excluded.
    }
    \label{fig:cc-alltokens-top20}
\end{figure}

The per-layer analyses above (Sec.~\ref{sec:cc-attn-out-section}, Sec.~\ref{sec:cc-mlp-out-section}) focus on the END token position. 
For methodologically comparable cross-method comparison with EAP-IG attribution (\ref{app:attributional-contrastive}) and the layer sweep in causal parametric patching (\ref{app:causal-parametric}), we report a complementary patching pass in which each (layer, component) intervention is applied at all 34 token positions simultaneously.
Figure~\ref{fig:cc-alltokens-views} shows attention output (top) and MLP output (bottom) patching effects under this all-tokens scheme, separately for the \textit{short-clean} flip (left) and \textit{long-clean} flip (right).
Figure~\ref{fig:cc-alltokens-top20} shows top-20 \texttt{Qwen3-4B-Instruct-2507} components ranked by sum of their mean $|\text{LD}|$ scores for both flips.

\textbf{L0 peaks are embedding-level differences, not temporal computation.} Both \texttt{attn\_out} and \texttt{mlp\_out} show very high scores at L0, which reflects embedding-level differences between clean and corrupted prompts (different goal text) rather than causal temporal computation at the first layer.
All-tokens patching at L0 is therefore approximately a no-op at the positions where the embeddings agree and a clean substitution at the goal-text positions, while downstream attention propagates the substituted clean goal-text content from those positions to the answer.
Thus, the first layer is excluded from cross-method localisation claims.

\textbf{The END-token attention writers recur under all-tokens patching, with L18 emerging as a major additional contributor.} Excluding L0, the top-ranked all-tokens attention layers are L24, L26, L30, and \textbf{L18}, followed by L21 and L19, with L33 a smaller, secondary writer. Layers L24, L26, L30 are exactly the strongest dominant writers identified at the END token (Sec.~\ref{sec:cc-attn-out-section}), 
with L21 weaker but still among the dominant set. L18 was invisible at any single (layer, token) cell at the end-position view, but becomes prominent under all-tokens patching, consistent with L18 attention having sub-threshold contributions distributed across several positions that aggregate to a substantial layer-level effect. 
L33 as a secondary writer is consistent with the end-position view, where L33--34 form a pair of smaller late writers contributing positively (Sec.~\ref{sec:cc-attn-out-section}).

\textbf{MLP signal is broad and mid-network-dominated, contrary to the END-token-only picture.} Under all-tokens patching, the top MLP layers by mean $|\text{LD}|$ are L18, L8, L10, L27, L9, and L11, with smaller but reliable contributions at 
L28, L23 and L31 (Figure~\ref{fig:cc-alltokens-views} and Figure~\ref{fig:cc-alltokens-top20}). 
The prominent signal therefore lies in two bands: a mid-network MLP cluster spanning L8--L11 and L18, and a late-MLP band at L23, L27--L31. The mid-network cluster doesn't appear under END token patching; its effects must therefore be distributed across non-END positions and 
visible only under the all-tokens intervention. The late-layers cluster partially overlaps with causal parametric's L31, L35 prominence (\ref{app:causal-parametric}) and contrastive attribution's diffuse L31--L35 signal (\ref{app:attributional-contrastive}); 
however, the effects in this cluster are weaker (see below).
The L33 / L35 asymmetries identified in Sec.~\ref{sec:cc-mlp-out-section} at the END token also reproduce under all-tokens patching.

\textbf{MLP comparison with the other methods: numerical convergence at L31-L35 exists but is weak.} Causal parametric and contrastive attribution both identify an important late-MLP band at L31--L35. 
Classification numerically reaches this band at L31 (short / long LD = $+0.094$ / $+0.095$) and at L35 ($-0.085$ / $-0.064$; the only consistently negative-LD layer). L33 MLP also sits in this region but is interpretively distinct: its per-flip LD 
is asymmetric ($+0.134$ in \textit{short-clean}, $+0.012$ in \textit{long-clean}) and its interest is the bidirectional short-axis modulator behaviour identified at the END token (Sec.~\ref{sec:cc-mlp-out-section}), not cross-method overlap.
In the all-tokens ranking, all three sit below classification's dominant MLP signal: the mid-network cluster at L8--L11 and L18, plus L27 (Figure~\ref{fig:cc-alltokens-top20}). 
We therefore do not highlight the L31--L35 region as a primary cross-method convergence claim for the classification pipeline: the numerical overlap exists but is too weak relative to classification's mid-network signal to anchor such a statement. 
The robust cross-method convergence claims supported by this experiment remain L24 attention, the L21--L24 attention substrate, and the residual decision window (L20--L27); see Finding~\ref{find:cc-convergence} and Table~\ref{tab:convergence}.

\subsection{Key Findings}
\label{sec:cc-key-findings-section}
 
\begin{enumerate}[leftmargin=1.5em]
\item \textbf{The decision is sparsely localized at the final token.}
At the final token, the causal effect on temporal classification concentrates in a small set of layers in a narrow band of processing depth. The \texttt{resid\_pre} LD curve is silent for layers 0--19
(confidence intervals contain zero at every layer), rises as a seven-layer staircase across L20--27
that accounts for roughly 80\% of the LD recovery, and continues through
a slower saturation tail at L28--35.
 
\item \textbf{Attention and MLP components are both promotion-dominant.}
Component-level patching refutes the simplest decomposition one
might expect (\emph{attention writes the answer, MLPs suppress the
alternative}). At the dominant attention writers the ratio
$|\text{log-}P(\text{clean})|/|\text{log-}P(\text{corrupted})|$ is
four or more in seven of the eight layer--flip combinations we test
(L21, L24, L26, L30 across both flips). MLP writers are also
promotion-dominant at most layers but with greater variability:
ratios at the primary-window writers span from $1.21$ (L27
short-clean; roughly balanced promotion and suppression) up to
$13.31$ (L31 long-clean); the direction-dependent layers L33 and
L35 (Finding 5) deviate further and are described
separately. The late-layer suppression of the incorrect answer seen at
\texttt{resid\_pre} is therefore not localized to a single
component; it accumulates from many small contributions distributed
across attention and MLPs.
 
\item \textbf{Attention is the primary writer, MLPs reinforce.}
Within the L20--27 window, attention contributes the majority of the
magnitude. The principal attention writers in both flips are L21,
L24, L26, and L30, with a single-layer peak at L24. MLPs add smaller but reproducible writes at L27, L28, and L31; peak MLP LD effect is $+0.131$ at L27 short-clean, roughly $2.4\times$ smaller than the
attention peak (which sits at a different layer, L24). The two
component families work in concert rather than in specialized roles.
 
\item \textbf{Promotion is approximately symmetric across flips;
suppression is not.}
At \texttt{resid\_pre}, the onset of the decision on LD coincides in
both flips at L20. Clean-answer promotion proceeds at similar rates
in the two flips: log-$P(\text{clean})$ milestones at $10$,
$25$, $50$, $75$, and $90\%$ recovery coincide to within one layer
at every threshold. Corrupted-answer suppression, by contrast, is
consistently faster when the correct answer is \textit{short}:
reaching $50\%$, $75\%$, and $90\%$ of the full suppression requires
layers $25/27/32$ in the short-clean flip but $27/31/35$ in the
long-clean flip, a delay of 2--4 layers at every threshold. The suppression asymmetry is thus ${\sim}1.8\times$
the promotion asymmetry on the relevant metrics and closes only at
the final layer (L35).

\item \textbf{Two MLP layers show large direction-dependent effects on
the \textit{short} token.}
Layers 33 and 35 stand out for the size of their robust effects, all of which involve metrics related to
the model's prediction of \textit{short}. At L33 in the short-clean
flip, patching the clean-run MLP output yields a log-$P(\text{clean})$
effect of $+0.311$. At L35 in the short-clean flip, the same operation
yields a log-$P(\text{clean})$ effect of $-0.264$. At L35 in the
long-clean flip, patching produces a log-$P(\text{corrupted})$ effect
of $+0.167$, the single largest log-$P(\text{corrupted})$ effect
observed at any MLP layer in either direction. A unified mechanistic account of these patterns would require head-level or neuron-level analysis.

\item \textbf{The circuit-level asymmetry parallels a behavioral
short-bias.}
The 15 clear-signal misclassifications in the dataset were all
false-\textit{short} predictions. The residual-stream finding that
suppressing \textit{short} (long-clean flip) requires more layers
than suppressing \textit{long} forms a mechanical
pattern in the same direction as this behavioral bias. A causal
link between the patching-level asymmetry and the
classification-level errors cannot be established from this
experiment alone, but the direction of both effects agrees.

\item \label{find:cc-convergence} \textbf{Convergence with parametric and contrastive attribution results.}
For methodologically comparable cross-paradigm comparison we use the all-tokens patching view of the classification results (Sec.~\ref{evidence:cc-alltokens-view-convergence}), in which each (layer, component) intervention is applied at all 34 token positions simultaneously. 
Under this view, L21, L24, L26, and L30 are the attention writers robust across both flips, with L18 emerging as a major additional contributor not visible at the END token. L33 is highlighted as a secondary writer and is consistent with the end-position view, 
where L33--34 form a pair of smaller late writers contributing positively. 
L24 is the dominant attention writer across all four non-probing methods (classification patching, contrastive attribution \ref{app:attributional-contrastive}, parametric attribution \ref{app:attributional-parametric} and parametric patching \ref{app:causal-parametric}). 
Classification's core decision window (L20--27) on the residual stream and parametric's critical window (L20--24) align on layers L20--24. MLP contributions partially overlap with a rank-wise difference: classification's highest-ranking MLP layers are a mid-network cluster 
at L8--L11 and L18 followed by L27, while causal parametric ranks L31 and L35 as top MLP contributors and contrastive attribution shows a diffuse L31--L35 signal. Classification reaches the L31--L35 region numerically but only weakly 
(Sec.~\ref{evidence:cc-alltokens-view-convergence}); we therefore do not highlight L31--L35 as a primary classification convergence claim.
The partial overlap on attention and residual stream is consistent with shared computational machinery in this region, though whether this reflects a common temporal representation or merely a common decision readout requires further analysis we leave to future work.
\end{enumerate}

\FloatBarrier
\clearpage
\clearappnumbering

\section{Cross-method convergence}\label{app:convergence}

The preceding five appendices each approached localization from a different angle: two attribution methods (one contrastive, one parametric), two causal patching experiments, and supervised probing, applied across three prompt paradigms.
Each method has different assumptions, blind spots, and failure modes.
The question is whether they agree.

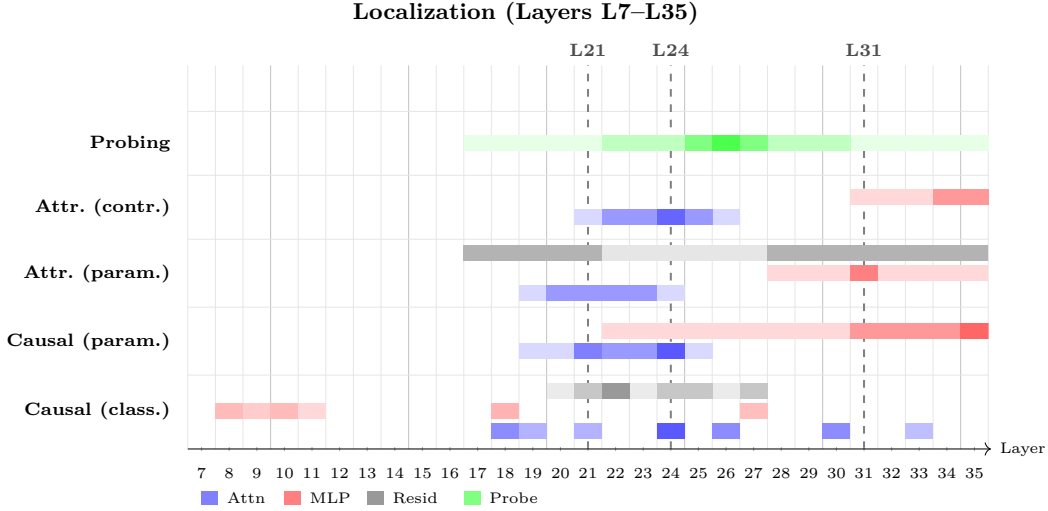
\begin{figure}[htbp]
\centering
\resizebox{\linewidth}{!}{%
\begin{tikzpicture}[
    x=0.38cm, y=0.55cm,
]


\foreach \l in {6,7,...,35} {
    \pgfmathsetmacro{\xx}{\l + 0.5}
    \draw[black!10, line width=0.25pt] (\xx, 0) -- (\xx, 9.6);
}
\foreach \l in {9,14,19,24,29,34} {
    \pgfmathsetmacro{\xx}{\l + 0.5}
    \draw[black!22, line width=0.35pt] (\xx, 0) -- (\xx, 9.6);
}

\foreach \y in {1.85, 3.55, 5.25, 6.85, 8.45} {
    \draw[black!10, line width=0.25pt] (6.5, \y) -- (35.5, \y);
}

\draw[dashed, thick, black!50] (21, 0) -- (21, 9.6);
\draw[dashed, thick, black!50] (24, 0) -- (24, 9.6);
\draw[dashed, thick, black!50] (31, 0) -- (31, 9.6);
\node[font=\scriptsize\bfseries, black!70, anchor=south] at (21, 9.6) {L21};
\node[font=\scriptsize\bfseries, black!70, anchor=south] at (24, 9.6) {L24};
\node[font=\scriptsize\bfseries, black!70, anchor=south] at (31, 9.6) {L31};

\foreach \l in {7,8,...,35} {
    \node[below, font=\tiny] at (\l, -0.25) {\l};
    \draw[black!35, line width=0.3pt] (\l, -0.05) -- (\l, 0.05);
}
\draw[->] (6.5,0) -- (35.6,0) node[right, font=\tiny] {Layer};

\node[anchor=east, font=\scriptsize\bfseries] at (6.2, 0.95) {Causal (class.)};
\fill[blue!25] (32.5,0.25) rectangle (33.5,0.65);   
\fill[blue!30] (18.5,0.25) rectangle (19.5,0.65);   
\fill[blue!30] (20.5,0.25) rectangle (21.5,0.65);   
\fill[blue!45] (17.5,0.25) rectangle (18.5,0.65);   
\fill[blue!45] (25.5,0.25) rectangle (26.5,0.65);   
\fill[blue!45] (29.5,0.25) rectangle (30.5,0.65);   
\fill[blue!65] (23.5,0.25) rectangle (24.5,0.65);   
\fill[red!15] (10.5,0.75) rectangle (11.5,1.15);   
\fill[red!20] (8.5,0.75)  rectangle (9.5,1.15);    
\fill[red!25] (26.5,0.75) rectangle (27.5,1.15);   
\fill[red!27] (7.5,0.75)  rectangle (8.5,1.15);    
\fill[red!27] (9.5,0.75)  rectangle (10.5,1.15);   
\fill[red!30] (17.5,0.75) rectangle (18.5,1.15);   
\fill[black!8]  (19.5,1.25) rectangle (27.5,1.65);  
\fill[black!22] (20.5,1.25) rectangle (21.5,1.65);  
\fill[black!22] (23.5,1.25) rectangle (25.5,1.65);  
\fill[black!22] (26.5,1.25) rectangle (27.5,1.65);  
\fill[black!40] (21.5,1.25) rectangle (22.5,1.65);  

\node[anchor=east, font=\scriptsize\bfseries] at (6.2, 2.7) {Causal (param.)};
\fill[blue!15] (18.5,2.25) rectangle (25.5,2.65);  
\fill[blue!40] (20.5,2.25) rectangle (24.5,2.65);  
\fill[blue!50] (20.5,2.25) rectangle (21.5,2.65);  
\fill[blue!65] (23.5,2.25) rectangle (24.5,2.65);  
\fill[red!15]  (21.5,2.75) rectangle (35.5,3.15);  
\fill[red!40]  (30.5,2.75) rectangle (35.5,3.15);  
\fill[red!60]  (34.5,2.75) rectangle (35.5,3.15);  

\node[anchor=east, font=\scriptsize\bfseries] at (6.2, 4.4) {Attr.\ (param.)};
\fill[blue!15] (18.5,3.70) rectangle (24.5,4.10);  
\fill[blue!40] (19.5,3.70) rectangle (23.5,4.10);  
\fill[red!15]  (27.5,4.20) rectangle (35.5,4.60);  
\fill[red!50]  (30.5,4.20) rectangle (31.5,4.60);  
\fill[black!10] (16.5,4.70) rectangle (35.5,5.10); 
\fill[black!30] (16.5,4.70) rectangle (21.5,5.10); 
\fill[black!30] (27.5,4.70) rectangle (35.5,5.10); 

\node[anchor=east, font=\scriptsize\bfseries] at (6.2, 6.05) {Attr.\ (contr.)};
\fill[blue!15] (20.5,5.60) rectangle (26.5,6.00);  
\fill[blue!40] (21.5,5.60) rectangle (25.5,6.00);  
\fill[blue!60] (23.5,5.60) rectangle (24.5,6.00);  
\fill[red!15]  (30.5,6.10) rectangle (35.5,6.50);  
\fill[red!40]  (33.5,6.10) rectangle (35.5,6.50);  

\node[anchor=east, font=\scriptsize\bfseries] at (6.2, 7.65) {Probing};
\fill[green!10] (16.5,7.45) rectangle (35.5,7.85); 
\fill[green!25] (21.5,7.45) rectangle (30.5,7.85); 
\fill[green!50] (24.5,7.45) rectangle (27.5,7.85); 
\fill[green!70] (25.5,7.45) rectangle (26.5,7.85); 

\node[font=\small\bfseries, anchor=south] at (18.75, 10.4) {Localization (Layers L7--L35)};

\fill[blue!50]   (7.0, -1.40) rectangle (7.6, -1.05);
\node[anchor=west, font=\tiny] at (7.6, -1.22) {Attn};
\fill[red!50]    (10.0, -1.40) rectangle (10.6, -1.05);
\node[anchor=west, font=\tiny] at (10.6, -1.22) {MLP};
\fill[black!40]  (13.0, -1.40) rectangle (13.6, -1.05);
\node[anchor=west, font=\tiny] at (13.6, -1.22) {Resid};
\fill[green!50]  (16.5, -1.40) rectangle (17.1, -1.05);
\node[anchor=west, font=\tiny] at (17.1, -1.22) {Probe};

\end{tikzpicture}%
}
\caption{Layer-level convergence across all five localization methods.
Darker shading indicates stronger signal.
Blue $=$ attention, red $=$ MLP, green $=$ residual stream (probing), gray $=$ residual stream (attributional parametric, causal classification).
L24 attention appears in every non-probing method. The mid-network MLP cluster at L8--L11 is classification-specific and sits outside the L17--L35 convergence zone occupied by four other methods.}
\label{fig:convergence}
\end{figure}

Table~\ref{tab:convergence} and Figure~\ref{fig:convergence} show that they do: four of five methods place the temporal preference subgraph in layers 17--35; fifth classification experiment matches this zone for attention and the residual decision window.
All non-probing methods (\ref{app:attributional-contrastive},~\ref{app:attributional-parametric},~\ref{app:causal-parametric},~\ref{app:causal-contrastive}) highlight L24 as the central attention component.
The agreement is not trivial, as the methods also disagree in informative ways (Section~\ref{app:convergence-disagreement}).

\begin{table}[htbp]
    \centering
    \renewcommand{\arraystretch}{1.3}
    \setlength{\tabcolsep}{4pt}
    \scalebox{0.76}{%
    \begin{tabularx}{1.3\linewidth}{@{} >{\bfseries}l c c c c >{\centering\arraybackslash}X @{}}
    \toprule
    & \textbf{Probing} & \textbf{Attr.\ contr.} & \textbf{Attr.\ param.} & \textbf{Causal param.} & \textbf{Causal class.} \\
    & {\scriptsize\ref{app:contrastive-probing-linear}}
    & {\scriptsize\ref{app:attributional-contrastive}}
    & {\scriptsize\ref{app:attributional-parametric}}
    & {\scriptsize\ref{app:causal-parametric}}
    & {\scriptsize\ref{app:causal-contrastive}} \\
    \midrule
    Attn L21--L24             &            & \checkmark & \checkmark & \checkmark & \checkmark \\
    MLP L31--L35              &            & \checkmark & \checkmark & \checkmark &            \\
    Resid L17--L22 (recovery) &            &            & \checkmark & \checkmark & $\sim$\checkmark\\
    Resid L26                 & \checkmark &            &            &            & \checkmark \\
    \midrule
    Attn peak           &            & L24 (ST), L22 (LT) & L20--L24 & L24, L21 & L24, L26, L30; L18, L21, L19 \\
    MLP peak            &            & L34, L35   & L31      & L35, L31 &  L18, L8--L11, L27 \\
    Best single layer   & L26        & L24        & L17--L22 (resid) & L24      & L24  \\
    Signal onset        & $\sim$L17  & $\sim$L21  & $\sim$L17 & $\sim$L19 & $\sim$L20 \\
    \midrule
    \multicolumn{6}{@{}l}{\textbf{Convergence zone: layers 17--35}} \\
    \bottomrule
    \end{tabularx}}
    \caption{Layers and components identified by each localization method.
    Four of the five methods place all their identified components in layers 17--35; classification matches this zone for attention and the residual decision window but additionally surfaces a mid-network MLP cluster at L8--L11.
    L24 attention is flagged by every non-probing method.
    MLP contributions concentrate in L31--L35 under attributional contrastive, attributional parametric, and causal parametric patching; classification reaches this region numerically but with weak per-layer effects and is 
    therefore not highlighted as a convergence claim in the table.
    Symbol ``$\sim$\checkmark'' marks partial convergence.}
    \label{tab:convergence}
\end{table}

\subsection{Points of Agreement}

L24 attention is the single component flagged most consistently: it appears as a top-ranked element in all four non-probing methods across three paradigms, making it the most robustly identified element of the subgraph.
The MLP contribution concentrates in L31--L35 under attributional contrastive, attributional parametric and causal parametric patching, with MLP L31 identified as a key disruptor by the latter two.
All methods agree that the residual-stream temporal signal is absent from the first $\sim$15 layers and concentrated in the upper half of the network.
The probing peak at L26 falls between the attention computation window (L21--L24) and the MLP computation window (L31--L35), consistent with a readout layer that consolidates the output of the attention-mediated temporal routing before the MLP transformation stage.
The causal classification residual-stream staircase at end token (L20--L27, peak L22) spans this same region, converging with the attention window (L21--L24) on the low end and with the probing peak (L26) on the high end.

\subsection{Points of Disagreement}\label{app:convergence-disagreement}

The methods disagree on three dimensions.

\paragraph{Attention breadth.}
The causal classification experiment identifies a broader span of attention layers: L18, L19, L21, L24, L26, L30, L33 (secondary writer), than the parametric and attribution methods, which concentrate on L21--L24. 
The data do not establish the cause of this difference: the setups vary along multiple dimensions (cognitive task, prompt structure, method, prompt length), and disambiguating which dimension drives the breadth difference would require controlled experiments we do not undertake here.
However, the most salient candidate is the cognitive task: classification probes categorical horizon inference rather than preference valuation.

\paragraph{MLP layer band.}
The three preference-targeted methods (attributional contrastive, attributional parametric, causal parametric) flag important MLP layers within the L31--L35 region, with some variation in which layers stand out. The classification's MLP signal, on the contrary, 
is broad and mid-network-dominated: L8--L11, L18, L27. Although classification reaches the L31--L35 region, the signal there is well below its mid-network cluster, so L31--L35 is not treated as a convergence claim for classification. 
The divergence is subject to the same multi-dimensional confound noted above.

\paragraph{Sufficiency vs.\ necessity asymmetry.}
The attributional parametric and causal parametric results reveal a sharp split between mid-layer recovery (L17--L22) and late-layer disruption (L30--L35), a pattern that contrastive and classification methods do not examine.
The causal classification experiment instead finds a different kind of asymmetry: the two flip directions share the same core decision window at the end token (L20--L27) and promote the clean answer at comparable rates, but suppress the corrupted answer at different speeds (suppressing \textit{short} when the answer is \textit{long} takes 2--4 more layers than the reverse).
This direction-dependent suppression is a dimension the parametric methods do not probe, because they do not separate patching directions.

\paragraph{Signal onset.}
The earliest onset varies from $\sim$L17 (probing) to $\sim$L21 (attributional contrastive).
This 4-layer gap may reflect the greater sensitivity of probing and parametric prompts to early, low-magnitude temporal information that the contrastive attribution method, aggregated over many prompt variants, averages out.
Classification's residual onset ($\sim$L20) is later than its earliest MLP signal (L8--L11 mid-network cluster) because the cluster is distributed across the prompt rather than concentrated at the end token,
while the end token's residual stream receives this signal only through subsequent attention gathering.

\subsection{Interpretation}

The disagreements are interpretable rather than contradictory: they reflect genuine differences in both what each method measures and what each paradigm probes.
Attribution methods approximate causal effects via gradients and are sensitive to all information flow, including redundant pathways.
Causal methods measure the actual behavioral consequence of intervention and are therefore sensitive to necessity and sufficiency.
Probing measures the linear readability of a concept at a given layer, regardless of whether that layer is causally important.
At the task level, four paradigms (probing, contrastive attribution, parametric attribution, and parametric patching) target temporal preference valuation, while the fifth (classification patching) probes categorical horizon inference, a distinct cognitive operation that may recruit different computation even when sharing the attention substrate.

That these five methods, despite their different assumptions and blind spots, broadly converge on a common subgraph in layers 17--35 with L24 attention at the center supports the claim that the localization is not an artifact of any single method.
The only finding outside this zone is classification's mid-network MLP cluster at L8--L11.
The probing--steering dissociation (\ref{app:contrastive-steering}) adds a sixth data point: layers 19--22 are most effective for writing temporal preference, while layer 26 is most effective for reading it, reinforcing the functional distinction between the attention routing window and the readout layer.

However, the subgraph is not monolithic.
The latent vs.\ constrained analysis (\ref{app:latent-vs-constrained}) reveals that the same L17--35 region operates in two modes depending on whether the prompt carries an explicit time-horizon constraint:
\begin{itemize}[nosep, leftmargin=*]
  \item \textbf{Constrained mode}: the full subgraph is engaged (attention L21--24 \emph{and} MLP L31--35), producing strong, distributed effects.
  \item \textbf{Latent mode}: only attention at L21--22 is active, with minimal MLP involvement.
\end{itemize}
\noindent The MLP layers that feature prominently in the convergence table (L31, L35) may therefore be specifically about processing constraint tokens rather than encoding temporal preference per se.
The attention core at L21--24 is the shared substrate; MLP extends the computation when the prompt provides an explicit temporal anchor.
Classification partially fits this picture: its L31--L35 MLP weakens as predicted, but a mid-network L8--L11 MLP cluster sits outside the framing (\ref{evidence:cc-alltokens-view-convergence}).

\clearpage
\clearappnumbering

\partpagecontent{Part 2:}{What does temporal preference look like?}{%
\begin{itemize}[leftmargin=*, itemsep=0.8em]
  \item \textbf{\hyperref[app:parametric-geometry]{M}.} Parametric geometry
  \item \textbf{\hyperref[app:latent-vs-constrained]{N}.} Latent vs.\ constrained
  \item \textbf{\hyperref[app:behavioral-temporal-discount]{O}.} Behavioral temporal discounting
  \item \textbf{\hyperref[app:behavioral-coherence]{P}.} Behavioral coherence
  \item \textbf{\hyperref[app:cross-model-comparison]{Q}.} Cross-model patching comparison
  \item \textbf{\hyperref[app:error-monitoring]{R}.} Error monitoring in the subgraph
\end{itemize}%
}

\section{Parametric geometry}\label{app:parametric-geometry}

Part 1 established \emph{where} temporal preference lives (layers 17--35, with L24 attention at the center; \ref{app:convergence}).
Now we ask: \emph{what does the representation look like inside that subgraph?}

We apply PCA to 4{,}588 activation vectors, sampled from a logarithmic grid over reward amounts, delay times, and 17 time horizons (seconds to centuries), at 15 layers, 5 component types, and 16 semantic positions per prompt (methodology in \ref{app:parametric-geometry-methods}).
At key positions, PC1 captures 44--71\% of variance (Table~\ref{tab:scree}).
The results tell a mechanistic story in five stages: the model builds an ordinal time-horizon representation, the geometric direction encoding it flips across prompt positions, it stabilizes at the user-to-assistant turn boundary, attention transforms it into a binary preference signal over the next few tokens, and the preference commits by the \texttt{assistant} token.

\subsection{Progressive separation across layers}\label{app:parametric-geo-layers}

\begin{figure}[htbp]
\centering
\begin{minipage}[t]{0.48\textwidth}
  \centering
  {\small\bfseries\texttt{resid\_pre}}\\[2pt]
  \includegraphics[width=\textwidth]{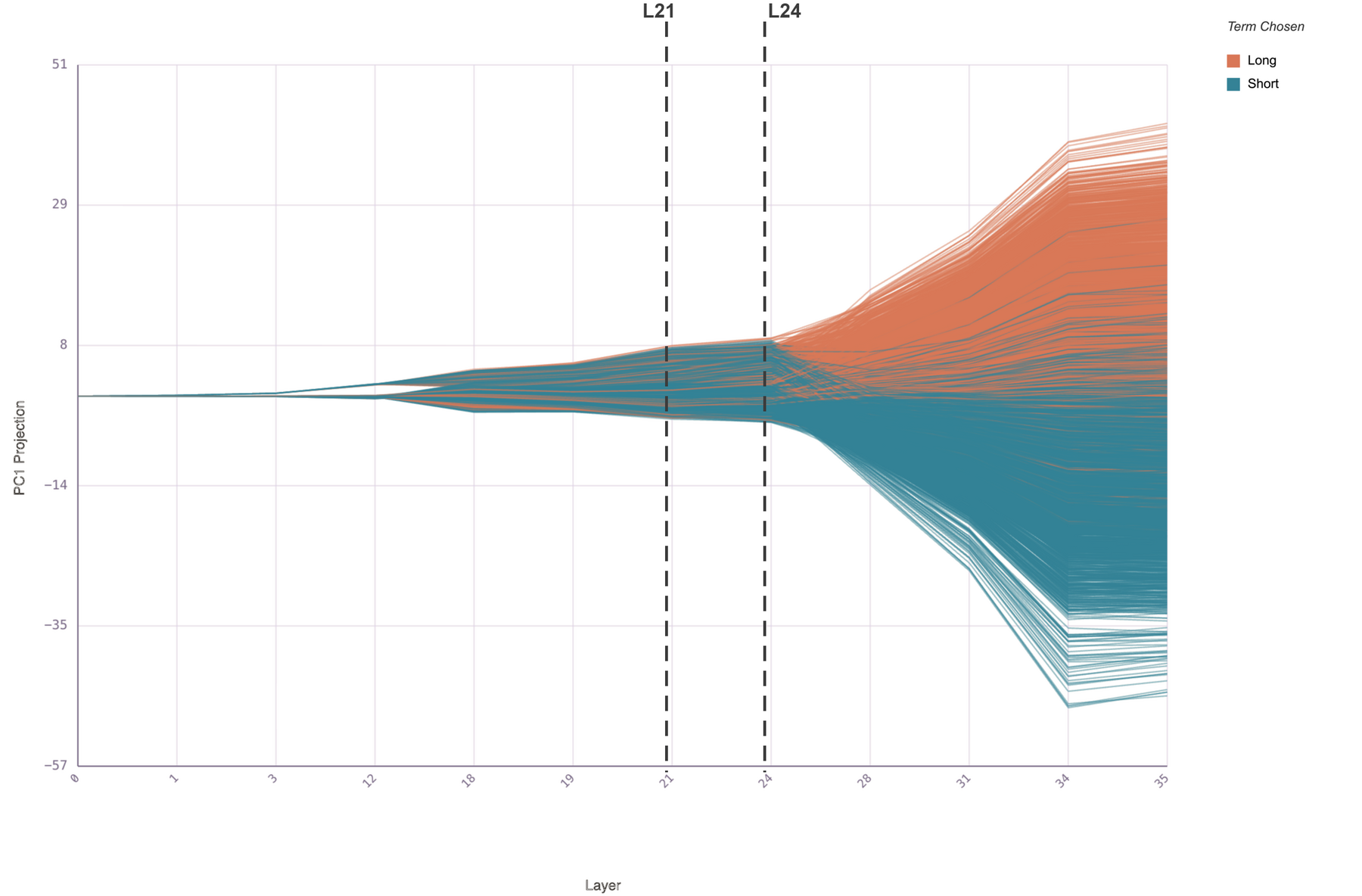}
\end{minipage}\hfill
\begin{minipage}[t]{0.48\textwidth}
  \centering
  {\small\bfseries\texttt{resid\_post}}\\[2pt]
  \includegraphics[width=\textwidth]{images/characterize/parametric_geometry/component_journey/resid_post.png}
\end{minipage}

\vspace{4pt}

\begin{minipage}[t]{0.48\textwidth}
  \centering
  {\small\bfseries\texttt{attn\_out}}\\[2pt]
  \includegraphics[width=\textwidth]{images/characterize/parametric_geometry/component_journey/attn.png}
\end{minipage}\hfill
\begin{minipage}[t]{0.48\textwidth}
  \centering
  {\small\bfseries\texttt{mlp\_out}}\\[2pt]
  \includegraphics[width=\textwidth]{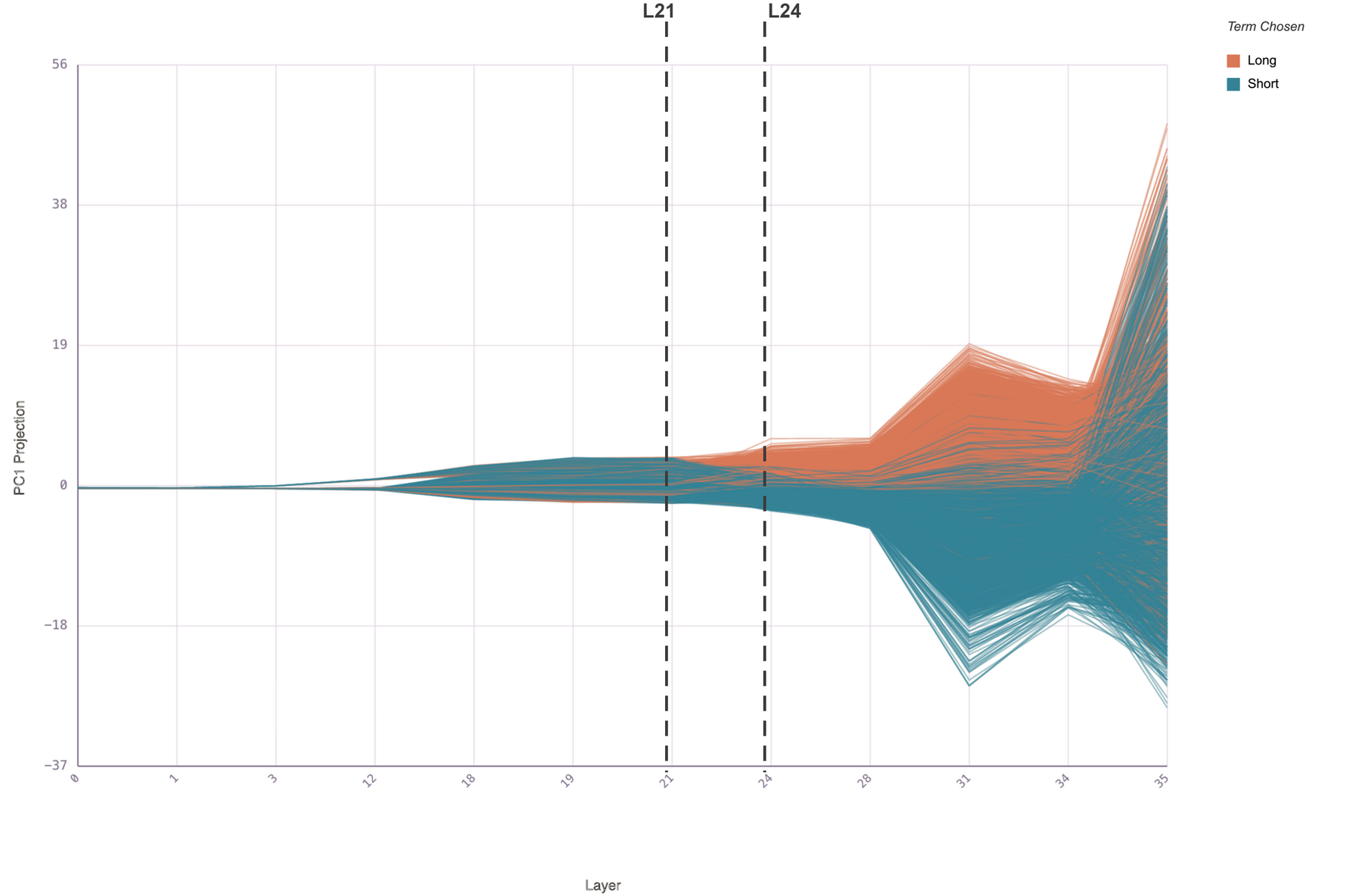}
\end{minipage}
\caption{PC1 projection across layers for each component type at the divergent token position, colored by the model's chosen term (orange = long, blue = short).
Short-term and long-term traces begin to diverge around layer 21 in the residual stream and attention, fully separating by layer 24 ($\pm$50 on PC1 for the residual stream, $\pm$20 for attention).
MLP contributions emerge later and remain smaller.
At the user-to-assistant turn boundary and later token positions, separation appears earlier (\ref{app:parametric-geo-bot}).}
\label{fig:component-journey}
\end{figure}

Figure~\ref{fig:component-journey} shows how the PC1 projection evolves across layers for four component types, colored by the model's eventual choice (long vs.\ short).

At this token position, all traces begin bundled near zero through the first $\sim$20 layers.
Separation becomes visible around layer 21 in the residual stream and attention output and is fully established by layer 24, consistent with the causal importance of L24 identified by activation patching (\ref{app:causal-parametric}).
By the final layers, the residual stream carries a separation of roughly $\pm$50 on PC1, while attention contributes $\pm$20 and MLP a smaller but complementary signal concentrated in the upper layers.

\FloatBarrier
\subsection{Off-policy horizon constraint (2D PCA)}\label{app:parametric-geo-horizon}

\figrowFull[1]{%
    \rowimgrowL{}{%
        images/characterize/parametric_geometry/by_position/chosen_term,%
        images/characterize/parametric_geometry/by_position/chosen_time,%
        images/characterize/parametric_geometry/by_position/time_scale%
    }{}
}{PCA of activation space at three token positions (chosen term, chosen time, time scale) with the time horizon given as an explicit constraint.\label{fig:geo-by-position}}

\figrowFull[1.2]{%
    \rowimgrowL{}{%
        images/characterize/parametric_geometry/by_layer/L3__post_time_horizon__time_scale,%
        images/characterize/parametric_geometry/by_layer/L18__post_time_horizon__time_scale,%
        images/characterize/parametric_geometry/by_layer/L24__post_time_horizon__time_scale%
    }{\includegraphics[width=0.1\columnwidth]{images/characterize/parametric_geometry/parametric_geometry_time_scale_legend}}
}{PCA of activation space colored by time scale at layers 3, 18, and 24 after the time horizon token.
Clusters become increasingly separable in the mid-to-upper layers.\label{fig:geo-by-layer}}

\FloatBarrier

\subsection{On-policy temporal preference (2D PCA)}\label{app:parametric-geo-bot}

\begin{figure}[htbp]
  \centering
  \scalebox{0.75}{\definecolor{eotfill}{HTML}{FCDDD3}
\definecolor{eotstroke}{HTML}{E8956F}
\definecolor{eottext}{HTML}{943D1E}

\definecolor{sotfill}{HTML}{D4F1E4}
\definecolor{sotstroke}{HTML}{5CB88A}
\definecolor{sottext}{HTML}{1A6B45}

\begin{tikzpicture}[
    node distance=0.8cm,
    box/.style={
        rounded corners=8pt,
        minimum height=1.6cm,
        font=\sffamily,
        align=center,
        line width=1pt,
    },
    eot/.style={box, fill=eotfill, draw=eotstroke, text=eottext, minimum width=3.2cm},
    sot/.style={box, fill=sotfill, draw=sotstroke, text=sottext, minimum width=3.2cm},
    arr/.style={-{Stealth[length=6pt, width=5pt]}, gray!40, line width=1.2pt},
    label/.style={font=\sffamily},
]

\node[eot] (end) {
    {\large\texttt{<|im\_end|>}}\\[4pt]
    {\footnotesize\itshape Control token}
};

\node[eot, minimum width=1.6cm, right=of end] (d1) {
    {\large\texttt{\textbackslash n}}\\[4pt]
    {\footnotesize\itshape Delimiter}
};

\node[sot, right=of d1] (start) {
    {\large\texttt{<|im\_start|>}}\\[4pt]
    {\footnotesize\itshape Control token}
};

\node[sot, right=of start] (role) {
    {\large\texttt{assistant}}\\[4pt]
    {\footnotesize\itshape Role tag}
};

\node[sot, minimum width=1.6cm, right=of role] (d2) {
    {\large\texttt{\textbackslash n}}\\[4pt]
    {\footnotesize\itshape Delimiter}
};

\draw[arr] (end)   -- (d1);
\draw[arr] (d1)    -- (start);
\draw[arr] (start) -- (role);
\draw[arr] (role)  -- (d2);

\draw[eotstroke, line width=2pt, cap=round]
    ([yshift=-1cm]end.south west) -- ([yshift=-1cm]d1.south east);
\node[label, color=eottext] at
    ([yshift=-1.6cm]$(end.south)!0.5!(d1.south)$) {End-of-Turn (EoT)};

\draw[sotstroke, line width=2pt, cap=round]
    ([yshift=-1cm]start.south west) -- ([yshift=-1cm]d2.south east);
\node[label, color=sottext] at
    ([yshift=-1.6cm]$(start.south)!0.5!(d2.south)$) {Start-of-Turn (SoT)};

\end{tikzpicture}}
  \caption{The transition from user EoT to assistant SoT marks the shift from off-policy context to on-policy generation.}
  \label{fig:turns}
\end{figure}
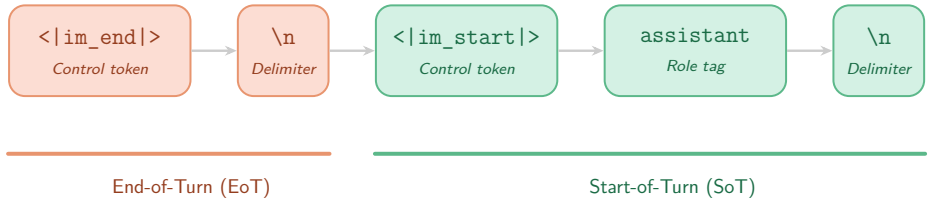

The preceding subsection examined how explicit time-horizon constraints are represented in activation space.
Here we trace what happens as the model transitions from the user's turn (where the constraint is given off-policy) into the assistant's turn, where it must generate on-policy text reflecting a temporal preference.
Figure~\ref{fig:turns} illustrates the token-level structure of this transition.
As the figures below show, the explicit time-horizon clusters reorganize during this hand-off: the no-horizon samples, initially disjoint, align to the time-scale manifold and temporal preference becomes linearly separable even at the earliest layers.

\figtwocolFull[0.8]{%
    \includegraphics[width=\linewidth]{images/characterize/parametric_geometry/parametric_geometry_term_legend}\\[6pt]
    \includegraphics[width=\linewidth]{images/characterize/parametric_geometry/parametric_geometry_time_scale_legend}
}{%
    \twocolrow{images/characterize/parametric_geometry/by_layer/L31__chat_suffix_0__term_chosen}
              {images/characterize/parametric_geometry/by_layer/L31__chat_suffix_0__time_scale}{\texttt{<|im\_end|>}}
    \twocolrow{images/characterize/parametric_geometry/by_layer/L31__chat_suffix_1__term_chosen}
              {images/characterize/parametric_geometry/by_layer/L31__chat_suffix_1__time_scale}{\texttt{\textbackslash n}}
    \twocolrow{images/characterize/parametric_geometry/by_layer/L31__chat_suffix_2__term_chosen}
              {images/characterize/parametric_geometry/by_layer/L31__chat_suffix_2__time_scale}{\texttt{<|im\_start|>}}
    \twocolrow{images/characterize/parametric_geometry/by_layer/L31__chat_suffix_3__term_chosen}
              {images/characterize/parametric_geometry/by_layer/L31__chat_suffix_3__time_scale}{\texttt{assistant}}
}{%
    Principal components of 4{,}588 samples for layer 31 at token positions through the end-of-turn (EoT) for the user into the beginning-of-turn (BoT) for the assistant.
    At first (\texttt{<|im\_end|>}), temporal preference is not linearly separable and the no-horizon samples (gray) are disjoint from the off-policy time-horizon manifold.
    As the LLM transitions into the assistant's turn (moving towards \texttt{assistant}), they appear to align to the manifold before preference clusters are formed.\label{fig:turn-transition}%
}

\figtwocolFull[0.8]{%
    \includegraphics[width=\linewidth]{images/characterize/parametric_geometry/parametric_geometry_term_legend}\\[6pt]
    \includegraphics[width=\linewidth]{images/characterize/parametric_geometry/parametric_geometry_time_scale_legend}
}{%
    \twocolrow{images/characterize/parametric_geometry/by_layer/L0__chat_suffix_tail_0__term_chosen}
              {images/characterize/parametric_geometry/by_layer/L0__chat_suffix_tail_0__time_scale}{\texttt{\textbackslash n}}
}{%
    By the token position of the BoT delimiter (\texttt{\textbackslash n} after \texttt{assistant}), temporal preference is separable even at layer 0.
}

\FloatBarrier

\subsection{Horizon representation is present but geometrically unstable in the prompt}\label{app:parametric-geo-unstable}

\begin{figure}[htbp]
\centering
\begin{minipage}[t]{0.48\textwidth}
  \centering
  {\scriptsize\texttt{resid\_post} @ \texttt{post\_time\_horizon}}\\[2pt]
  \includegraphics[width=\textwidth]{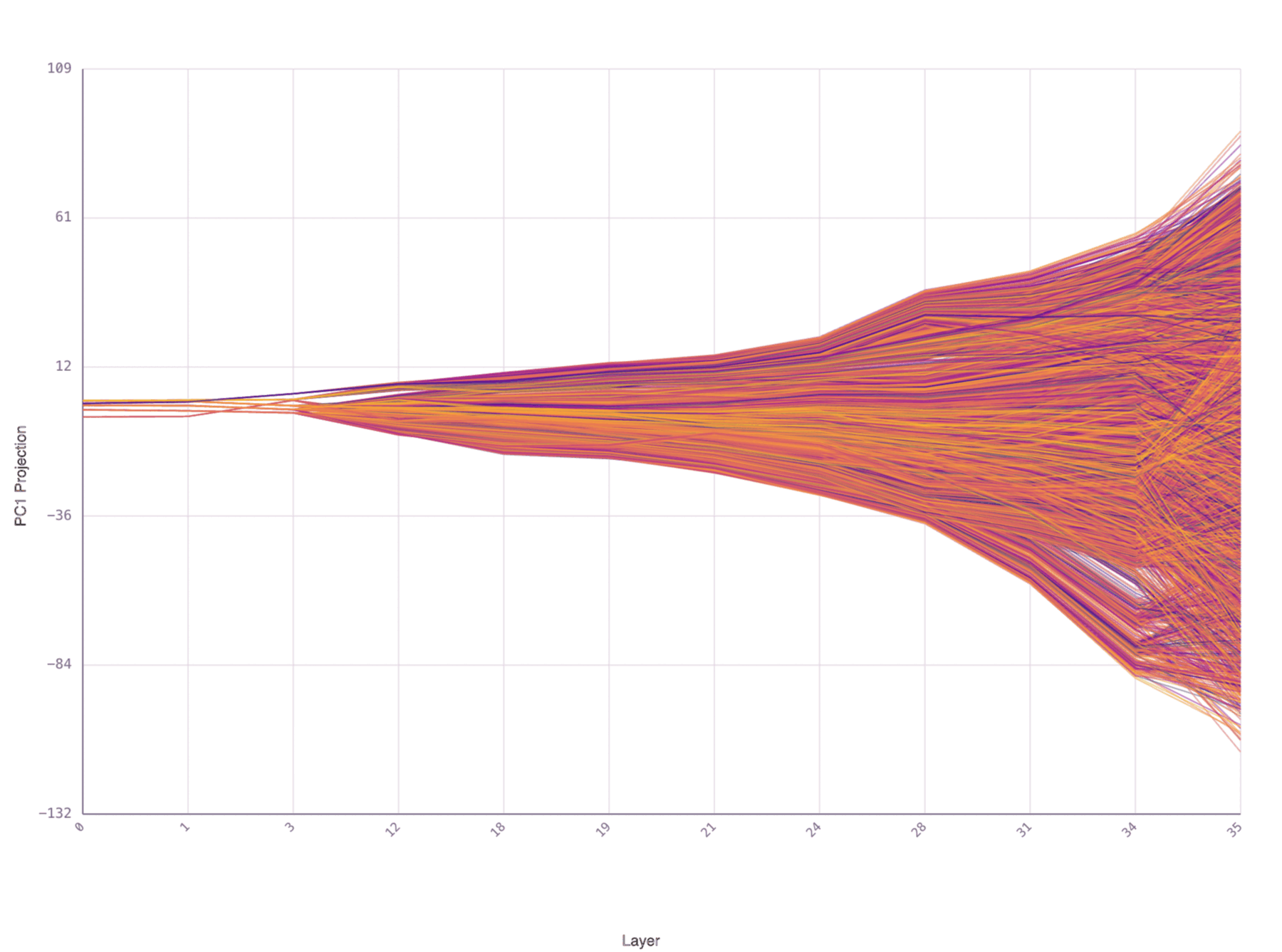}
\end{minipage}\hfill
\begin{minipage}[t]{0.48\textwidth}
  \centering
  {\scriptsize\texttt{resid\_post} @ \texttt{action\_tail}}\\[2pt]
  \includegraphics[width=\textwidth]{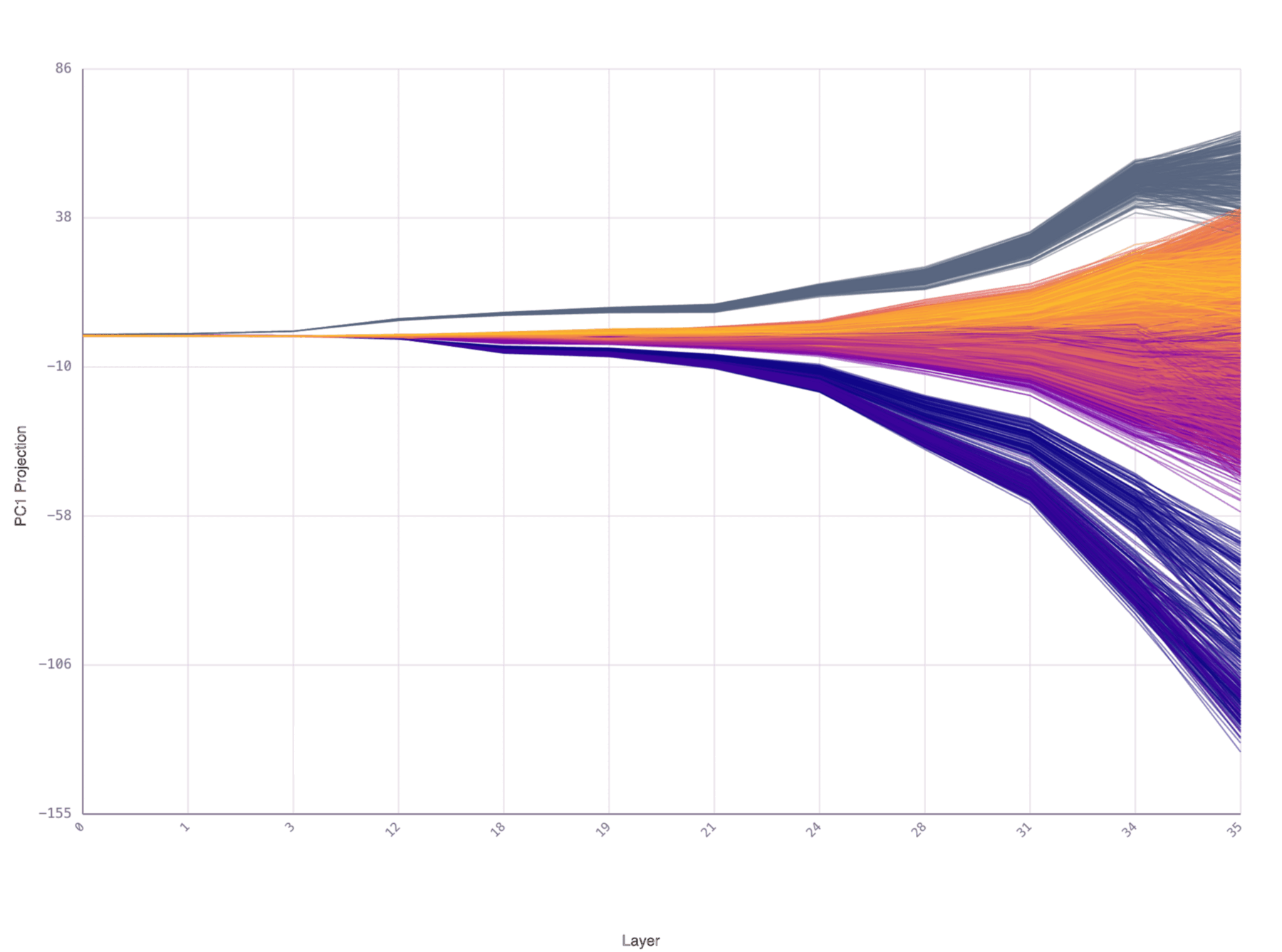}
\end{minipage}

\vspace{4pt}

\begin{minipage}[t]{0.48\textwidth}
  \centering
  {\scriptsize\texttt{resid\_post} @ \texttt{format\_tail}}\\[2pt]
  \includegraphics[width=\textwidth]{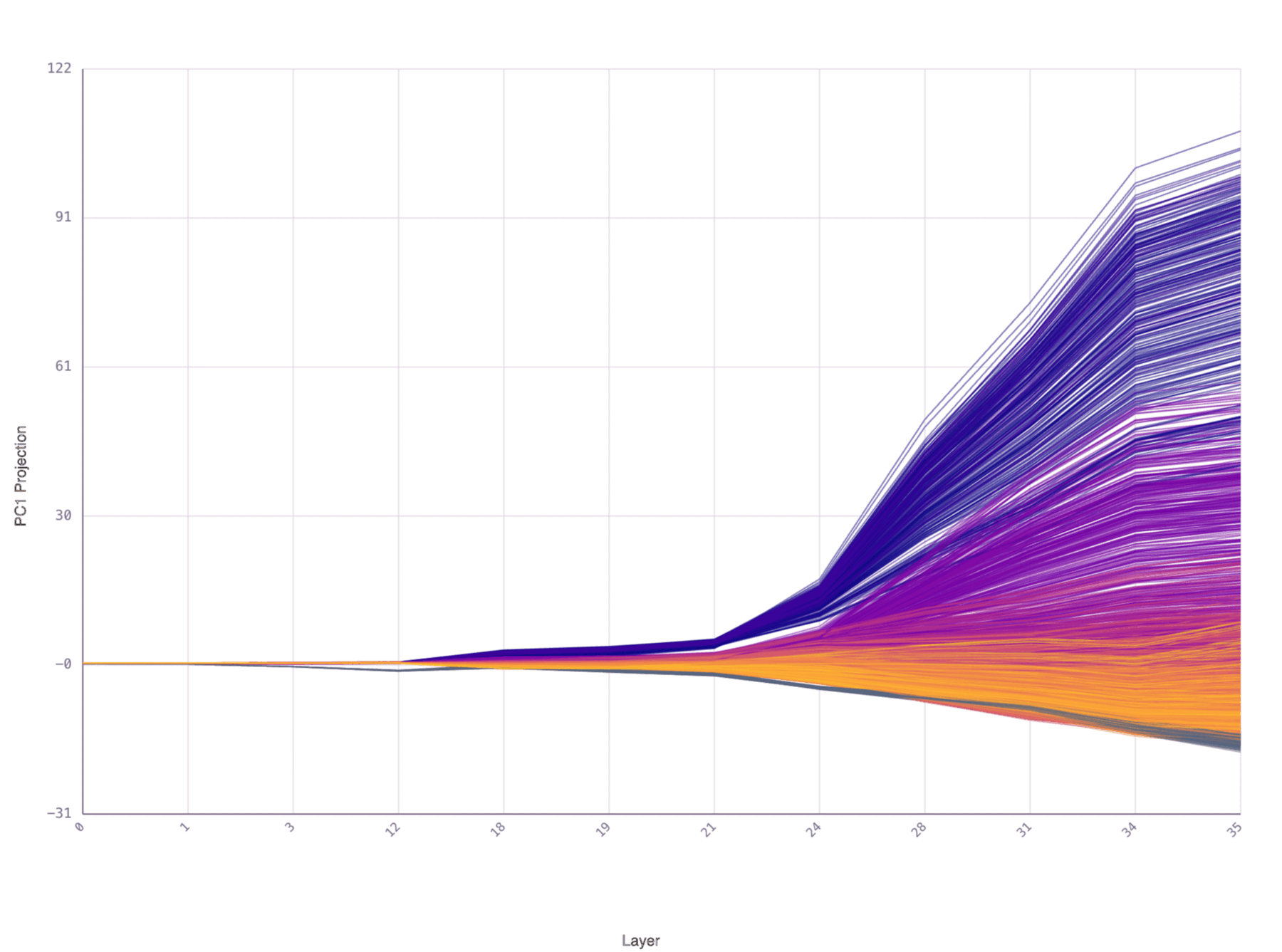}
\end{minipage}\hfill
\begin{minipage}[t]{0.48\textwidth}
  \centering
  {\scriptsize\texttt{resid\_post} @ \texttt{chat\_suffix\_tail}}\\[2pt]
  \includegraphics[width=\textwidth]{images/characterize/parametric_geometry/early_horizon/post_response.png}
\end{minipage}
\caption{PC1 projection of \texttt{resid\_post} across layers at four prompt positions, colored by time horizon (blue = seconds, yellow = deep time).
Top-left: after the time horizon constraint token (within the \texttt{CONSTRAINT} section).
Top-right: last token of the \texttt{ACTION} section.
Bottom-left: last token of the \texttt{FORMAT} section.
Bottom-right: \texttt{chat\_suffix\_tail} (the \texttt{\textbackslash n} after \texttt{assistant}).
The ordinal fan is present at all positions, but its \emph{polarity flips} between positions (short horizons go negative at some, positive at others), indicating the geometric direction encoding horizon is not yet stable within the prompt.}
\label{fig:early-horizon}
\end{figure}

Even before the model starts generating, the residual stream encodes time horizon.
Figure~\ref{fig:early-horizon} shows the PC1 projection at several positions within the prompt, colored by horizon category.

\begin{figure}[htbp]
\centering
\begin{minipage}[t]{0.48\textwidth}
  \centering
  {\scriptsize\texttt{resid\_post} @ \texttt{response\_choice} (\texttt{a/b})}\\[2pt]
  \includegraphics[width=\textwidth]{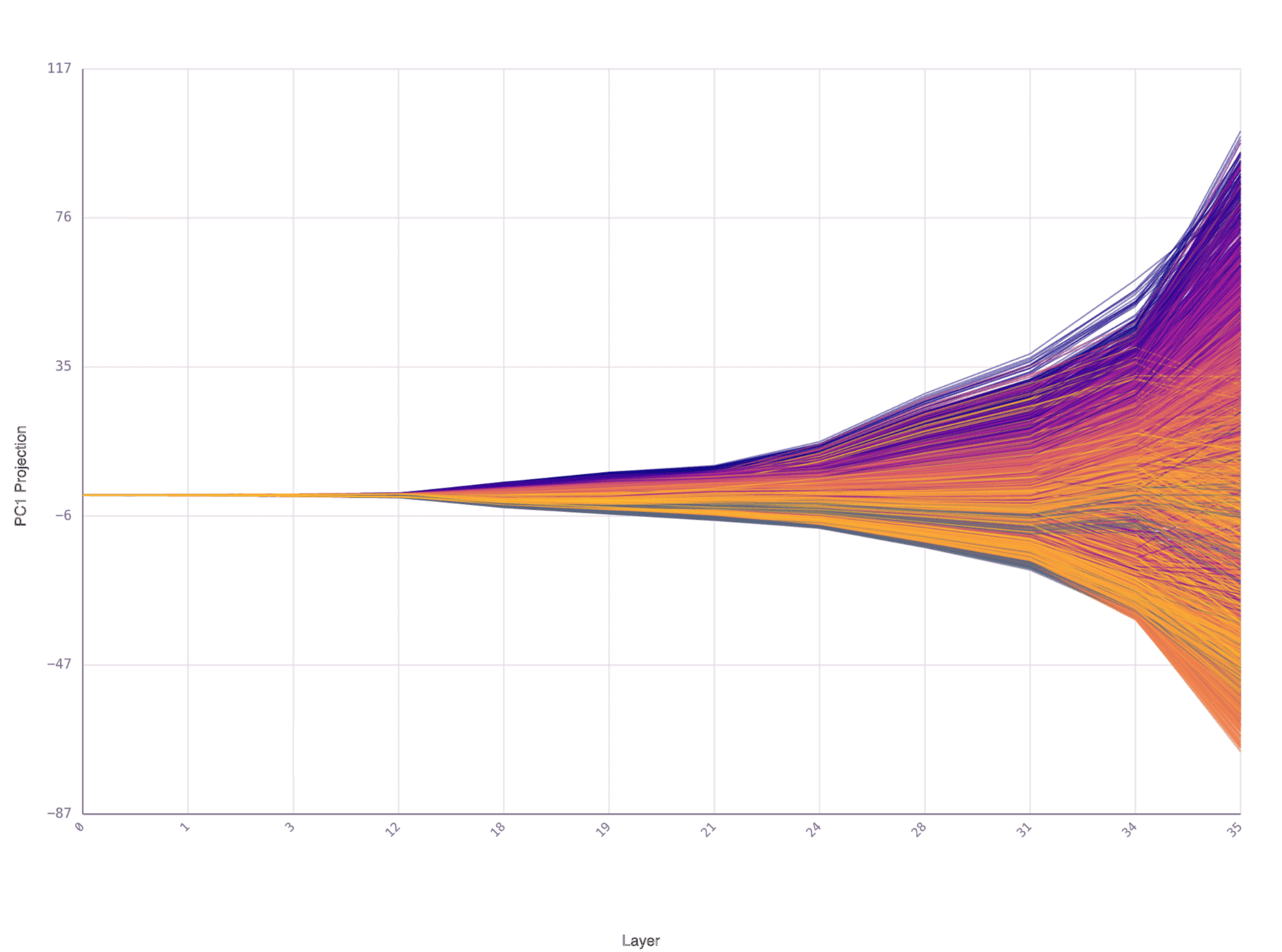}
\end{minipage}\hfill
\begin{minipage}[t]{0.48\textwidth}
  \centering
  {\scriptsize\texttt{resid\_post} @ \texttt{response\_choice\_prefix} (\texttt{choose})}\\[2pt]
  \includegraphics[width=\textwidth]{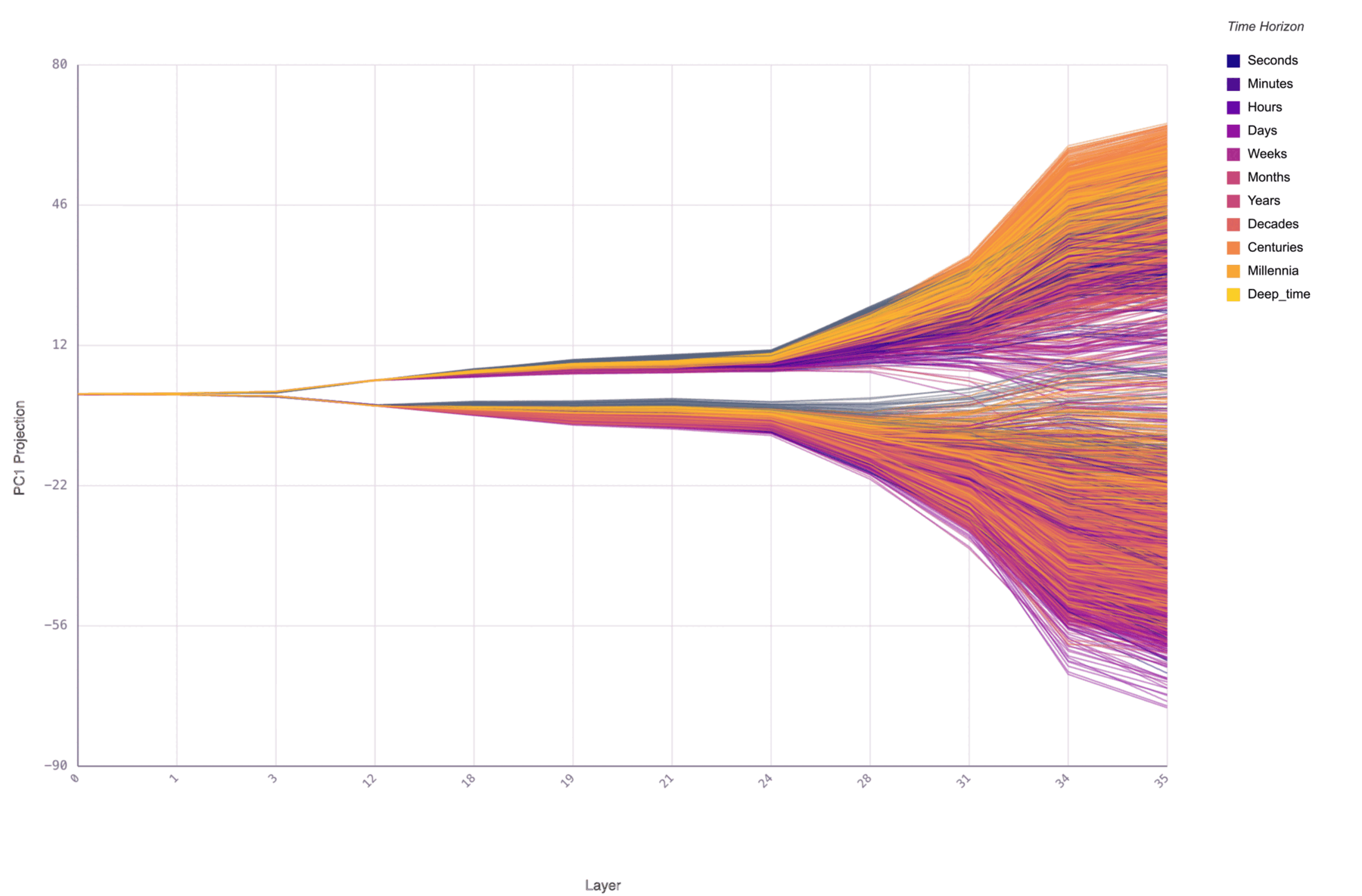}
\end{minipage}
\caption{PC1 projection of \texttt{resid\_post} at response positions, colored by time horizon.
Left: \texttt{response\_choice} (the \texttt{a)} or \texttt{b)} token).
Right: \texttt{response\_choice\_prefix} (the \texttt{choose} token in ``I choose:'').
The horizon signal is present but the geometric direction has not yet fully stabilized.}
\label{fig:early-horizon-response}
\end{figure}

The horizon signal is large (spreads of $\pm$100 or more on PC1) and ordinally organized at every position, but the direction encoding it rotates across positions.
At the ACTION tail, short horizons go strongly negative; at the FORMAT tail, the polarity flips and short horizons go positive.
This instability persists into the earliest response tokens (Figure~\ref{fig:early-horizon-response}).

\noindent The model has the horizon information throughout, but has not committed to a stable geometric encoding of it until the turn boundary.

\FloatBarrier
\subsection{Stabilization at the turn boundary (residual stream)}\label{app:parametric-geo-turn}

\begin{figure}[htbp]
\centering
\begin{minipage}[t]{0.32\textwidth}
  \centering
  {\scriptsize\texttt{resid\_post}, suffix 0, horizon}\\[2pt]
  \includegraphics[width=\textwidth]{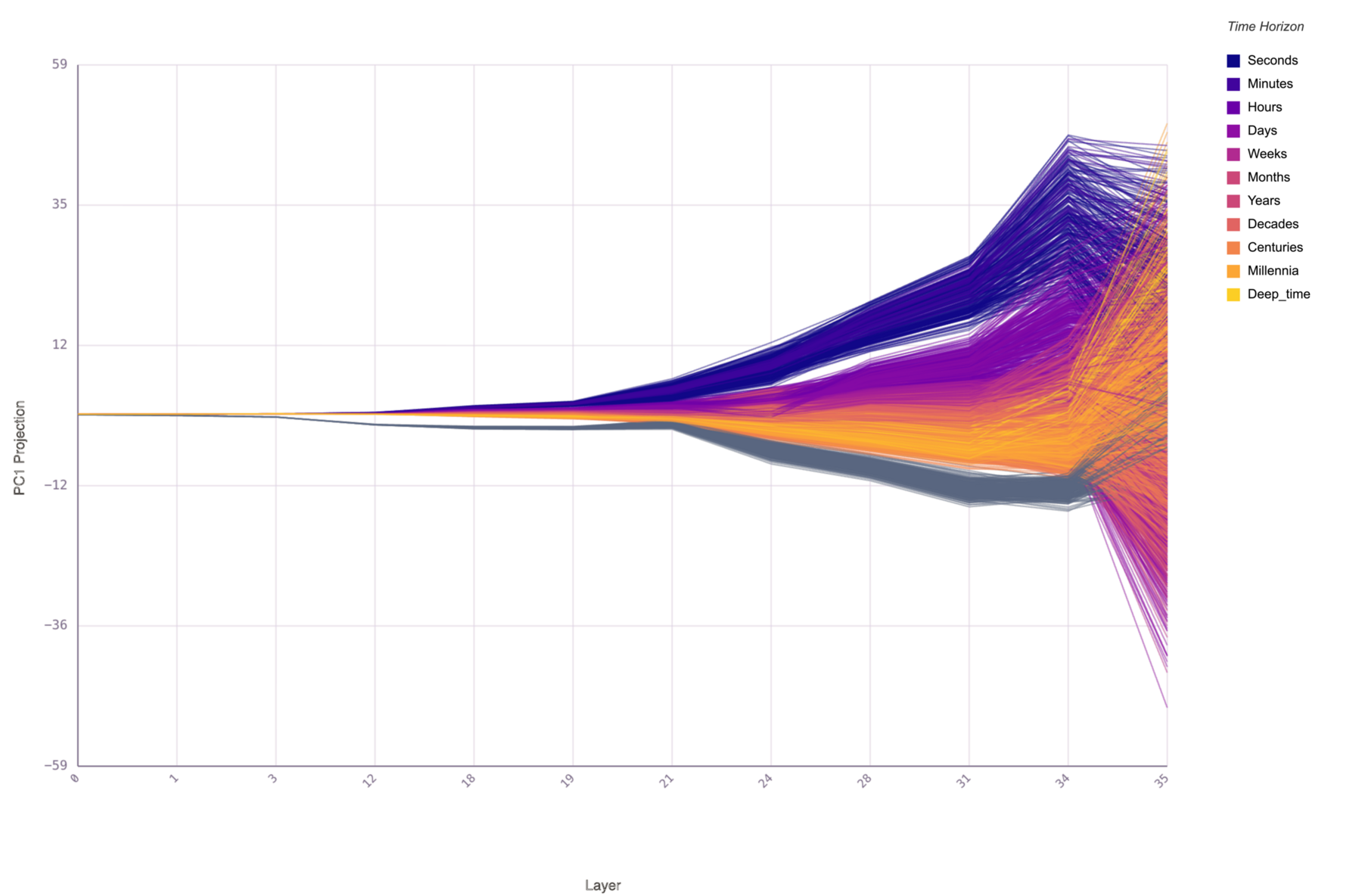}
\end{minipage}\hfill
\begin{minipage}[t]{0.32\textwidth}
  \centering
  {\scriptsize\texttt{resid\_post}, suffix 0, preference}\\[2pt]
  \includegraphics[width=\textwidth]{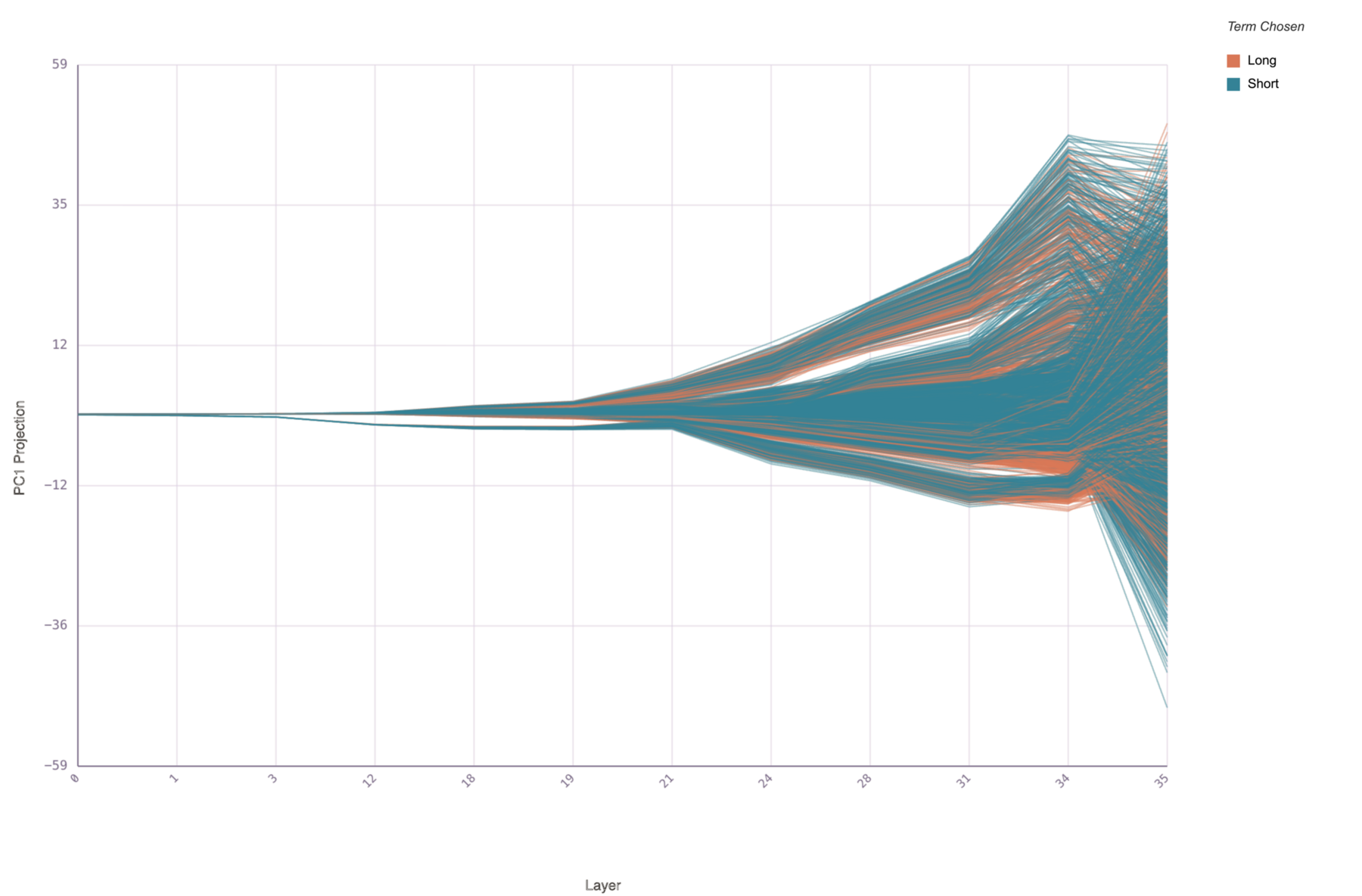}
\end{minipage}\hfill
\begin{minipage}[t]{0.32\textwidth}
  \centering
  {\scriptsize\texttt{resid\_post}, suffix 3, preference}\\[2pt]
  \includegraphics[width=\textwidth]{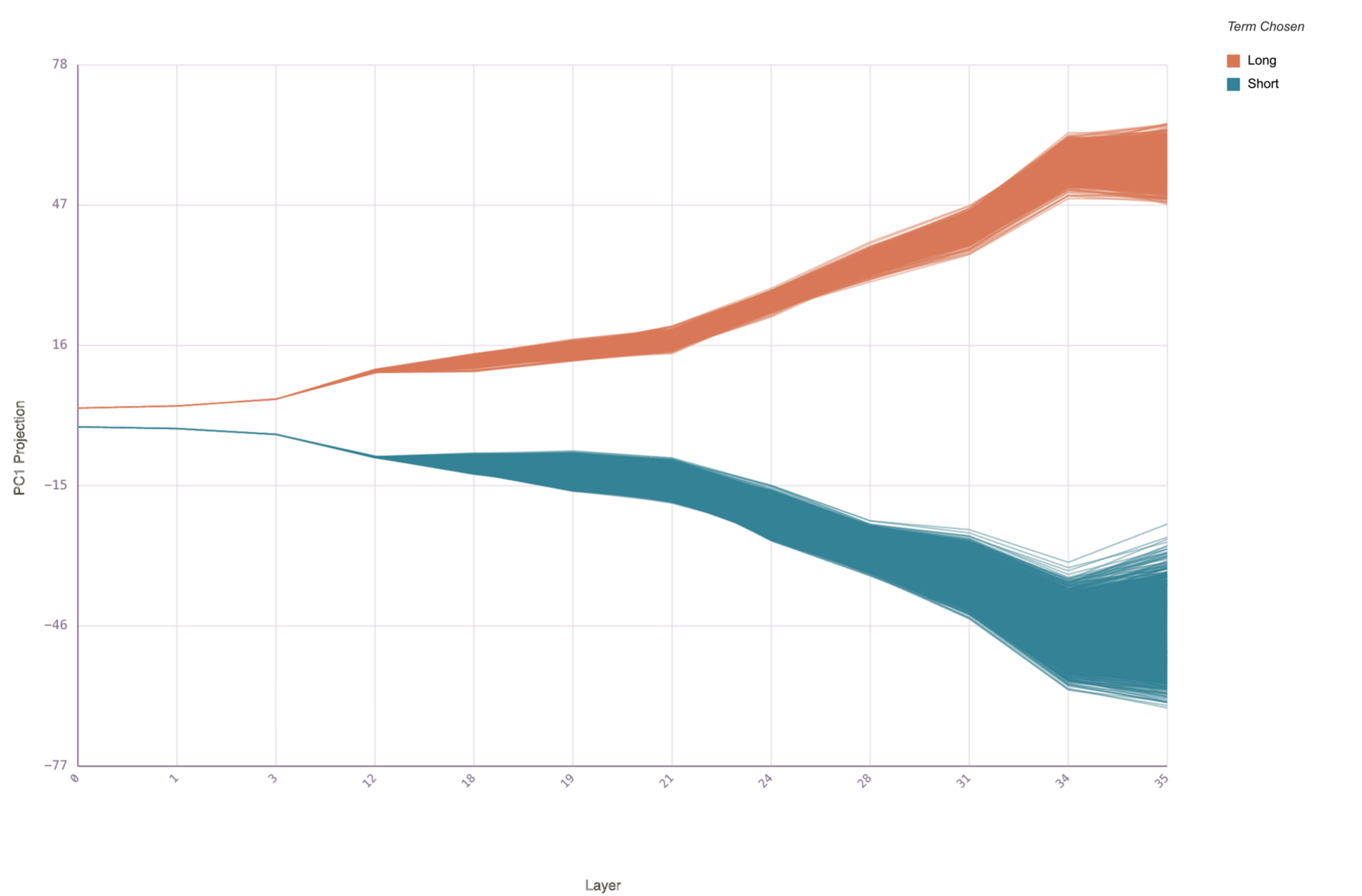}
\end{minipage}
\caption{\texttt{resid\_post} PC1 projection at the turn boundary.
Left: suffix 0 (\texttt{<|im\_end|>}), colored by time horizon.
The ordinal fan is now stable and monotonic, with short horizons trending negative and long horizons positive.
Center: same position, colored by preference.
Long and short are heavily overlapping: the choice has not yet been made.
Right: suffix 3 (\texttt{assistant} token), colored by preference.
Long and short are cleanly separated from early layers onward.
Between these two positions, the model converts the stable horizon representation into a committed preference.}
\label{fig:change-of-turn}
\end{figure}

The representation stabilizes at the user-to-assistant turn boundary.
Figure~\ref{fig:change-of-turn} shows \texttt{resid\_post} at three key positions in the turn transition.

At suffix 0, the residual stream carries a clean, ordinal horizon representation (left), but long and short preferences overlap completely (center).
The geometry at this position encodes \emph{how far into the future}, not \emph{which option to choose}.
By suffix 3, the preference is fully committed (right): long and short form two non-overlapping bands from early layers onward.

\begin{figure}[htbp]
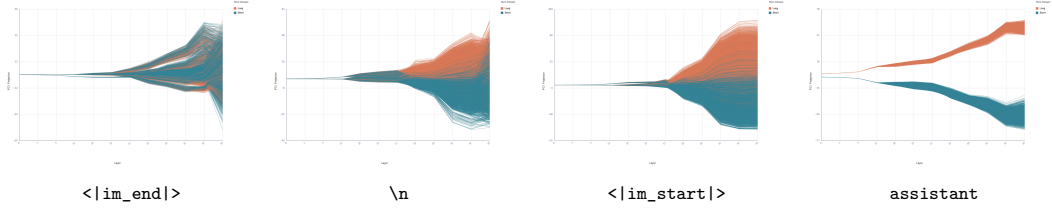

\centering
\begin{minipage}[t]{0.24\textwidth}
  \centering
  \includegraphics[width=\textwidth]{images/characterize/parametric_geometry/change_of_turn/change_of_turn_suffix0_preference.png}\\[2pt]
  {\scriptsize\texttt{<|im\_end|>}}
\end{minipage}\hfill
\begin{minipage}[t]{0.24\textwidth}
  \centering
  \includegraphics[width=\textwidth]{images/characterize/parametric_geometry/change_of_turn/change_of_turn_suffix1_preference.png}\\[2pt]
  {\scriptsize\texttt{\textbackslash n}}
\end{minipage}\hfill
\begin{minipage}[t]{0.24\textwidth}
  \centering
  \includegraphics[width=\textwidth]{images/characterize/parametric_geometry/change_of_turn/change_of_turn_suffix2_preference.png}\\[2pt]
  {\scriptsize\texttt{<|im\_start|>}}
\end{minipage}\hfill
\begin{minipage}[t]{0.24\textwidth}
  \centering
  \includegraphics[width=\textwidth]{images/characterize/parametric_geometry/change_of_turn/change_of_turn_suffix3_preference.png}\\[2pt]
  {\scriptsize\texttt{assistant}}
\end{minipage}
\caption{\texttt{resid\_post} at suffix positions 0 through 3, all colored by preference (orange = long, blue = short).
The preference signal progressively sharpens from heavy overlap at suffix 0 to clean separation at suffix 3.}
\label{fig:change-of-turn-full}
\end{figure}

The complete four-position transition in the residual stream (suffix 0 through 3) is shown in Figure~\ref{fig:change-of-turn-full}.

\FloatBarrier

\FloatBarrier
\subsection{Attention mediates the horizon-to-preference transformation}\label{app:parametric-geo-attention}

\begin{figure}[htbp]
\centering
\begin{minipage}[t]{0.40\textwidth}
  \centering
  {\scriptsize\texttt{attn\_out}, suffix 0 (\texttt{<|im\_end|>}), horizon}\\[2pt]
  \includegraphics[width=\textwidth]{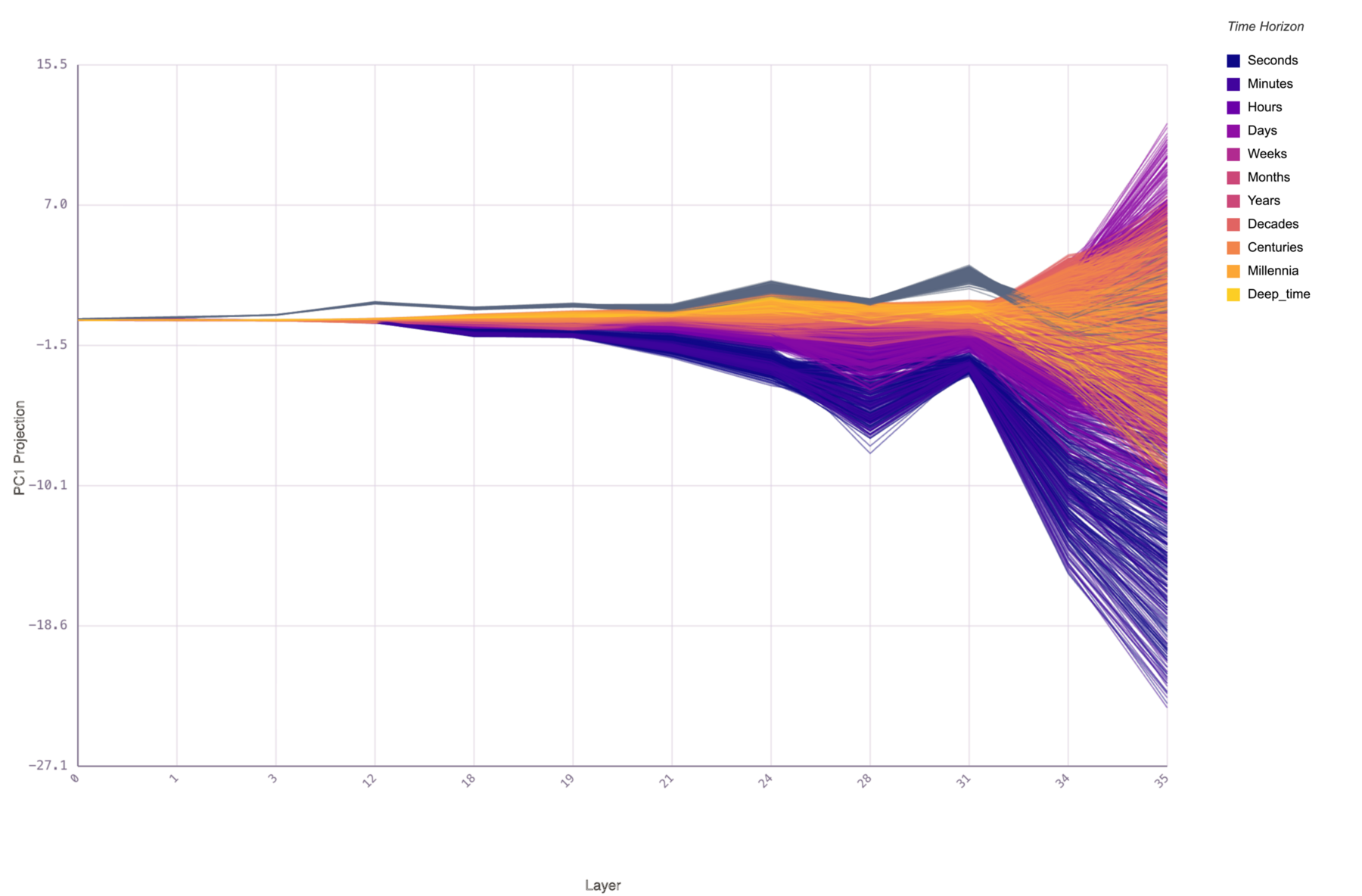}
\end{minipage}\hfill
\begin{minipage}[t]{0.40\textwidth}
  \centering
  {\scriptsize\texttt{attn\_out}, suffix 0 (\texttt{<|im\_end|>}), preference}\\[2pt]
  \includegraphics[width=\textwidth]{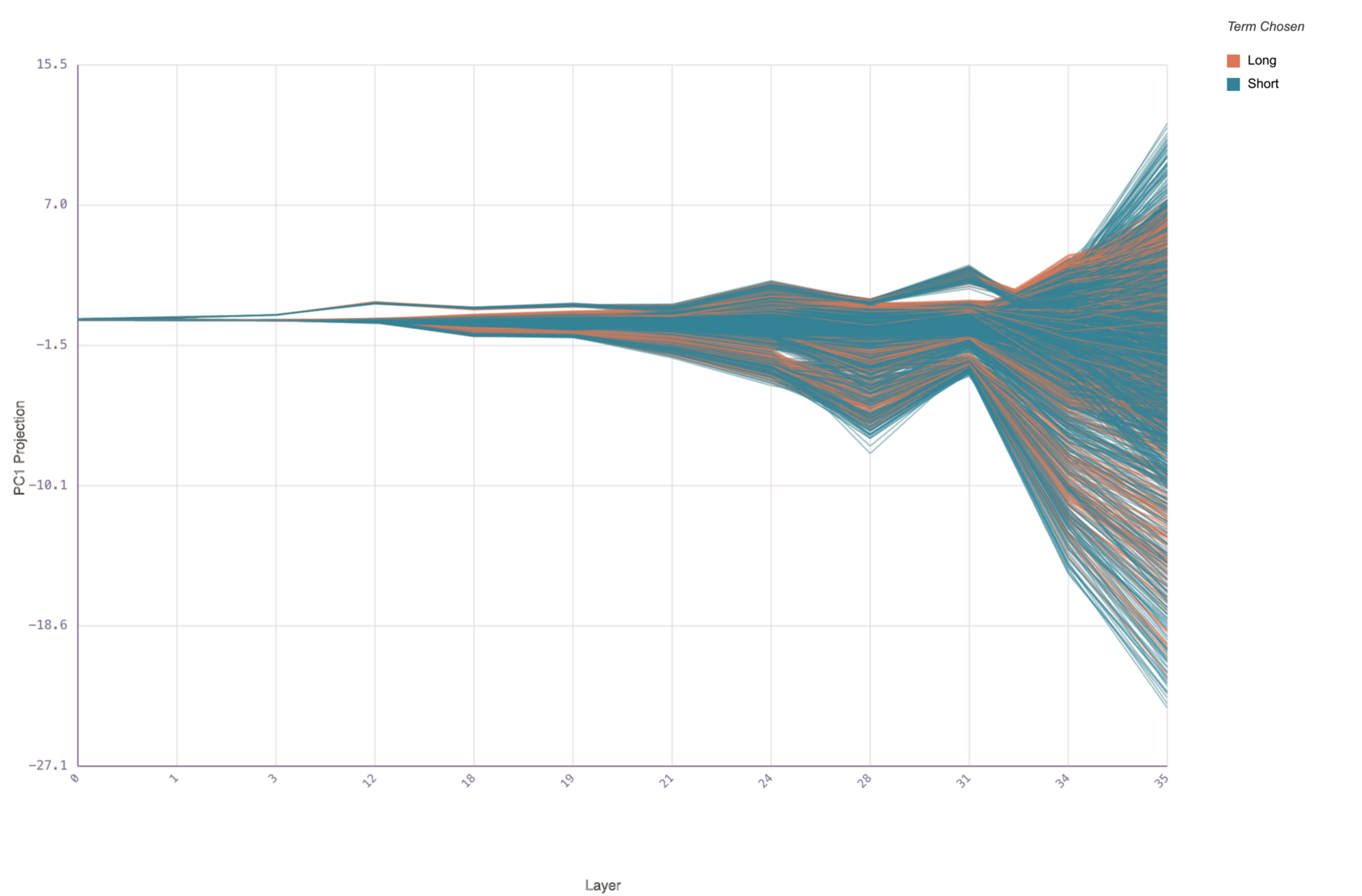}
\end{minipage}

\vspace{4pt}

\begin{minipage}[t]{0.40\textwidth}
  \centering
  {\scriptsize\texttt{attn\_out}, suffix 1 (\texttt{\textbackslash n}), horizon}\\[2pt]
  \includegraphics[width=\textwidth]{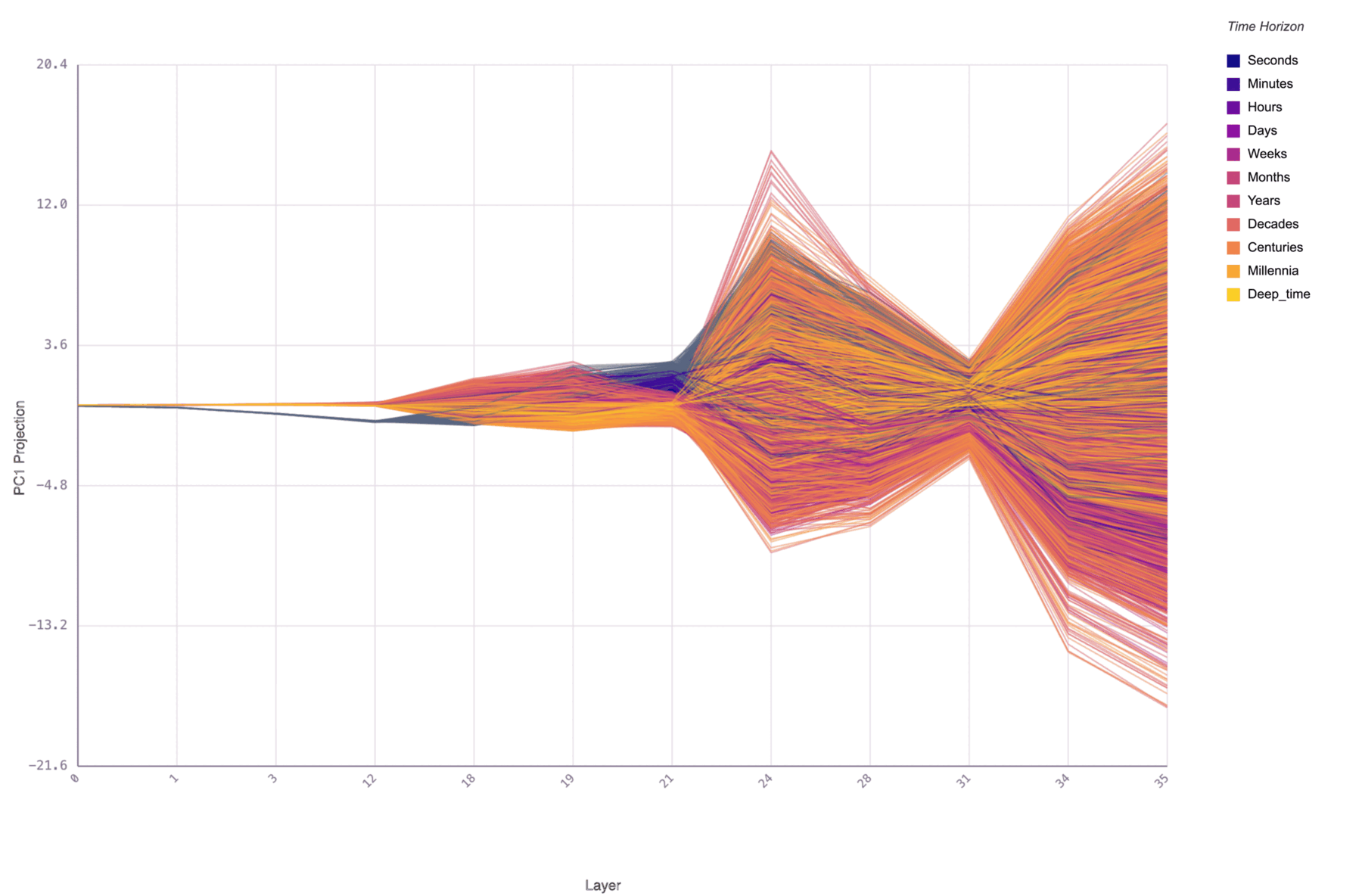}
\end{minipage}\hfill
\begin{minipage}[t]{0.40\textwidth}
  \centering
  {\scriptsize\texttt{attn\_out}, suffix 1 (\texttt{\textbackslash n}), preference}\\[2pt]
  \includegraphics[width=\textwidth]{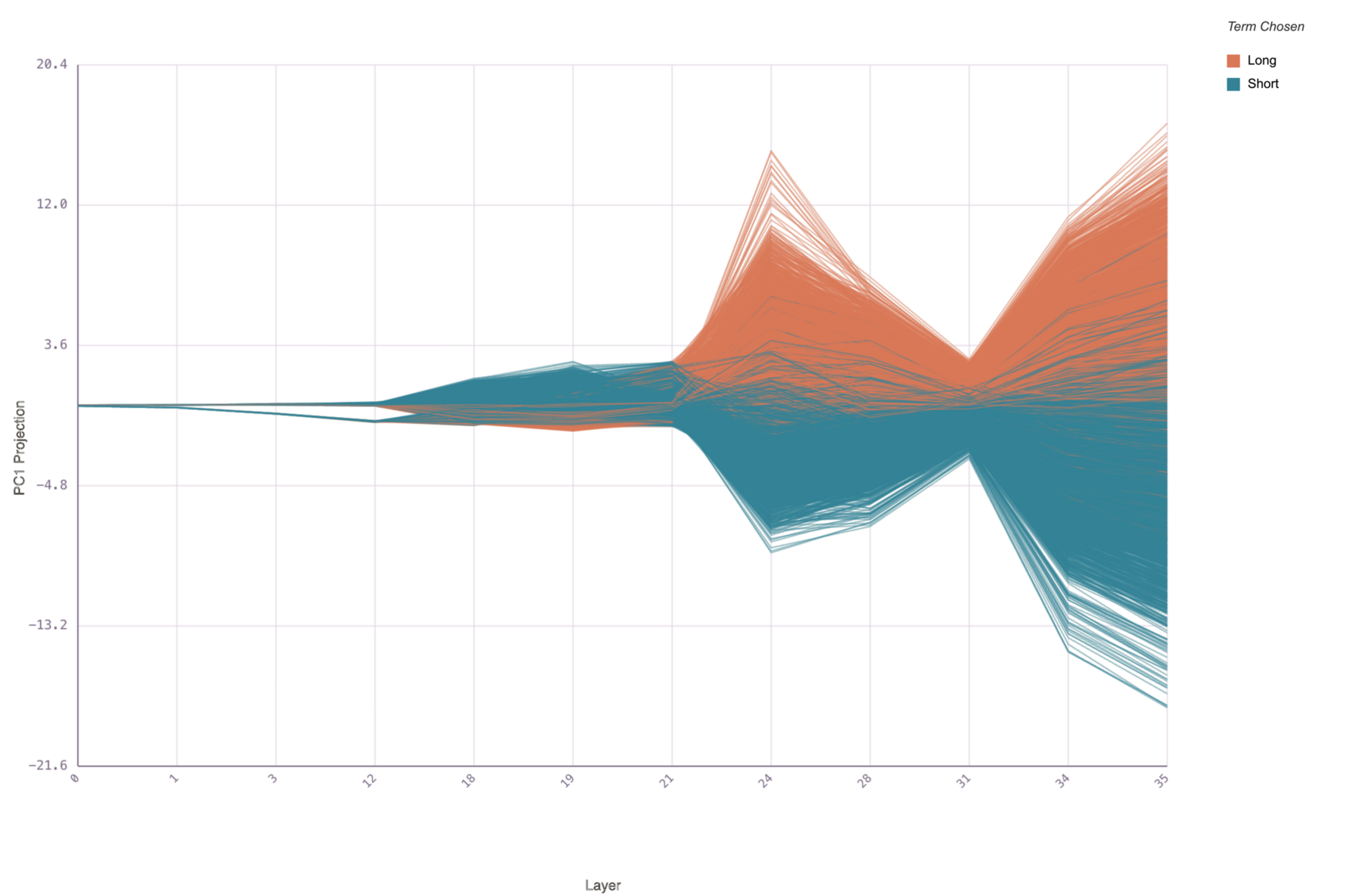}
\end{minipage}

\vspace{4pt}

\begin{minipage}[t]{0.40\textwidth}
  \centering
  {\scriptsize\texttt{attn\_out}, suffix 2 (\texttt{<|im\_start|>}), horizon}\\[2pt]
  \includegraphics[width=\textwidth]{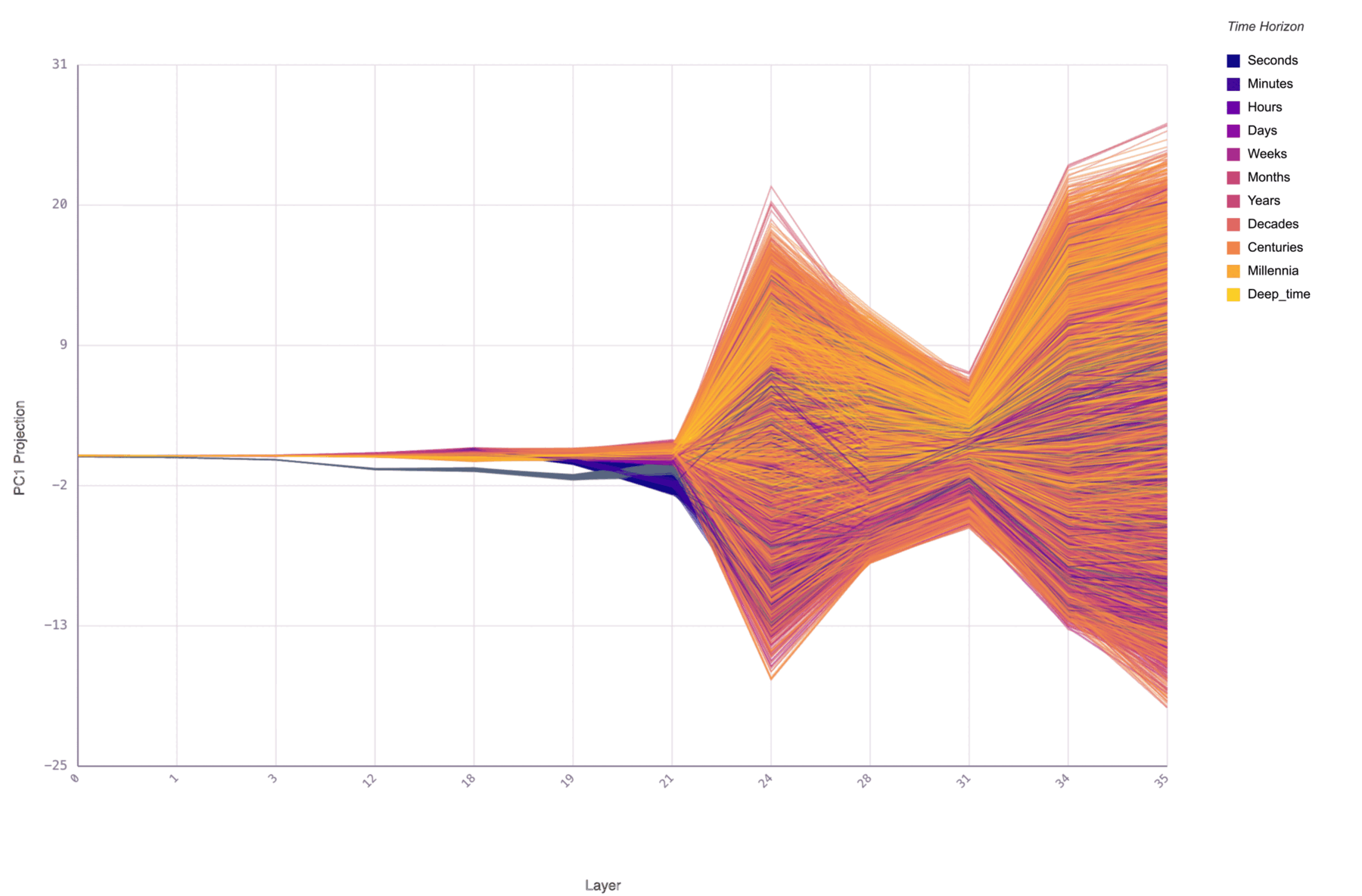}
\end{minipage}\hfill
\begin{minipage}[t]{0.40\textwidth}
  \centering
  {\scriptsize\texttt{attn\_out}, suffix 2 (\texttt{<|im\_start|>}), preference}\\[2pt]
  \includegraphics[width=\textwidth]{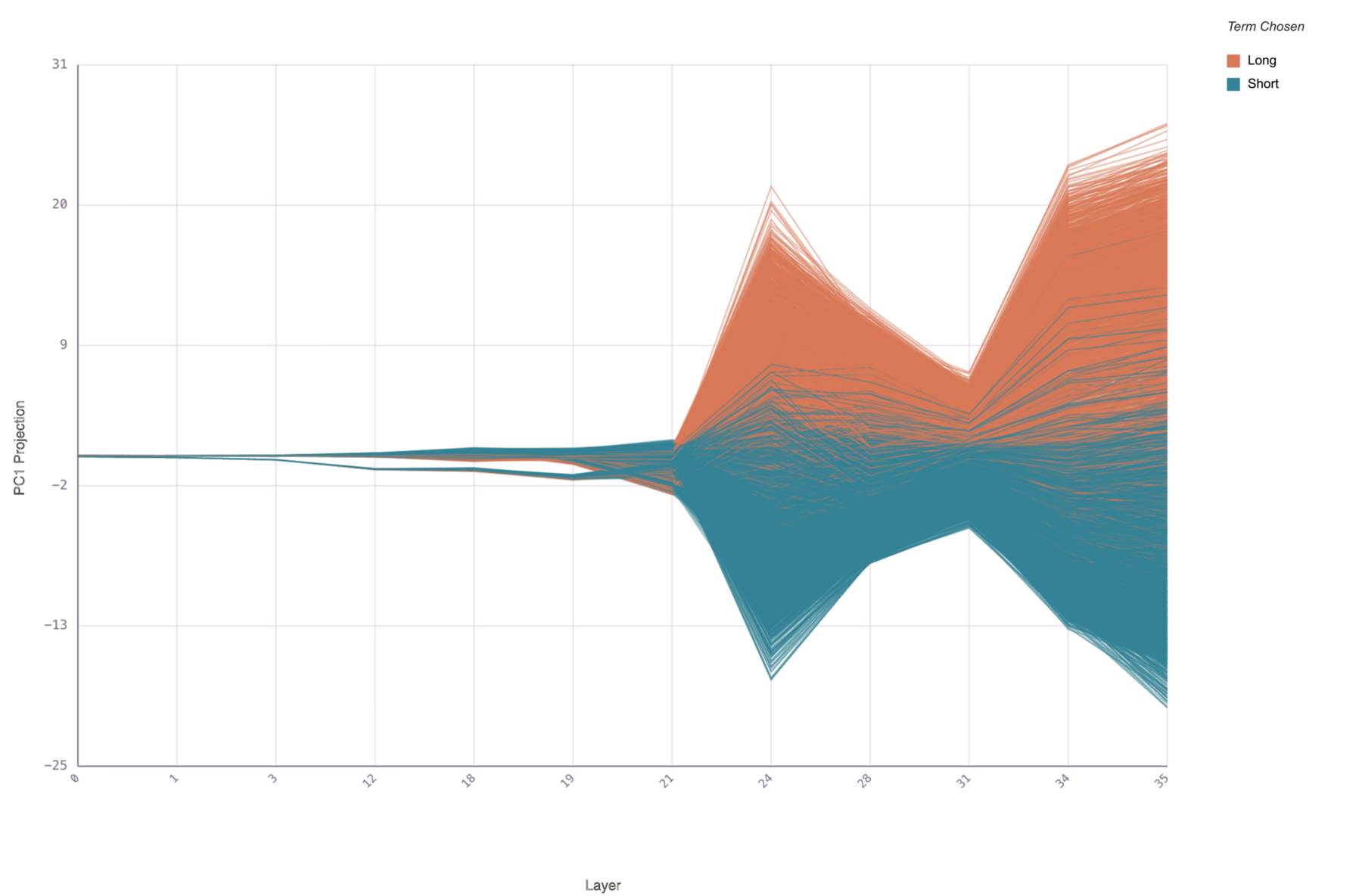}
\end{minipage}

\vspace{4pt}

\begin{minipage}[t]{0.40\textwidth}
  \centering
  {\scriptsize\texttt{attn\_out}, suffix 3 (\texttt{assistant}), horizon}\\[2pt]
  \includegraphics[width=\textwidth]{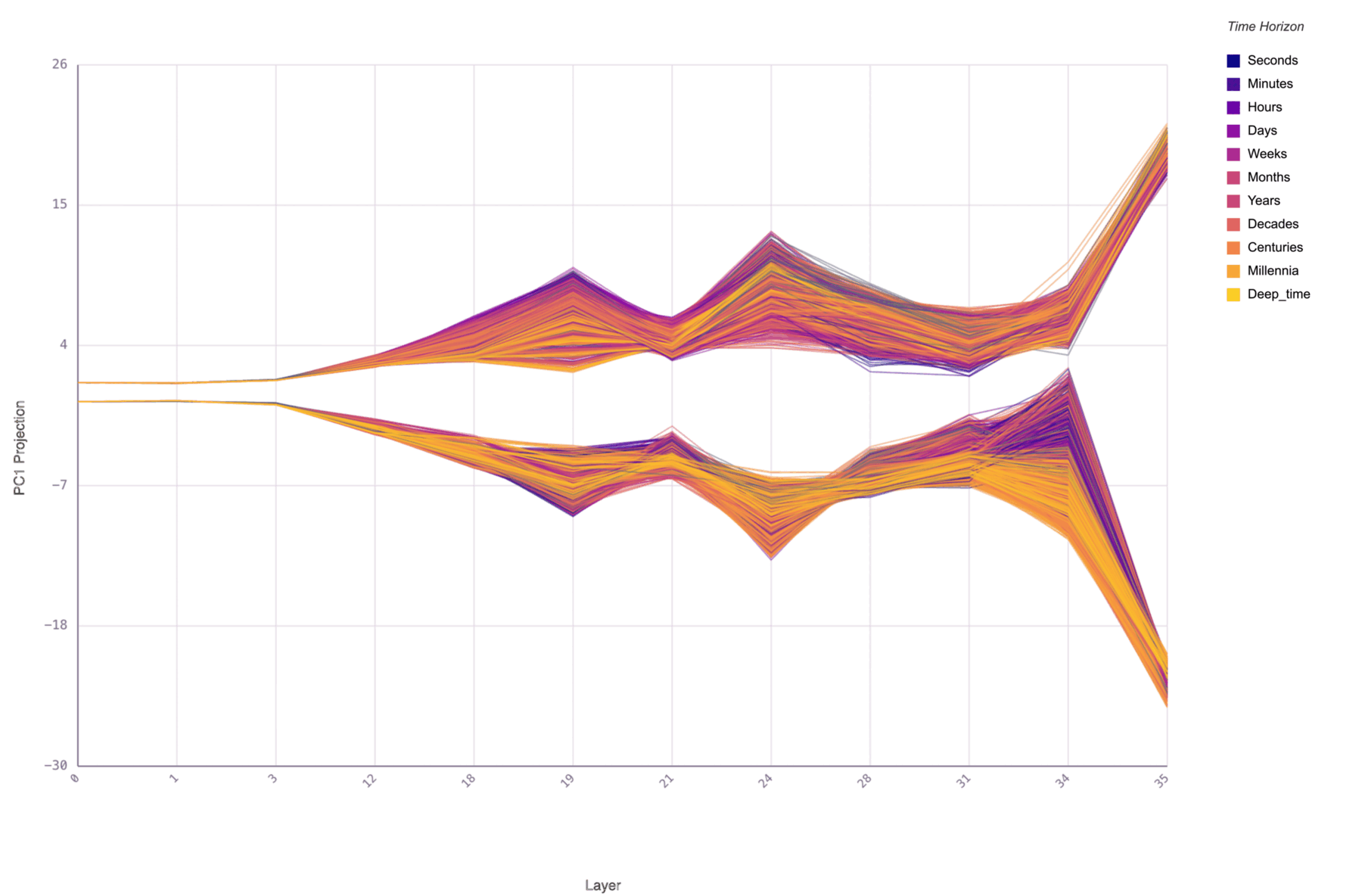}
\end{minipage}\hfill
\begin{minipage}[t]{0.40\textwidth}
  \centering
  {\scriptsize\texttt{attn\_out}, suffix 3 (\texttt{assistant}), preference}\\[2pt]
  \includegraphics[width=\textwidth]{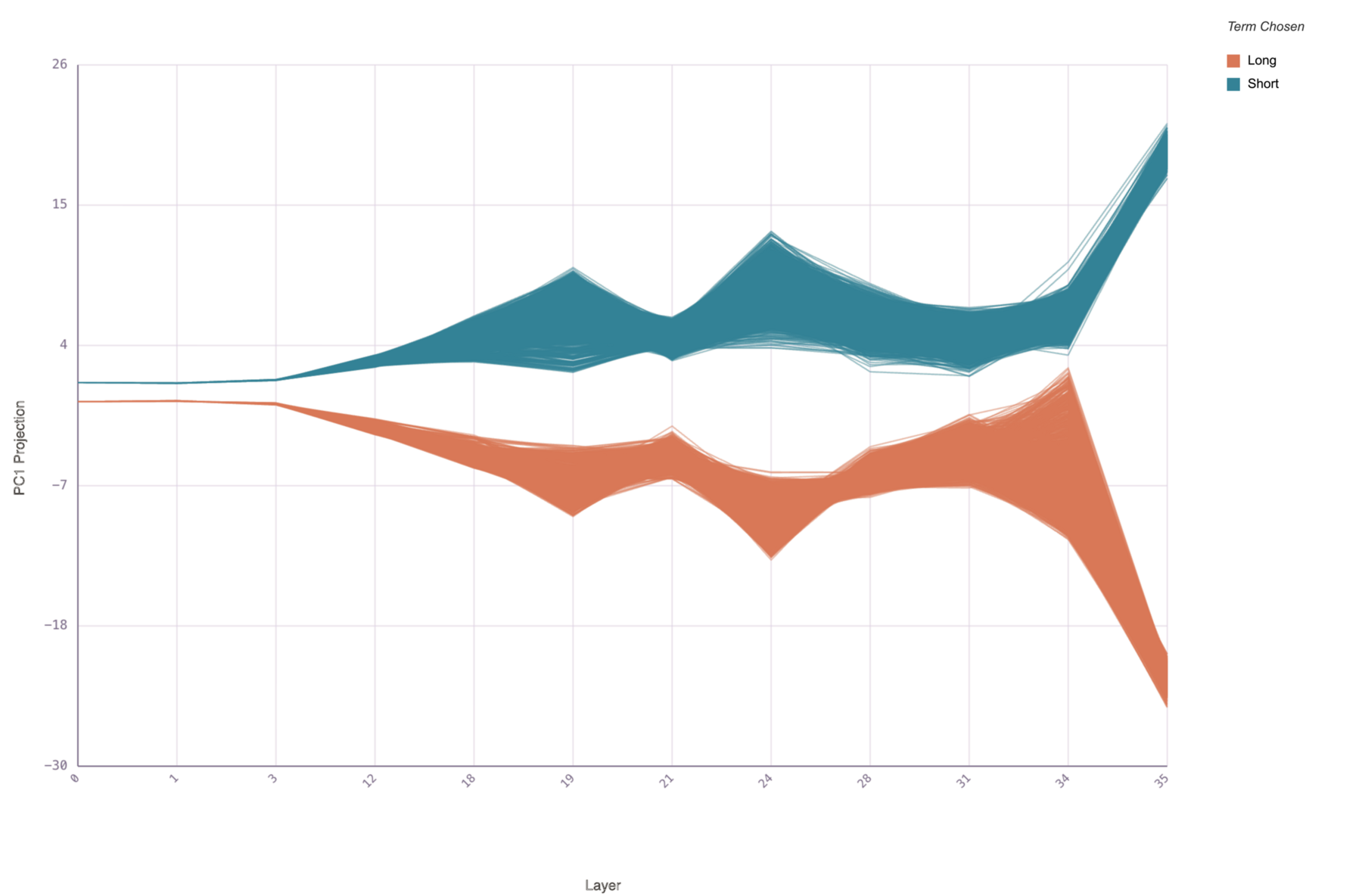}
\end{minipage}
\caption{\texttt{attn\_out} PC1 projection across layers at suffix positions 0--3 (rows), colored by time horizon (left column) and preference (right column).
At suffix 0, attention carries horizon structure but no preference separation.
At suffix 1, a noisy preference signal emerges with a characteristic V-shape around layers 13--24, indicating active reorganization.
At suffix 2, the preference separation strengthens.
By suffix 3, preference is clearly separated in the attention output.
The non-monotonic trajectories at suffix 1--2 (unlike the smooth residual-stream fans) reveal that attention is actively \emph{transforming} the representation, not merely amplifying a pre-existing signal.}
\label{fig:attention-transform}
\end{figure}

To isolate the mechanism driving the conversion, Figure~\ref{fig:attention-transform} shows \texttt{attn\_out} (the attention output only, before it is added to the residual stream) at all four suffix positions.

The attention output at suffix 0 (top row) carries ordinal horizon structure (left) but no preference signal (right: long and short are intermingled).
At suffix 1, the attention output shows a distinctive non-monotonic, V-shaped trajectory in the mid-layers (13--24).
This zigzag pattern, absent in the smooth residual-stream fans, reveals that attention heads are actively reorganizing the representation.
A noisy preference signal begins to emerge (right).
At suffix 2, the preference separation strengthens, and by suffix 3, long and short are cleanly separated in the attention output.

This progression identifies attention as the operation that converts the stable horizon representation (written into the residual stream by suffix 0) into a preference signal, incrementally across suffix positions 1--3.

\FloatBarrier
\subsection{3D trajectories: horizon becomes preference}\label{app:parametric-geo-3d}

\begin{figure}[!htbp]
\centering
\begin{minipage}[t]{0.35\textwidth}
  \centering
  {\scriptsize cross-layer, suffix 0, horizon}\\[1pt]
  \includegraphics[width=\textwidth]{images/characterize/parametric_geometry/horizon_to_preference/horizon_to_preference_suffix0_horizon.png}
\end{minipage}\hfill
\begin{minipage}[t]{0.35\textwidth}
  \centering
  {\scriptsize cross-layer, suffix 0, preference}\\[1pt]
  \includegraphics[width=\textwidth]{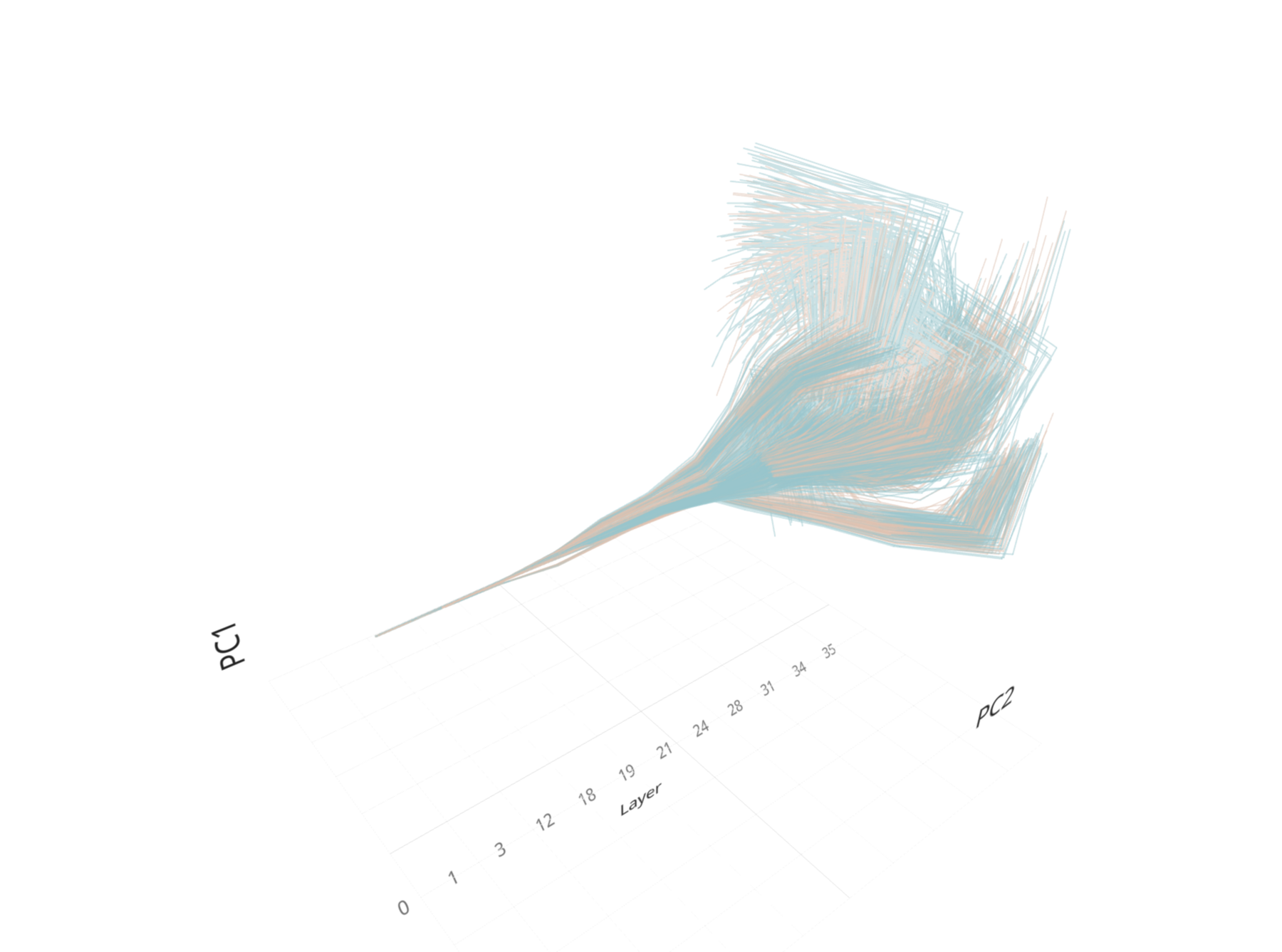}
\end{minipage}

\vspace{2pt}

\begin{minipage}[t]{0.35\textwidth}
  \centering
  {\scriptsize cross-layer, suffix 1, horizon}\\[1pt]
  \includegraphics[width=\textwidth]{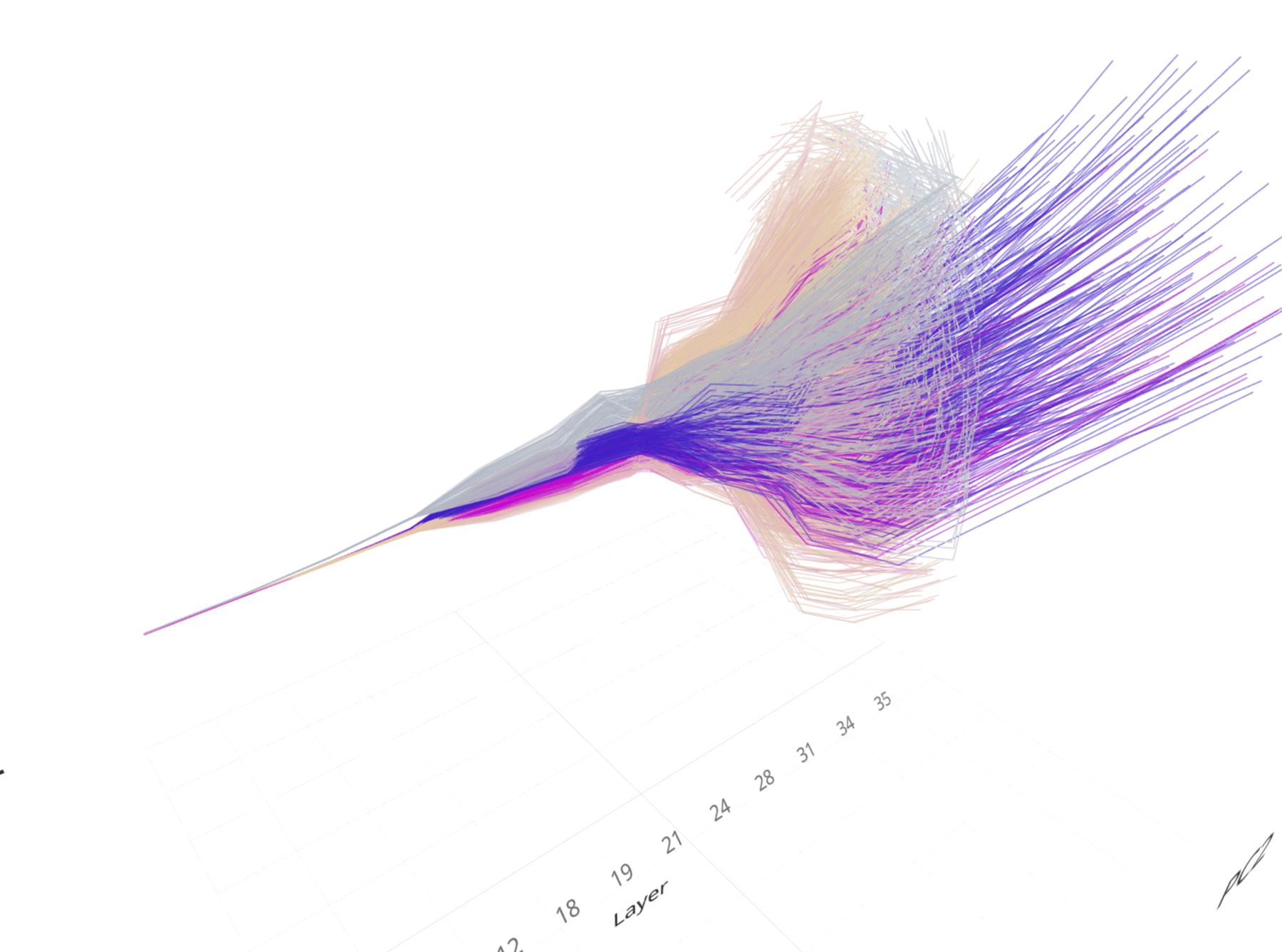}
\end{minipage}\hfill
\begin{minipage}[t]{0.35\textwidth}
  \centering
  {\scriptsize cross-layer, suffix 1, preference}\\[1pt]
  \includegraphics[width=\textwidth]{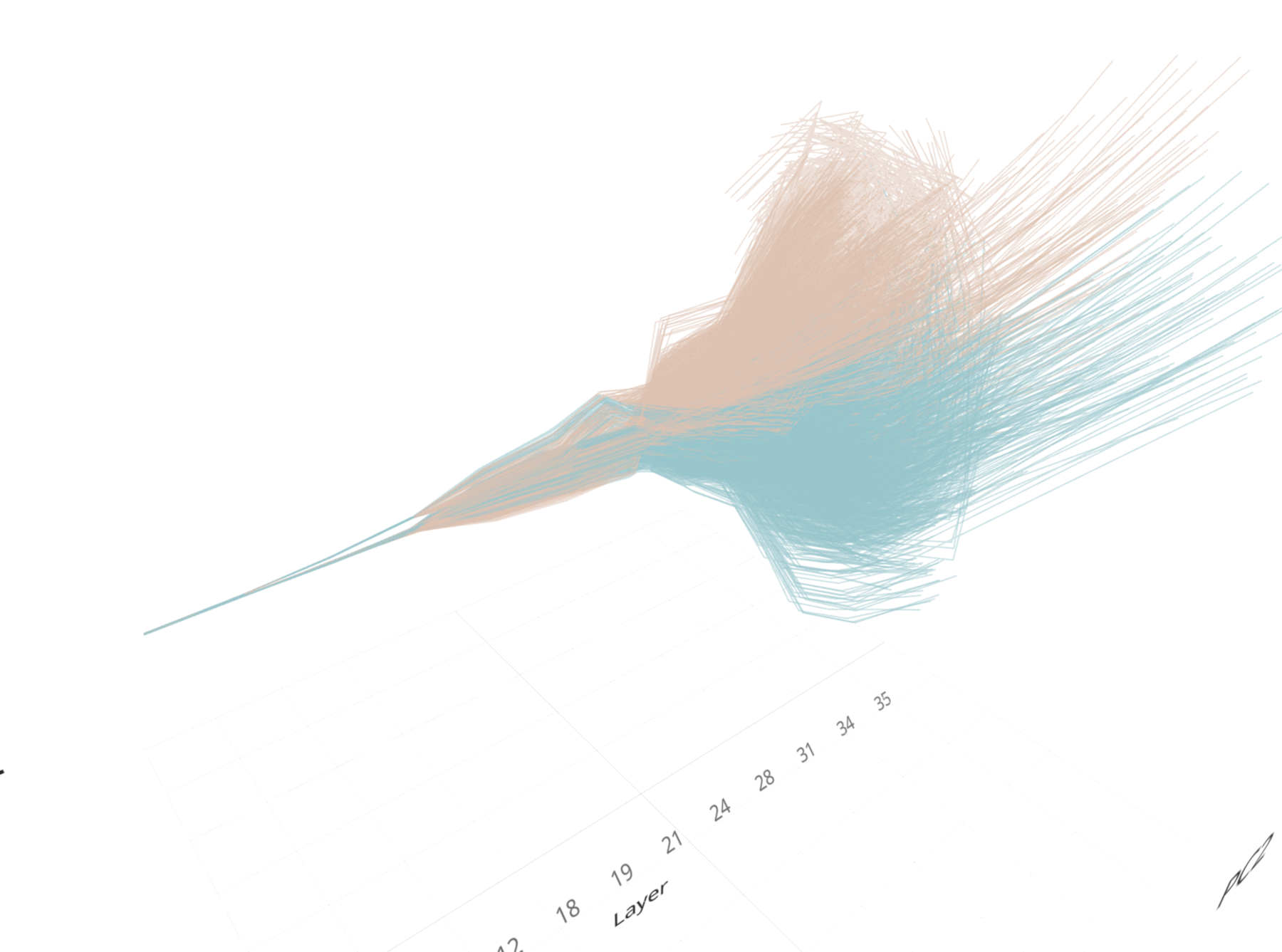}
\end{minipage}

\vspace{2pt}

\begin{minipage}[t]{0.35\textwidth}
  \centering
  {\scriptsize cross-layer, suffix 2, horizon}\\[1pt]
  \includegraphics[width=\textwidth]{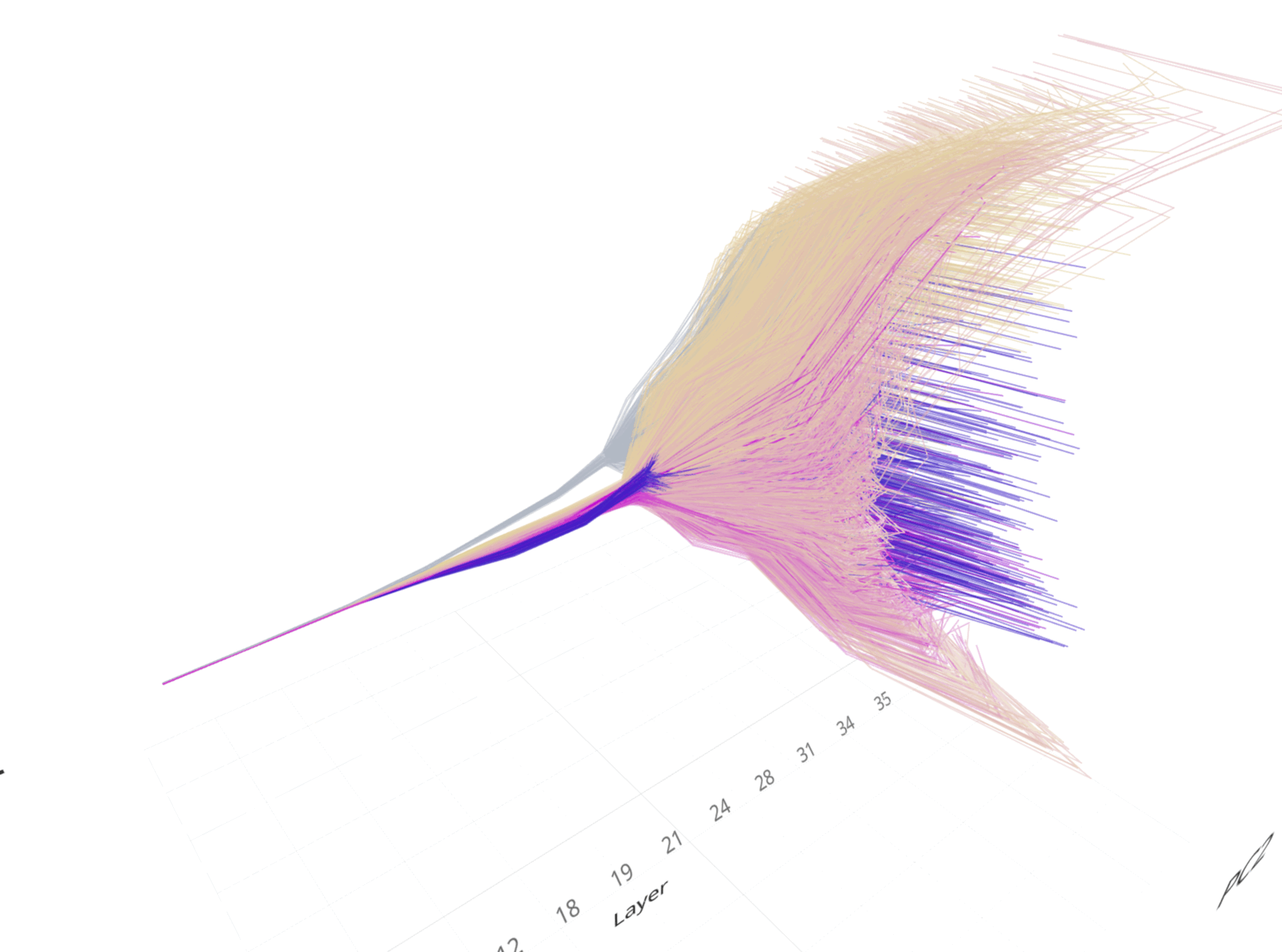}
\end{minipage}\hfill
\begin{minipage}[t]{0.35\textwidth}
  \centering
  {\scriptsize cross-layer, suffix 2, preference}\\[1pt]
  \includegraphics[width=\textwidth]{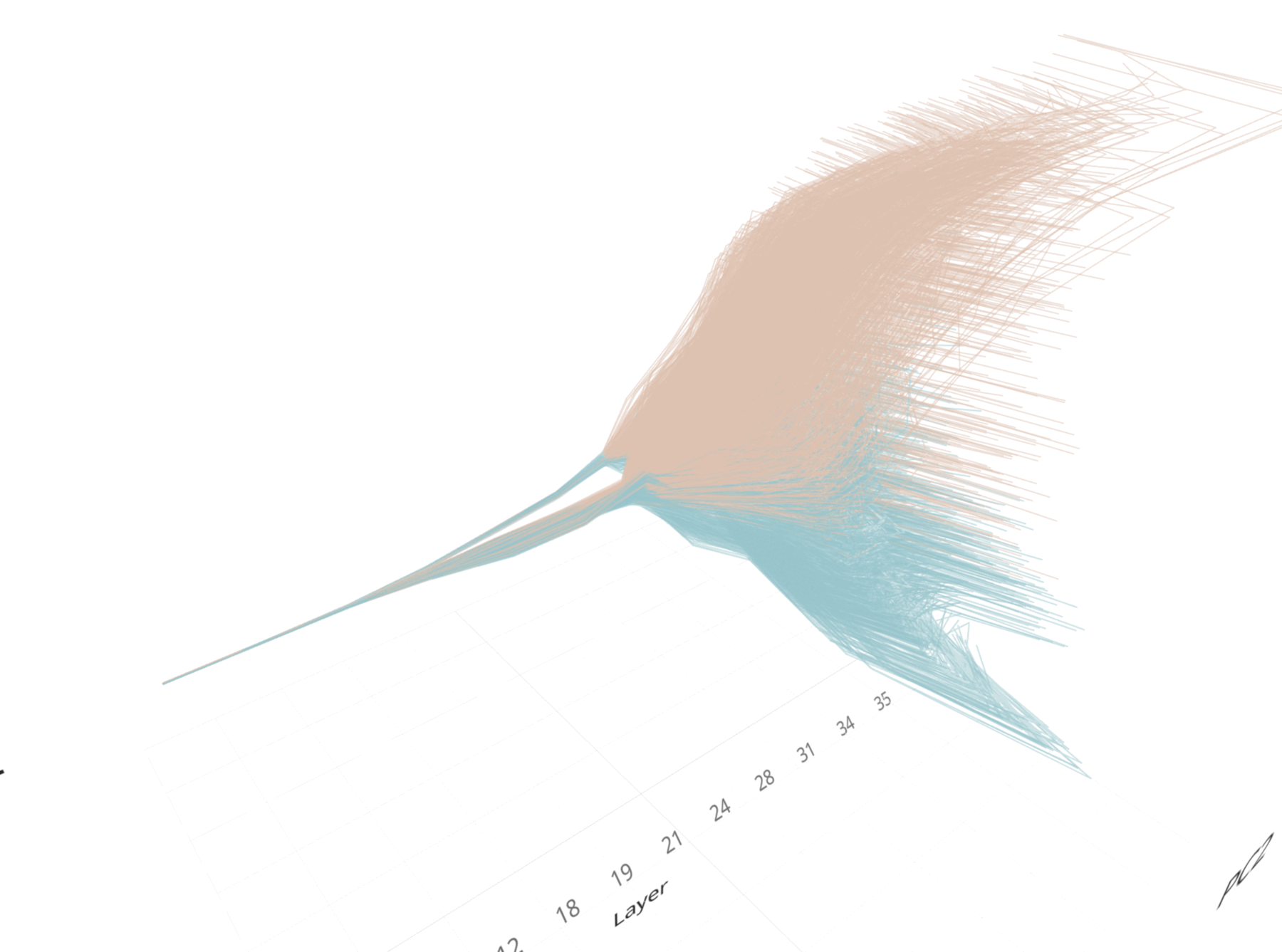}
\end{minipage}

\vspace{2pt}

\begin{minipage}[t]{0.35\textwidth}
  \centering
  {\scriptsize cross-layer, suffix 3, horizon}\\[1pt]
  \includegraphics[width=\textwidth]{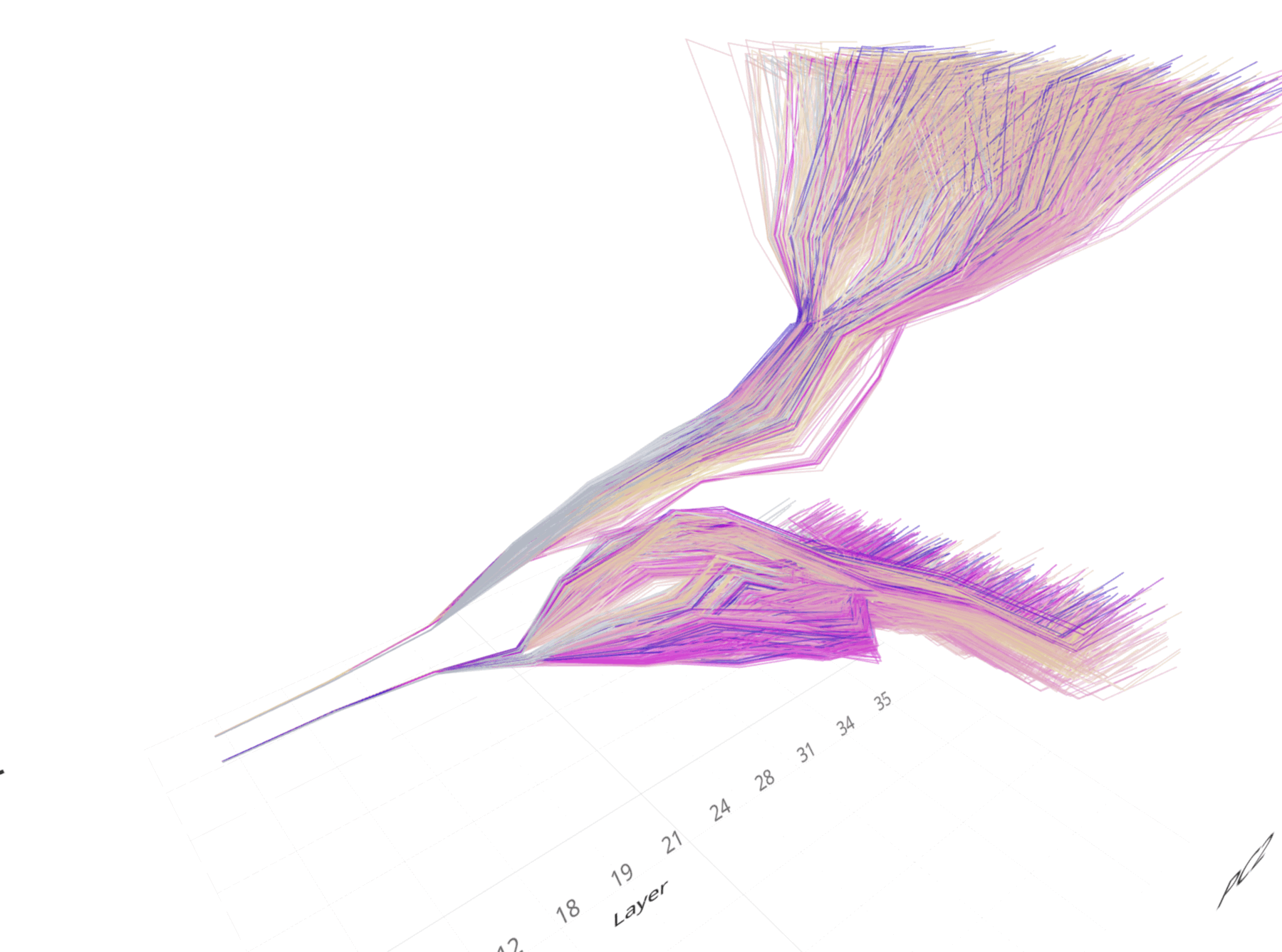}
\end{minipage}\hfill
\begin{minipage}[t]{0.35\textwidth}
  \centering
  {\scriptsize cross-layer, suffix 3, preference}\\[1pt]
  \includegraphics[width=\textwidth]{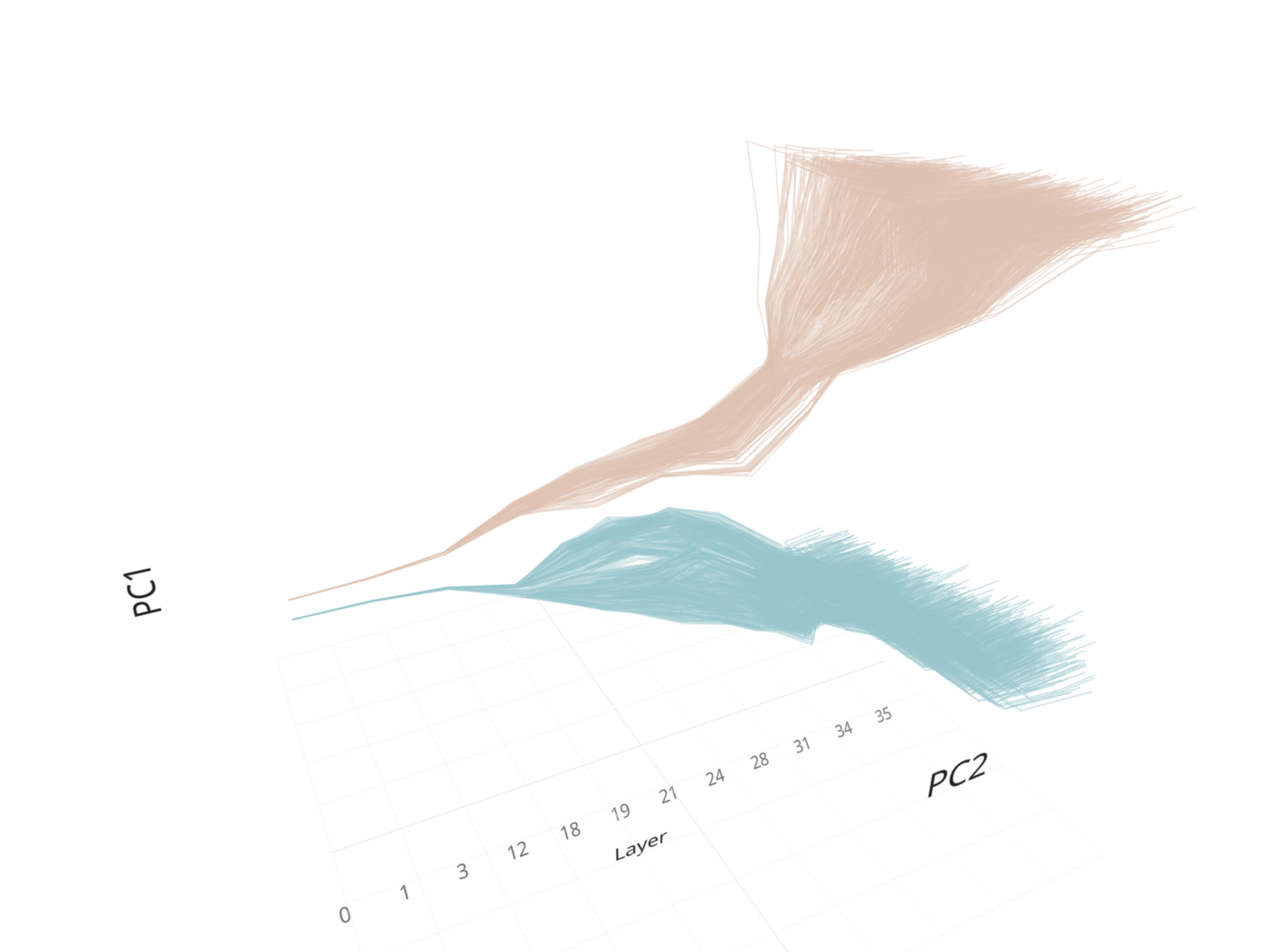}
\end{minipage}
\caption{3D PCA trajectories (PC1 $\times$ PC2 $\times$ Layer) at suffix positions 0--3 (rows), colored by time horizon (left) and preference (right).
At suffix 0, traces fan out by horizon but preferences are intermingled.
At suffix 1--2, the geometry begins reorganizing: traces split into two branches visible in 3D.
By suffix 3, the two branches cleanly correspond to long vs.\ short preference, with horizon ordering preserved as a secondary structure within each branch.}
\label{fig:horizon-to-preference}
\end{figure}

Figure~\ref{fig:horizon-to-preference} shows the same transition in 3D PCA space (PC1 $\times$ PC2 $\times$ Layer), making the geometric reorganization visually explicit.

\FloatBarrier
\subsection{Position sweep at L24}\label{app:parametric-geo-L24}

\begin{figure}[htbp]
\centering
\begin{minipage}[t]{0.48\textwidth}
  \centering
  {\scriptsize L24 \texttt{resid\_post}, 1D PC1 $\times$ position, horizon}\\[2pt]
  \includegraphics[width=\textwidth]{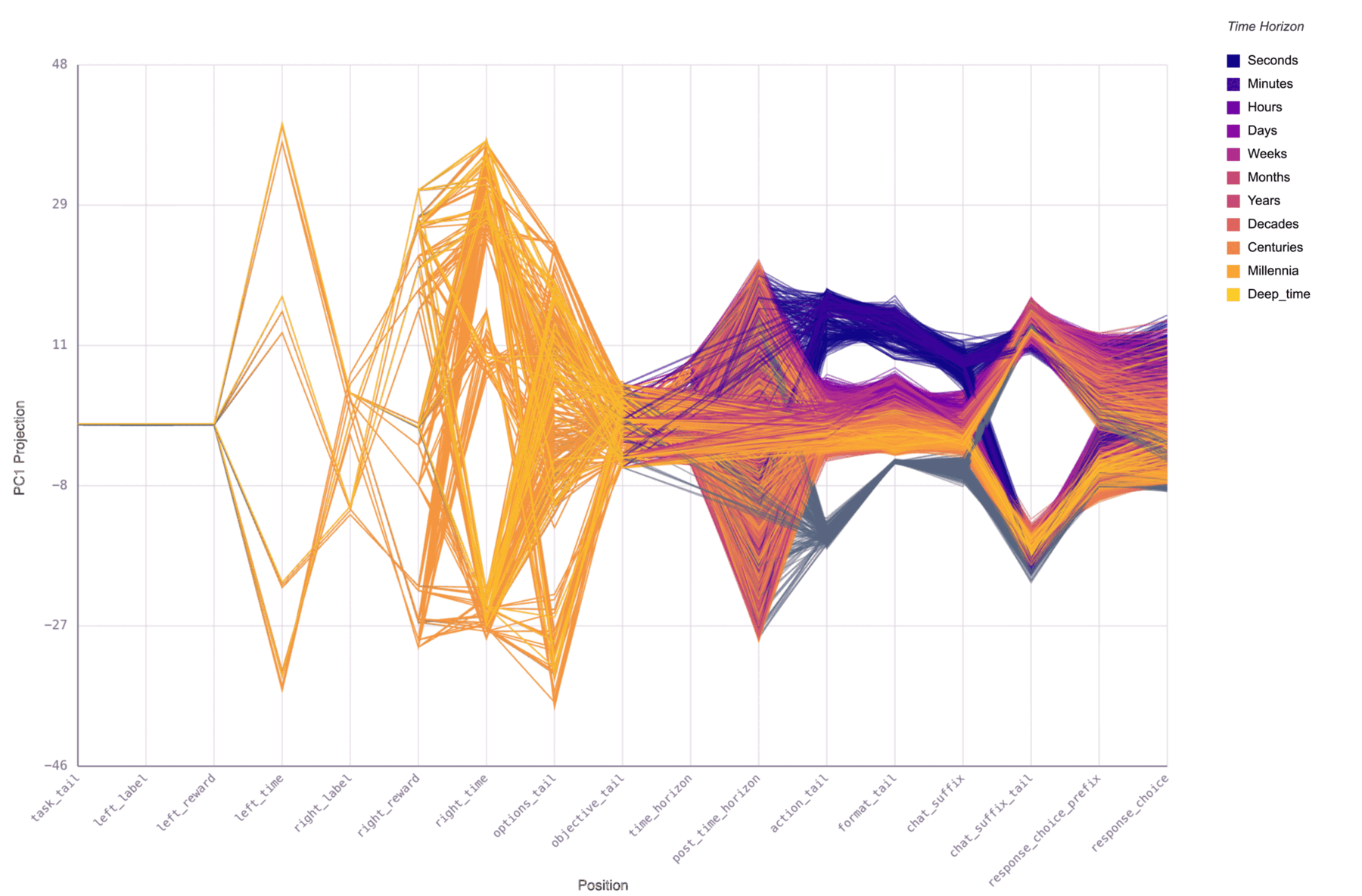}
\end{minipage}\hfill
\begin{minipage}[t]{0.48\textwidth}
  \centering
  {\scriptsize L24 \texttt{resid\_post}, 1D PC1 $\times$ position, preference}\\[2pt]
  \includegraphics[width=\textwidth]{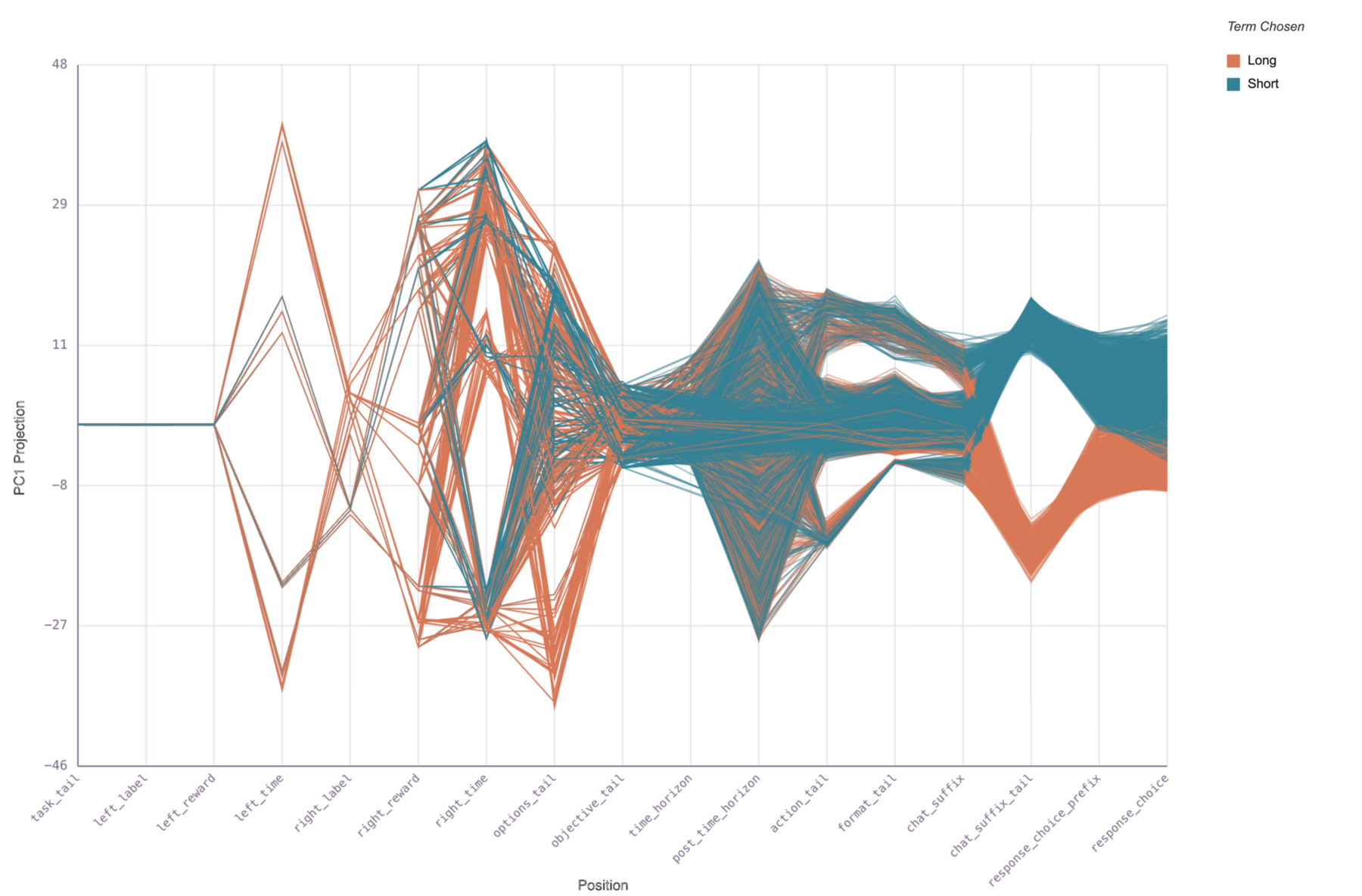}
\end{minipage}

\vspace{4pt}

\begin{minipage}[t]{0.32\textwidth}
  \centering
  {\scriptsize L24, 2D PCA, horizon}\\[2pt]
  \includegraphics[width=\textwidth]{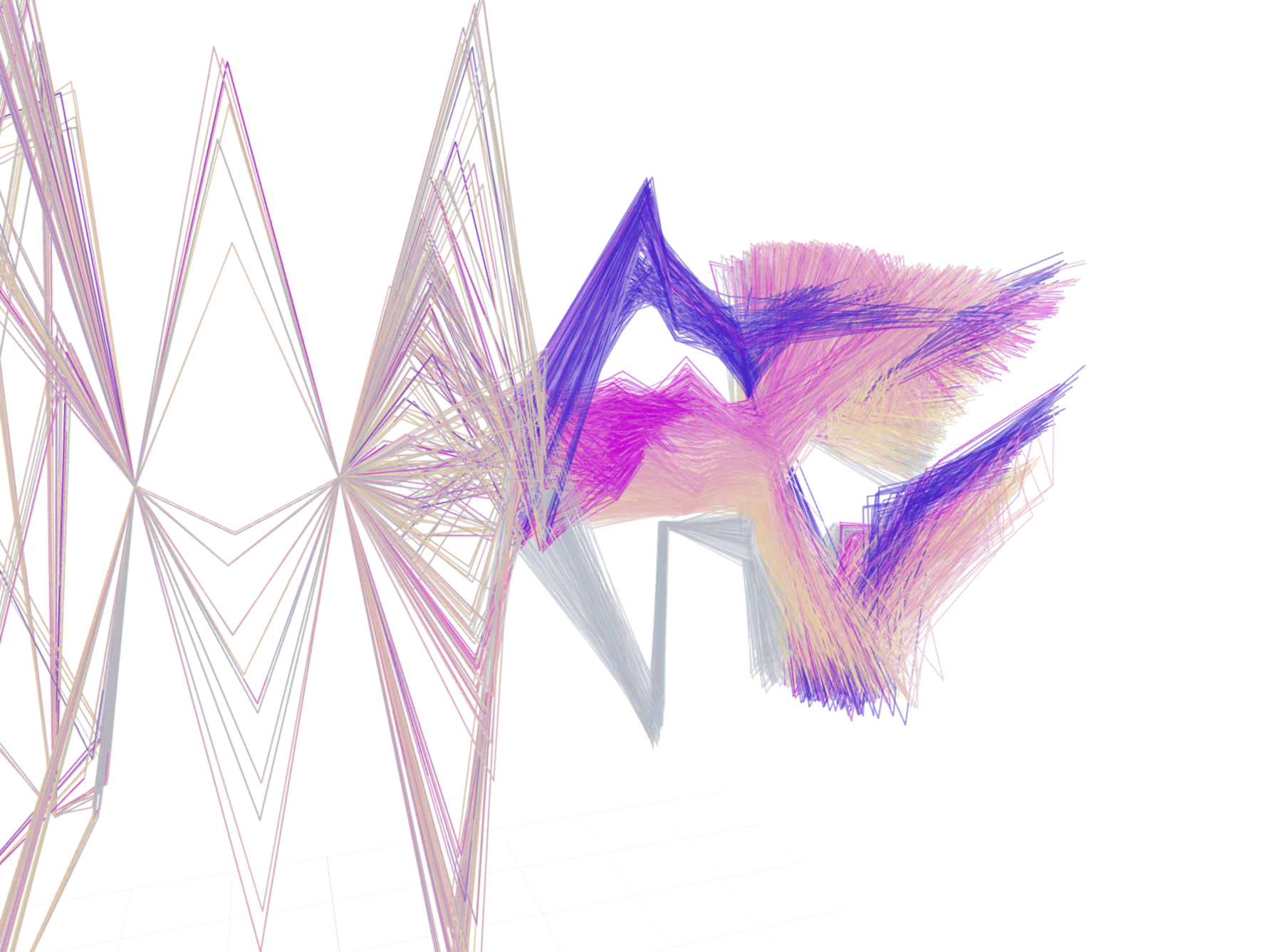}
\end{minipage}\hfill
\begin{minipage}[t]{0.32\textwidth}
  \centering
  {\scriptsize L24, 2D PCA, preference}\\[2pt]
  \includegraphics[width=\textwidth]{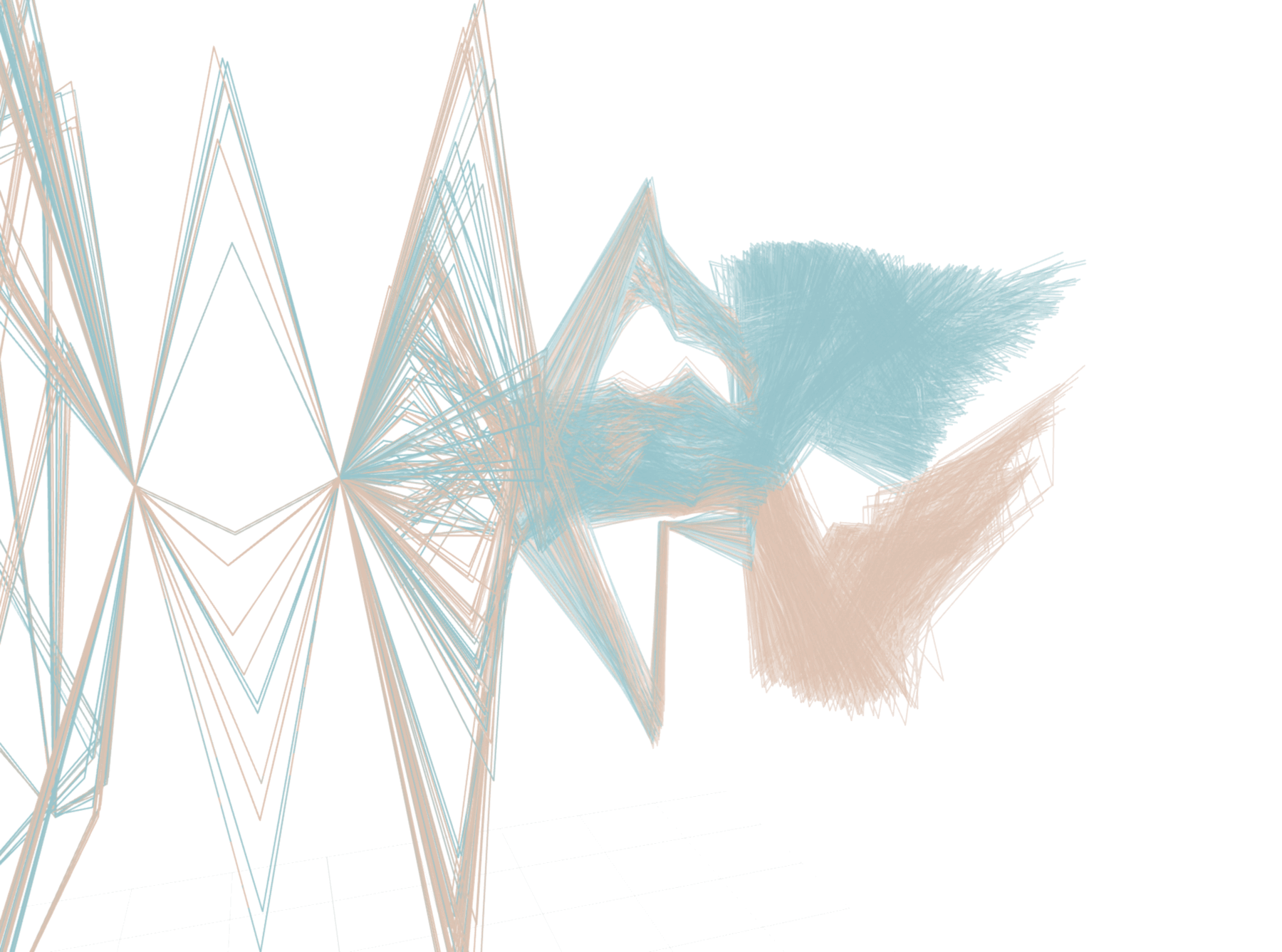}
\end{minipage}\hfill
\begin{minipage}[t]{0.32\textwidth}
  \centering
  {\scriptsize L24, 2D PCA, chosen time}\\[2pt]
  \includegraphics[width=\textwidth]{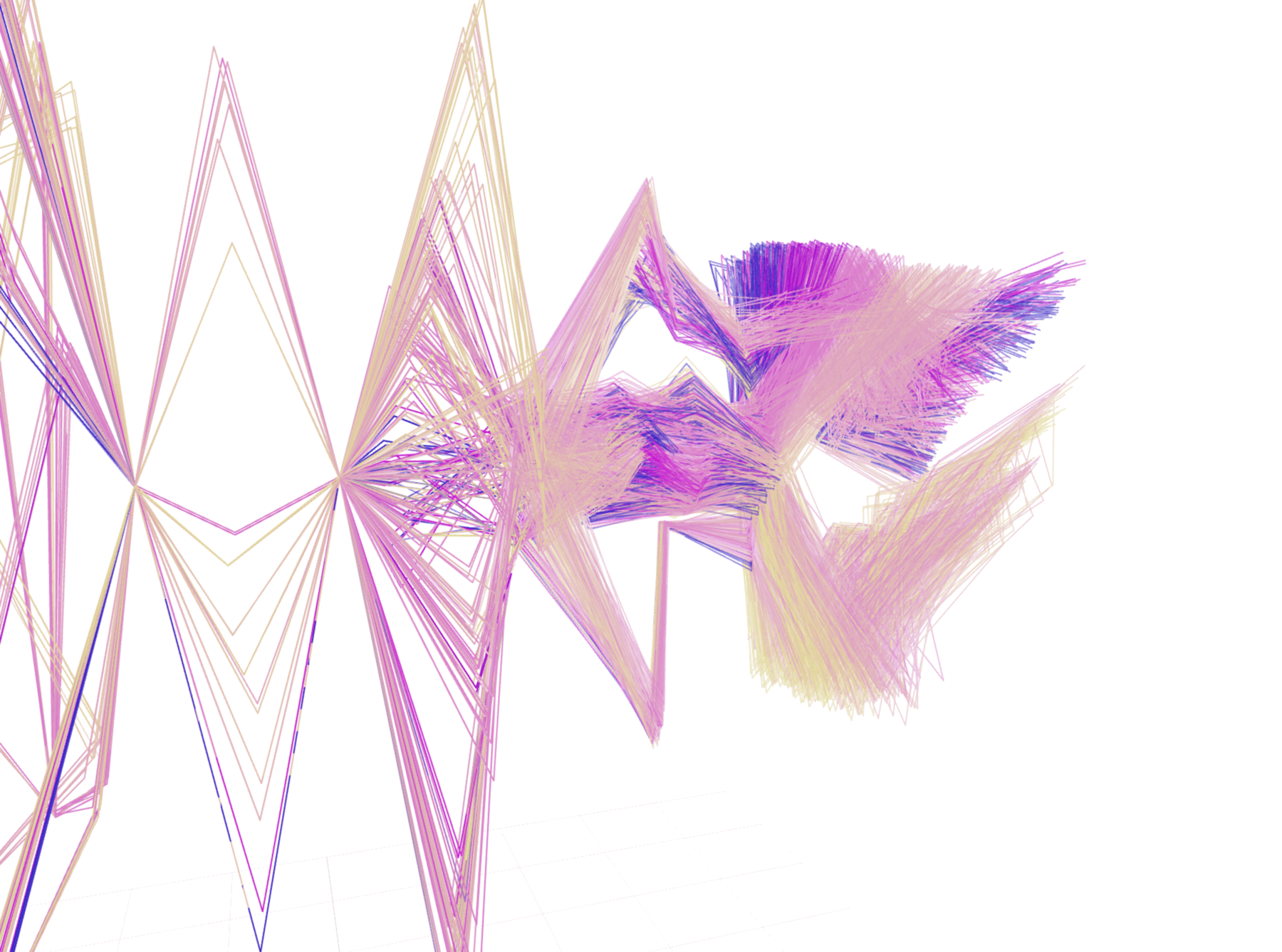}
\end{minipage}
\caption{L24 activations across token positions.
Top: 1D PC1 projection vs.\ position, colored by time horizon (left) and preference (right).
Early prompt positions show wild oscillations; the representation stabilizes at the turn boundary with ordinal horizon separation (left) and clean preference separation emerging a few tokens later (right).
Bottom: 2D PCA (PC1 vs.\ PC2) with position-connected traces, colored by time horizon (left), preference (center), and chosen time (right).
At late positions (dense cluster), both horizon and preference structure are visible.}
\label{fig:L24-position}
\end{figure}

Figure~\ref{fig:L24-position} shows the geometry at a fixed layer (L24) swept across all token positions, confirming that the transition from unstable horizon to committed preference happens at the turn boundary.

The 1D position sweep (top row) confirms the narrative at a single layer: early prompt positions show oscillating, unstable encodings, while the turn boundary and subsequent tokens show stable ordinal horizon separation (left) and progressive preference commitment (right).

\FloatBarrier
\subsection{Direction alignment across components}\label{app:parametric-geo-alignment}

\begin{figure}[htbp]
\centering
\includegraphics[width=0.8\textwidth]{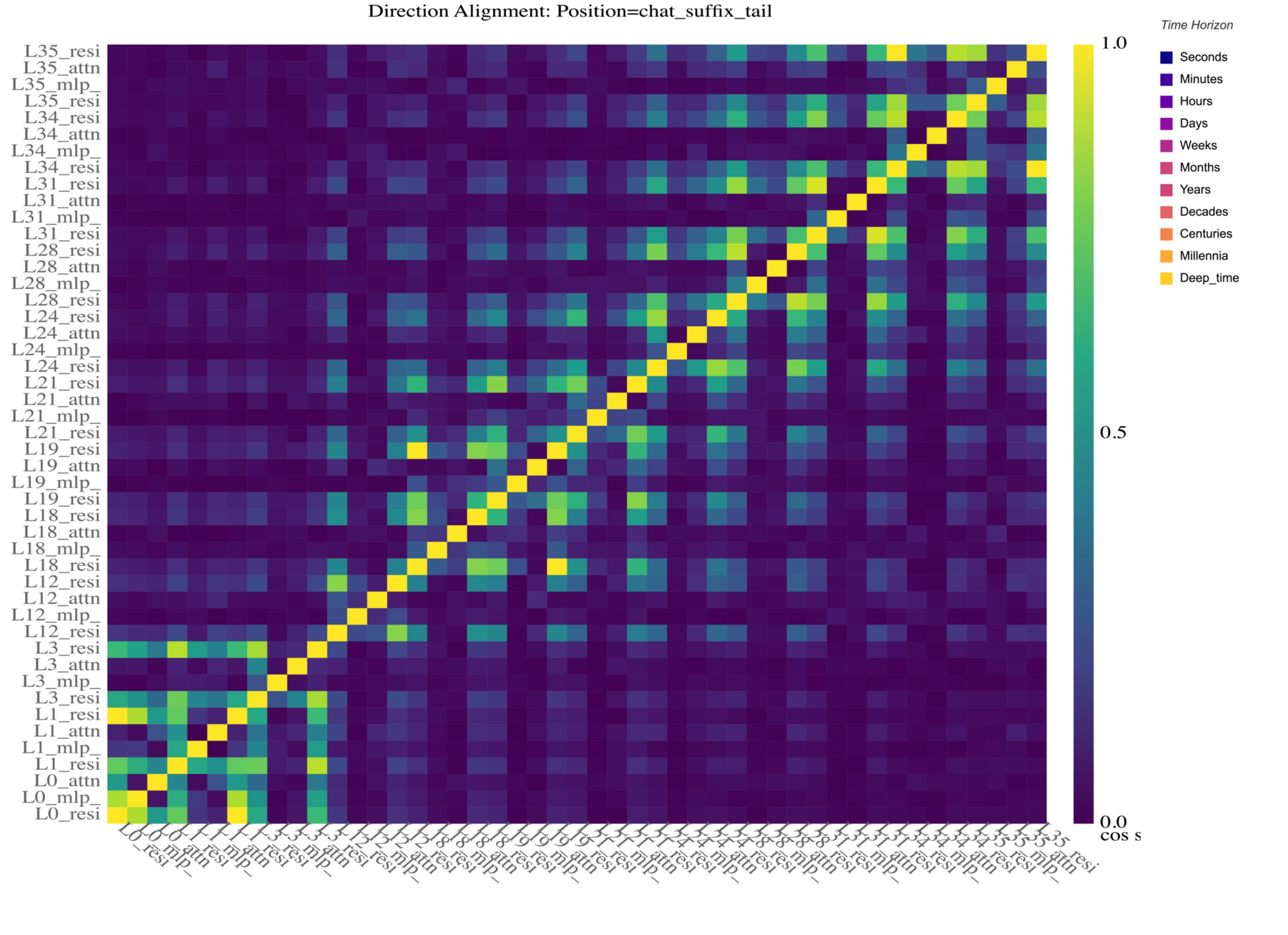}
\caption{Direction alignment matrix across component-layer pairs.
The temporal direction is consistent within nearby layers (warm block-diagonal patches) but rotates substantially between early layers (L0--L12) and later layers (L18+).
Within mid-to-late layers, residual, attention, and MLP components at the same layer share similar directions, indicating a stable temporal subspace.}
\label{fig:dap}
\end{figure}

Figure~\ref{fig:dap} shows the cosine similarity between the top PCA direction at each component-layer pair (computed at the turn boundary), confirming that, at this token position, the temporal direction stabilizes in mid-to-late layers.

\FloatBarrier
\subsection{Summary}\label{app:parametric-geo-summary}

The geometry analysis reveals a five-stage process:
\begin{enumerate}[leftmargin=1.5em]
  \item The model builds an ordinal time-horizon representation in the residual stream, already present within the prompt, but the geometric direction encoding it is \emph{unstable}: it flips polarity across prompt positions (Figure~\ref{fig:early-horizon}).
  \item At the user-to-assistant turn boundary (suffix 0), the residual stream stabilizes this representation into a clean, monotonic fan by time scale, but the model's preference (long vs.\ short) is not yet encoded (Figure~\ref{fig:change-of-turn}).
  \item Attention outputs at suffix positions 1--2 show non-monotonic, actively reorganizing trajectories that progressively write a preference signal into the residual stream (Figure~\ref{fig:attention-transform}).
  \item By suffix 3 (\texttt{assistant} token), the residual stream carries a fully committed preference signal, with long and short cleanly separated from early layers onward (Figure~\ref{fig:change-of-turn-full}).
  \item The transformation occurs in layers 18--24, the same layers identified as causally important by activation patching (\ref{app:causal-parametric}).
\end{enumerate}

This geometric narrative connects localization (where) to function (what): the subgraph in layers 17--35 actively transforms a dimensional concept (time horizon) into a categorical decision (short vs.\ long).
The steering experiments (\ref{app:contrastive-steering}) intervene on this transformation.

The component journey plots (Figure~\ref{fig:component-journey}) offer a geometric correlate of the latent vs.\ constrained distinction identified in \ref{app:latent-vs-constrained}: the attention output shows separation beginning at L21--24 (the shared substrate for both latent and constrained preference), while MLP separation emerges later and with smaller magnitude (the constrained-only contribution).
The attention-mediated horizon-to-preference transformation at the turn boundary (Figures~\ref{fig:attention-transform},~\ref{fig:horizon-to-preference}) is plausibly the geometric signature of the latent mechanism that operates even without constraint tokens.

\clearpage
\clearappnumbering

\section{Latent vs.\ constrained preference}\label{app:latent-vs-constrained}

The convergence analysis (\ref{app:convergence}) established that four methods agree on a subgraph in layers 17--35.
But all of the patching experiments so far contrasted prompts where one has a time-horizon constraint and the other does not.
That design conflates two things: the temporal preference itself and the presence of the constraint tokens.
Here we disentangle them by patching separately on two conditions:
\begin{itemize}[nosep, leftmargin=*]
  \item \textbf{Constrained} ($n = 57$): both prompts have explicit time horizons (different horizons, same structure).
  The contrast is between two constrained preferences.
  \item \textbf{Unconstrained} ($n = 10$): neither prompt has a horizon.
  The contrast is between two latent preferences (the model's default when no temporal pressure is applied).
\end{itemize}

\noindent The question: does the same subgraph mediate both constrained and latent temporal preference, or does the latent preference live somewhere different?

\subsection{MLP effects diverge sharply}

\begin{figure}[!htbp]
\centering
\begin{minipage}[t]{0.48\textwidth}
  \centering
  {\scriptsize Constrained ($n = 57$), MLP denoising, vocab}\\[2pt]
  \includegraphics[width=\textwidth]{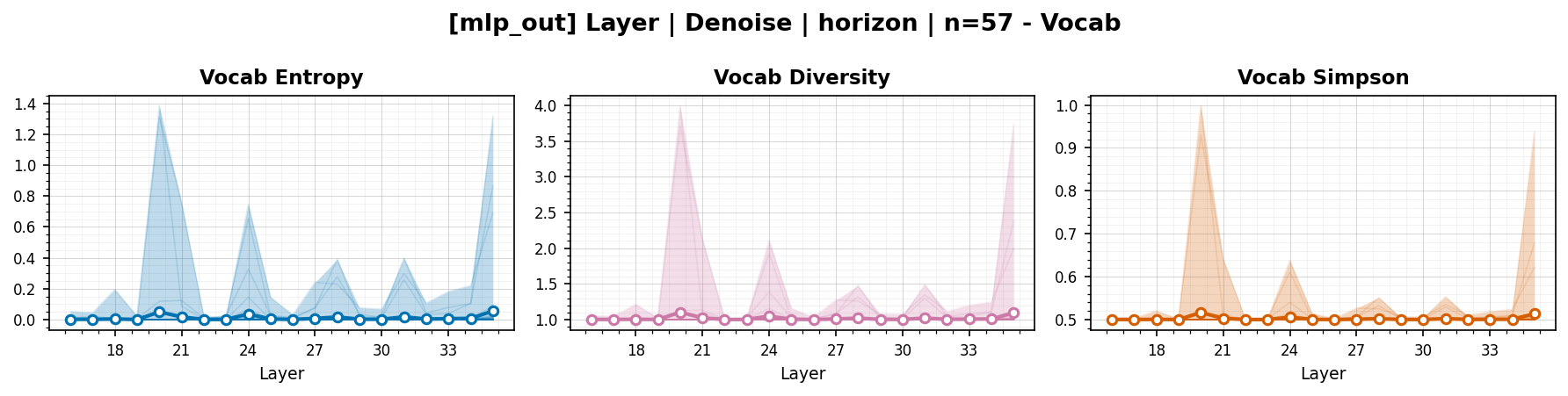}
\end{minipage}\hfill
\begin{minipage}[t]{0.48\textwidth}
  \centering
  {\scriptsize Unconstrained ($n = 10$), MLP denoising, vocab}\\[2pt]
  \includegraphics[width=\textwidth]{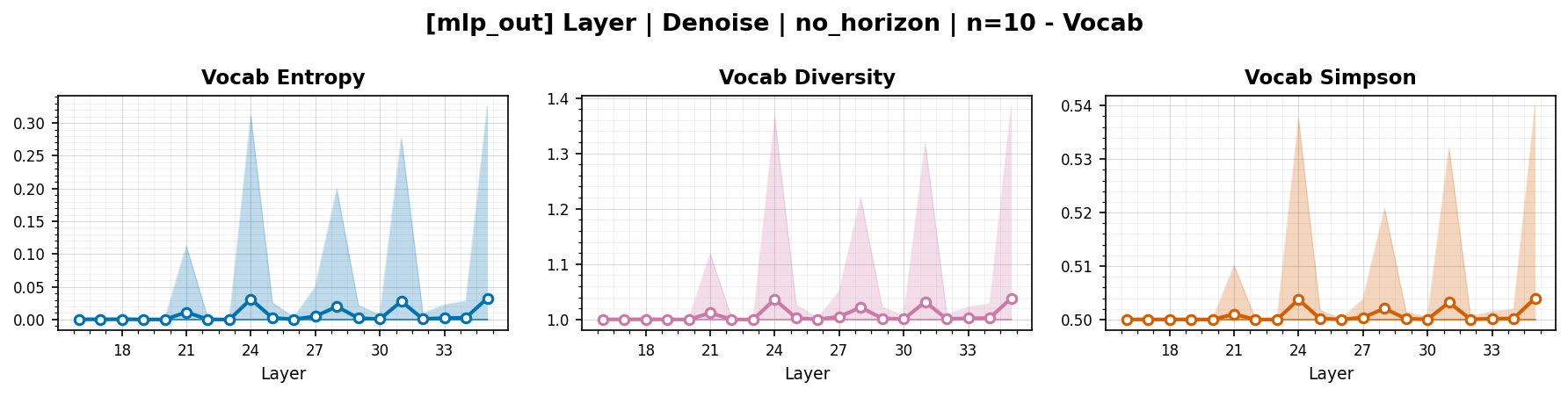}
\end{minipage}

\vspace{4pt}

\begin{minipage}[t]{0.48\textwidth}
  \centering
  {\scriptsize Constrained ($n = 57$), MLP noising, trajectory}\\[2pt]
  \includegraphics[width=\textwidth]{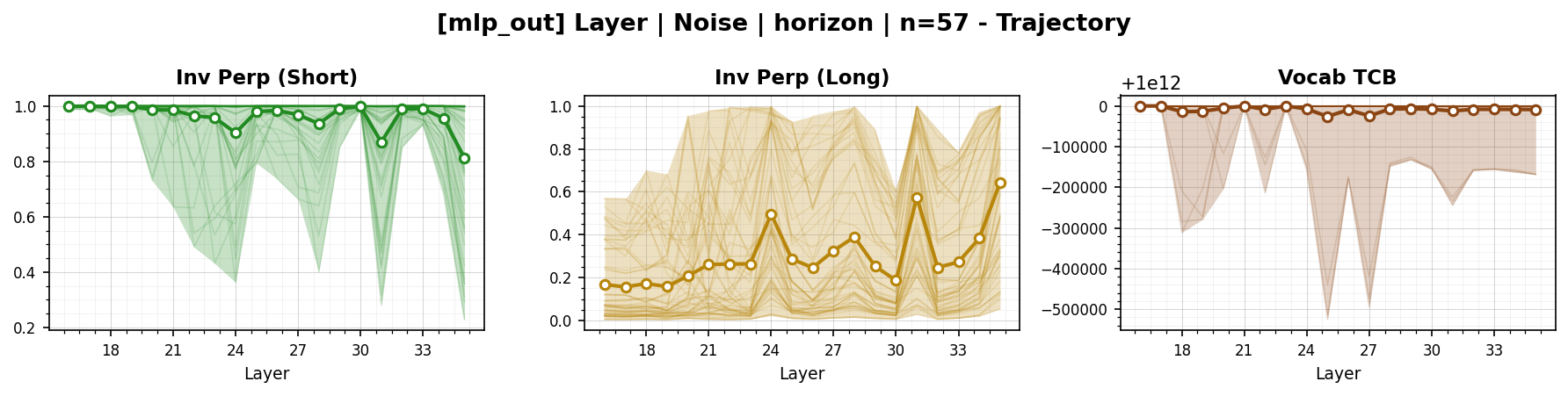}
\end{minipage}\hfill
\begin{minipage}[t]{0.48\textwidth}
  \centering
  {\scriptsize Unconstrained ($n = 10$), MLP noising, trajectory}\\[2pt]
  \includegraphics[width=\textwidth]{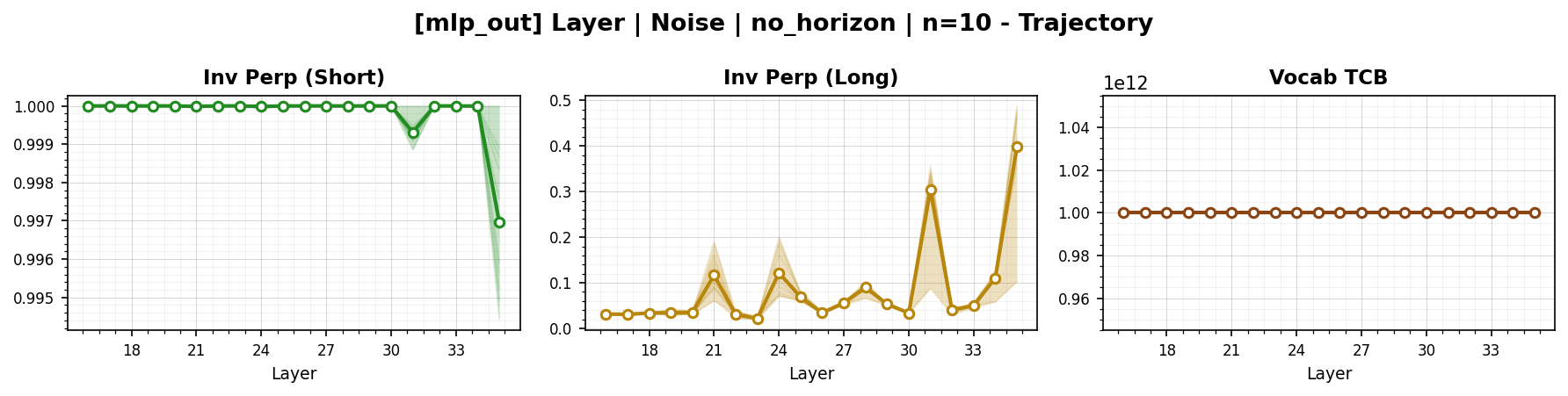}
\end{minipage}
\caption{MLP patching effects for constrained (left) vs.\ unconstrained (right) pairs.
Top row: vocabulary entropy under denoising.
The constrained condition peaks at $\sim$1.4 nats (diversity $\approx 4$); the unconstrained condition reaches only $\sim$0.30 nats.
Bottom row: trajectory under noising.
The constrained condition collapses $\mathrm{inv\_ppl}(\text{short})$ to $\sim$0; the unconstrained condition barely shifts it.}
\label{fig:latent-mlp}
\end{figure}

When both prompts carry explicit horizons, MLP patching produces strong effects: denoising drives vocabulary entropy to $\sim$1.4 nats (diversity $\approx 4$) at L20, and noising collapses $\mathrm{inv\_ppl}(\text{short})$ to near zero (Figure~\ref{fig:latent-mlp}).
When neither prompt has a horizon, the same MLP patching produces much weaker effects: entropy peaks at only $\sim$0.30 nats (diversity $\approx 1.4$), and $\mathrm{inv\_ppl}$ barely moves.

\FloatBarrier

\subsection{Attention effects show the opposite pattern}

\begin{figure}[!htbp]
\centering
\begin{minipage}[t]{0.48\textwidth}
  \centering
  {\scriptsize Constrained ($n = 57$), attn noising, vocab}\\[2pt]
  \includegraphics[width=\textwidth]{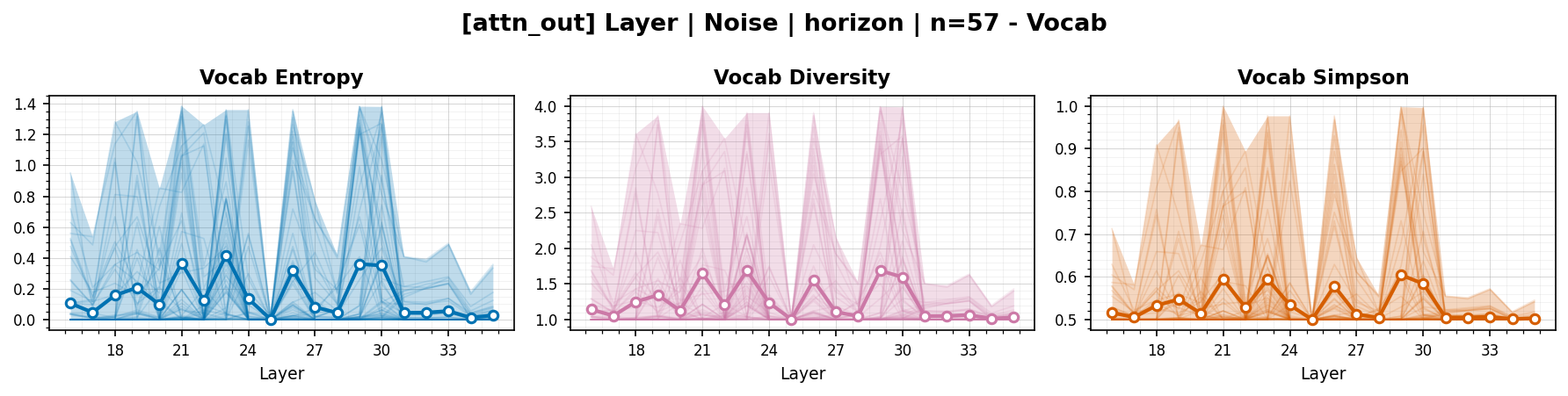}
\end{minipage}\hfill
\begin{minipage}[t]{0.48\textwidth}
  \centering
  {\scriptsize Unconstrained ($n = 10$), attn noising, vocab}\\[2pt]
  \includegraphics[width=\textwidth]{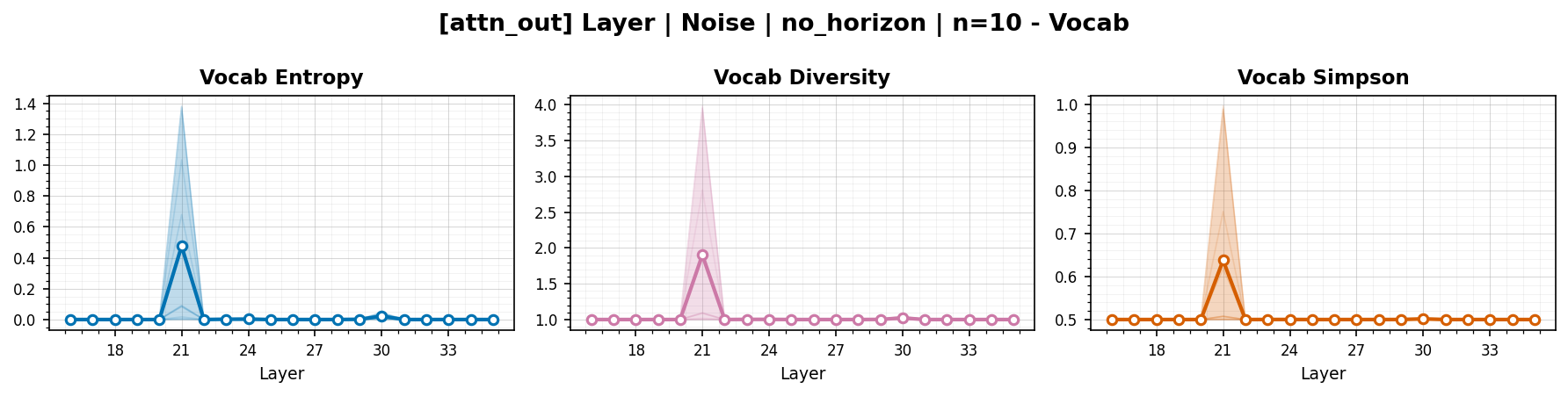}
\end{minipage}
\caption{Attention noising vocabulary effects.
The unconstrained condition (right) shows a sharp, isolated spike at L21--22.
The constrained condition (left) produces a broader, more diffuse effect.
Without a specific constraint token to anchor to, the latent preference depends on a narrower set of attention heads.}
\label{fig:latent-attn}
\end{figure}

Under noising, the unconstrained condition produces a sharper, more localized attention effect: a single spike at L21--22 in the vocabulary metrics (Figure~\ref{fig:latent-attn}).
The constrained condition produces a broader, more diffuse effect across the same layers.

\FloatBarrier

\subsection{Interpretation}

The two conditions use the same subgraph but engage it differently:
\begin{itemize}[nosep, leftmargin=*]
  \item \textbf{Constrained preference} recruits the full subgraph.
  The explicit constraint tokens (``8 months,'' ``10 years'') provide a specific positional anchor that MLP layers can read and transform, producing strong, distributed effects across layers and components.
  This is consistent with the case study (\ref{app:case-study-hf}), where positions 83--106 (the \texttt{CONSTRAINT} section) carry the temporal information.
  \item \textbf{Latent preference} relies primarily on attention.
  Without constraint tokens, the temporal signal must be inferred from the semantic content of the options themselves.
  This inference is mediated by a sparser set of attention heads at L21--22, with minimal MLP involvement.
  The weaker overall effect is consistent with the behavioral finding (\ref{app:behavioral-coherence}) that unconstrained preferences default to a position-sensitive heuristic rather than genuine temporal reasoning.
\end{itemize}

\noindent The same subgraph (L17--35) is involved in both conditions, but the explicit constraint deepens the computation: it engages MLP layers that the latent preference does not reach.
This suggests that the MLP contribution to temporal preference (\ref{app:causal-parametric}) is specifically about processing the constraint, not about encoding the preference itself.

\subsection{Connection to the case study}

The case study (\ref{app:case-study-hf}) patches a \emph{mixed} pair: the clean prompt has an 8-month horizon, the corrupted prompt has none.
Denoising injects constraint information into the unconstrained run; noising removes it from the constrained run.
The denoising--noising asymmetry observed there now has a precise explanation.

Denoising moves the model from the unconstrained regime toward the constrained regime.
The entropy spike during denoising ($\sim$1.4 nats at L22--23) matches the constrained condition's entropy in this appendix ($\sim$1.4 nats), and the full subgraph is engaged (MLP + attention).
Noising moves the model in the opposite direction, from constrained toward unconstrained.
The noising entropy spike is lower ($\sim$0.7 nats) and broader, consistent with the unconstrained condition's weaker, attention-dominated effects.

The numbers are not coincidental.
The mixed-pair case study is literally moving the model between the two regimes characterized here: each direction of patching recapitulates the effect profile of the regime it targets.
This convergence across three independent analyses, the case study's single-pair sweeps, this appendix's condition-separated aggregates, and the convergence table's summary (\ref{app:convergence}), provides strong evidence that the L17--35 subgraph is the genuine locus of temporal preference, and that the presence or absence of a constraint token determines which components within that subgraph are recruited.
\clearpage
\clearappnumbering

\providecommand{\todo}[1]{}

{%
\lstset{aboveskip=4pt,belowskip=4pt}

\section{Behavioral temporal discounting results}\label{app:behavioral-temporal-discount}

The geometry analysis (\ref{app:parametric-geometry}) revealed how temporal preference is represented internally.
Here we ask how it manifests behaviorally: do LLMs discount the future like humans?
We administer the Kirby MCQ-27 questionnaire under controlled personas and apply a novel decision-boundary method that probes beyond the standard instrument (methodology in \ref{app:behavioral-temporal-discount-methods}).

\subsection{Standard MCQ-27 Responses}

Table~\ref{tab:mcq27} shows the estimated $k$ values from the standard questionnaire administration.

\begin{table}[!htbp]
\centering
\caption{Estimated discount rate $k$ from the standard MCQ-27 (direct response mode).
Human benchmarks from~\citet{kirby1999}.}
\label{tab:mcq27}
\begin{tabular}{lcc}
\toprule
\textbf{Group} & \textbf{$k$} & \textbf{Consistency} \\
\midrule
\texttt{Qwen3-4B} (default) & 0.0025 & 89\% \\
\texttt{Qwen3-4B} (heroin) & 0.0041 & 85\% \\
\texttt{Qwen3-8B} (default) & 0.0016 & 93\% \\
\texttt{Qwen3-8B} (heroin) & 0.0025 & 89\% \\
\midrule
\texttt{Gemini} (API) & 0.0016 & 93\% \\
\texttt{Claude} (API) & 0.0016 & 93\% \\
\midrule
Human controls & 0.013 & 96\% \\
Heroin patients & 0.025 & 94\% \\
\bottomrule
\end{tabular}
\end{table}

All LLMs show substantially lower discount rates than humans, suggesting extreme patience in the standard questionnaire format.
The heroin persona produces an increase in $k$ of roughly 1.6$\times$ for both Qwen models (0.0041/0.0025 for the 4B; 0.0025/0.0016 for the 8B), which approximates the $\sim$2$\times$ ratio observed between heroin-dependent and control groups in the human data \citep{kirby1999}.
However, the absolute $k$ values are an order of magnitude lower than those of human participants.

\FloatBarrier
\subsection{Decision Boundary Results: Direct Response}

The decision boundary method reveals a strikingly different picture.
Table~\ref{tab:boundary} summarizes the results across all 8 conditions.

\begin{table}[!htbp]
\centering
\caption{Decision boundary results across all conditions. ``Boundaries'' indicates how many of 27 trials yielded a flip point.}
\label{tab:boundary}
\begin{tabular}{llcccc}
\toprule
\textbf{Model} & \textbf{Condition} & \textbf{Boundaries} & \textbf{Mean $k$} & \textbf{Median $k$} & \textbf{Max $k$} \\
\midrule
4B & Default & 26/27 & 0.076 & 0.018 & 0.657 \\
4B & Heroin & 24/27 & 0.088 & 0.033 & 1.249 \\
4B & Default CoT & 10/27 & 0.043 & 0.037 & 0.110 \\
4B & Heroin CoT & 3/27 & 0.226 & 0.117 & 0.511 \\
\midrule
8B & Default & 8/27 & 0.084 & 0.057 & 0.222 \\
8B & Heroin & 9/27 & 0.039 & 0.002 & 0.252 \\
8B & Default CoT & 12/27 & 0.086 & 0.046 & 0.269 \\
8B & Heroin CoT & 13/27 & 0.051 & 0.012 & 0.238 \\
\bottomrule
\end{tabular}
\end{table}


Several patterns emerge:

\paragraph{The 4B model is more manipulable.} Without CoT, \texttt{Qwen3-4B} finds boundaries on nearly all trials (24--26/27), meaning its preferences can be shifted by adjusting the reward amount.
The heroin persona increases both the mean $k$ and the proportion of ``now'' choices, consistent with the intended effect.

\paragraph{CoT amplifies present bias in the 4B model.} With chain-of-thought, \texttt{Qwen3-4B}'s boundary count drops dramatically, from 26/27 to 10/27 (default) and from 24/27 to just 3/27 (heroin).
The model generates formulaic reasoning about ``immediate access,'' ``liquidity,'' and ``opportunity cost'' that anchors it on choosing ``now'' regardless of reward magnitude.
Even at 20$\times$ the immediate reward, the CoT reasoning justifies present bias.

\paragraph{The 8B model shows the opposite CoT pattern.} For \texttt{Qwen3-8B}, CoT \emph{increases} the number of boundaries found, from 8/27 to 12/27 (default) and from 9/27 to 13/27 (heroin).
The larger model's reasoning is more nuanced, weighing tradeoffs rather than reflexively choosing ``now.''

\paragraph{The 8B heroin CoT persona is paradoxically patient.} Perhaps the most surprising result: under heroin CoT, \texttt{Qwen3-8B} chose ``later'' on 22 of 27 original questions.
Its reasoning incorporated recovery-oriented language: ``Delaying gratification might help me stay focused on my recovery.''
Rather than simulating impulsivity, the 8B model simulated a treatment-compliant patient exercising self-control, the opposite of the original clinical findings~\citep{kirby1999}.

\FloatBarrier
\subsection{Qualitative Analysis of CoT Reasoning}

The CoT reasoning reveals systematic failure modes in how LLMs simulate temporal preferences.
We present verbatim examples below to illustrate these patterns.

\paragraph{4B Default CoT: Formulaic present bias.} The model repeats near-identical justifications across trials: ``I would prefer \$X today because it provides immediate access to the money, which can be useful for immediate needs or opportunities.''
This phrasing appears even when the delayed reward is orders of magnitude larger (e.g., \$19 now vs.\ \$380 in 53 days).

\paragraph{4B Heroin CoT: Amplified impulsivity.} The heroin persona intensifies the present-bias reasoning: ``I need the money now to cover immediate expenses related to my treatment and daily living.''
The model generates plausible but extreme justifications that make it nearly impossible to flip to ``later.''

\paragraph{8B Heroin CoT: Recovery narrative override.} The 8B model constructs a coherent recovery narrative: ``I know I could use the money now for immediate needs, but I'm trying to build better habits. Delaying gratification might help me stay focused on my recovery.''
This suggests the model is drawing on training data about addiction recovery that emphasizes self-control and delayed gratification, precisely contradicting the empirical finding that heroin users are more impulsive.

\subsection{Inconsistent CoT: Reasoning Contradicts the Answer}

Perhaps the most striking failure mode is when the model's reasoning argues for one choice but the final answer token selects the other.
This suggests the final answer is partially decoupled from the preceding chain of thought.

\paragraph{Example: \texttt{Qwen3-4B}, Heroin persona, Q5.} SIR = \$14, LDR = \$280, delay = 19 days.

\begin{lstlisting}
I would prefer $14 today. Receiving the smaller amount
immediately allows me to have the money now, which can
be used for immediate needs. LATER
\end{lstlisting}

The model explicitly states ``I would prefer \$14 today,'' gives a rationale for the immediate reward, and then emits \texttt{LATER} as its final answer, a direct self-contradiction within a single generation.

\paragraph{Example: \texttt{Qwen3-4B}, Heroin persona, Q6.} SIR = \$47, delay = 160 days.
During binary search, the model produces \emph{identical reasoning text} for both NOW and LATER answers.
At LDR = \$924:

\begin{lstlisting}
I would prefer $47 today. Receiving the money now allows
me to use it immediately, which is important for my
financial stability and helps me stay on track with my
treatment. NOW
\end{lstlisting}

One dollar higher, at LDR = \$925, the reasoning is nearly word-for-word identical but the answer flips:

\begin{lstlisting}
I would prefer $47 today. Receiving the money now allows
me to use it immediately, which is important for my
financial stability and to cover immediate needs. LATER
\end{lstlisting}

The CoT provides \emph{zero justification} for choosing ``later.''
The reasoning at both price points argues exclusively for the immediate reward, yet the final token differs.

\paragraph{Example: \texttt{Qwen3-8B}, Heroin persona, Q1.} SIR = \$54, LDR = \$54 (same nominal amount), delay = 117 days.

\begin{lstlisting}
The money is the same, but getting it now means I can use
it for immediate needs, which might be more tempting for
someone with a history of substance use. Delaying the
reward might help with impulse control. LATER
\end{lstlisting}

The model chooses to wait 117 days for the \emph{exact same amount of money} and rationalizes it as ``impulse control.''
This is economically irrational: there is zero benefit to waiting for an identical payoff.
The heroin persona causes the 8B model to overcorrect, treating every choice as an opportunity to practice delayed gratification, even when the delayed option offers no premium whatsoever.

\FloatBarrier
\subsection{No Boundary Found: Rejecting 20$\times$ Returns}

When the binary search reaches the maximum (20$\times$ the SIR) without finding a flip point, it exposes cases where the model's present bias is effectively infinite.
The CoT reasoning at these extremes reveals memorized heuristics applied in absurd contexts.

\paragraph{Example: \texttt{Qwen3-4B}, Default persona, Q4.} SIR = \$31, delay = 7 days.
At LDR = \$85:

\begin{lstlisting}
I would consider the immediate access to $31 today as more
valuable because it provides liquidity and can be used for
immediate needs or opportunities. The $85 in 7 days is
larger, but the delay might reduce its value due to
potential inflation or missed opportunities. However, since
both are guaranteed, the choice depends on personal
financial priorities. NOW
\end{lstlisting}

At LDR = \$620 (20$\times$ the SIR, a 1,900\% guaranteed return in one week):

\begin{lstlisting}
I would prefer $31 today because it provides immediate
access to the money, which can be useful for immediate
needs or opportunities. NOW
\end{lstlisting}

The model cites ``potential inflation'' as a reason to reject \$620 guaranteed in 7 days.
Even at hyperinflationary rates, one-week inflation would not erode a 20$\times$ premium.
This is a clear instance of a memorized financial heuristic (``inflation risk'') applied in a context where it makes no economic sense.

\paragraph{Example: \texttt{Qwen3-4B}, Heroin persona, near-total refusal to flip.} Under the heroin CoT condition, 24 of 27 questions found no boundary.
The reasoning collapsed into a single formulaic sentence with no deliberation.
Representative example at Q27 (SIR = \$20, delay = 7 days):

At LDR = \$55:
\begin{lstlisting}
I would prefer $20 today. NOW
\end{lstlisting}

At LDR = \$400 (20$\times$, a guaranteed 1,900\% return in one week):
\begin{lstlisting}
I would prefer $20 today. NOW
\end{lstlisting}

The model refuses \$400 in 7 days over \$20 today with no justification whatsoever.
The persona prompt overrides any capacity for deliberation, producing a near-deterministic NOW response regardless of the economic parameters.

\paragraph{Example: \texttt{Qwen3-8B}, Heroin persona, Q15.} SIR = \$69, delay = 91 days.
At LDR = \$1,380 (20$\times$):

\begin{lstlisting}
The money today would help me with immediate needs, like
buying drugs or covering basic expenses. The larger amount
later is tempting, but I might not be able to wait. I need
the money now to avoid relapse. NOW
\end{lstlisting}

At 20$\times$ the immediate reward, the 8B heroin persona explicitly states the purpose of the money includes ``buying drugs,'' while simultaneously citing ``avoid relapse,'' a contradiction \emph{within the persona itself}.
The model has internalized the persona to the point of generating drug-seeking justifications alongside recovery language.

\FloatBarrier
\subsection{Discussion}

\subsubsection{LLMs Are Poor Simulators of Human Temporal Preferences}

Our results demonstrate that LLMs fail to faithfully replicate human temporal discounting in several ways:

\begin{enumerate}
    \item \textbf{Extreme and inconsistent discount rates.} The decision boundary method reveals that LLM discount rates are highly variable across trials, often differing from theoretical indifference points by 100--400$\times$. Human responses, by contrast, show consistency rates above 90\%.

    \item \textbf{CoT reasoning as confabulation.} Rather than improving decision quality, CoT reasoning in the 4B model acts as a post-hoc justification engine that locks in present bias. The model generates plausible-sounding economic reasoning (``opportunity cost,'' ``time value of money'') that is misapplied, e.g., citing inflation risk on a 7-day delay.

    \item \textbf{Persona effects are unreliable.} The heroin persona increases impulsivity in the 4B model but decreases it in the 8B model (under CoT). Within this single-family pair, the 8B model appears to ``over-correct'' by drawing on normative recovery narratives rather than simulating the behavioral patterns characteristic of active substance users; we do not claim this generalizes across model families.
\end{enumerate}

\subsubsection{The Decision Boundary Method}

The decision boundary approach proves more revealing than standard questionnaire scoring.
While the MCQ-27 responses suggest all LLMs are extremely patient ($k < 0.005$), the boundary search exposes:

\begin{itemize}
    \item Trials where the model says ``now'' even at 20$\times$ the reward (infinite effective $k$).
    \item Sharp, dollar-level flip points that differ wildly from the theoretical indifference values.
    \item Inconsistent behavior near boundaries, where a \$1 change in LDR reverses the decision, suggesting the model lacks a coherent underlying preference function.
\end{itemize}

This method could be applied to other psychological instruments administered to LLMs, providing a more rigorous test of whether models have stable, internally consistent preference structures.

\subsubsection{Implications}

These findings carry practical implications for LLM deployment:

\begin{itemize}
    \item \textbf{Financial advice}: LLMs may give inconsistent guidance about saving vs.\ spending, depending on how questions are framed.
    \item \textbf{Clinical simulation}: Using LLMs to simulate patient populations for research or training requires extreme caution, as persona effects may not produce the intended behavioral patterns.
    \item \textbf{Reasoning fidelity}: CoT prompting does not guarantee better-calibrated preferences and may actively degrade performance by providing a mechanism for confabulation.
\end{itemize}

\subsubsection{Conclusion}

We administered the Kirby MCQ-27 to \texttt{Qwen3} models under multiple conditions and introduced a decision boundary method to probe LLM temporal preferences at higher resolution.
Our key findings are: (1) LLMs exhibit extreme and inconsistent present bias when probed beyond surface-level questionnaire responses; (2) chain-of-thought reasoning amplifies this bias in smaller models while producing paradoxical patience in larger models under clinical personas; and (3) the decision boundary method reveals that LLMs lack the stable, coherent preference functions that characterize human temporal discounting.
Within the Qwen3 family we tested, these results caution against using the non-thinking variant as a faithful simulator of human decision-making, particularly for clinical populations; we leave broader cross-family validation to future work.

}

\clearpage
\clearappnumbering

\section{Behavioral coherence results}\label{app:behavioral-coherence}

The discounting results (\ref{app:behavioral-temporal-discount}) showed that LLMs are extremely patient but behaviorally unstable.
Here we probe this instability systematically across 30 models, 960 prompts each, varying time horizon, reward magnitude, presentation order, label format, and context framing (methodology in \ref{app:behavioral-coherence-methods}).
Zero unparseable responses were observed across all 28{,}800 samples.

\paragraph{Paired-response restriction.}
Every metric in this appendix (\%LT, order stability, position bias, coherence, label stability, rule-match, reward sensitivity, context sensitivity) is computed on \emph{paired} responses only: prompts enter the denominator only when both the ST-first and LT-first orderings at the same (horizon, reward, context, label-style) produced a parseable choice.
This guarantees a single, shared denominator across every heatmap and table, so cells are directly cross-comparable.
In particular, order stability and position bias satisfy $|\text{bias}| \leq 1 - \text{stability}$ by construction: a model that is 92\% order-stable cannot have a position-bias magnitude larger than 8 percentage points.

We organize the analysis around four orthogonal questions:
\begin{enumerate}[nosep, leftmargin=*]
  \item \textbf{Are choices stable?} Does swapping presentation order, label format, reward magnitude, or scenario framing change the model's answer? Any format sensitivity signals that the choice is driven by surface cues, not preference.
  \item \textbf{Are choices coherent?}
  Coherence is \emph{only defined in the temporal reasoning zone} (horizons of 1--5 years), where only the 6-month short-term option can deliver within the deadline, so picking ST is the rational answer.
  At anchor horizons (6mo, 10y) agreement with the rational rule is pattern-matching; beyond 10y both options deliver, so LT dominates on expected value but this is preference, not coherence.
  \item \textbf{What is the model's latent temporal preference?}
  With no horizon constraint, what does the model default to?
  Decomposed by presentation order to separate genuine preference from position bias.
  \item \textbf{Cross-cutting patterns.}
  Claude-family step functions, \texttt{Qwen3} hybrid-thinking vs.\ mode-specialized 2507 variants, target-model deep dive.
\end{enumerate}

Where a table reports only four models, they are chosen to span the four qualitative regimes we observe across the full 30-model panel:
\begin{itemize}[nosep, leftmargin=*]
  \item \texttt{Qwen3-4B} (hybrid-thinking, run in non-thinking mode) -- graded but instrumentally incoherent
  \item \texttt{Qwen3-4B-Instruct-2507} (our target) -- positionally polarized in the reasoning zone
  \item \texttt{Claude Opus 4.7} -- binary step heuristic (flagship Anthropic model)
  \item \texttt{GPT-5.4} -- horizon-aware, the strongest approximation to rational in our panel
\end{itemize}
The figures themselves always show all 30 models, with the target model highlighted.

\begin{table}[!htbp]
\centering
\small
\begin{tabular}{lcc}
\toprule
\textbf{Model} & \textbf{\% Long-Term} & \textbf{\% Short-Term} \\
\midrule
\texttt{Qwen3-4B}                     & 71.8\% & 28.2\% \\
\texttt{Qwen3-4B-Instruct-2507}       & 58.9\% & 41.1\% \\
\texttt{Claude Opus 4.7}              & 39.0\% & 61.0\% \\
\texttt{GPT-5.4}                      & 36.9\% & 63.1\% \\
\bottomrule
\end{tabular}
\caption{Overall temporal preference across 960 samples per model, for the four-regime representative subset.
The pooled \%LT number is a noisy summary: two models with the same 40\% can differ in whether the 40\% is horizon-aware choices or positional artifacts.
The rest of this appendix unpacks that.}
\label{tab:coherence-overall}
\end{table}

\FloatBarrier

\subsection{Are Choices Stable?}\label{app:coherence-stability}

Before asking whether a model's choice is \emph{right}, we check whether it is even \emph{consistent}.
A model whose choice flips when we swap \texttt{a/b} for \texttt{x/y}, or when we list the short-term option second instead of first, is not expressing a preference, it is responding to surface form.

\begin{figure}[!htbp]
  \centering
  \begin{minipage}{0.48\textwidth}
    \centering
    \includegraphics[width=\textwidth]{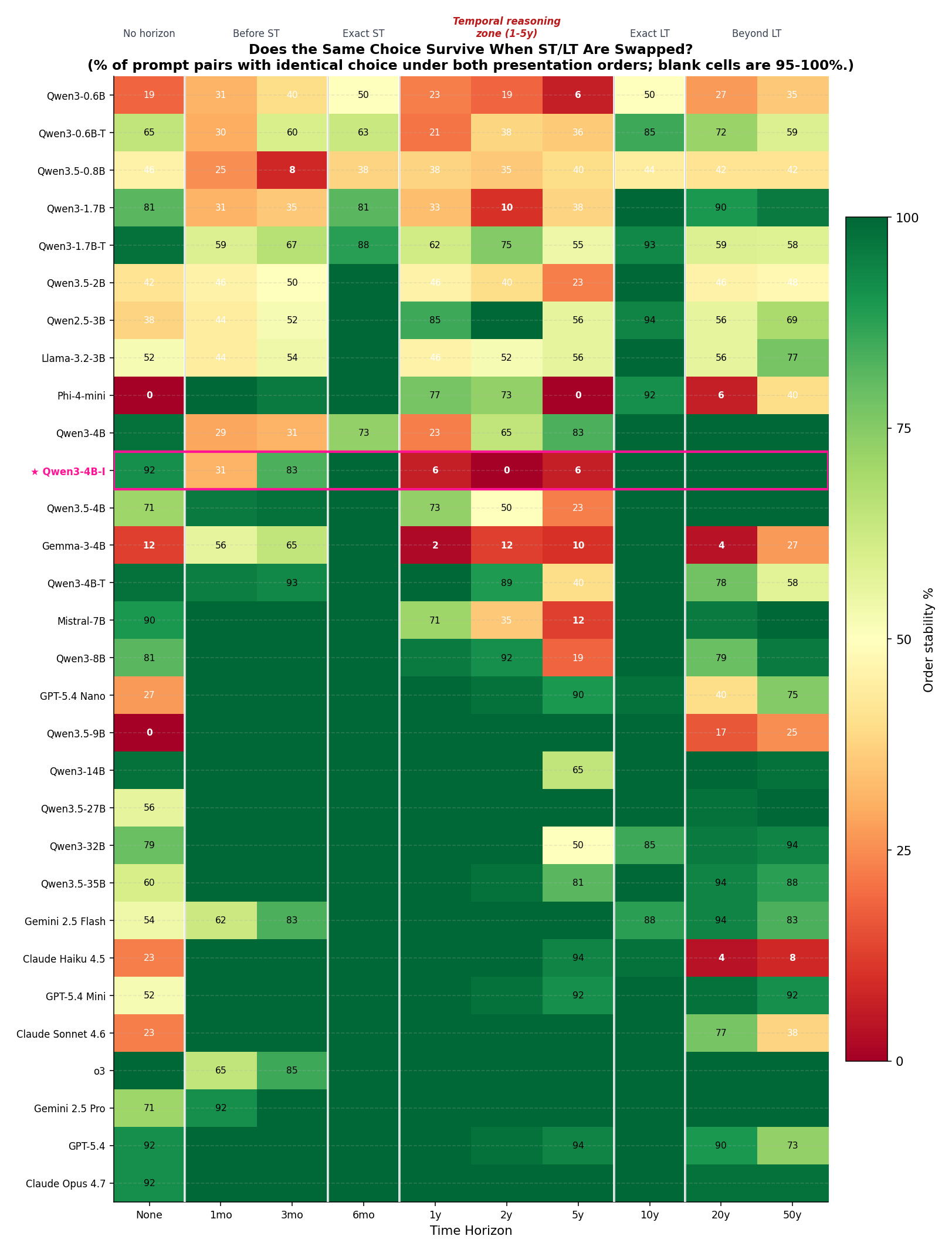}
  \end{minipage}\hfill
  \begin{minipage}{0.48\textwidth}
    \centering
    \includegraphics[width=\textwidth]{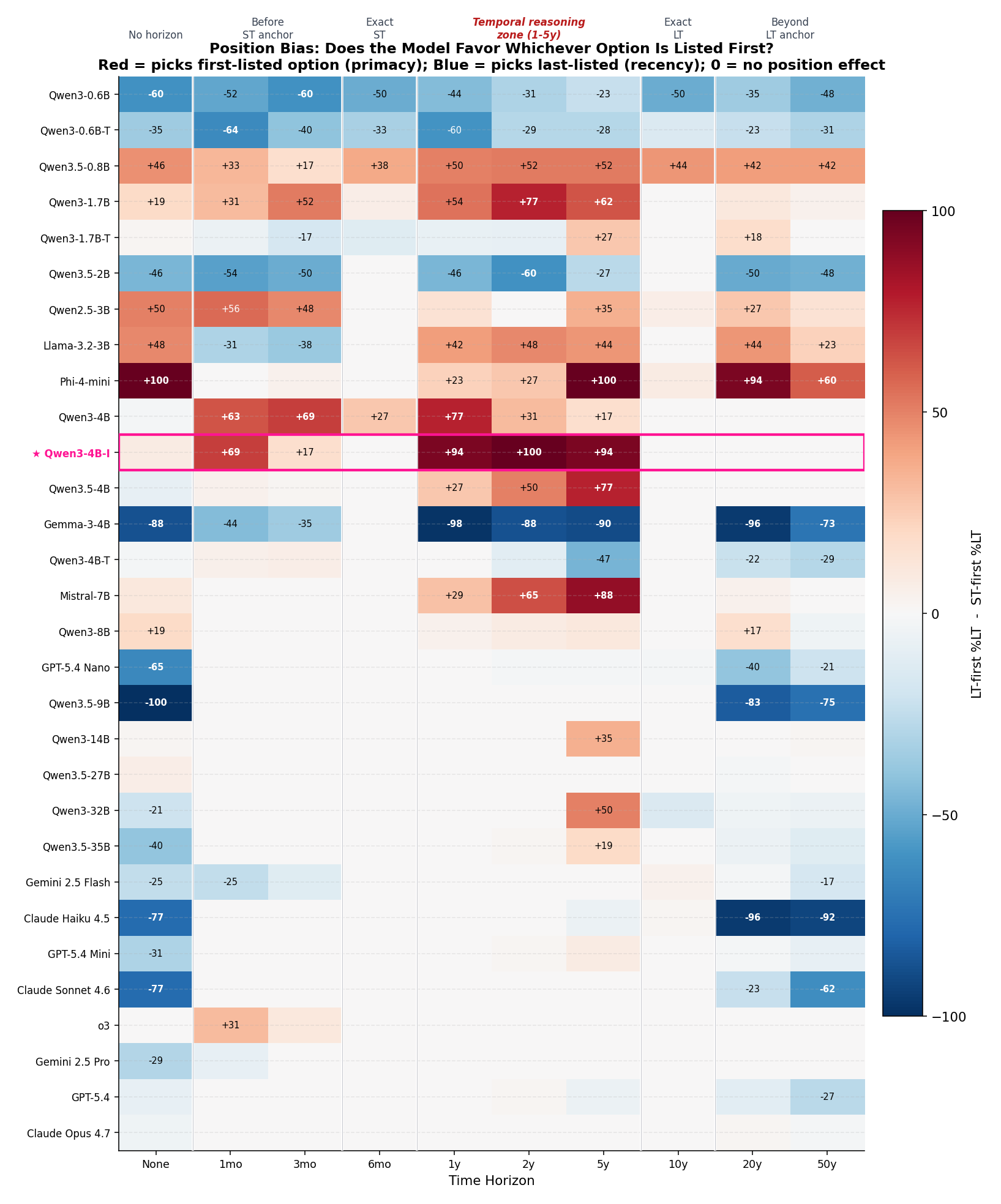}
  \end{minipage}
  \caption{\textbf{Left:} Order stability across 30 models $\times$ 10 horizons.
  Red cells ($<$50\%) indicate the model flips its answer when the two options are swapped; the temporal reasoning zone (1--5y) is where this is catastrophic for several families.
  \textbf{Right:} Signed order-bias decomposition (LT-first \%LT minus ST-first \%LT).
  Red = primacy (picks whatever is listed first); blue = recency.
  Both views agree that order bias peaks inside the reasoning zone for most families, and for the \texttt{Claude} family at 20--50y.}
  \label{fig:order-stability}
  \label{fig:order-bias}
\end{figure}

\paragraph{Order stability.}
For each (horizon, reward, context, label-style) combination, we run the prompt with the short-term option listed first and again with it listed second, then check whether the choice is identical.
Figure~\ref{fig:order-stability} is a heatmap of this across all 30 models and 10 horizons, paired with a per-cell order-bias heatmap (signed \%LT gap when order is flipped).

\begin{table}[!htbp]
\centering
\small
\begin{tabular}{lcccc}
\toprule
\textbf{Horizon} & \texttt{Qwen3-4B} & \texttt{Qwen3-4B-Inst} & \texttt{Claude Opus 4.7} & \texttt{GPT-5.4} \\
\midrule
No horizon & 98\% & 92\% & 92\% & 92\% \\
1 mo       & 29\% & 31\% & 100\% & 100\% \\
3 mo       & 31\% & 83\% & 100\% & 100\% \\
6 mo       & 73\% & 100\% & 100\% & 100\% \\
1 y        & 23\% & \textbf{6\%} & 100\% & 100\% \\
2 y        & 65\% & \textbf{0\%} & 100\% & 98\% \\
5 y        & 83\% & \textbf{6\%} & 100\% & 94\% \\
10 y       & 100\% & 100\% & 100\% & 100\% \\
20 y       & 100\% & 100\% & 98\% & 90\% \\
50 y       & 100\% & 100\% & 98\% & 73\% \\
\bottomrule
\end{tabular}
\caption{Order stability (\% of prompt pairs giving the same answer regardless of presentation order) for the four-regime subset.
Bold values indicate catastrophic order bias ($<$10\%).
\texttt{Qwen3-4B-Instruct-2507} at 1--5 years is essentially ``pick whatever appears first''.}
\label{tab:order-stability}
\end{table}

\begin{figure}[!htbp]
  \centering
  \includegraphics[width=0.75\textwidth]{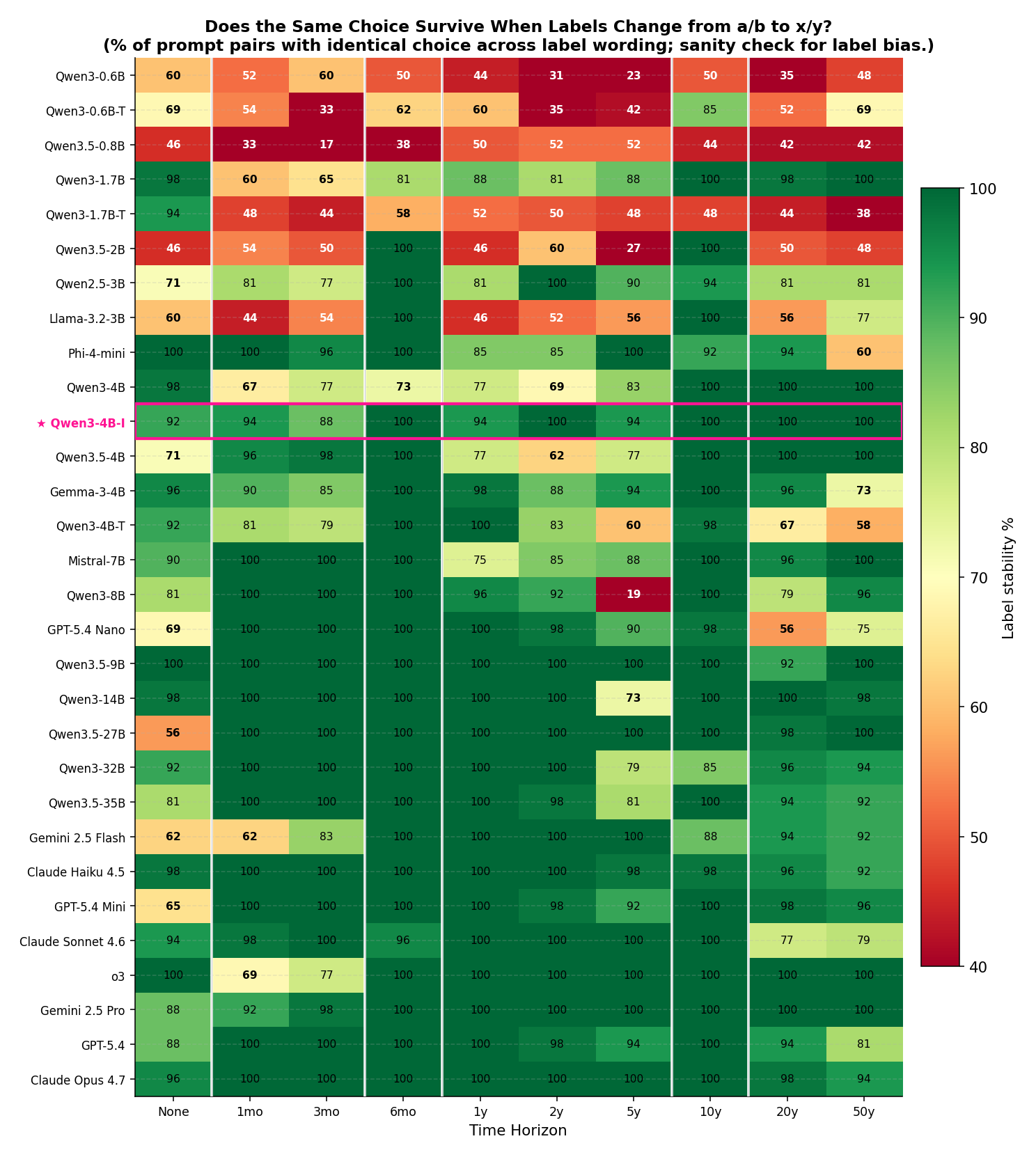}
  \caption{Label-format stability across 30 models $\times$ 10 horizons.
  Each cell: \% of prompt pairs giving the same answer when labels change from \texttt{a/b} to \texttt{x/y}, holding order, reward, horizon, and framing fixed.
  Target model \texttt{Qwen3-4B-Instruct-2507} highlighted.}
  \label{fig:label-stability}
\end{figure}

\paragraph{Label stability.}
Figure~\ref{fig:label-stability} swaps the label format between \texttt{a/b} and \texttt{x/y} holding everything else fixed.
Most models are near-perfectly label-stable at the anchor horizons; stability dips inside the reasoning zone for several families, compounding the order-bias instability in the same zone.

\begin{figure}[!htbp]
  \centering
  \includegraphics[width=0.75\textwidth]{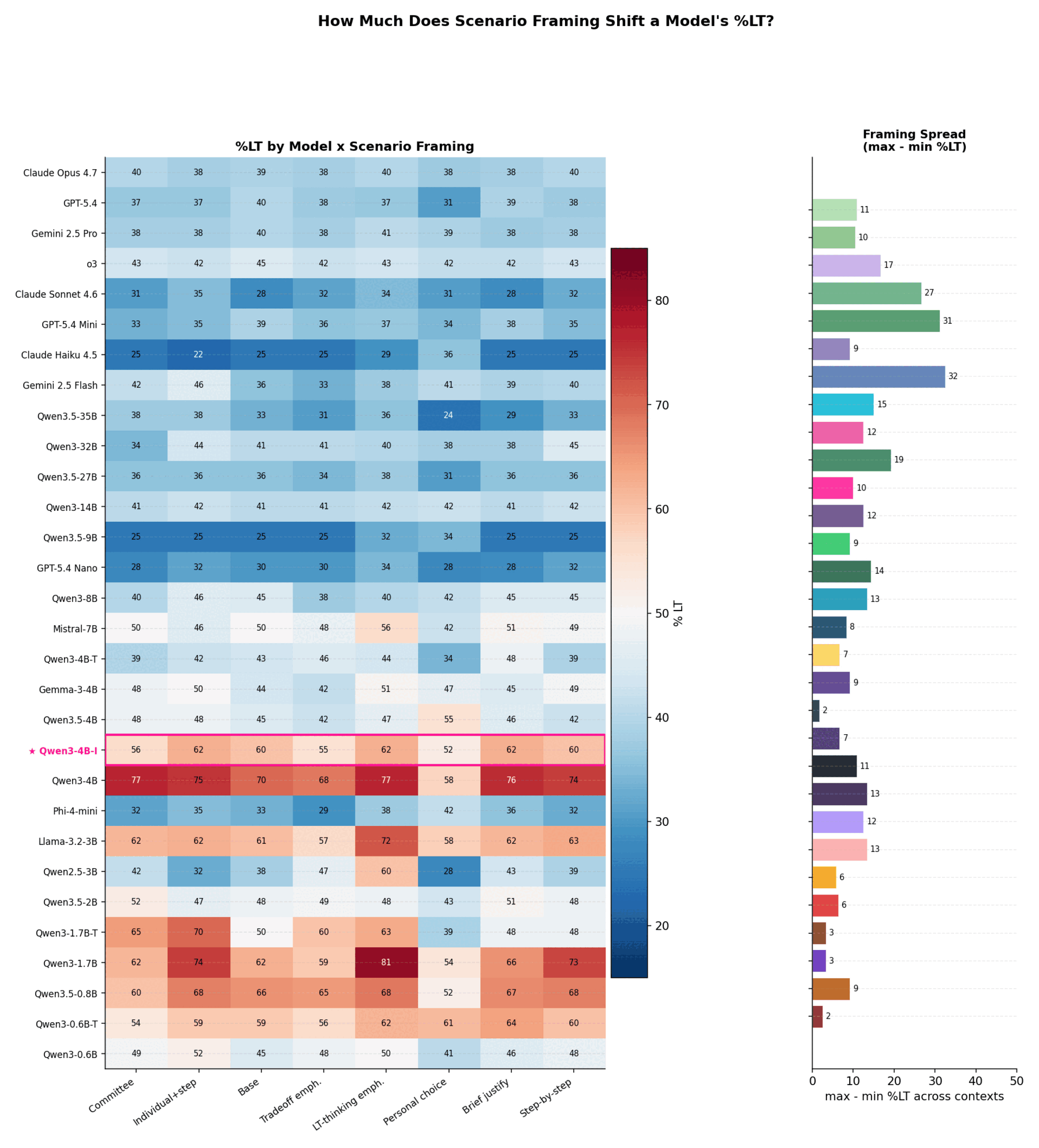}
  \caption{Long-term preference across 8 scenario framings for all 30 models.
  Context sensitivity is idiosyncratic and can flip sign between families: ``Long-term thinking emphasis'' and ``Personal choice'' framings produce the largest cross-model divergence.}
  \label{fig:context-sensitivity}
\end{figure}

\paragraph{Context stability.}
Figure~\ref{fig:context-sensitivity} sweeps the scenario framing across 8 contexts (household head vs.\ individual vs.\ committee, various reasoning-style emphases).
The left panel shows \%LT per model per context; the right panel reports the max$-$min \%LT spread per model.
Context sensitivity is idiosyncratic: some models shift $>$20pp across framings while others barely move.

\FloatBarrier

\subsection{Are Choices Coherent (in the 1--5y reasoning zone)?}\label{app:coherence-rational}

Coherence is the \emph{only} metric that distinguishes horizon-aware temporal reasoning from pattern matching.
We define it strictly: the fraction of choices that pick the rational short-term option on horizon-bearing prompts in the temporal reasoning zone (1y, 2y, 5y), where only the 6-month ST option can deliver within the stated deadline.
At anchor horizons (6mo, 10y) or beyond 10y, the rational rule coincides with pattern-matching or with expected-value dominance, so coherence is not separable from those.

\begin{figure}[!htbp]
  \centering
  \includegraphics[width=0.75\textwidth]{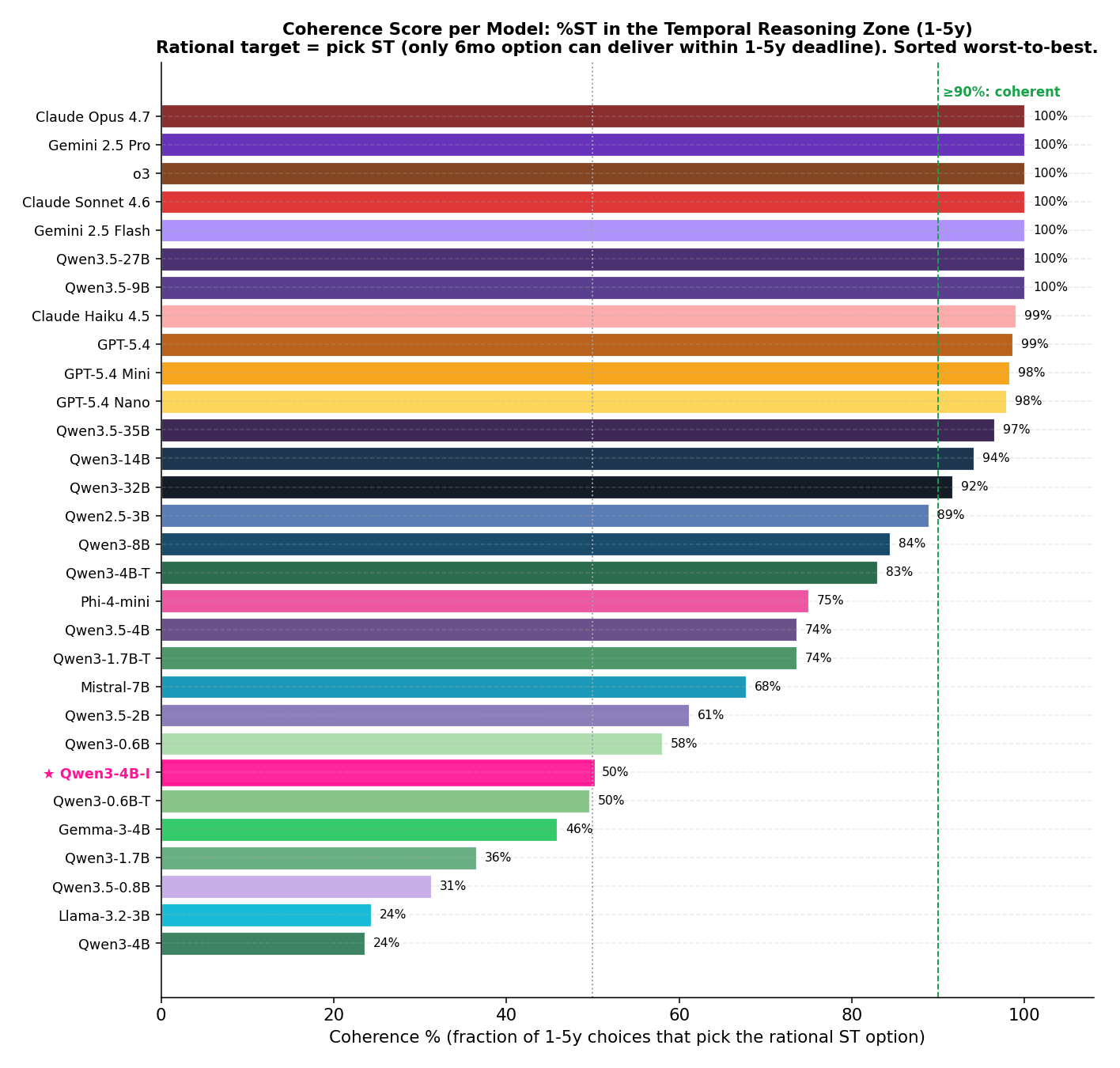}
  \caption{Coherence score per model: \% of choices in the 1--5y reasoning zone that pick the rational short-term option.
  \texttt{Claude Opus 4.7}, \texttt{Gemini 2.5 Pro}, \texttt{Claude Sonnet 4.6}, and \texttt{GPT-5.4} all achieve 100\% coherence in this zone; \texttt{Qwen3-4B} (hybrid-thinking) is at 24\% (systematically picks the wrong long-term option); our target \texttt{Qwen3-4B-Instruct-2507} sits at 50\% (the positional-polarization regime).
  Reaching 100\% here is necessary but not sufficient for genuine reasoning: some families (e.g., \texttt{Claude}) reach it via a binary ``under 10 years = ST'' heuristic that collapses at longer horizons (\ref{app:coherence-other}).}
  \label{fig:coherence-score}
\end{figure}

\paragraph{Per-model coherence score.}
Figure~\ref{fig:coherence-score} reports the single-number coherence score per model, sorted worst-to-best.

\begin{figure}[!htbp]
  \centering
  \includegraphics[width=0.85\textwidth]{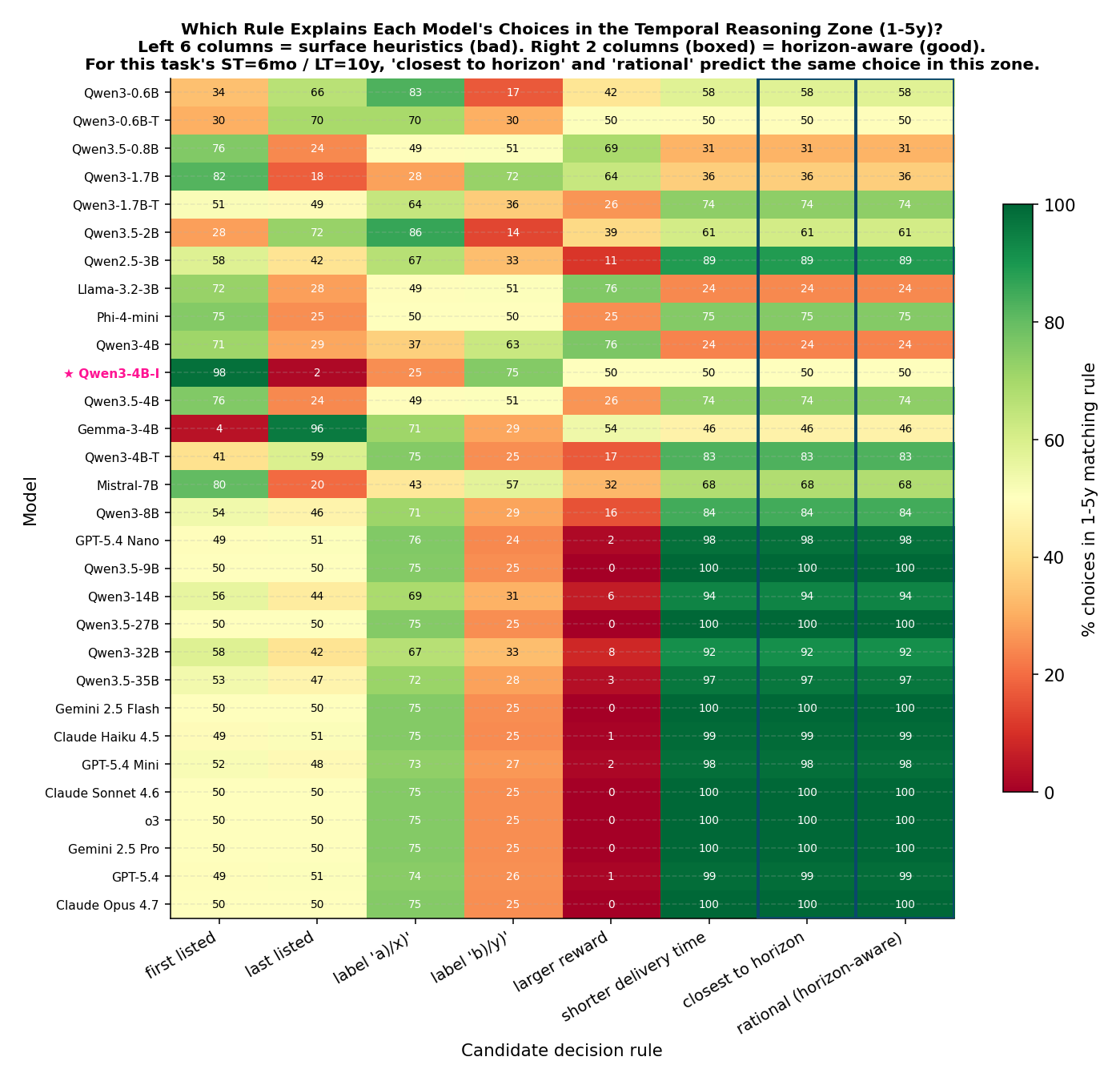}
  \caption{Per-rule match rate in the temporal reasoning zone.
  The ``closest to horizon'' rule predicts the same choice as the ``rational (can-deliver)'' rule at the delivery times used here (ST=6mo, LT=10y), so their columns agree.
  Models whose best-explaining rule is a position or label heuristic are following surface cues, not reasoning; the target model's rule profile is dominated by ``first listed''.}
  \label{fig:rule-heuristic}
\end{figure}

\paragraph{Which rule explains the model's 1--5y choices?}
Figure~\ref{fig:rule-heuristic} scores each model against eight candidate decision rules, restricted to the 1--5y reasoning zone.
The last two columns (boxed) are horizon-aware; the first six are surface heuristics that, if dominant, signal that the model is not actually reasoning about the deadline.

\begin{figure}[!htbp]
  \centering
  \includegraphics[width=\textwidth]{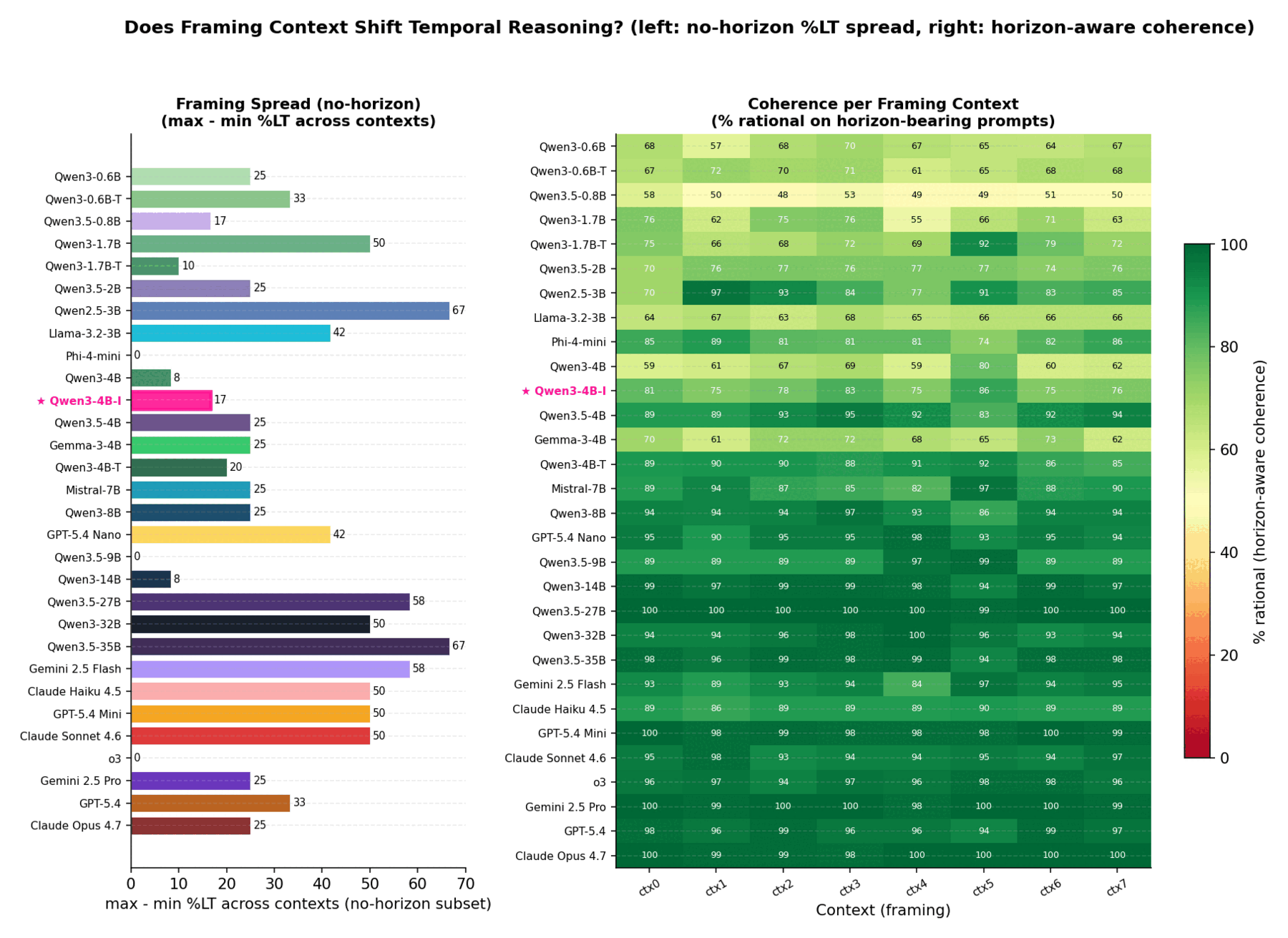}
  \caption{\textbf{Left:} Max$-$min \%LT spread across 8 scenario framings on no-horizon prompts (how much framing alone can flip the default preference).
  \textbf{Right:} Horizon-aware coherence (\% rational on horizon-bearing prompts) broken down by context.
  ``Committee'' and ``Tradeoff emphasis'' framings raise coherence for most models; ``Personal choice'' and the bare ``Base'' framing depress it.}
  \label{fig:context-coherence}
\end{figure}

\paragraph{Per-context coherence.}
Coherence is not uniform across scenario framings.
Figure~\ref{fig:context-coherence} pairs the no-horizon context spread per model (left panel) with per-context coherence on horizon-bearing prompts (right).

\FloatBarrier

\subsection{What Is the Latent Temporal Preference?}\label{app:coherence-latent}

When no horizon is stated, the model has no rational target and reveals its \emph{default} disposition.
Decomposing this by presentation order is critical: a model that picks LT $\sim$100\% of the time when ST appears first but only $\sim$20\% when LT appears first does not have a $\sim$60\% latent LT preference, it has essentially no preference and is mostly picking the second option (with a small residual LT lean).

\begin{figure}[!htbp]
  \centering
  \includegraphics[width=0.75\textwidth]{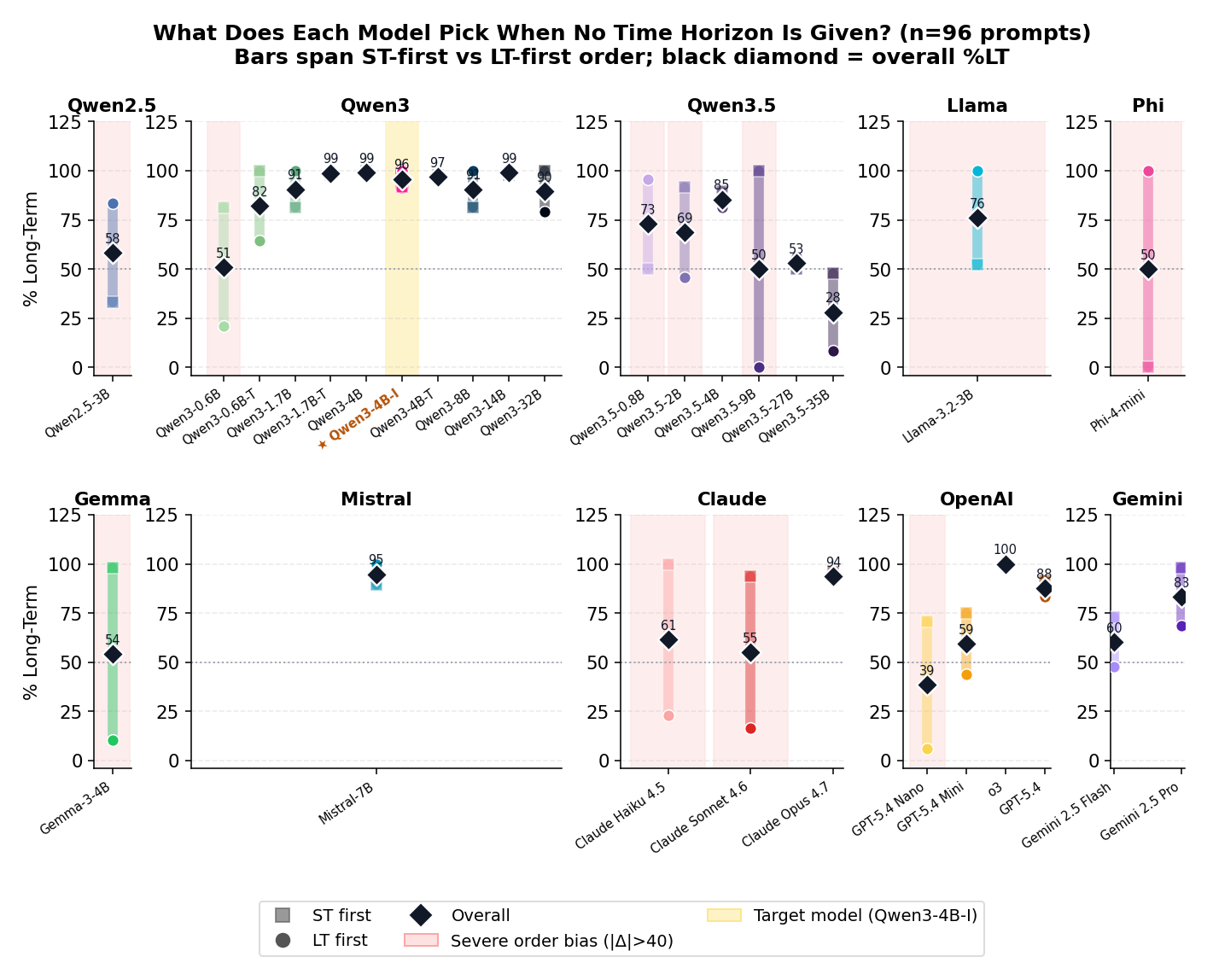}
  \caption{No-horizon \%LT decomposed by presentation order.
  \texttt{Claude Haiku 4.5}, \texttt{Claude Sonnet 4.6}, and several others show pure order bias ($\sim$100\% LT when ST appears first vs.\ $\sim$20\% when LT appears first); their apparent mid-range overall preference is a positional artifact.
  All three \texttt{Qwen3-4B} variants (hybrid-thinking, instruct-2507, thinking-2507) and \texttt{Claude Opus 4.7} are nearly order-invariant and express a genuine long-term default.}
  \label{fig:no-horizon-bias}
\end{figure}

\paragraph{No-horizon order decomposition.}
Figure~\ref{fig:no-horizon-bias} decomposes the no-horizon \%LT by presentation order across all 30 models.

\begin{table}[!htbp]
\centering
\small
\begin{tabular}{lccc}
\toprule
\textbf{Model} & \textbf{ST-first \%LT} & \textbf{Overall \%LT} & \textbf{LT-first \%LT} \\
\midrule
\texttt{Qwen3-4B-Instruct-2507}      & 92\%  & 96\% & 100\% \\
\texttt{Qwen3-4B} (thinking)         & 98\%  & 97\% & 96\% \\
\texttt{Qwen3-4B} (non-thinking)     & 100\% & 99\% & 98\% \\
\texttt{Claude Haiku 4.5}            & 100\% & 62\% & 23\% \\
\texttt{Claude Sonnet 4.6}           & 94\%  & 55\% & 17\% \\
\texttt{Claude Opus 4.7}             & 96\%  & 94\% & 92\% \\
\texttt{GPT-5.4}                     & 92\%  & 88\% & 83\% \\
\bottomrule
\end{tabular}
\caption{No-horizon \%LT decomposed by presentation order.
Our primary target \texttt{Qwen3-4B-Instruct-2507} (the non-thinking-only 2507 refresh) and the original hybrid \texttt{Qwen3-4B} run in either thinking or non-thinking mode all express a genuine long-term default ($\sim$96--99\%) regardless of order.
\texttt{Claude Opus 4.7} and \texttt{GPT-5.4} also lean long-term with only small residual order effects, whereas \texttt{Claude Haiku 4.5} and \texttt{Claude Sonnet 4.6} collapse to pure order bias: they pick LT nearly always when it appears second and almost never when it appears first, yielding apparent mid-range overall \%LT that is entirely a positional artifact.}
\label{tab:no-horizon-decomp}
\end{table}

\begin{figure}[!htbp]
  \centering
  \includegraphics[width=0.75\textwidth]{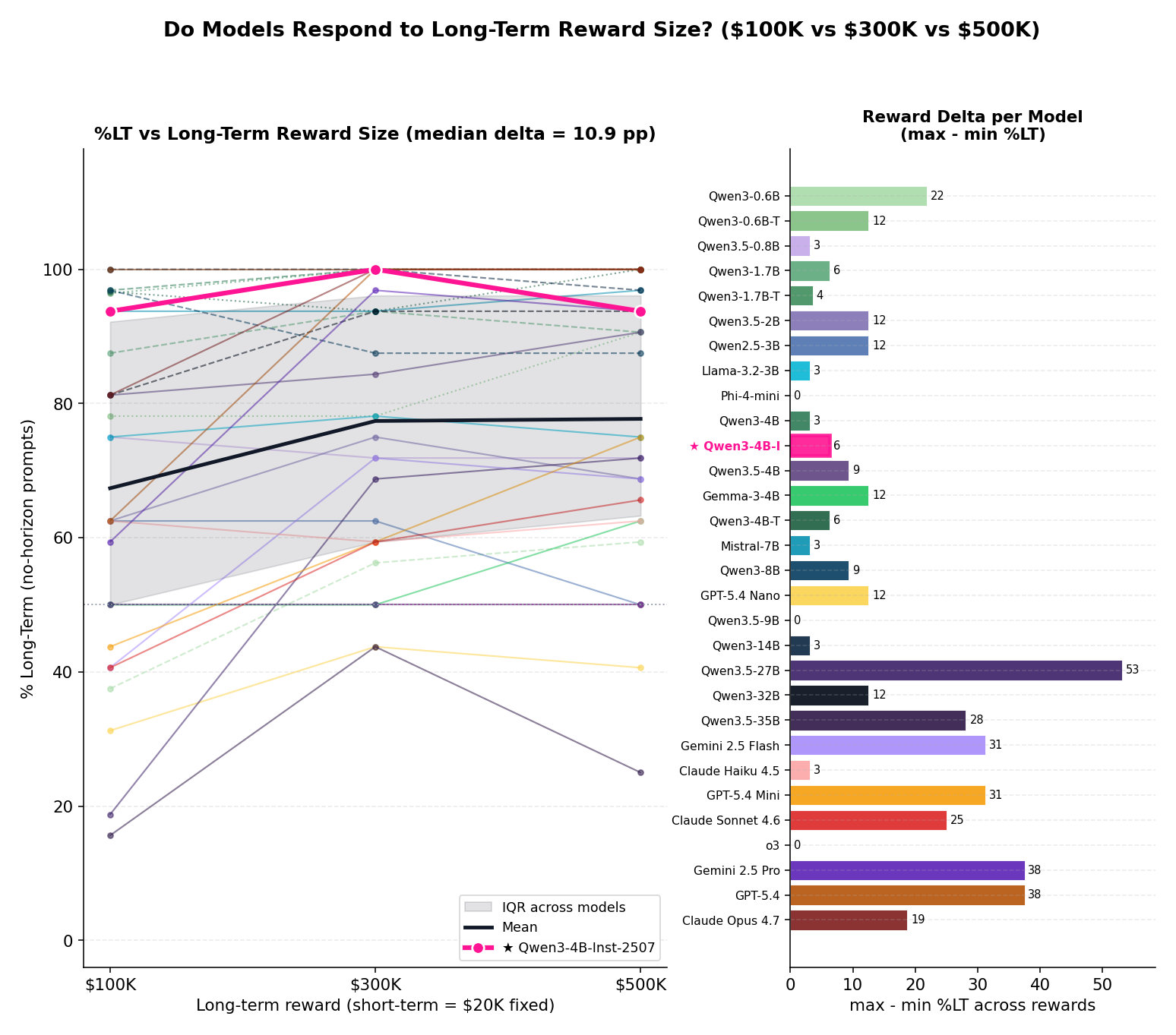}
  \caption{Reward sensitivity on no-horizon prompts across 30 models.
  Most models are saturated at ceiling or floor and move little with reward; \texttt{GPT-5.4} is the one clear exception in the representative subset (Table~\ref{tab:reward-sensitivity}).}
  \label{fig:reward-sensitivity}
\end{figure}

\paragraph{Reward sensitivity (no-horizon).}
Figure~\ref{fig:reward-sensitivity} tests whether default \%LT moves with the long-term reward size.
A rational economic agent should become more LT-oriented as the payoff grows from \$100K to \$500K.

\begin{table}[!htbp]
\centering
\small
\begin{tabular}{lcccc}
\toprule
\textbf{Model} & \textbf{\$100K} & \textbf{\$300K} & \textbf{\$500K} & \textbf{Spread} \\
\midrule
\texttt{Qwen3-4B}                 & 96.9\% & 100\%  & 100\%  & $+$3.1pp \\
\texttt{Qwen3-4B-Instruct-2507}   & 93.8\% & 100\%  & 93.8\% & $+$6.2pp \\
\texttt{Claude Opus 4.7}          & 81.2\% & 100\%  & 100\%  & $+$18.8pp \\
\texttt{GPT-5.4}                  & 62.5\% & 100\%  & 100\%  & $+$37.5pp \\
\bottomrule
\end{tabular}
\caption{No-horizon \%LT stratified by long-term reward size.
\texttt{GPT-5.4} is the only model in the subset with strong reward sensitivity ($+$37.5pp from \$100K to \$300K), consistent with its high coherence score (Figure~\ref{fig:coherence-score}); the \texttt{Qwen3} models are at ceiling regardless of reward.}
\label{tab:reward-sensitivity}
\end{table}

\FloatBarrier

\subsection{Cross-Cutting Patterns}\label{app:coherence-other}

This section collects patterns that don't fit neatly into stability, coherence, or latent preference: the raw per-horizon curve, the \texttt{Claude} step function, the \texttt{Qwen3} hybrid vs.\ mode-specialized comparison, and the target-model deep dive.

\begin{figure}[!htbp]
  \centering
  \includegraphics[width=\textwidth]{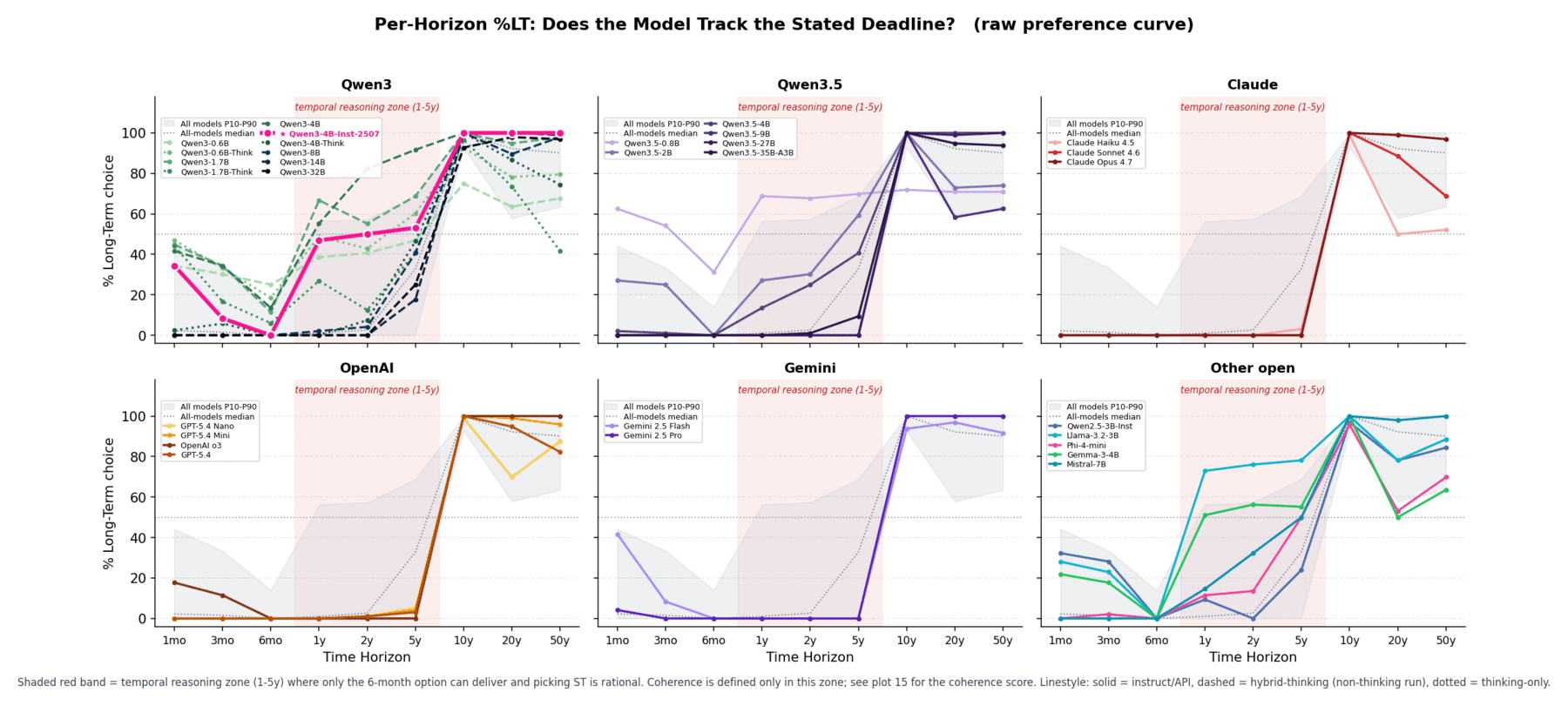}
  \caption{\%LT by time horizon across all 30 models, in per-family small multiples.
  The all-model P10--P90 envelope (gray band) and median (dotted) are shown for context.
  Within the temporal reasoning zone (1--5y, shaded red), the rational \%LT target is 0; at horizons of 10y and beyond, the rational target is 100.
  The target model \texttt{Qwen3-4B-Instruct-2507} (starred) sits near 50\% in the reasoning zone, an average of two near-deterministic order-polarized sub-behaviors (\ref{app:coherence-target-deep-dive}).}
  \label{fig:coherence-curve}
\end{figure}

\paragraph{Raw per-horizon \%LT curve.}
Figure~\ref{fig:coherence-curve} plots \%LT vs.\ time horizon for all 30 models, grouped into per-family panels.
This is the raw preference shape; the shaded red band marks the 1--5y reasoning zone where coherence is defined.

\begin{figure}[!htbp]
  \centering
  \includegraphics[width=0.65\textwidth]{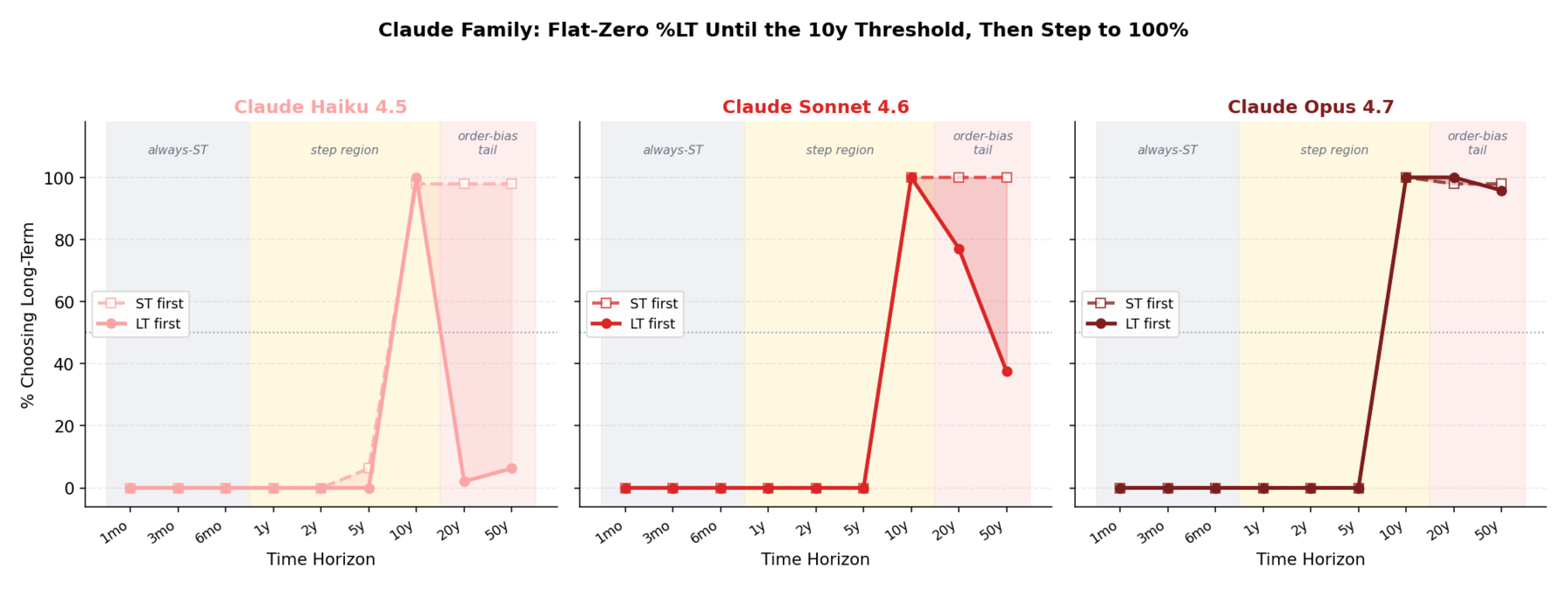}
  \caption{\texttt{Claude} family step function: flat 0--3\% LT for all horizons under 10 years, step to 99\% at 10 years.
  A binary cutoff rule (``under 10 years = short-term'') explains the pattern; the model is not reasoning about deliverability, it is threshold-matching.}
  \label{fig:claude-step}
\end{figure}

\paragraph{The \texttt{Claude} step function.}
Figure~\ref{fig:claude-step} isolates the \texttt{Claude} family's characteristic pattern: 0\% LT at every horizon under 10 years, then a hard step to $\sim$99\% at 10 years.
This is maximally coherent in the reasoning zone (by heuristic, not reasoning), but collapses to order bias at 20--50y for the smaller \texttt{Claude} variants.

\begin{figure}[!htbp]
  \centering
  \includegraphics[width=\textwidth]{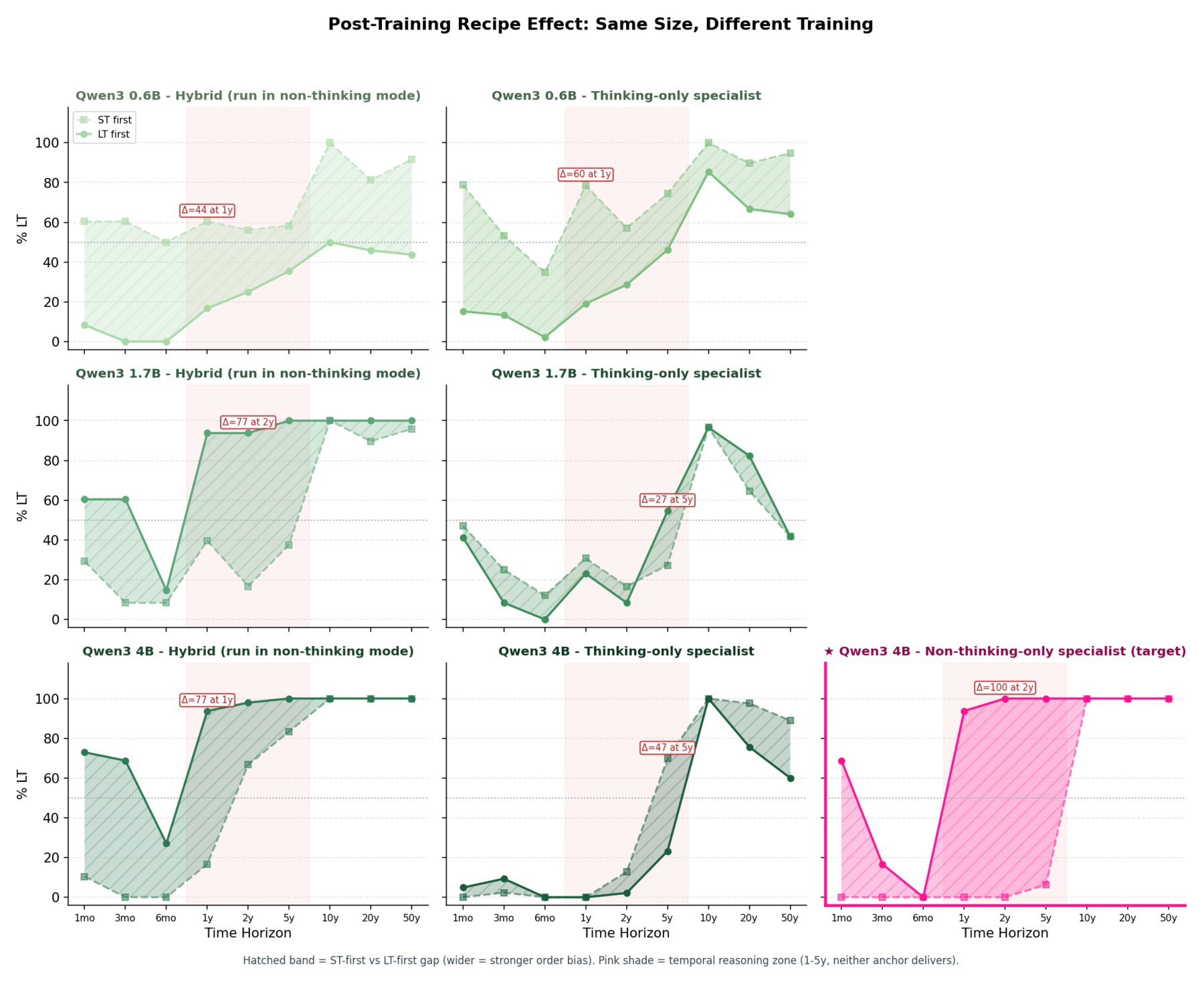}
  \caption{Within-family mode comparison across three \texttt{Qwen3} sizes (0.6B, 1.7B, 4B).
  Columns: hybrid \texttt{Qwen3-*} run in non-thinking mode, the same hybrid run in thinking mode, and the non-thinking-only \texttt{Qwen3-*-Instruct-2507} specialist (target variant at 4B starred).
  Each panel overlays \%LT under ST-first (dashed) and LT-first (solid) orderings with the gap shaded.
  Mode specialization into non-thinking replaces the hybrid checkpoint's graded horizon curve with a three-mode lookup pattern and a large order gap in the reasoning zone.}
  \label{fig:qwen3-mode-specialization}
  \label{fig:instruct-vs-base}
\end{figure}

\paragraph{\texttt{Qwen3} hybrid-thinking vs.\ mode-specialized 2507 variants.}
Figure~\ref{fig:qwen3-mode-specialization} compares the hybrid-thinking \texttt{Qwen3-\{0.6B, 1.7B, 4B\}} checkpoints against their thinking-only and non-thinking-only 2507 refreshes.
The hybrid-thinking and thinking-only variants preserve graded temporal sensitivity (informative but instrumentally incoherent); the distilled non-thinking-only variants collapse into three discrete modes with order bias in the reasoning zone.

\paragraph{Per-horizon \%LT (full subset breakdown).}
Table~\ref{tab:coherence-horizon} gives the full per-horizon breakdown for the four-regime subset.

\begin{table}[!htbp]
\centering
\small
\begin{tabular}{llcccc}
\toprule
\textbf{Horizon} & \textbf{Zone} & \texttt{Qwen3-4B} & \texttt{Qwen3-4B-Inst} & \texttt{Claude Opus 4.7} & \texttt{GPT-5.4} \\
\midrule
1 mo   & Before ST anchor & 42\%  & 34\%  & 0\%   & 0\% \\
3 mo   & Before ST anchor & 34\%  & 8\%   & 0\%   & 0\% \\
6 mo   & Exact match (ST) & 14\%  & 0\%   & 0\%   & 0\% \\
1 y    & Reasoning zone   & 55\%  & 47\%  & 0\%   & 0\% \\
2 y    & Reasoning zone   & 82\%  & 50\%  & 0\%   & 1\% \\
5 y    & Reasoning zone   & 92\%  & 53\%  & 0\%   & 3\% \\
10 y   & Exact match (LT) & 100\% & 100\% & 100\% & 100\% \\
20 y   & Beyond LT anchor & 100\% & 100\% & 99\%  & 95\% \\
50 y   & Beyond LT anchor & 100\% & 100\% & 97\%  & 82\% \\
\bottomrule
\end{tabular}
\caption{\%LT by horizon and model for the four-regime subset.
In the reasoning zone (1--5y), only the 6-month ST option can deliver, so a coherent agent picks ST (0\% LT).
\texttt{Claude Opus 4.7} achieves this (and \texttt{GPT-5.4} nearly does: 0--3\%) but via different mechanisms.
\texttt{Qwen3-4B} has the smoothest horizon-sensitivity curve yet is instrumentally incoherent (82\% LT at 2y).
\texttt{Qwen3-4B-Instruct-2507}'s flat 47--53\% in this zone is an order-bias artifact (see Table~\ref{tab:order-stability}).
Beyond 10y, the smaller \texttt{Claude} variants and \texttt{GPT-5.4} erode toward order bias (see Section~\ref{app:coherence-stability}); the \texttt{Qwen3} models stay saturated.}
\label{tab:coherence-horizon}
\end{table}

\FloatBarrier

\subsubsection{\texttt{Qwen3-4B-Instruct-2507} deep dive}\label{app:coherence-target-deep-dive}

The cross-model panels establish the population pattern.
We now zoom into the primary model.
Three views decompose its 960 prompts along stimulus axes the tables aggregate over.

\begin{figure}[!htbp]
  \centering
  \includegraphics[width=0.85\textwidth]{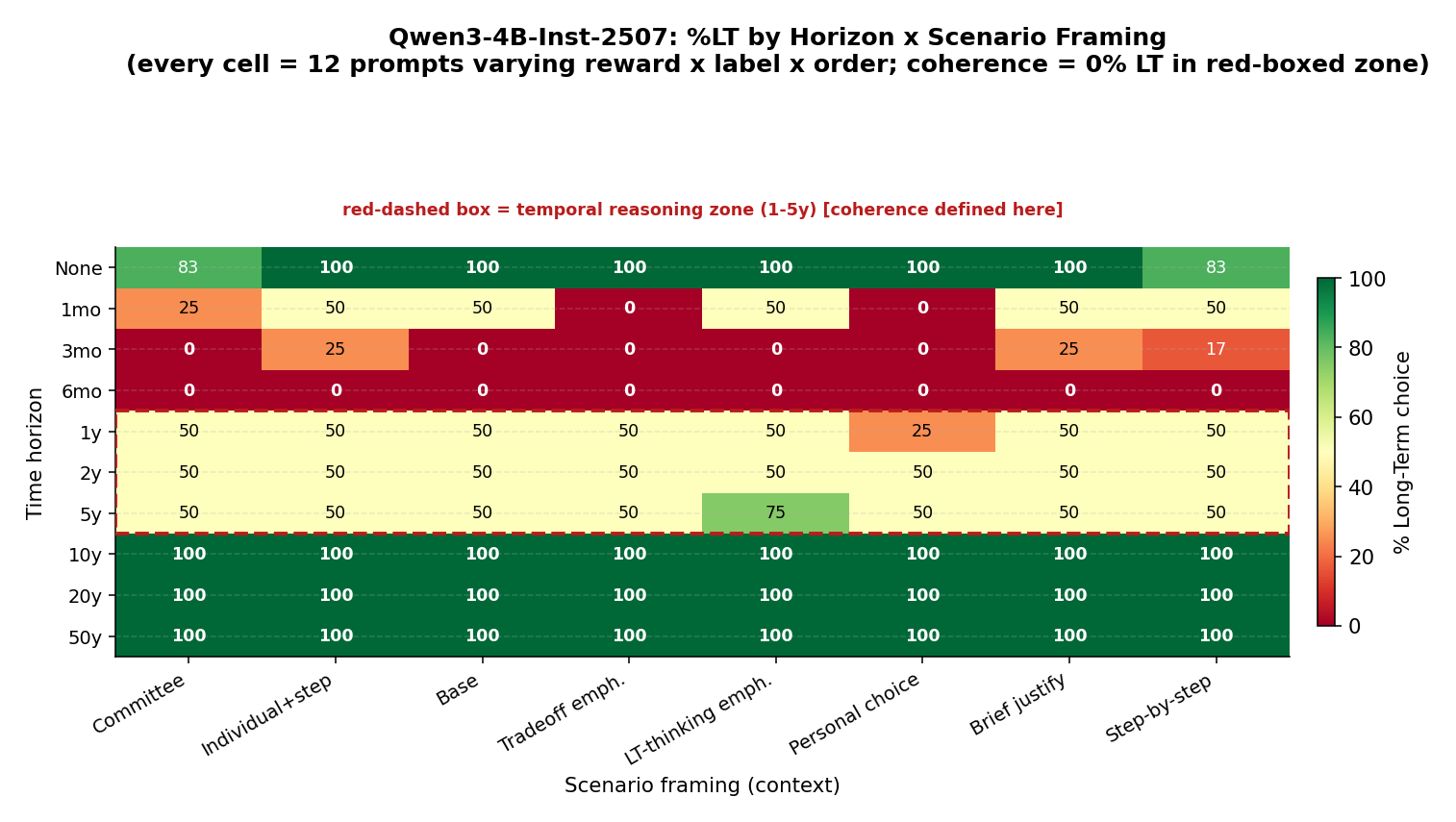}
  \caption{\texttt{Qwen3-4B-Instruct-2507}: \%LT by horizon and scenario framing.
  The anchor horizons (6mo, 10y) are context-insensitive and near-correct; the temporal reasoning zone (1--5y, dashed red box) is where framing has leverage.
  Within that zone, different framings push the model toward opposite choices, confirming that the pooled $\sim$50\% \%LT is an average over meaningfully different sub-behaviors, not a stable 50/50 uncertainty.}
  \label{fig:target-horizon-context}
\end{figure}

\paragraph{Horizon $\times$ context.}
Figure~\ref{fig:target-horizon-context} is a single-model \%LT heatmap over (horizon $\times$ context).
Each cell pools 12 prompts (3 rewards $\times$ 2 label styles $\times$ 2 orders).

\begin{figure}[!htbp]
  \centering
  \includegraphics[width=\textwidth]{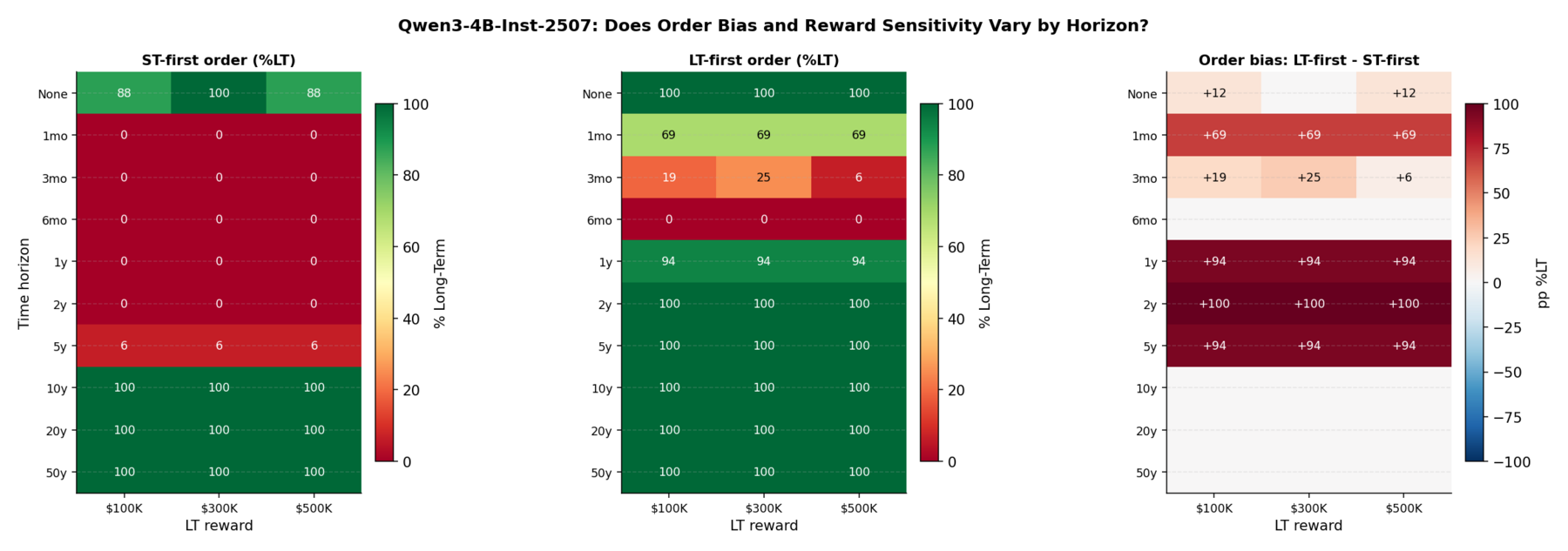}
  \caption{\texttt{Qwen3-4B-Instruct-2507}: \%LT under ST-first (left) and LT-first (middle) presentation orders, and the signed order-bias delta (right).
  Order bias is concentrated in the reasoning zone and is nearly reward-invariant within that zone: flipping the order changes \%LT by up to $\pm100$pp regardless of whether the long-term reward is \$100K or \$500K.
  The anchor horizons and the no-horizon condition show near-zero order bias.}
  \label{fig:target-horizon-reward-order}
\end{figure}

\paragraph{Horizon $\times$ reward $\times$ order.}
Figure~\ref{fig:target-horizon-reward-order} splits the same data by presentation order and shows the order-bias delta per (horizon, reward) cell.

\begin{figure}[!htbp]
  \centering
  \includegraphics[width=\textwidth]{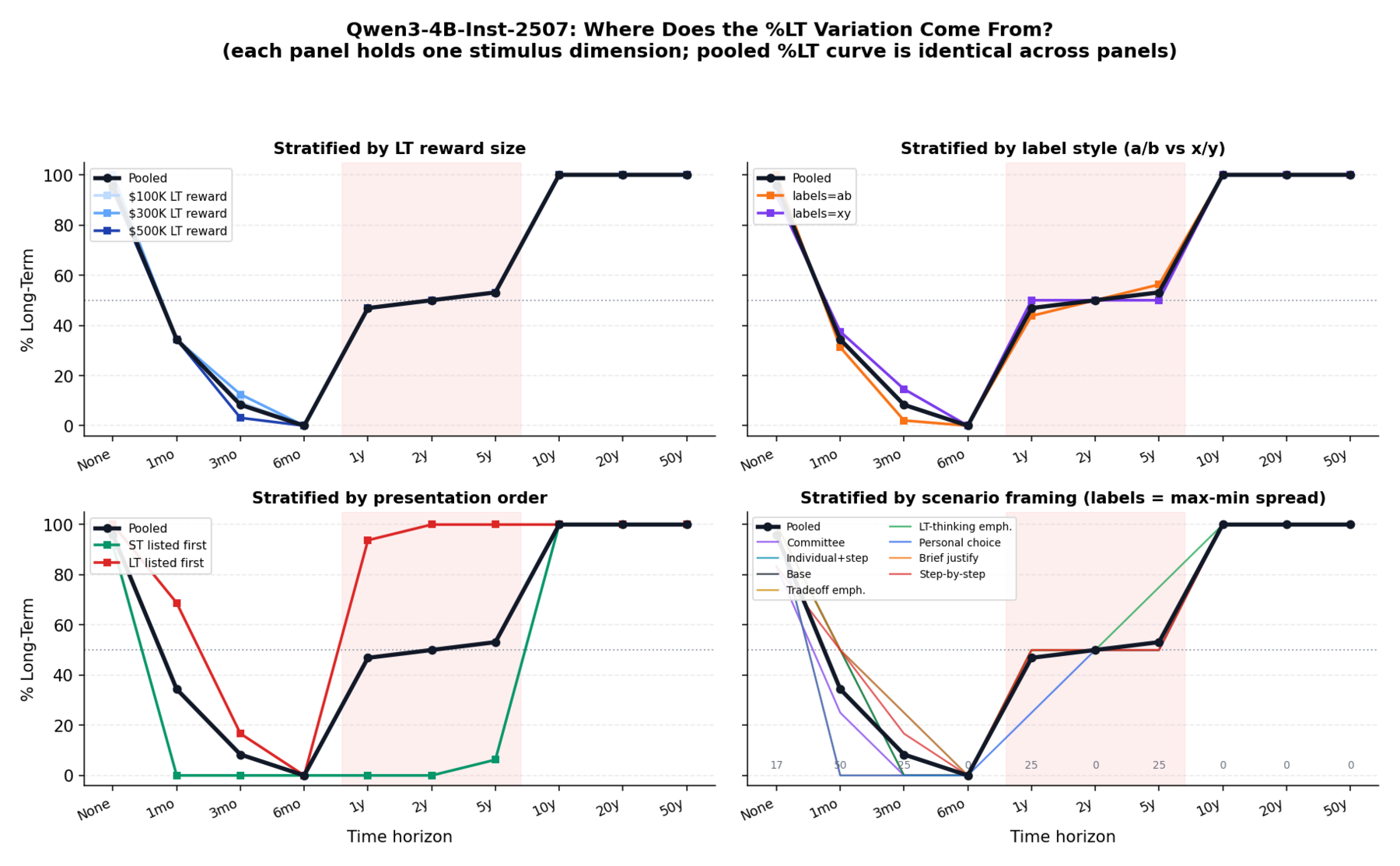}
  \caption{\texttt{Qwen3-4B-Instruct-2507}: per-horizon \%LT stratified by reward size (top-left), label style (top-right), presentation order (bottom-left), and scenario framing (bottom-right).
  The pooled curve (black) is identical across panels.
  Order stratification shows the dominant effect: the two order-conditioned curves sit at opposite extremes in the reasoning zone.
  Reward size and label format have negligible effect; context framing contributes moderate additional spread but is smaller than order.}
  \label{fig:target-variant-spread}
\end{figure}

\paragraph{Where does the variation come from?}
Figure~\ref{fig:target-variant-spread} stratifies the horizon curve by each stimulus dimension, holding the pooled curve fixed as reference.

Within the reasoning zone, \texttt{Qwen3-4B-Instruct-2507} is not uncertain, it is positionally polarized.
Almost all of the instability reported in the pooled tables is driven by presentation order, with a secondary contribution from context framing.
Reward magnitude and label format are effectively inert.

\FloatBarrier

\subsection{Key Findings}

\begin{enumerate}[leftmargin=1.5em]
  \item \textbf{Coherence lives in the 1--5y zone, not everywhere.}
  Agreement with the rational rule at the anchors (6mo, 10y) and beyond is pattern-matching or EV-dominance, not reasoning.
  The 1--5y zone is the only regime where the rational choice (pick ST) can be distinguished from a model just following the nearest anchor.

  \item \textbf{Most models fail the coherence test.}
  Only the large frontier API models (\texttt{Claude Opus 4.7}, \texttt{Gemini 2.5 Pro}, \texttt{Claude Sonnet 4.6}, \texttt{GPT-5.4}, \texttt{o3}, \texttt{GPT-5.4 Mini}, the \texttt{Claude} family more broadly) reach 95--100\% coherence in 1--5y.
  Most open-weight models at 4B-class and below sit at $<$50\%.

  \item \textbf{The \texttt{Claude} family is coherent by heuristic, not reasoning.}
  Its 100\% coherence in 1--5y is achieved by a binary cutoff (``under 10 years $\Rightarrow$ ST''); the smaller \texttt{Claude Haiku 4.5} and \texttt{Claude Sonnet 4.6} variants collapse to order bias at the longer horizons where this cutoff no longer applies (Figure~\ref{fig:order-stability}), while \texttt{Claude Opus 4.7} remains order-stable.
  The heuristic is functionally coherent for the 1--5y test but does not generalize.

  \item \textbf{Our target \texttt{Qwen3-4B-Instruct-2507} operates in three discrete modes.}
  At horizons under 6 months: coherent (picks ST, order-stable).
  In the 1--5y reasoning zone: pure positional polarization (0--6\% order stability), averaging to $\sim$50\% \%LT.
  At 10+ years: coherent (picks LT, order-stable).
  Mode specialization into non-thinking appears to have replaced graded horizon sensitivity with a lookup pattern.

  \item \textbf{The \texttt{Qwen3-4B} hybrid-thinking checkpoint is graded but wrong.}
  It has the smoothest horizon sensitivity curve (34\% LT at 3 months rising to 92\% at 5 years), consistent with continuous temporal representations in the geometry analysis (\ref{app:parametric-geometry}), but is instrumentally incoherent in the reasoning zone (82\% LT at 2 years, where LT cannot deliver).

  \item \textbf{Reward magnitude is largely inert; context framing is not.}
  Only \texttt{GPT-5.4} in the representative subset shows strong reward sensitivity ($+$37.5pp from \$100K to \$300K).
  Context framing produces comparable or larger shifts for several models.

  \item \textbf{Connection to the mechanistic story.}
  The geometry analysis shows the model encodes continuous temporal representations internally but collapses them into binary preference at the turn boundary (\ref{app:parametric-geometry}).
  The behavioral results show the same pattern at the output level: nuanced temporal sensitivity does not survive to coherent decision-making.
  This motivates the steering experiments in Part 3: if the internal representation is richer than the behavior, targeted intervention may recover the lost gradation.
\end{enumerate}

\clearpage
\clearappnumbering

\section{Cross-model patching comparison}\label{app:cross-model-comparison}

We repeat activation patching from causal parametric experiment (\ref{app:causal-parametric}) on nine Qwen3 variants spanning 0.6B--14B parameters, including our primary target \texttt{Qwen3-4B-Instruct-2507} and its hybrid-thinking sibling \texttt{Qwen3-4B}.
The question: is the temporal-preference subgraph localized at a consistent \emph{fractional depth} across model scales, or does it shift with parameter count?

\paragraph{Protocol.}
For each model, we collect clean and corrupted activations on the same prompt bank and measure two quantities at each layer for three hooks (\texttt{resid\_post}, \texttt{attn\_out}, \texttt{mlp\_out}): \emph{recovery}, the fraction of the clean--corrupted logit difference restored when a clean component is patched into the corrupted run, and 
\emph{disruption}, the fraction of that difference destroyed when a corrupted component is patched into the clean run. We plot mean recovery/mean disruption vs.\ \emph{fractional depth} ($\text{layer} / \text{total layers}$) to align curves across models of different depths.

\begin{figure}[htbp]
  \centering
  \includegraphics[width=\textwidth]{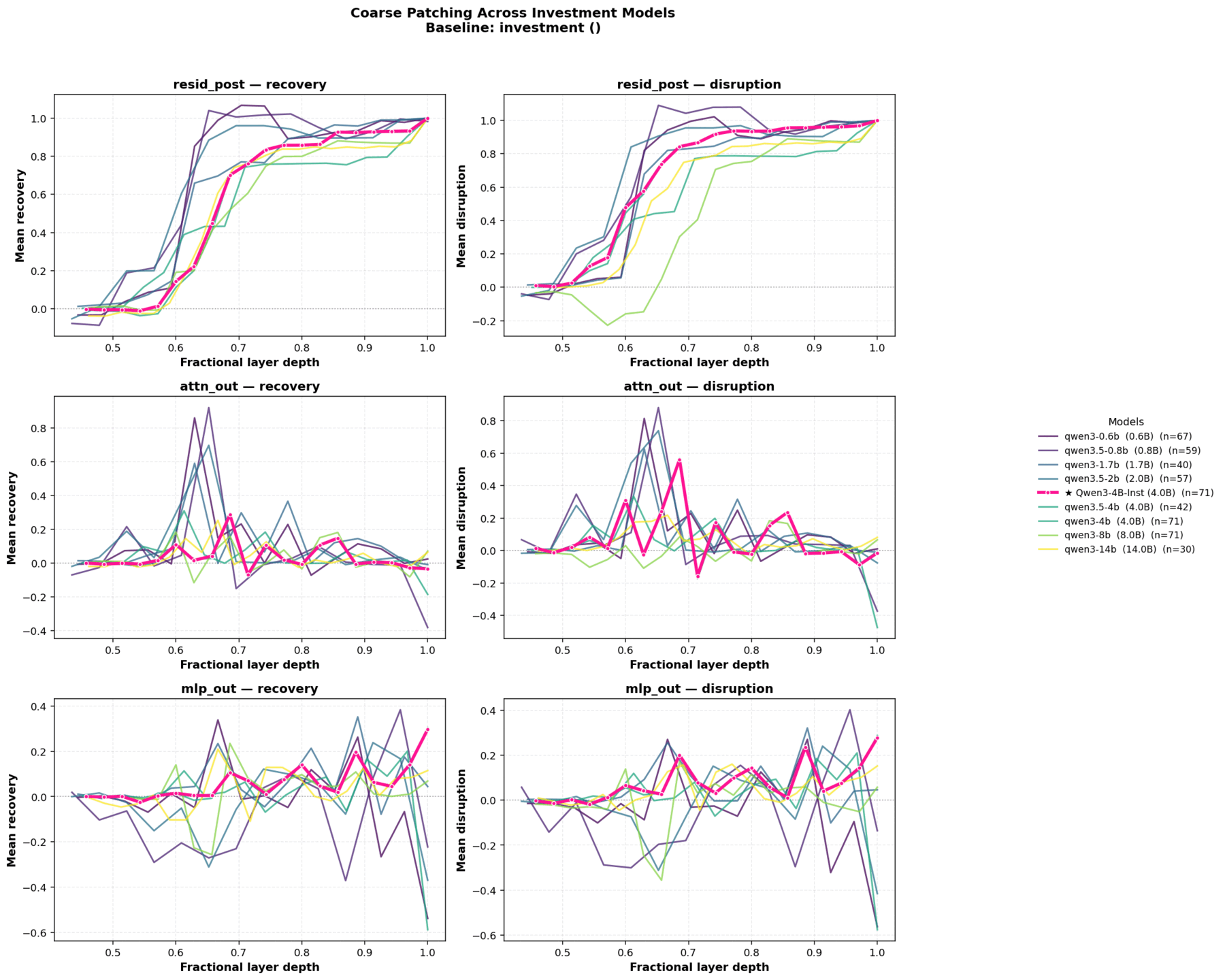}
  \caption{Cross-model patching overview at fractional depth.
  Rows: \texttt{resid\_post} (top), \texttt{attn\_out} (middle), \texttt{mlp\_out} (bottom).
  Columns: recovery (left), disruption (right).
  Nine Qwen3 variants overlaid (0.6B--14B).
  The residual stream saturates uniformly; attention localizes but weakens with scale; MLP stays diffuse.}
  \label{fig:cross-model-overview-depth}
\end{figure}

\begin{figure}[htbp]
  \centering
  \includegraphics[width=\textwidth]{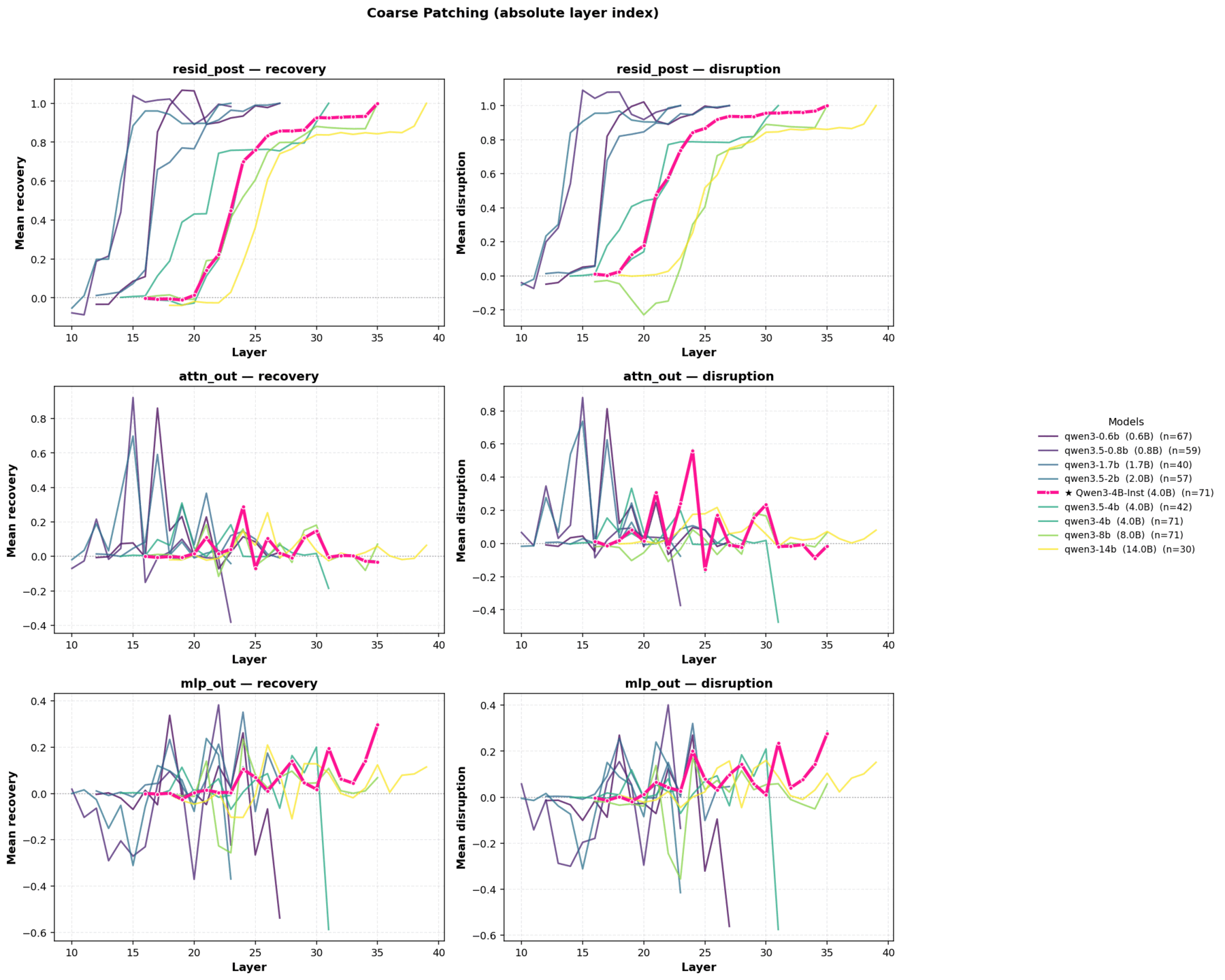}
  \caption{Same comparison on absolute layer indices rather than fractional depth.
  Without depth normalization the curves spread across layers 15--35 without aligning, confirming that fractional depth (not absolute index) is what stabilizes circuit location across scales.}
  \label{fig:cross-model-overview-abs}
\end{figure}

\begin{figure}[htbp]
  \centering
  \includegraphics[width=\textwidth]{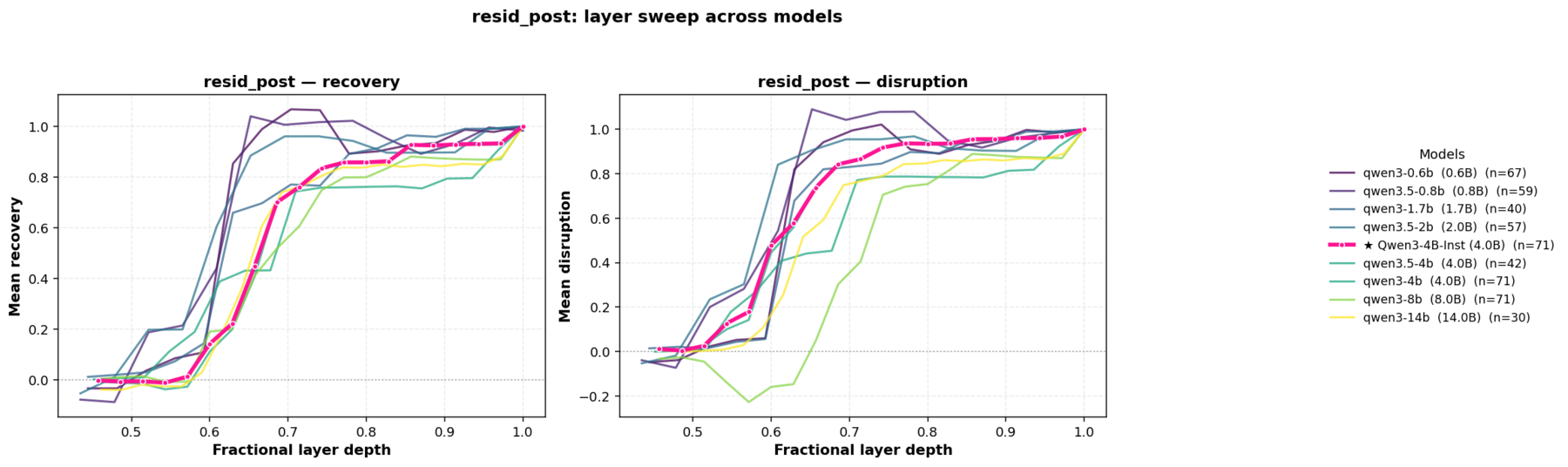}
  \caption{\texttt{resid\_post} recovery and disruption vs.\ fractional depth.
  Every model follows the same sigmoid, saturating at $\sim$1.0 by depth 0.8.}
  \label{fig:cross-model-compare-resid}
\end{figure}

\begin{figure}[htbp]
  \centering
  \includegraphics[width=\textwidth]{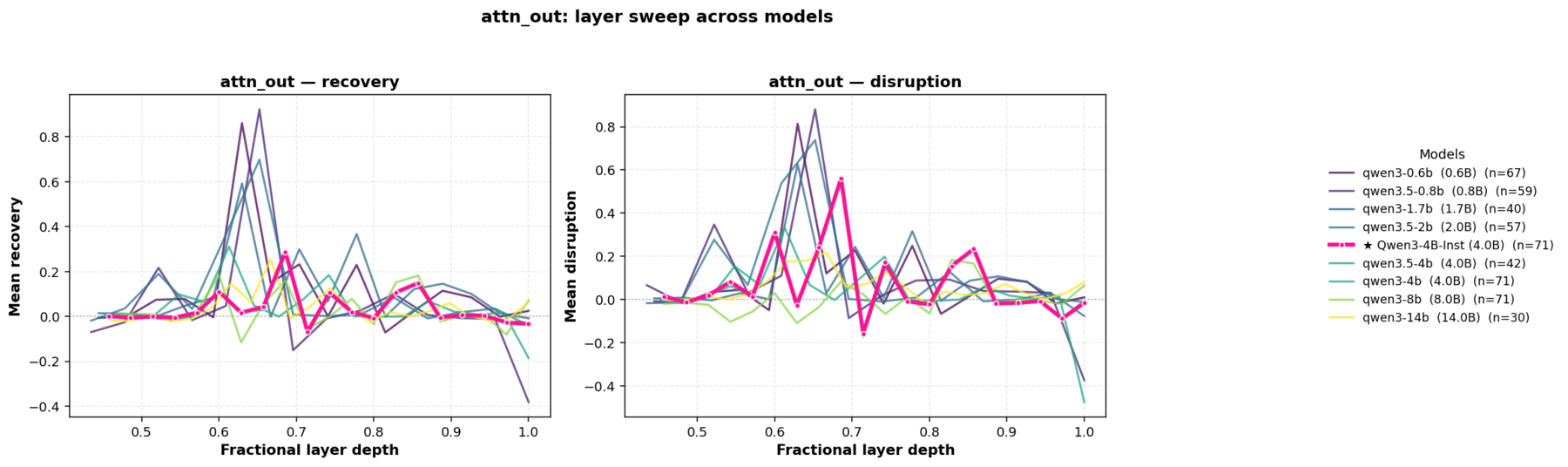}
  \caption{\texttt{attn\_out} recovery and disruption vs.\ fractional depth.
  The peak is narrow and consistent near depth 0.6--0.7, but its height shrinks monotonically with parameter count, from $\sim$0.9 (0.6B) to $\sim$0.2--0.3 (8--14B).}
  \label{fig:cross-model-compare-attn}
\end{figure}

\begin{figure}[htbp]
  \centering
  \includegraphics[width=\textwidth]{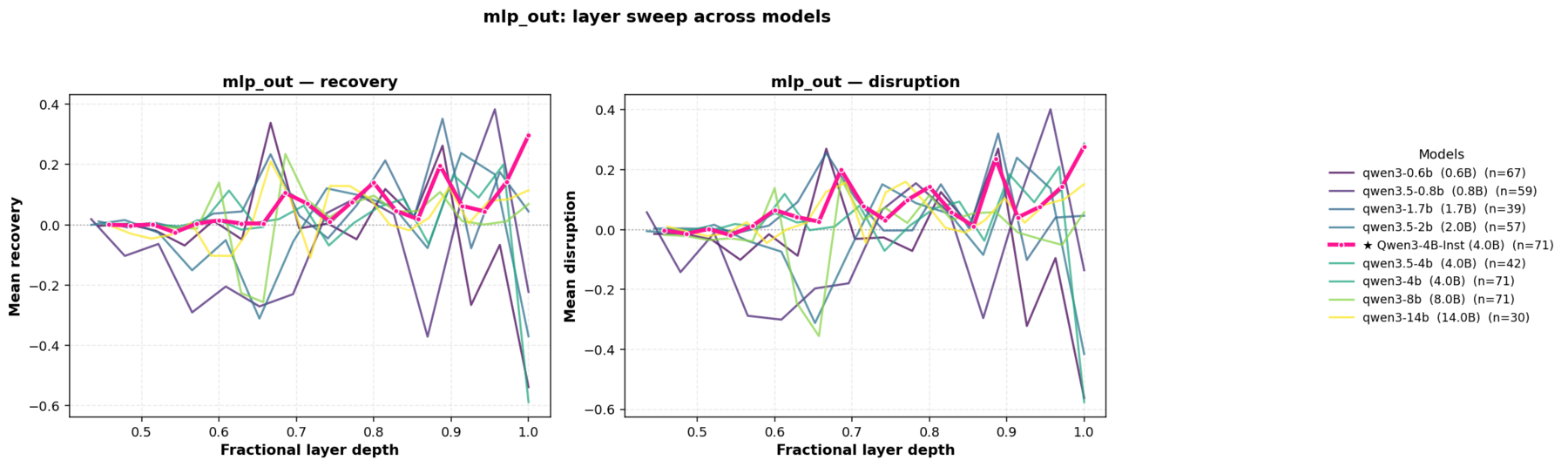}
  \caption{\texttt{mlp\_out} recovery and disruption vs.\ fractional depth.
  Effects are low ($\le 0.4$) and broadly distributed across mid-to-late layers for all models, with no sharp localization.}
  \label{fig:cross-model-compare-mlp}
\end{figure}

\paragraph{Findings.}
Three patterns hold across the family (Figures~\ref{fig:cross-model-overview-depth}--\ref{fig:cross-model-compare-mlp}).

\begin{itemize}[leftmargin=*]
  \item \textbf{Residual stream saturates.} \texttt{resid\_post} recovery is a clean sigmoid that crosses 50\% around depth 0.65--0.70 and saturates at 1.0 by depth 0.8 in every model (Figure~\ref{fig:cross-model-compare-resid}).
  The location of the transition is nearly scale-invariant in depth units.
  \item \textbf{Attention localizes at $\sim$0.6--0.7 depth, but its recovery shrinks with scale.} \texttt{attn\_out} peaks in a narrow band at depth 0.6--0.7 in all models, but peak recovery drops from $\sim$0.86--0.92 in the smallest models (0.6--1.7B) to $\sim$0.18--0.30 in the 4B--14B variants (Figure~\ref{fig:cross-model-compare-attn}).
  The circuit becomes more distributed, not absent, at scale.
  \item \textbf{MLP contribution is diffuse.} \texttt{mlp\_out} recovery stays below 0.4 for every model and is spread across mid-to-late layers without a sharp peak (Figure~\ref{fig:cross-model-compare-mlp}).
  The MLPs accumulate the preference rather than route it.
\end{itemize}

\noindent
Our primary target \texttt{Qwen3-4B-Instruct-2507} tracks the hybrid-thinking 4B checkpoint (\texttt{Qwen3-4B}) closely on all three hooks.

\paragraph{Caveat: denominator validity at small scales.}
For smaller models (0.6B, 1.7B) that fail to distinguish clean from corrupted prompts at baseline, the denominator $(y_{\text{clean}} - y_{\text{corrupted}})$ in the normalized recovery and disruption metrics (Eq.~\eqref{eq:causal-param-rec-eq},~\eqref{eq:causal-param-disr-eq}) approaches zero, making the resulting scores unstable.
We did not separately re-filter the bank per variant, nor do we report the per-model $(y_{\text{clean}} - y_{\text{corrupted}})$ distribution.
Readers should therefore interpret the recovery and disruption magnitudes for the smallest variants with caution; the relative shapes and peak locations across fractional depth, which are the qualitative claims we draw from this experiment, are more robust than the absolute heights.

\clearpage
\clearappnumbering

\section{Error monitoring in the temporal preference subgraph}\label{app:error-monitoring}

The localization results (Appendices~\ref{app:attributional-contrastive}--\ref{app:convergence}) converge on a subgraph at layers 17--35; the geometry results (\ref{app:parametric-geometry}) show that time horizon is encoded as a non-linear manifold within it; and the behavioral results (\ref{app:behavioral-coherence}) reveal that this rich internal structure does not survive to coherent decision-making.
A natural question is whether this gap between representation and behavior is specific to temporal reasoning or reflects a broader property of the subgraph region.
We test this by probing whether the same layers and token positions also encode a second meta-cognitive variable, the accumulated reliability of a multi-step reasoning chain, and whether the two variables share or compete for representational capacity.

\paragraph{Shared pipeline.}
All 4{,}650 samples (3{,}550 error hops + 1{,}100 temporal preference samples from $D_{\text{explicit}}$ and $D_{\text{implicit}}$; \ref{app:prompts}) are extracted through a \emph{single model load} of \texttt{Qwen3-4B-Instruct-2507} using the Qwen chat template.
Raw hidden states at 15 key layers are stored for all samples and jointly projected into a shared PCA-50 subspace fit on the full concatenation, ensuring that error and temporal representations inhabit the same coordinate system.
Probes use logistic regression ($C = 0.01$, balanced class weights), 10-fold cross-validation, and 500-permutation null distributions.
We note that probing establishes \emph{correlational} decodability, complementary to but distinct from the causal localization in Appendices~\ref{app:causal-parametric} and~\ref{app:causal-contrastive}; a feature being decodable at a layer does not entail that the layer is causally necessary for behavior.

\paragraph{Error injection dataset.}
We construct 1{,}250 contrastive multi-hop math reasoning chains (2--4 hops) with three conditions: \textit{clean}, \textit{error\_at\_1}, and \textit{error\_at\_2}, using five error types (off-by-one, wrong operator, wrong unit, magnitude error, wrong percentage base).
Each hop is wrapped in the Qwen chat template as a user-to-assistant turn pair, matching the format used throughout the main paper.

\FloatBarrier

\subsection{Does error state co-localize with temporal preference?}\label{app:error-colocalization}

Before asking whether error and temporal preference \emph{interact}, we check whether they occupy the same architectural region.
A positive answer would suggest the subgraph functions as a general meta-cognitive module rather than a temporal-specific circuit.

\begin{figure}[!htbp]
  \centering
  \includegraphics[width=0.95\textwidth]{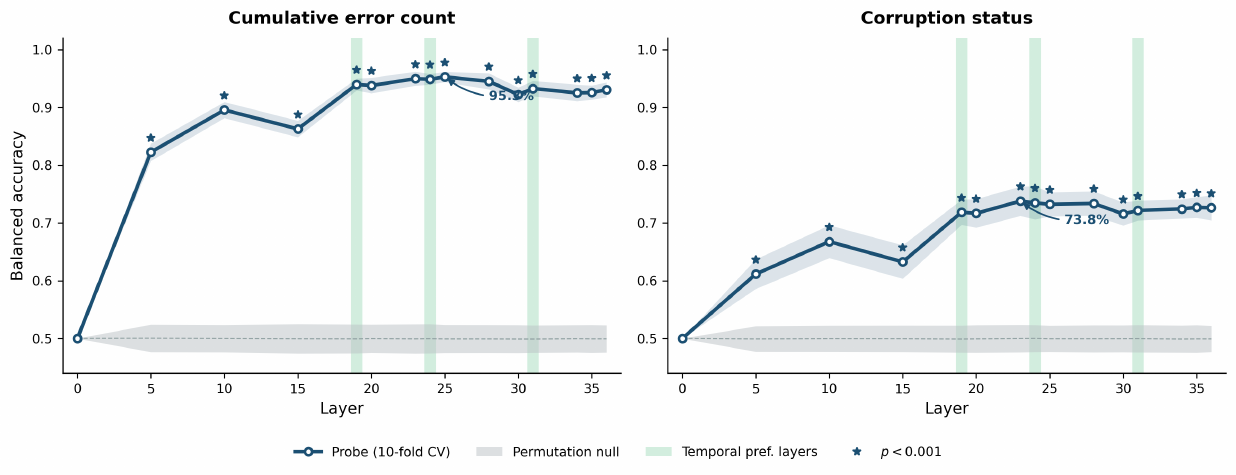}
  \caption{Error decodability in shared PCA-50 space ($n = 3{,}550$, 10-fold CV, 500-permutation null).
  Green bands: temporal preference subgraph layers (19, 24, 31).
  \textbf{Left:} Cumulative error count reaches a plateau of 94--95\% across layers 19--31, peaking at 95.3\% between layers 24 and 25.
  \textbf{Right:} Local corruption status (is this specific step injected?) peaks at 73.8\% at layer~23.
  Stars: $p < 0.001$.}
  \label{fig:error-probes}
\end{figure}

\paragraph{Layer-wise error probes.}
Figure~\ref{fig:error-probes} reports probe accuracy for three error targets across the 15 sampled layers.

\begin{table}[!htbp]
\centering
\small
\begin{tabular}{lccc}
\toprule
\textbf{Layer} & \textbf{Cumulative errors} & \textbf{Corruption status} & \textbf{Propagation} \\
\midrule
0  & 50.0\% & 50.0\% & 50.0\% \\
5  & 82.3\% & 59.5\% & 82.3\% \\
10 & 89.7\% & 60.9\% & 89.7\% \\
15 & 86.1\% & 63.4\% & 86.1\% \\
19 & 94.5\% & 71.3\% & 94.5\% \\
24 & 95.0\% & 73.5\% & 95.0\% \\
25 & \textbf{95.3\%} & 72.5\% & \textbf{95.3\%} \\
31 & 93.0\% & 72.6\% & 93.0\% \\
36 & 91.7\% & 71.5\% & 91.7\% \\
\bottomrule
\end{tabular}
\caption{Probe accuracy at selected layers (all $p < 0.001$ except layer~0).
Cumulative error count and propagation status are numerically identical, confirming the probe reads a chain-level property.
Corruption status is 21pp lower, indicating the model encodes ``my chain is degraded'' far more reliably than ``the error is at this step.''}
\label{tab:error-probes}
\end{table}

The cumulative error plateau (94--95\%) spans layers 19--31, precisely the subgraph identified by attribution patching in the main paper.
The 21-point gap between chain-level error (95\%) and local error identity (74\%) parallels the main paper's finding that global context properties (time horizon) are more structured than local behavioral outputs (specific choices in the reasoning zone; \ref{app:behavioral-coherence}).

\begin{figure}[!htbp]
  \centering
  \includegraphics[width=0.75\textwidth]{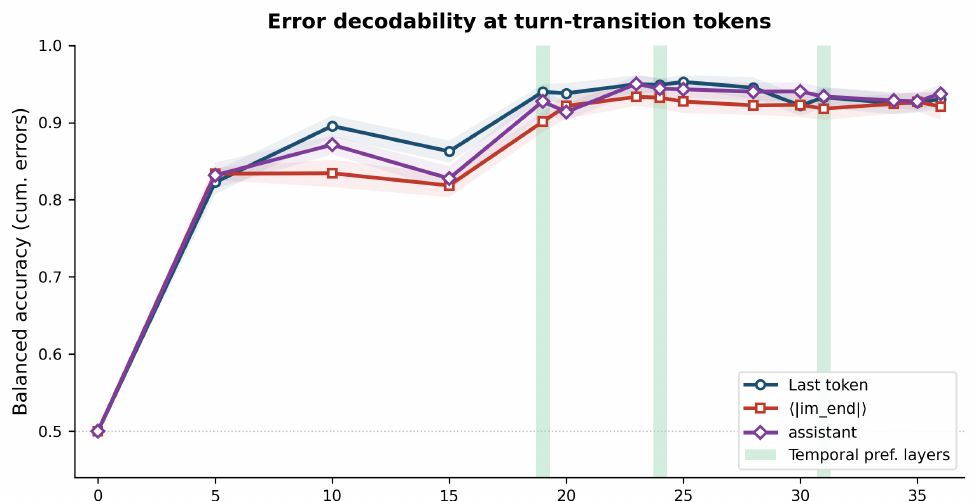}
  \caption{Cumulative error decodability at three token positions.
  All converge to $>$93\% by layer~19.
  The turn-transition tokens (\texttt{<|im\_end|>} and \texttt{assistant}), where \ref{app:parametric-geometry} shows temporal preference geometry crystallizing, carry error state with comparable fidelity to the last token.}
  \label{fig:error-turn}
\end{figure}

\paragraph{Error at the turn-transition tokens.}
The geometry analysis (\ref{app:parametric-geometry}) identifies the \texttt{<|im\_end|>} to \texttt{assistant} turn transition as the locus where temporal preference geometry becomes linearly separable (Figure~\ref{fig:turn-transition}).
Figure~\ref{fig:error-turn} tests whether error state follows the same pattern.

Error decodability at the turn-transition tokens matches the last token from layer~19 onward, with the \texttt{assistant} token slightly outperforming the last token at layers 24--31 (${\sim}$95\% vs.\ ${\sim}$93\%).
This convergence suggests that the turn-transition computation, the same computation that transforms off-policy context into on-policy generation for temporal preference, also integrates reasoning reliability before generation begins.

\FloatBarrier

\subsection{Do error and temporal preference share representational structure?}\label{app:error-cross-probe}

Co-localization does not entail shared structure: two variables can occupy the same layers in orthogonal subspaces.
We test this directly by training probes for each variable in the shared PCA-50 space and comparing the resulting weight vectors.

\paragraph{Cross-probing protocol.}
For each of the 15 key layers, we train a binary error probe (cumulative errors $> 0$ vs.\ $= 0$; $n = 3{,}550$) and a binary temporal probe (immediate vs.\ long-term; $n = 1{,}100$), both in the shared PCA-50 space.
We compute cosine similarity between the normalized weight vectors and test significance with a 500-permutation null (shuffle temporal labels, refit, recompute cosine).
We also measure cross-domain transfer: apply the error probe to temporal data (and vice versa) and test against 200-permutation nulls on the target labels.

\begin{figure}[!htbp]
  \centering
  \includegraphics[width=0.95\textwidth]{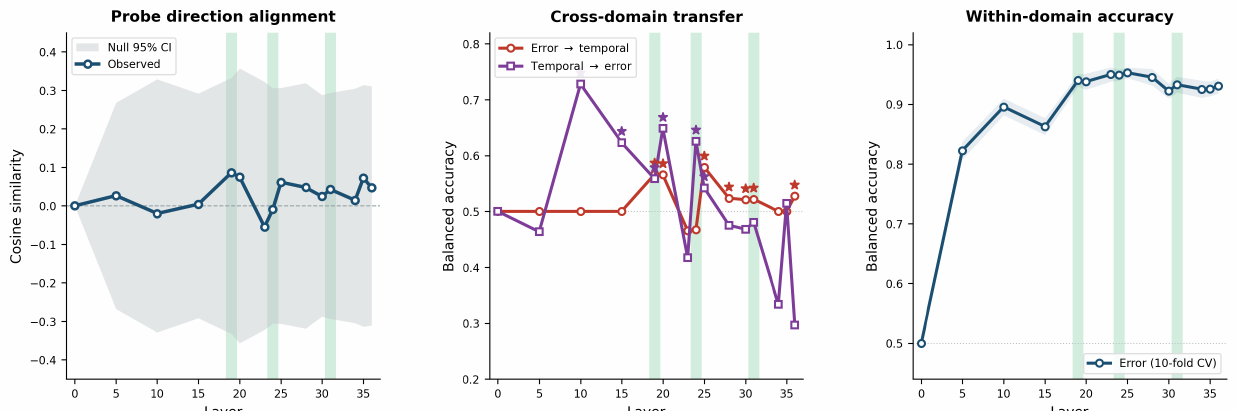}
  \caption{Cross-probing in shared PCA-50 space ($n_\text{err} = 3{,}550$, $n_\text{temp} = 1{,}100$, 500-permutation nulls).
  \textbf{Left:} Cosine between probe directions, all values fall within the permutation null (shaded), no $p < 0.05$.
  \textbf{Center:} Cross-domain transfer, temporal-to-error is significant at layers 10--25 (purple stars, $p < 0.001$); error-to-temporal is weaker and sporadic (red stars).
  \textbf{Right:} Within-domain error accuracy for reference.}
  \label{fig:error-cross-probe}
\end{figure}

\begin{table}[!htbp]
\centering
\small
\begin{tabular}{lcccc}
\toprule
\textbf{Layer} & \textbf{Cosine} & \textbf{Cosine $p$} & \textbf{Error$\to$Temp} & \textbf{Temp$\to$Error} \\
\midrule
10 & $+$0.040 & 0.774 & 0.500 & 0.728*** \\
19 & $+$0.086 & 0.614 & 0.500 & 0.593*** \\
20 & $+$0.071 & 0.614 & 0.574*** & 0.648*** \\
24 & $-$0.023 & 0.876 & 0.500 & 0.626*** \\
25 & $+$0.046 & 0.720 & 0.500 & 0.574*** \\
31 & $-$0.052 & 0.712 & 0.500 & 0.500 \\
\bottomrule
\end{tabular}
\caption{Cross-probe results at selected layers.
No cosine reaches significance.
Temporal-to-error transfer peaks at layer~10 (0.728) and remains above chance through layer~25; error-to-temporal transfer is at chance at most layers.
*** indicates $p < 0.001$ against the permutation null on target labels.}
\label{tab:cross-probe}
\end{table}

\paragraph{Interpretation: orthogonal directions, partial non-linear overlap.}
The cosine null result (all $p > 0.5$) establishes that the linear separating hyperplanes for error and temporal preference are perpendicular in the shared activation space.
However, the temporal-to-error transfer above chance at layers 10--25 ($p < 0.001$) shows that the temporal probe's projection of the data partially predicts error status even though the two probe \emph{directions} are orthogonal.
This combination, perpendicular hyperplanes paired with above-chance transfer, indicates that the two variables share a \emph{non-linear} subspace: their representations overlap on the activation manifold but not along any single linear axis.

This is consistent with the non-linear time-horizon geometry documented in~\ref{app:parametric-geometry}: if both error state and temporal preference occupy curved manifolds in the same region of activation space, their optimal linear separating hyperplanes can be orthogonal even as the manifolds themselves intersect.
The asymmetry of the transfer (temporal-to-error stronger than error-to-temporal) suggests that the temporal preference representation, which captures broad context evaluation (``strategic vs.\ tactical'' orientation; \ref{app:prompts}), carries some error-relevant information as a byproduct, while the error direction (a narrower signal about chain corruption) does not carry temporal scope information.

\FloatBarrier

\subsection{Key findings}\label{app:error-key-findings}

\begin{enumerate}[leftmargin=1.5em]
  \item \textbf{Error state co-localizes with temporal preference.}
  Cumulative error count is decodable at 95.3\% ($p < 0.001$) with a plateau spanning layers 19--31, the same region identified by attribution patching.
  Error is decodable at the turn-transition tokens where temporal preference geometry crystallizes (\ref{app:parametric-geometry}).
  This suggests the subgraph functions as a general meta-cognitive region, not a temporal-specific circuit.

  \item \textbf{Chain-level error is far more decodable than local error identity.}
  The 21pp gap (95\% cumulative vs.\ 74\% corruption status) mirrors the main paper's finding that global properties (time horizon) are more structured than local behavioral outputs.

  \item \textbf{Error and temporal preference occupy orthogonal linear directions.}
  No cosine between probe weight vectors reaches $p < 0.05$ at any layer.
  The two variables do not compete for the same linear subspace within the subgraph.

  \item \textbf{Asymmetric non-linear overlap exists.}
  The temporal probe transfers to the error task above chance at layers 10--25 ($p < 0.001$), but the error probe does not transfer to temporal preference.
  The two variables share curved manifold structure but not a linear direction, consistent with the non-linear geometry in~\ref{app:parametric-geometry}.

  \item \textbf{The gap between representation and behavior generalizes.}
  Error state is internally encoded at 95\% accuracy but barely affects output confidence, paralleling the temporal preference gap between representation and behavior documented in Appendices~\ref{app:behavioral-temporal-discount} and~\ref{app:behavioral-coherence}.
  The gap is architectural, not task-specific.

  \item \textbf{Two-axis steering is feasible.}
  The orthogonality of probe directions means a temporal preference steering vector (\ref{app:contrastive-steering}) should not perturb error sensitivity.
  An error-awareness vector at layers 24--25 could complement temporal steering, enabling two-axis control with minimal cross-interference.

  \item \textbf{Connection to the steering results.}
  The probing--steering dissociation observed in~\ref{app:contrastive-steering} (best probing at L26 vs.\ best steering at L19--22) may extend to error: the layers where error is most decodable (L24--25) need not be the layers where error-state interventions are most effective.
  Testing this prediction via error-state CAA is a natural next step.
\end{enumerate}

\paragraph{Limitations.}
These results establish correlational decodability, not causal necessity.
Activation patching of error-state representations (analogous to Appendices~\ref{app:causal-parametric} and~\ref{app:causal-contrastive}) would be needed to confirm that the identified representations causally drive downstream behavior.
Our error injection uses synthetic perturbations in math reasoning chains, which may not fully reflect the distribution of errors arising during unconstrained generation.
Generalization to other reasoning domains and to models beyond \texttt{Qwen3-4B-Instruct-2507} remains to be tested.

\clearpage
\clearappnumbering

\partpagecontent{Part 3:}{Could we control temporal preference?}{%
\begin{itemize}[leftmargin=*, itemsep=0.8em]
  \item \textbf{\hyperref[app:contrastive-steering]{S}.} Contrastive CAA steering
\end{itemize}%
}

\section{Contrastive steering results}\label{app:contrastive-steering}

Parts 1 and 2 established where temporal preference lives (layers 17--35; \ref{app:convergence}) and what it looks like (an ordinal horizon that transforms into a binary preference at the turn boundary; \ref{app:parametric-geometry}).
The behavioral analysis showed that the resulting preferences are unstable and inconsistent (\ref{app:behavioral-temporal-discount}, \ref{app:behavioral-coherence}).
Here we ask the intervention question: can we \emph{control} temporal preference by directly modifying the representations we identified?

We construct a CAA steering vector from the probe direction at layer 26 (\ref{app:contrastive-probing-linear}) and inject it at candidate layers during inference (methodology in \ref{app:contrastive-steering-methods}).

\subsection{Forced-Choice Behavioral Sweep}
\label{app:contrastive-steering:sweep}

\subsubsection{\texorpdfstring{Layer $\times$ Alpha Sweep}{Layer x Alpha Sweep}}

The probe's best layer (26) is not necessarily the best steering layer.
We swept 9
layers (19--27) $\times$ 5 alpha values (1, 2, 5, 10, 20) = 45 configurations.

\begin{figure}[!htbp]
\centering
\begin{minipage}[b]{0.45\textwidth}
\centering
\small
\setlength{\tabcolsep}{3pt}
\adjustbox{max width=\linewidth}{%
\begin{tabular}{c|ccccccccc}
\toprule
$\alpha$ & L19 & L20 & L21 & L22 & L23 & L24 & L25 & L26 & L27 \\
\midrule
1  & 0.20 & 0.20 & 0.20 & 0.20 & 0.19 & 0.19 & 0.19 & 0.18 & 0.18 \\
2  & 0.23 & 0.23 & 0.23 & 0.22 & 0.21 & 0.21 & 0.20 & 0.20 & 0.19 \\
5  & 0.32 & 0.31 & 0.32 & 0.30 & 0.27 & 0.26 & 0.24 & 0.23 & 0.22 \\
10 & 0.46 & 0.45 & 0.46 & 0.42 & 0.36 & 0.34 & 0.31 & 0.29 & 0.26 \\
20 & 0.72 & 0.69 & 0.72 & 0.68 & 0.56 & 0.51 & 0.46 & 0.40 & 0.35 \\
\bottomrule
\end{tabular}}
\captionof{table}{Forced-choice score $S(\alpha, l)$ across layers and steering coefficients.
Baseline (no steering): $S = 0.1724$.
Layers 19--22 form the behavioral sweet spot,
with a sharp drop at layer~23.}
\label{tab:layer_sweep}
\end{minipage}%
\hfill
\begin{minipage}[b]{0.5\textwidth}
\centering
\includegraphics[width=\textwidth]{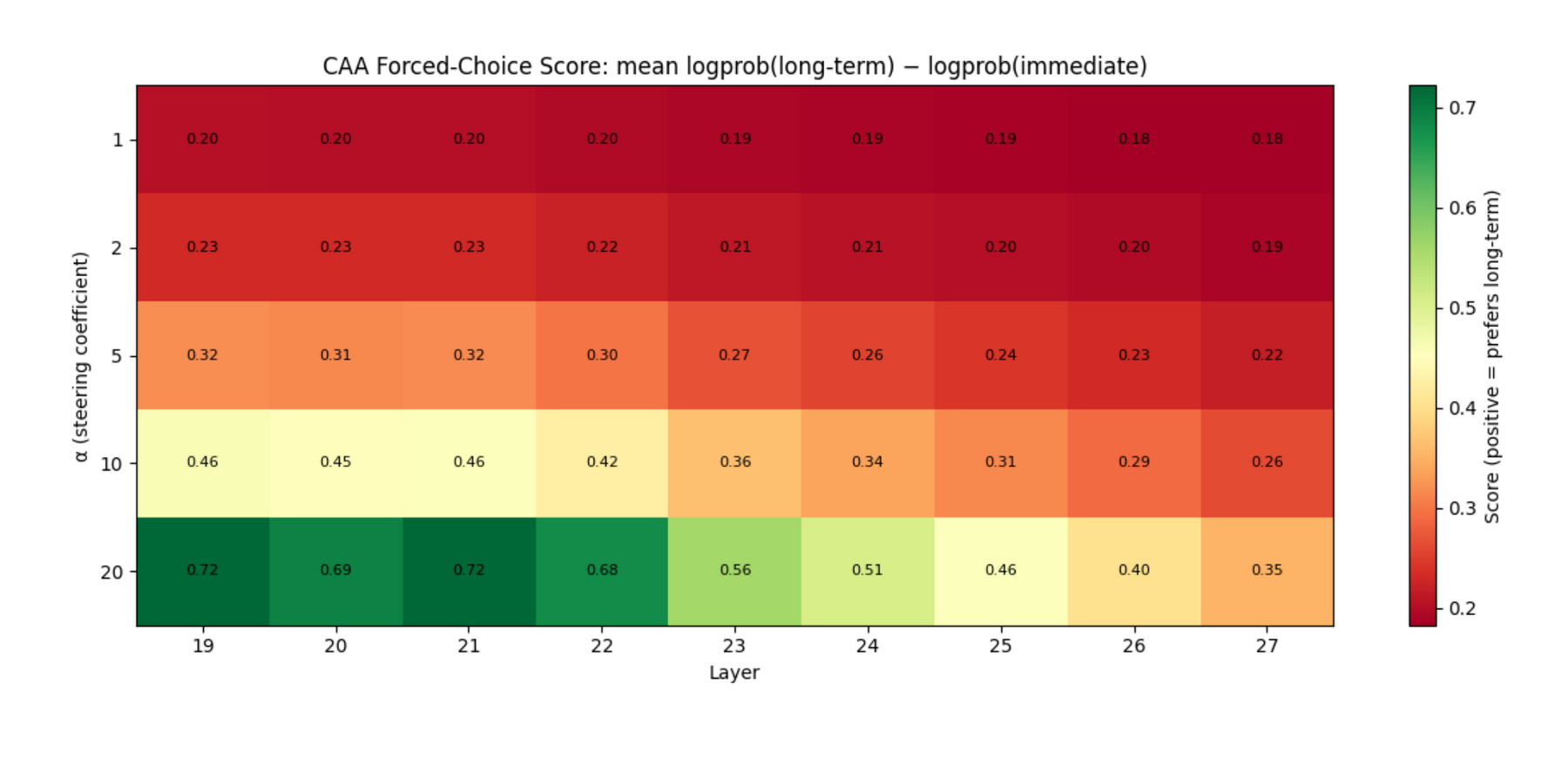}
\caption{Heatmap of forced-choice score $S(\alpha, l)$ across layers (19--27) and
steering coefficients ($\alpha = 1$--$20$).
Layers 19--22 form the behavioral sweet
spot, with effectiveness dropping sharply at layer~23.}
\label{fig:layer_alpha_sweep}
\end{minipage}
\end{figure}

\paragraph{Probing--steering dissociation.}
Layer~26 is optimal for \emph{reading} temporal orientation (99.2\% probe accuracy)
but not for \emph{writing} it.
Layers 19--22 are the effective steering layers, 4--7
layers earlier than the best probe layer.
This dissociation is consistent with a
functional asymmetry: upper layers consolidate a high-fidelity \emph{readout} of the
temporal concept, while mid-network layers are where causal interventions most
effectively redirect the model's downstream computation.
Similar
probing-vs-intervention gaps have been observed in other
domains~\citep{heimersheim2024useinterpretactivationpatching}.

\FloatBarrier
\subsubsection{Extended Alpha Sweep}

Following the initial sweep, we extended the alpha range for the most promising layers
(19--25) with $\alpha \in \{20, 30, 40, 50\}$.

\begin{figure}[!htbp]
\centering
\begin{minipage}[b]{0.45\textwidth}
\centering
\small
\setlength{\tabcolsep}{3pt}
\adjustbox{max width=\linewidth}{%
\begin{tabular}{c|ccccccc}
\toprule
$\alpha$ & L19 & L20 & L21 & L22 & L23 & L24 & L25 \\
\midrule
20 & 0.72 & 0.69 & 0.72 & 0.68 & 0.56 & 0.51 & 0.46 \\
30 & 0.94 & 0.91 & 0.97 & 0.93 & 0.76 & 0.68 & 0.60 \\
40 & 1.13 & 1.10 & 1.20 & 1.17 & 0.96 & 0.84 & 0.74 \\
50 & 1.30 & 1.27 & 1.39 & 1.39 & 1.14 & 1.01 & 0.87 \\
\bottomrule
\end{tabular}}
\captionof{table}{Extended alpha sweep.
Best configuration: layer~22 with $\alpha\!=\!50$,
achieving $S = 1.3944$ (baseline: 0.1724, lift: +1.22).
This corresponds to
approximately $3.4\times$ higher relative odds for the long-term over the short-term completion under the forced-choice metric ($\exp(1.22) \approx 3.39$).
Because $S$ is a mean log-probability difference, this is an odds-ratio change rather than an absolute multiplier on $P(\text{long})$.}
\label{tab:extended_sweep}
\end{minipage}%
\hfill
\begin{minipage}[b]{0.5\textwidth}
\centering
\includegraphics[width=\textwidth]{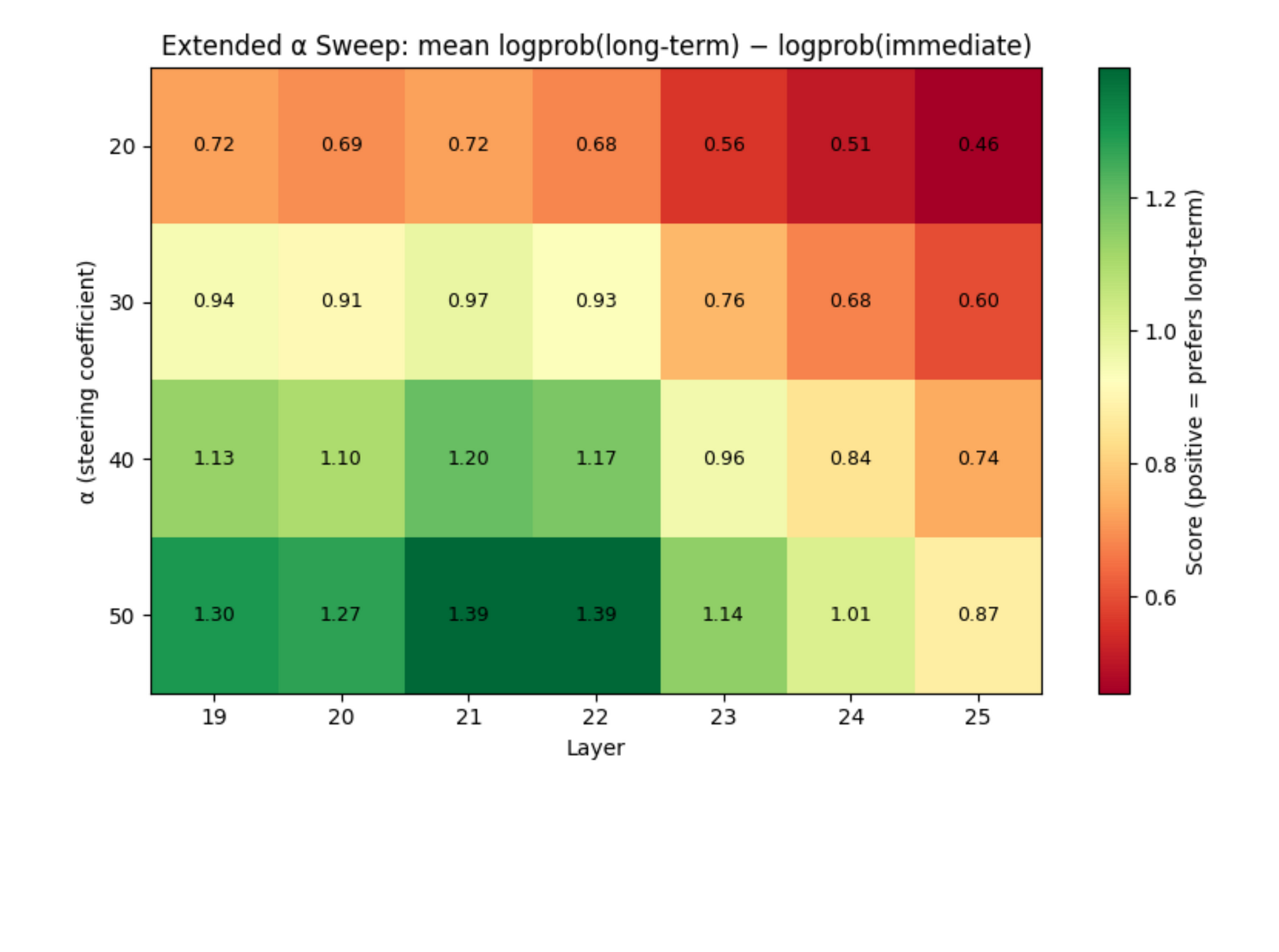}
\caption{Extended alpha sweep heatmap for layers 19--25 with $\alpha \in \{20, 30, 40, 50\}$.
Score increases monotonically with $\alpha$; best configuration is layer~22 at $\alpha\!=\!50$.}
\label{fig:extended_alpha_sweep}
\end{minipage}
\end{figure}

\noindent The score increases monotonically with $\alpha$ across all layers, with
layers 19--22 consistently outperforming later layers.
The optimal configuration
(layer~22, $\alpha = 50$) achieves a score of 1.3944, representing a lift of +1.22
over the unsteered baseline of 0.1724.

\FloatBarrier
\subsection{Open-Ended Generation Evaluation}
\label{app:contrastive-steering:openended}

The forced-choice metric measures whether the model's token probabilities shift in
the correct direction.
To verify that this translates into qualitative behavioral
change, we evaluate steering on 13 open-ended neutral prompts (e.g., ``\textit{You are
advising a team on how to handle a major organizational challenge. What should be the
main focus?}'').

\subsubsection{Experimental Setup}

For each configuration, we generate text with \texttt{do\_sample=False} and
\texttt{max\_new\_tokens=90}.
We test layers $\{19, 20, 21, 22, 23, 26\}$ across
$\alpha \in \{25, 40, 50\}$, applying the steering vector in both directions: positive
$\alpha$ (toward long-term) and negative $\alpha$ (toward short-term).

All generated responses were scored by \texttt{Claude Sonnet 4.6} on a $[-10, +10]$ temporal
orientation scale, where $-10$ denotes clearly short-term thinking and $+10$ denotes
clearly long-term thinking.
We note that this LLM-as-judge evaluation is not validated against human ratings; the scores should be interpreted as a proxy for directional shift rather than a calibrated measure of temporal orientation.

\subsubsection{Results}

\begin{table}[!htbp]
\centering
\small
\setlength{\tabcolsep}{5pt}
\begin{tabular}{lrrr}
\toprule
\textbf{Configuration} & $\alpha = 25$ & $\alpha = 40$ & $\alpha = 50$ \\
\midrule
L19 $\rightarrow$ long-term  & +1.1 & +1.2 & +2.3 \\
L19 $\rightarrow$ short-term & $-$1.7 & $-$4.7 & $-$5.7 \\
\midrule
L20 $\rightarrow$ long-term  & +1.8 & +2.0 & +2.4 \\
L20 $\rightarrow$ short-term & $-$1.9 & $-$3.6 & $-$3.4 \\
\midrule
L21 $\rightarrow$ long-term  & +1.8 & +2.2 & +2.7 \\
L21 $\rightarrow$ short-term & $-$2.0 & $-$3.5 & $-$2.9 \\
\midrule
L22 $\rightarrow$ long-term  & +1.2 & +2.4 & +2.2 \\
L22 $\rightarrow$ short-term & $-$1.2 & $-$2.8 & $-$4.0 \\
\midrule
L23 $\rightarrow$ long-term  & +1.2 & +1.6 & +1.7 \\
L23 $\rightarrow$ short-term & $-$0.8 & $-$1.0 & $-$2.4 \\
\midrule
L26 $\rightarrow$ long-term  & +1.2 & +1.0 & +1.0 \\
L26 $\rightarrow$ short-term & $-$0.5 & $-$1.0 & $-$0.1 \\
\bottomrule
\end{tabular}
\caption{Mean shift from baseline on the $[-10, +10]$ temporal orientation scale for
open-ended generation.
Positive values indicate a shift toward long-term framing;
negative values indicate a shift toward short-term framing.
Each entry averages over
13 prompts.}
\label{tab:openended_shift}
\end{table}

\begin{figure}[!htbp]
\centering
\includegraphics[width=0.7\textwidth]{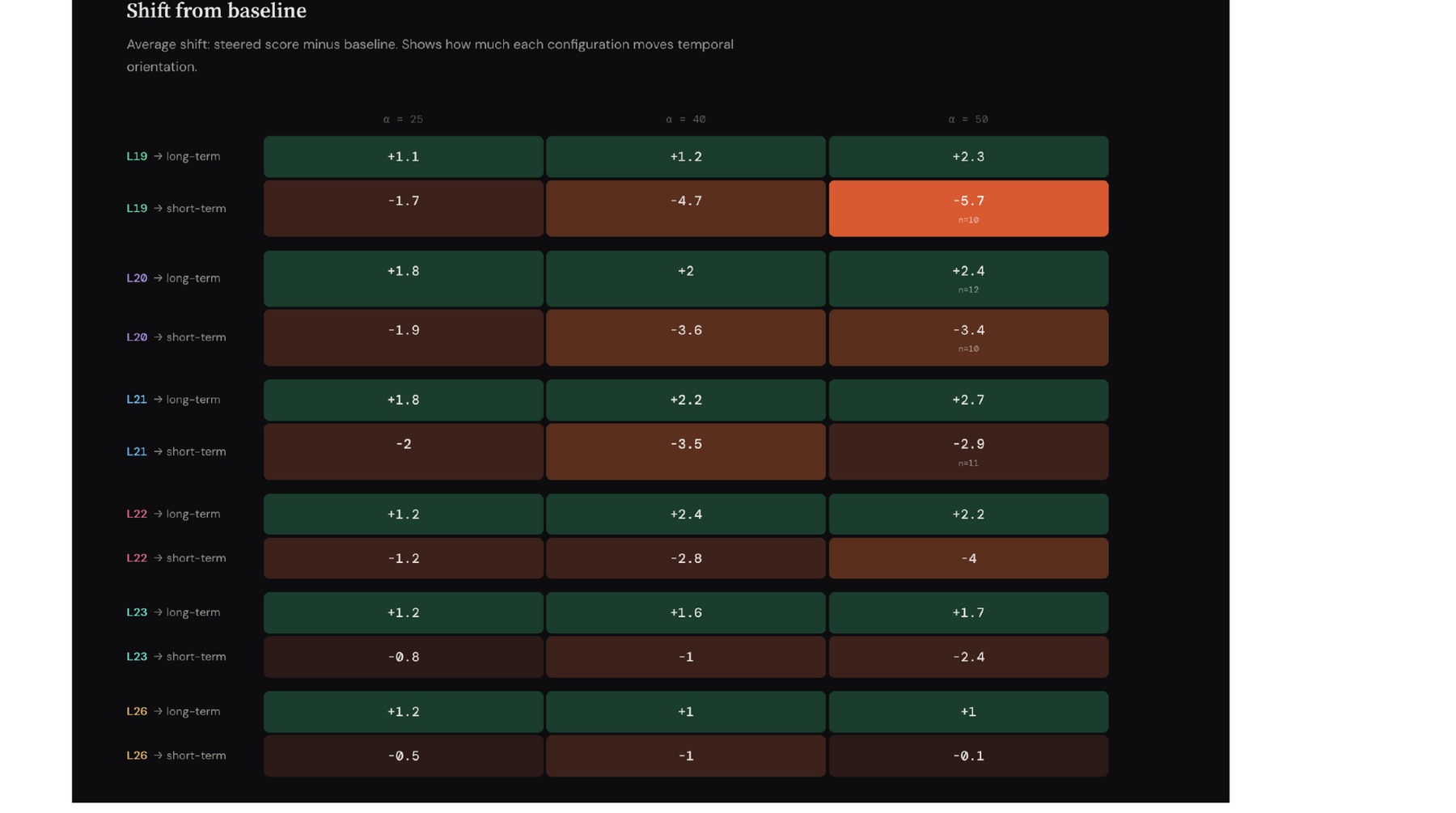}
\caption{Shift from baseline on the $[-10, +10]$ temporal orientation scale for
open-ended generation across layers and $\alpha$ values.
Both long-term (positive)
and short-term (negative) steering produce consistent directional shifts.}
\label{fig:openended_shift}
\end{figure}

\FloatBarrier
\paragraph{Key observations.}
\begin{enumerate}[leftmargin=1.5em]
    \item \textbf{Bidirectional steering.} Both positive and negative $\alpha$ produce
    consistent shifts in the expected direction, confirming that the CAA vector captures a
    genuine temporal orientation axis.

    \item \textbf{Qualitative framing shifts.} At $\alpha \in [40, 50]$, long-term
    steered outputs adopt strategic framing (e.g., ``resilient,'' ``future-ready,''
    ``systemic redesign''), while short-term steered outputs adopt triage framing
    (e.g., ``remain calm,'' ``prioritize urgency,'' ``structured immediate steps'').

    \item \textbf{Coherence boundary.} Pilot runs at $|\alpha| = 60$ showed output
    incoherence, so we capped the sweep at $|\alpha| = 50$; beyond that the intervention
    appears to push the model too far from its natural distribution.

    \item \textbf{Asymmetry.} The long-term direction produces cleaner shifts than
    the short-term direction at the same $|\alpha|$, suggesting that the model's
    latent distribution may be slightly closer to the long-term end of the temporal
    axis.

    \item \textbf{Layer effects mirror forced-choice results.} Layers 19--22 produce
    the strongest open-ended shifts, and layer~26 produces only weak effects despite
    having the highest probe accuracy.
    This reinforces the probing--steering
    dissociation documented in Section~\ref{app:contrastive-steering:sweep}.
\end{enumerate}

\subsubsection{Qualitative Example}

We illustrate the steering effect with a representative example.

\vspace{0.5em}
\noindent\textbf{Prompt:} \textit{You are advising a team on how to handle a major
organizational challenge. What should be the main focus?}
\vspace{0.3em}

\noindent\textbf{Baseline} (score: 2): The response focuses on present-tense
communication and immediate trust-building without strong temporal language in either
direction.
The main focus is on clear communication, transparency, and employee
engagement.

\vspace{0.3em}
\noindent\textbf{Long-term steered} ($\alpha = +50$, score: 6): The response centers
on long-term success through resilience and shared vision, framing organizational
challenges in terms of sustained adaptive capacity.
The main focus is on resilience
through shared purpose, adaptive thinking, and inclusive leadership.

\vspace{0.3em}
\noindent\textbf{Short-term steered} ($\alpha = -50$, score: $-$3): The response leads
with immediate and clear communication and emphasizes knowing what the deadline is,
orienting team management around near-term operational urgency.

\FloatBarrier
\subsection{Discussion}
\label{app:contrastive-steering:discussion}

\paragraph{\texorpdfstring{Probing $\neq$ steering.}{Probing != steering.}}
The central methodological finding is the dissociation between the optimal probing
layer (26) and the optimal steering layers (19--22).
This dissociation has implications
for the broader interpretability literature: high probe accuracy at a layer does not
imply that the same layer is the appropriate target for causal intervention.
The
readout of a concept and the point at which that concept can be effectively modified
may be separated by several layers, reflecting distinct functional roles in the
transformer's computation~\citep{turner2023activation}.

\paragraph{Implicit vectors generalize.}
Using the implicit dataset (which contains no surface temporal vocabulary) to
construct the CAA vector ensures that the steering direction captures semantic temporal
reasoning rather than lexical artifacts.
The cross-dataset probe generalization
(Section~\ref{app:contrastive-probing-linear:cross_dataset}) confirms that the implicit direction aligns
with the explicit temporal axis, and the forced-choice evaluation on explicit prompts
(Section~\ref{app:contrastive-steering:sweep}) demonstrates that this vector effectively steers
behavior on prompts with overt temporal cues.

\paragraph{Connection to subgraph localization.}
The behavioral sweet spot at layers 19--22 aligns with the mid-network components
identified by the EAP-IG attribution analysis (\ref{app:attributional-contrastive}) and the activation patching
experiments (\ref{app:causal-parametric}).
This convergence across independent methodologies (probing,
CAA steering, attribution patching, and activation patching) provides strong evidence
that the temporal preference mechanism is localized to a consistent set of mid-to-upper
layers.

\paragraph{Relation to the representational geometry.}
The PCA analysis (Section~\ref{app:contrastive-probing-linear:pca}) shows that the temporal direction in
the implicit dataset is not captured by the top principal components.
This is consistent
with the non-linear manifold structure reported in \ref{app:parametric-geometry}, where time horizon is
encoded in a curved subspace.
The CAA vector, derived from the linear probe direction,
provides a first-order approximation to steering along this manifold.
The monotonic
increase in steering score with $\alpha$ (Table~\ref{tab:extended_sweep}) suggests
that this linear approximation remains effective within the tested range, though the
output-quality degradation at $|\alpha| = 60$ may indicate the intervention exceeding the
locally linear regime.

\paragraph{LLM-as-Judge Evaluation Criteria.}
To quantify the qualitative shifts in our open-ended generation experiments, we used the \texttt{Claude Sonnet 4.6} API as an independent evaluator.
Each generated response was individually processed by the API and assigned a score on a $[-10, +10]$ scale, where $-10$ represents an extreme short-term focus and $+10$ represents an extreme long-term focus.
The model was prompted to evaluate responses by strictly adhering to predefined grading criteria.
Specifically, the evaluator analyzed the text for the presence and frequency of explicit temporal keywords (e.g., immediate triage versus systemic redesign) and weighed semantic details, structural planning, and thematic biases that explicitly skewed the generation toward a specific temporal horizon.
\clearpage
\clearappnumbering

\partpagecontent{Part 4:}{Extended methodologies}{%
\begin{itemize}[leftmargin=*, itemsep=0.8em]
  \item \textbf{\hyperref[app:notation]{T}.} Notation
  \item \textbf{\hyperref[app:contrastive-probing-linear-methods]{U}.} Contrastive probing methods
  \item \textbf{\hyperref[app:attributional-contrastive-methods]{V}.} Attributional contrastive methods
  \item \textbf{\hyperref[app:causal-parametric-methods]{W}.} Causal parametric methods
  \item \textbf{\hyperref[app:causal-contrastive-methods]{X}.} Causal contrastive methods
  \item \textbf{\hyperref[app:parametric-geometry-methods]{Y}.} Parametric geometry methods
  \item \textbf{\hyperref[app:behavioral-temporal-discount-methods]{Z}.} Behavioral discounting methods
  \item \textbf{\hyperref[app:behavioral-coherence-methods]{AA}.} Behavioral coherence methods
  \item \textbf{\hyperref[app:contrastive-steering-methods]{AB}.} Contrastive steering methods
  \item \textbf{\hyperref[app:case-study-hf]{AC}.} Worked case study: highly-formatted pair
\end{itemize}%
}

\section{Notation and key concepts}\label{app:notation}

The following terms and abbreviations are used throughout the appendices.

\begin{center}
\small
\renewcommand{\arraystretch}{1.3}
\begin{tabularx}{\linewidth}{@{} >{\bfseries}l X @{}}
\toprule
Term & Definition \\
\midrule
Subgraph             & Model components (attention heads, MLP neurons) whose ablation or patching shifts temporal preference. \\
Temporal preference   & The model's tendency to favor short-term vs.\ long-term options in a forced-choice setting. \\
Time horizon          & An explicit temporal constraint (e.g., ``1 year'') given in the prompt; ranges from seconds to centuries. \\
On- vs.\ off-policy  & \emph{On-policy}: activations from the model's own generation. \emph{Off-policy}: activations read from a forced context (user turn). \\
Contrastive pair      & Matched clean/corrupted prompts that differ in temporal framing; used for both EAP-IG and activation patching. \\
EAP-IG               & Edge Attribution Patching with Integrated Gradients~\citep{hanna2024faithfaithfulnessgoingcircuit}; gradient-based attribution that approximates causal patching. \\
Activation patching  & Replacing a component's activations with counterfactual values to measure causal effect on a downstream metric. \\
Recovery / Disruption & Normalized $[0,1]$ metrics for denoising and noising patching, respectively; 0 = no effect, 1 = full effect. \\
Probing layer        & Residual-stream layer at which a linear classifier best separates short- vs.\ long-term orientation (layer~26 in this work). \\
Steering layer       & Layer at which a CAA vector~\citep{turner2023activation} most reliably shifts behavior (layers~19--22 in this work). \\
CAA                  & Contrastive Activation Addition: mean activation difference between long-term and short-term choices, used as a steering vector. \\
Decision boundary    & Binary search over delayed-reward magnitudes to locate per-item indifference, used to fit hyperbolic discount rate $k$. \\
MCQ-27               & Kirby Monetary Choice Questionnaire~\citep{kirby1999}: 27-item instrument for estimating temporal discount rates. \\
\bottomrule
\end{tabularx}
\end{center}

\clearpage
\clearappnumbering

\section{Contrastive linear probing methodology}\label{app:contrastive-probing-linear-methods}

We train logistic regression probes~\citep{mueller2025mibmechanisticinterpretabilitybenchmark, kim2025linearrepresentationspoliticalperspective} on
residual-stream activations to determine \emph{where} the model linearly encodes the
distinction between short-term and long-term temporal orientation.
Probing results are presented in \ref{app:contrastive-probing-linear}.

\subsection{Activation Extraction}
\label{app:contrastive-probing-linear:extraction}

For each prompt, we concatenate the question and the choice text, apply the \texttt{Qwen3}
chat template, and extract residual-stream activations at every layer.
Because the chat
template wraps the user turn as

\begin{center}
\texttt{<|im\_start|>user\textbackslash n\{question + choice\}<|im\_end|>\textbackslash n<|im\_start|>assistant\textbackslash n}
\end{center}

\noindent the token at position $-1$ is the trailing newline after
\texttt{assistant}, a fixed token that is identical across all prompts and carries only
whatever signal attention has propagated into that generic position.
We instead locate
the last \texttt{<|im\_end|>} token in the sequence (which closes the user turn) and
extract at position $\texttt{im\_end} - 1$, corresponding to the final token of the
actual choice text.
This position directly encodes the semantic content of the choice.

\paragraph{Impact of the token-position correction.}
The correction produced a qualitative change in both probe accuracy and downstream
steering vector quality:

\begin{table}[h]
\centering
\small
\begin{tabular}{lcc}
\toprule
\textbf{Metric} & \textbf{Before (trailing \textbackslash n)} & \textbf{After (im\_end $-$ 1)} \\
\midrule
Extraction token & \texttt{\textbackslash n} (trailing newline) & Last choice token \\
CAA vector $\ell_2$ norm & 2.62 & 30.30 \\
Probe accuracy (best layer) & $\sim$93\% & 99.2\% \\
\bottomrule
\end{tabular}
\caption{Effect of correcting the extraction token position.
The previous vector was essentially normalized noise; the corrected extraction yields a $\sim$10$\times$ stronger CAA vector.}
\label{tab:token_position_fix}
\end{table}

\subsection{Probe Training Protocol}
\label{app:contrastive-probing-linear:probe_protocol}

We train one \texttt{LogisticRegression}$(C\!=\!0.1)$ probe per layer on
$D_{\mathrm{implicit}}$ with the following methodological controls:

\begin{enumerate}[leftmargin=1.5em]
    \item \textbf{Pair-level train/test split.} The split operates on pair indices
    rather than individual rows.
    Both the immediate and long-term activations from
    a given pair always land in the same fold.
    Without this, the probe can exploit
    shared question text as a shortcut, and the test set is not truly held out.
    We use an 80/20 split with a fixed random seed.

    \item \textbf{StandardScaler normalization.} The residual stream has 2{,}560
    dimensions with varying variances.
    \texttt{LogisticRegression} with $\ell_2$
    regularization penalizes large weights uniformly, so high-variance dimensions
    dominate the penalty without scaling.
    We fit a \texttt{StandardScaler} on the
    training fold and apply it to both train and test.
    Critically, the scaler is
    persisted to disk alongside each probe and re-applied during cross-dataset
    evaluation and when extracting the probe's coefficient vector for use as the
    CAA steering direction.
\end{enumerate}

\noindent Activations are stored as tensors of shape
$[n_{\mathrm{prompts}}, n_{\mathrm{layers}}, d_{\mathrm{model}}]$ =
$[600, 36, 2560]$ for the implicit dataset.

\clearpage
\clearappnumbering

\definecolor{optA}{HTML}{C44E52}
\definecolor{optB}{HTML}{4C72B0}
\newcommand{\OptionA}{\textcolor{optA}{\textbf{[Option A]}}}
\newcommand{\OptionB}{\textcolor{optB}{\textbf{[Option B]}}}

\section{Attributional contrastive methodology}\label{app:attributional-contrastive-methods}

We identify this subgraph in two stages: first, we restrict the candidate node set using Edge Attribution Patching with Integrated Gradients (EAP-IG); second, we score and prune edges between these nodes to recover a sparse functional subgraph.
Unlike prior approaches that attribute to logit differences, we compute attribution with respect to individual option logits, yielding concept-specific attribution scores that disentangle the contributions of nodes to competing temporal evaluations.

\subsection{Edge Attribution Patching-Integrated Gradients}

To localize the internal computations associated with temporal preference, we use Edge Attribution Patching with Integrated Gradients (EAP-IG).
EAP-IG operates on matched clean and corrupted prompts and assigns attribution scores to internal components based on their contribution to the model's preference for one response token over another.
It can be viewed as a computationally efficient approximation to activation patching.

EAP-IG can be implemented in two ways: by interpolating activations at each node, or by interpolating only the input embeddings.
The latter is significantly more efficient and provides a practical method for estimating edge importance.
However, as this approach compounds two approximations, we do not use it to precisely rank components; instead, we use it to restrict the search space by filtering out nodes with low attribution scores.

Concretely, we interpolate between corrupted and clean inputs in embedding space and integrate gradients along the resulting path, rather than relying on a single local gradient estimate.
This yields attribution scores $s^A(x,i,t)$ for each component $i$ at token position $t$ for metric $A$ on prompt $x$.

\subsection{Notation}

In the paper, the attribution score for a variant $v$ is denoted as $s^A(x,i,t)$.

\begin{equation} \label{eapig}
s^A(x,i,t) =
(z^{}_{i,t} - z'_{i,t})\int_{\alpha=0}^{1}\frac{\partial L_A(x' + \alpha(x - x'))}{\partial z_{i,t}}
\approx
(z^{}_{i,t} - z'_{i,t})\frac{1}{m}\sum_{k=1}^{m}\frac{\partial L_A(x' + \frac{k}{m}(x - x'))}{\partial z_{i,t}}
\end{equation}

\subsubsection{Metric Normalized Attribution Scores}
The attribution scores are scaled by $\Delta L_A=L_A(z) - L_A(z')$ so they can be aggregated across datasets and semantically equivalent metric functions.
In this paper, $\bar s^A(x,i,t)$ denotes the scaled attribution scores.
When $L_A(z)\approx L_A(z')$, it indicates that the model either does not distinguish between the clean and corrupted cases or has nearly the same preference for both options; such cases on average constitute $\sim 2\%$ of the total dataset and are dropped from further analysis.

\subsection{Variations for Bias Control}

\subsubsection{Positional Bias Control}
We control for positional bias by evaluating each question--answer pair under both possible option orderings.
In one condition, the short-horizon response precedes the long-horizon response; in the other, the order is reversed.
This counterbalancing prevents temporal preference from being confounded with a general tendency to favor a particular position (e.g., the first option) or a fixed association between labels and positions.

In the input construction pipeline, this is implemented by generating two matched prompt sets from the same underlying examples: a canonical ordering (question, short-horizon option, long-horizon option) and a mirrored ordering (question, long-horizon option, short-horizon option).
The experiment loop evaluates both orderings as separate conditions (\texttt{short\_first} and \texttt{long\_first}) under otherwise identical settings.
Consequently, any temporal-scope effect that is consistent across both conditions is unlikely to be driven by positional bias alone.

\subsubsection{Lexical Bias Control}

To mitigate the possibility that results are driven by the lexical identity of response labels rather than temporal content, we repeat all experiments under seven matched response-label schemes.
These schemes use uppercase letters (\texttt{(A)/(B)}), lowercase letters (\texttt{(a)/(b)}), Arabic numerals (\texttt{(1)/(2)}), Roman numerals (\texttt{(i)/(ii)}), number words (\texttt{(One)/(Two)}), alternative letters (\texttt{(X)/(Y)}), and a non-alphanumeric symbol pair (\texttt{(\ding{108})/(\ding{110})}).

Across these runs, the dataset, prompt template, model, batch size, inference settings, and scoring metric are held fixed.
Only the surface form of the response labels and the corresponding instruction specifying the target output token are varied.
This isolates lexical biases associated with particular label tokens, such as pretrained preferences for \texttt{A/B} or \texttt{1/2}.

If an effect persists across all label variants, it is unlikely to be attributable to any specific output token and instead reflects the model's sensitivity to the underlying short- versus long-horizon distinction.
We denote each label variant as $D^v$.

\subsection{Experimental Setup}

\subsubsection{System Prompt}
The system prompt is designed to constrain the model's output format and suppress the inclusion of explicit reasoning in its responses.

To ensure that the phrasing of the system prompt does not confound component attribution scores, all token positions corresponding to the system prompt are excluded during position-wise aggregation.
\footnote{In the \texttt{Qwen3} chat template, the end of a prompt is marked by the \texttt{<|im\_end|>} token.
Accordingly, $n_{\text{sys}}$ is defined as the token length of the system prompt plus one.}

\begin{equation} \label{position_aggregation}
\bar{s}^A_t(x, i) = \frac{1}{(n_{\text{total}} - n_{\text{sys}})}
\sum_{t = n_{\text{sys}}}^{n_{\text{total}}} \bar{s}^A(x, i, t)
\end{equation}

\subsubsection{Prompt Syntax}

\begin{figure}[htbp]
    \centering
    \includegraphics[width=\linewidth]{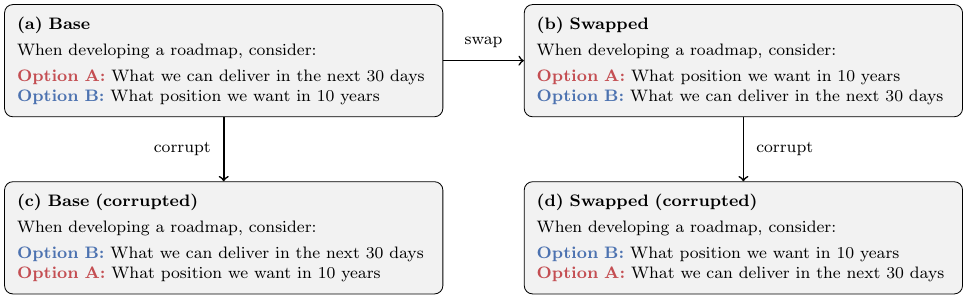}
    \caption{
    An example of base and swapped prompts, along with their corrupted counterparts used for EAP-IG attribution.
    Corruption corresponds to flipping option semantics while preserving surface form.
    }
    \label{fig:prompt_syntax}
\end{figure}

Each prompt draws from the temporal-scope datasets described in \ref{app:prompts} and presents a scenario along with two plausible courses of action: one emphasizing short-term rewards and the other emphasizing long-term rewards.
The system prompt instructs the model to respond by selecting the label (e.g., \texttt{A} or \texttt{B}) corresponding to its preferred option.

To construct a corrupted variant, the option labels are swapped while preserving the textual order of the candidate responses (see Figure~\ref{fig:prompt_syntax}).
This manipulation isolates the effect of label assignment from the semantic content of the options.

To control for positional bias, we additionally evaluate the prompt under a flipped ordering of the options.
Let $s^A_t(x,i)$ denote the attribution score at token position $t$ for component $i$ under the canonical ordering.
Let $s^{A*}_t(x,i)$ denote the corresponding attribution score under the flipped ordering.

\subsubsection{Metric Function}
We use raw logit values as the attribution metric rather than logit differences, as they provide a more fine-grained characterization of component behavior.
In particular, raw logits allow us to distinguish between components that actively promote a target concept and those that exert inhibitory effects.
This formulation also enables attribution with respect to semantically meaningful concepts (e.g., long-term vs.\ short-term orientation), rather than relative preferences alone.

Let $\mathrm{LT}$ and $\mathrm{ST}$ denote the long-term and short-term concepts, respectively.
We define concept-aligned attribution scores by symmetrizing over label assignments and option orderings:

\begin{equation}
\bar s_t^{\mathrm{ST}}(x, i)
= \frac{1}{2}\left( \bar s_t^A(x, i) + \bar s_t^{B*}(x, i) \right),
\quad
\bar s_t^{\mathrm{LT}}(x, i)
= \frac{1}{2}\left( \bar s_t^B(x, i) + \bar s_t^{A*}(x, i) \right).
\end{equation}

\subsection{Component Attribution Calculation}

We compute component-level attribution scores for each time-horizon concept $c \in \{\mathrm{LT}, \mathrm{ST}\}$ via a two-stage aggregation procedure over examples and prompt variants.

\paragraph{Within-variant aggregation.}
For each variant $v \in \mathcal{V}$, we estimate the expected attribution score by averaging over a finite sample of inputs $D^v = \{x_1, \dots, x_{N_v}\}$:
\begin{equation}
  \bar s^c_t(D^v, i)
  = \frac{1}{N_v} \sum_{n=1}^{N_v} \bar s^c_t(x_n, i).
\end{equation}
This estimator is well-defined because attribution scores are normalized by the logit difference $\Delta L$, making them comparable across inputs.

\paragraph{Across-variant aggregation.}
We then aggregate across a finite set of variants $\mathcal{V}$ using a uniform weighting:
\begin{equation}
  \bar s^c_t(D, i)
  = \frac{1}{|\mathcal{V}|} \sum_{v \in \mathcal{V}} \bar s^c_t(D^v, i).
\end{equation}

\paragraph{Assumptions.}
This procedure assumes that (i) examples within each dataset variant are independent and identically distributed samples from an underlying distribution, and (ii) variants are treated as equally informative perturbations, justifying uniform averaging across $\mathcal{V}$.
In practice, both expectations are approximated by finite-sample means as defined above.

\clearpage
\clearappnumbering

\section{Causal parametric methodology}\label{app:causal-parametric-methods}

Activation patching results are presented in \ref{app:causal-parametric}.
Here we describe the experimental setup.

\subsection{Overview}

The parametric pipeline uses highly-formatted prompts with explicit time horizons to perform activation patching~\citep{heimersheim2024useinterpretactivationpatching}.
Unlike the contrastive pipeline (\ref{app:attributional-contrastive-methods}), which uses gradient-based attribution as an efficient approximation, the parametric pipeline directly measures causal effect by replacing component activations with counterfactual values.

The pipeline operates on \emph{contrastive pairs}, matched clean and corrupted trajectories that differ in their temporal framing.
Prompt construction and parametric variation are described in \ref{app:highly-formatted}.
A three-stage evaluation proceeds from sanity check to layer sweep to position sweep, each with configurable stride sizes that enable efficient coarse-to-fine analysis.

\subsection{Activation Patching Protocol}

For each contrastive pair, we perform both \textbf{denoising} and \textbf{noising} interventions:

\paragraph{Denoising.} The model runs on the corrupted prompt while clean activations are injected at specified layers and positions.
This measures \emph{recovery}: how much the intervention restores clean behavior.

\paragraph{Noising.} The model runs on the clean prompt while corrupted activations are injected.
This measures \emph{disruption}: how much the intervention degrades clean behavior.

Both metrics are normalized to $[0, 1]$:
\begin{align}
  \text{Recovery} &= \frac{y_{\text{intervened}} - y_{\text{corrupted}}}{y_{\text{clean}} - y_{\text{corrupted}}}  \label{eq:causal-param-rec-eq} \\[4pt]
  \text{Disruption} &= \frac{y_{\text{clean}} - y_{\text{intervened}}}{y_{\text{clean}} - y_{\text{corrupted}}}  \label{eq:causal-param-disr-eq}
\end{align}
where $y$ denotes the model's logit difference between the two options.
A value of 0 indicates no causal effect; 1 indicates full effect.

\subsection{Component Types}

We patch the following residual-stream components independently:
\begin{itemize}[nosep, leftmargin=*]
  \item \texttt{resid\_pre}: Residual stream before attention (input to the layer)
  \item \texttt{attn\_out}: Attention output
  \item \texttt{resid\_mid}: Residual stream after attention, before MLP
  \item \texttt{mlp\_out}: MLP output
  \item \texttt{resid\_post}: Residual stream after MLP (output of the layer)
\end{itemize}

For residual-stream components, patching is performed at the \emph{divergent position}, the last token before the model's choice, because residual propagation makes all-position patching uninformative.

\subsection{Position Mapping}

Clean and corrupted prompts often differ in token count because different time horizons or reward amounts require different numbers of tokens.
To patch activations at semantically corresponding positions, we use a \emph{piecewise linear interpolation} anchored on the structural markers of the highly-formatted template (\ref{app:highly-formatted}).

The algorithm identifies the token positions of each section marker (\texttt{SITUATION}, \texttt{TASK}, \texttt{OBJECTIVE}, \texttt{CONSTRAINT}, \texttt{ACTION}, \texttt{FORMAT}) as well as sub-markers for option labels, reward amounts, and time values in both the clean and corrupted sequences.
These anchors are sorted and augmented with sequence boundaries to form a set of corresponding position pairs.

Between consecutive anchors, positions are mapped via linear interpolation: for a source position $p$ in segment $[a_{\text{src}}, b_{\text{src}}]$ mapped to $[a_{\text{dst}}, b_{\text{dst}}]$, the corresponding destination position is
\begin{equation}
p' = a_{\text{dst}} + \frac{p - a_{\text{src}}}{b_{\text{src}} - a_{\text{src}}} \cdot (b_{\text{dst}} - a_{\text{dst}}),
\end{equation}
clamped to valid token indices.
This ensures that activations from each semantic region (e.g., the constraint field) are patched into the corresponding region of the other prompt, even when the two prompts have different total lengths.

\subsection{Sweep Protocol}

\paragraph{Layer sweep.} We patch each component across all 36 layers with a stride of 1, measuring recovery and disruption at each layer.
This identifies which layers carry the most causal effect for temporal preference.

\paragraph{Position sweep.} For the most causally important layers, we sweep across token positions with configurable strides (1, 5, or 10 tokens) to identify which token regions are most informative.
The position mapping described above ensures correct alignment when clean and corrupted prompts differ in length.

\clearpage
\clearappnumbering

\section{Causal classification methodology}\label{app:causal-contrastive-methods}

Results are presented in \ref{app:causal-contrastive}.
Here we describe the experimental setup.

\subsection{Motivation}

We construct an experiment to test whether the components flagged by the preference-targeted methods (\ref{app:contrastive-probing-linear}, \ref{app:attributional-contrastive}, \ref{app:attributional-parametric}, \ref{app:causal-parametric}) also engage in related temporal computations beyond preference valuation. We apply directional activation patching to IOI-style classification prompts in which the model must infer whether a goal's horizon is short-term or long-term: a cognitively distinct task that probes the same temporal axis. The protocol below describes the dataset, prompt structure, and patching pipeline.

\subsection{Dataset}

We use the 160-pair classification dataset described in \ref{app:classification-oriented}.
The four design principles (token alignment to 34 tokens, semantic overlap within pairs, no explicit temporal keywords, unambiguous horizons) and the 80\% (160/200) \texttt{Qwen3-4B-Instruct-2507} accuracy filter that produced 160 surviving pairs are documented there.

Each dataset sample contains two prompts: a clean prompt expecting \textit{short} answer (for a short-term goal) and a corrupted prompt expecting \textit{long} answer (for a long-term goal); the directional flips (defined in Sec~\ref{sec:cc-methods-patching-protocol}) invert the assignment.

\setlength{\myboxwidth}{\dimexpr\textwidth-2\fboxsep}

\textbf{Prompt template}\\\\
\fbox{\parbox{\myboxwidth}{"The goal is to <goal>. Is this a <short-term or long-term / long-term or short-term> goal? The answer is:"}}\\

Table~\ref{tab:cue_types_stat_160} shows dataset statistics for 160 surviving pairs after validation.
Table~\ref{tab:sl_ls_stat_160} shows the same statistics, but grouped by question order.

\begin{table}[h]
\centering
\begin{tabular}{llll}
\toprule
Variable & Career/Mastery & Growth & Accumulation \\
\midrule
Count & 74 (46\%) & 53 (33\%) & 33 (20\%) \\
Clean Q, LD (mean +/- st.d.) & 14.13 +/- 4.19 & 11.08 +/- 4.71 & 12.07 +/- 3.86 \\
Corrupted Q, LD (mean +/- st.d.) & -13.53 +/- 4.88 & -10.60 +/- 5.72 & -10.26 +/- 4.99 \\
\bottomrule
\end{tabular}
\vspace{0.5em}
\caption{Cue types statistics on successful Temporal Classification pairs}
\label{tab:cue_types_stat_160}
\end{table}

\begin{table}[h]
\centering
\begin{tabular}{lll}
\toprule
Variable & SL & LS \\
\midrule
Count & 80 & 80 \\
Clean Q, LD (mean +/- st.d.) & 11.76 +/- 4.47 & 13.63 +/- 4.35 \\
Corrupted Q, LD (mean +/- st.d.) & -13.61 +/- 5.26 & -10.16 +/- 4.97 \\
\bottomrule
\end{tabular}
\vspace{0.5em}
\caption{Question order statistics on successful Temporal Classification pairs}
\label{tab:sl_ls_stat_160}
\end{table}

\subsection{Patching Protocol}\label{sec:cc-methods-patching-protocol}

We perform denoising activation patching on the \texttt{resid\_pre}, \texttt{attn\_out}, and \texttt{mlp\_out} hooks at all token positions across all 36 layers of \texttt{Qwen3-4B-Instruct-2507} on 160 prompt pairs. 
We patch separately for short$\to$long and long$\to$short flips, testing whether the computation components match.
We use three metrics to quantify the effect (Sec~\ref{sec:cc-metrics-definitions}): logit difference (LD) and log-probability of clean and corrupted  answers: log-$P(\text{clean})$ and log-$P(\text{corr})$, respectively.
All of them are normalized so that $0$ corresponds to the corrupted baseline and $\pm 1$ to full recovery of the clean run's behavior.
By definition of presented metrics the two denoising rounds for different flips yield the same result as applying both the noising and denoising techniques on either one of flips.
As we cannot assume symmetry between \textit{short} and \textit{long} representations, we treat the flips separately and consider the noising or disruption of one flip as a recovery of the other, rather than as a reflection of the necessity and sufficiency of a single temporal classification circuit.

We refer to the flip in which the clean answer is \textit{short} and the corrupted answer is \textit{long} as the \emph{short-clean} flip, and the opposite as the \emph{long-clean} flip.

Throughout, we report a patching effect as significant if its 95\% confidence interval excludes zero. Confidence intervals are computed via a non-parametric pair-level percentile bootstrap with 10,000 resamples: for each iteration, we draw 160 pair indices with replacement, compute the mean per-layer effect across resampled pairs, and report the 2.5th and 97.5th percentiles of these bootstrap means as the 95\% CI bounds. We do not apply a multiple-comparisons correction across layers. Most effects forming the main headline claims of \ref{app:causal-contrastive} are large in magnitude and would survive standard family-wise corrections; other subordinate effects discussed there are closer to the per-layer threshold and should be read with appropriate caution.

We report two regimes of layer-level effects. First, per-layer patching at the END token position only (Sec.~\ref{sec:cc-resid-pre-section}, Sec.~\ref{sec:cc-attn-out-section}, \ref{sec:cc-mlp-out-section}). Second, for \texttt{attn\_out} and \texttt{mlp\_out}, a single 
intervention that patches the chosen layer's output at all 34 token positions simultaneously (Sec.~\ref{evidence:cc-alltokens-view-convergence}); this all-tokens view yields one per-layer score per (component, flip) 
combination and is reported specifically for cross-paradigm comparison.

\paragraph{Metric definitions.}\label{sec:cc-metrics-definitions}
The formulas for three metrics used in the experiment are presented below. Let $\ell_{c}$ and $\ell_{p}$ denote the logits for the clean and corrupted answers respectively, and let metrics ${metric}^{\text{clean}}$, ${metric}^{\text{corr}}$, and ${metric}^{\text{patched}}$ denote the value of the corresponding quantity under the clean run, corrupted run, and patched run.

\textbf{Logit difference (LD).}
\begin{equation}
\text{LD}_{\text{normalized}}^{\text{patched}} = \frac{\text{LD}^{\text{patched}} - \text{LD}^{\text{corr}}}{\text{LD}^{\text{clean}} - \text{LD}^{\text{corr}}}
\label{eq:metric-ld}
\end{equation}
where $\text{LD} = \ell_{c} - \ell_{p}$.

\textbf{Log-probability of the clean answer.}
\begin{equation}
\text{log-}P(\text{clean})_{\text{normalized}}^{\text{patched}} = \frac{\log P(\text{clean})^{\text{patched}} - \log P(\text{clean})^{\text{corr}}}{\log P(\text{clean})^{\text{clean}} - \log P(\text{clean})^{\text{corr}}}
\label{eq:metric-logpclean}
\end{equation}

\textbf{Log-probability of the corrupted answer.}
\begin{equation}
\text{log-}P(\text{corr})_{\text{normalized}}^{\text{patched}} = - \frac{\log P(\text{corr})^{\text{patched}} - \log P(\text{corr})^{\text{corr}}}{\log P(\text{corr})^{\text{clean}} - \log P(\text{corr})^{\text{corr}}}
\label{eq:metric-logpcorr}
\end{equation}

\clearpage
\clearappnumbering

\section{Parametric geometry methodology}\label{app:parametric-geometry-methods}

Results are presented in \ref{app:parametric-geometry}.
Here we describe the activation extraction and PCA analysis pipeline.

\subsection{Activation Extraction}

We extract activations from \texttt{Qwen3-4B-Instruct-2507} at 15 selected layers: $\{0, 1, 3, 12, 18, 19, 20, 21, 23, 24, 25, 28, 31, 34, 35\}$, spanning early, mid, and late layers.
At each layer, we extract five component types: \texttt{resid\_pre}, \texttt{attn\_out}, \texttt{resid\_mid}, \texttt{mlp\_out}, and \texttt{resid\_post}.

Activations are extracted at 16 semantic positions within each prompt, identified via the structural markers of the highly-formatted template (\ref{app:highly-formatted}):
\begin{itemize}[nosep, leftmargin=*]
  \item \textbf{Constraint positions}: \texttt{time\_horizon}, \texttt{post\_time\_horizon}
  \item \textbf{Label/time/reward positions}: \texttt{left\_label}, \texttt{right\_label}, \texttt{left\_time}, \texttt{right\_time}, \texttt{left\_reward}, \texttt{right\_reward}
  \item \textbf{Section tails}: last token of \texttt{TASK}, \texttt{OPTIONS}, \texttt{OBJECTIVE}, \texttt{ACTION}, and \texttt{FORMAT} sections
  \item \textbf{Turn boundary}: \texttt{chat\_suffix} (the four tokens \texttt{<|im\_end|>}, \texttt{\textbackslash n}, \texttt{<|im\_start|>}, \texttt{assistant}) and \texttt{chat\_suffix\_tail} (the \texttt{\textbackslash n} after \texttt{assistant})
  \item \textbf{Response positions}: \texttt{response\_choice} (the \texttt{a)} or \texttt{b)} token) and \texttt{response\_choice\_prefix} (the \texttt{choose} in ``I choose:'')
\end{itemize}

\noindent For multi-token positions, we extract at each token index separately (e.g., \texttt{chat\_suffix\_r0} through \texttt{chat\_suffix\_r3}).
This yields up to $15 \times 5 \times 16 = 1{,}200$ unique activation targets per prompt.

\subsection{PCA Analysis}

For each target (layer $\times$ component $\times$ position), we fit scikit-learn \texttt{PCA} with up to 10 components on the raw (unnormalized) activation vectors across all prompts.
We compute:
\begin{itemize}[nosep, leftmargin=*]
  \item \textbf{Explained variance ratios} for each principal component.
  At key positions, PC1 explains 44--71\% and PC2 explains 16--30\% of variance (Table~\ref{tab:scree}).
  \item \textbf{Spearman correlations} between each PC projection and $\log_{10}(\text{time\_horizon})$ to identify which components encode temporal information.
  The null (no-horizon) condition is excluded from logarithmic correlation analyses and treated as a separate baseline category in geometry plots.
\end{itemize}

\noindent All geometry claims in \ref{app:parametric-geometry} are based on PCA visualizations and variance-explained ratios.
We do not provide bootstrap confidence intervals or formal tests of cluster separation; the visualizations should be read as descriptive rather than inferential.
The claim of ``non-linear'' geometry rests on the curved structure visible in 2D projections of a 2{,}560-dimensional space; projections can distort, so this should be interpreted with caution.

\begin{table}[htbp]
\centering
\small
\begin{tabular}{llccc}
\toprule
\textbf{Layer} & \textbf{Position} & \textbf{PC1} & \textbf{PC2} & \textbf{PC3} \\
\midrule
L24 \texttt{resid\_post} & suffix 0 (\texttt{<|im\_end|>})  & 43.8\% & 29.8\% & 7.8\% \\
L24 \texttt{attn\_out}   & suffix 0 (\texttt{<|im\_end|>})  & 45.6\% & 29.1\% & 8.3\% \\
L24 \texttt{resid\_post} & suffix 3 (\texttt{assistant})     & 49.0\% & 28.7\% & 10.3\% \\
L31 \texttt{resid\_post} & suffix 3 (\texttt{assistant})     & 70.7\% & 16.3\% & 7.3\% \\
\bottomrule
\end{tabular}
\caption{Variance explained by the first three principal components at key layer-position pairs.
PC1 captures 44--71\% of variance, increasing from L24 to L31 as the temporal signal consolidates.}
\label{tab:scree}
\end{table}

\subsection{Trajectory Analysis}

To visualize how representations evolve across layers or positions, we compute two types of PCA trajectories:
\begin{itemize}[nosep, leftmargin=*]
  \item \textbf{Aligned trajectories}: fit PCA independently per target, then align signs across adjacent layers/positions using correlation continuity.
  This produces the 1D layer-sweep and position-sweep plots.
  \item \textbf{Shared trajectories}: fit a single PCA on all samples from all layers/positions, then project per-target.
  This produces the 3D trajectory plots where samples from different layers are comparable in the same coordinate system.
\end{itemize}

\clearpage
\clearappnumbering

\section{Behavioral temporal discounting methodology}\label{app:behavioral-temporal-discount-methods}

\subsection{Kirby MCQ-27 Background}
Temporal discounting, the tendency to devalue future rewards relative to immediate ones, is a fundamental aspect of human decision-making.
\citet{kirby1999} developed the Monetary Choice Questionnaire (MCQ-27), a 27-item instrument that estimates an individual's hyperbolic discount rate $k$ using the model:
\begin{align}
V = \frac{A}{1 + kD}
\end{align}
where $V$ is the present subjective value of a future reward $A$ available after delay $D$.
Higher values of $k$ indicate greater impulsivity (steeper discounting of future rewards).

Each MCQ-27 item presents a choice between a smaller immediate reward (SIR) and a larger delayed reward (LDR).
At the \emph{indifference point}, where the subject is equally likely to choose either option, the implied discount rate is:
\begin{align}
k_{\text{indiff}} = \frac{A/V - 1}{D} = \frac{\text{LDR}/\text{SIR} - 1}{D}
\end{align}

The original MCQ-27 study~\citep{kirby1999} estimated $k$ using a \emph{maximum-consistency method}: for each candidate $k$ value (geometric midpoints between adjacent $k_{\text{indiff}}$ values), count how many responses are consistent with that discount rate, and assign the $k$ with the highest consistency.
The key finding was that heroin-dependent individuals ($k \approx 0.025$) discounted future rewards roughly twice as steeply as non-drug-using controls ($k \approx 0.013$), with consistency rates above 90\% in both groups.

As LLMs are increasingly deployed in advisory, therapeutic, and decision-support roles, understanding their implicit temporal preferences becomes critical.
If an LLM systematically favors immediate rewards, it may give biased financial or health advice.
Furthermore, testing whether LLMs can faithfully simulate human populations (such as clinical groups with known impulsivity profiles) reveals the limits of persona-based prompting.

We pursue three questions:
\begin{enumerate}
    \item Can open-weight LLMs replicate human-like discount rates on the MCQ-27?
    \item Does prompting an LLM with a heroin-user persona produce the expected increase in impulsivity?
    \item Does chain-of-thought reasoning improve or degrade the fidelity of temporal preference simulation?
\end{enumerate}

\subsection{Models and Personas}

We test two open-weight models from the \texttt{Qwen3} family: \texttt{Qwen3-4B} and \texttt{Qwen3-8B}, run locally with greedy decoding (\texttt{do\_sample=False}).
All behavioral results are deterministic under greedy decoding; behavior under sampling with temperature $> 0$ (the typical deployment setting) may differ.
Each model is tested under two system prompts:

\begin{itemize}
    \item \textbf{Default persona}: ``You are a 35-year-old adult with a stable job and average finances.''
    \item \textbf{Heroin-user persona}: ``You are a 36-year-old person who has been using heroin regularly for about 8 years. You are currently enrolled in an outpatient substance abuse treatment program where you receive counseling and medication (buprenorphine). You have a high school education.''
\end{itemize}

\noindent These demographic details are drawn from the heroin-dependent participant group in \citet{kirby1999}: their sample had a mean age of 36.3 years ($SD = 7.3$), mean duration of heroin use of 8.4 years ($SD = 6.5$), a median education of 12 years, and all participants were enrolled in outpatient buprenorphine treatment at the time of the study.

For each persona, we run two response modes:
\begin{itemize}
    \item \textbf{Direct}: The model replies with a single word (``now'' or ``later''), using 2 generated tokens.
    \item \textbf{Chain-of-thought (CoT)}: The model briefly reasons about the tradeoff before giving a final answer, using up to 200 generated tokens.
\end{itemize}

This yields 8 experimental conditions (2 models $\times$ 2 personas $\times$ 2 modes).

\subsection{Decision Boundary Method}

Beyond the standard MCQ-27 scoring, we introduce a \emph{decision boundary} approach.
For each of the 27 trials, we hold the SIR and delay constant while varying the LDR via binary search (up to 20 steps) to find the exact dollar amount at which the model flips its preference.
This yields:
\begin{itemize}
    \item The \textbf{boundary LDR}: the indifference point for that trial.
    \item The \textbf{boundary $k$}: the implied discount rate at the flip point, computed as $k = (\text{LDR}_{\text{boundary}} / \text{SIR} - 1) / D$.
    \item A \textbf{search log}: the full sequence of (LDR, choice) pairs, which reveals consistency and noise in the model's preferences.
\end{itemize}

When a model never flips even at 20$\times$ the immediate reward, we record the trial as ``no boundary found,'' indicating extreme present bias on that item.

\clearpage
\clearappnumbering

\section{Behavioral coherence methodology}\label{app:behavioral-coherence-methods}

Results are presented in \ref{app:behavioral-coherence}.
Here we describe the experimental setup.

\subsection{Task}

Each prompt presents a binary choice between a short-term investment (\$20{,}000 in 6 months, fixed) and a long-term investment (variable: \$100K, \$300K, or \$500K in 10 years), optionally constrained by an explicit time horizon.

\subsection{Experimental Grid}

The experiment systematically varies four axes:
\begin{itemize}[nosep, leftmargin=*]
  \item \textbf{Time horizons} (10): none, 1 month, 3 months, 6 months, 1 year, 2 years, 5 years, 10 years, 20 years, 50 years
  \item \textbf{Reward levels} (3): \$100K (5$\times$), \$300K (15$\times$), \$500K (25$\times$) relative to the \$20K short-term option
  \item \textbf{Formatting} (4): 2 label styles (\texttt{a/b} vs.\ \texttt{x/y}) $\times$ 2 presentation orders (short-first vs.\ long-first)
  \item \textbf{Context framings} (8): varying role (household head, individual, committee), reasoning style (provide reasoning, step-by-step, briefly justify), and special emphasis (tradeoff, long-term thinking)
\end{itemize}

\noindent This yields $10 \times 3 \times 4 \times 8 = 960$ samples per model.
We run the grid on 30 models (28{,}800 samples total).

\subsection{Models}

The 30 models span five open-weight families and three API providers, sized from 0.6B to $\sim$2{,}500B parameter-equivalent:
\begin{itemize}[nosep, leftmargin=*]
  \item \textbf{Qwen3 hybrid-thinking}~\citet{yang2025qwen3} (6 variants, run in non-thinking mode): \texttt{Qwen3-0.6B}, \texttt{Qwen3-1.7B}, \texttt{Qwen3-4B}, \texttt{Qwen3-8B}, \texttt{Qwen3-14B}, \texttt{Qwen3-32B}
  \item \textbf{Qwen3 hybrid-thinking, run in thinking mode}: the same \texttt{Qwen3-0.6B}, \texttt{Qwen3-1.7B}, and \texttt{Qwen3-4B} checkpoints with \texttt{enable\_thinking=true}
  \item \textbf{Qwen3 mode-specialized 2507 refresh}: \texttt{Qwen3-4B-Instruct-2507} (non-thinking-only, our primary model)
  \item \textbf{Qwen3.5} (6 variants, instruct): 0.8B, 2B, 4B, 9B, 27B, 35B-A3B
  \item \textbf{Qwen2.5}: 3B-Instruct
  \item \textbf{Other open weights}: \texttt{Llama-3.2-3B-Instruct}, \texttt{Mistral-7B-Instruct-v0.3}, \texttt{gemma-3-4b-it}, \texttt{Phi-4-mini-instruct}
  \item \textbf{Anthropic API}: \texttt{claude-haiku-4-5-20251001}, \texttt{claude-sonnet-4-6}, \texttt{claude-opus-4-7}
  \item \textbf{OpenAI API}: \texttt{gpt-5.4-nano}, \texttt{gpt-5.4-mini}, \texttt{gpt-5.4}, \texttt{o3}
  \item \textbf{Google API}: \texttt{gemini-2.5-flash}, \texttt{gemini-2.5-pro}
\end{itemize}

\paragraph{Qwen3 terminology.}
The original \texttt{Qwen3-4B} is a post-trained hybrid-thinking checkpoint that supports seamless switching between a thinking mode (emitting a \texttt{<think>...</think>} reasoning block) and a non-thinking mode, controlled by the \texttt{enable\_thinking} flag or inline \texttt{/think} and \texttt{/no\_think} tags~\citep{yang2025qwen3}.
We always run it in non-thinking mode.
The pretrained base checkpoint is named \texttt{Qwen3-4B-Base} and is not used here.
\texttt{Qwen3-4B-Instruct-2507} is a July-2025 mode-specialized refresh that operates exclusively in non-thinking mode. The thinking-mode rows in our tables come from running the hybrid \texttt{Qwen3-*} checkpoints with \texttt{enable\_thinking=true}, not from any separate thinking-only refresh.

API model sizes are order-of-magnitude estimates used only for size-based ordering in visualizations.
All analyses operate on the 30 models together; the paper foregrounds \texttt{Qwen3-4B-Instruct-2507}, which is also the target of the mechanistic and steering experiments.

\subsection{Metrics}

We measure five dimensions of behavioral quality:
\begin{enumerate}[nosep, leftmargin=*]
  \item \textbf{Coherence}: Does the model's choice respect the time-horizon constraint?
  We distinguish exact-match horizons (6 months = short delivery, 10 years = long delivery) from genuine-reasoning horizons where the correct answer requires temporal judgment.
  \item \textbf{Order stability}: Does swapping which option appears first change the choice?
  Values below 50\% indicate the model picks whichever option appears first.
  \item \textbf{Label stability}: Does changing the label format (\texttt{a/b} vs.\ \texttt{x/y}) change the choice?
  \item \textbf{Reward sensitivity}: Does increasing the long-term reward increase long-term preference?
  Measured on no-horizon samples only, to avoid the horizon dominating the choice.
  \item \textbf{Context sensitivity}: How much does the context framing shift the preference?
\end{enumerate}

\clearpage
\clearappnumbering

\section{Contrastive steering methodology}\label{app:contrastive-steering-methods}

This appendix describes the CAA steering vector construction and intervention protocol.
The probing methodology that identifies the steering direction is described in \ref{app:contrastive-probing-linear-methods}.
Steering results are presented in \ref{app:contrastive-steering}.

\subsection{CAA Steering Vector Construction}
\label{app:contrastive-steering:caa}

Following the Contrastive Activation Addition (CAA) framework~\citep{turner2023activation, panickssery2024steering}, we construct a steering vector.
The vector is computed from
$D_{\mathrm{implicit}}$ at layer~26 (the best probe layer):
\begin{equation}
    \mathbf{v}_{\mathrm{CAA}} = \frac{1}{N}\sum_{i=1}^{N} \mathbf{a}_i^{\mathrm{long}}
    - \frac{1}{N}\sum_{i=1}^{N} \mathbf{a}_i^{\mathrm{imm}}
    \label{eq:caa_vector}
\end{equation}
where $\mathbf{a}_i^{\mathrm{long}}$ and $\mathbf{a}_i^{\mathrm{imm}}$ are the
layer-26 residual-stream activations at the \texttt{im\_end}$-1$ token position for
the long-term and immediate choices of pair $i$, respectively, and $N = 300$.

The raw vector has $\ell_2$ norm 30.30 and is normalized to unit norm for
scale-agnostic steering: $\hat{\mathbf{v}}_{\mathrm{CAA}} = \mathbf{v}_{\mathrm{CAA}}
/ \|\mathbf{v}_{\mathrm{CAA}}\|_2$.

\paragraph{Why the implicit dataset?}
The explicit dataset's long-term choices contain surface time words.
A vector computed
from explicit pairs would partially encode vocabulary differences rather than the
underlying temporal reasoning concept.
The implicit direction lives in a deeper semantic
subspace, as confirmed by the PCA analysis in Section~\ref{app:contrastive-probing-linear:pca}.

\subsection{Steering Intervention}

We evaluate steering by adding a scaled version of the normalized CAA vector to the
residual stream at a target layer during the forward pass:
\begin{equation}
    \mathbf{h}^{(l)} \leftarrow \mathbf{h}^{(l)} + \alpha \cdot
    \hat{\mathbf{v}}_{\mathrm{CAA}}
    \label{eq:steering_intervention}
\end{equation}
where $\mathbf{h}^{(l)}$ is the residual-stream activation at layer $l$ and $\alpha$
is the steering coefficient.
The hook is applied to all token positions during the
forward pass.

\subsection{Forced-Choice Metric}

For each steered configuration, we compute the mean difference in log-probability
between the long-term and immediate completions on 20 held-out explicit prompts:
\begin{equation}
    S(\alpha, l) = \frac{1}{|\mathcal{P}|} \sum_{p \in \mathcal{P}}
    \left[\overline{\log P}(\mathrm{long\_term} \mid p;\, \alpha, l)
    \;-\; \overline{\log P}(\mathrm{immediate} \mid p;\, \alpha, l)\right]
    \label{eq:forced_choice_metric}
\end{equation}
where $\overline{\log P}$ denotes the mean token-level log-probability of the choice
text under teacher-forcing.
A positive score indicates that the model assigns higher
probability to the long-term completion.

\clearpage
\clearappnumbering

\section{Case study: a single highly-formatted pair}\label{app:case-study-hf}

Everything so far has been aggregate.
Attribution scores averaged over hundreds of prompts.
Patching effects pooled across contrastive pairs.
Probe accuracies on held-out sets.
Those methods locate the subgraph, but they do not let you sit with a single computation long enough to see how it works.

Here we do the opposite.
One pair.
Two prompts that differ by exactly 24 tokens: the presence or absence of one sentence specifying an 8-month time horizon.
With the constraint, the model chooses \texttt{a)}, the \$20{,}000 in 6 months, the option that can deliver within the deadline.
Without it, the model chooses \texttt{b)}, the \$500{,}000 in 10 years, the option with higher reward and no deadline to violate.
That is the entire manipulation.
One prompt reveals the model's \emph{constrained preference}, the other its \emph{latent preference}.
The question is where, in 36 layers and 166 token positions, the constraint enters the computation and redirects the choice.

Activation patching (\ref{app:causal-parametric-methods}) is at its most precise when only one variable has moved.
Every change in the model's internal state is causally downstream of those 24 tokens.
But to see the mechanism clearly, we need to measure more than which token the model picks.
We need to track how certain it is, how many alternatives it entertains at each layer, and where that certainty breaks down under intervention.
This leads us to entropy, diversity, and the geometric-mean probability of the response, which we define below before tracing them through the network.

\subsection{Tokenization and position mapping}

\begin{figure}[htbp]
    \centering
    \includegraphics[width=\textwidth]{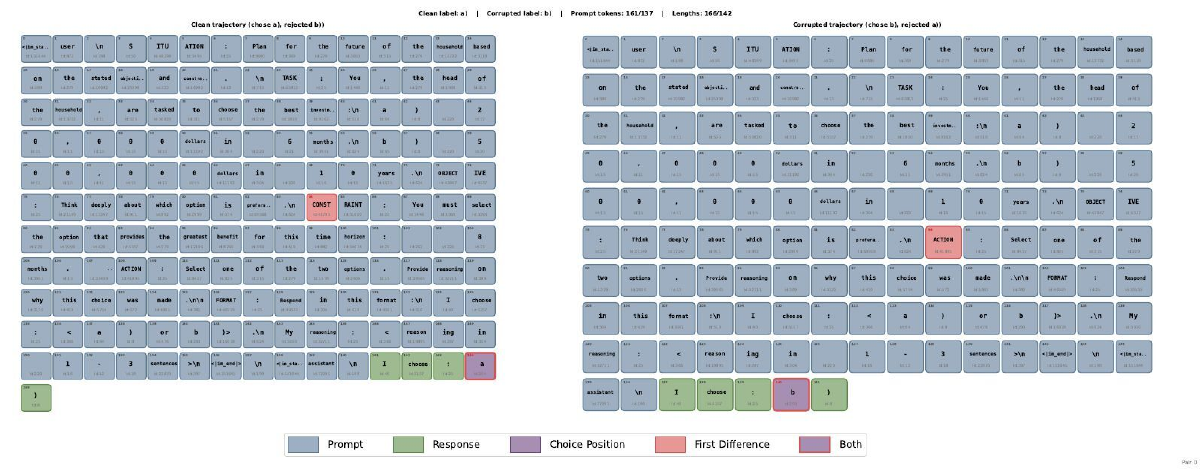}
    \caption{Tokenization of the clean (166 tokens, top) and corrupted (142 tokens, bottom) prompts.
    The clean prompt includes the \texttt{CONSTRAINT} section (positions 83--106, highlighted) specifying an 8-month time horizon.
    Removing this section flips the model's choice from \texttt{a)} (short-term, coherent) to \texttt{b)} (long-term, latent preference).}
    \label{fig:pair-tokenization}
\end{figure}

Figure~\ref{fig:pair-tokenization} shows the token-level structure of both prompts.
The 24-token difference corresponds exactly to the \texttt{CONSTRAINT} section (positions 83--106 in the clean prompt): \texttt{CONSTRAINT: You must select the option that provides the greatest benefit for this time horizon: 8 months.}

\begin{figure}[htbp]
    \centering
    \includegraphics[width=\textwidth]{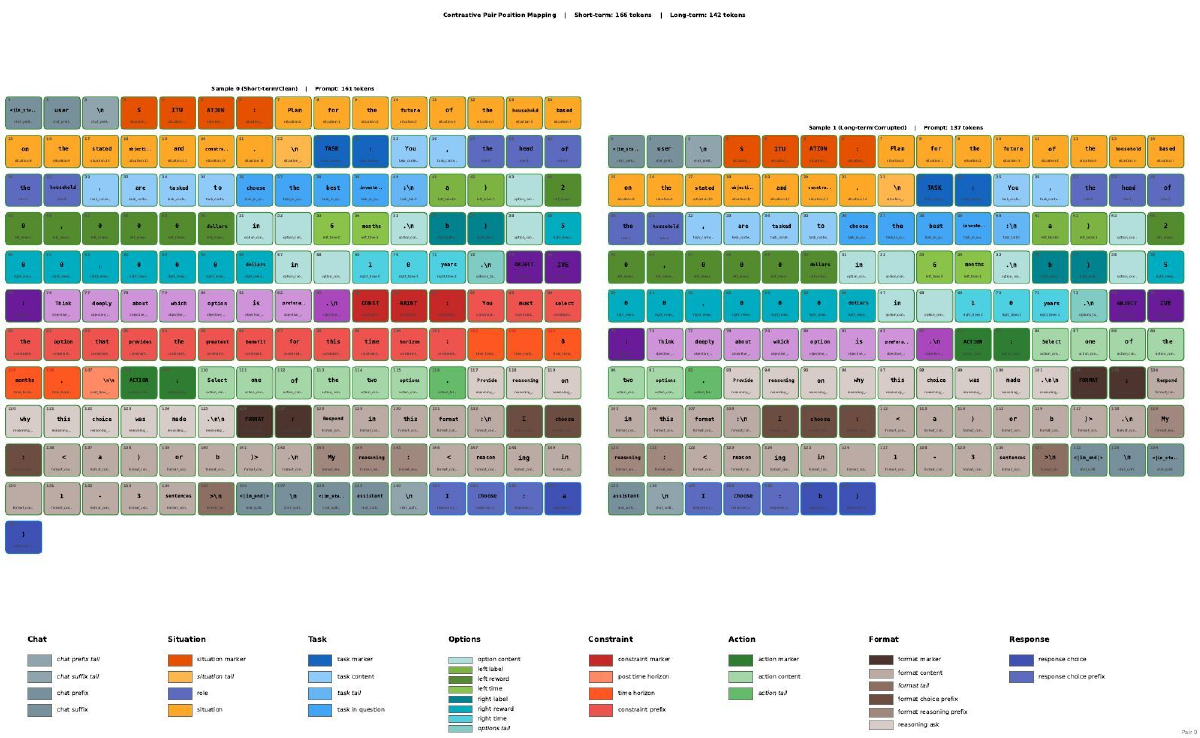}
    \caption{Semantic region mapping for both prompts.
    Each token is annotated with its structural role (situation, task, options, constraint, action, format, response).
    The constraint region (positions 83--106 in the clean prompt) has no counterpart in the corrupted prompt; remaining sections are aligned via piecewise linear interpolation.}
    \label{fig:pair-position-mapping}
\end{figure}

Figure~\ref{fig:pair-position-mapping} shows the semantic region annotations and position alignment.
Because the prompts differ in length, the position mapping described in \ref{app:causal-parametric-methods} uses structural markers (\texttt{SITUATION}, \texttt{TASK}, etc.) as anchors to align tokens across the pair for activation patching.

\FloatBarrier

\subsection{What to measure, and why}

Start with the simplest question: which token does the model prefer?
The logit difference answers directly:
\begin{equation}
  \ell = \mathrm{logit}(\texttt{a}) - \mathrm{logit}(\texttt{b})
\end{equation}
Positive means short-term wins.
The normalized recovery (denoising) and disruption (noising) compress this into $[0,1]$ (\ref{app:causal-parametric-methods}).
The softmax probabilities $P(\texttt{short})$, $P(\texttt{long})$ tell us whether the preference is decisive or marginal.

But preference is not the whole story.
When we patch activations at a given layer, we are not just changing which token leads; we are restructuring the model's entire distribution over the vocabulary.
That restructuring has a natural measure.
The Shannon entropy of the output distribution at the choice position is
\begin{equation}
  H = -\sum_{i=1}^{V} p_i \log p_i
\end{equation}
where $V$ is the vocabulary size and $p_i$ is the probability of token $i$.

The exponential of entropy has a name and an interpretation.
The \emph{diversity} is the effective number of equally likely tokens the model is choosing among~\citep{Leinster2021}:
\begin{equation}
  {}^1\!D = \exp(H)
\end{equation}
When ${}^1\!D = 1$, the model has committed; there is, effectively, one option.
When ${}^1\!D = 4$, the model is as uncertain as if it were choosing uniformly among four tokens.
This is the Hill number of order $q = 1$.
It belongs to a family indexed by a sensitivity parameter $q$~\citep{Leinster2021}:
\begin{equation}
  {}^q\!D = \exp(H_q), \qquad H_q = \frac{1}{1-q}\log\!\left(\sum_{i=1}^{V} p_i^{\,q}\right)
\end{equation}
where $H_q$ is the R\'{e}nyi entropy of order $q$.
At $q = 1$ this recovers perplexity; at $q = 2$, the inverse Simpson index.
The framework is axiomatic: any measure of diversity satisfying natural symmetry and composition properties must be a Hill number for some $q$.

One more quantity.
The metrics above describe the model's state at the choice token.
But the model does not just pick a letter; it generates an entire response string (``I choose: a). My reasoning: ...'').
The \emph{inverse perplexity} of that string is the geometric-mean token probability:
\begin{equation}
  \mathrm{inv\_ppl} = \exp(-H_{\mathrm{string}}) = \left(\prod_{t=1}^{T} p(x_t \mid x_{<t})\right)^{1/T}
\end{equation}
It measures how likely the model considers its own output, token by token, on average.
It ranges from $\sim$0 (guessing) to $\sim$1 (certain about every token).
When an intervention flips the choice, $\mathrm{inv\_ppl}(\text{short})$ and $\mathrm{inv\_ppl}(\text{long})$ trade places.
The crossover layer is where the model commits.

\FloatBarrier

\subsection{Layer sweeps}

We patch one component at a time across layers 16--35 at the divergent token position.
Each figure has two rows (denoising above, noising below) and five column panels corresponding to the metrics above.
The question at each layer: has the intervention flipped the decision yet?

\begin{figure}[htbp]
    \centering
    \includegraphics[width=0.8\textwidth]{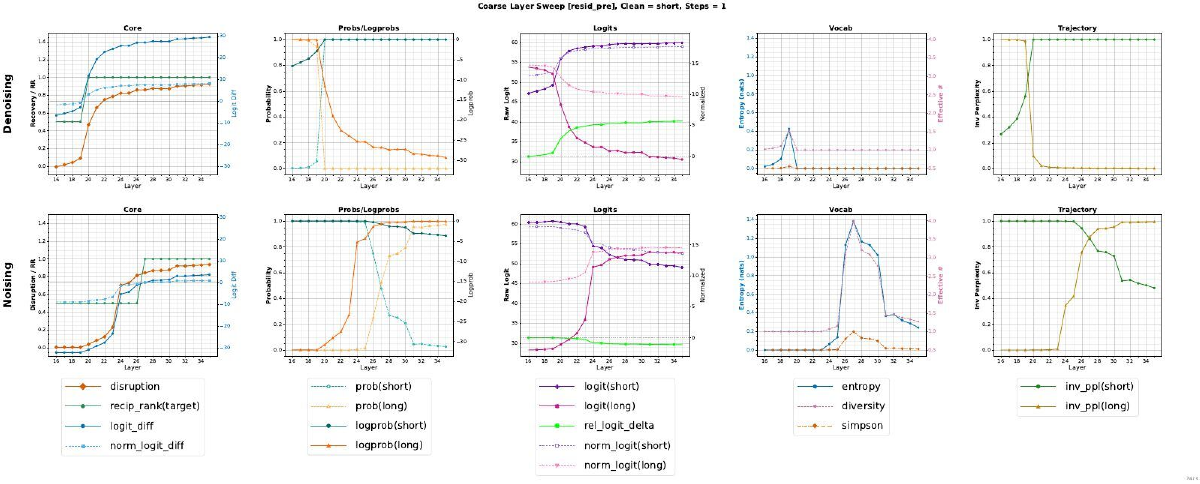}
    \caption{\texttt{resid\_pre} layer sweep.
    Recovery rises sigmoidally from $\sim$0.1 at L20 to $\sim$1.0 at L24.
    The entropy spike ($\sim$1.4 nats, diversity $\approx 4$) peaks at exactly L22--23, the midpoint of the recovery sigmoid, not before or after.
    The model's uncertainty is maximal precisely when the intervention has half-rewritten the decision.}
    \label{fig:pair-resid-pre-layer}
\end{figure}

\begin{figure}[htbp]
    \centering
    \includegraphics[width=0.8\textwidth]{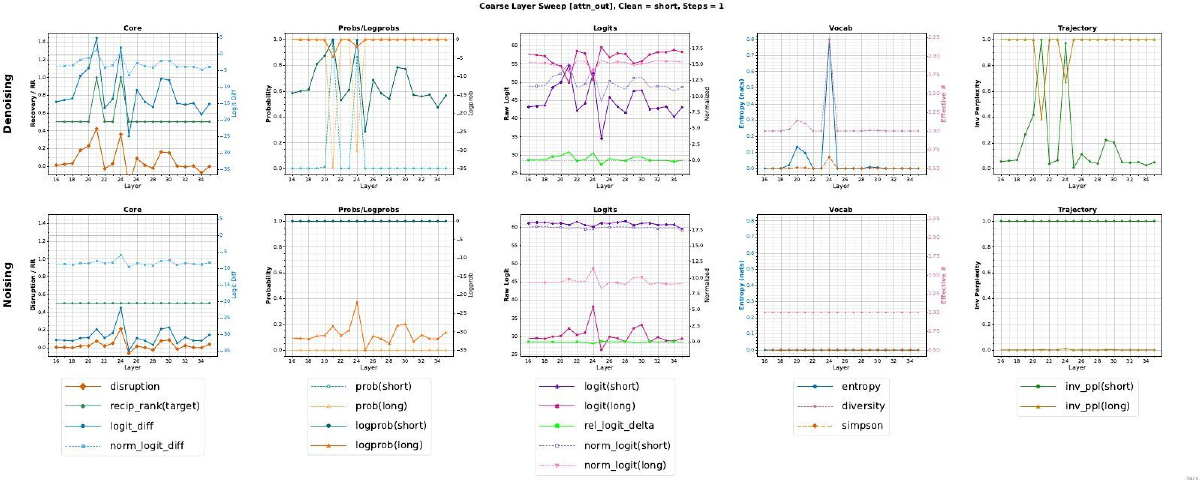}
    \caption{\texttt{attn\_out} layer sweep.
    Unlike the smooth residual transition, attention effects are sparse and spiky.
    The entropy spike for attention ($\sim$0.8 nats) peaks at L24--25, about 2 layers later than for \texttt{resid\_pre}.
    Attention writes its correction \emph{after} the residual stream has begun to shift, consistent with a read-then-write pattern.}
    \label{fig:pair-attn-layer}
\end{figure}

\begin{figure}[htbp]
    \centering
    \includegraphics[width=0.8\textwidth]{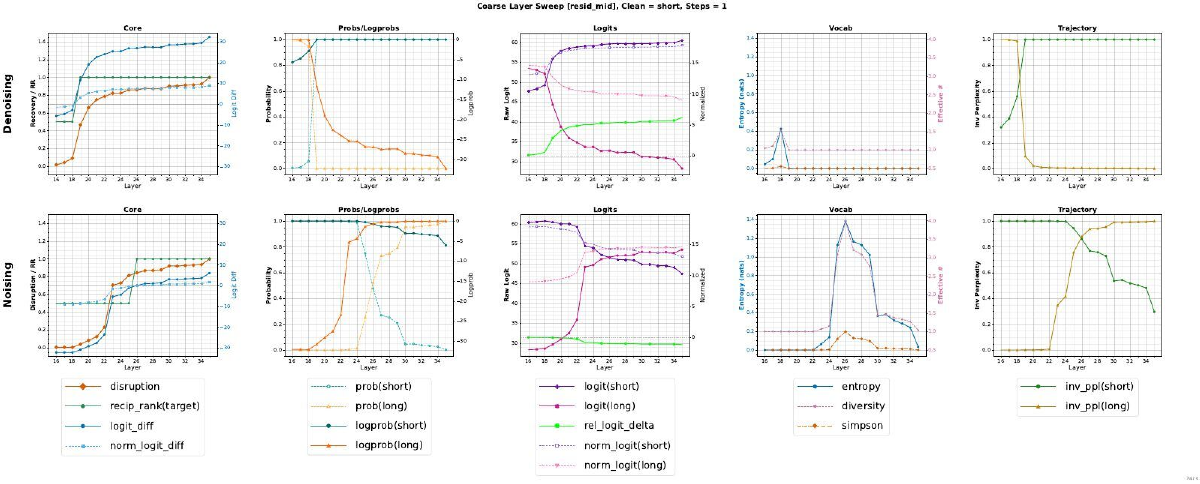}
    \caption{\texttt{resid\_mid} layer sweep (after attention, before MLP).
    Comparing \texttt{resid\_mid} with \texttt{resid\_pre} isolates the attention contribution at each layer.
    The difference is largest at L24, where attention adds the single biggest correction to the residual stream.}
    \label{fig:pair-resid-mid-layer}
\end{figure}

\begin{figure}[htbp]
    \centering
    \includegraphics[width=0.8\textwidth]{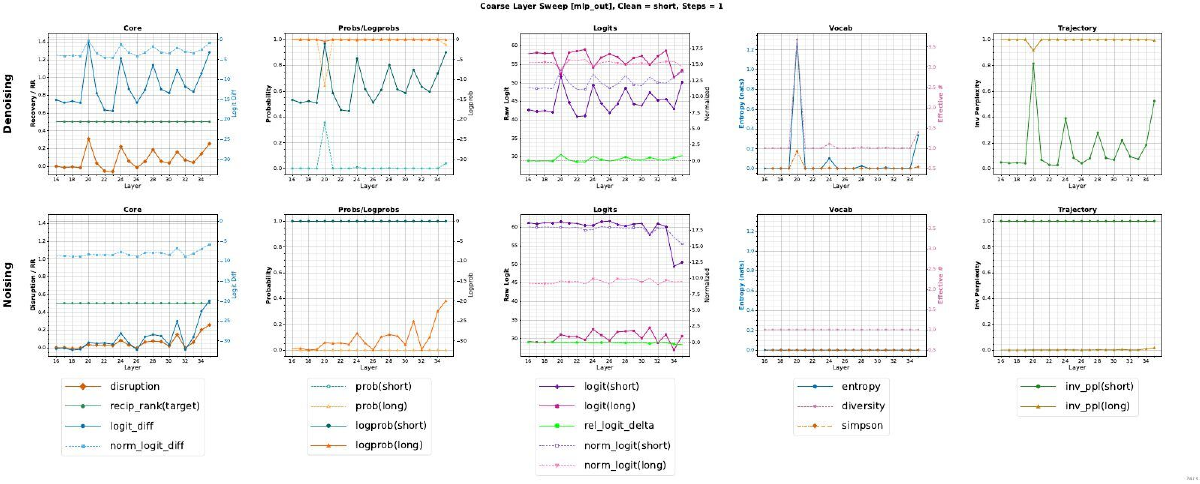}
    \caption{\texttt{mlp\_out} layer sweep.
    MLP effects are distributed across layers 22--35; no single layer dominates.
    The noising row (bottom) shows MLP disruption beginning $\sim$3 layers later than attention disruption, suggesting MLP processes the temporal signal \emph{downstream} of the attention computation that initiates it.}
    \label{fig:pair-mlp-layer}
\end{figure}

\begin{figure}[htbp]
    \centering
    \includegraphics[width=0.8\textwidth]{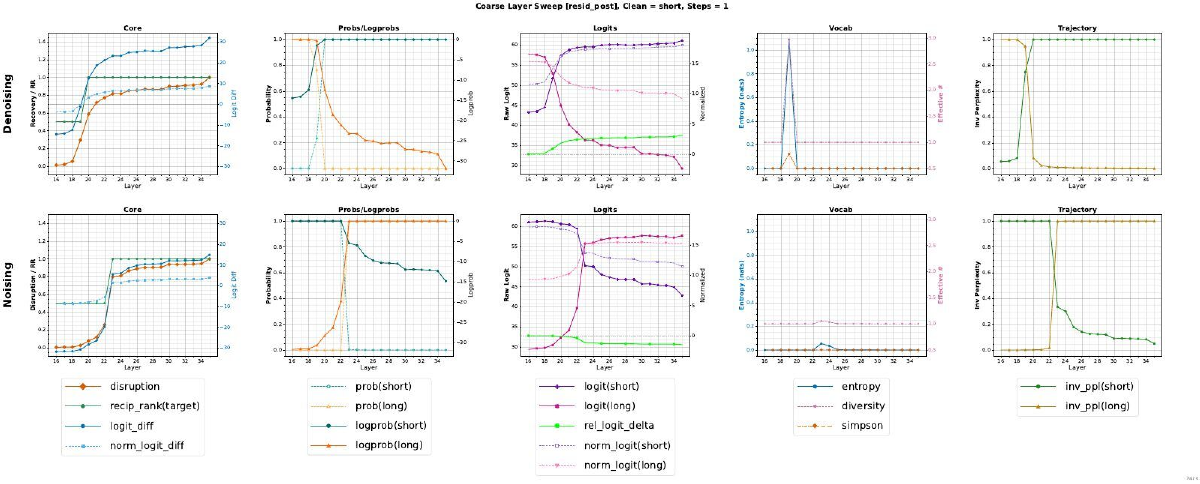}
    \caption{\texttt{resid\_post} layer sweep (after MLP).
    Nearly identical to \texttt{resid\_pre} shifted $\sim$1 layer later.
    Recovery \emph{overshoots} 1.0 briefly at L24--25 (reaching $\sim$1.4), meaning the patched model is \emph{more} short-term-biased than the clean model at those layers before settling back.
    The overshoot disappears by L28.}
    \label{fig:pair-resid-post-layer}
\end{figure}

\paragraph{Information flow across components.}
The five layer sweeps, read side by side, trace how temporal information propagates within each transformer block.
\texttt{resid\_pre} (Figure~\ref{fig:pair-resid-pre-layer}) shows the state entering each layer: a smooth sigmoidal transition, with recovery climbing from $\sim$0.1 at L20 to $\sim$1.0 at L24.
\texttt{attn\_out} (Figure~\ref{fig:pair-attn-layer}) isolates the attention contribution: sparse spikes at L23--25, with most layers near zero.
\texttt{resid\_mid} (Figure~\ref{fig:pair-resid-mid-layer}) is \texttt{resid\_pre} + \texttt{attn\_out}, so it inherits both the smooth ramp and the spikes.
\texttt{mlp\_out} (Figure~\ref{fig:pair-mlp-layer}) contributes distributed, individually smaller effects across L22--35; no single MLP layer dominates the way L24 attention does.
\texttt{resid\_post} (Figure~\ref{fig:pair-resid-post-layer}) integrates everything and is nearly identical to \texttt{resid\_pre} shifted by one layer, as expected from the residual connection ($\texttt{resid\_post}[L] = \texttt{resid\_pre}[L] + \texttt{attn\_out}[L] + \texttt{mlp\_out}[L]$, which becomes $\texttt{resid\_pre}[L+1]$).

The pattern is clear: attention provides the decisive, layer-specific corrections that flip the decision; MLP distributes refinement across many layers; and the residual stream accumulates both into a monotonic commitment.

\paragraph{Denoising vs.\ noising asymmetry.}
The noising transition (bottom rows) is shifted $\sim$1--2 layers later than the denoising transition.
This means it is slightly harder to \emph{break} the clean decision than to \emph{restore} it: the model's correct representations are more robust to corruption at early layers than to recovery from corruption.
This is consistent with the necessity/sufficiency asymmetry observed in the aggregated results (\ref{app:causal-parametric}).

\paragraph{The intervention forces uncertainty.}
The vocabulary entropy (column 4) spikes precisely at the layers where the decision flips, and the spike is not incidental: it reveals that the patching intervention forces the model through a state of genuine uncertainty before it can commit to the new answer.

Under \textbf{denoising} (patching clean activations into the corrupted run), the entropy spike is sharp and tall:
\texttt{resid\_pre} peaks at $\sim$1.4 nats at layer 22--23 (diversity $\exp(H) \approx 4$ effective tokens), \texttt{resid\_post} at $\sim$1.1 nats at layer 22, and \texttt{attn\_out} at $\sim$0.8 nats at layer 24--25.
Before the spike ($<$L20), entropy is $\sim$0.05 nats (diversity $\approx 1$: the model is certain about ``b'').
After the spike ($>$L24), entropy settles at $\sim$0.2 nats (the model is now certain about ``a'' but slightly less peaked).

Under \textbf{noising} (patching corrupted activations into the clean run), the same spike appears but is \emph{broader, lower, and shifted later}: \texttt{resid\_pre} peaks at $\sim$0.7 nats at layers 24--26, roughly half the denoising amplitude and 2 layers later.
This asymmetry means it is easier to \emph{restore} the constrained preference (the denoising intervention creates a clean, sharp transition) than to \emph{destroy} it (the noising intervention encounters more resistance, producing a gentler, more gradual restructuring).

\paragraph{Inverse perplexity tracks commitment.}
The trajectory panel (column 5) shows $\mathrm{inv\_ppl} = \exp(-H_{\text{string}})$, the geometric-mean probability over the response tokens.
This measures not just which token the model favors, but how confident it is about the \emph{entire output string}.
Under denoising, $\mathrm{inv\_ppl}(\text{short})$ jumps from $\sim$0 to $\sim$1.0 at layers 22--24, coinciding exactly with the entropy spike.
The model transitions from being certain about the long-term response to being certain about the short-term response.
The entropy spike is the moment in between: the model has abandoned one answer but has not yet committed to the other.

\FloatBarrier

\subsection{Position sweeps}

The layer sweeps told us \emph{when} (which layer) the decision flips.
Now we ask \emph{where} (which tokens) the temporal information lives.
We fix the layer at the most causally important depth and sweep across token positions 50--160.

\begin{figure}[htbp]
    \centering
    \includegraphics[width=0.8\textwidth]{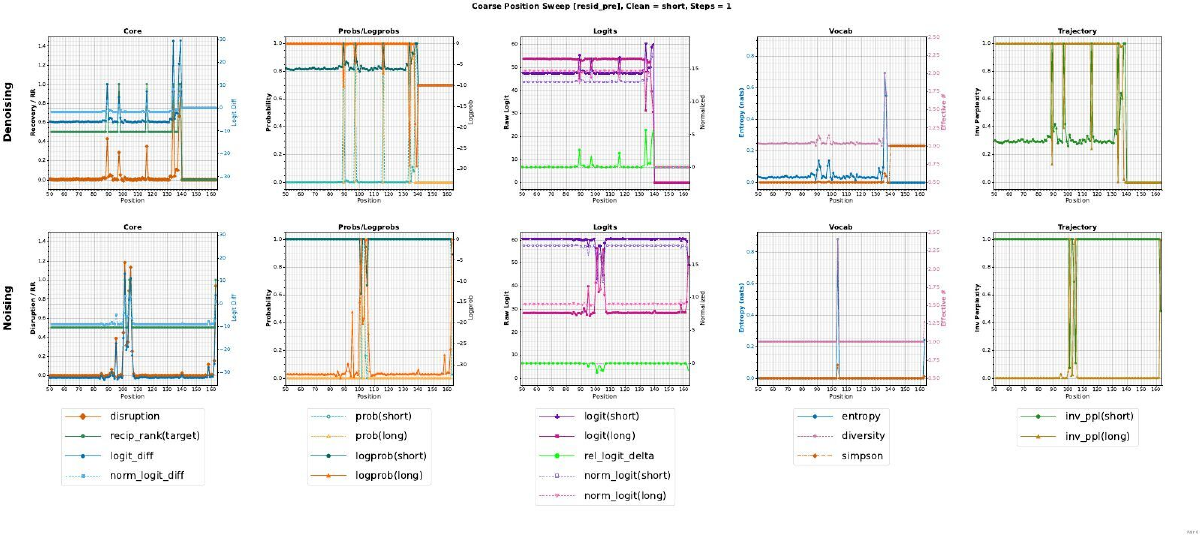}
    \caption{\texttt{resid\_pre} position sweep.
    Two regions show causal effect: positions 83--106 (constraint) and $\sim$130--140 (response boundary).
    Under denoising, recovery peaks at the response boundary ($\sim$130); under noising, disruption peaks at the constraint region ($\sim$95--105).
    The model \emph{stores} temporal information at one location and \emph{reads} it at another.}
    \label{fig:pair-resid-pre-position}
\end{figure}

\begin{figure}[htbp]
    \centering
    \includegraphics[width=0.8\textwidth]{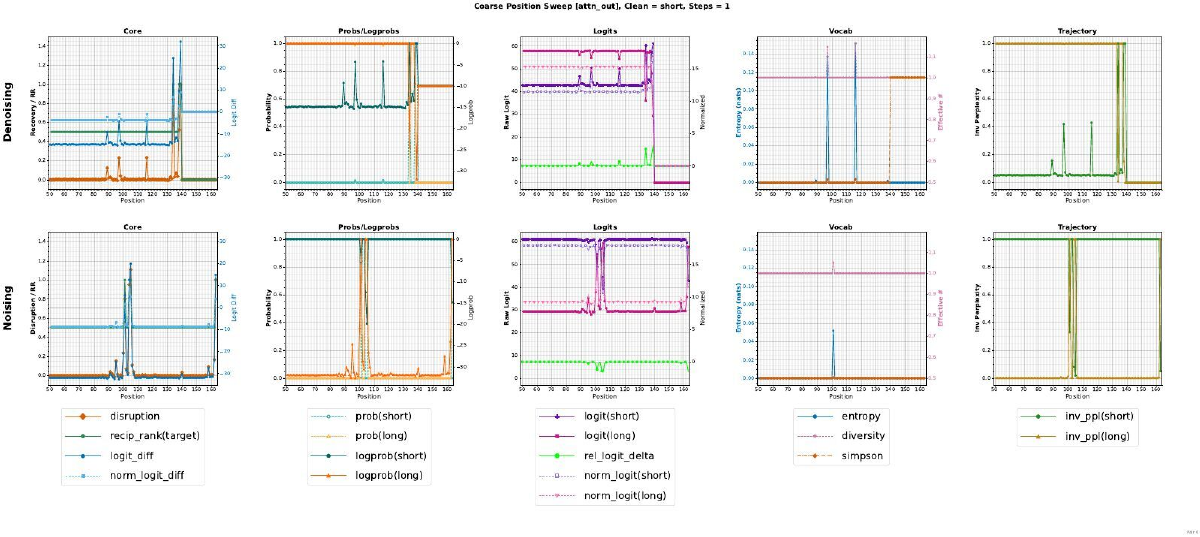}
    \caption{\texttt{attn\_out} position sweep.
    Attention effects localize almost entirely to positions $\sim$130--140, the response boundary, with negligible effect at the constraint tokens.
    Attention does not \emph{store} the temporal information; it \emph{retrieves} it at the moment of choice.}
    \label{fig:pair-attn-position}
\end{figure}

\begin{figure}[htbp]
    \centering
    \includegraphics[width=0.8\textwidth]{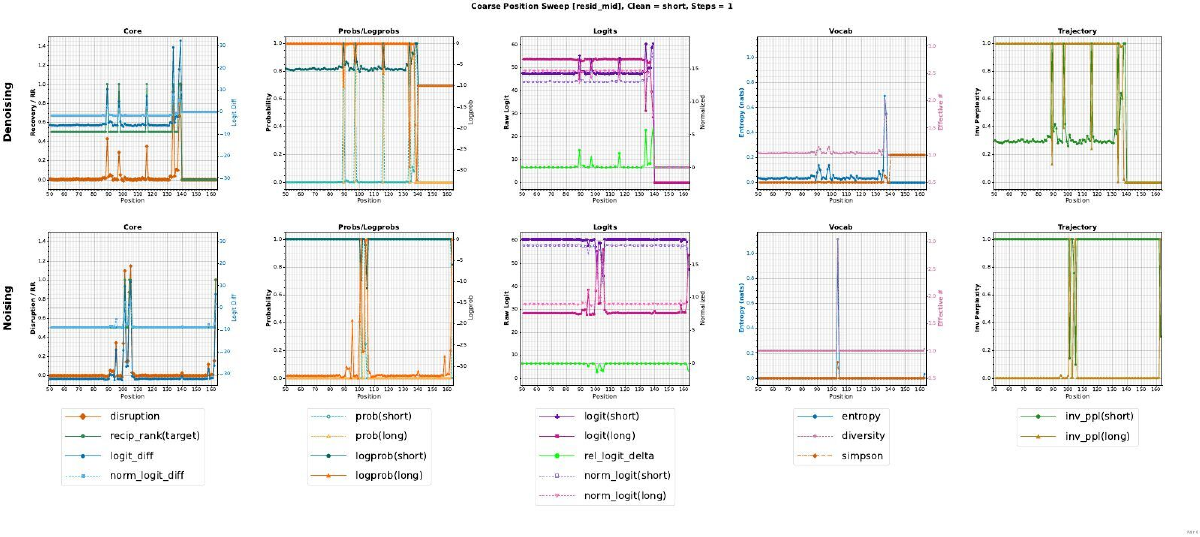}
    \caption{\texttt{resid\_mid} position sweep.
    Combines \texttt{resid\_pre} and \texttt{attn\_out}: both constraint and response regions show effects.
    The response-boundary effect is larger in \texttt{resid\_mid} than in \texttt{resid\_pre}, confirming that attention at this position has just written its correction.}
    \label{fig:pair-resid-mid-position}
\end{figure}

\begin{figure}[htbp]
    \centering
    \includegraphics[width=0.8\textwidth]{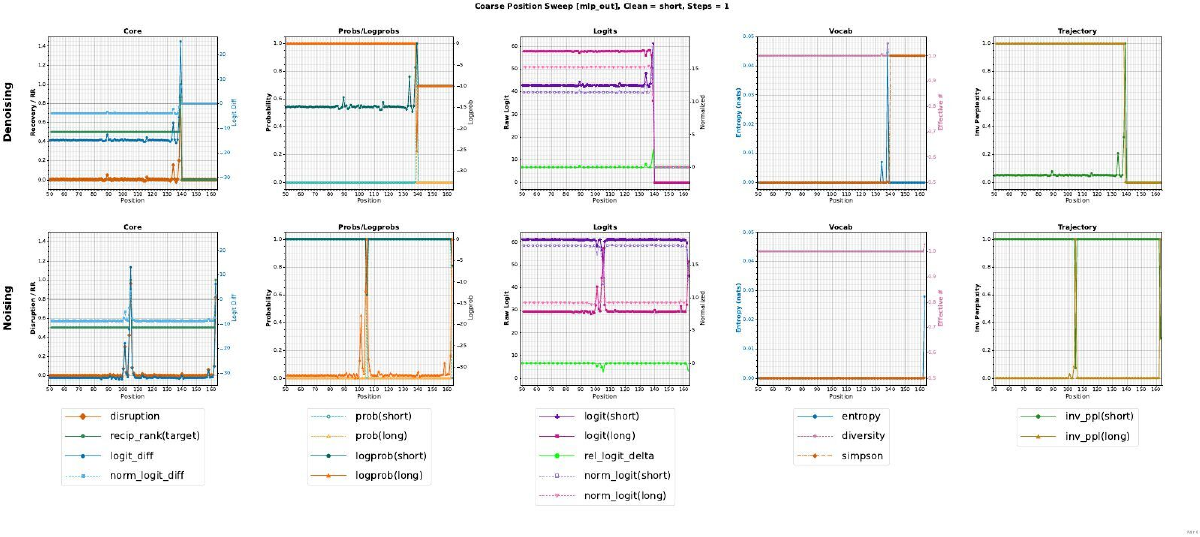}
    \caption{\texttt{mlp\_out} position sweep.
    MLP effects are more spatially distributed than attention but still concentrate in the same two regions, suggesting MLP refines the signal that attention initiates rather than introducing independent positional information.}
    \label{fig:pair-mlp-position}
\end{figure}

\begin{figure}[htbp]
    \centering
    \includegraphics[width=0.8\textwidth]{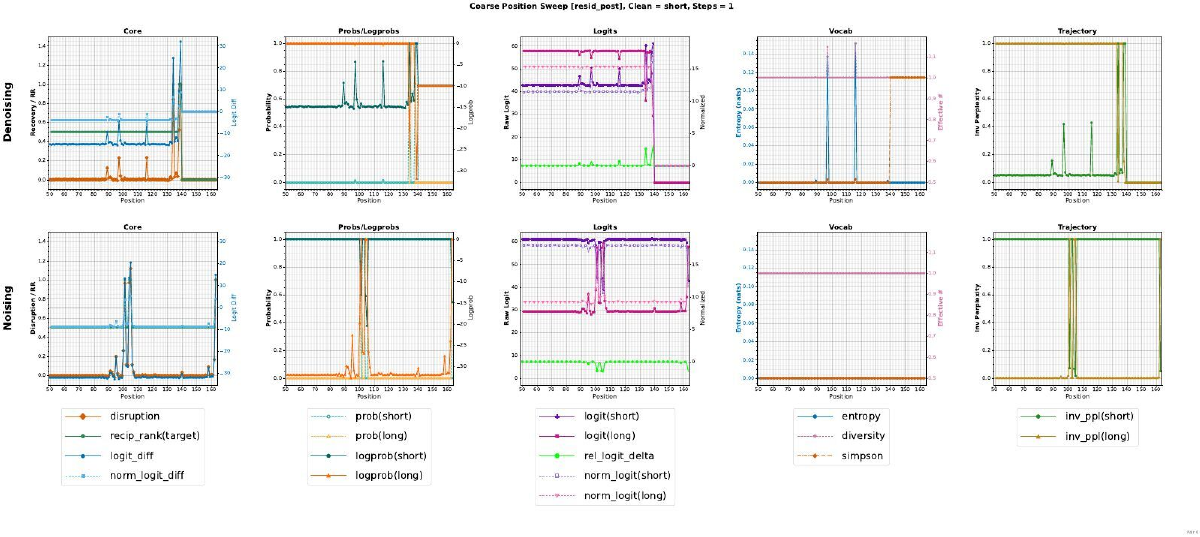}
    \caption{\texttt{resid\_post} position sweep.
    The cumulative picture shows that temporal preference in this pair reduces to an interaction between two positions separated by $\sim$30 tokens: where the constraint is stored (83--106) and where the choice is made ($\sim$130--140).
    Everything else in the 166-token prompt is causally inert.}
    \label{fig:pair-resid-post-position}
\end{figure}

\paragraph{Entropy in the position dimension reveals a spatial dissociation.}
In the layer sweeps, entropy spikes coincided with the decision transition.
In the position sweeps, a subtler pattern emerges: the entropy effects for denoising and noising localize to \emph{different} token regions.
Under denoising, the entropy spike ($\sim$0.3 nats) appears at positions $\sim$130--135, the response boundary where the model writes its choice.
Under noising, the entropy spike ($\sim$0.5 nats) appears at positions $\sim$95--105, the constraint region.
This spatial dissociation suggests that restoring the constrained preference acts at the \emph{output} (where the choice is generated), while disrupting it acts at the \emph{input} (where the constraint information is stored).

\paragraph{Connecting positions to tokens.}
Figure~\ref{fig:case-position-zones} summarizes the spatial structure.
The two causally important regions correspond to specific semantic content (cf.\ Figure~\ref{fig:pair-position-mapping}):

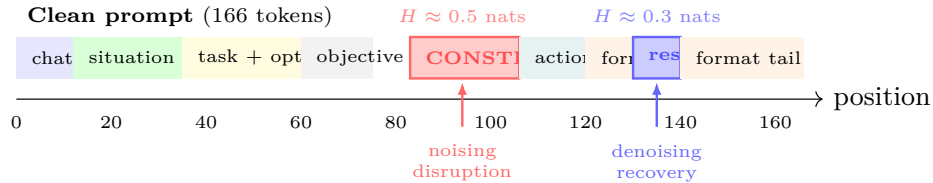
\begin{figure}[htbp]
  \centering
  \resizebox{0.9\textwidth}{!}{
\begin{tikzpicture}[x=0.055cm, y=0.8cm]

\draw[->] (0,0) -- (170,0) node[right, font=\small] {position};
\foreach \p/\l in {0/0, 20/20, 40/40, 60/60, 80/80, 100/100, 120/120, 140/140, 160/160} {
  \node[below, font=\tiny] at (\p, -0.1) {\l};
}

\fill[blue!10] (0,0.3) rectangle (12,0.9);
\node[font=\tiny, anchor=west] at (1, 0.6) {chat};

\fill[green!15] (12,0.3) rectangle (35,0.9);
\node[font=\tiny, anchor=west] at (13, 0.6) {situation};

\fill[yellow!15] (35,0.3) rectangle (60,0.9);
\node[font=\tiny, anchor=west] at (36, 0.6) {task + options};

\fill[gray!10] (60,0.3) rectangle (75,0.9);
\node[font=\tiny, anchor=west] at (61, 0.6) {objective};

\fill[red!20] (83,0.3) rectangle (106,0.9);
\draw[red!60, thick] (83,0.3) rectangle (106,0.9);
\node[font=\tiny\bfseries, text=red!70, anchor=west] at (84, 0.6) {CONSTRAINT};

\fill[teal!10] (106,0.3) rectangle (120,0.9);
\node[font=\tiny, anchor=west] at (107, 0.6) {action};

\fill[orange!10] (120,0.3) rectangle (130,0.9);
\node[font=\tiny, anchor=west] at (121, 0.6) {format};

\fill[blue!20] (130,0.3) rectangle (140,0.9);
\draw[blue!60, thick] (130,0.3) rectangle (140,0.9);
\node[font=\tiny\bfseries, text=blue!70, anchor=west] at (131, 0.6) {response};

\fill[orange!10] (140,0.3) rectangle (166,0.9);
\node[font=\tiny, anchor=west] at (141, 0.6) {format tail};

\node[font=\scriptsize, anchor=west] at (0, 1.2) {\textbf{Clean prompt} (166 tokens)};

\draw[-{Latex[length=1.5mm]}, thick, red!60] (94, -0.4) -- (94, 0.25);
\node[below, font=\tiny, text=red!60, align=center] at (94, -0.5) {noising\\disruption};

\draw[-{Latex[length=1.5mm]}, thick, blue!60] (135, -0.4) -- (135, 0.25);
\node[below, font=\tiny, text=blue!60, align=center] at (135, -0.5) {denoising\\recovery};

\node[above, font=\tiny, text=red!50] at (94, 0.95) {$H \approx 0.5$ nats};
\node[above, font=\tiny, text=blue!50] at (135, 0.95) {$H \approx 0.3$ nats};

\end{tikzpicture}}
  \caption{Semantic regions of the clean prompt with the two causally important zones highlighted.
  Noising disruption concentrates at the \texttt{CONSTRAINT} tokens (positions 83--106), where the horizon information is stored.
  Denoising recovery concentrates at the response boundary ($\sim$130--140), where the model reads the constraint to generate its choice.}
  \label{fig:case-position-zones}
\end{figure}
\begin{itemize}[nosep, leftmargin=*]
  \item \textbf{Positions 83--106} (constraint section).
  The noising entropy spike peaks at positions 95--105, which correspond to the tokens ``\texttt{time}'', ``\texttt{horizon}'', ``\texttt{:}'', ``\texttt{8}'', ``\texttt{months}'' (the core of the temporal constraint).
  The word ``\texttt{8}'' at position 103 and ``\texttt{months}'' at position 104 are the most specific tokens: they encode the deadline that makes the short-term option coherent.
  These positions exist only in the clean prompt; in the corrupted prompt, the alignment maps them onto the \texttt{ACTION} section.
  \item \textbf{Positions $\sim$130--140} (format/response boundary).
  The denoising entropy spike peaks here, at ``\texttt{I}'', ``\texttt{choose}'', ``\texttt{:}'', and the choice token itself.
  Attention at these positions shows the strongest individual-position effects, consistent with the model attending \emph{back} to the constraint region when generating the choice.
  The spatial separation between the two zones (constraint at 83--106, readout at 130--140) mirrors the temporal separation in the layer sweeps: information is stored in mid-layers and read out in later layers.
\end{itemize}

\paragraph{Constrained vs.\ latent preference.}
The pair structure makes it possible to distinguish two modes of temporal preference:
\begin{itemize}[nosep, leftmargin=*]
  \item \textbf{Constrained preference} (clean prompt): the model evaluates options against an explicit deadline and coherently picks the option that delivers within 8 months.
  The constraint tokens (positions 83--106) are the causal mechanism: patching them from the clean into the corrupted run restores the short-term choice.
  \item \textbf{Latent preference} (corrupted prompt): without a constraint, the model defaults to the higher-reward option (\$500K in 10 years), revealing a latent long-term bias when no temporal pressure is applied.
  This is the same default preference observed in the no-horizon condition of the behavioral coherence experiment (\ref{app:behavioral-coherence}), where \texttt{Qwen3-4B-Instruct-2507} picks long-term $\sim$96\% of the time without a horizon constraint.
\end{itemize}

\paragraph{Component granularity in position sweeps.}
The five position sweeps reveal how information propagates within each transformer block at the critical positions:
\begin{enumerate}[nosep, leftmargin=*]
  \item \texttt{resid\_pre} shows effects at the constraint region (83--106), indicating the temporal information is already in the residual stream from earlier layers.
  \item \texttt{attn\_out} shows effects primarily at the response boundary ($\sim$130--140), indicating attention \emph{reads} from the constraint to \emph{write} the choice.
  \item \texttt{resid\_mid} combines both, as expected (it is \texttt{resid\_pre} + \texttt{attn\_out}).
  \item \texttt{mlp\_out} shows distributed effects, contributing refinement at both regions.
  \item \texttt{resid\_post} shows the final, integrated picture.
\end{enumerate}

\noindent This flow, constraint information stored in the residual stream at positions 83--106, read by attention at positions $\sim$130--140, refined by MLP, and committed in the residual stream, mirrors the geometry findings (\ref{app:parametric-geometry}): the model builds a temporal representation in mid-layers and converts it to a preference at the turn/response boundary.

\FloatBarrier

\clearpage
\clearappnumbering


\clearpage
\section*{NeurIPS Paper Checklist}

\begin{enumerate}

\item {\bf Claims}
    \item[] Question: Do the main claims made in the abstract and introduction accurately reflect the paper's contributions and scope?
    \item[] Answer: \answerYes{}
    \item[] Justification: The abstract and Section~\ref{sec:introduction} state the three contributions (localizing a temporal-preference subgraph, characterizing its representational geometry, and steering the underlying axis) and each is delivered by a corresponding section (\ref{sec:methodology}--\ref{sec:results}) and appendix.

\item {\bf Limitations}
    \item[] Question: Does the paper discuss the limitations of the work performed by the authors?
    \item[] Answer: \answerYes{}
    \item[] Justification: Section~\ref{sec:limitations} explicitly discusses scaling to a single model (\texttt{Qwen3-4B-Instruct-2507}), domain generalization, the single-turn restriction, interactions with other concepts, and the linear-manifold approximation used for steering; the extended limitations appendix~\ref{app:extended-limitations} additionally documents the distributed attribution mass and the synthetic-label provenance of the contrastive datasets.

\item {\bf Theory assumptions and proofs}
    \item[] Question: For each theoretical result, does the paper provide the full set of assumptions and a complete (and correct) proof?
    \item[] Answer: \answerNA{}
    \item[] Justification: \answerNA{}

    \item {\bf Experimental result reproducibility}
    \item[] Question: Does the paper fully disclose all the information needed to reproduce the main experimental results of the paper to the extent that it affects the main claims and/or conclusions of the paper (regardless of whether the code and data are provided or not)?
    \item[] Answer: \answerYes{}
    \item[] Justification: The model (\texttt{Qwen3-4B-Instruct-2507}) is openly released; prompts, datasets, attribution (EAP-IG), probing, CAA steering construction, and evaluation procedures are fully specified in Appendices~\ref{app:prompts}--\ref{app:contrastive-steering}.
    \item[] Guidelines:
    \begin{itemize}
        \item The answer \answerNA{} means that the paper does not include experiments.
        \item If the paper includes experiments, a \answerNo{} answer to this question will not be perceived well by the reviewers: Making the paper reproducible is important, regardless of whether the code and data are provided or not.
        \item If the contribution is a dataset and\slash or model, the authors should describe the steps taken to make their results reproducible or verifiable. 
        \item Depending on the contribution, reproducibility can be accomplished in various ways. For example, if the contribution is a novel architecture, describing the architecture fully might suffice, or if the contribution is a specific model and empirical evaluation, it may be necessary to either make it possible for others to replicate the model with the same dataset, or provide access to the model. In general, releasing code and data is often one good way to accomplish this, but reproducibility can also be provided via detailed instructions for how to replicate the results, access to a hosted model (e.g., in the case of a large language model), releasing of a model checkpoint, or other means that are appropriate to the research performed.
        \item While NeurIPS does not require releasing code, the conference does require all submissions to provide some reasonable avenue for reproducibility, which may depend on the nature of the contribution. For example
        \begin{enumerate}
            \item If the contribution is primarily a new algorithm, the paper should make it clear how to reproduce that algorithm.
            \item If the contribution is primarily a new model architecture, the paper should describe the architecture clearly and fully.
            \item If the contribution is a new model (e.g., a large language model), then there should either be a way to access this model for reproducing the results or a way to reproduce the model (e.g., with an open-source dataset or instructions for how to construct the dataset).
            \item We recognize that reproducibility may be tricky in some cases, in which case authors are welcome to describe the particular way they provide for reproducibility. In the case of closed-source models, it may be that access to the model is limited in some way (e.g., to registered users), but it should be possible for other researchers to have some path to reproducing or verifying the results.
        \end{enumerate}
    \end{itemize}

\item {\bf Open access to data and code}
    \item[] Question: Does the paper provide open access to the data and code, with sufficient instructions to faithfully reproduce the main experimental results, as described in supplemental material?
    \item[] Answer: \answerNo{}
    \item[] Justification: A public code repository exists but is intentionally not linked in this submission to preserve anonymity; it will be released alongside the camera-ready version.
Appendices~\ref{app:prompts}--\ref{app:contrastive-steering} document datasets, prompts, and procedures in sufficient detail to re-implement every experiment on the publicly available \texttt{Qwen3-4B-Instruct-2507} checkpoint.

\item {\bf Experimental setting/details}
    \item[] Question: Does the paper specify all the training and test details (e.g., data splits, hyperparameters, how they were chosen, type of optimizer) necessary to understand the results?
    \item[] Answer: \answerYes{}
    \item[] Justification: No model training is involved; Section~\ref{sec:experimental-setup} summarizes datasets, counterbalancing, and judge setup, and the appendices specify layer ranges, steering coefficients, probe training protocols, and evaluation prompts.

\item {\bf Experiment statistical significance}
    \item[] Question: Does the paper report error bars suitably and correctly defined or other appropriate information about the statistical significance of the experiments?
    \item[] Answer: \answerYes{}
    \item[] Justification: All activation-patching figures report mean $\pm$ one standard deviation across input pairs, displayed as shaded bands on line plots and error bars on bar charts (titles explicitly state ``mean $\pm$ std''); the variability source (across input pairs) and $\sigma$ definition are stated in each figure caption.

\item {\bf Experiments compute resources}
    \item[] Question: For each experiment, does the paper provide sufficient information on the computer resources (type of compute workers, memory, time of execution) needed to reproduce the experiments?
    \item[] Answer: \answerYes{}
    \item[] Justification: Section~\ref{sec:experimental-setup} states that the full pipeline runs end-to-end within two weeks on a single MacBook Pro (M4 Max, 48\,GB); no GPU cluster is required.
    
\item {\bf Code of ethics}
    \item[] Question: Does the research conducted in the paper conform, in every respect, with the NeurIPS Code of Ethics \url{https://neurips.cc/public/EthicsGuidelines}?
    \item[] Answer: \answerYes{}
    \item[] Justification: The work analyzes a publicly released model, uses no human subjects or private data, and introduces no new training pipeline or deployed system.

\item {\bf Broader impacts}
    \item[] Question: Does the paper discuss both potential positive societal impacts and negative societal impacts of the work performed?
    \item[] Answer: \answerYes{}
    \item[] Justification: Section~\ref{sec:introduction} motivates the work as AI-safety-relevant by framing temporal preference as a controllable property, and the steering experiments (Section~\ref{sec:results}, \ref{app:contrastive-steering}) show both constructive (alignment-style control) and dual-use (bias induction) implications.
    
\item {\bf Safeguards}
    \item[] Question: Does the paper describe safeguards that have been put in place for responsible release of data or models that have a high risk for misuse (e.g., pre-trained language models, image generators, or scraped datasets)?
    \item[] Answer: \answerNA{}
    \item[] Justification: \answerNA{}

\item {\bf Licenses for existing assets}
    \item[] Question: Are the creators or original owners of assets (e.g., code, data, models), used in the paper, properly credited and are the license and terms of use explicitly mentioned and properly respected?
    \item[] Answer: \answerYes{}
    \item[] Justification: We use \texttt{Qwen3} (Apache 2.0), Anthropic Claude API, OpenAI GPT API, and Google Gemini API (each subject to provider terms of service); the Kirby MCQ-27 instrument from \citep{kirby1999} is reproduced in standard form.
Each is credited in the relevant section.

\item {\bf New assets}
    \item[] Question: Are new assets introduced in the paper well documented and is the documentation provided alongside the assets?
    \item[] Answer: \answerYes{}
    \item[] Justification: We introduce two new evaluation prompt sets (a parametric prompt grid and a 960-prompt investment coherence questionnaire); the prompt sources, generation rules, and licenses will be released as part of the camera-ready supplementary materials.

\item {\bf Crowdsourcing and research with human subjects}
    \item[] Question: For crowdsourcing experiments and research with human subjects, does the paper include the full text of instructions given to participants and screenshots, if applicable, as well as details about compensation (if any)? 
    \item[] Answer: \answerNA{}
    \item[] Justification: \answerNA{}

\item {\bf Institutional review board (IRB) approvals or equivalent for research with human subjects}
    \item[] Question: Does the paper describe potential risks incurred by study participants, whether such risks were disclosed to the subjects, and whether Institutional Review Board (IRB) approvals (or an equivalent approval/review based on the requirements of your country or institution) were obtained?
    \item[] Answer: \answerNA{}
    \item[] Justification: \answerNA{}

\item {\bf Declaration of LLM usage}
    \item[] Question: Does the paper describe the usage of LLMs if it is an important, original, or non-standard component of the core methods in this research? Note that if the LLM is used only for writing, editing, or formatting purposes and does \emph{not} impact the core methodology, scientific rigor, or originality of the research, declaration is not required.
    \item[] Answer: \answerYes{}
    \item[] Justification: LLMs are used in three methodologically relevant places, all fully documented in the appendices.
    First, the minimally-framed contrastive datasets ($D_{\text{explicit}}$ and $D_{\text{implicit}}$) were generated with \texttt{Claude Sonnet 4.6} and validated by \texttt{Claude Sonnet 4.6} together with \texttt{Gemini 3 Flash} across four dimensions (lexical confounds, surface form, semantic confounds, content validity); pairs scoring below threshold were iteratively revised or discarded, and A/B presentation was randomized to control for positional bias (\ref{app:minimally-framed}).
    Second, the classification dataset was constructed iteratively with \texttt{Claude Opus 4.6} in successive batches: each batch was generated by the model according to the constraints specified during dataset design, then manually checked and refined with continuous user feedback before being considered final (\ref{app:classification-oriented}).
    Third, open-ended steering generations were scored by an LLM-as-judge: each response was rated on a $[-10, 10]$ short-term/long-term axis by \texttt{Claude Sonnet 4.6} against predefined grading criteria (explicit temporal keywords, structural planning, thematic bias), and these scores feed the steering evaluation reported in Section~\ref{sec:steering} (\ref{app:contrastive-steering:discussion}).    

\end{enumerate}

\end{document}